\pdfoutput=1
\documentclass{article}

\usepackage[preprint,nonatbib]{neurips_2023}

\usepackage[utf8]{inputenc} 
\usepackage[T1]{fontenc}    
\usepackage{hyperref}       
\usepackage{url}            
\usepackage{booktabs}       
\usepackage{amsfonts}       
\usepackage{nicefrac}       
\usepackage{microtype}      
\usepackage{xcolor}         
\usepackage{graphicx}
\hypersetup{
    colorlinks=true,
    anchorcolor=blue,
    citecolor={olive},
    linkcolor=blue
}

\usepackage{amsmath}
\usepackage{bm}
\usepackage{algorithm}
\usepackage{algpseudocode}
\usepackage{multirow}
\usepackage{caption}
\usepackage{bbm}
\usepackage{xcolor,colortbl}
\usepackage{wrapfig}

\title{Improving and Benchmarking Offline Reinforcement Learning Algorithms}

\author{
Bingyi Kang$^1$\thanks{equal contribution}~~~ Xiao Ma$^{1*}$~~~ Yirui Wang$^{12}$~~~ Yang Yue$^{13}$~~~ Shuicheng Yan$^{1}$ \\
  {$^1$Sea AI Lab~~~ $^2$National University of Singapore~~~ $^3$Tsinghua University} \\
  \texttt{\{bingykang, yusufma555, yirui.wang0212\}@gmail.com}~ \\
  \texttt{le-y22@mails.tsinghua.edu.cn}~ \texttt{yansc@sea.com}
}
\definecolor{darkgreen}{rgb}{0,0.5,0}
\definecolor{fullred}{rgb}{0.95,.0,.1}
\definecolor{brown}{rgb}{0.65,0.16,0.16}
\definecolor{orange}{rgb}{1,0.5,0}
\newcommand{\comment}[3]{\ifthenelse{\boolean{}}
{
  \marginpar{\tiny\noindent{\raggedright{{\colorbox{#3}{\sffamily\textcolor{white}{#1
            [\arabic{0}]}}}}} \color{#3}{#2} \par}}{}}

\newcommand{\xseq}[1]{$\{x^{1:N}_{t}\}_{t=0}^T$}
\newcommand{\dynamic}[1]{\ensuremath{T}}
\newcommand{\cvae}[1]{ControlVAE}

\newcommand{\keep}[1]{}
\newcommand{\old}[1]{}

\algnewcommand{\LeftComment}[1]{\Statex \(\triangleright\) #1}

\begin{document}


\maketitle
\begin{abstract}
Recently, Offline Reinforcement Learning (RL) has achieved remarkable progress with the emergence of various algorithms and datasets. However, these methods usually focus on algorithmic advancements, ignoring that many low-level implementation choices considerably influence or even drive the final performance. As a result, it becomes hard to attribute the progress in Offline RL as these choices are not sufficiently discussed and aligned in the literature. In addition, papers focusing on a  dataset (\textit{e.g.}, D4RL) often ignore algorithms proposed on another dataset (\textit{e.g.}, RL Unplugged), causing isolation among the algorithms, which might slow down the overall progress. Therefore, this work aims to bridge the gaps caused by low-level choices and datasets. To this end, we empirically investigate 20 implementation choices using three representative algorithms (\textit{i.e.}, CQL, CRR, and IQL) and present a guidebook for choosing implementations. Following the guidebook, we find two variants CRR$^+$ and CQL$^+$, achieving new state-of-the-art on D4RL. Moreover, we benchmark eight popular offline RL algorithms across datasets under unified training and evaluation framework. The findings are inspiring: the success of a learning paradigm severely depends on the data distribution, and some previous conclusions are biased by the dataset used. 
Our code is available at \href{https://github.com/sail-sg/offbench}{https://github.com/sail-sg/offbench}.
\end{abstract}
\section{Introduction}
Deep Reinforcement Learning (RL) is of significant importance to solving sequential decision-making tasks, ranging from game playing~\cite{mnih2013playing,silver2017mastering,berner2019dota} to robot control~\cite{levine2016end,kahn2018self,savva2019habitat}. However, interacting with the environment is prohibitively expensive and dangerous in real-world safety-sensitive scenarios, which limits the applications of RL methods outside of simulators. Therefore, offline RL, targeting learning agents from pre-collected experiences by arbitrary agents to avoid online interaction, is receiving increasing attention. 
As a result, remarkable achievements have been made in recent years. Most of them aim to solve the distributional shift problem ~\cite{levine2020offline}, by introducing constraints or regularizations in either policy evaluation step~\cite{kumar2020conservative,yu2021combo} or policy improvement step~\cite{wang2020critic,fujimoto2021minimalist}.

The rapid progress brings new challenges in benchmarking the advances in offline RL.
First, offline RL algorithms contain many low-level design choices that are often not well-discussed or aligned in the literature. This makes it impossible to assess whether improvements are due to the algorithms or due to their implementations. Similar observations have been made by various studies~\cite{andrychowicz2020matters,Engstrom2020Implementation} in online RL that low-level choices play a critical role in driving performance and, thus, should not be overlooked.
Second, multiple datasets are released to facilitate offline RL research, among which RL Unplugged~\cite{gulcehre2020rl} and D4RL~\cite{fu2020d4rl} are the most popular ones. However, there is apparent isolation between them.
That is,
algorithms evaluated on one dataset~\cite{wang2020critic,schrittwieser2021online,gulcehre2020rl} are often ignored by papers focusing on another~\cite{kumar2020conservative,kostrikov2022offline,yu2021combo}, and vice versa. As a result, the conclusions drawn in a paper might be highly biased by the dataset used. In addition, evaluation metrics might not be aligned and not directly comparable. For example,~\cite{wang2022diffusion} considers the best score at training, while \cite{kostrikov2022offline} reports the ending performance of the training process. 

The key goal of this work is thus two-fold: 1) to investigate low-level algorithm choices in depth to better attribute the progress in offline RL; 2) to benchmark offline RL algorithms across datasets with a unified evaluation protocol to facilitate future research. We first summarize and implement 20 choices from the literature and select three representative algorithms for low-level implementation study, including  Critic Regularized Regression (CRR)~\cite{wang2020critic}, Conservative Q-Learning (CQL)~\cite{kumar2020conservative}, and Implicit Q-Learning (IQL)~\cite{kostrikov2022offline}. Through careful alignment and ablation, we provide a guidebook for making low-level decisions in offline RL algorithms. Moreover, we develop two variants of CRR and CQL (which we refer to as CRR$^+$ and CQL$^+$) based on the guidebook, which significantly improves upon their original implementations (CQL$^+$ by $5\% $ and CRR$^+$ by $33.8\%$), and outperform the current state-of-the-art (SOTA) method.

Then, to benchmark and validate the generalization ability across datasets with different distributions. We select two algorithms (Muzero Unplugged~\cite{schrittwieser2021online} 
and CRR) from RL Unplugged, and six (BC, SAC~\cite{haarnoja2018soft}, Onestep RL~\cite{brandfonbrener2021offline}, TD3+BC~\cite{fujimoto2021minimalist}, CQL, and IQL) from D4RL. 
To eliminate the effect of codebases, we carefully re-implement all eight algorithms by strictly following their official implementations under a unified offline RL training framework. Then we conduct experiments on 26 tasks from 5 domains across D4RL and RL Unplugged. 
D4RL resembles the case where the data is generated by one or more \textit{fixed} agents, while RL Unplugged, on the other hand, considers the abundant replays generated by prior RL agents that can be used to learn a new agent efficiently. Our findings are insightful: algorithms with policy constraints demonstrate better transferability across datasets, while value constraints, which produce lower-bounded values, generally produce a worse performance on replay data. In addition, unconstrained off-policy algorithms, e.g., SAC and MuZero Unplugged, often fail on datasets generated by a mixture of agents, which is contradictory to the previous conclusion made by \cite{schrittwieser2021online}.
In summary, we make the following contributions: 
\begin{itemize}
    \item A unified training framework for various offline RL algorithms.
    \item A guidebook for low-level implementation choices in offline RL and two improved algorithms (CRR$^+$ and CQL$^+$) that have never been recorded before. 
    \item Insightful observations on dataset distributions and algorithmic designs and practical recommendations for algorithm selection.
\end{itemize}





\section{Related Works}
\textbf{Offline RL.}
The biggest challenge of offline RL is the distribution shift between the learning policy and the behavior policy, \textit{i.e.}, the policy used to collect the offline dataset~\cite{levine2020offline}. Due to the unconstrained Bellman's update, the extrapolation error of unseen Q-values $Q(s, a)$ will accumulate during training and eventually produce an erroneous policy. Therefore, most of the existing offline RL methods consider a conservative learning framework implemented as an additional soft constraint upon the RL objective. The key to conservative learning is encouraging the learning policy to stay close to the behavior policy. It will query the out-of-distribution (OOD) actions less frequently. Such a constraint can be imposed directly on the policy improvement step, \textit{i.e.}, on the policy~\cite{fujimoto2019off,wu2019behavior,fujimoto2021minimalist,Siegel2020Keep,wang2020critic}. For example, TD3+BC~\cite{fujimoto2021minimalist} adds an additional behavior cloning term as a regularizer for the policy to stay in the dataset manifold. Also, the conservative constraints can be indirectly imposed on the policy evaluation step, \textit{i.e.}, the Q-functions~\cite{kumar2020conservative}. Differently, MuZero Unplugged has demonstrated 
its applicability to both online and offline RL settings without any modifications to its algorithmic structures~\cite{schrittwieser2021online}. However, as prior algorithms are either evaluated only on D4RL or RL Unplugged without intersections, it is hard to compare their performance directly. In this work, we unify eight popular algorithms under the same framework, ranging from simple CQL and CRR~\cite{wang2020critic}, \textit{etc}., to highly complex MuZero Unplugged~\cite{schrittwieser2021online}, and experiment on both D4RL and RL Unplugged dataset to provide a better understanding of the progress. For algorithms that require a significant change to the network architecture or computation cost, e.g., Decision Transformer~\cite{chen2021decision} or SAC-n~\cite{an2021uncertainty}, we leave them for future study.

\textbf{Benchmarking RL Algorithms.}
Reinforcement learning provides a principled way to solve sequential decision-making problems.
However, it is notorious for its instability and sensitivity to hyper-parameters and low-level implementation choices~\cite{tucker2018mirage,duan2016benchmarking,islam2017reproducibility}. Such a phenomenon commonly exists in model-based RL~\cite{wang2019benchmarking}, off-policy RL~\cite{voloshin2019empirical,furuta2021co}, and on-policy RL~\cite{andrychowicz2020matters,Engstrom2020Implementation}. 
In addition, as discussed in~\cite{agarwal2021deep}, naive point estimation of RL returns might be statistically unstable. However, existing works on offline RL mainly compare their point estimations with results from prior published papers without careful tuning of hyper-parameters and implementation choices. Such a scheme may hinder understanding the real progress in the offline RL field. Hence, we present a comprehensive benchmarking for offline RL algorithms, covering eight popular algorithms ranging from constrained policy improvement~\cite{fujimoto2021minimalist,wang2020critic,kostrikov2022offline} to constrained policy evaluation~\cite{kumar2020conservative}. We evaluate all algorithms under a unified framework with three different metrics which report the extreme performance as well as the stability of algorithms.
\section{Preliminaries}
We consider a Markov Decision Process (MDP) denoted as a tuple $\mathcal{M} = (\mathcal{S}, \mathcal{A}, p_0(s), p(s'\mid s, a), r(s, a), \gamma)$, where $\mathcal{S}$ and $\mathcal{A}$ are the state and action spaces, $p_0(s)$ is the initial state distribution, $p(s'\mid s, a)$ is the transition function, $r(s, a)$ is the reward function, and $\gamma$ is the discount factor. The target of reinforcement learning is to find a policy $\pi^*(a\mid s)$ that maximizes the accumulative return
\begin{align}
    &\pi^* = \arg\max_{\pi}\mathbb{E}\left[ \sum\limits_{t=0}^{\infty}\gamma^t r(s_t, a_t)\right]\nonumber\\
    &s_0\sim p_0(s), s'\sim p(\cdot\mid s, a), a\sim \pi(\cdot\mid s). 
\end{align}
In an actor-critic framework, policy optimization follows Bellman's expectation operator $\mathcal{B}^\pi Q(s, a) = r(s, a) + \gamma \mathbb{E}_{s'\sim p(\cdot\mid s, a), a'\sim\pi(\cdot\mid s')}\left[Q(s', a')\right]$, which alternates between policy evaluation and policy improvement. Given a policy $\pi_\theta(a\mid s)$ and Q function $Q_\phi(s, a)$, policy evaluation aims to learn the Q function by minimizing the prediction error $\mathbb{E}_{\mu_{\pi_\theta}\pi_\theta(a\mid s)}\left[(Q_\phi(s, a) - \mathcal{B}^{\pi_\theta}Q_\phi(s, a))^2\right]$, where $\mu_{\pi_\theta}$ is the stationary distribution induced by $\pi_\theta(a\mid s)$~\cite{schulman2015trust}. On the other hand, policy improvement focuses on learning the optimal policy by maximizing the approximated accumulative return by Q functions $\mathbb{E}_{\mu_{\pi_\theta}(s)\pi_\theta(a\mid s)}\left[Q_\phi(a\mid s)\right]$. 

However, as querying OOD actions is inevitable when sampling from $\pi_\theta(a\mid s)$, both the policy improvement and evaluation steps are affected in offline RL setups. To alleviate this issue, conservative RL methods impose additional constraints on either the policy improvement step or the policy evaluation step to encourage the learning policy $\pi_\theta$ to stay close to the behavior policy $\pi_\beta$ that generates the dataset. Concretely, conservative policy improvement and conservative policy evaluation can be written as
\begin{align}
    &\max_\theta \mathbb{E}_{\mu_{\pi_\theta}(s)\pi_\theta(a\mid s)}\left[Q_\phi(a\mid s) - \alpha_1 C^\pi_{\theta, \phi, \beta}\right]\nonumber\\
    &\min_\phi\mathbb{E}_{\mu_{\pi_\theta}\pi_\theta(a\mid s)}\left[(Q_\phi(s, a) - \mathcal{B}^{\pi_\theta}Q_\phi(s, a))^2 + \alpha_2 C^Q_{\theta, \phi, \beta} \right]\nonumber
\end{align}
where $\alpha_1$ and $\alpha_2$ are hyper-parameters, and $C^\pi_{\theta, \phi, \beta}$ and $C^Q_{\theta, \phi, \beta}$ are conservative constraints for the policy and value functions respectively.
In the later part of this paper, we refer to these two approaches as the conservative policy evaluation and the conservative policy improvement.
\section{Implementation Choices for Offline RL}
RL algorithms are notorious for their instability and sensitivity to hyper-parameters and implementation choices~\cite{andrychowicz2020matters,Engstrom2020Implementation}. As a particular case of RL, besides this issue, offline RL also 
suffers from other implementation difficulties on the conservative constraints terms during optimization. 
We first pick two baseline algorithms for case studies, CRR for conservative policy improvement and CQL for conservative policy evaluation, investigating how better low-level choices enable baselines to outperform the SOTA algorithm, IQL. In addition, we perform ablations on IQL. Our results demonstrate that the success of IQL highly depends on the choice of implementations.

\subsection{Study Design}\label{sect:study_design}
\begin{figure*}[htb]
	\centering
	\begin{tabular}{ccc}
    \includegraphics[height=68pt]{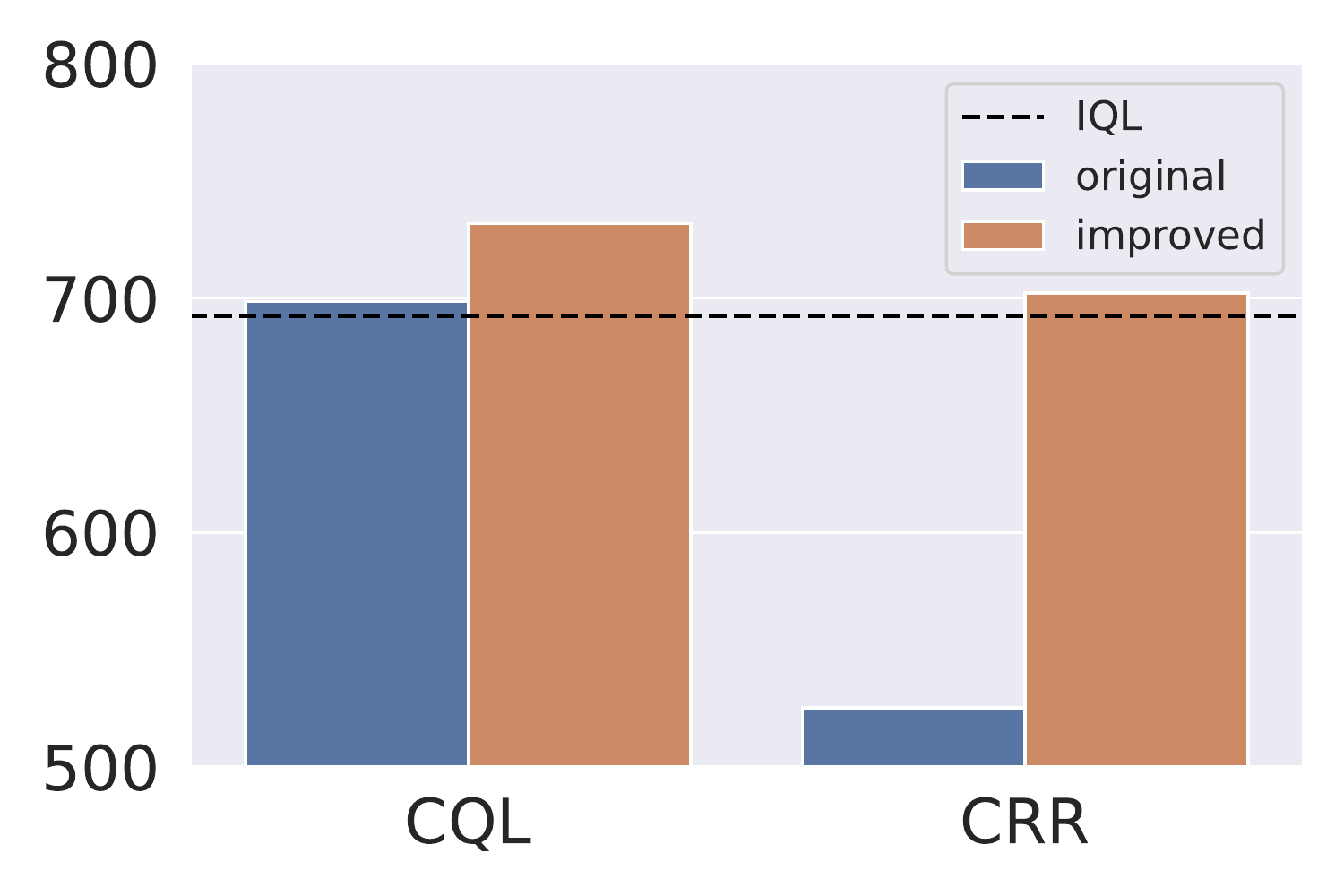} &
    \includegraphics[height=68pt]{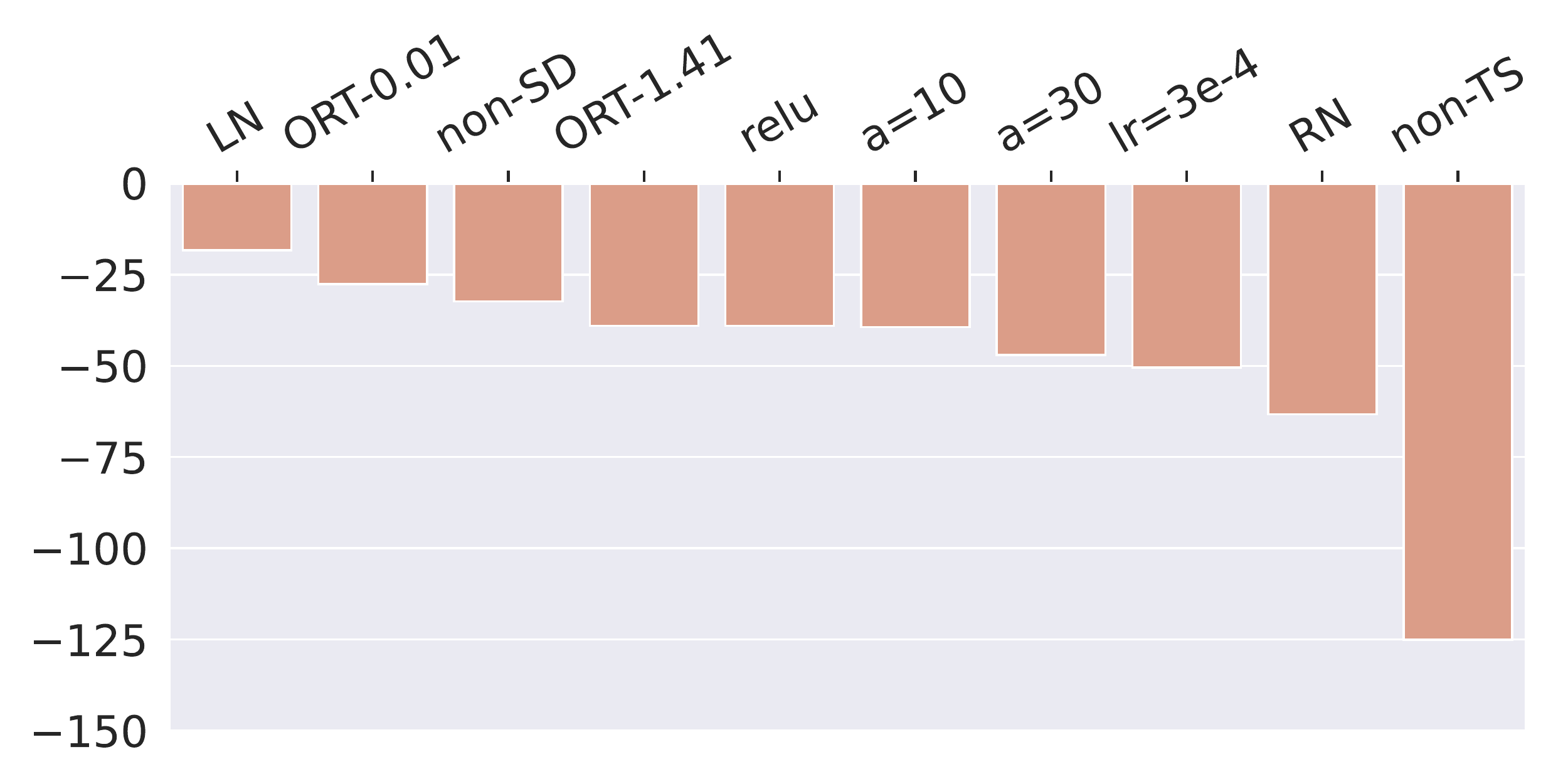} &
    \includegraphics[height=68pt]{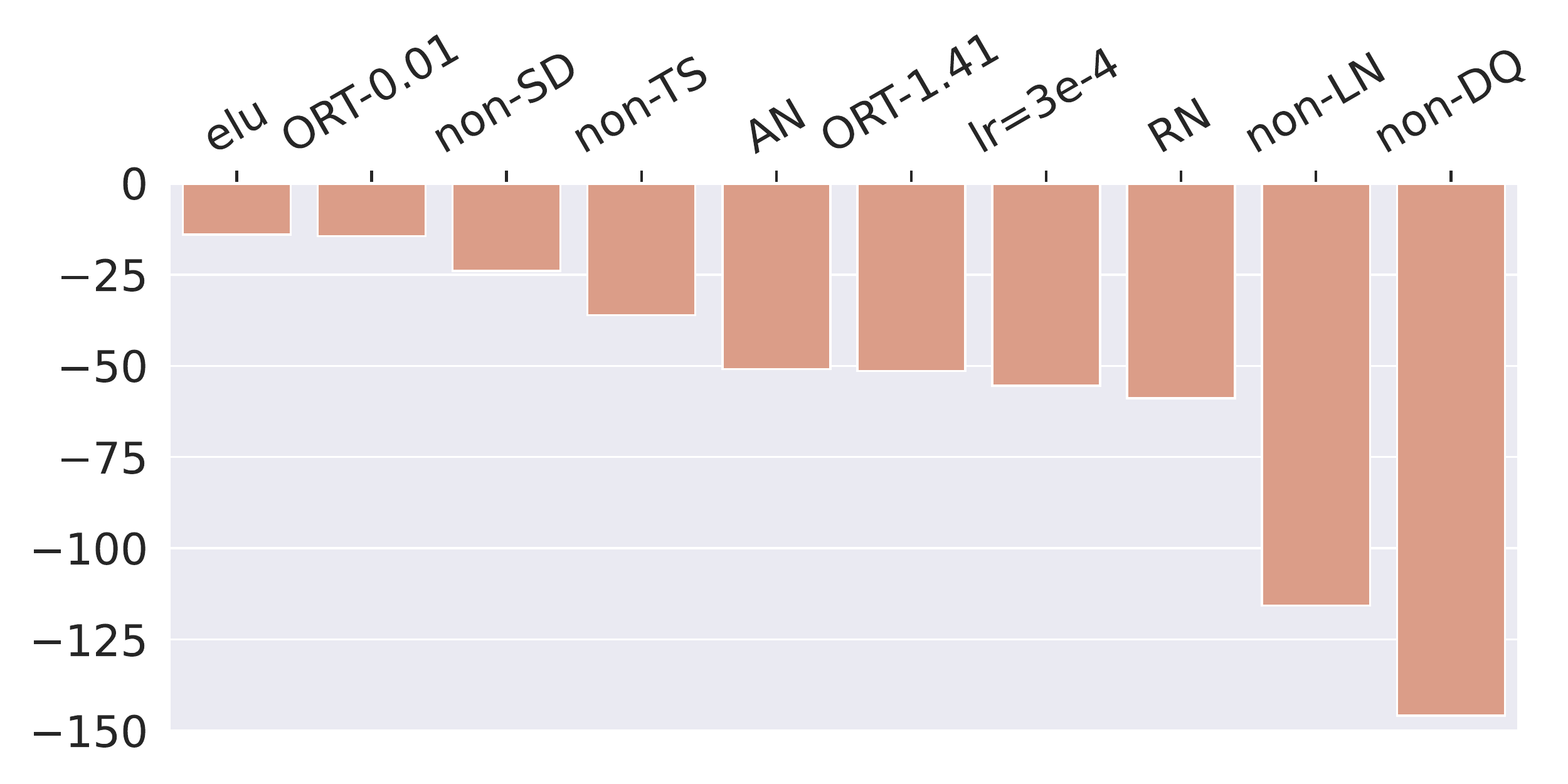} \\
    \small (a) Improving the baselines & \small (b) CQL Ablations  &\small (c) CRR Ablations
	\end{tabular}
	\centering
	\caption{\small Influence of implementation choices. We conduct ablation studies for two common baselines, CRR and CQL, on the gym-locomotion-v2 tasks of the D4RL benchmark. (a) We sweep the implementation choices and hyper-parameters of CQL and CRR and report the average \textbf{last step return}. We show that with careful choices, CQL$^+$ achieves a total score of 731.9 (5\% improvement) and CRR$^+$ reaches 702.3 (33.8\% improvement), both of which outperform the SOTA method IQL (692.4). (b, c) We report the performance drop of ablations compared with the optimal configuration in (a). More details and discussions are available in Sect.~\ref{sect:cql} and Sect.~\ref{sect:crr}. }
	\label{fig:ablations}
\end{figure*}

We split the implementation choices into two categories, general RL choices, and algorithmic-specific choices. In this section, we discuss the general RL choices. Specifically, we focus on the gym-locomotion tasks (v2) of the D4RL dataset. For a fair comparison with reported results from~\cite{kostrikov2022offline}, we report the average last step return over 3 seeds and 100 episodes. We pick a subset of the implementation choices for investigation, where the abbreviation used in Fig.~\ref{fig:ablations} is bolded. 

\textit{Weight initialization scheme.} The initialization scheme of the last output layer of the network has a huge impact on the final performance~\cite{andrychowicz2020matters}. We study three variants: orthogonal initialization with scale $\sqrt{2}$ (\textbf{ORT-1.41}), orthogonal initilaization with scale 0.01 (\textbf{ORT-0.01}), and the default Lecun normal initialization (\textbf{non-ORT}).

\textit{Policy learning rate and scheduler}. For the Q function learning rate, we fix a commonly adopted value of $3e^{-4}$. For policy learning rate, we examine two configurations, $1e^{-4}$ (\textbf{lr=1e-4}) and $3e^{-4}$ (\textbf{lr=3e-4}) with cosine learning rate scheduler.

\textit{Reward normalization.} Reward normalization is one of the most important factors in RL~\cite{Engstrom2020Implementation}. We evaluate two settings: without reward normalization (\textbf{non-RN}) and reward normalization (\textbf{RN}) as $r'=r / (\max R - \min R) * 1000$, where $\max R$ and $\min R$ denote the maximum and minimum trajectory returns of the dataset.

\textit{Policy distribution parameterization.} We consider two variants of policy representation, tanh-squashed Gaussian $a \sim \mathrm{tanh}(\mathcal{N}(\mu_a, \sigma_a\mid s))$ (\textbf{TS}), or a clipped Gaussian distribution with tanh-squashed mean $a\sim \mathrm{clip}(\mathcal{N}(\mathrm{tanh}(\mu_a), \sigma_a\mid s), -1, 1)$ (\textbf{non-TS}). In addition, we also evaluate the influence of variance parameterization by either making it state-dependent (\textbf{SD}) or independent parameters (\textbf{non-SD}).

\textit{Layer normalization.} As Q-value over-estimation is a common issue in offline RL, we add Layer Normalization~\cite{ba2016layer} to the policy and Q value networks (\textbf{LN}) and examine if it improves the numerical stability.

\textit{Activation functions.} We choose two different activation functions, \textbf{relu} and \textbf{elu}~\cite{clevert2015fast}. The activation function is applied after layer normalization.

\begin{figure*}[htb]
	\centering
	\begin{tabular}{cc}
    \includegraphics[height=80pt]{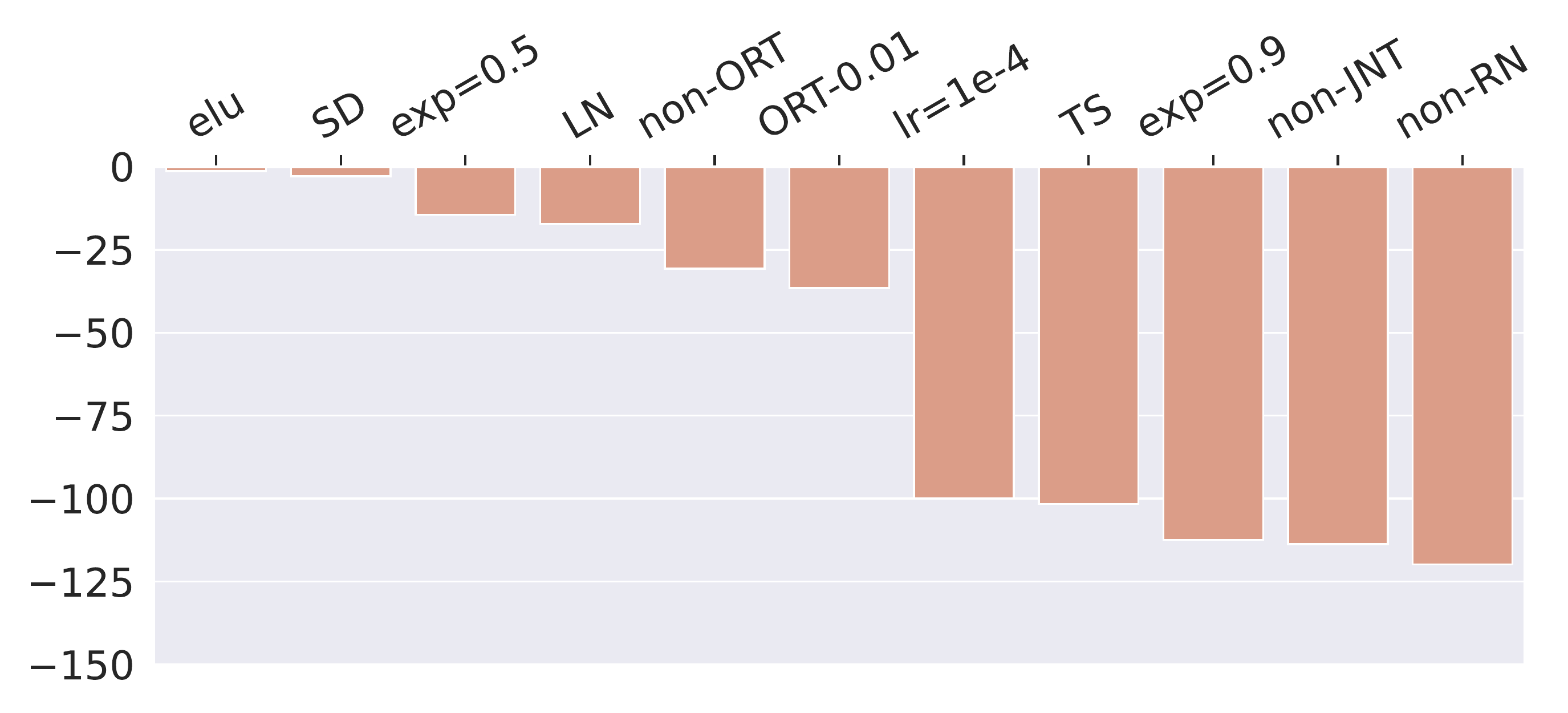} &
    \includegraphics[height=90pt]{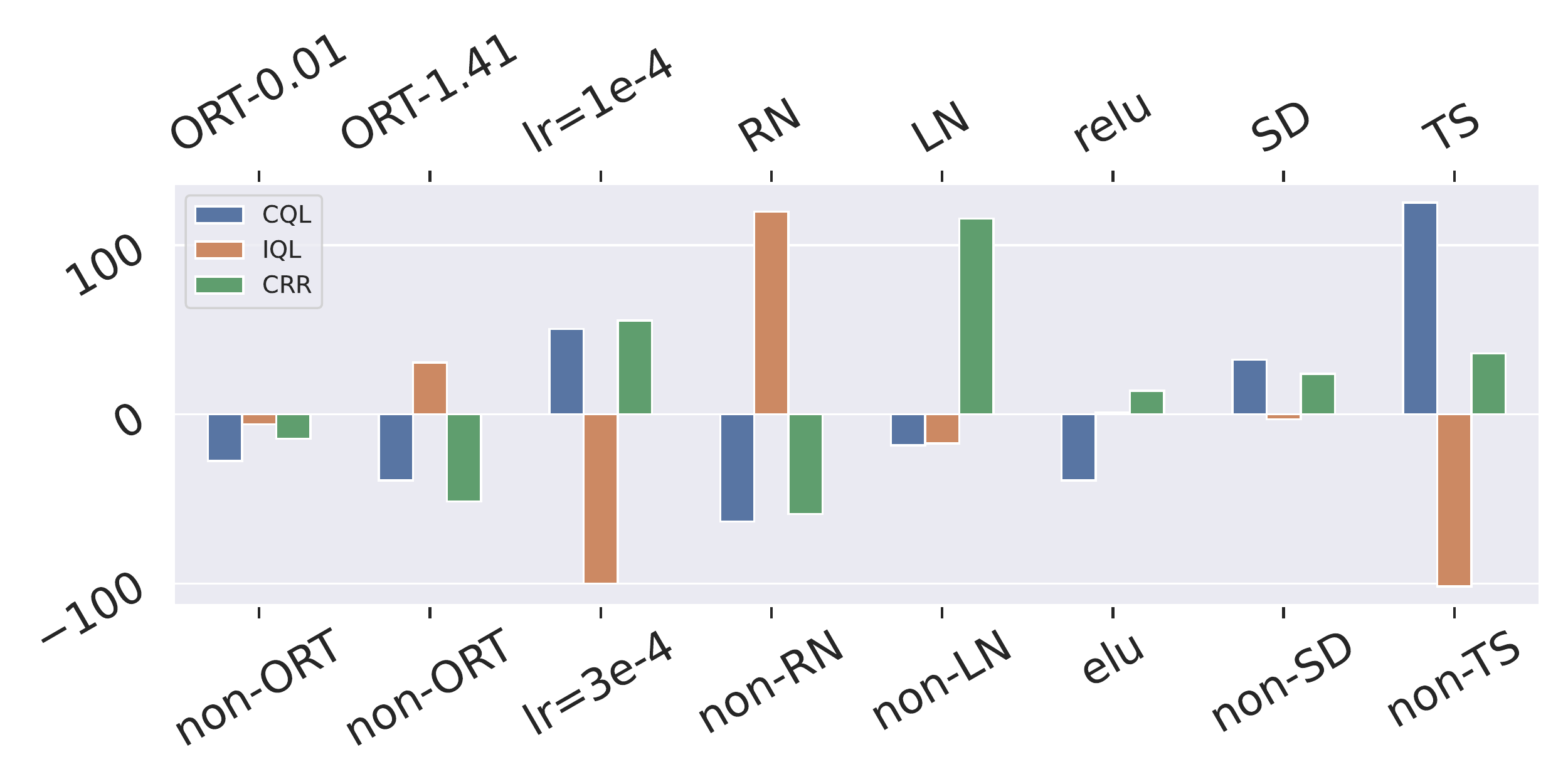} \\
    \small (a) IQL Ablations & \small (b) Summary of Implementation Ablations
	\end{tabular}
	\centering
	\caption{\small Influence of implementation choices. (a) We report the performance drop of IQL ablations compared with the optimal configuration in its official implementation~\cite{kostrikov2022offline} (more details in Sect.~\ref{sect:iql}). (b) For shared implementation choices, we visualize the performance differences between an implementation choice (top half) and its baseline (bottom half).}
	\label{fig:iql}
\end{figure*}

\subsection{Case Study: Conservative Q Learning}\label{sect:cql}
Conservative Q Learning (CQL)~\cite{kumar2020conservative} focuses on directly regularizing the Q-value functions during optimization. CQL learns a lower-bound of the ground-truth Q values by implementing $C_{\theta, \phi, \beta}^Q$ as
\begin{equation}
    C_{\theta, \phi, \beta}^Q = \mathbb{E}_{s\sim \mathcal{D}}\left[\log\sum\limits_{a}\exp Q(s, a) - \mathbb{E}_{a\sim \pi_\beta (a\mid s)}\left[ Q(s, a) \right]\right].
\end{equation}
Intuitively, CQL encourages the agent to produce high Q-values for in-distribution actions (positive sample), while suppressing the Q-value of OOD actions (negative samples). 

For CQL, we further ablate its \textit{number of actions} for negative examples.
In practice, we use $\pi_\theta(a\mid s)$, $\pi_\theta(a\mid s')$, and $\mathcal{U}(-1, 1)$ to generate negative samples, where $s'$ is the next state and $\mathcal{U}$ is a uniform distribution. For each distribution $\mathbf{a=N}$ actions are sampled, where $N\in \{10, 30, 50\}$.

\textbf{CQL$^+$}. The official implementation of CQL adopts $\mathbf{a=10}$, a policy learning rate of $3e-4$, and relu activation for all networks. After sweeping, we observe that using $\mathbf{a=50}$, policy learning of $1e-4$, and elu activation, CQL$^+$ significantly improves the performance of CQL. In Fig.~\ref{fig:ablations}(a), CQL$^+$ achieves a total score of \textbf{731.9} on gym-locomotion-v2 tasks of D4RL benchmark, while the original implementation has a score of 698.5~\cite{kumar2020conservative}. Fig.~\ref{fig:ablations}(b) also details the ablation results of CQL. We observe that tanh-squashed distribution is the most critical component of CQL, without which CQL would suffer a significant 17\% performance drop. In addition, reward normalization hurts the performance of CQL. Moreover, a proper number of sampled actions, learning rate, and activation function also contribute significantly to the performance of CQL$^+$, while layer-norm and weight initialization schemes have a relatively minor impact on its performance.

\subsection{Case Study: Critic Regularized Regression}\label{sect:crr}
Critic Regularized Regression (CRR)~\cite{wang2020critic} handles offline RL with conservative policy improvement. Concretely, it learns the policy by
\begin{equation}
    \arg\max_\pi \mathbb{E}_{s, a\sim \mathcal{D}}\left[f(Q_\phi, \pi, s, a)\log\pi_\theta (a\mid s))\right]\label{eqn:crr}
\end{equation}
where $\mathcal{D}$ is the dataset, and $f$ is a non-negative scalar function whose value is monotonically increasing in $Q_\phi$. One common choice for $f$ is $f(Q_\phi, \pi, s, a) = \exp\left[Q_\phi(s, a) - \mathbb{E}_{a'\sim \pi(a\mid s)}Q(s, a')\right]$.
Such a formulation follows Advantage Weighted Regression (AWR)~\cite{peng2019advantage}, where Eqn.~(\ref{eqn:crr}) is derived as the closed-form solution to an optimization problem with $C_{\theta, \phi, \beta}^\pi: D_{KL}\left[\pi_{\theta}(\cdot | s)\parallel \pi_{\beta}(\cdot \mid s)\right]\leq \epsilon$ as the constraint.

For CRR, we additionally ablate the following two components:

\textit{Double Q learning.} As the original CRR adopts a single Q value structure, we additionally implement the double Q learning (\textbf{DQ}). Specifically, we use $\min(Q_1, Q_2)$ as the final prediction of the Q value.

\textit{Advantage normalization.} Advantage normalization (\textbf{AN}) normalizes the advantages in a batch~\cite{andrychowicz2020matters} for numerical stability of advantages. In our CRR implementation, we implement this by normalizing the exponential advantage $\exp A(s,a)$ over a batch.

\textbf{CRR$^+$}. The original CRR implementation considers only a single Q-value network with ResNet~\cite{he2016deep}-based architectures. For a fair comparison with other baselines, we focus on a simple 3-layer network. After sweeping, we observe double Q learning and layer normalization are the two most critical items for its performance boost, without which the performance would drop by 20.8\% and 16.5\%. On the other hand, the choices of activation functions, weight initialization schemes, and policy representations are less important considerations for CRR. As suggested by Fig.~\ref{fig:ablations}(a), CRR$^+$ achieves a significant performance boost from a total score of 525.2 to \textbf{702.3} (33.8\% improvement), and it even outperforms the SOTA method IQL (692.4).



\subsection{Case Study: Implicit Q Learning}\label{sect:iql}
Implicit Q-Learning (IQL)~\cite{kostrikov2022offline} is one of the SOTA methods in offline RL. Similar to OnestepRL~\cite{brandfonbrener2021offline}, IQL learns the policy in SARSA~\cite{sutton2018reinforcement} style without querying OOD actions during policy evaluation. To better approximate the maximum Q-value to allow multi-step dynamic programming, IQL performs expectile regression during policy evaluation. Specifically, it introduces an additional value function $V_\psi(s)$, and performs policy evaluation by
\begin{align}
    &\min_\psi \mathbb{E}_{s, a\sim \mathcal{D}}\left[ L_2^\tau (Q_\phi(s, a)) - V_\psi(s) \right]\nonumber\\
    &\min_\phi \mathbb{E}_{s, a, s'\sim \mathcal{D}}\left[ (r(s, a) +\gamma V_\psi(s') - Q_\phi(s, a))^2 \right],\label{eqn:iql}
\end{align}
where $L_2^\tau$ is the expectile regression loss defined as $L_2^\tau(x) = |\tau - \mathbbm{1}(x < 0)|x^2$ with hyper-paramter $\tau \in (0, 1)$ and the indicator function $\mathbbm{1}$. 
Intuitively, larger $\tau$ allows $V_\psi(s)$ to better approximate $\max\limits_a Q(s, a)$. As a result, IQL performs Q learning without querying OOD actions. For policy improvement, IQL follows AWR as described in Eqn.~(\ref{eqn:crr}). 

For IQL, we ablate its \textit{expectile rate} $\tau$ (\textbf{exp=$\tau$}) and its \textit{training scheme}. 
In particular, although in its original paper, the IQL algorithm follows OnestepRL which performs policy improvement with a fixed learned value network (\textbf{non-JNT}), IQL jointly learns its policy and value networks (\textbf{JNT}) in its official implementation.

We present the ablation results of IQL in Fig.~\ref{fig:iql}(a). We observe that the official configuration of IQL already gives the optimal performance and it is highly sensitive to the choice of implementations. 
Five alternative implementations will cause a significant performance drop of more than 100 points in its total return,
including the policy learning rate (drop by 14.4\%), the use of state-independent variance in policy distribution (drop by 14.6\%), high expectile rate $\tau$ (drop by  16.2\%), the absence of joint policy and value training scheme (drop by 16.3\%), and the absence of reward normalization (drop by 17.2\%).
Thus, careful implementations are critical to the strong performance of IQL.

\subsection{Recommendations of Implementation Choices}
Through careful ablations over implementation choices and hyper-parameter configurations, we observe that all three algorithms require a careful choice of implementations. In addition to the algorithm-specific choices, for shared implementation choices as described in Sect.~\ref{sect:study_design}, we summarize their influences across three algorithms in Fig.~\ref{fig:iql}(b). Specifically, CRR and CQL have a similar trend regarding implementation choices (6 out of 8), while IQL requires particular tuning on its own.
We make a list of recommended implementation choices for prototyping new algorithms.

\textit{Weight initialization schemes.} 
In contrast to on-policy algorithms~\cite{andrychowicz2020matters}, we observe that orthogonal initialization generally performs worse than Lecun initialization, regardless of the last layer weight scale (5 out of 6 cases).

\textit{Policy learning rate.} In Fig.~\ref{fig:iql}(b), we observe that although both CRR and CQL benefit from a smaller learning rate of 1e-4, IQL suffers from it with a sharp performance drop (-14.4\%). We would recommend trying both learning rates ($1e-4$ and $3e-4$) when implementing new algorithms.

\textit{Reward normalization scheme.} Similar to learning rates, reward normalization has diverged influence over algorithms. Both CRR and CQL observe clear performance drops on normalized rewards (-8\% and -8.7\%), while it significantly improves the performance of IQL (17.2\%). We would recommend trying out both choices for prototyping.

\textit{Layer normalization.} Layer normalization contributes significantly to the success of CRR on continuous control tasks (18.8\% improvement). Although both CQL and IQL encounter a performance drop, they are relatively minor (2.5\% and 2.4\%). Thus, we would recommend directly adding layer normalization to algorithms for prototyping.

\textit{Activation function.} According to our ablation results, the choice of activation function has a relatively smaller impact than other choices, where only CQL encounters a 5\% improvement switching from relu to elu. We would recommend directly starting with elu activation as a safer choice.

\textit{Policy distribution parameterization.} We observe that using state-dependent variance learning generally improves CQL and CRR, and has a negligible influence on IQL. Nevertheless, tanh-squashed distributions have a strong impact on all three algorithms. CRR and CQL benefit from the tanh transformations, while IQL suffers from it. To this end, we would recommend taking state-dependent variance as a start, while trying both tanh and non-tanh squashed distributions.

We do not exhaust all the implementation choices but we believe the above-mentioned ablations could help prototype offline RL algorithms more easily.
\begin{table*}[ht]
\fontsize{7}{7}\selectfont
\centering
\caption{Running average returns on D4RL Dataset. Specifically, we observe a discrepancy between the reproduced results and the reported results of CQL on antmaze-v0 tasks, where the results are formatted as reproduced/reported.}
\label{tab:d4rl}
\begin{tabular}{cccccccccc}
\toprule
                             & BC             & 10\% BC & Onestep        & MuZero & SAC  & TD3+BC        & IQL            & CRR$^+$           & CQL$^+$           \\
                             \midrule
halfcheetah-medium-v2        & 42.6           & 41.6    & 47.9           & 25.0   & 36.3 & \textbf{48.1} & 47.5           & 47.8           & \textbf{48.1}  \\
hopper-medium-v2             & 56.8           & 56.2    & 61.1           & 2.2    & 1.6  & 54.7          & 62.2           & 65.4           & \textbf{69.1}  \\
walker2d-medium-v2           & 69.5           & 71.4    & 78.1           & 0.1    & -0.2 & 76.3          & 81.3           & \textbf{84.2}  & 83.9           \\
halfcheetah-medium-replay-v2 & 36.6           & 37.1    & 37.1           & 40.8   & 24.7 & 43.8          & 43.6           & 45.2           & \textbf{46.0}  \\
hopper-medium-replay-v2      & 45.1           & 70.0    & 91.1           & 30.3   & 18.1 & 45.5          & 94.2           & 72.3           & \textbf{97.1}  \\
walker2d-medium-replay-v2    & 23.3           & 54.4    & 54.3           & 41.5   & 0.7  & 42.6          & 78.9           & \textbf{85.2}  & 81.7           \\
halfcheetah-medium-expert-v2 & 47.0           & 86.4    & \textbf{93.7}  & -1.2   & 7.9  & 92.4          & 89.3           & 91.6           & 83.7           \\
hopper-medium-expert-v2      & 55.4           & 102.1   & \textbf{102.4} & 1.8    & 1.7  & 87.6          & 84.1           & 101.7          & 98.2           \\
walker2d-medium-expert-v2    & 93.3           & 108.7   & 109.9          & -0.1   & 0.3  & 106.7         & 109.2          & \textbf{110.4} & 110.2          \\
\rowcolor[HTML]{EFEFEF} 
total (gym-locomotion)       & 469.6          & 627.9   & 675.6          & 140.2  & 91.1 & 597.7         & 690.2          & 703.9          & \textbf{717.9} \\
\midrule
pen-human-v0                 & \textbf{79.6}  & 1.9     & 72.4           & 0.81   & 1.8  & 5.9           & 75.0           & 67.8           & 77.4           \\
pen-cloned-v0                & 33.5           & -0.7    & 26.9           & 7.13   & -0.4 & 17.2          & \textbf{36.9}  & 36.8           & 22.9           \\
\rowcolor[HTML]{EFEFEF} 
total (adroit)               & \textbf{113.1} & 1.2     & 99.2           & 7.9    & 1.4  & 23.1          & 111.9          & 104.6          & 100.3          \\
\midrule
kitchen-complete-v0          & 64.9           & 3.8     & 66.0           & 0      & 0.9  & 2.2           & 67.4           & \textbf{72.5}  & 42.0           \\
kitchen-partial-v0           & 35.8           & 65.1    & 59.3           & 0.17   & 0.0  & 0.7           & 36.9           & 39.9           & \textbf{40.7}  \\
kitchen-mixed-v0             & 49.7           & 46.0    & 56.5           & 0      & 0.4  & 0.0           & 49.2           & \textbf{50.1}  & 45.7           \\
\rowcolor[HTML]{EFEFEF} 
total (kitchen)              & 150.4          & 114.9   & 181.7          & 0.2    & 1.4  & 2.9           & 153.4          & \textbf{162.4} & 128.5          \\
\midrule
antmaze-umaze-v0             & 52.0           & 61.3    & 62.4           & 0.0    & 0.1  & 40.2          & \textbf{81.0}  & 0.0            & 47.0 / 74.0    \\
antmaze-umaze-diverse-v0     & 46.1           & 52.4    & 44.7           & 0.0    & 0.0  & 58.0          & \textbf{59.6}  & 41.9           & 41.4 / 84.0    \\
antmaze-medium-play-v0       & 0.1            & 8.2     & 5.4            & 0.0    & 0.0  & 0.2           & \textbf{75.4}  & 0.0            & 0.3 / 61.2     \\
antmaze-medium-diverse-v0    & 0.3            & 3.2     & 1.8            & 0.0    & 0.0  & 0.0           & \textbf{74.8}  & 0.0            & 0.13 / 53.7    \\
antmaze-large-play-v0        & 0.0            & 1.2     & 0.1            & 0.0    & 0.0  & 0.0           & \textbf{47.3}  & 0.0            & 0.17 / 15.8    \\
antmaze-large-diverse-v0     & 0.0            & 2.1     & 0.9            & 0.0    & 0.0  & 0.0           & \textbf{45.9}  & 0.0            & 0 / 14.9       \\
\rowcolor[HTML]{EFEFEF} 
total (antmaze)              & 98.5           & 128.5   & 115.2          & 0.0    & 0.1  & 98.4          & \textbf{384.0} & 41.9           & 89.0 / 303.6   \\
\bottomrule
\end{tabular}
\end{table*}

\begin{table*}[ht]
\centering
\fontsize{8}{8}\selectfont
\caption{Running average returns on RL Unplugged Dataset.}
\label{tab:rlup}
\begin{tabular}{cccccccccc}
\toprule
          & BC     & 10\% BC & Onestep & MuZero          & SAC    & TD3+BC & IQL    & CRR$^+$   & CQL$^+$   \\
          \midrule
cartpole swingup & 280.2  & 534.5   & 198.2   & 427.1           & 816.6  & 405.1  & 787.8  & 437.9  & 318.4  \\
Walker Walk      & 327.4  & 352.0   & 485.2   & 894.6           & 849.8  & 312.4  & 866.7  & 905.2  & 905.2  \\
Finger Turn Hard & 131.9  & 236.9   & 137.9   & 358.8           & 528.5  & 225.6  & 337.8  & 163.7  & 134.9  \\
Cheetah Run      & 328.1  & 97.7    & 243.1   & 166.0           & 239.8  & 754.5  & 189.2  & 641.6  & 358.8  \\
Walker Stand     & 344.8  & 357.1   & 384.4   & 930.5           & 568.3  & 378.7  & 666.1  & 659.7  & 273.1  \\
Fish Swim        & 303.4  & 419.3   & 414.6   & 382.5           & 85.5   & 329.5  & 195.8  & 173.6  & 85.3   \\
\midrule
\rowcolor[HTML]{EFEFEF} 
total            & 1730.4 & 2036.8  & 2038.4  & \textbf{3159.5} & 3089.7 & 2425.7 & 3076.9 & 3000.3 & 2076.6\\
\bottomrule
\end{tabular}
\end{table*}

\section{Cross-Dataset Evaluation}
One clear difference between Offline RL and online RL is that the environment interactions are replaced with a fixed dataset. An agent's performance highly depends on the dataset used, and the generalization across datasets should be an essential consideration for the performance evaluation.

\subsection{Evaluation Setups}
\begin{wrapfigure}{r}{0.5\textwidth}
	\centering
	\begin{tabular}{cc}
    \includegraphics[width=0.45\linewidth]{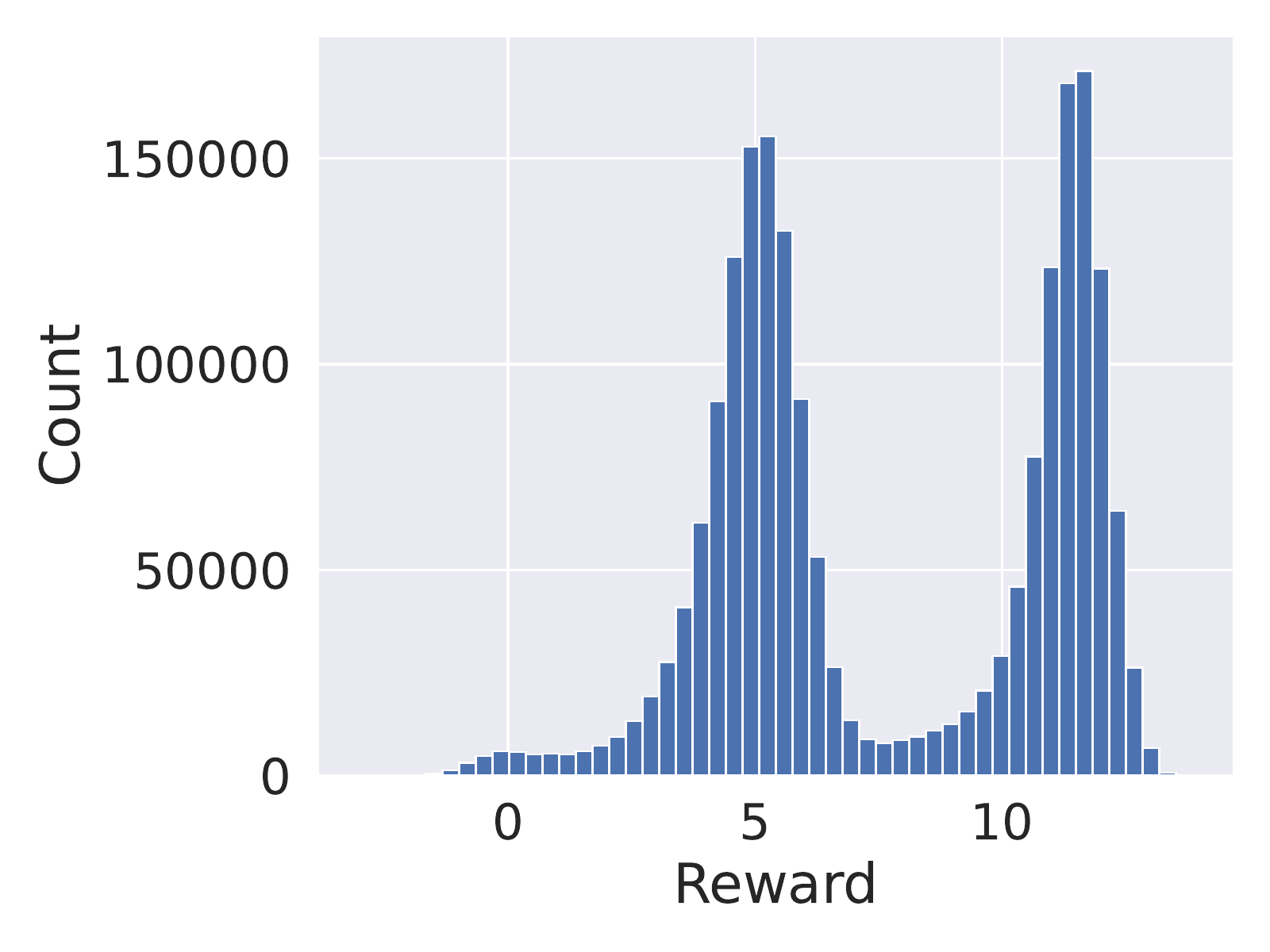} &
    \includegraphics[width=0.45\linewidth]{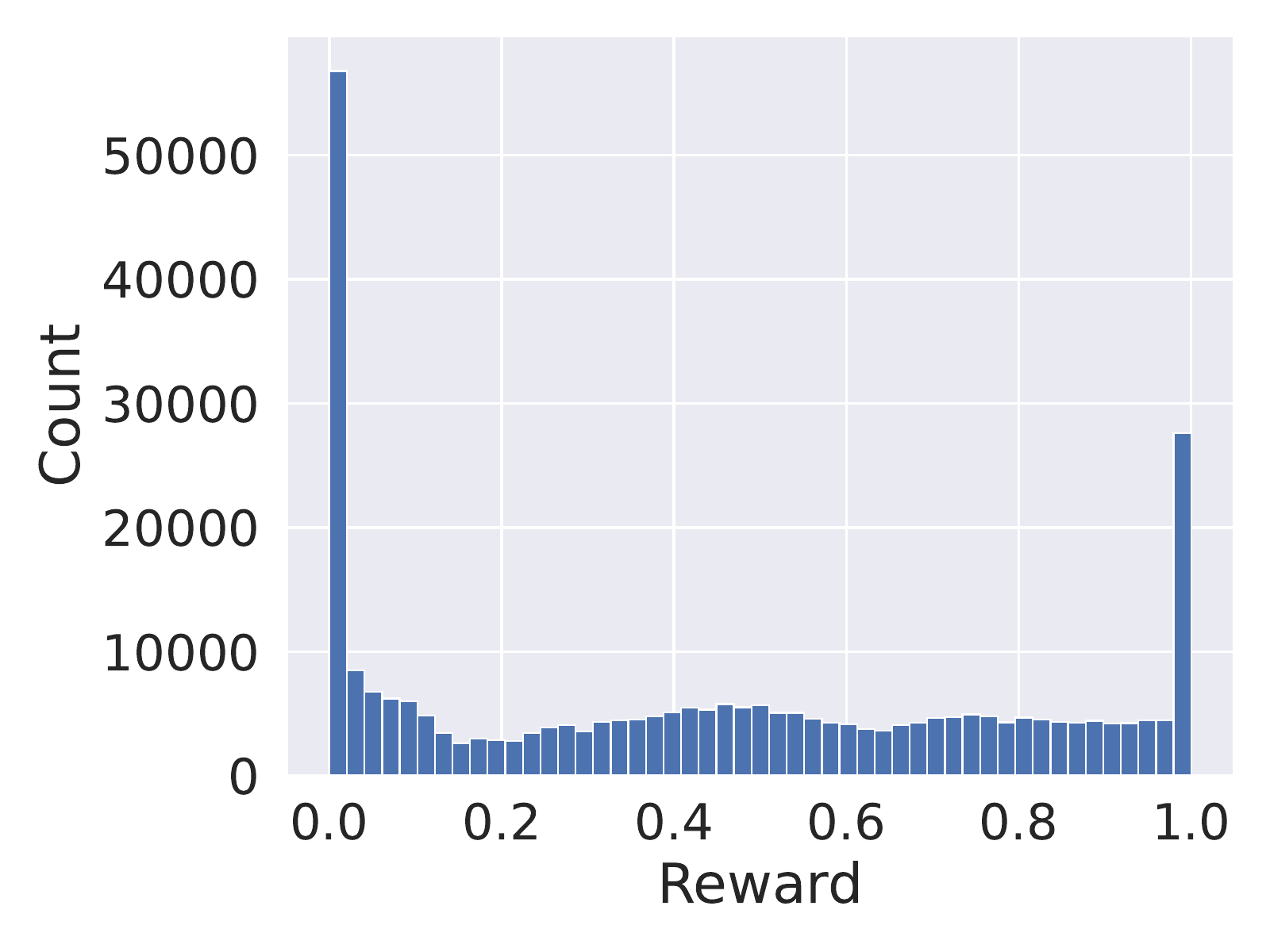} \\
    \small (a) halfcheetah-m-e-v2 & \small (b) cheetah-run
	\end{tabular}
	\centering
	\caption{\small Reward distributions of (a) halfcheetah-medium-expert-v2 from D4RL dataset and (b) cheetah-run from RL Unplugged dataset. These two tasks are conceptually similar but possess distinct data distributions.}
	\label{fig:reward_dist_main}
\end{wrapfigure}
\textbf{Dataset.}
We further evaluate a wide range of algorithms across two datasets with distinct distributions, RL Unplugged~\cite{gulcehre2020rl} and D4RL~\cite{fu2020d4rl}. RL Unplugged considers a wide range of tasks covering both continuous, \textit{e.g.}, DeepMind Control Suite~\cite{tassa2018deepmind}, and discrete actions, \textit{e.g.}, Atari games. The dataset is generated by down-sampling the replay buffer of an online agent, which contains a continuous spectrum of the past agent, \textit{i.e.}, the complete exploration process during the RL. Such a setup is useful for the fast iteration of new agents directly from previously collected experience, where the environment is slow to interact with. Differently, most tasks in D4RL focus on continuous control tasks with data generated by one or more fixed agents, except for the full-replay tasks. 
Hence, the data distribution of D4RL is significantly different from RL Unplugged, where data tends to be a multi-modal distribution. We visualize the reward and return distribution of the cheetah-run dataset of RL Unplugged and halfcheetah-medium-expert-v2 task in Fig.~\ref{fig:reward_dist_main}. We observe that halfcheetah-medium-expert-v2 has two modes, while the cheetah-run is closer to a uniform distribution. 

\textbf{Evaluation Protocol.}
One critical issue of the current offline RL research is that there exists no standard evaluation protocol. Some prior works consider the average of last return~\cite{kostrikov2022offline,kumar2020conservative,brandfonbrener2021offline} as their metrics, which, however, often fail to capture the stability of an algorithm. Others might consider the best evaluation return throughout the training. Nevertheless, it often overestimates the real performance of offline RL and is naturally infeasible for offline RL as an evaluation environment is missing in realistic setups. In this work, 
we propose to consider the \textit{last running average return} for offline RL evaluation. Specifically, we maintain a sliding window of size $L$, and for time step $t$, the average return is calculated as $\hat{R}_t = \frac{1}{L}\sum_{t'=t-L}^t R_{t'}$. The running average return at the last step $T$ is used as the final metric for evaluation. 
Running average return captures the stability of algorithms, and the use of the last step running average return better fits the offline RL.

\subsection{Algorithms for Evaluation}
Nevertheless, there is a clear isolation between algorithms evaluated on D4RL~\cite{kumar2020conservative,brandfonbrener2021offline,fujimoto2021minimalist,kostrikov2022offline} and the ones evaluated on RL Unplugged~\cite{wang2020critic,schrittwieser2021online}.
As a result, it is yet unclear how an algorithm generalizes across different data distributions.

We benchmark 8 algorithms across datasets. Specifically, in addition to the three algorithms discussed in the previous section, we consider the following algorithms in our empirical study: (1) Soft Actor-Critic (SAC)~\cite{haarnoja2018soft}, a popular off-policy RL algorithm. (2) MuZero Unplugged~\cite{schrittwieser2021online}, an offline RL algorithm without conservative constraints. MuZero Unplugged has demonstrated strong performance on RL Unplugged, but untested on D4RL. For a fair comparison with other algorithms, we use flat MLPs for MuZero, instead of ResNets as in its original implementation. (3) $x$\% BC, which stands for behavior cloning with trajectories of the top $x$\% accumulative return. Here, we choose $x=10$ following~\cite{kostrikov2022offline}. (4) TD3+BC~\cite{fujimoto2021minimalist}, which is a conservative policy improvement method by constraining the learning policy with a simple behavior cloning term. (5) OnestepRL~\cite{brandfonbrener2021offline}, which implements implicit conservative policy improvement with SARSA-style Q-learning.

We strictly follow the official implementations and re-implement all algorithms in JAX~\cite{jax2018github} with the neural network library Flax~\cite{flax2020github}. We search the hyper-parameters for all algorithms and fix the parameters with the best overall performance. More details are available in the appendix.

\subsection{Results and Discussions: D4RL}
We present the \textit{last step running average return} on the D4RL dataset in Tab.~\ref{tab:d4rl}.

On gym-locomotion tasks, CQL$^+$ achieves the best total running average return (717.9) across all algorithms, outperforming the SOTA method, IQL, by a large margin. 
Interestingly, CRR$^+$, a direct variant of the off-policy algorithm AWR, produces a stable and strong performance with careful implementation tuning on locomotion tasks and outperforms IQL.
However, TD3+BC achieves a much worse performance (597.7) than its best-achieved performance (737.8, in the appendix). Such a phenomenon is caused by the instability of TD3+BC. We suspect the underlying reason is that directly regularizing the learning policy by behavior cloning leads to a conflicting optimization objective and eventually hinders policy improvement.
For adroit and kitchen, we observe that OnestepRL, CRR$^+$, and IQL are significantly better than the other algorithms. All these three algorithms follow the policy improvement of AWR, \textit{i.e.}, weighted behavior cloning (Eqn.~\ref{eqn:crr}), which naturally avoids querying the OOD actions and better fits these environments with sparser reward.

We observe on antmaze environments with long-delayed rewards, IQL consistently outperforms all other algorithms. Most algorithms suffer from a performance drop as training proceeds. In particular, we observe an apparent discrepancy between the reproduced and reported results on CQL on antmaze. We followed the official configurations and implementations and performed a careful hyper-parameter sweeping but failed to reproduce the results.
Interestingly, although MuZero Unplugged claims its generality on both online and offline setups, we observe that MuZero generally fails on D4RL tasks, except for medium-replay environments where the replay data is present.

\subsection{Results and Discussions: RL Unplugged}
The \textit{last step running average return} on RL Unplugged dataset
are presented in Tab.~\ref{tab:rlup}. 

Contradictory results to D4RL are observed on RL Unplugged. MuZero has achieved relatively strong performance,  even with the much smaller network re-implemented than its original implementations. However, on D4RL, MuZero failed except for the medium-replay tasks. To our surprise, as an off-policy algorithm, SAC observes a similar trend: it fails on D4RL but achieves strong performance on RL Unplugged, outperforming all other algorithms except MuZero. CRR$^+$ and IQL also transfer their strong performance to RL Unplugged, while CQL$^+$ and OnestepRL achieve much worse performances than others. 

\subsection{Discussions and Recommendations}
We have observed distinct results on D4RL and RL Unplugged datasets. Specifically, algorithms that succeed on D4RL, including CQL$^+$ and OnestepRL, fail to transfer their success to RL Unplugged. On the contrary, algorithms that fail on D4RL, including SAC and MuZero, achieve strong performance in RL Unplugged. By summarizing their shared properties, we draw the following conclusions.


\textit{MuZero is closer to an off-policy algorithm}. As we have observed on both the D4RL dataset and RL Unplugged dataset, MuZero has shown the same trend in its performance with SAC, rather than other offline RL algorithms. It fails to handle the data distributions generated by fixed agents, which provides no coverage of the exploration process of a normal RL agent. To fix this issue, conservative constraints could be considered during the Monte-Carlo Tree Search (MCTS) process of MuZero.

\textit{Overly conservative constraints will have adverse effects on replay data.} On the contrary to D4RL, for replay datasets, overly conservative algorithms, \textit{e.g.}, CQL, work worse than the standard off-policy RL algorithms, \textit{e.g.}, SAC. This is potentially because they estimate an overly loose lower bound of the actual policy/value, and thus fail to exploit the rare but good data in the dataset effectively.

\textit{AWR-style policy improvement gives the best generalization across data distributions.} Different from standard Q-learning, algorithms with AWR-style policy improvement, \textit{e.g.}, IQL, and CRR, achieve stable performance on both RL Unplugged and D4RL datasets. This is potentially because the AWR-style policy improvement naturally encodes an implicit conservative constraint for policy and thus minimizes the need for additional constraints. As a result, they are minimally conservative and can adapt to different data distributions with minimal effort.

\textit{Recommendation of algorithms.}  When starting a practical project with offline RL, given the above analysis, we would recommend CRR and IQL as the go-to algorithms, which are efficient, general across different data distributions, and have strong performance on most of the tasks. Although in all experiments, we observe IQL still gives the best overall best performance, which demonstrates its algorithmic level advancements, IQL is more sensitive to hyper-parameters and would need careful implementation and hyper-parameter tuning in practical use.
\section{Conclusions}
We conduct a large-scale empirical study on the low-level implementation choices of offline RL algorithms and benchmark 8 algorithms across D4RL and RL Unplugged datasets. We show that low-level implementations are crucial to the performance of offline RL algorithms. By sweeping 20 low-level implementation choices, we present a guidebook for implementing new offline algorithms, as well as CRR$^+$ and CQL$^+$ that outperform the SOTA algorithm following the guidebook. Lastly, we benchmark 8 popular algorithms across RL Unplugged and D4RL, and show that conservative policy improvement algorithms demonstrate the best transferrability across datasets. We hope this work would shed light on the future research of offline RL.

\clearpage
\bibliographystyle{plain}

\newpage
\appendix
\section*{Broader Impact}
In this work, we conduct a thorough empirical study about the low-level implementation choices and a cross-dataset benchmark for offline RL algorithms. Although the existing experiments are based on simulated tasks, offline RL provides a principled framework for learning intelligent agents from real-world data, which could potentially cause negative effects to safety and democratic privacy. 

\section{Experiment Details}
In this section, we introduce the benchmark and the corresponding experiment details.

\subsection{Experiment Setups}
To build a unified and efficient framework, we build the codebase with JAX for its high computational efficiency and simplicity. For a fair comparison, all algorithms share the same data pre-processing pipeline, network architecture, evaluation pipeline, and etc. For each environment, each evaluation result is computed by averaging the accumulative return over 100 evaluation episodes and 3 random seeds. We now give detailed descriptions of algorithms and their implementation details as follows.

\subsection{Algorithms Details}
For all algorithms, we follow~\cite{kostrikov2022offline} where a three-layer network with hidden dimension 256 is used. Specifically, the network has an architecture of (256-ACT-256-ACT-$2\times D$), where ACT, \textit{e.g.}, relu or elu, is the activation function and $D$ is the output dimension. For algorithms that require layer normalization, it is applied before the activation function. For Q-value networks or value networks, the output dimension $D=1$; Gaussian policies $\pi_\theta(a\mid s ) = \mathcal{N}(\mu_\theta, \sigma_\theta)$ are used for policy networks, where the output dimension $D=\mbox{action dimension}$ and both the mean and variance are computed as parameters. As for all actions, $a\in (-1, 1)$, transformations are required for the output Gaussian policy. Specifically, we consider two approaches: 1) tanh-squashed distribution $\pi_\theta(a\mid s) = \mathrm{tanh}(\mathcal{N}(\mu, \sigma))$, which can be implemented as a Transformed distribution; 2) clipped distribution with tanh-transformed mean: $\pi_\theta(a\mid s) = \mathrm{clip}(\mathcal{N}(\mathrm{tanh}(\mu), \sigma), -1, 1)$. Besides, IQL requires a state-independent variance, and in this case, the output of the network architecture of the policy network is changed to (256-ACT-256-ACT-$D$), with an additional $D$-dimensional learnable vector as the standard deviation. We train all algorithms with Adam optimizer~\cite{kingma2014adam}. Specifically, we use a learning rate of $3e-4$ for value functions and for policy, we either use $1e-4$ or $3e-4$ with a cosine scheduler.


\subsubsection{Behavior Cloning}
Behavior cloning simply corresponds to the maximum likelihood estimation of the dataset actions. Specifically, given a parameterized policy $\pi_\theta(a\mid s)$, we optimize the policy by
\begin{equation}
    \pi^* = \arg\max_{\pi_\theta}\mathbb{E}_{s,a\sim\mathcal{D}}\left[ \log\pi_\theta(a\mid s) \right]\label{eqn:bc}
\end{equation}
There are two ways to implement the behavior cloning: 1) directly apply Eqn.~\ref{eqn:bc}. 2) optimize the mean squared error of the Monte-Carlo samples and optimize the policy via reparameterization trick~\cite{kingma2013auto}, \textit{i.e.}, $\pi^* = \arg\min_{\pi_\theta}\mathbb{E}_{s, a\sim\mathcal{D}}\left[(\hat{a} - a)^2\right], \hat{a}\sim \pi_\theta(a\mid s)$. Theoretically, these two approaches are equivalent but empirically, we found the second approach generally works better. We adopt the second approach across our experiment. 

As a commonly adopted baselines, $x\%$BC is trained with the trajectories with the top $x\%$ highest returns. Intuitively, such an approach filters out the low-quality data but will be bounded by the highest achievable performance of the dataset.

\subsubsection{TD3+BC}
TD3~\cite{fujimoto2018addressing} is a popular off-policy RL algorithm for continuous control. The core idea of TD3 is to address the value overestimation problem by computing the TD target by taking the minimum of two isolated value networks. Specifically, TD3 learns the Q-value network with
\begin{align}
    &Q^* = \arg\min_Q \mathbb{E}_{\tau}\left[ (Q(s, a) - (r(s, a) + \gamma \hat{Q}(s', a')))^2 \right], \quad a'\sim \pi (\cdot\mid s')\nonumber\\
    &\hat{Q}(s', a') = \min(Q_1(s', a'), Q_2(s', a')),
\end{align}
where $Q_1$ and $Q_2$ are two independent Q networks, and in particular, $\pi(\cdot\mid s)$ is a deterministic policy. Together with the policy evaluation step that learns the Q value networks, policy network is trained by maximizing the Q values. Specifically, TD3 takes \textbf{only} $Q_1$ for policy improvement
\begin{equation}
    \pi^* = \arg\max_\pi \mathbb{E}_\tau \left[ Q_1(s, a) \right], \quad a\sim \pi(\cdot\mid s)\label{eqn:td3_pi}
\end{equation}

On top of TD3, TD3+BC~\cite{fujimoto2021minimalist} directly extends to the offline RL setups by adding an additional behavior cloning term to the Eqn.~\ref{eqn:td3_pi}
\begin{equation}
    \pi^* = \arg\max_\pi \mathbb{E}_{s,a\sim\mathcal{D}} \left[ Q_1(s, a') - \alpha(a' - a)^2 \right], \quad a'\sim \pi(\cdot\mid s).
\end{equation}
To stabilize the training, during policy improvement, TD3+BC normalizes its Q values by $\frac{1}{n}\sum\limits_{i=1}^N |Q_1^i(s, a)|$, where $N$ is the batch size.

\subsubsection{Soft Actor-Critic}
Soft Actor-Critic (SAC) is a popular off-policy RL method. SAC considers entropy regularized reinforcement learning, where it changes the RL problem to
\begin{equation*}
    \pi^* = \arg\max_\pi\mathbb{E}_{\tau\sim\pi}\left[r(s, a) + \alpha H(\pi(\cdot \mid s))\right], \tau = (s, a, s', ...).
\end{equation*}
Intuitively, entropy-regularized RL encourages an agent to maximize the accumulative return of trajectories and, at the same time, maximize the entropy of its policy. In addition, this can be interpreted as RL with an entropy-regularized reward function $r'(s, a) = r(s, a) + \alpha H(\pi(\cdot\mid s))$.

SAC follows TD3 on its general framework but makes the following modifications. First, SAC adopts a stochastic policy, where $\pi(a\mid s) = \mathcal{N}(\mu, \sigma)$ and as a result, the entropy can be computed. Second, instead  of using $Q_1$ for policy improvement, SAC performs policy improvement with 
\begin{equation}
    \pi^* = \arg\max_\pi \mathbb{E}_\tau \left[ \min(Q_1(s, a), Q_2(s, a)) + H(\pi(\cdot \mid s))\right], \quad a\sim \pi(\cdot\mid s)\label{eqn:sac_pi}
\end{equation}
    
\subsubsection{Conservative Q Learning}
As discussed in Sect.~\ref{sect:cql}, CQL is one of the most popular offline RL algorithms. CQL implements a conservative policy evaluation step to handle the value overestimation issue in offline RL
\begin{equation}
    C_{\theta, \phi, \beta}^Q = \mathbb{E}_{s\sim \mathcal{D}}\left[\log\sum\limits_{a}\exp Q(s, a) - \mathbb{E}_{a\sim \pi_\beta (a\mid s)}\left[ Q(s, a) \right]\right].
\end{equation}
In practical implementations, CQL is developed based on SAC, but with a modified policy evaluation step
\begin{equation}
\begin{aligned}
    \arg\min\limits_{Q}\mathbb{E}_{s,a, s'\sim \mathcal{D}}\left[\log\sum\limits_{a}\exp Q(s, a) - Q(s, a) + \right. \\
    \left. \left(Q(s, a) - (r(s, a) + \gamma \min(Q_1(s', a'), Q_2(s', a')))\right)^2\right]
\end{aligned}
\end{equation}
for both $Q_1$ and $Q_2$, where $a'\sim \pi (\cdot\mid s')$. In addition, to approximate $\log\sum\limits_a\exp Q(s, a)$, CQL adopts importance sampling
\begin{equation}
\begin{aligned}
    \log\sum\limits_a\exp Q(s, a)\approx \log \left( \sum\limits_{a\sim\pi(a\mid s)} \frac{1}{\pi(a\mid s)}\exp Q(s, a) + \sum\limits_{a\sim\pi(a\mid s')} \frac{1}{\pi(a\mid s')}\exp Q(s, a) \right. \\
    \left. + \sum\limits_{a\sim\mathcal{U}(a)} \frac{1}{\mathcal{U}(a)}\exp Q(s, a) \right),
\end{aligned}
\end{equation}
where $\pi(a\mid s)$ is the current policy, $\pi(a\mid s')$ is the next policy, $\mathcal{U}(a)$ is a uniform distribution with range $(-1, 1)$. Specifically, for each distribution, the original CQL samples 10 actions, while CQL$^+$ samples 50 actions, which turns out to produce a better performance.

\subsubsection{Critic Regularized Regression}
As discussed in Sect.~\ref{sect:crr}, CRR follows AWR~\cite{peng2019advantage} for its policy improvement. Specifically, it considers a constrained policy improvement problem
\begin{align}
    \arg\max_\pi \int_s d_\mu (s) \int_a \pi(a\mid s)A(s, a)da ds, \quad\mathrm{s.t.}\quad \int_s d_\mu(s)D_{KL}\left[ \pi(\cdot\mid s)\parallel\mu(\cdot\mid s) \right]\leq \epsilon,
\end{align}
where $d_\mu(s)$ is the state distribution induced by a behavior policy $\mu$. By solving this constrained optimization problem, we obtain a \textit{one-step improved} policy
\begin{equation}
     \pi^*(a\mid s) = \frac{1}{Z(s)}\mu(a\mid s)\exp\left( \frac{1}{\beta}A(s, a) \right),
\end{equation}
with $Z(s)$ being the partition function and $A(s, a)$ being the advantage. As a result, AWR performs the policy improvement by projecting, \textit{i.e.}, distilling, the one-step improved policy $\pi^*$ onto the learning policy $\pi$ by
\begin{align}
    &\arg\min_\pi \mathbb{E}_{s\sim \mathcal{D}}\left[ D_{KL}\left[ \pi^*(\cdot\mid s)\parallel \pi(\cdot\mid s) \right] \right]\nonumber\\
    =&\arg\max_\pi \mathbb{E}_{s\sim \mathcal{D}, a\sim \mu(\cdot\mid s)}\left[ \log\pi(a\mid s)\exp A(s, a) \right]\label{eqn:awr}
\end{align}

Following AWR, CRR performs policy improvement in the form of Eqn.~\ref{eqn:awr}. Specifically, it computes the advantage with $A(s, a) = Q(s, a) - \mathbb{E}_{a'\sim \pi(\cdot\mid s)}\left[Q(s, a')\right]$. Besides, in the context of offline RL, $\mu(a\mid s)$ corresponds to the behavior policy $\pi_\beta(a\mid s)$ that generates the dataset. As a result, by adopting the AWR-style policy improvement, CRR naturally imposes an implicit constraint for conservative policy improvement. 

However, in its original implementation, CRR adopts a single Q value network, and according to our ablation study, it produces a poor performance on D4RL dataset. CRR$^+$, on the contrary, significantly improves its performance by simply applying double Q learning. Specifically, we follow the TD3 on computing the TD target, and during policy improvement, we compute the advantage as
\begin{equation}
\label{eqn:adv-value-sample}
    A(s, a) = \min(Q_1(s, a), Q_2(s, a)) - \frac{1}{N}\sum\limits_{i=1}^N\min(Q_1(s, a^i), Q_2(s, a^i)), \quad a^i\sim \pi(\cdot\mid s)
\end{equation}

\subsubsection{OnestepRL}
OnestepRL adopts a similar policy improvement scheme with CRR, \textit{i.e.}, performing policy improvement in the way of AWR. 
OnestepRL focuses on on-policy Q evaluation of the behavior policy and only performs one step of policy improvement. Specifically, OnestepRL fits a behavior policy $\beta$ through maximum likelihood and trains the policy evaluation steps to estimate $Q^\beta$, only querying in-distribution actions. 
During the policy improvement, the advantage is computed using value sample, which is similar to Eqn.~\ref{eqn:adv-value-sample} but without the use of double Q learning.
In its original implementation, OnestepRL adopts a network that is larger than the standard network, which is a three layer network with 1024 hidden units.

\subsubsection{Implicit Q Learning}
IQL also follows the AWR for its policy improvement. Unlike OnestepRL which considers Bellman's expectation equation, IQL aims to perform policy evaluation with Bellman's optimality equation that enables multi-step Q learning, while querying only in-distribution samples. Specifically, as discussed in Sect.~\ref{sect:iql}, IQL performs policy evaluation follows Eqn.~\ref{eqn:iql}. However, in practice, IQL also adopts double Q learning. Besides, the training scheme of IQL requires joint optimization of the Q value function, value function, and policy, instead of the isolated training scheme of OnestepRL. 

\subsubsection{MuZero Unplugged}

Muzero Unplugged mainly depends on the \textit{Reanalyse} algorithm in ~\cite{schrittwieser2021online} and ~\cite{schrittwieser2020mastering}. Specifically, Muzero unplugged contains three parameterized models:
\begin{align}
    \text{State Encoder}: &\quad {z}_t = h_\theta ({s}_t) \\
    \text{Transition Model}: &\quad {z}_{t+1} = g_\theta ({s}_t, {a}_t) \\
    \text{Prediction Model}: &\quad p_t, r_t, v_t = f_\theta ({z}_t).
\end{align}
where $z_t$ is the latent embedding of state $s_t$, $p_t$ is the policy prediction, $r_t$ and $v_t$ is the predicted reward and value. Muzero unplugged jointly optimizes its parameters $\theta$ to minizine the following loss at every timestep $t$, by unrolling a model for $K$ steps in the future, 
\begin{equation}
    \ell_t(\theta) = \sum_{k=0}^K \ell^p(\pi_{t+k}, p_t^k) + \sum_{k=0}^K \ell^v(z_{t+k}, v^k_t) + \sum_{k=1}^K \ell^r(u_{t+k}, r^k_t),
\end{equation}
where $p_t^k, v_t^k,$ and $r_t^k$ are the policy, value, and reward prediction produced by the k-step unrolled model. $p=\pi_{t+k}$ is the improved policy generated by MCTS, $z_{t+k}$ is the n-step return starting from timestep $t+k$. $u_{t+k}$ is the ground-truth reward signal at the corresponding timestep. All these targets are from the real trajectory. 

Since we are working with continuous action spaces, we follow the sample-based MCTS introduced in ~\cite{hubert2021learning}. Instead of searching over the whole action space, it samples $N$ actions from the policy network and performs MCTS over the sampled $N$ actions. To this end, it modifies the UCB as follows: 
\begin{equation}
    a^* = \arg\max_{a} \left[Q(s,a) + \frac{\hat{\beta}(a|s)}{\beta(a|s)} \pi(a|s) \cdot \frac{\sqrt{\sum_b N(s,b)}}{1 + N(s,a)} \left(c_1 + \log \left(\frac{\sum_b N(s,b) + c_2 + 1}{c_2} \right) \right) \right],
\end{equation}
where $\beta$ is a sampling policy and  $\hat{\beta}(a|s) = \frac{1}{N} \sum_i \delta_{a=a_i}$ is the corresponding empirical distribution which is non-zero only on the sampled actions $\{a_i\}_{i=1}^N$. $\pi$ is the output of the policy prediction. In our implementation, $\beta$ is chosen to be $\pi$. 

We use a three-layer MLP with relu activation and a hidden size of 256 as the state encoder. For the transition model and prediction model, a two-layer MLP is used. The policy prediction is parameterized with a Tanh squashed Gaussian distribution. We set the number of samples $N$ to 20 throughout all experiments.


\section{Additional Results}
In this section, we present the additional results of the benchmark. 

\subsection{Alternative Metrics}
\begin{table*}[ht]
\fontsize{7}{8}\selectfont
\centering
\caption{Last average returns on D4RL Dataset.}
\label{tab:d4rl_last}
\begin{tabular}{cccccccccc}
\toprule
                             & BC            & 10\% BC & Onestep        & MuZero & SAC  & TD3+BC        & IQL            & CRR$^+$           & CQL$^+$           \\
                             \midrule
halfcheetah-medium-v2        & 42.6          & 41.8    & 47.9           & 27.7   & 37.2 & \textbf{48.2} & 47.5           & 47.8           & \textbf{48.2}  \\
hopper-medium-v2             & 56.8          & 56.7    & 61.1           & 2.1    & 1.5  & 55.9          & 63.4           & 64.4           & \textbf{73.7}  \\
walker2d-medium-v2           & 69.5          & 71.4    & 70.6           & 1.5    & -0.1 & 78.3          & 80.7           & \textbf{84.0}  & 83.8           \\
halfcheetah-medium-replay-v2 & 36.6          & 37.1    & 37.1           & 43.1   & 26.8 & 43.5          & 43.1           & 45.0           & \textbf{46.4}  \\
hopper-medium-replay-v2      & 45.1          & 70.0    & 91.1           & 25.7   & 13.4 & 68.1          & 94.3           & 67.7           & \textbf{96.0}  \\
walker2d-medium-replay-v2    & 23.3          & 54.4    & 54.3           & 63.4   & 1.2  & 48.1          & 78.9           & 84.7           & \textbf{84.9}  \\
halfcheetah-medium-expert-v2 & 45.8          & 86.4    & 93.7           & -1.2   & 9.4  & 92.2          & 89.9           & \textbf{92.5}  & 86.3           \\
hopper-medium-expert-v2      & 55.4          & 102.1   & 102.4          & 1.7    & 1.3  & 75.1          & 89.5           & \textbf{105.2} & 102.9          \\
walker2d-medium-expert-v2    & 86.0          & 108.5   & 109.9          & -0.4   & -0.2 & 77.3          & 109.2          & \textbf{110.8} & 109.6          \\
\rowcolor[HTML]{EFEFEF} 
total (gym-locomotion)       & 461.0         & 628.2   & 668.1          & 163.7  & 90.3 & 586.8         & 696.5          & 702.3          & \textbf{731.9} \\
\midrule
pen-human-v0                 & 66.9          & -0.4    & 72.7           & -1.865 & -0.5 & 5.2           & 74.3           & 67.4           & \textbf{74.6}  \\
pen-cloned-v0                & \textbf{38.5} & -1.0    & 31.2           & 5.29   & -0.7 & 18.4          & 36.8           & 34.9           & 26.2           \\
\rowcolor[HTML]{EFEFEF} 
total (adroit)               & 105.4         & -1.4    & 103.9          & 3.4    & -1.2 & 23.5          & \textbf{111.1} & 102.3          & 100.9          \\
\midrule
kitchen-complete-v0          & 62.7          & 4.1     & 62.9           & 0      & 0.0  & 3.8           & 67.3           & 78.1           & 39.7           \\
kitchen-partial-v0           & 34.2          & 65.6    & \textbf{59.1}  & 0      & 0.0  & 0.0           & 36.5           & 38.7           & 50.8           \\
kitchen-mixed-v0             & 50.6          & 45.5    & \textbf{55.8}  & 0      & 0.0  & 0.0           & 49.4           & 51.8           & 42.6           \\
\rowcolor[HTML]{EFEFEF} 
total (kitchen)              & 147.4         & 115.2   & \textbf{177.8} & 0.0    & 0.0  & 3.8           & 153.3          & 168.5          & 133.0          \\
\midrule
antmaze-umaze-v0             & 51.3          & 56.3    & 68.7           & 0.0    & 0.3  & 45.0          & \textbf{82.7}  & 0.0            & 67.0           \\
antmaze-umaze-diverse-v0     & 42.7          & 53.0    & 41.0           & 0.0    & 0.0  & \textbf{71.3} & 52.3           & 37.7           & 0.0            \\
antmaze-medium-play-v0       & 0.3           & 6.0     & 10.3           & 0.0    & 0.0  & 0.3           & \textbf{76.7}  & 0.0            & 0.0            \\
antmaze-medium-diverse-v0    & 0.0           & 1.7     & 3.0            & 0.0    & 0.0  & 0.0           & \textbf{78.3}  & 0.0            & 0.0            \\
antmaze-large-play-v0        & 0.0           & 1.7     & 0.0            & 0.0    & 0.0  & 0.0           & \textbf{46.3}  & 0.0            & 0.0            \\
antmaze-large-diverse-v0     & 0.0           & 1.7     & 3.0            & 0.0    & 0.0  & 0.0           & \textbf{51.3}  & 0.0            & 0.0            \\
\rowcolor[HTML]{EFEFEF} 
total (antmaze)              & 94.3          & 120.3   & 126.0          & 0.0    & 0.3  & 116.7         & \textbf{387.7} & 37.7           & 67.0    \\
\bottomrule
\end{tabular}
\end{table*}

\begin{table*}[ht]
\centering
\fontsize{8}{8}\selectfont
\caption{Last average returns on RL Unplugged Dataset.}
\label{tab:rlup_last}
\begin{tabular}{cccccccccc}
\toprule
          & BC     & 10\% BC & Onestep & MuZero          & SAC    & TD3+BC & IQL    & CRR$^+$   & CQL$^+$   \\
          \midrule
cartpole swingup & 262.7  & 522.5   & 196.5   & 399.6  & 848.3           & 380.1  & 788.2  & 531.3  & 380.9  \\
Walker Walk      & 306.4  & 360.1   & 516.3   & 902.6  & 841.4           & 310.6  & 861.9  & 892.5  & 450.8  \\
Finger Turn Hard & 164.5  & 266.9   & 140.6   & 298.9  & 530.8           & 233.2  & 316.5  & 149.8  & 97.5   \\
Cheetah Run      & 343.7  & 95.2    & 258.3   & 169.8  & 218.2           & 767.4  & 213.9  & 646.4  & 337.2  \\
Walker Stand     & 341.2  & 344.6   & 382.7   & 938.2  & 575.1           & 402.4  & 674.1  & 667.9  & 257.2  \\
Fish Swim        & 296.7  & 448.5   & 417.5   & 328.2  & 78.8            & 312.9  & 187.2  & 167.5  & 93.9   \\
\rowcolor[HTML]{EFEFEF} 
total            & 1730.5 & 2073.5  & 2096.4  & 3037.3 & \textbf{3093.7} & 2426.9 & 3073.6 & 3075.0 & 1618.3\\
\bottomrule
\end{tabular}
\end{table*}

\begin{table*}[ht]
\fontsize{7}{8}\selectfont
\centering
\caption{Best average returns on D4RL Dataset.}
\label{tab:d4rl_best}
\begin{tabular}{cccccccccc}
\toprule
                             & BC            & 10\% BC & Onestep        & MuZero & SAC  & TD3+BC        & IQL            & CRR$^+$           & CQL$^+$           \\
                             \midrule
halfcheetah-medium-v2        & 43.2          & 43.0    & 48.1           & 34.0   & 45.9  & \textbf{48.5}  & 47.7           & 48.4           & \textbf{48.5} \\
hopper-medium-v2             & 68.3          & 74.6    & 75.6           & 3.2    & 14.6  & 64.0           & 80.0           & 79.6           & \textbf{82.6} \\
walker2d-medium-v2           & 75.9          & 78.8    & 73.5           & 7.5    & 7.9   & 82.2           & 83.9           & 85.4           & \textbf{85.7} \\
halfcheetah-medium-replay-v2 & 38.3          & 39.3    & 38.8           & 44.0   & 49.8  & 44.7           & 45.4           & 45.8           & \textbf{46.8} \\
hopper-medium-replay-v2      & 73.5          & 85.2    & 97.3           & 37.1   & 80.0  & 95.0           & 99.5           & \textbf{101.4} & 100.2         \\
walker2d-medium-replay-v2    & 46.0          & 68.3    & 60.4           & 76.4   & 20.9  & 85.7           & 86.9           & \textbf{90.8}  & 88.4          \\
halfcheetah-medium-expert-v2 & 46.8          & 90.1    & 94.5           & 3.8    & 13.7  & 95.6           & 91.2           & \textbf{94.7}  & 92.5          \\
hopper-medium-expert-v2      & 66.1          & 111.5   & 110.7          & 2.9    & 19.7  & 110.2          & 107.3          & \textbf{111.6} & 111.3         \\
walker2d-medium-expert-v2    & 106.3         & 109.5   & 110.8          & 0.6    & 9.4   & \textbf{111.9} & 111.3          & 111.3          & 111.2         \\
\rowcolor[HTML]{EFEFEF} 
total (gym-locomotion)       & 564.2         & 700.3   & 709.7          & 209.6  & 261.9 & 737.8          & 753.1          & \textbf{769.0} & 767.2         \\
\midrule
pen-human-v0                 & \textbf{97.5} & 12.4    & 89.5           & 31.27  & 11.5  & 42.3           & 91.6           & 82.9           & 92.5          \\
pen-cloned-v0                & 48.5          & 5.2     & 45.5           & 33.27  & 28.1  & 51.8           & \textbf{71.7}  & 64.2           & 51.9          \\
\rowcolor[HTML]{EFEFEF} 
total (adroit)               & 146.1         & 17.6    & 134.9          & 64.5   & 39.6  & 94.1           & 163.3          & \textbf{147.1} & 144.4         \\
\midrule
kitchen-complete-v0          & 78.1          & 31.1    & 89.3           & 17.5   & 24.7  & 24.9           & 79.7           & \textbf{89.4}  & 68.5          \\
kitchen-partial-v0           & 42.2          & 73.5    & \textbf{67.3}  & 9.167  & 15.5  & 17.2           & 47.2           & 65.2           & 64.1          \\
kitchen-mixed-v0             & 54.2          & 53.8    & \textbf{65.4}  & 4.167  & 14.7  & 22.0           & 55.2           & 61.3           & 65.3          \\
\rowcolor[HTML]{EFEFEF} 
total (kitchen)              & 174.4         & 158.4   & \textbf{222.1} & 30.8   & 54.8  & 64.1           & 182.0          & 215.9          & 197.9         \\
\midrule
antmaze-umaze-v0             & 67.7          & 71.3    & 78.3           & 0.0    & 0.3   & \textbf{96.0}  & 87.7           & 44.3           & 72.0          \\
antmaze-umaze-diverse-v0     & 60.3          & 62.7    & 57.7           & 0.0    & 0.0   & \textbf{92.3}  & 73.7           & 60.3           & 47.0          \\
antmaze-medium-play-v0       & 0.7           & 15.7    & 12.3           & 0.0    & 0.0   & 2.0            & \textbf{82.3}  & 0.0            & 7.3           \\
antmaze-medium-diverse-v0    & 2.0           & 6.7     & 4.7            & 0.0    & 0.0   & 1.7            & \textbf{84.7}  & 0.0            & 4.3           \\
antmaze-large-play-v0        & 0.0           & 6.3     & 0.7            & 0.0    & 0.0   & 0.0            & \textbf{59.3}  & 0.0            & 4.3           \\
antmaze-large-diverse-v0     & 0.0           & 9.0     & 3.0            & 0.0    & 0.0   & 0.3            & \textbf{62.3}  & 0.3            & 2.3           \\
\rowcolor[HTML]{EFEFEF} 
total (antmaze)              & 130.7         & 171.7   & 156.7          & 0.0    & 0.3   & 192.3          & \textbf{450.0} & 105.0          & 137.3   \\
\bottomrule
\end{tabular}
\end{table*}

\begin{table*}[ht]
\centering
\fontsize{8}{8}\selectfont
\caption{Best average returns on RL Unplugged Dataset.}
\label{tab:rlup_best}
\begin{tabular}{cccccccccc}
\toprule
          & BC     & 10\% BC & Onestep & MuZero          & SAC    & TD3+BC & IQL    & CRR$^+$   & CQL$^+$   \\
          \midrule
cartpole swingup & 477.7  & 811.8   & 456.8   & 649.4           & 879.6  & 700.9           & 879.2  & 834.6  & 739.8  \\
Walker Walk      & 418.0  & 525.1   & 581.0   & 961.4           & 954.5  & 921.6           & 940.4  & 951.9  & 747.9  \\
Finger Turn Hard & 195.5  & 329.4   & 219.2   & 618.1           & 629.4  & 316.3           & 441.1  & 239.7  & 210.2  \\
Cheetah Run      & 422.1  & 157.2   & 419.3   & 187.7           & 332.6  & 833.0           & 258.1  & 729.4  & 416.1  \\
Walker Stand     & 400.7  & 611.2   & 472.2   & 971.2           & 907.6  & 726.5           & 821.1  & 717.8  & 469.2  \\
Fish Swim        & 377.7  & 493.8   & 473.4   & 544.3           & 171.5  & 404.4           & 250.1  & 213.6  & 119.1  \\
\rowcolor[HTML]{EFEFEF} 
total            & 2309.3 & 2977.8  & 2860.0  & \textbf{3932.2} & 3877.1 & \textbf{3933.4} & 3630.9 & 3710.6 & 2715.4\\
\bottomrule
\end{tabular}
\end{table*}

\subsection{Training Curves}
In addition, we also visualize the training curves of all algorithms. We show that offline RL algorithms are fairly unstable and a point estimation, \textit{e.g.}, best return or last return, might not be a good indicator for their performances. Specifically, for OnestepRL, as its training is splitted into different stages, we only visualize the last stage.

\begin{figure*}[htb]
\centering
\begin{tabular}{cccccc}
\includegraphics[width=0.225\linewidth]{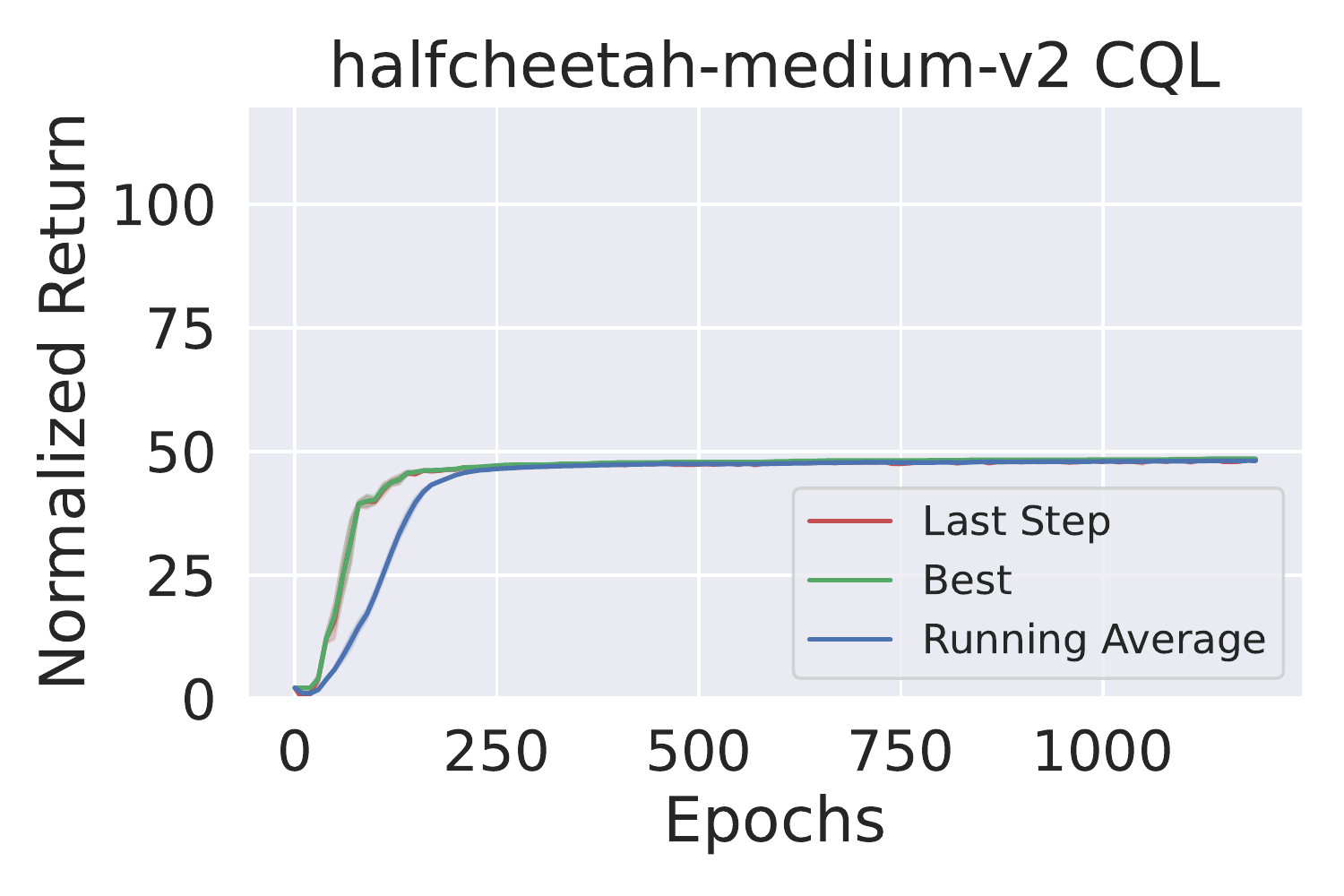}&\includegraphics[width=0.225\linewidth]{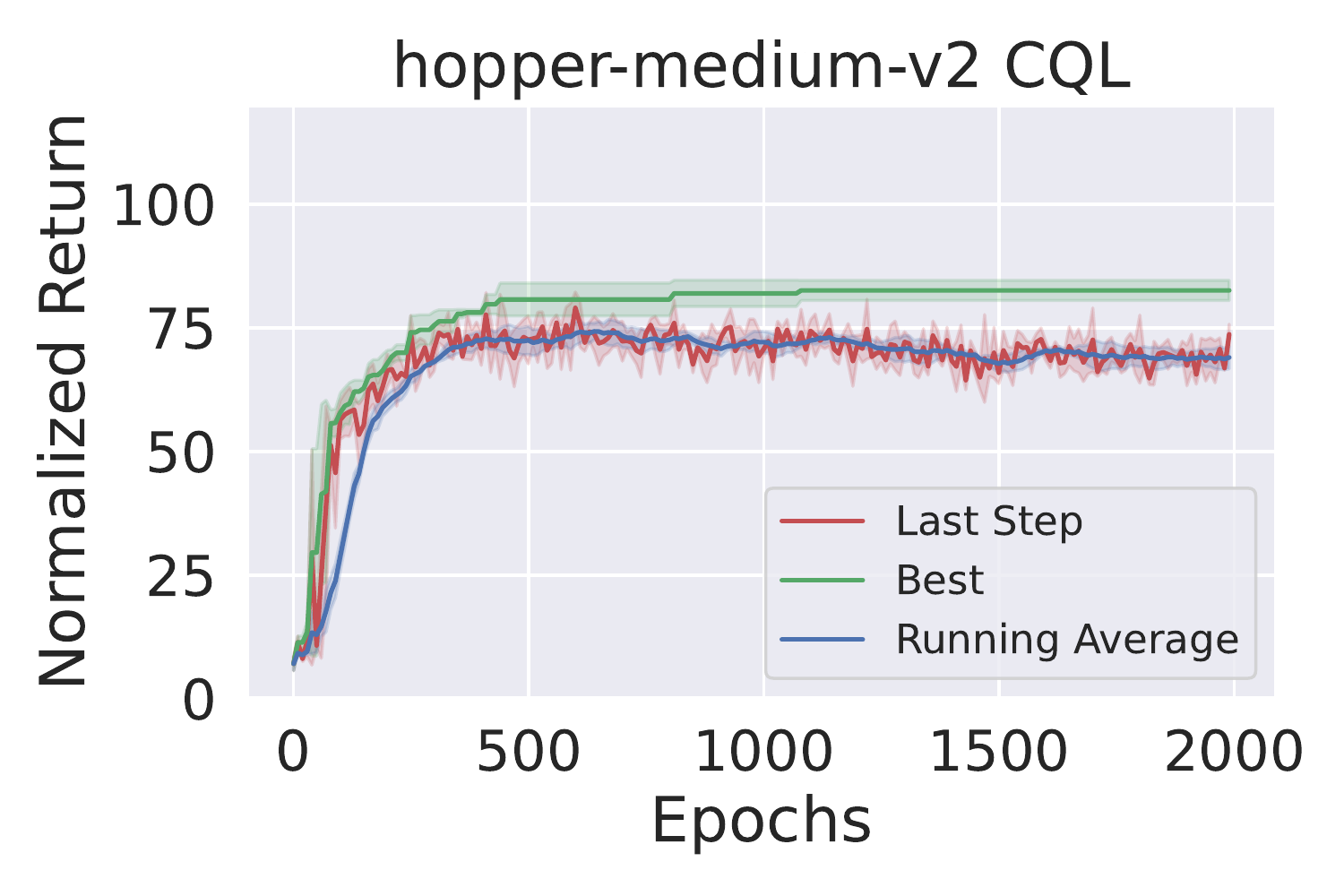}&\includegraphics[width=0.225\linewidth]{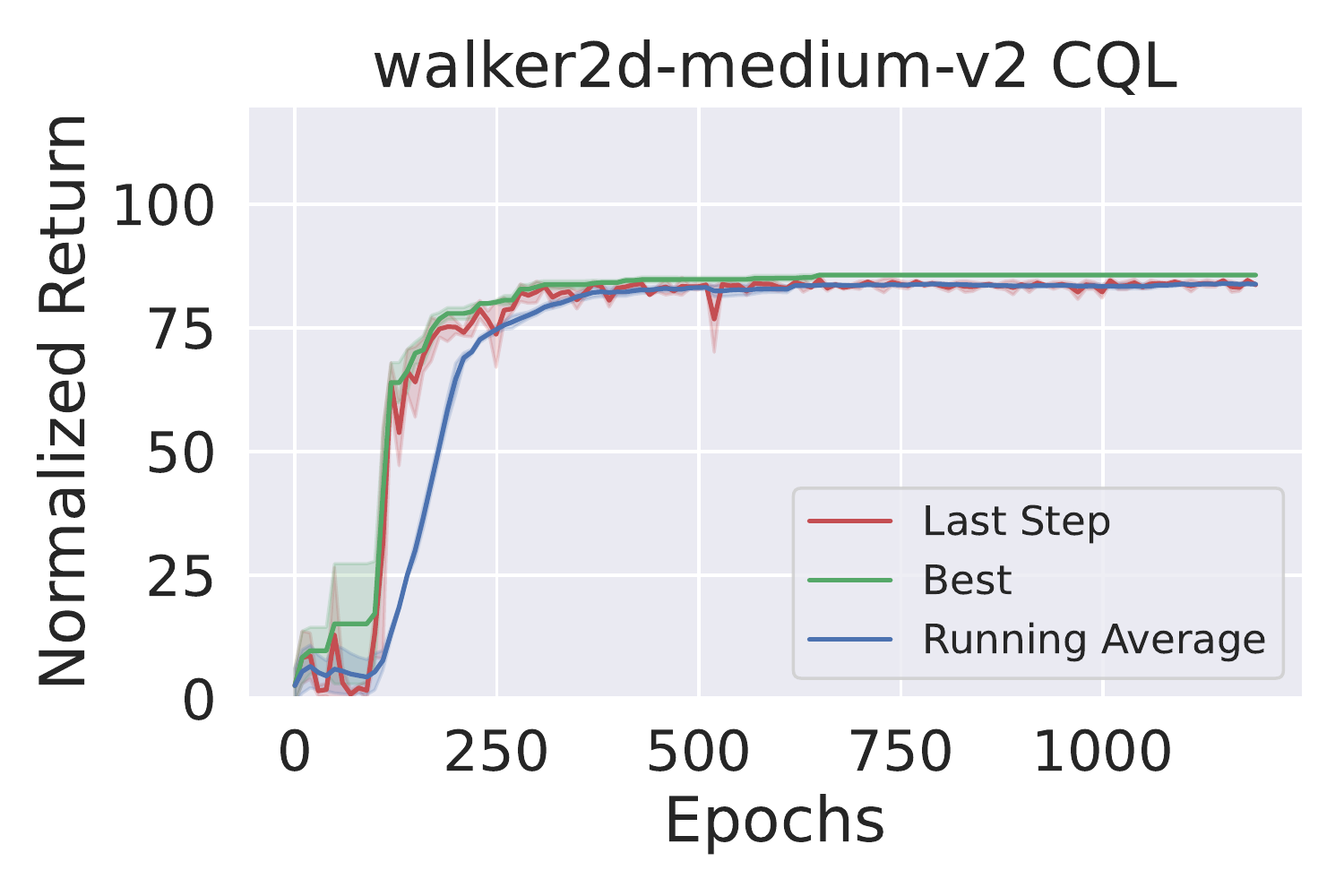}&\includegraphics[width=0.225\linewidth]{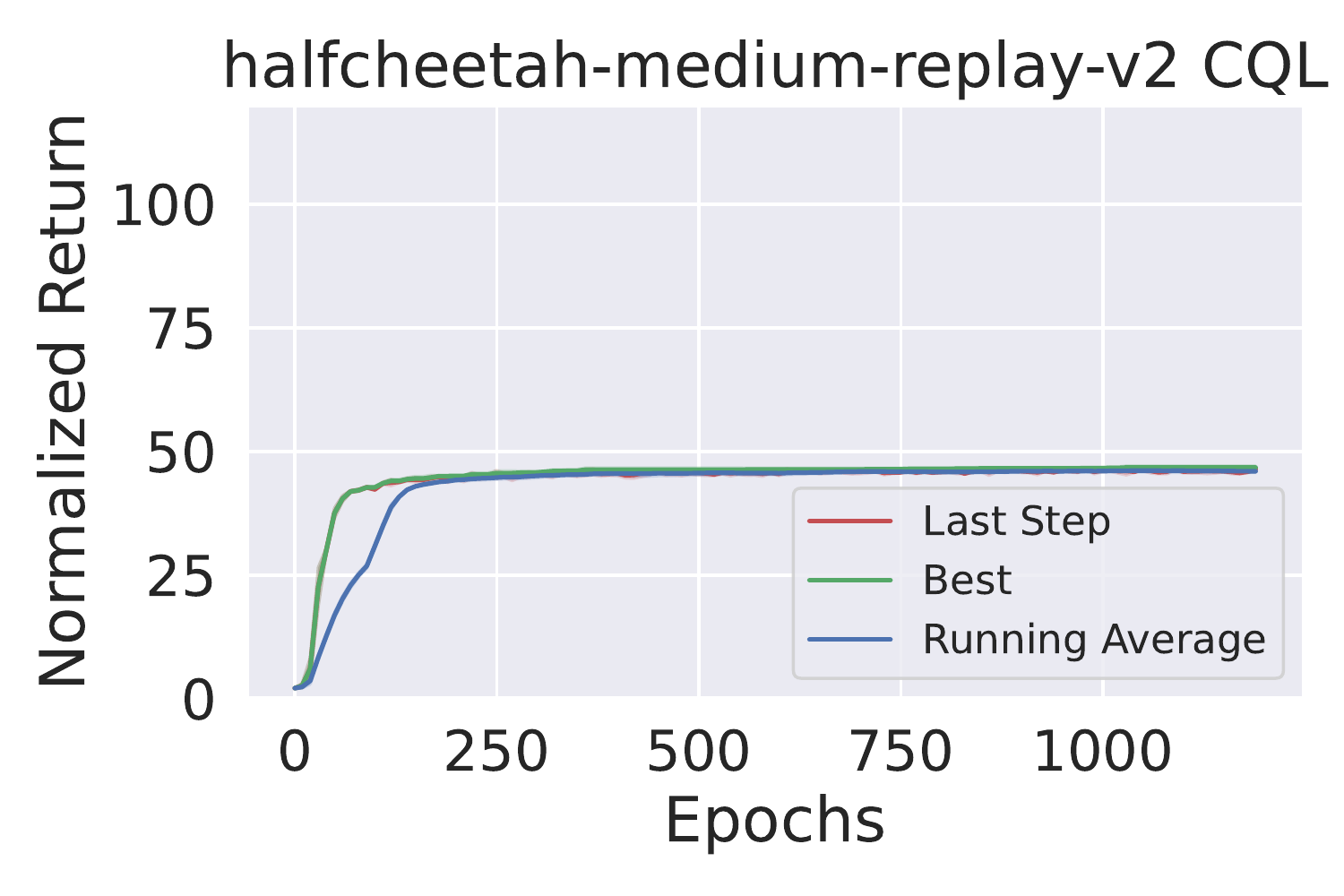}\\\includegraphics[width=0.225\linewidth]{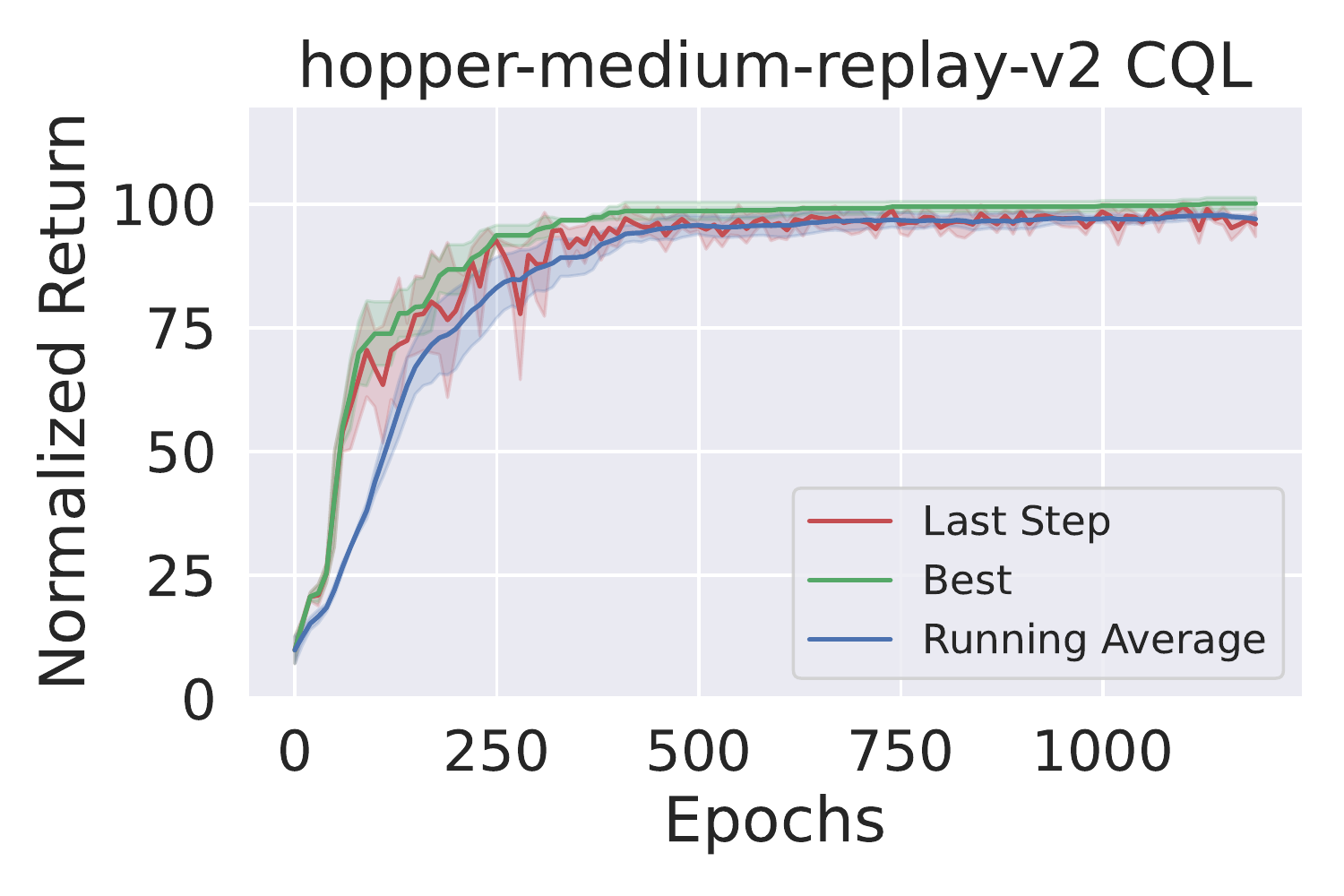}&\includegraphics[width=0.225\linewidth]{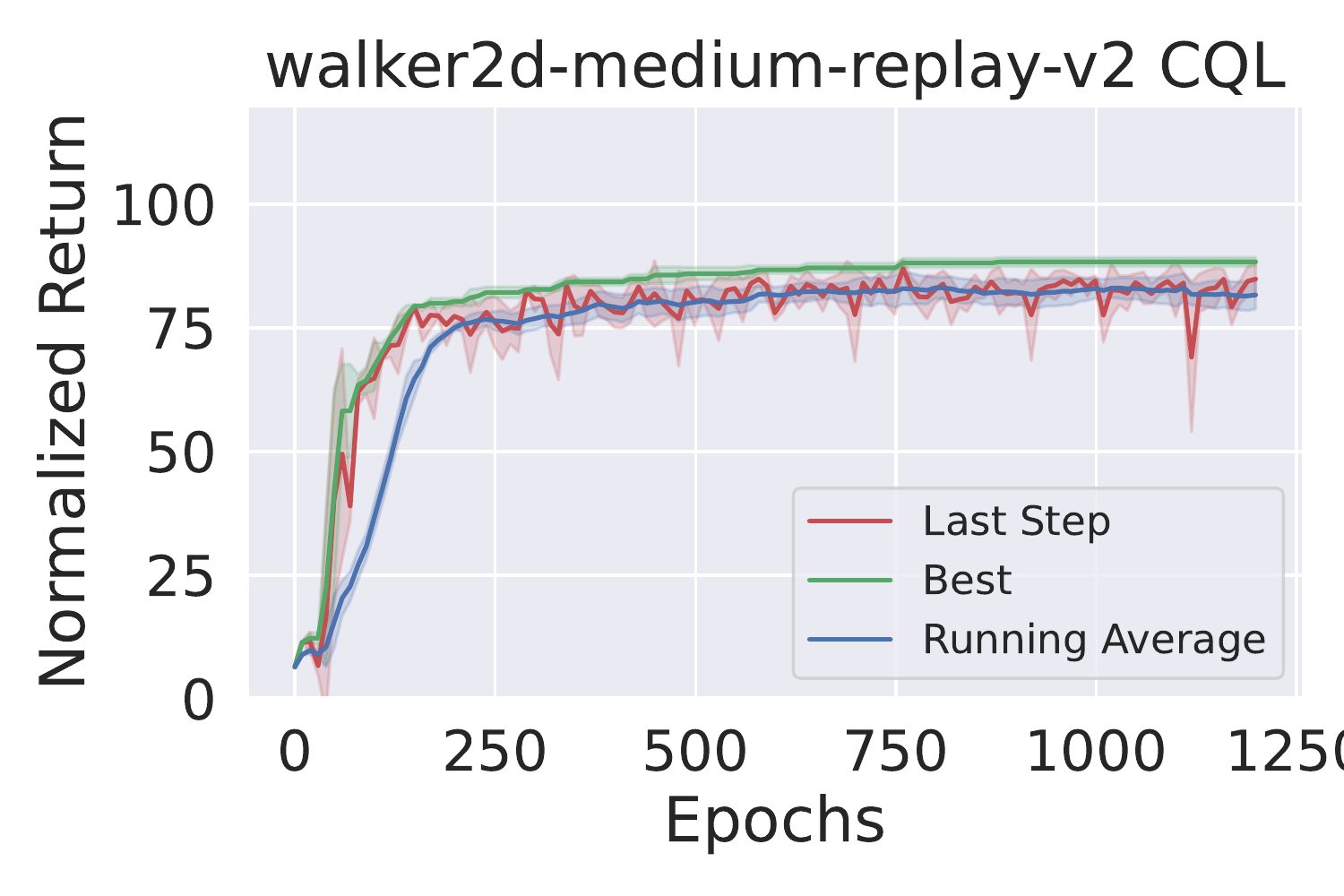}&\includegraphics[width=0.225\linewidth]{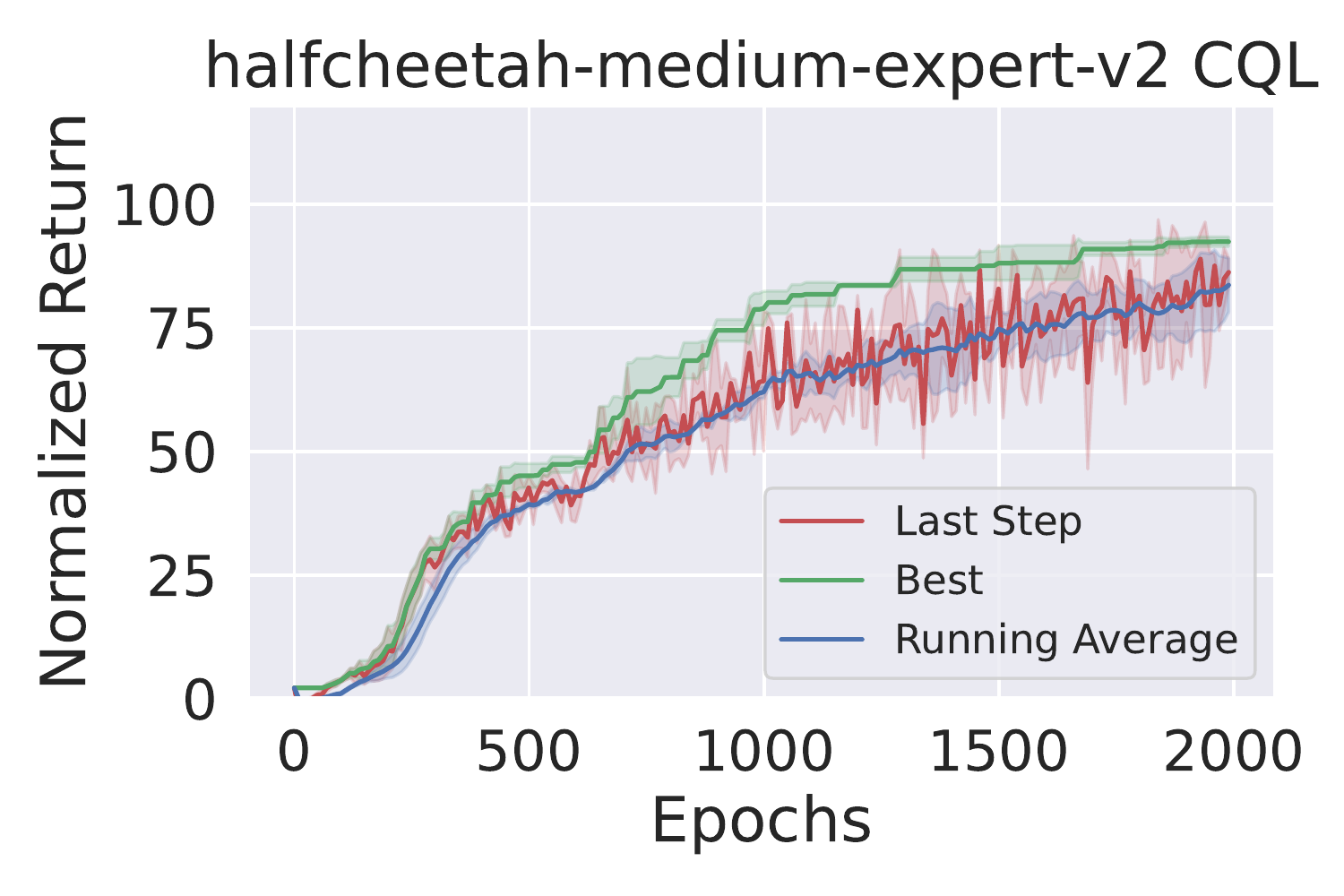}&\includegraphics[width=0.225\linewidth]{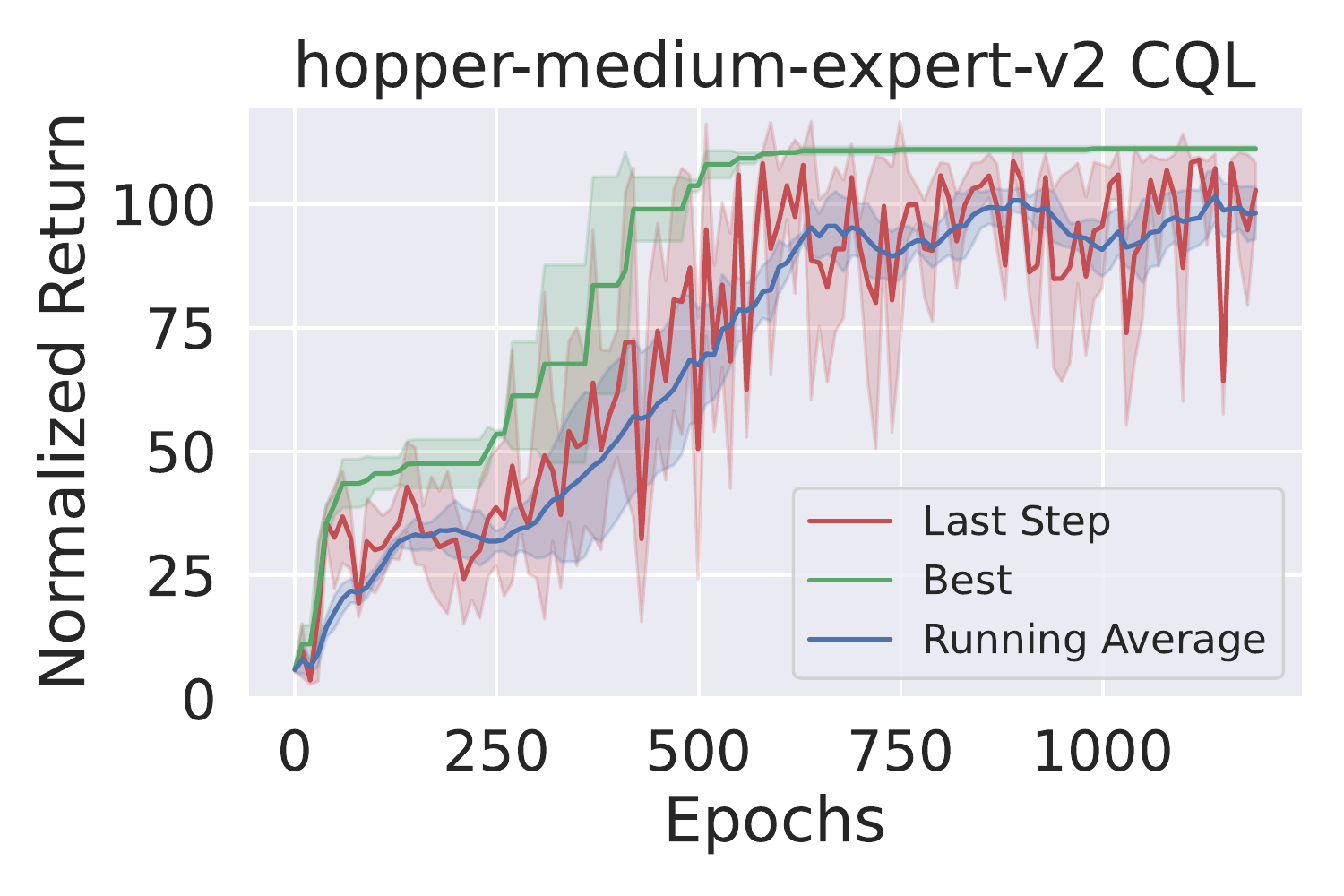}\\\includegraphics[width=0.225\linewidth]{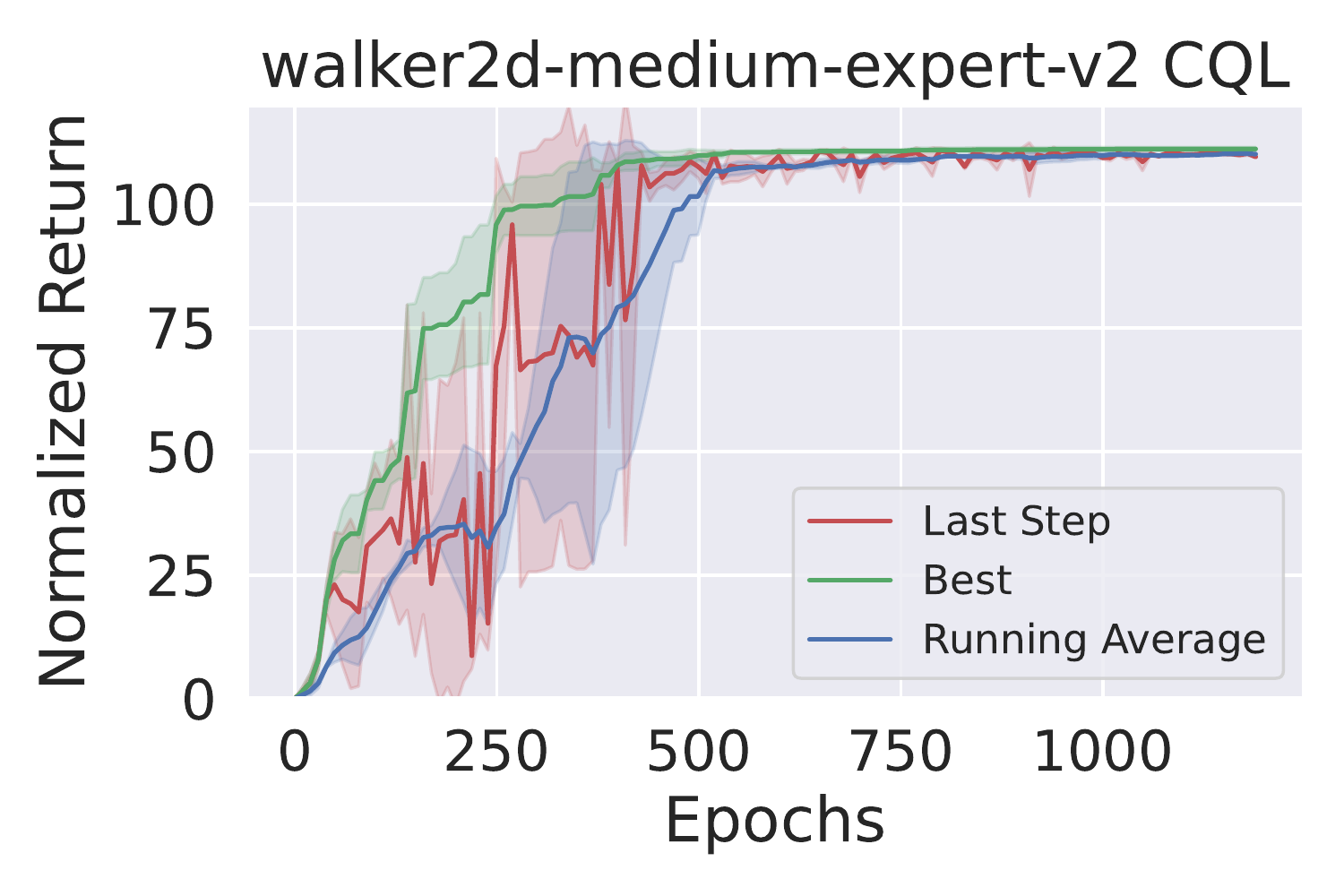}&\includegraphics[width=0.225\linewidth]{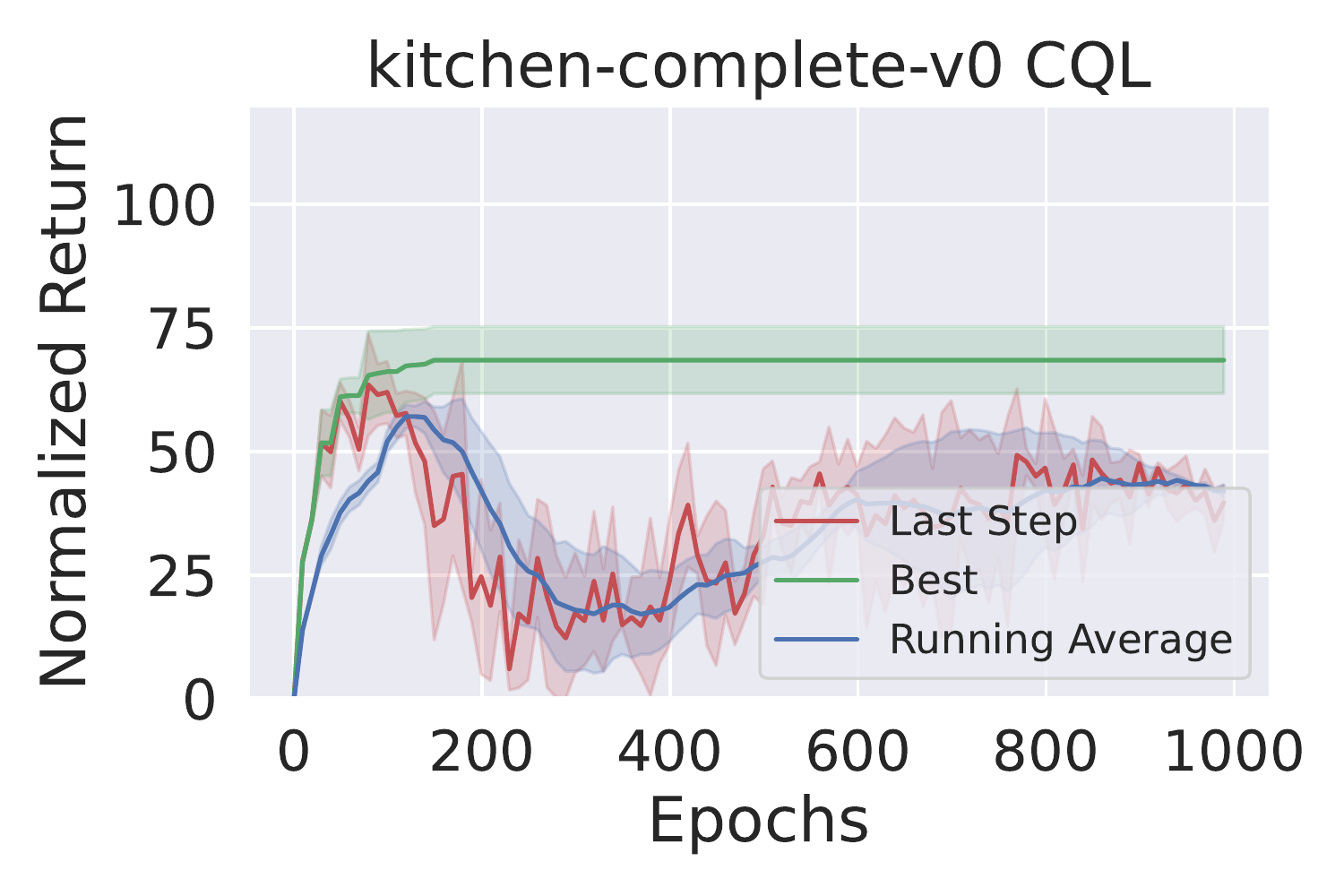}&\includegraphics[width=0.225\linewidth]{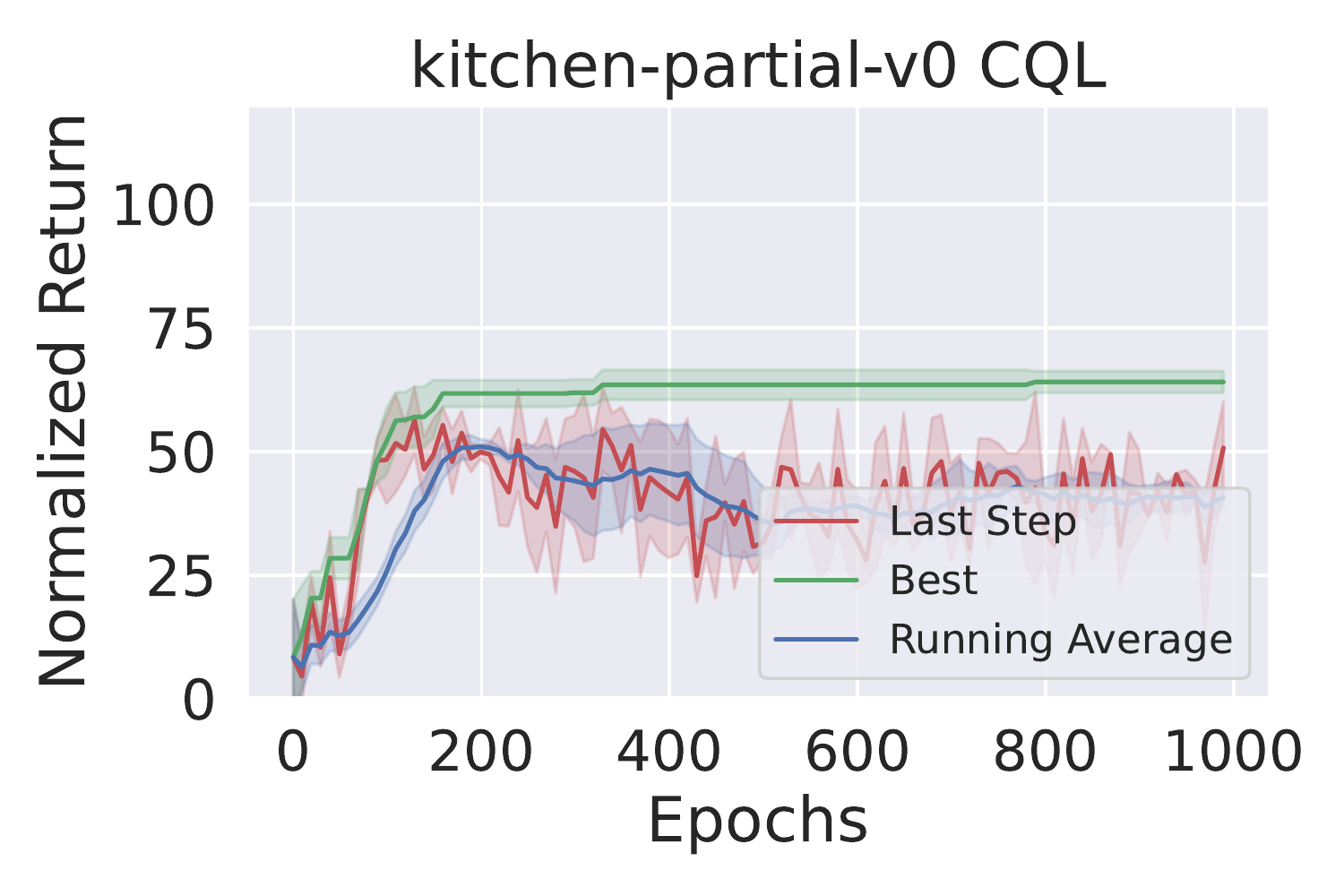}&\includegraphics[width=0.225\linewidth]{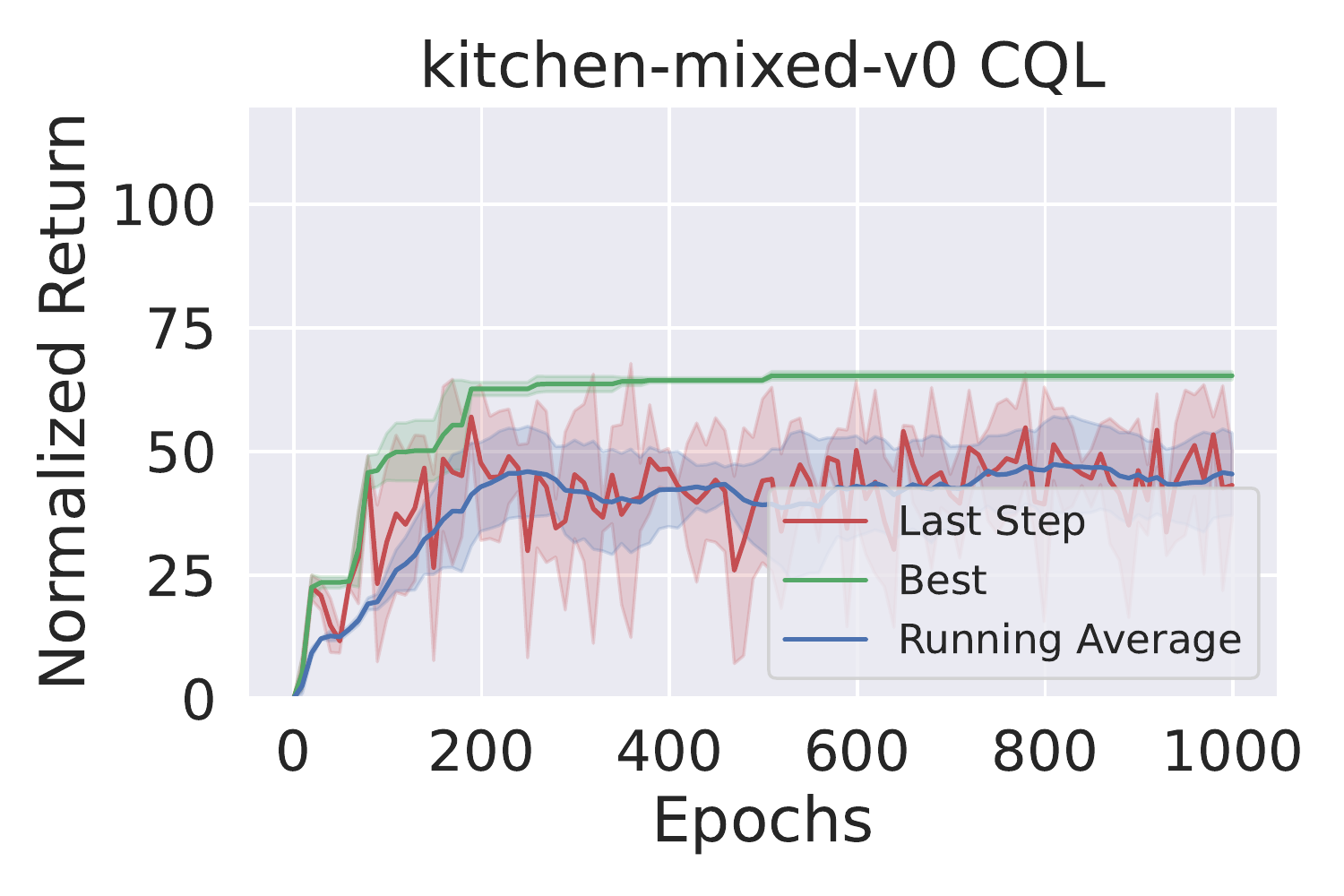}\\\includegraphics[width=0.225\linewidth]{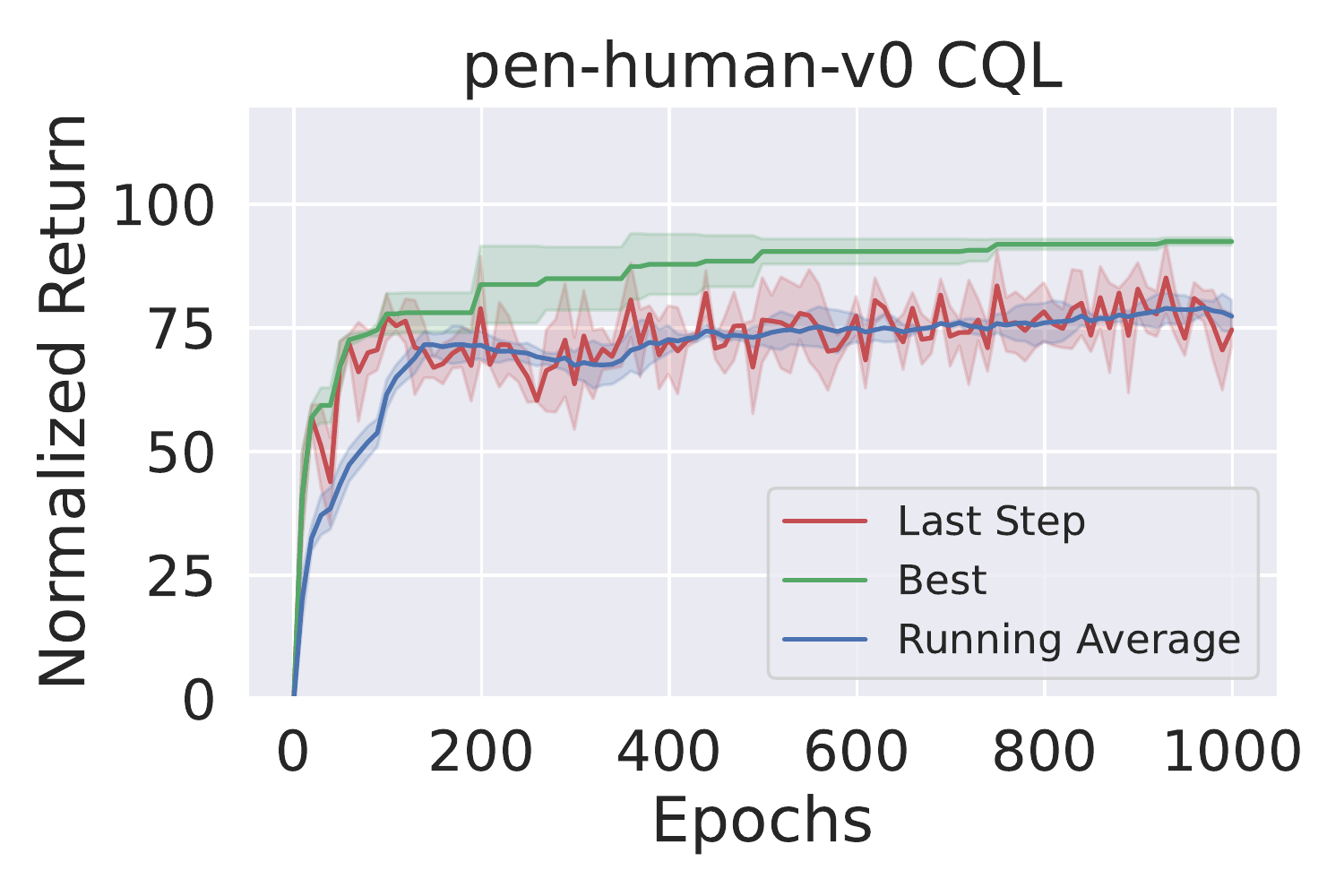}&\includegraphics[width=0.225\linewidth]{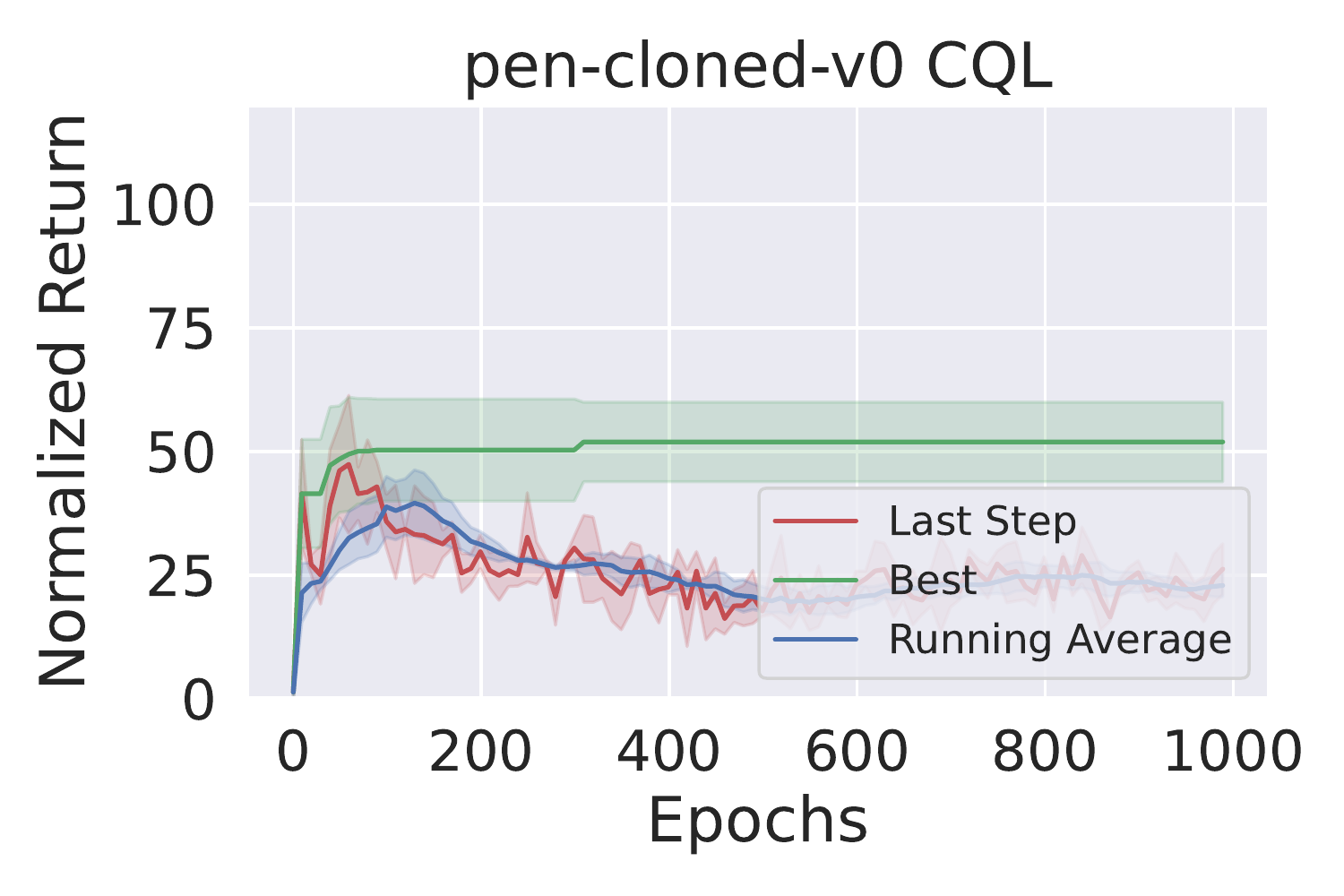}&\includegraphics[width=0.225\linewidth]{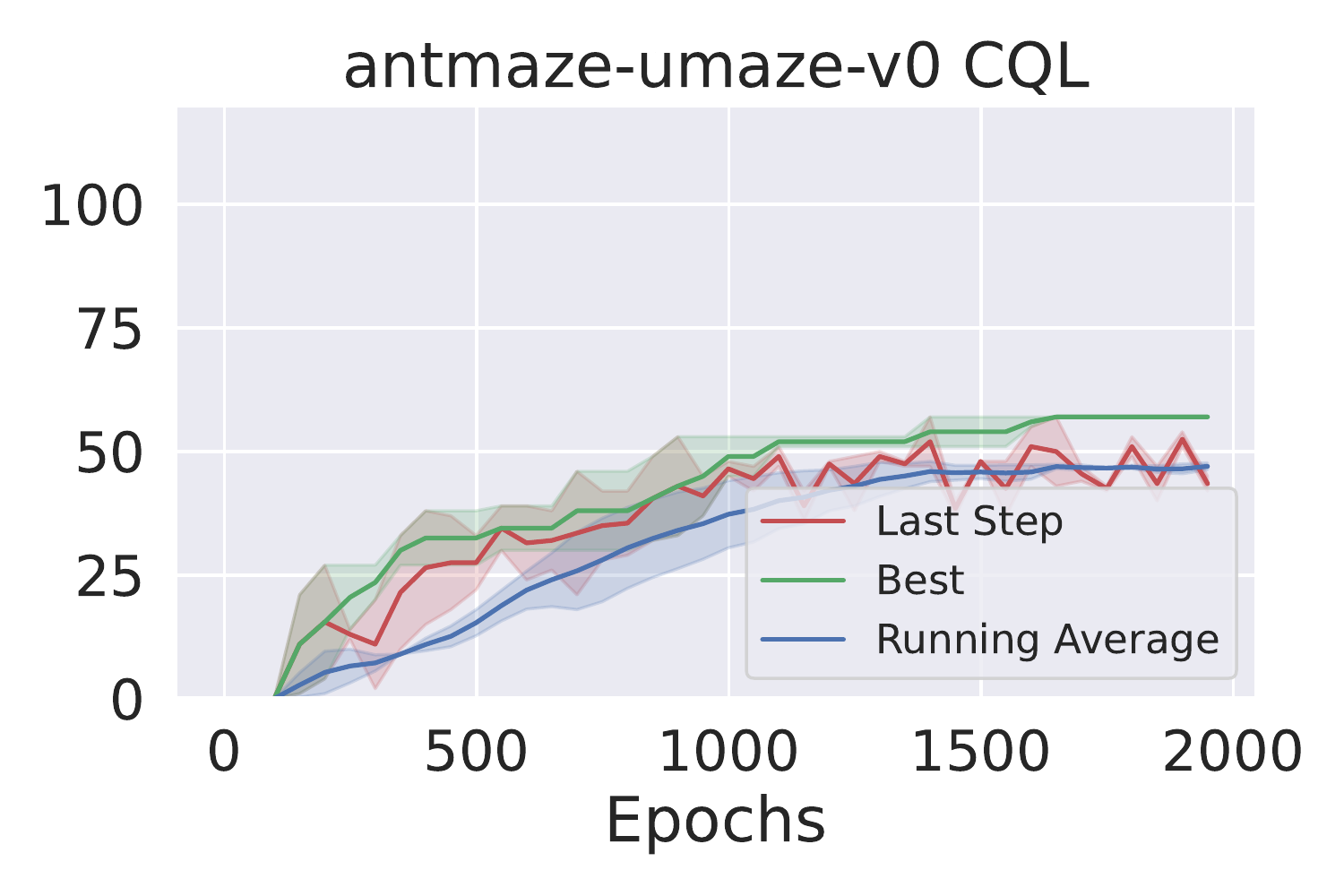}&\includegraphics[width=0.225\linewidth]{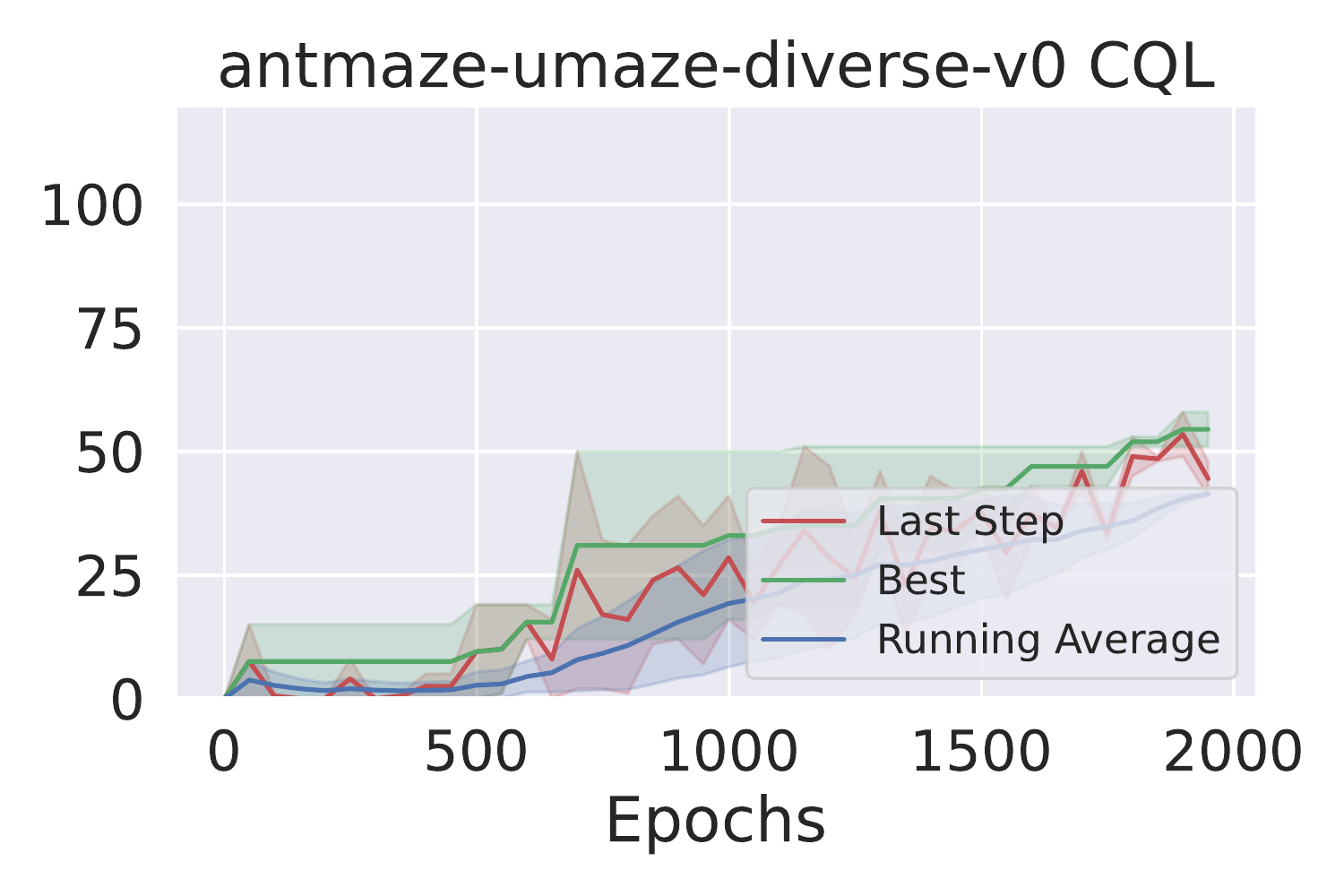}\\\includegraphics[width=0.225\linewidth]{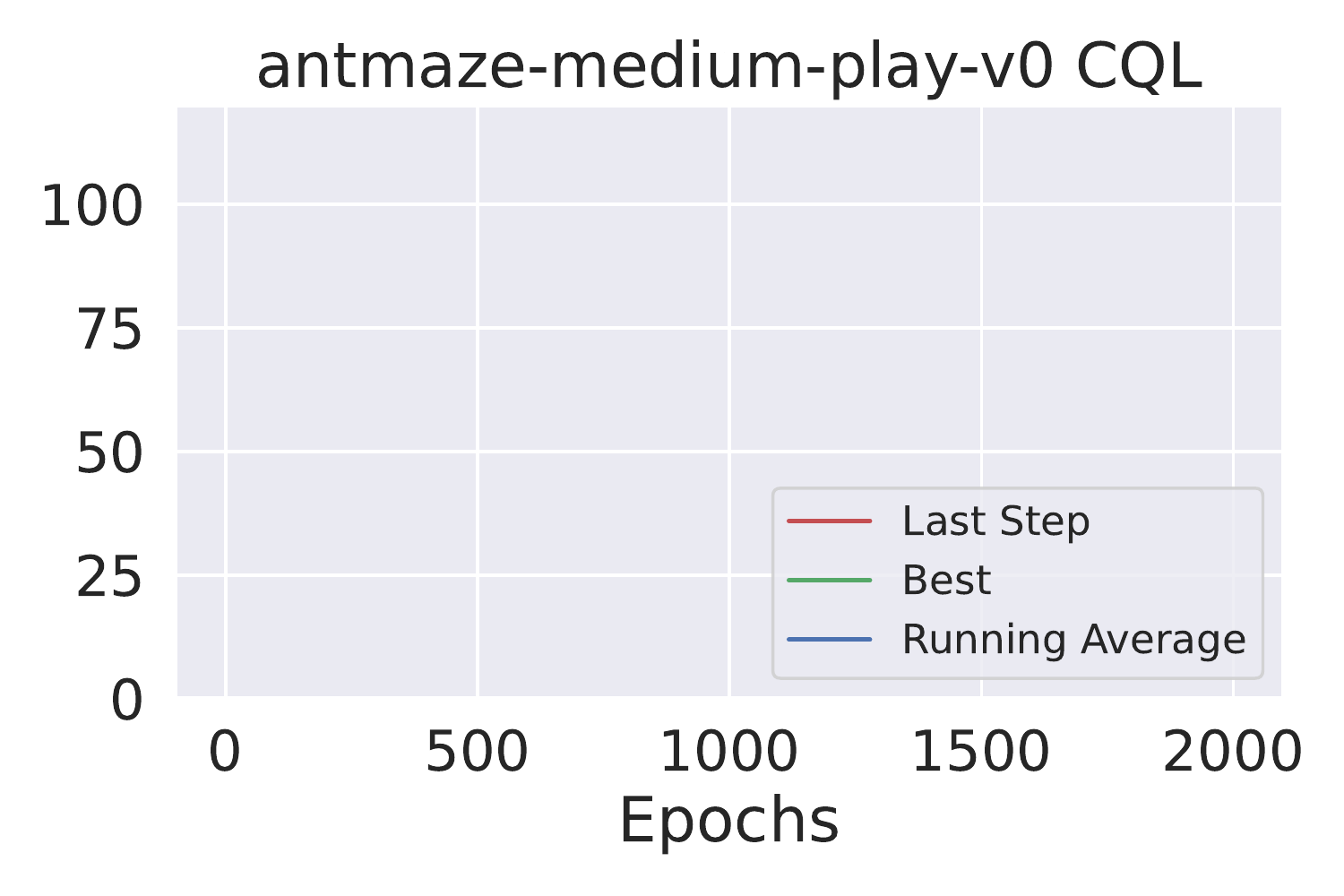}&\includegraphics[width=0.225\linewidth]{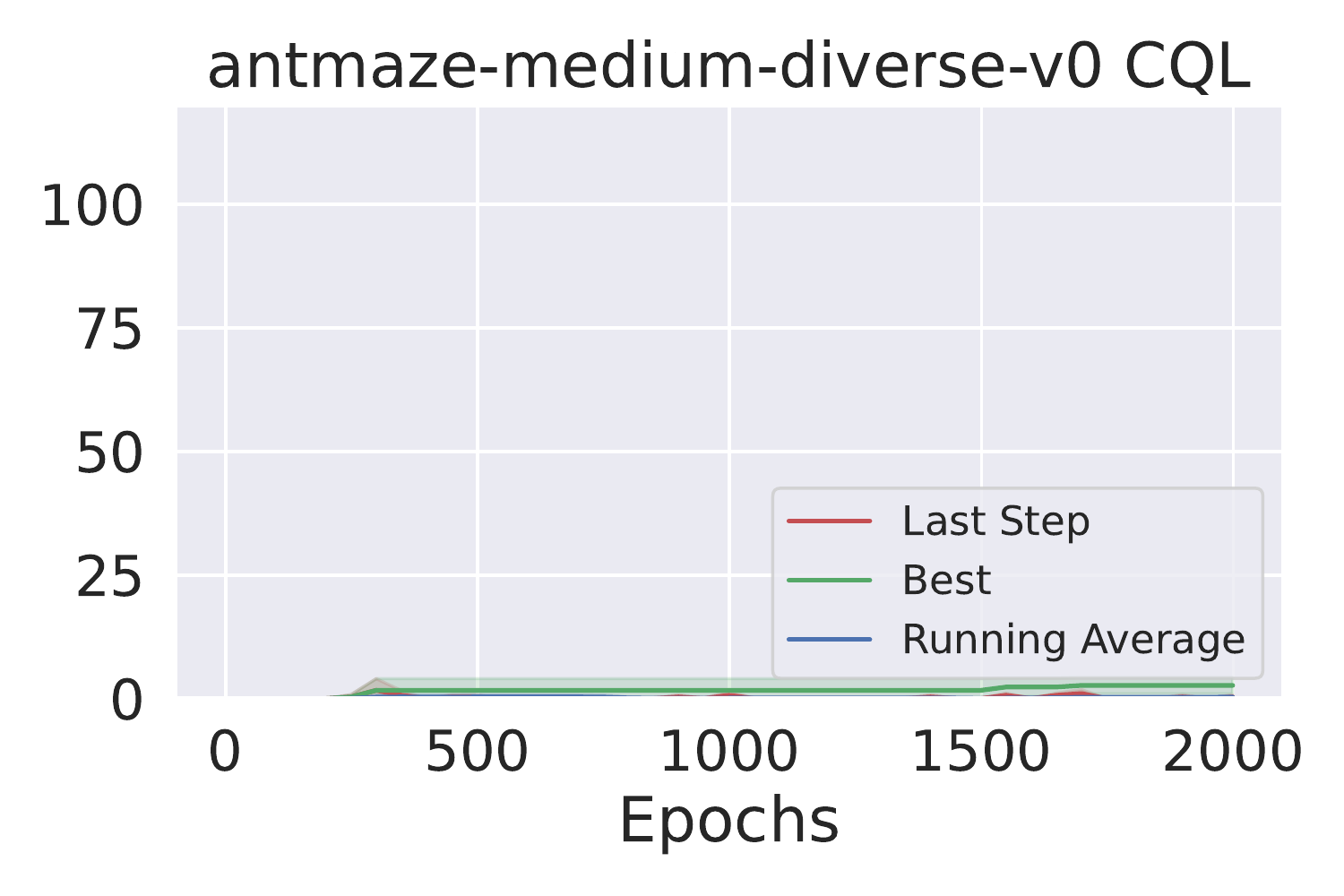}&\includegraphics[width=0.225\linewidth]{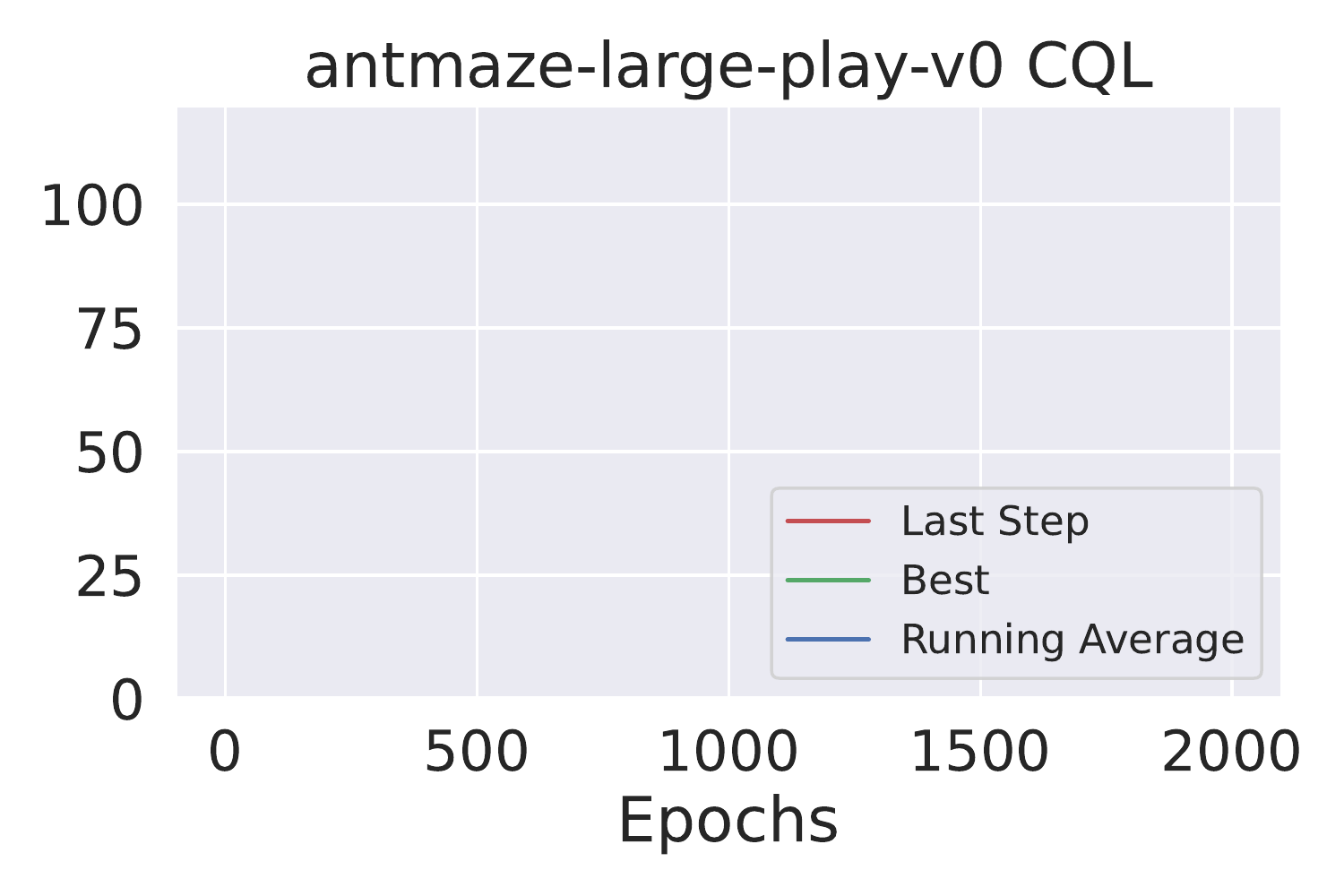}&\includegraphics[width=0.225\linewidth]{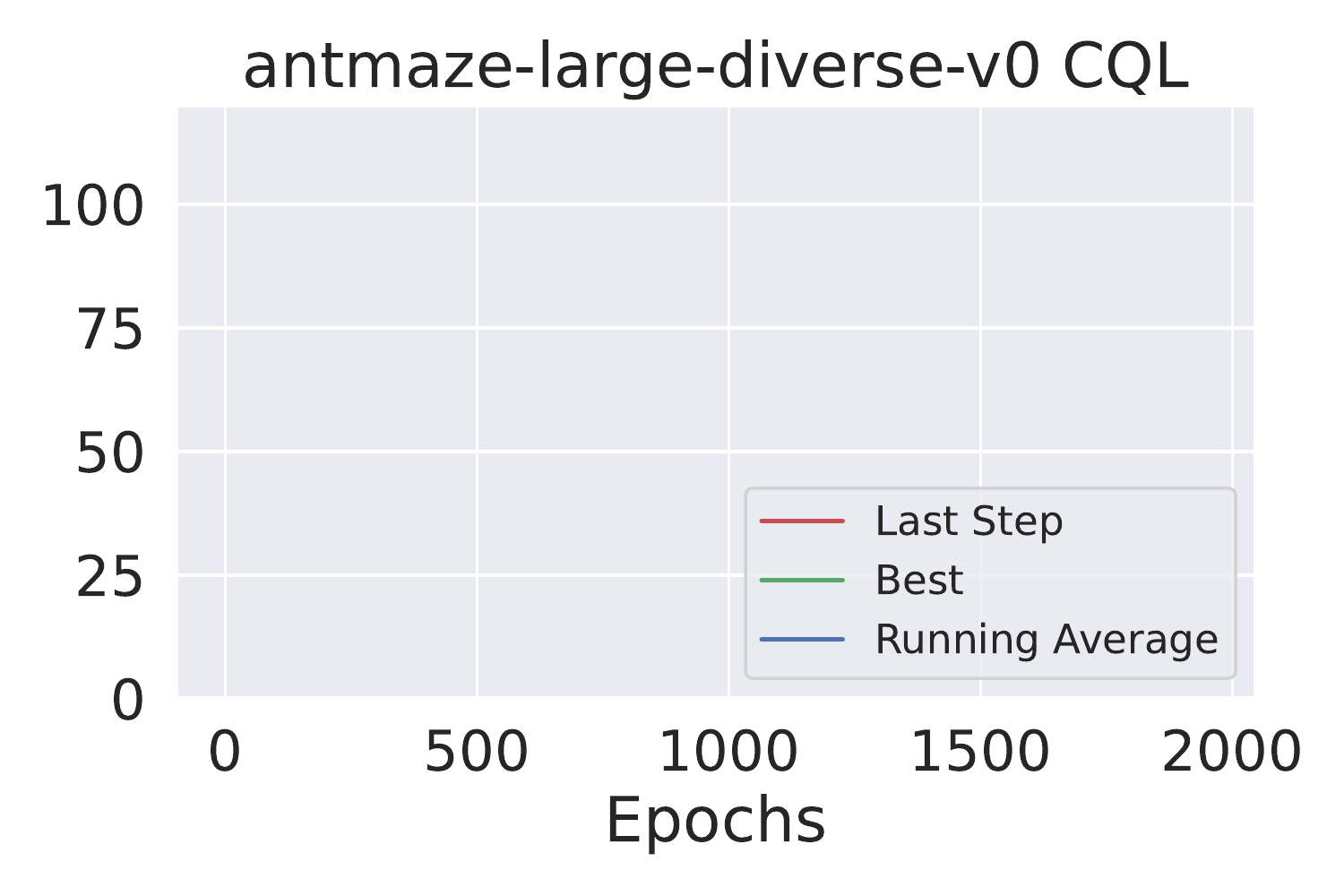}\\\end{tabular}
\centering
\caption{Training Curves of CQL on D4RL}
\end{figure*}
\begin{figure*}[htb]
\centering
\begin{tabular}{cccccc}
\includegraphics[width=0.225\linewidth]{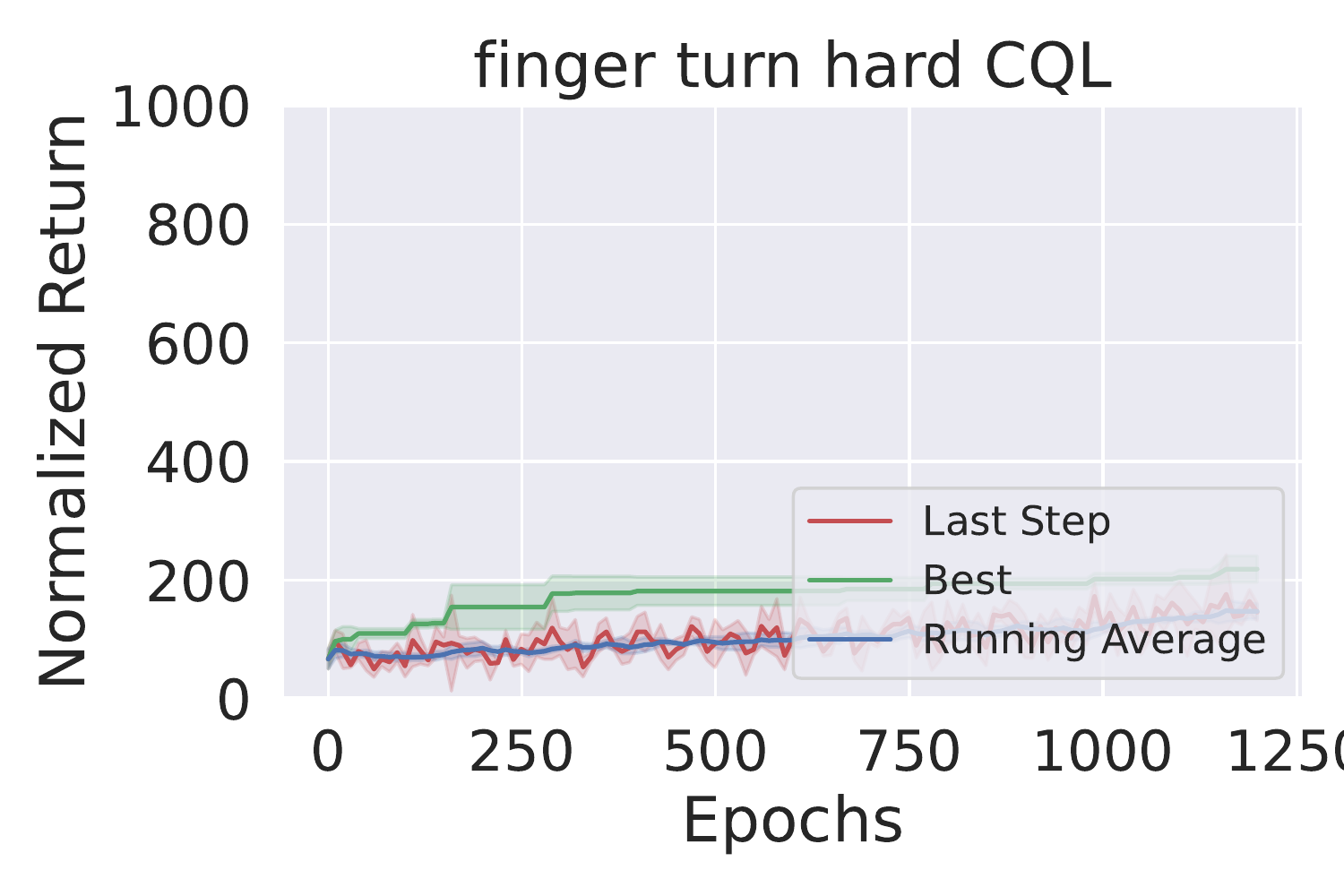}&\includegraphics[width=0.225\linewidth]{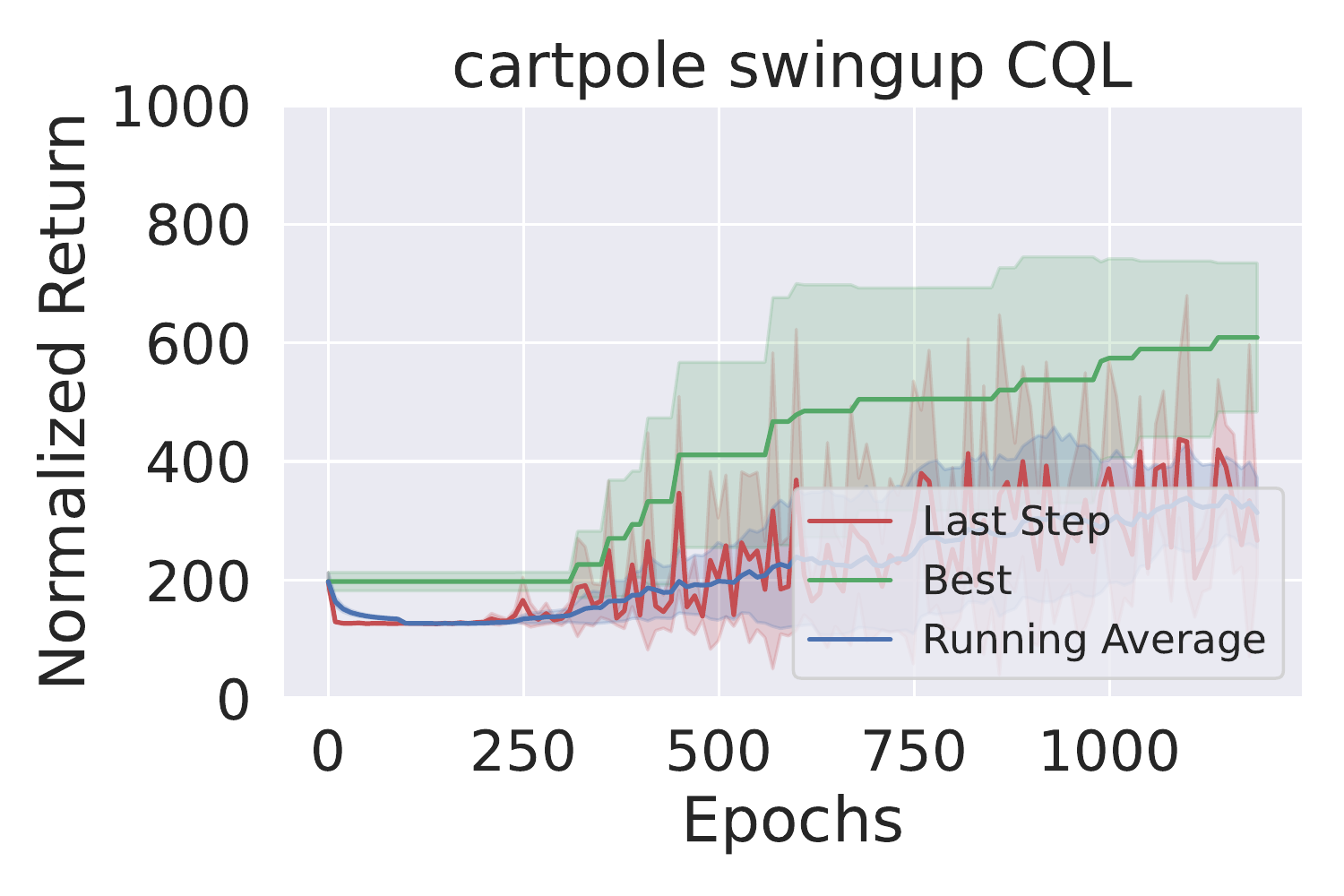}&\includegraphics[width=0.225\linewidth]{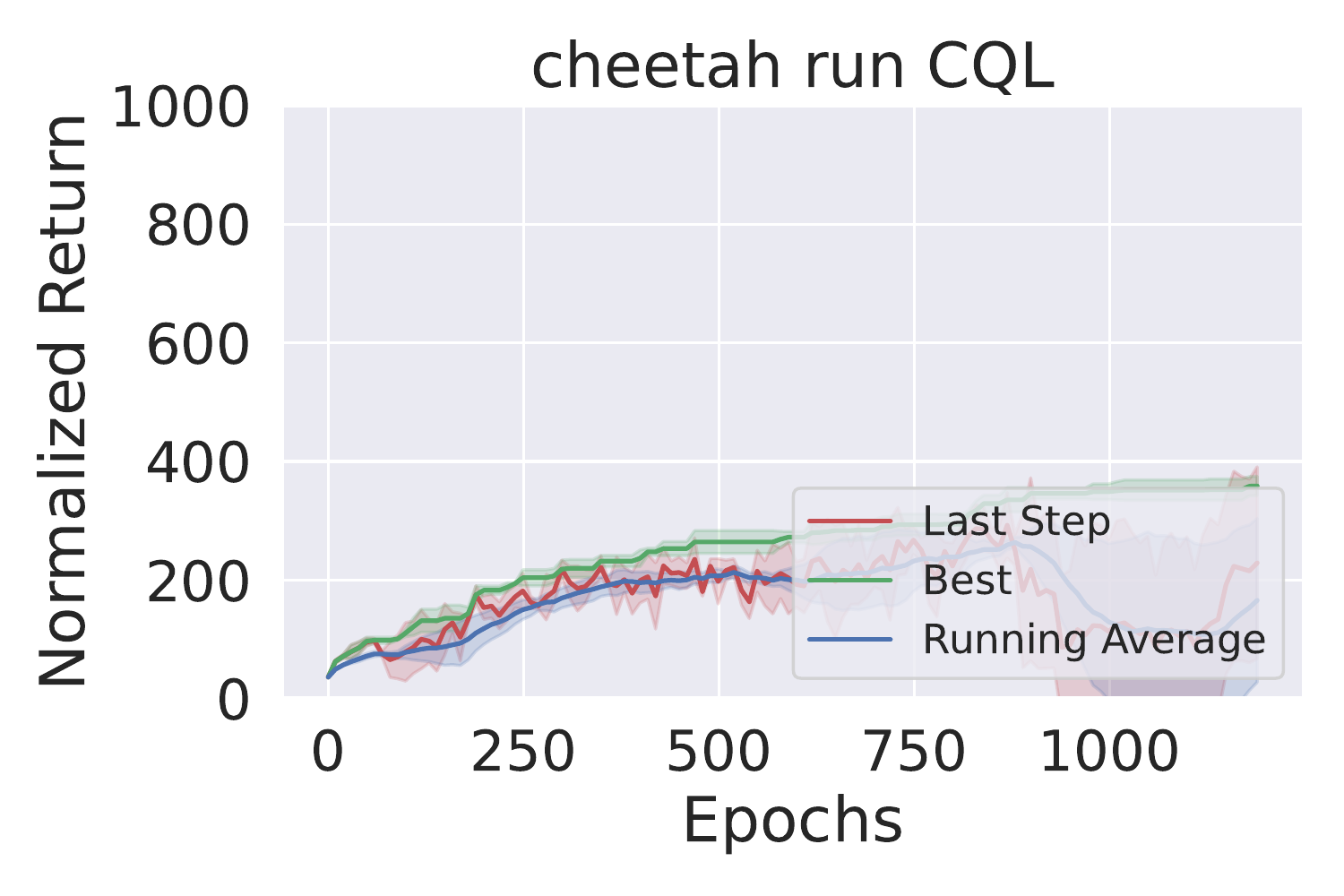}&\includegraphics[width=0.225\linewidth]{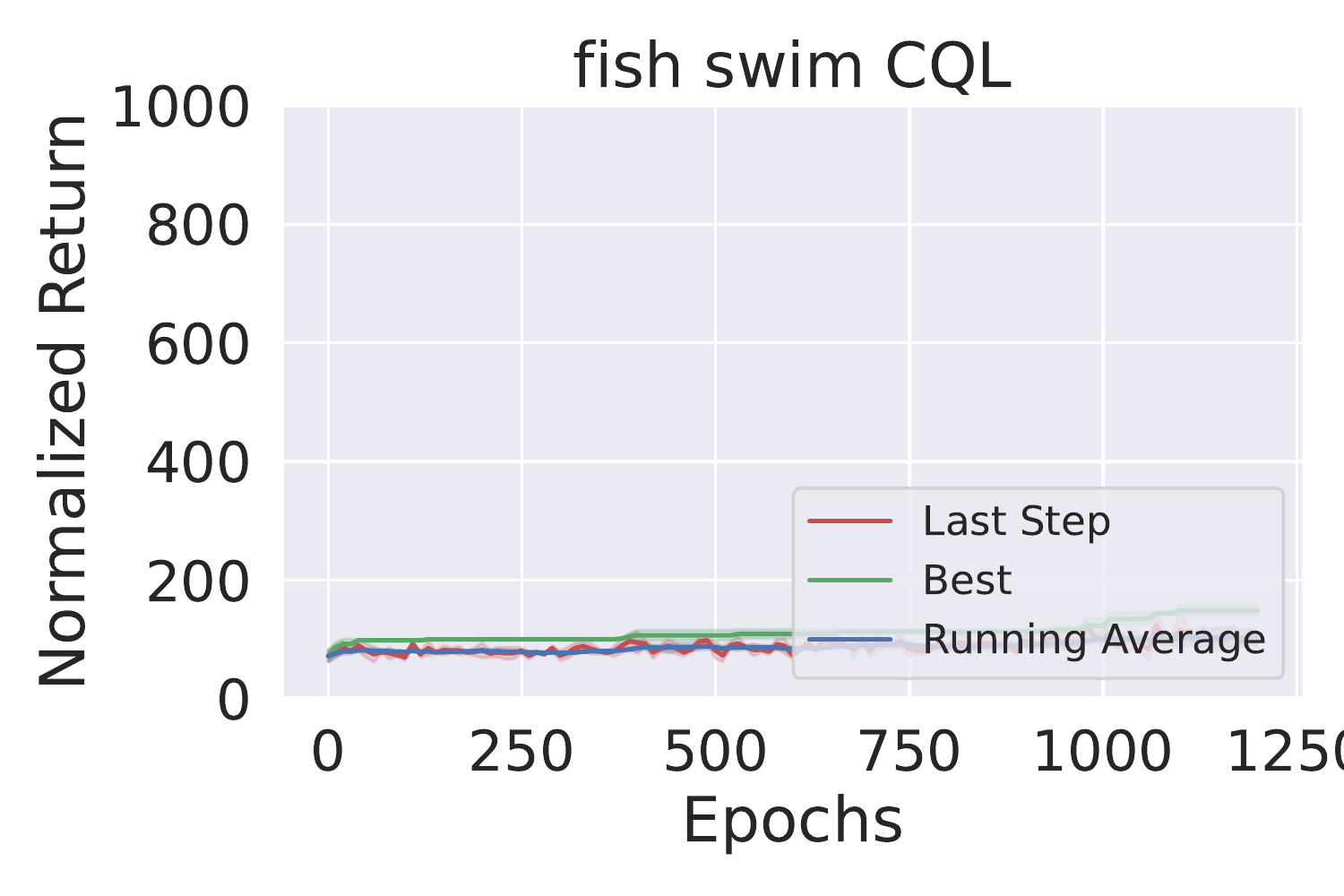}\\\includegraphics[width=0.225\linewidth]{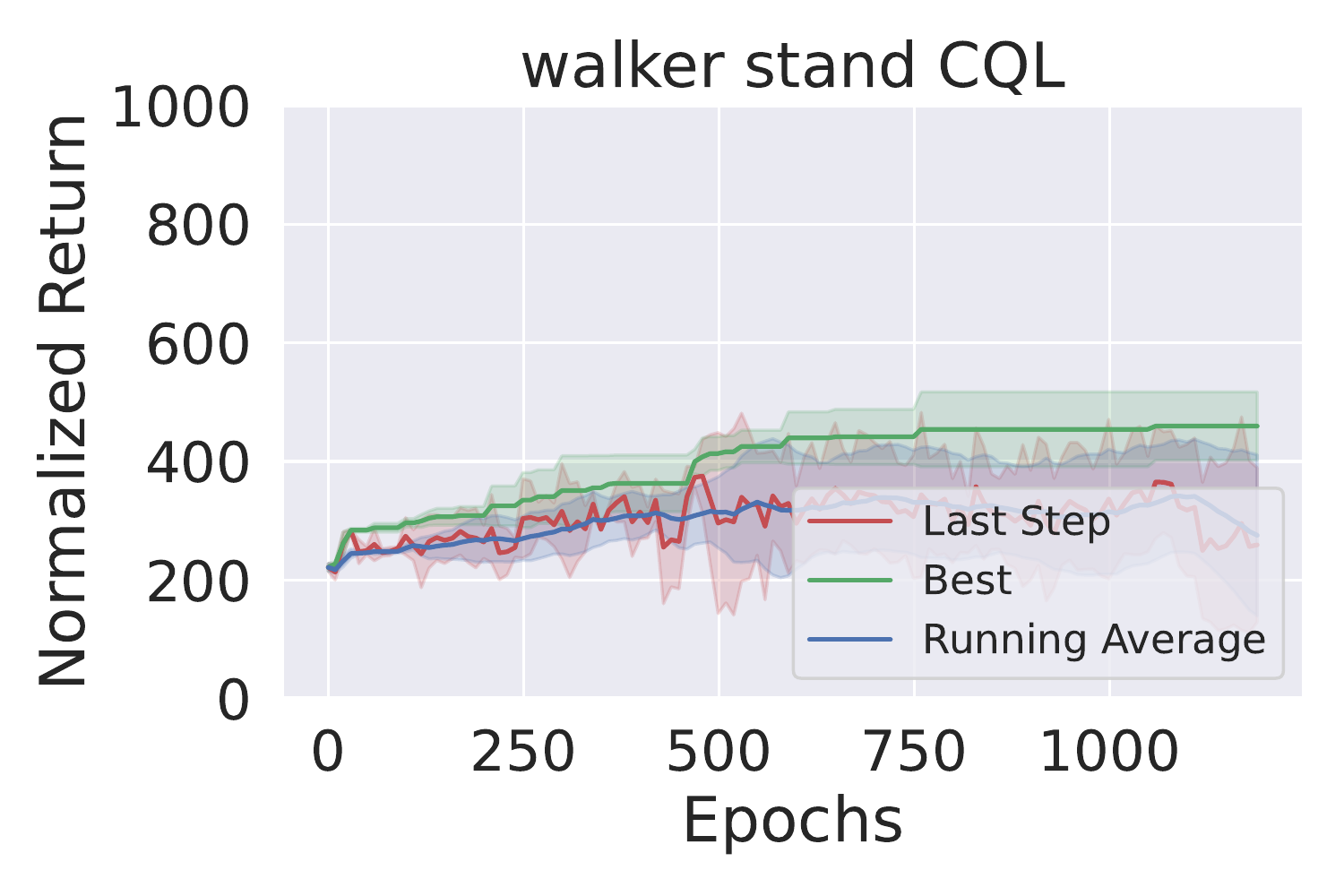}&\includegraphics[width=0.225\linewidth]{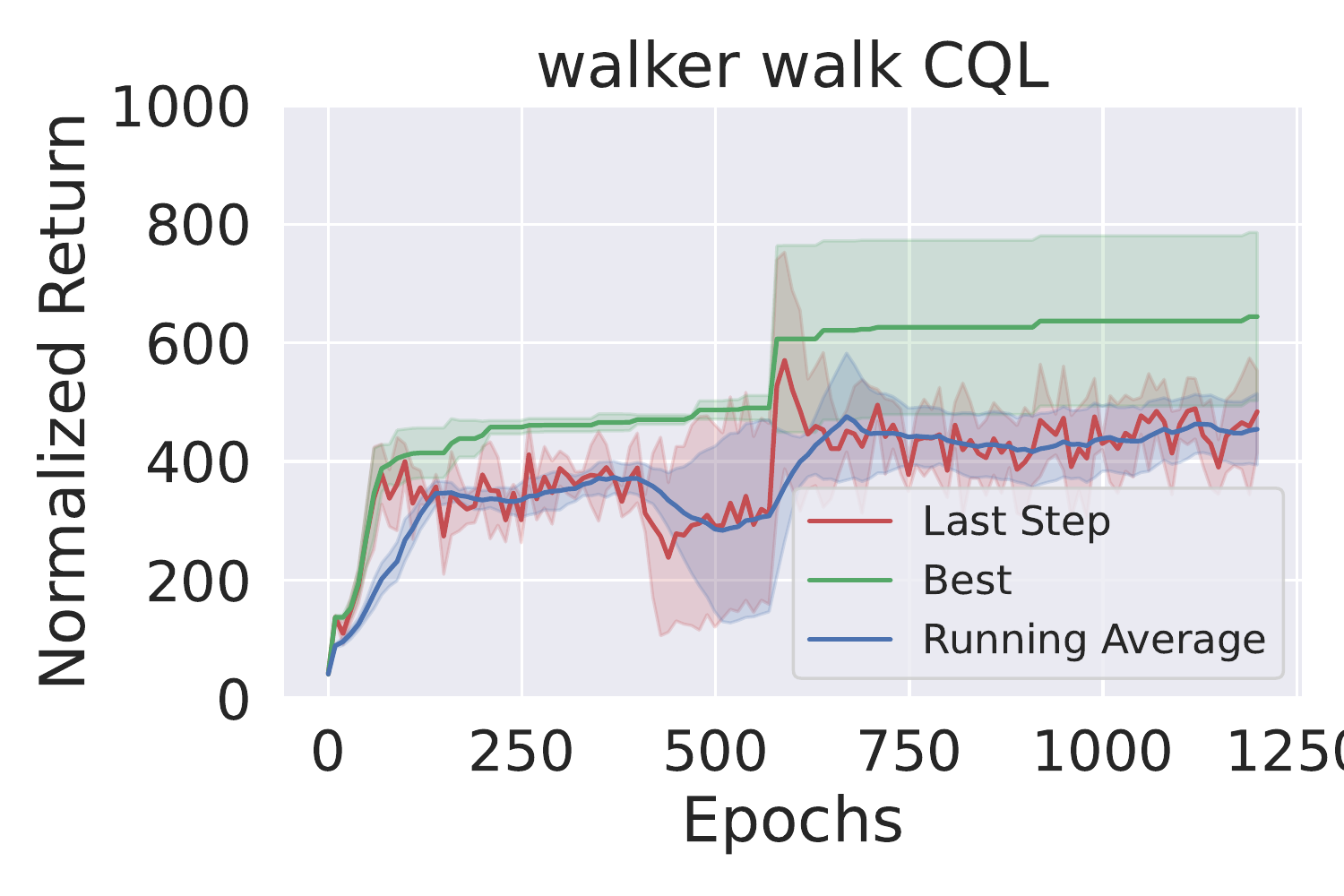}&&\\\end{tabular}
\centering
\caption{Training Curves of CQL on RLUP}
\end{figure*}
\begin{figure*}[htb]
\centering
\begin{tabular}{cccccc}
\includegraphics[width=0.225\linewidth]{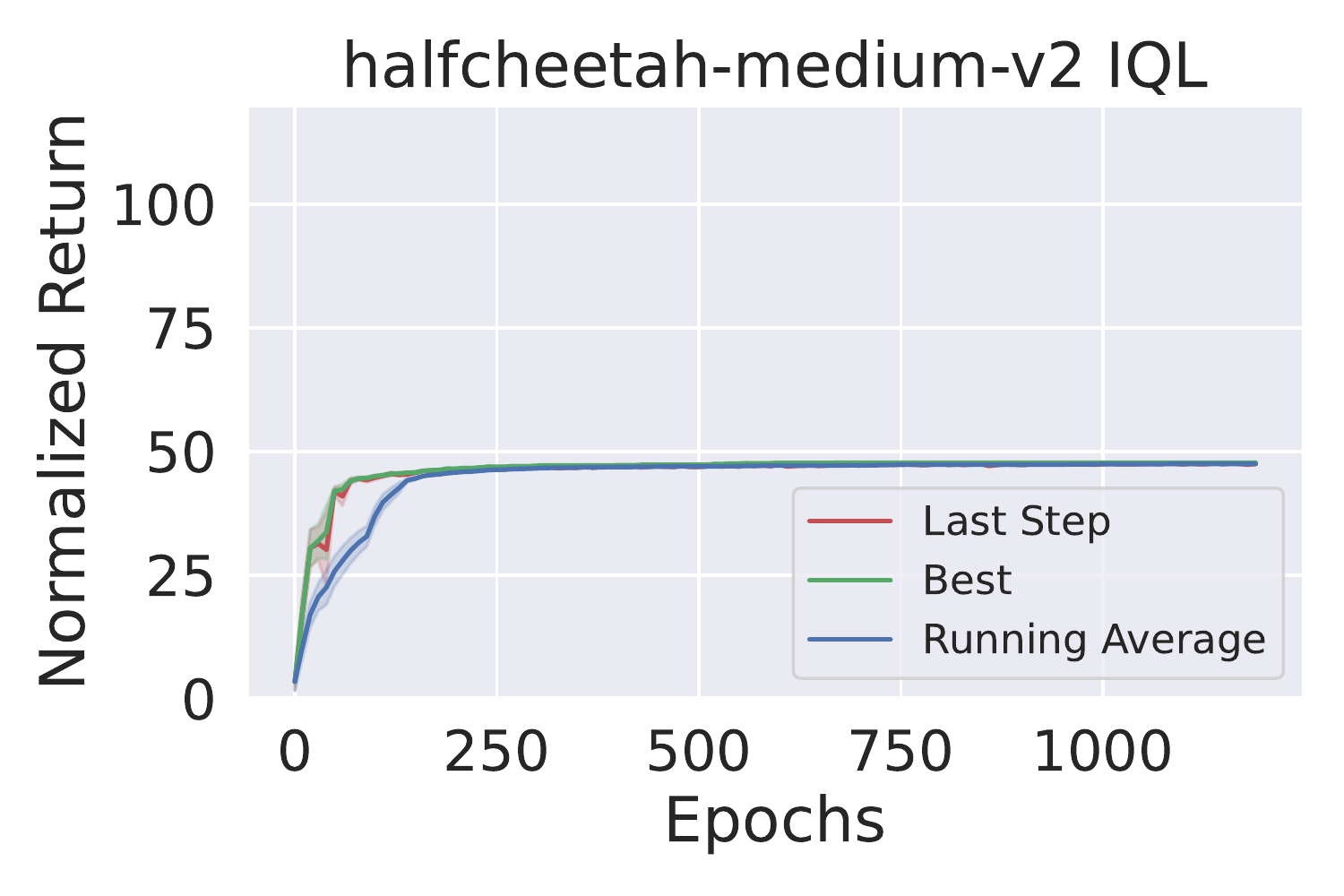}&\includegraphics[width=0.225\linewidth]{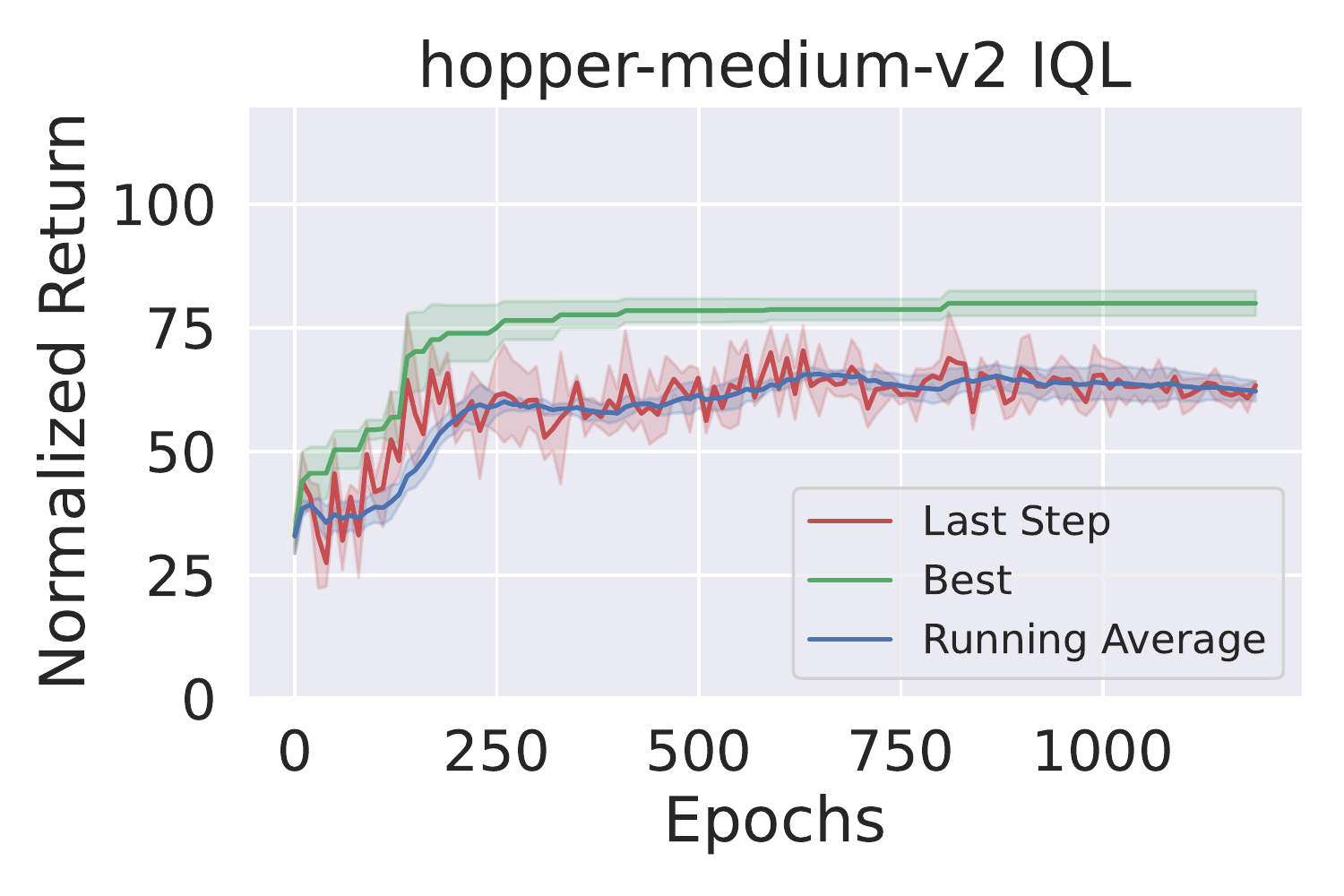}&\includegraphics[width=0.225\linewidth]{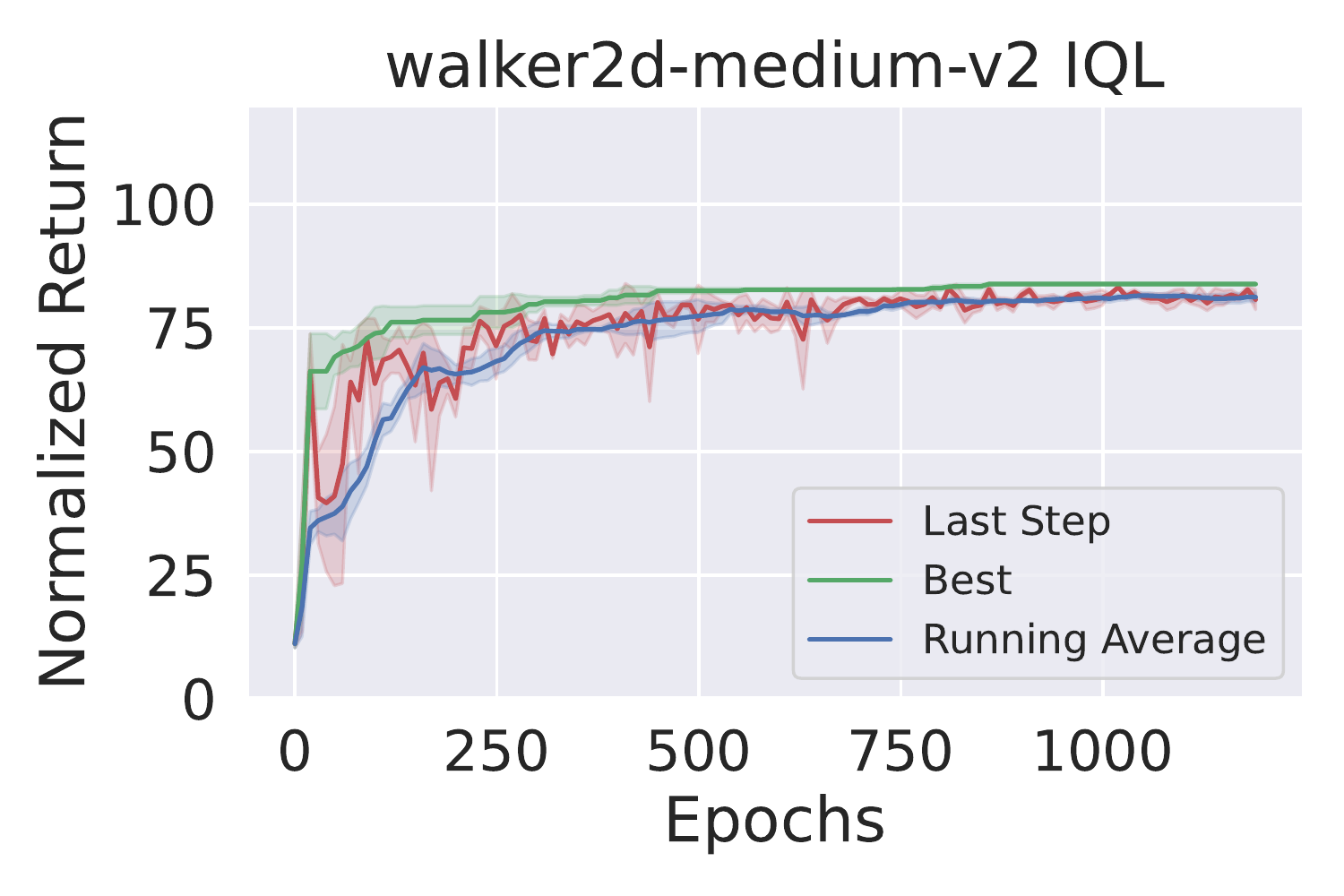}&\includegraphics[width=0.225\linewidth]{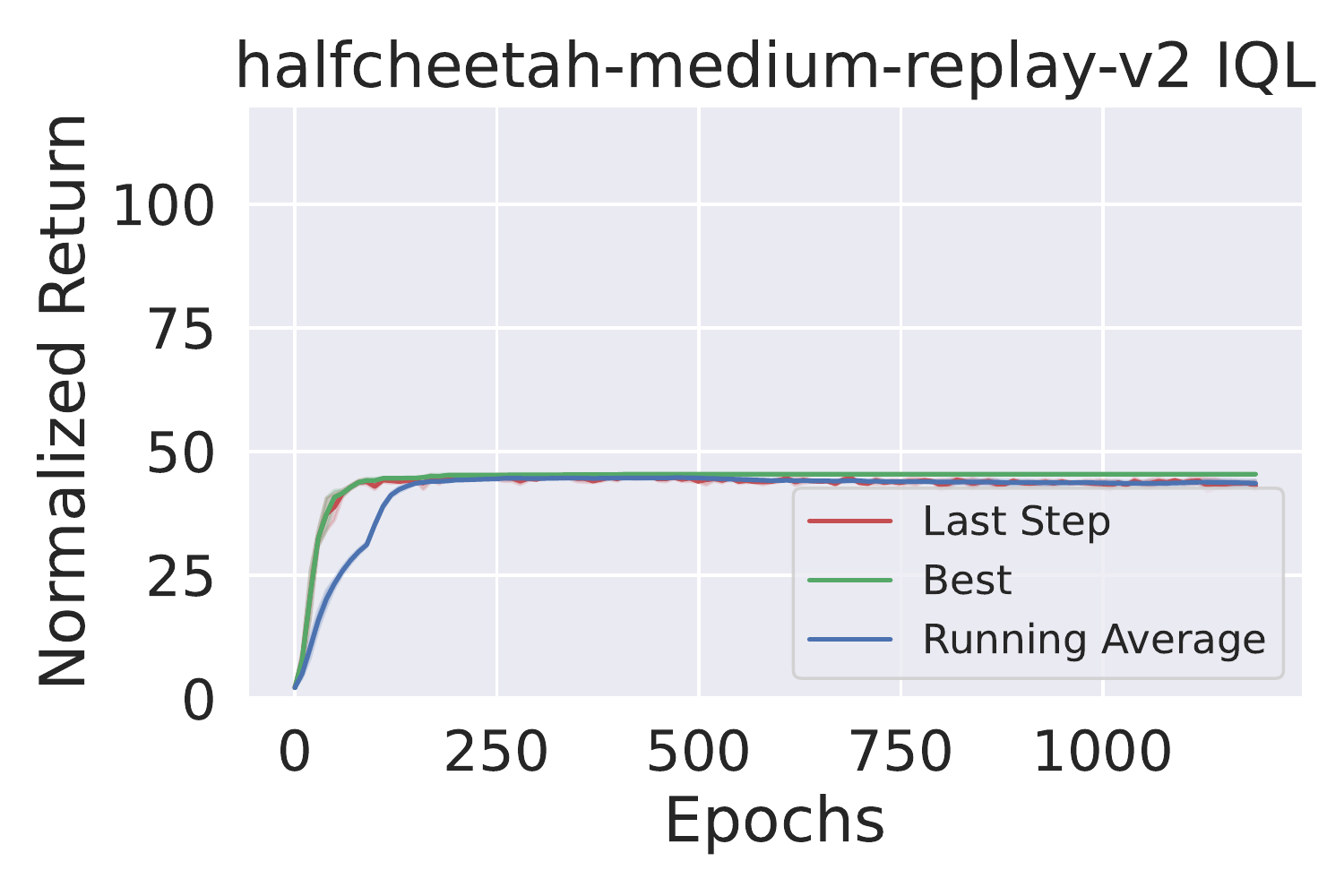}\\\includegraphics[width=0.225\linewidth]{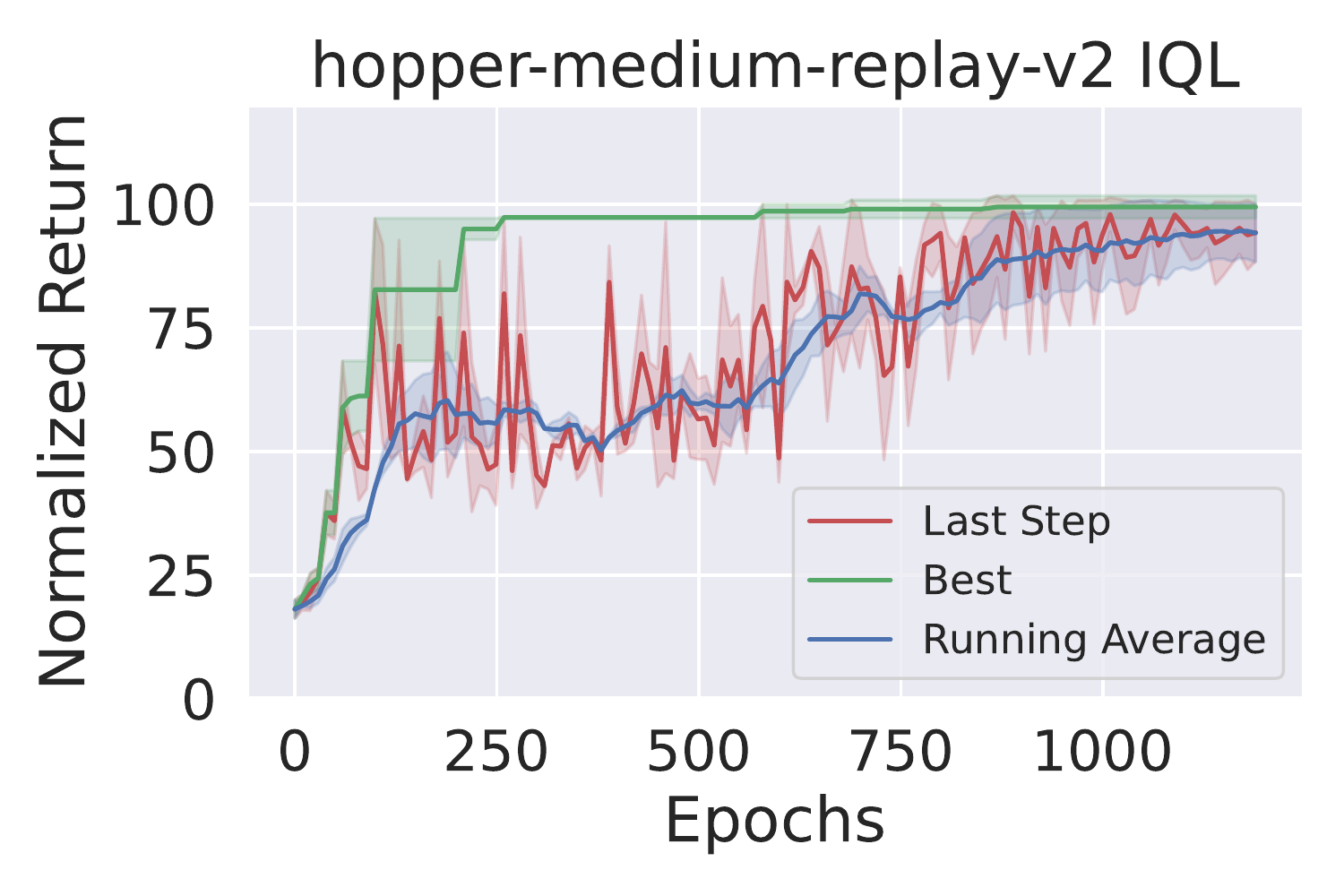}&\includegraphics[width=0.225\linewidth]{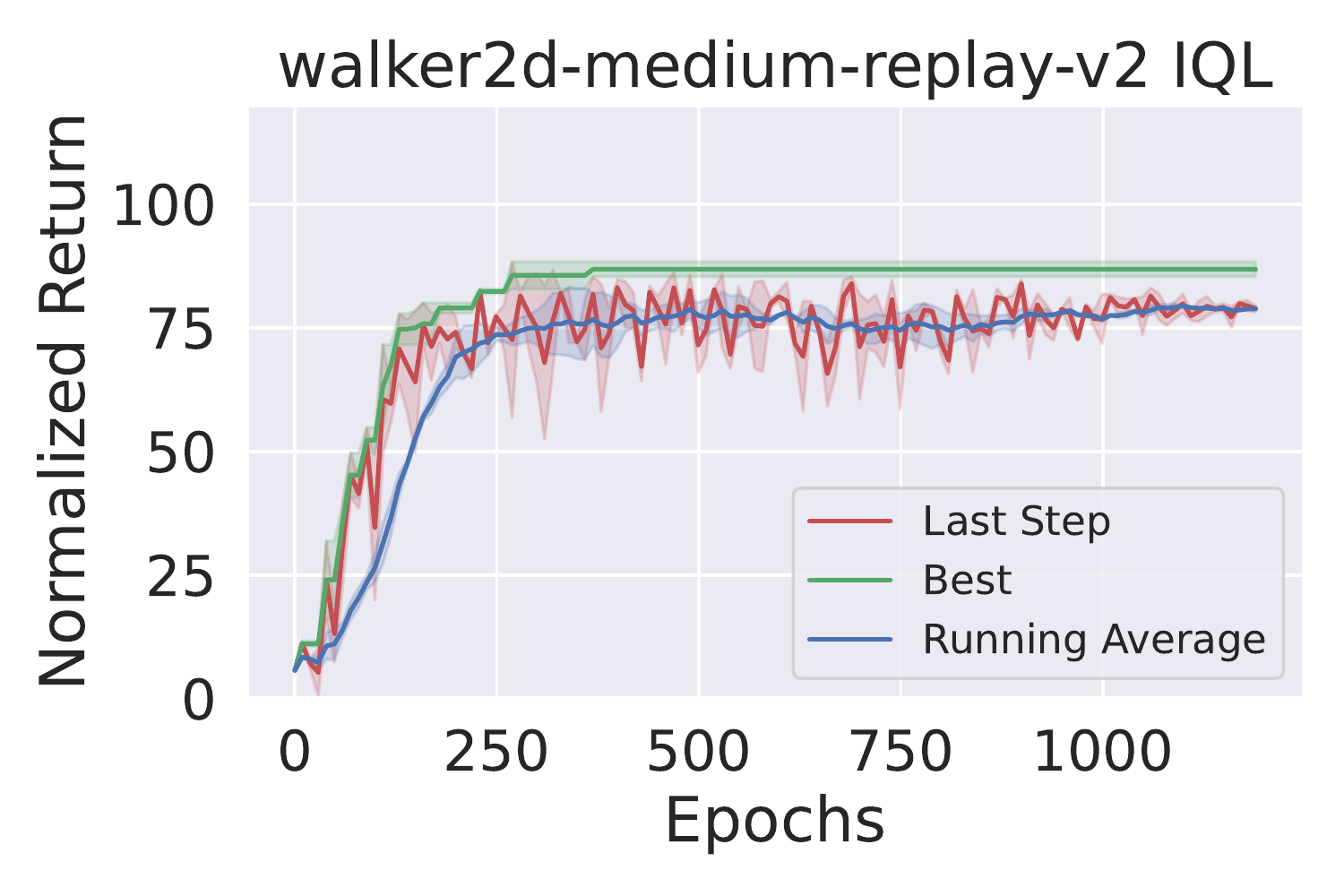}&\includegraphics[width=0.225\linewidth]{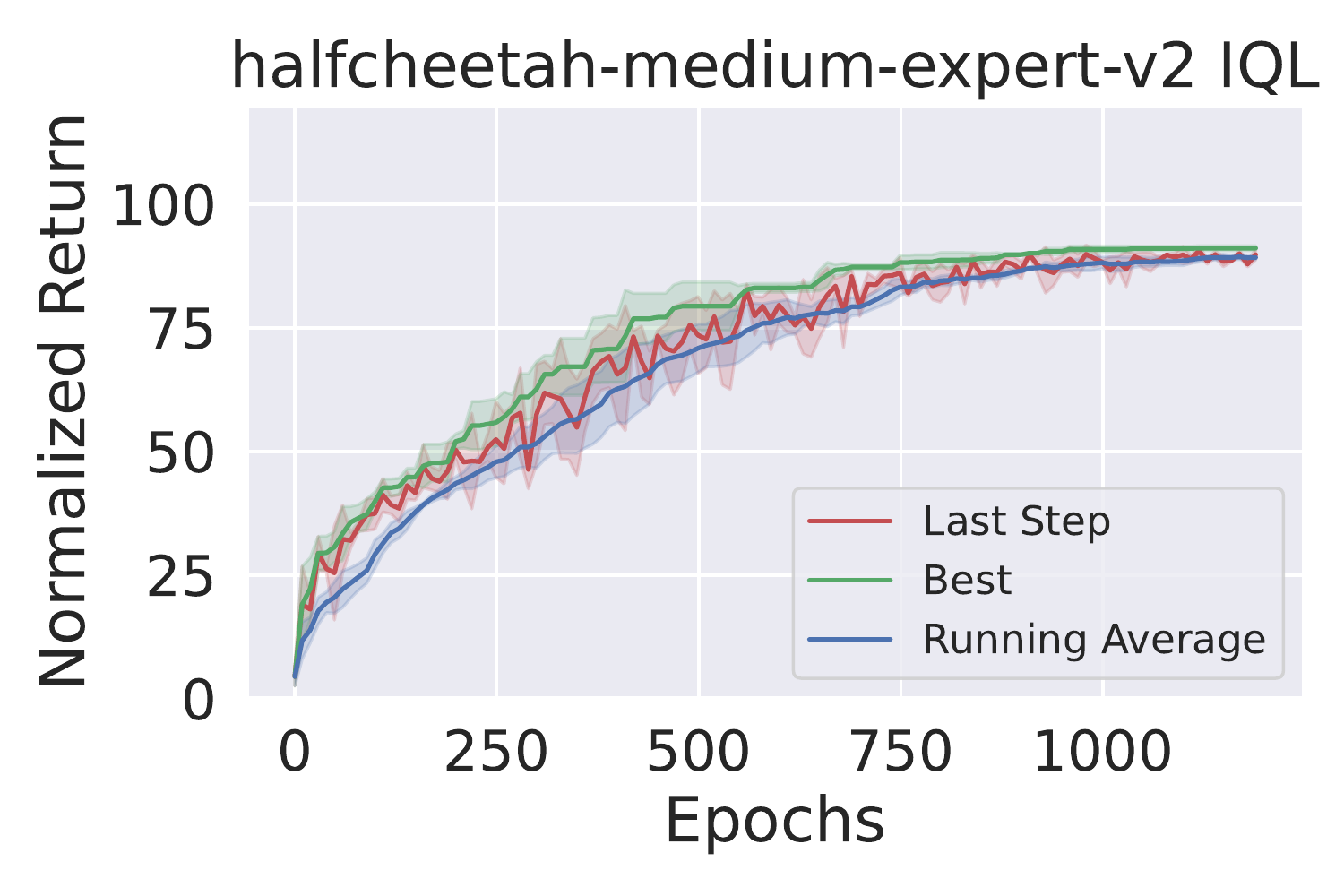}&\includegraphics[width=0.225\linewidth]{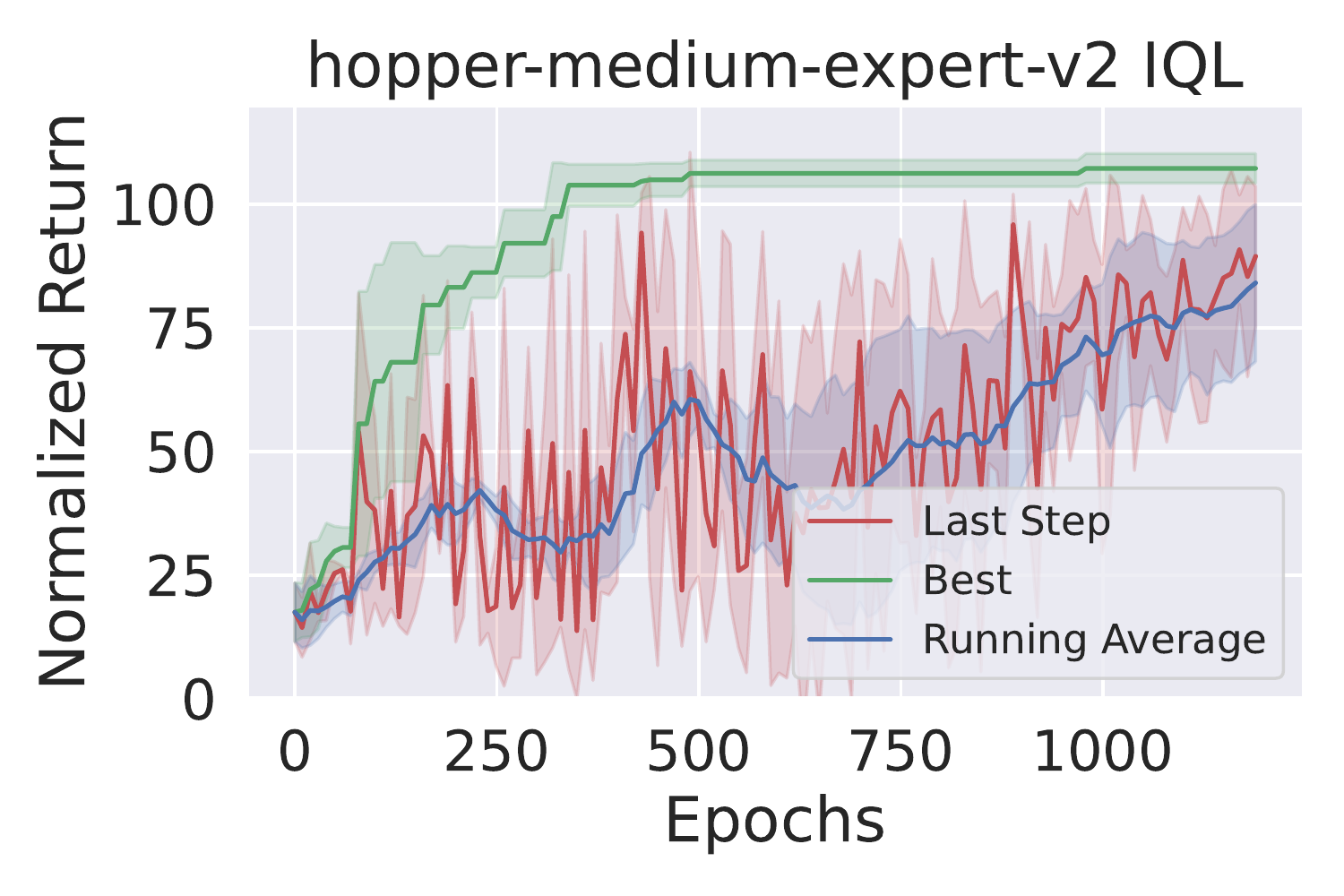}\\\includegraphics[width=0.225\linewidth]{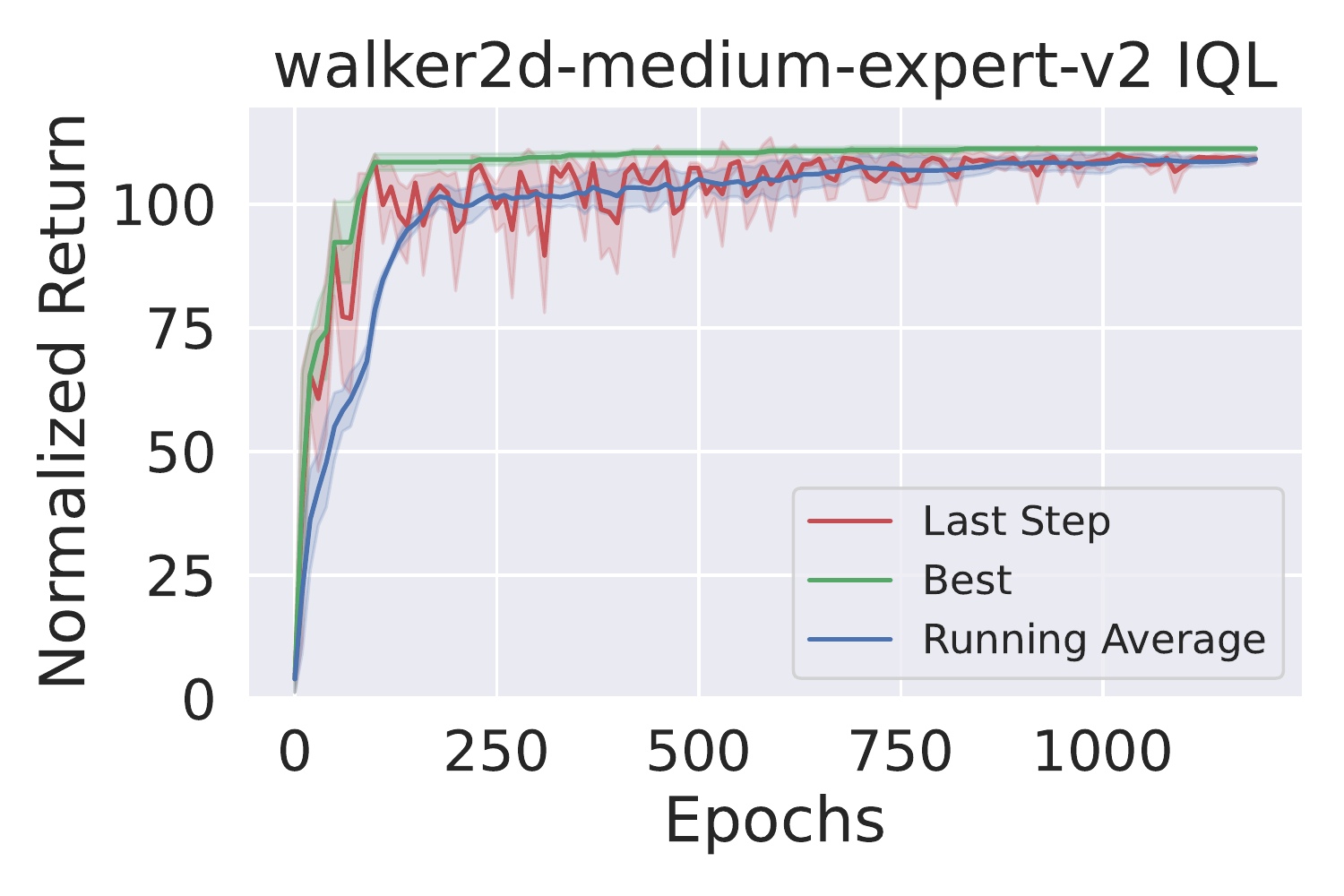}&\includegraphics[width=0.225\linewidth]{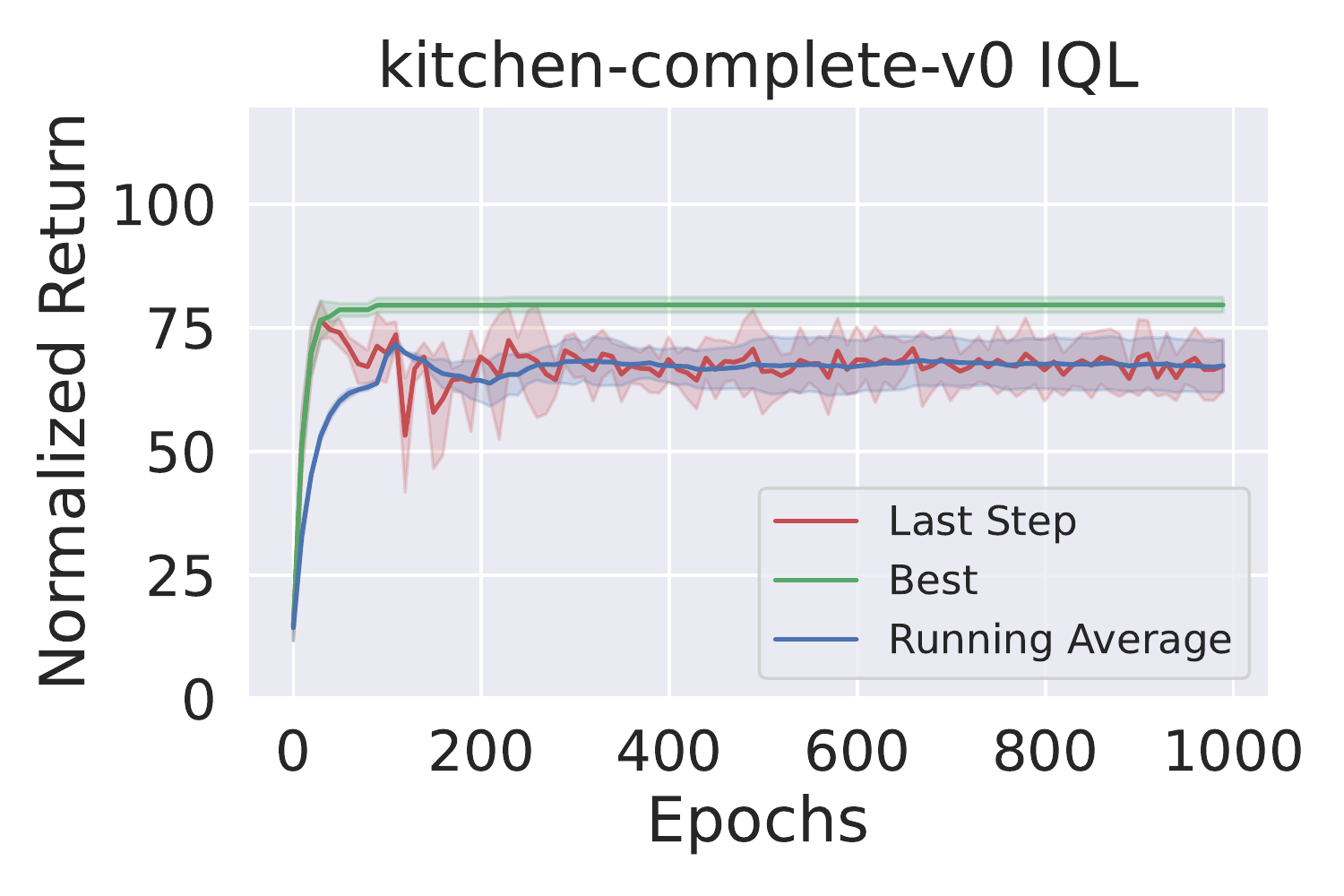}&\includegraphics[width=0.225\linewidth]{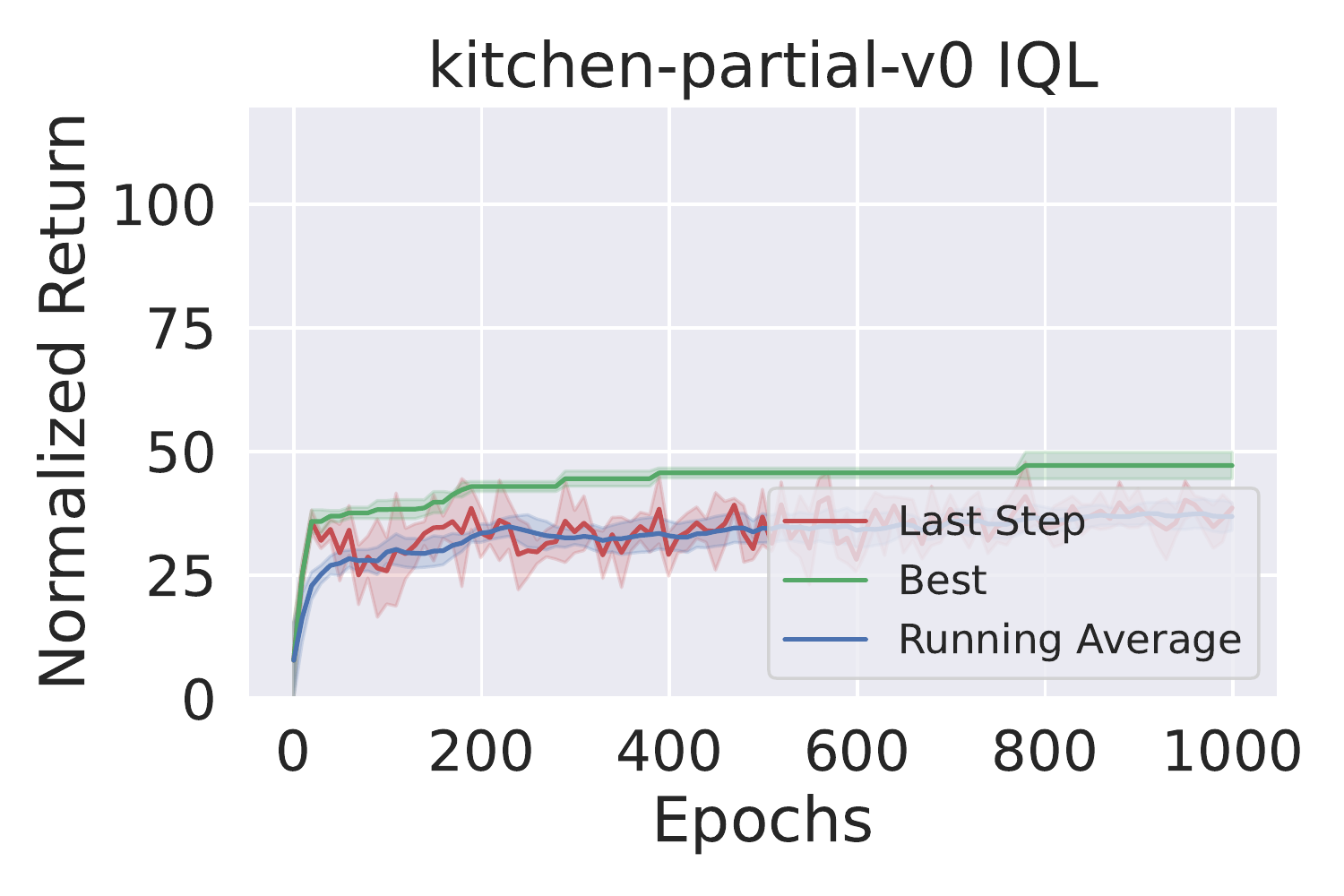}&\includegraphics[width=0.225\linewidth]{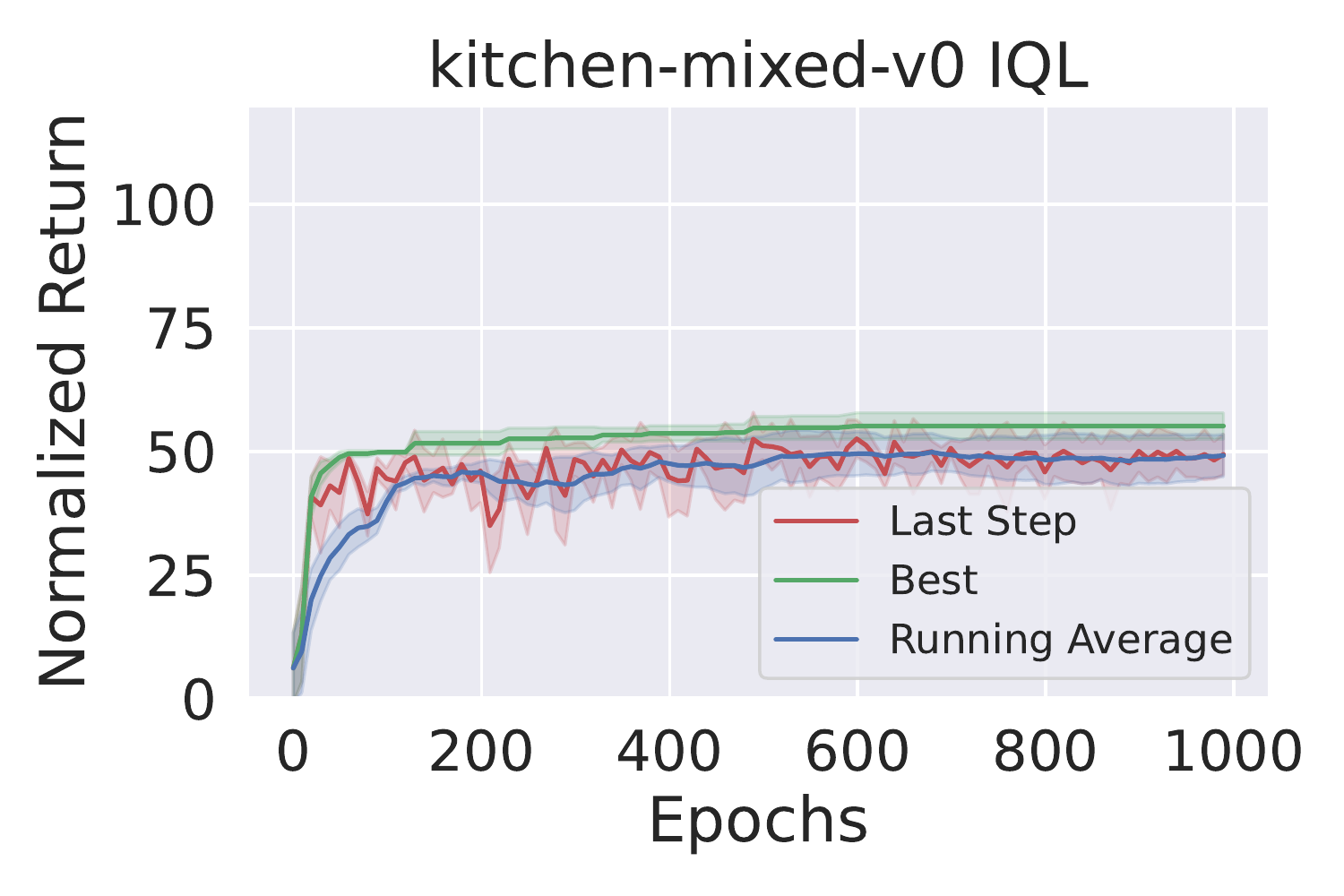}\\\includegraphics[width=0.225\linewidth]{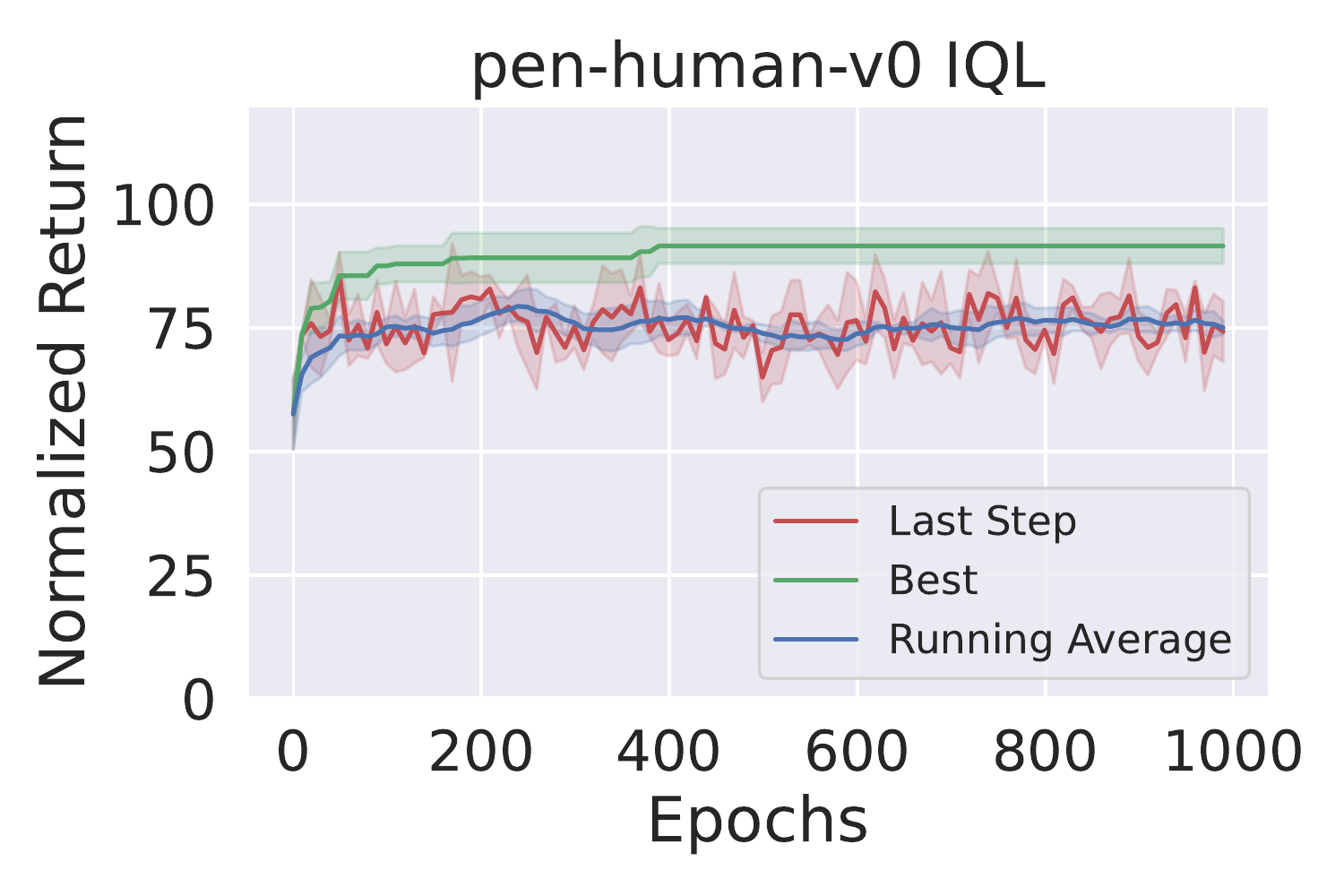}&\includegraphics[width=0.225\linewidth]{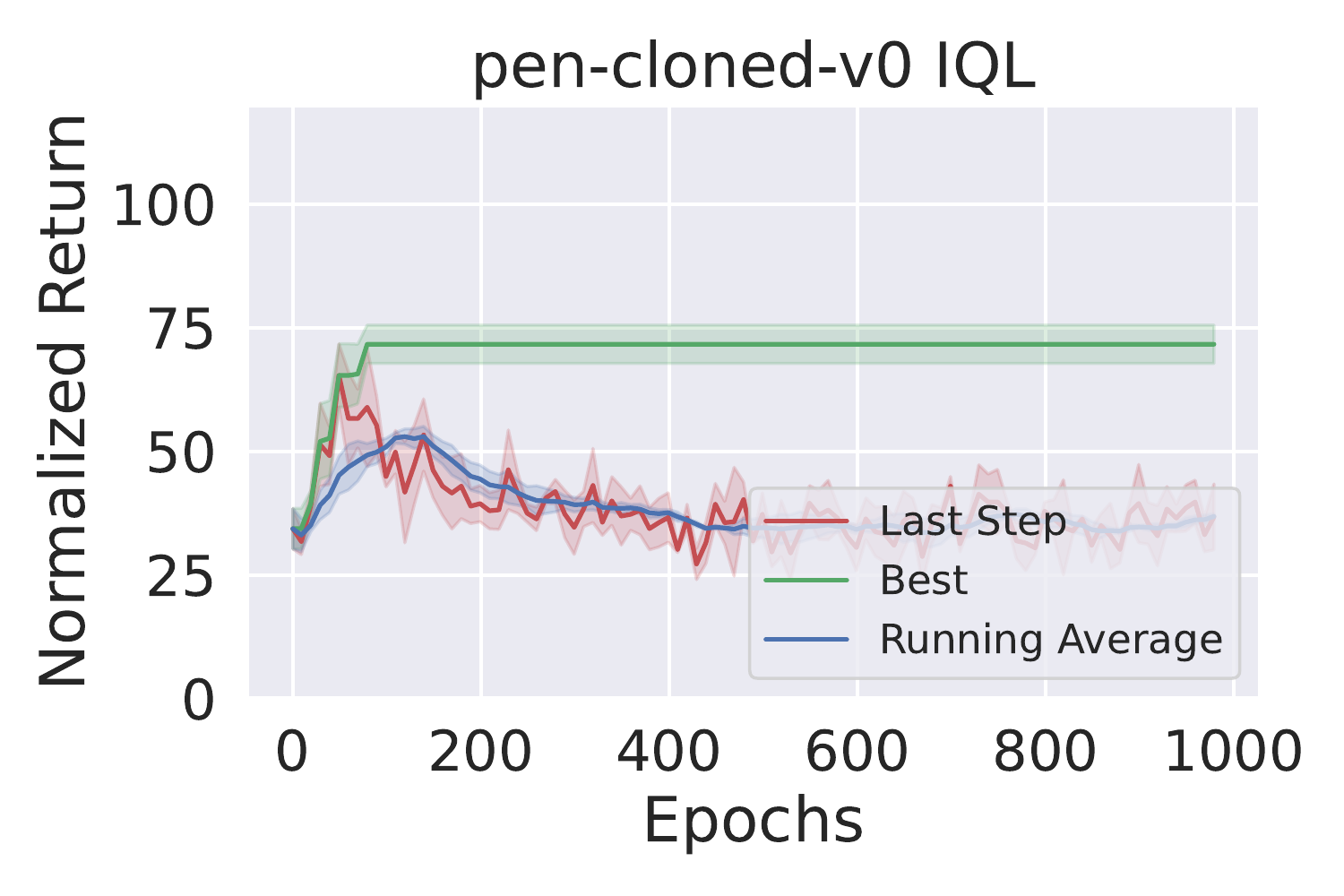}&\includegraphics[width=0.225\linewidth]{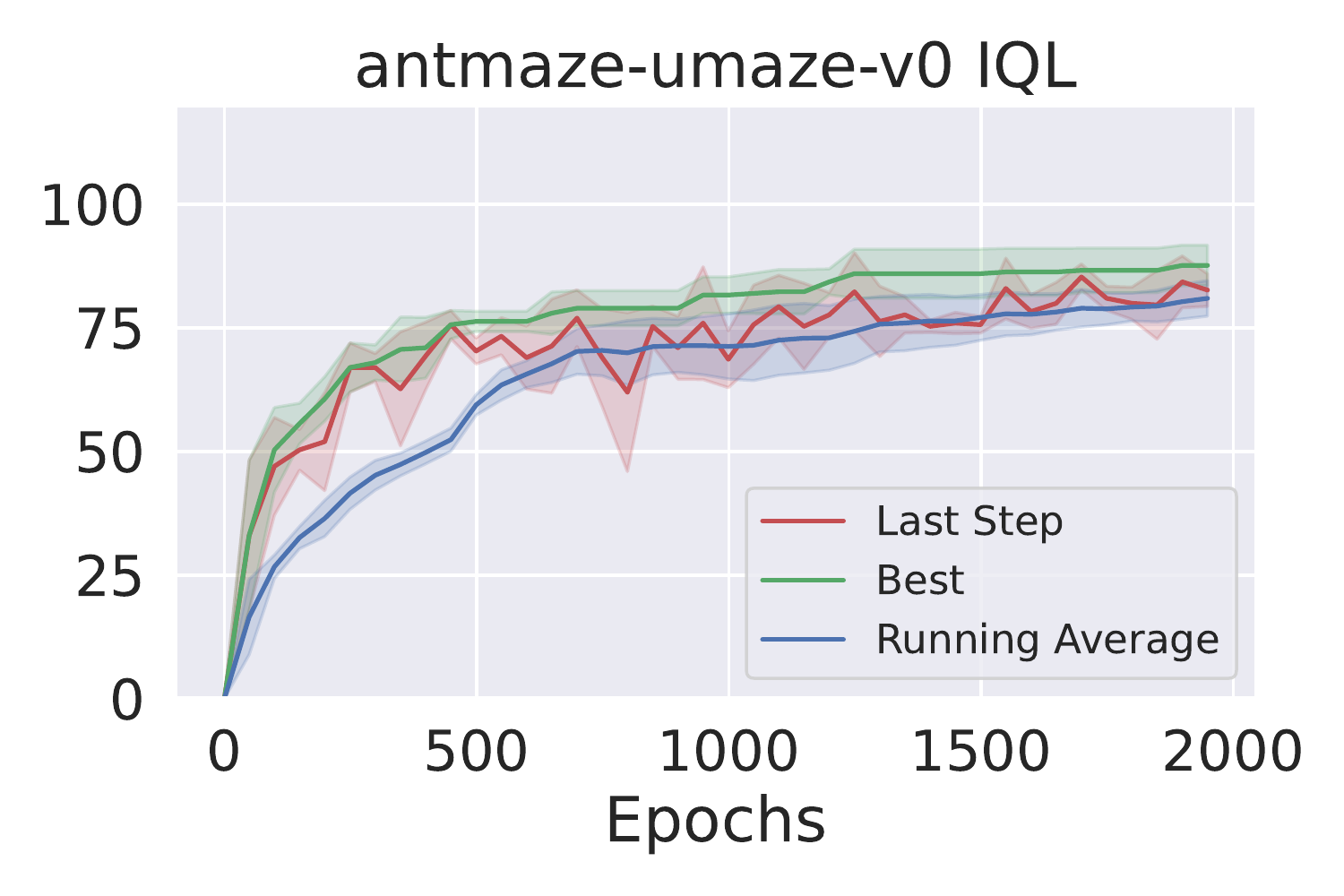}&\includegraphics[width=0.225\linewidth]{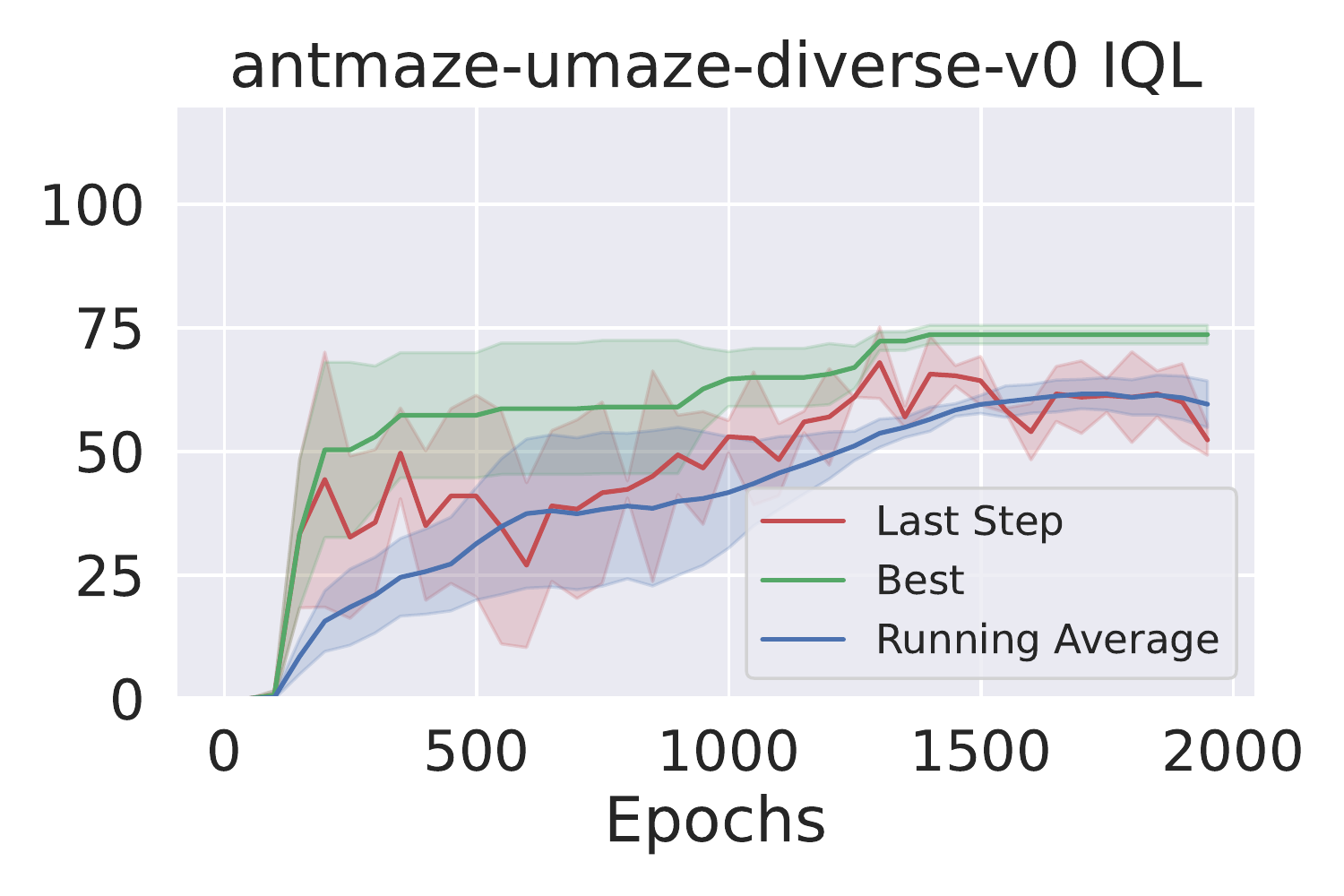}\\\includegraphics[width=0.225\linewidth]{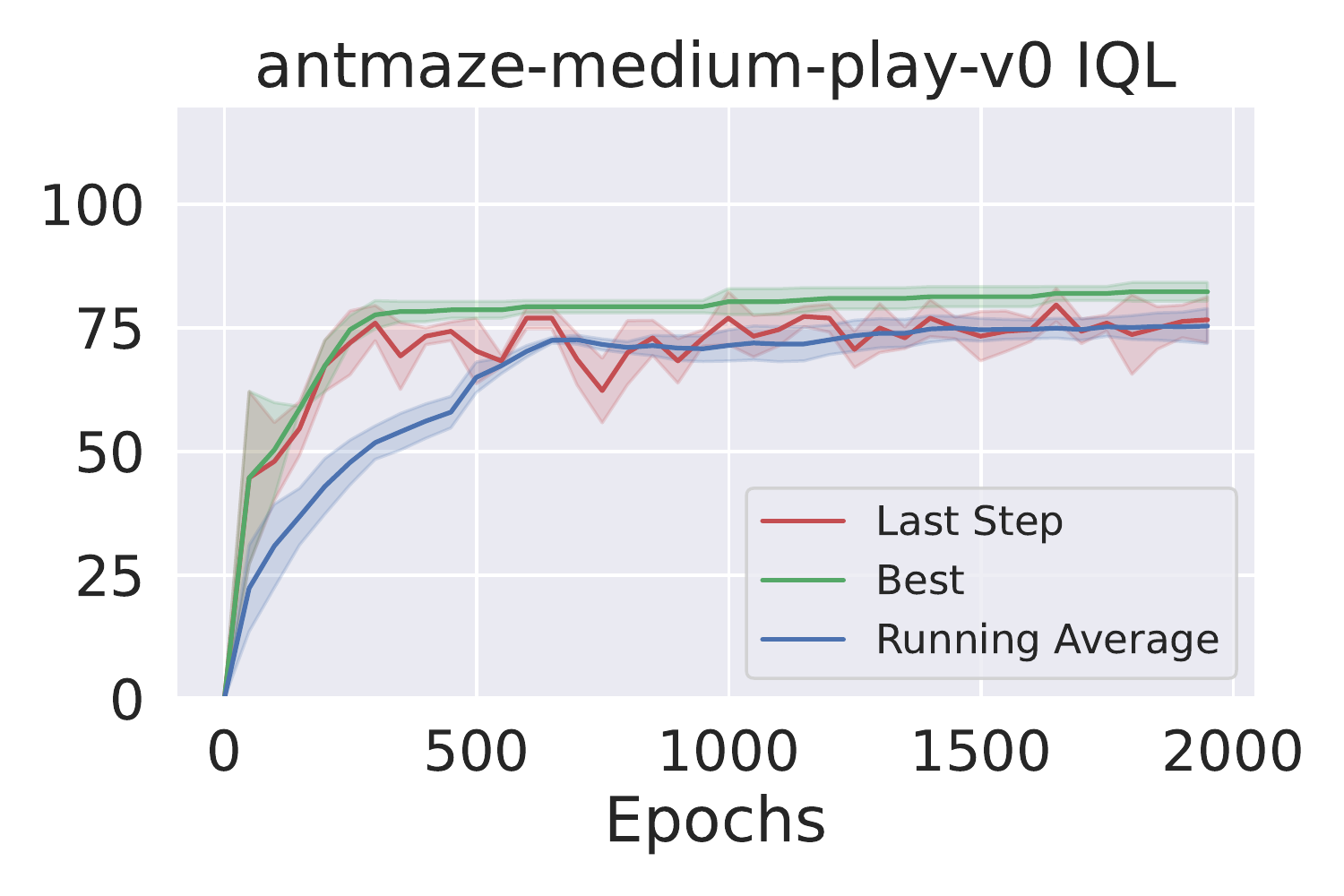}&\includegraphics[width=0.225\linewidth]{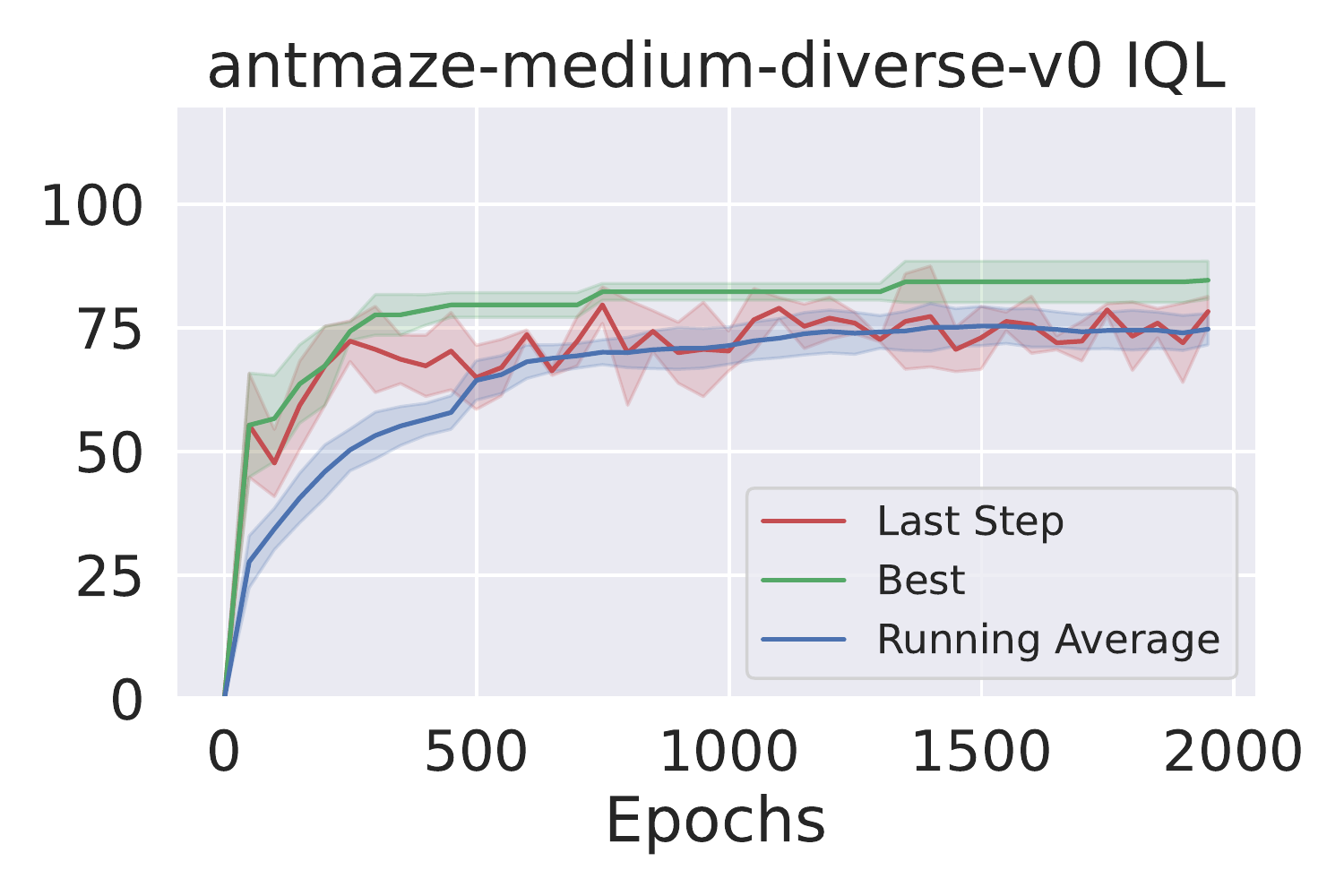}&\includegraphics[width=0.225\linewidth]{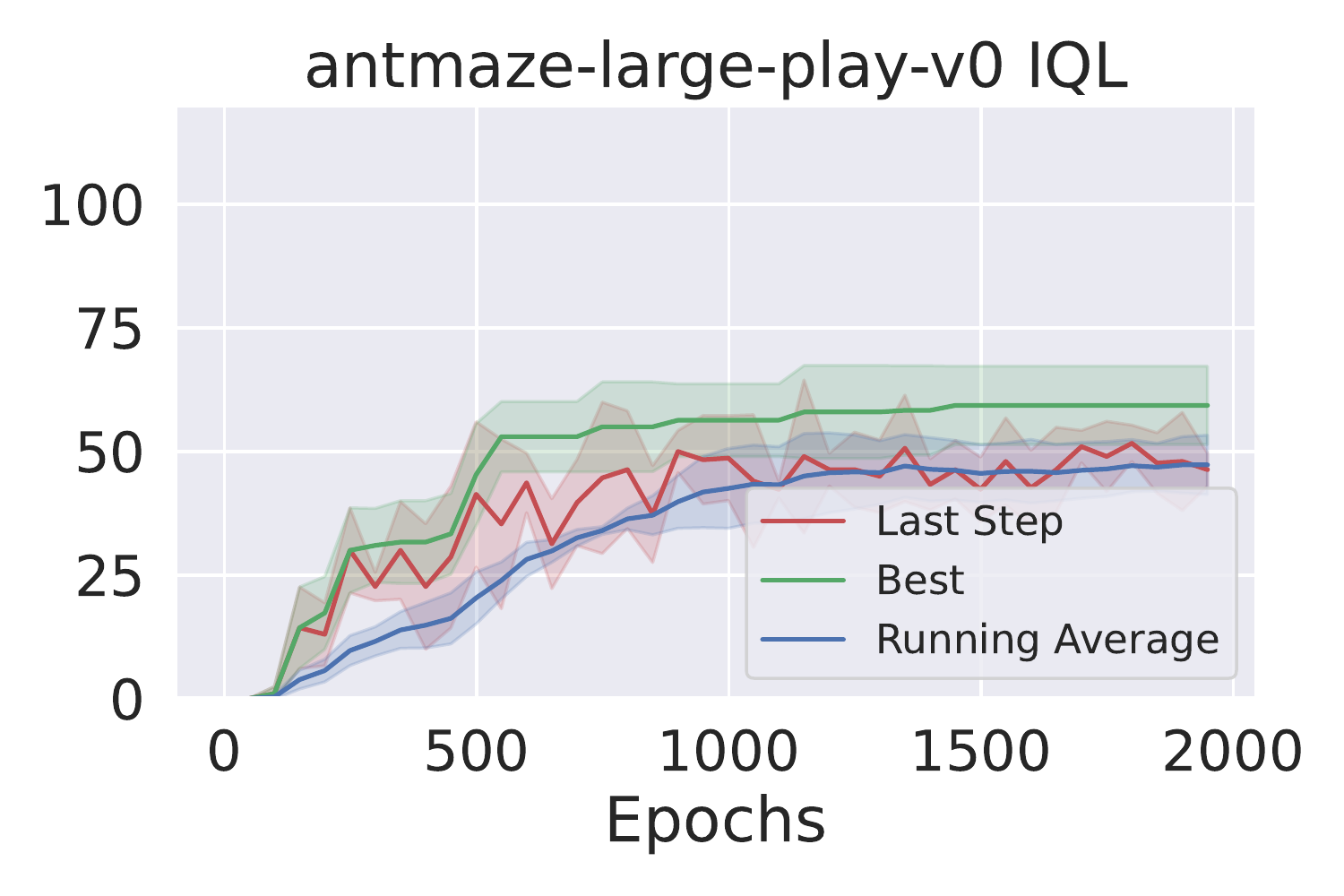}&\includegraphics[width=0.225\linewidth]{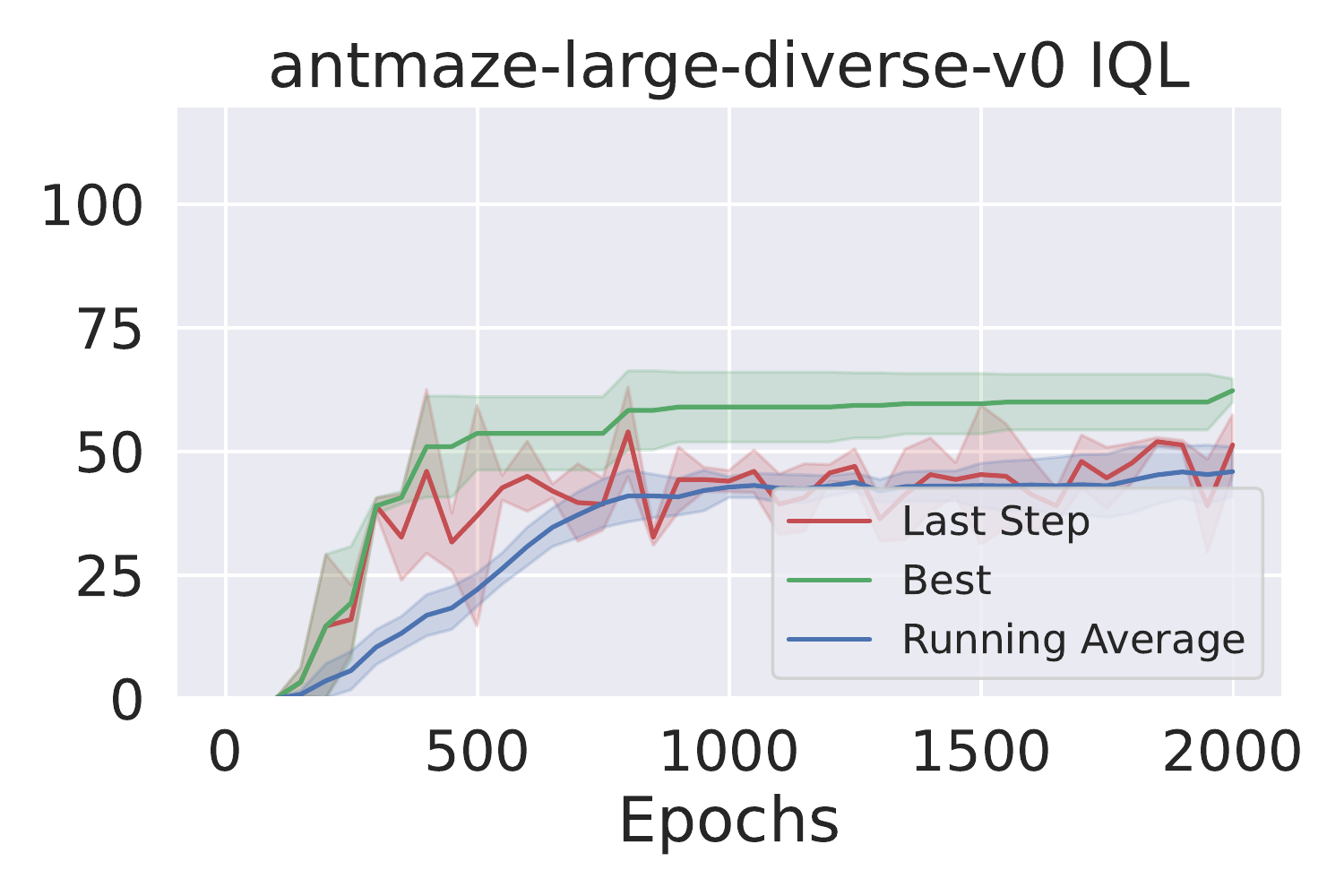}\\\end{tabular}
\centering
\caption{Training Curves of IQL on D4RL}
\end{figure*}
\begin{figure*}[htb]
\centering
\begin{tabular}{cccccc}
\includegraphics[width=0.225\linewidth]{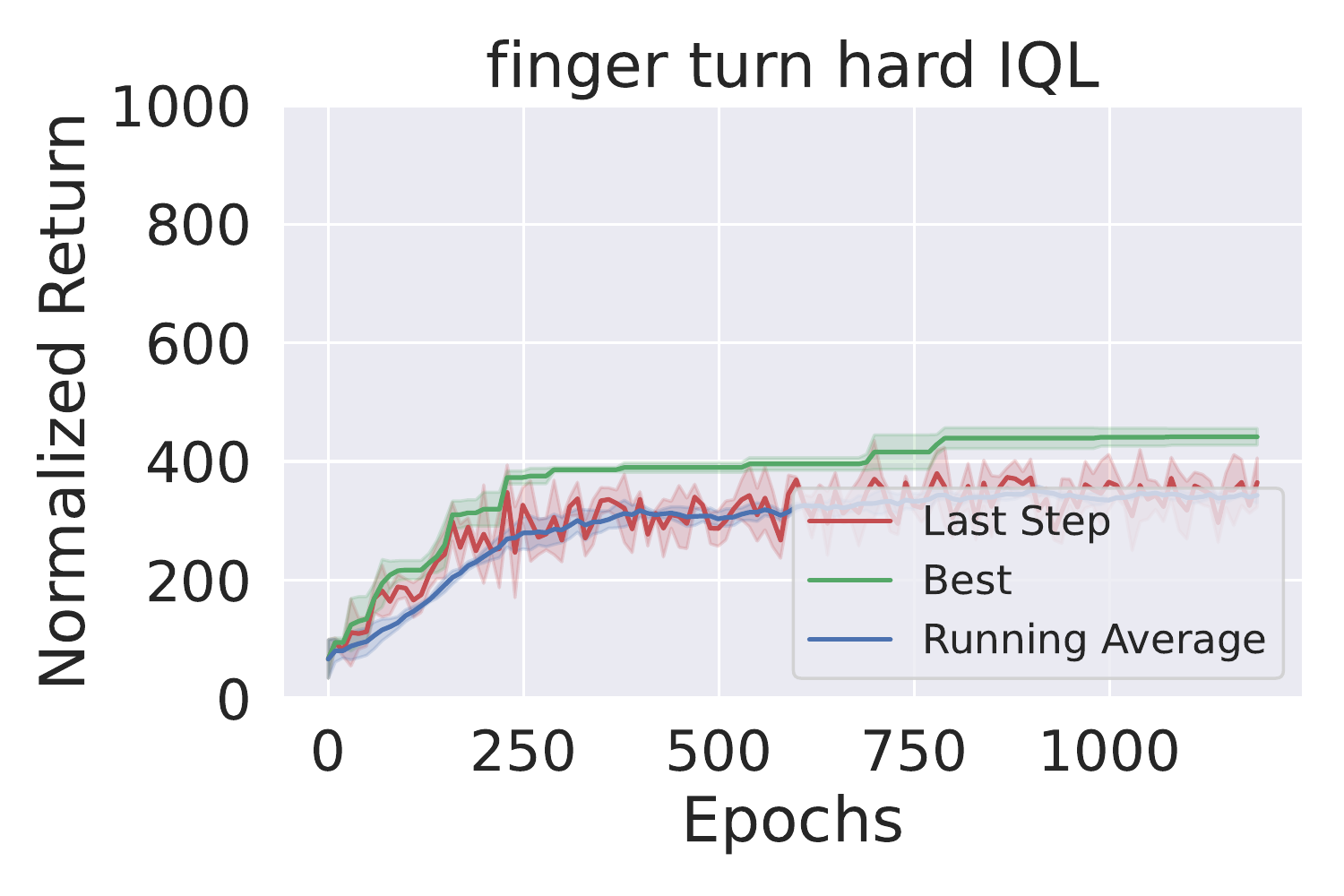}&\includegraphics[width=0.225\linewidth]{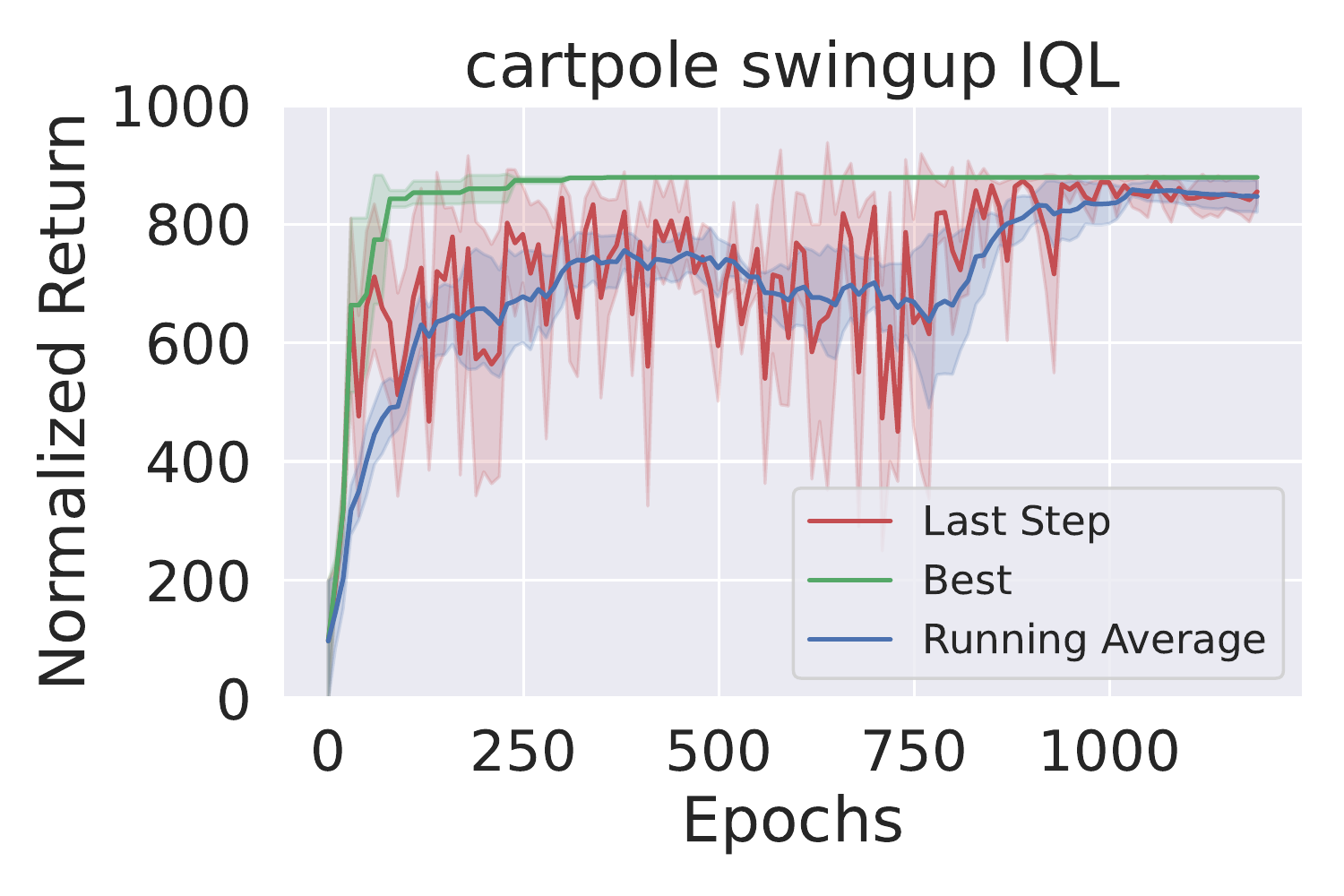}&\includegraphics[width=0.225\linewidth]{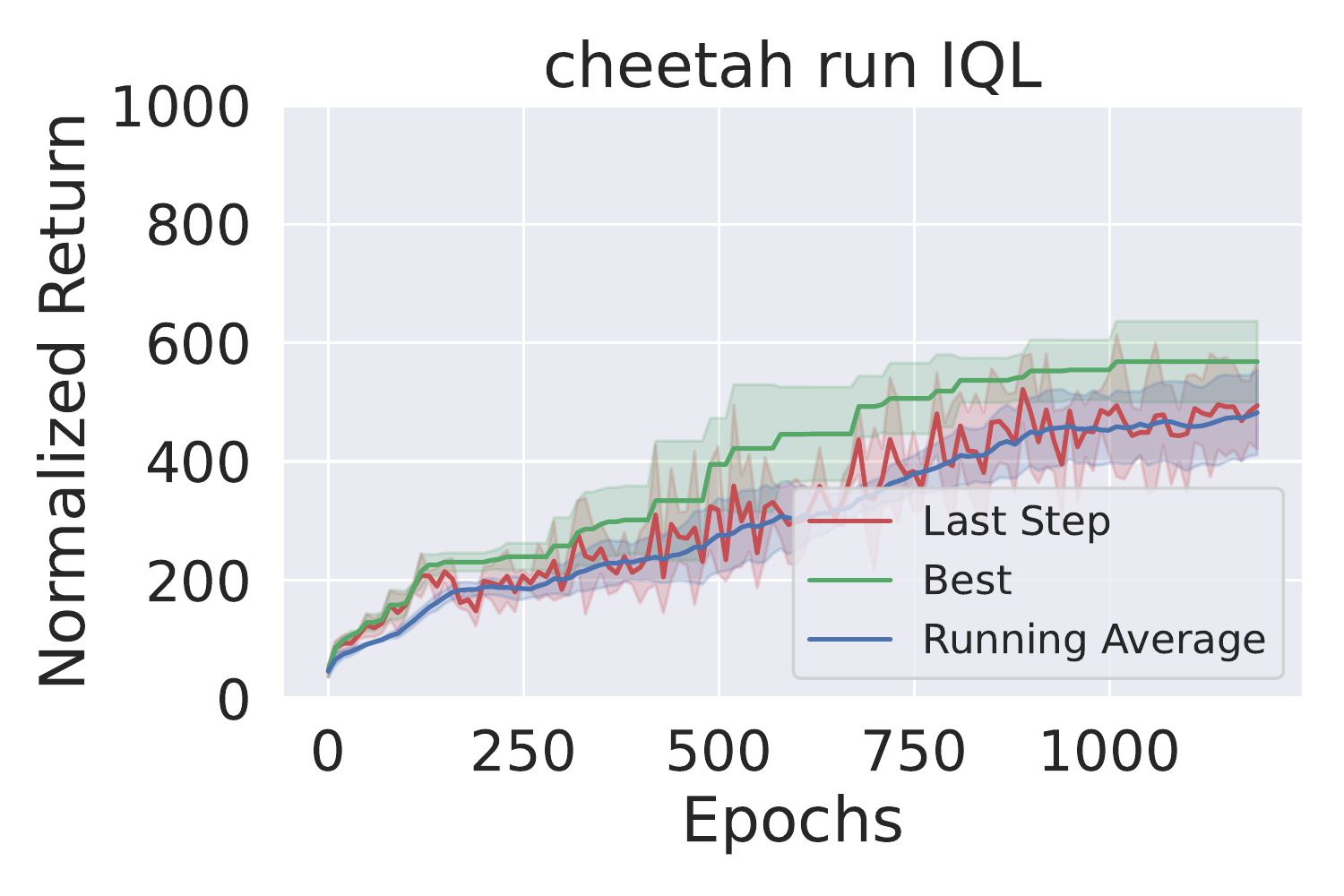}&\includegraphics[width=0.225\linewidth]{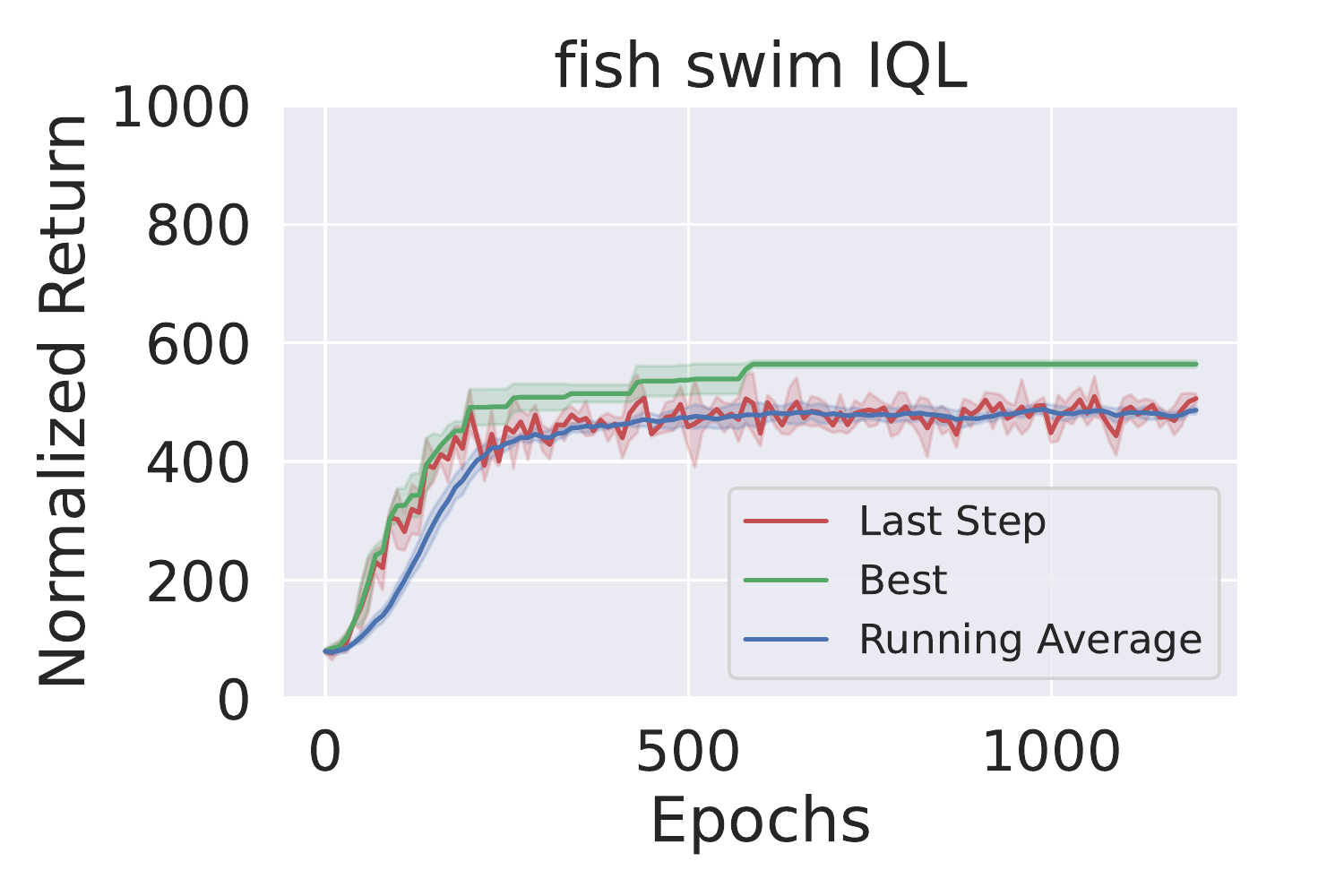}\\\includegraphics[width=0.225\linewidth]{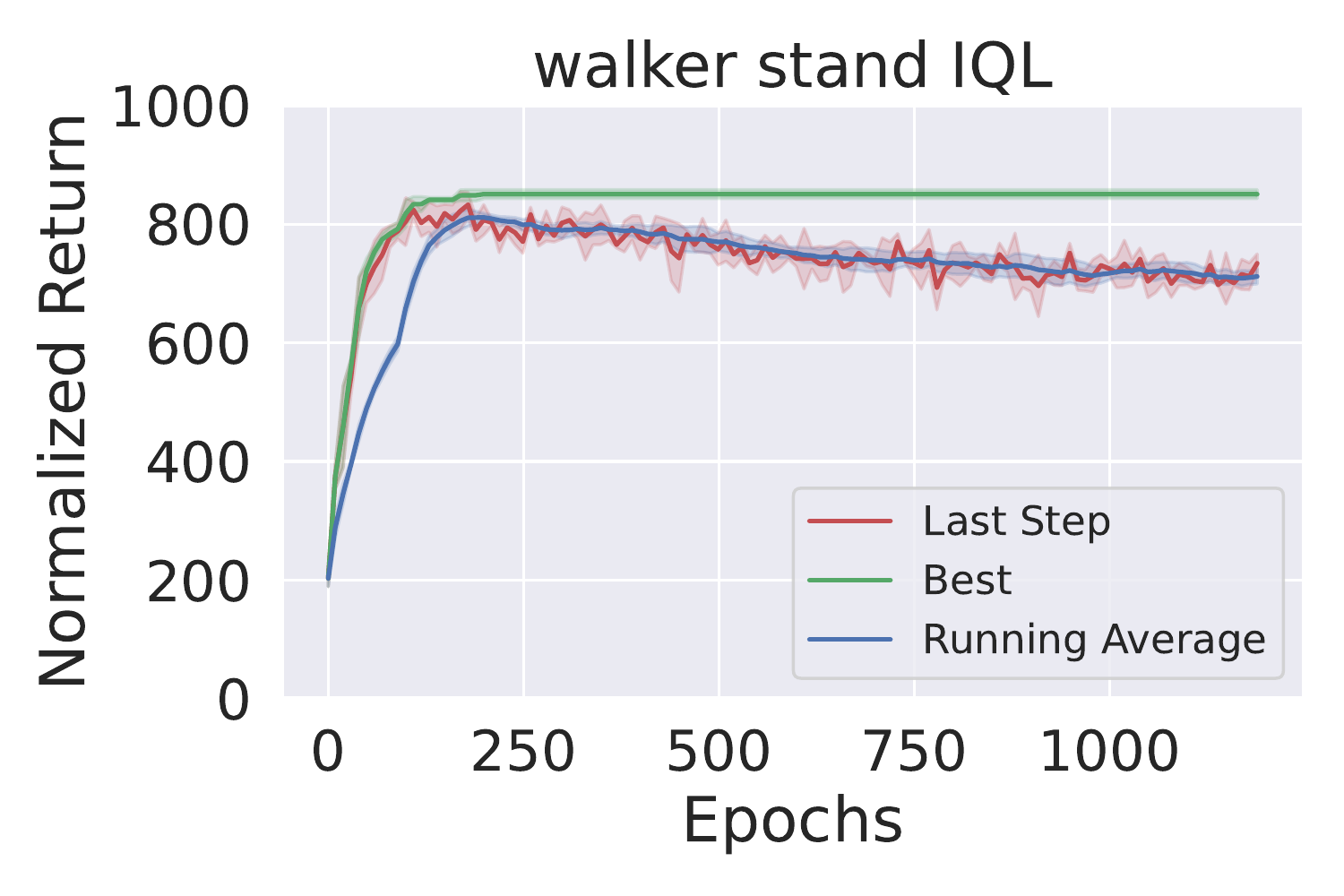}&\includegraphics[width=0.225\linewidth]{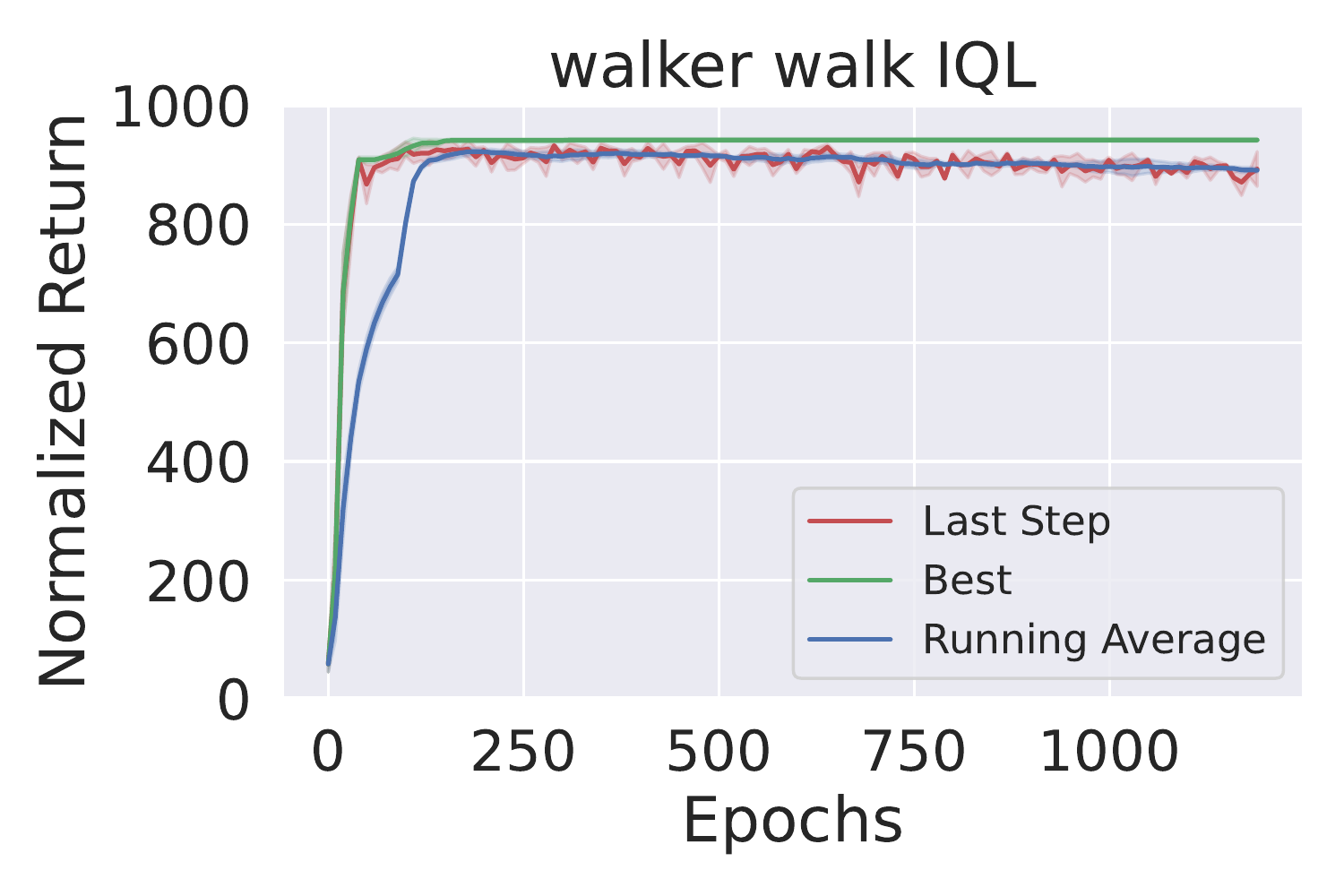}&&\\\end{tabular}
\centering
\caption{Training Curves of IQL on RLUP}
\end{figure*}
\begin{figure*}[htb]
\centering
\begin{tabular}{cccccc}
\includegraphics[width=0.225\linewidth]{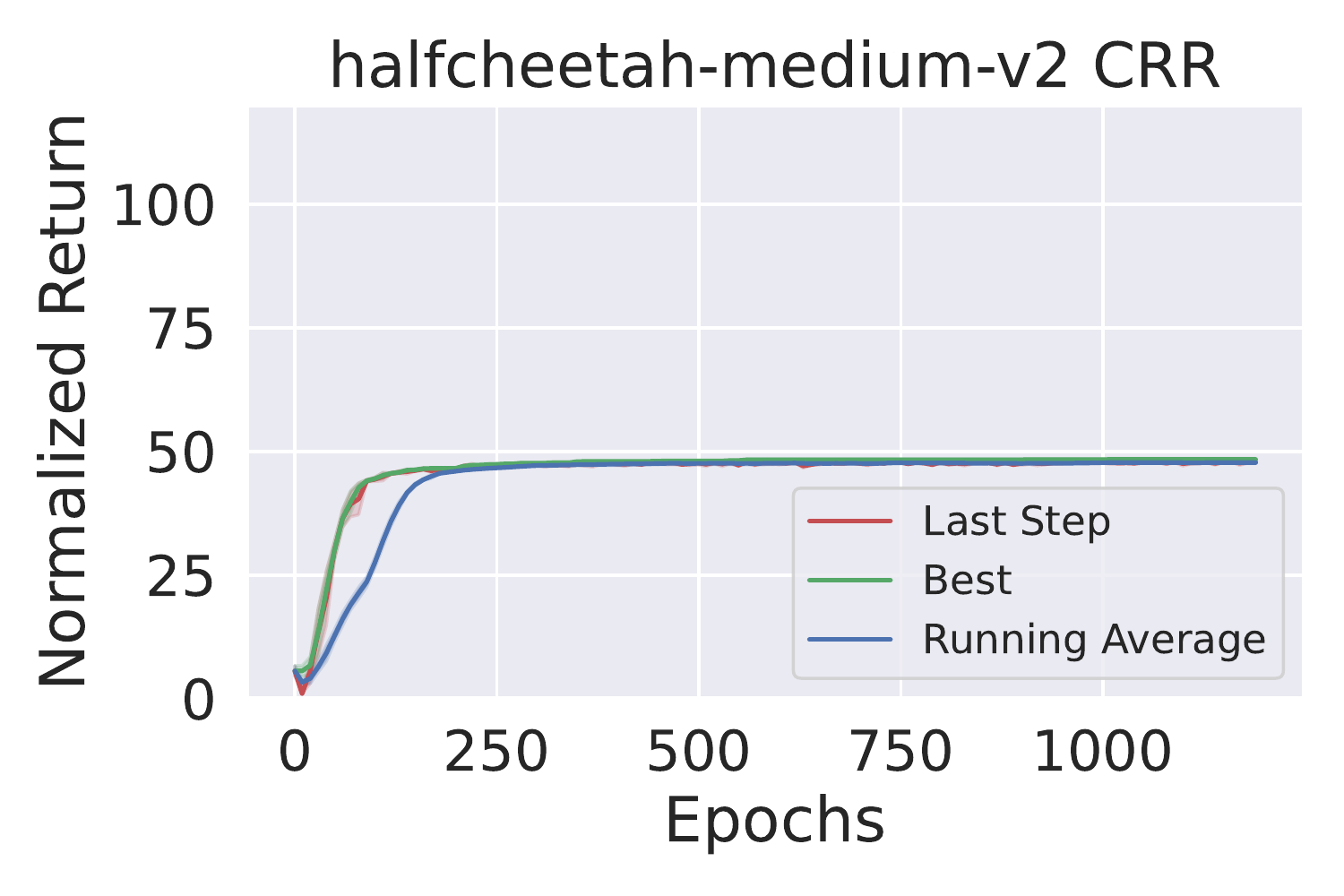}&\includegraphics[width=0.225\linewidth]{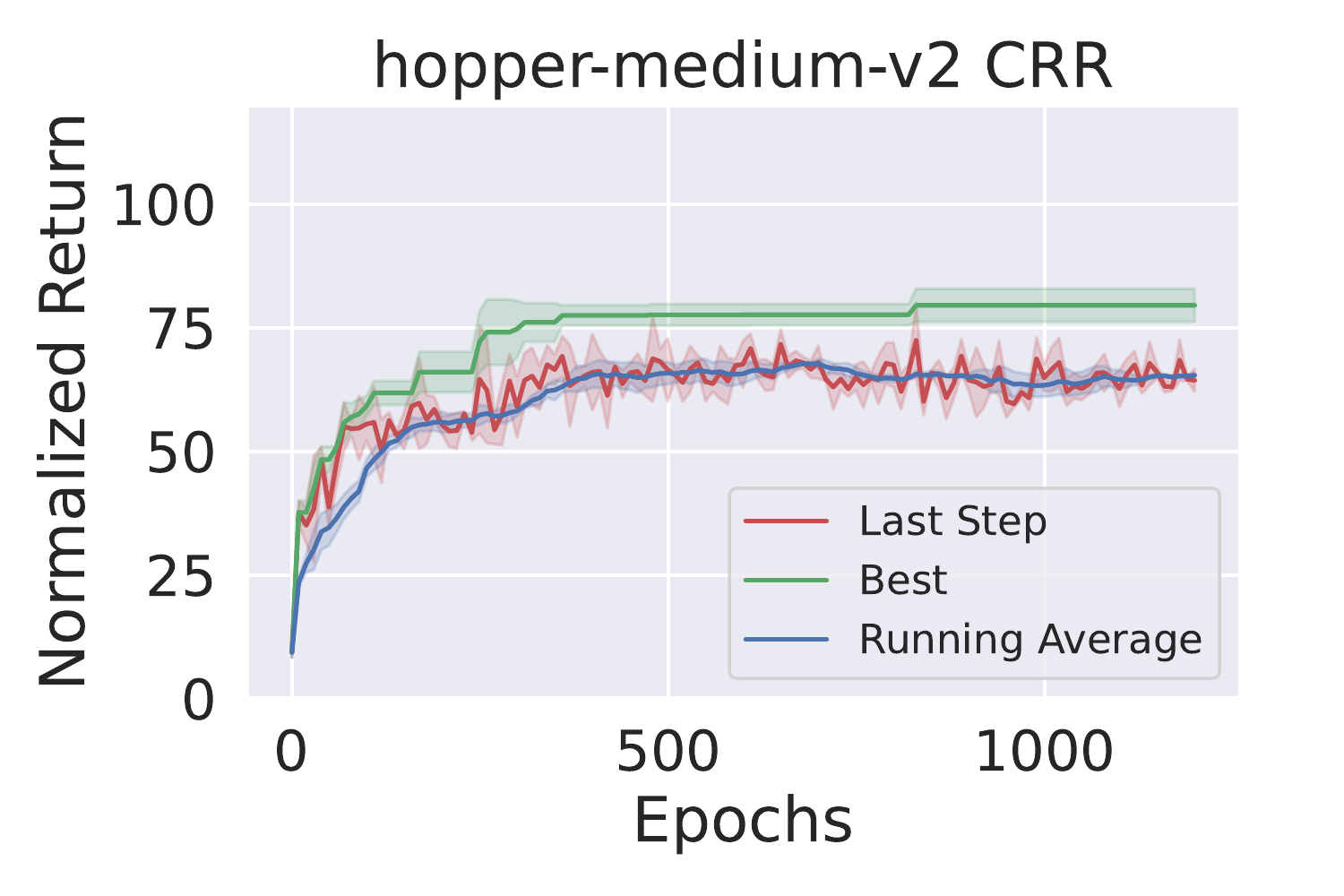}&\includegraphics[width=0.225\linewidth]{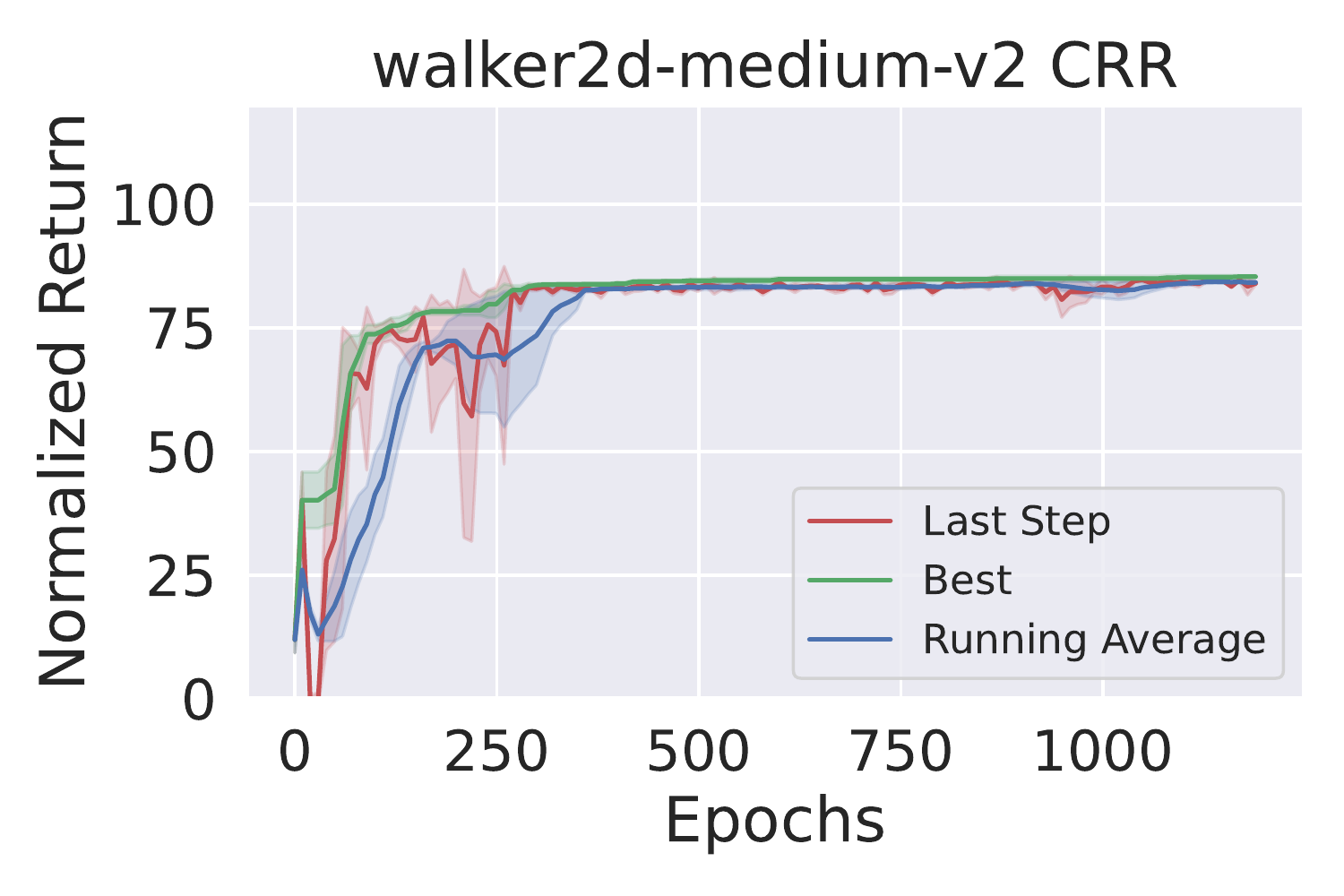}&\includegraphics[width=0.225\linewidth]{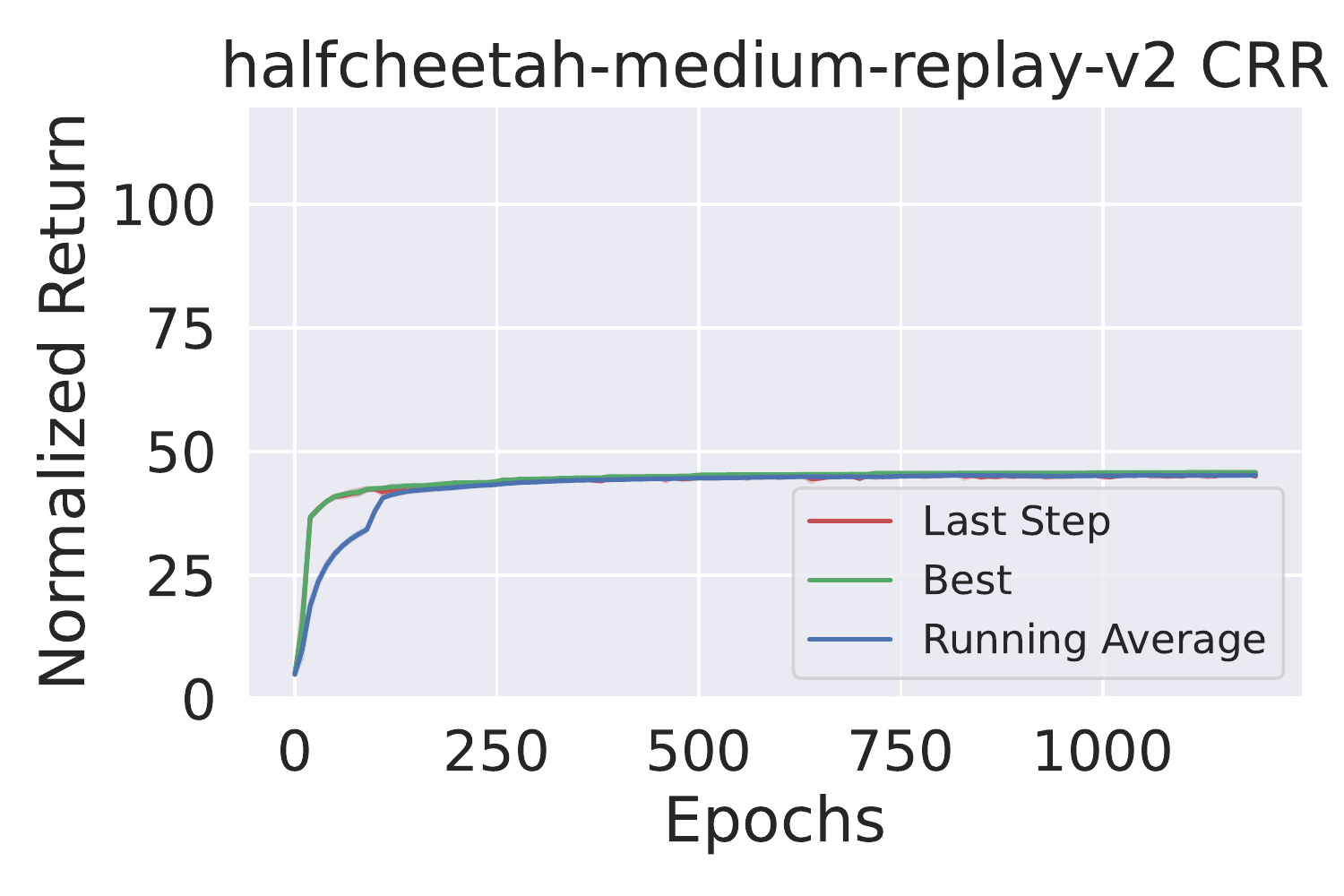}\\\includegraphics[width=0.225\linewidth]{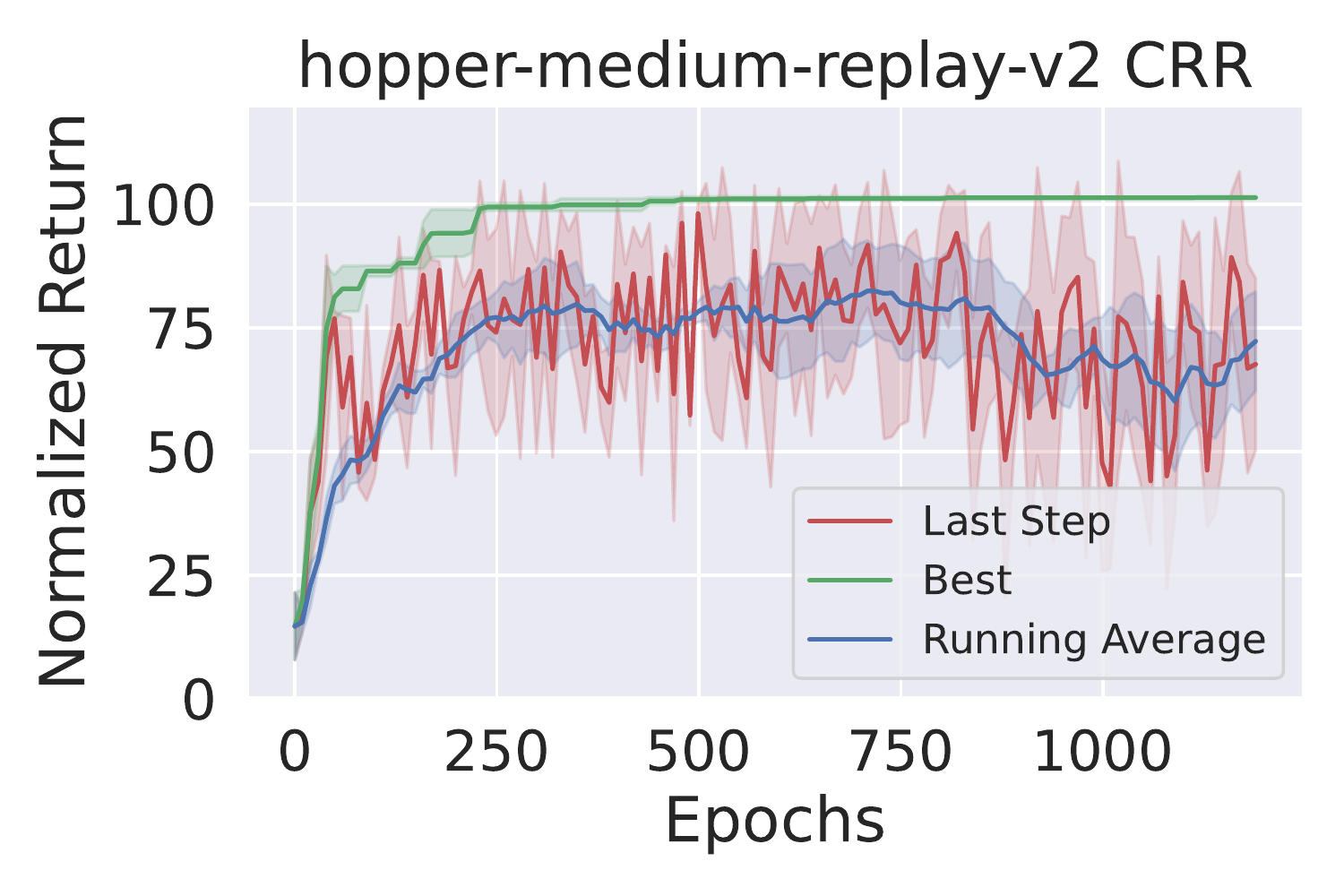}&\includegraphics[width=0.225\linewidth]{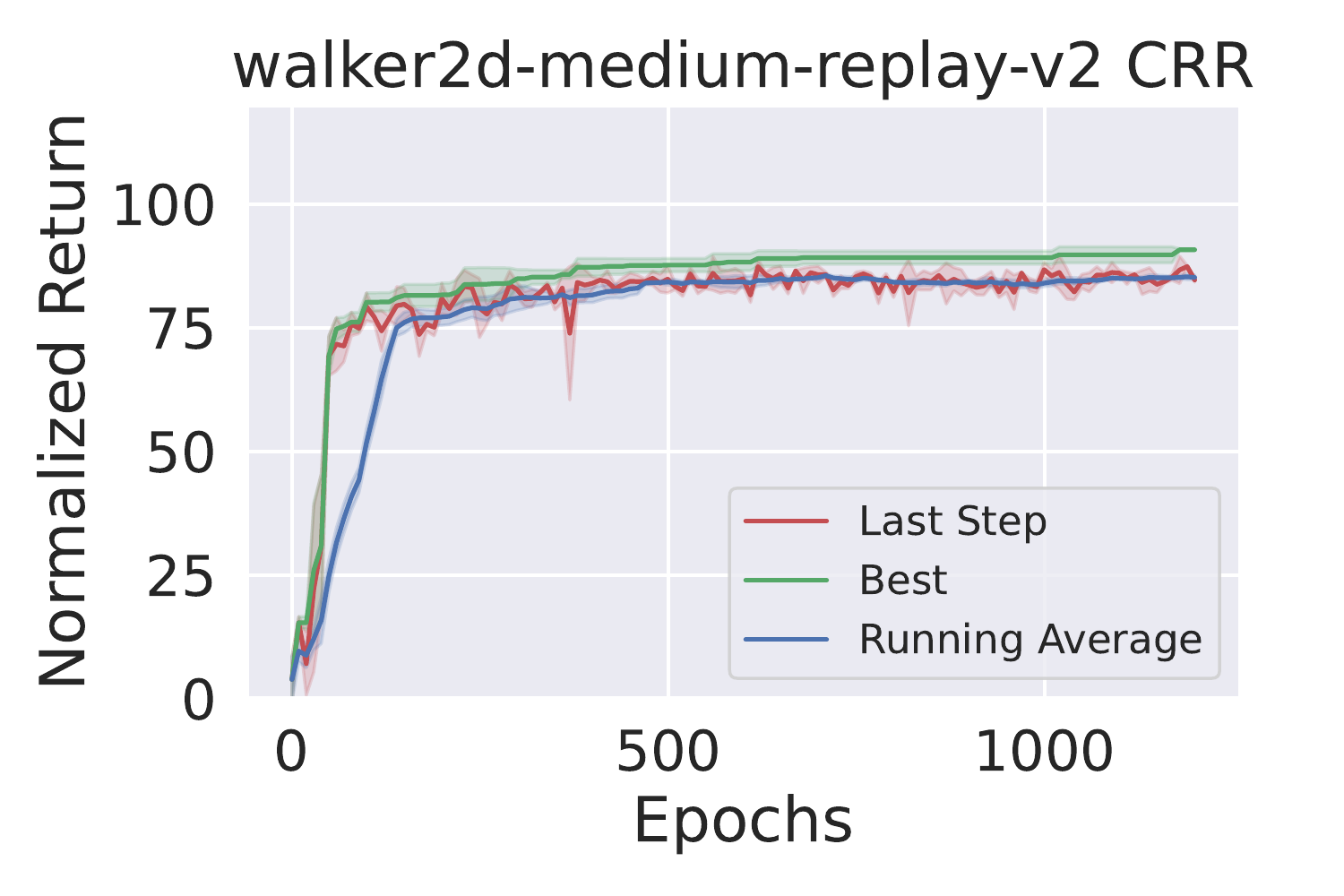}&\includegraphics[width=0.225\linewidth]{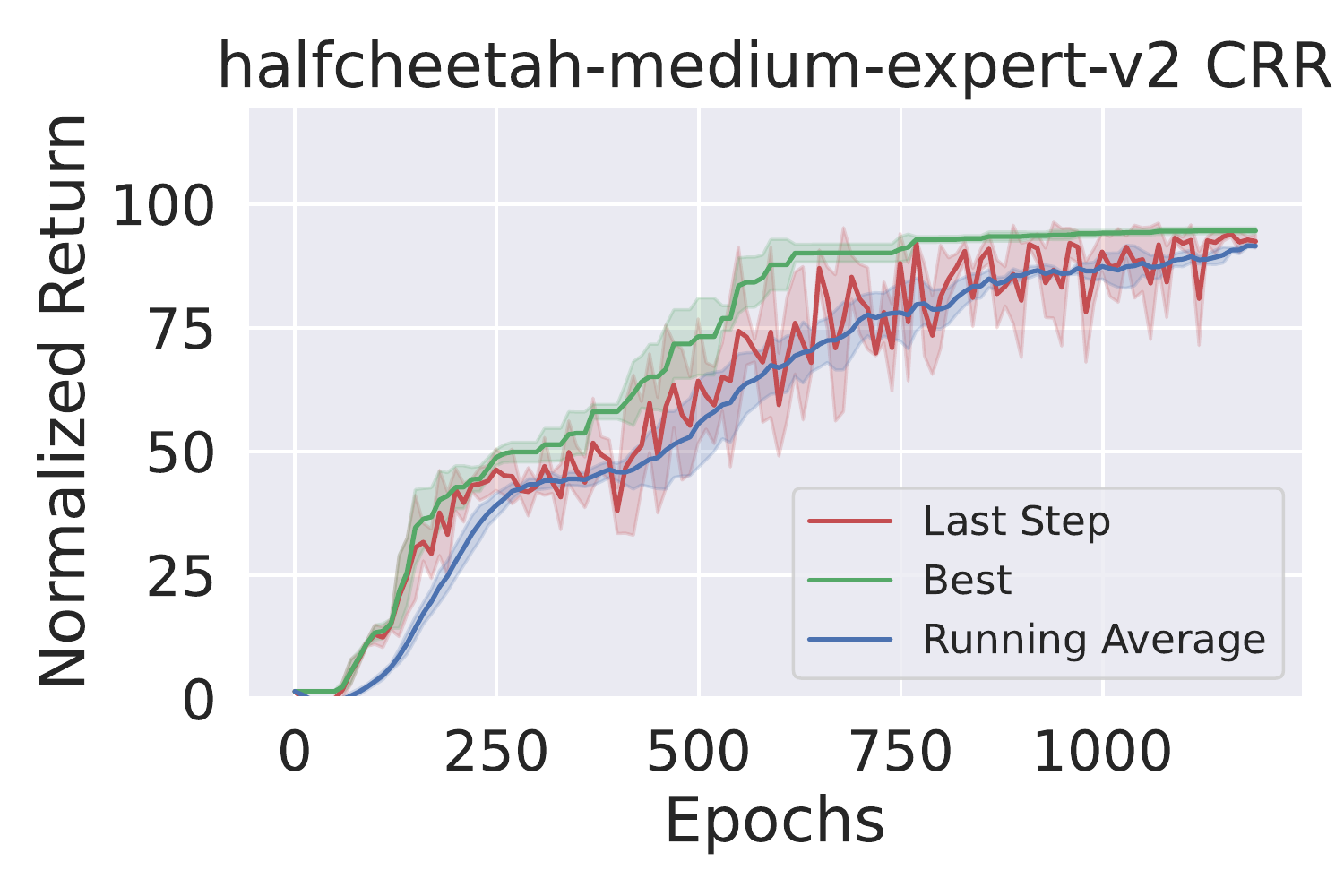}&\includegraphics[width=0.225\linewidth]{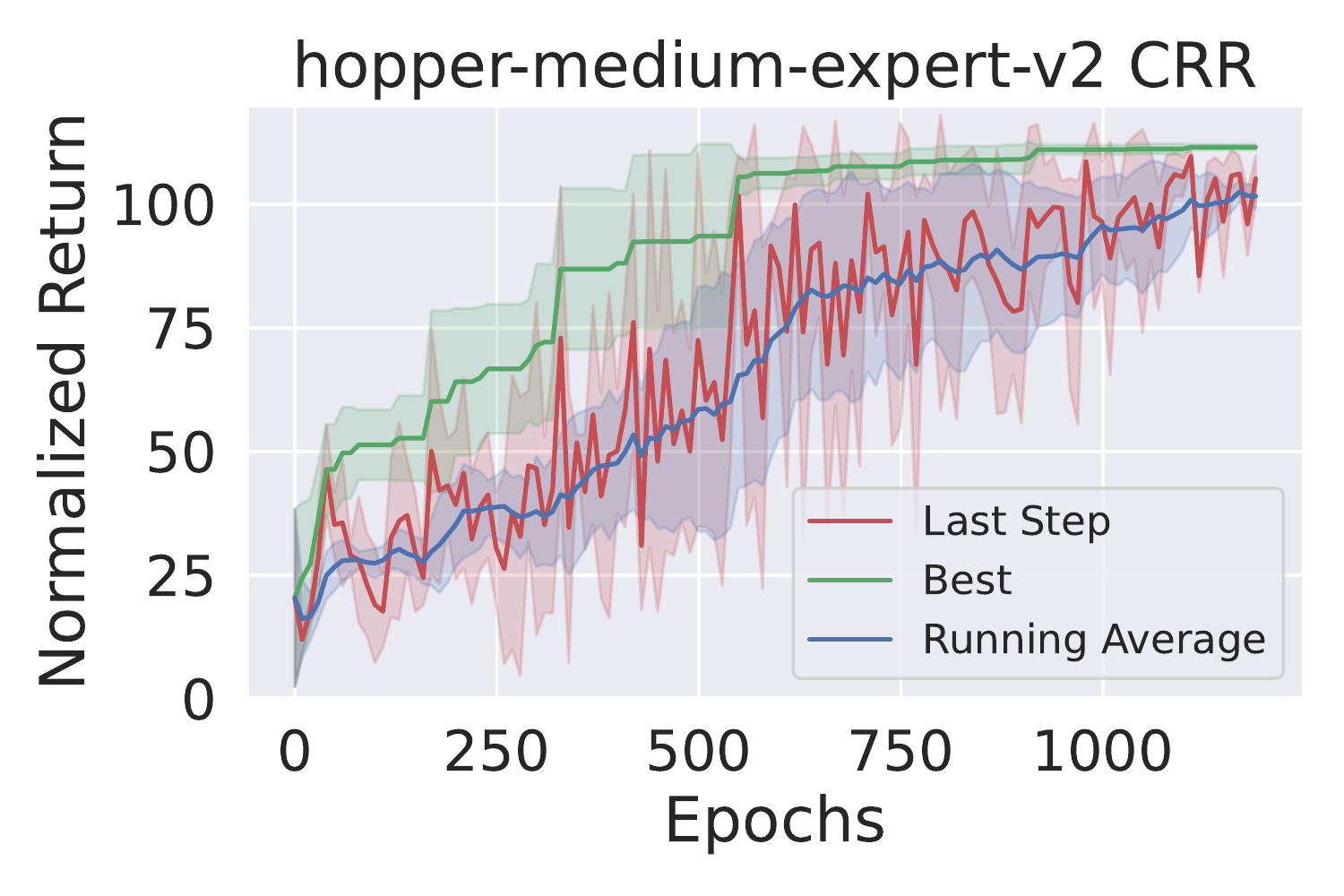}\\\includegraphics[width=0.225\linewidth]{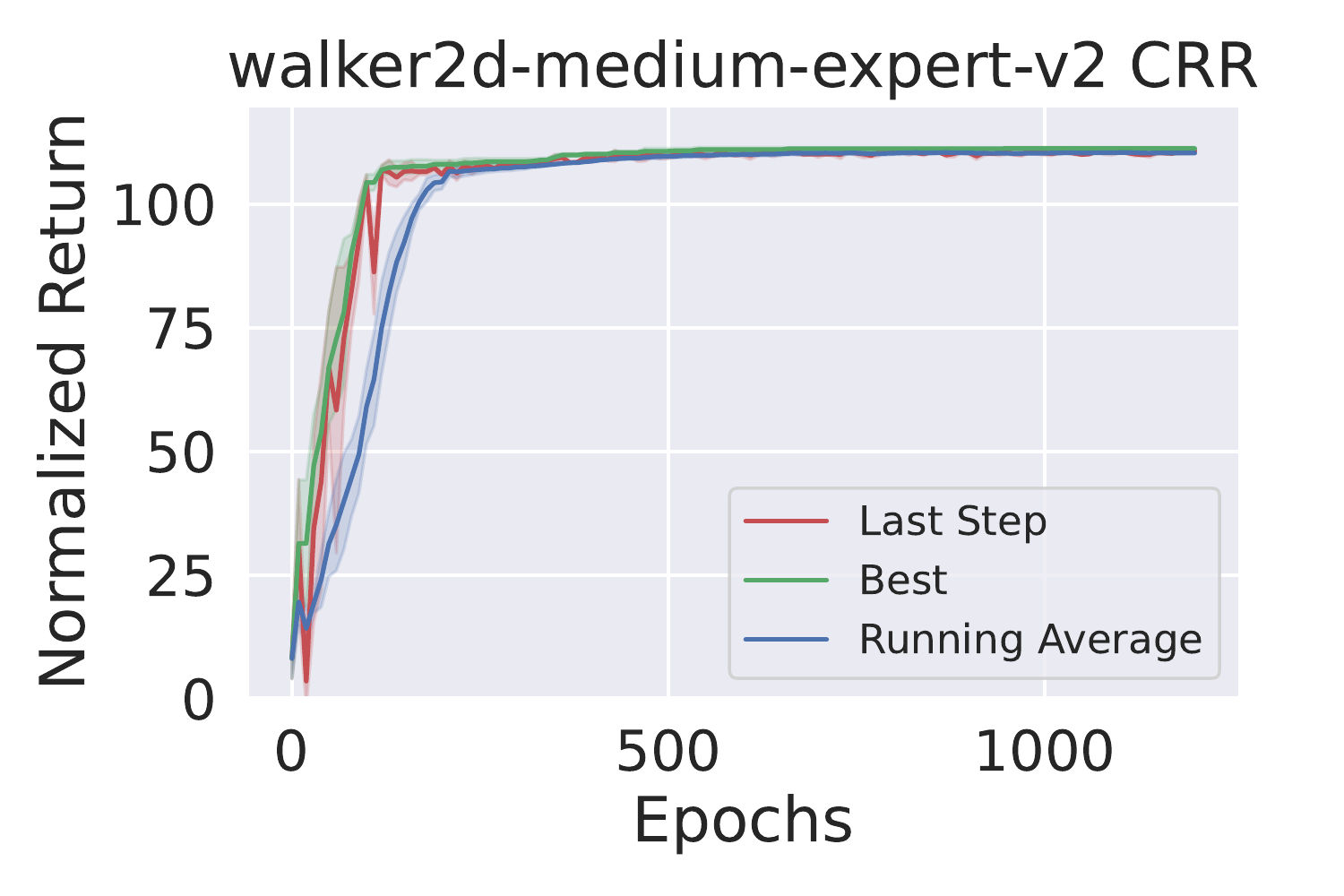}&\includegraphics[width=0.225\linewidth]{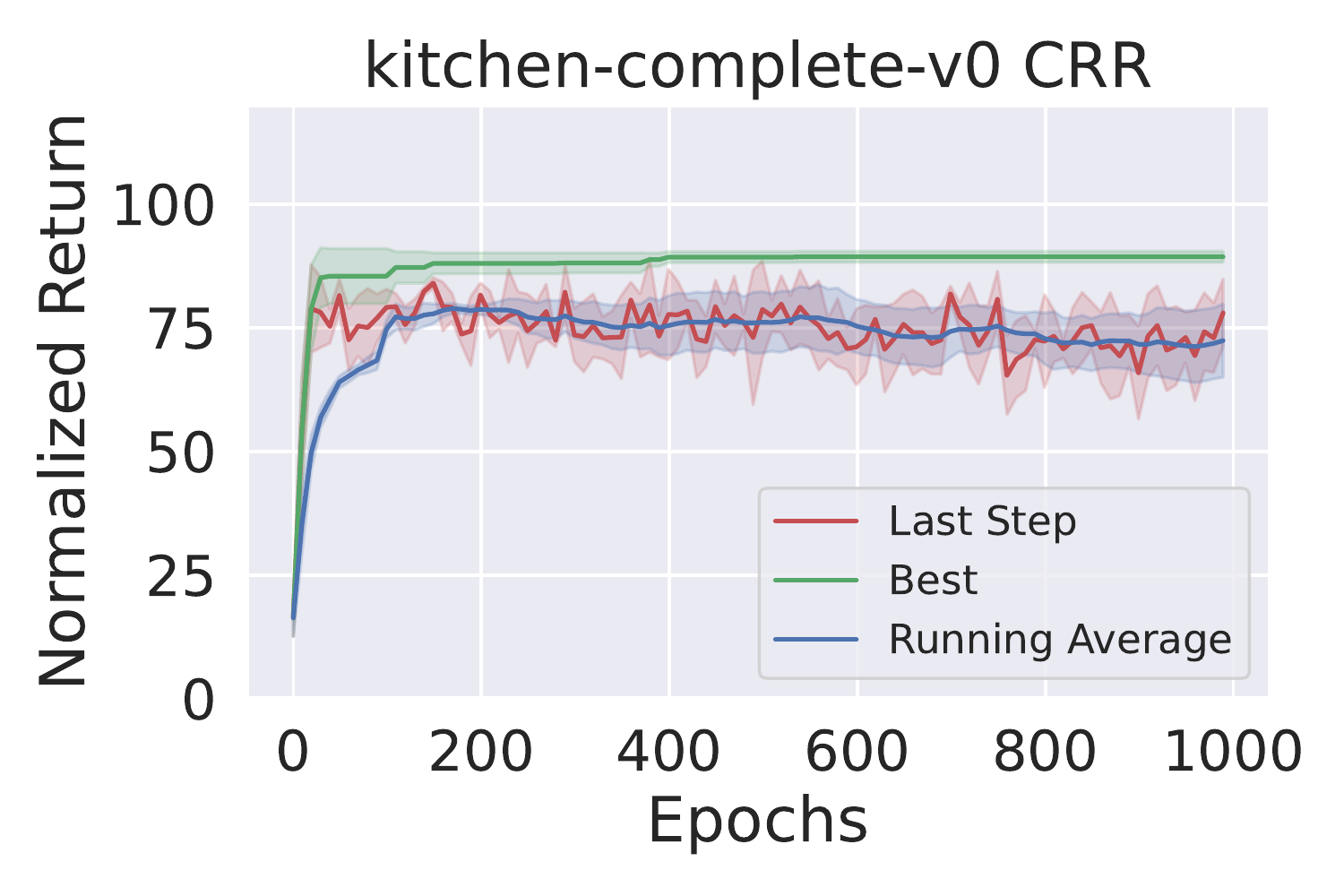}&\includegraphics[width=0.225\linewidth]{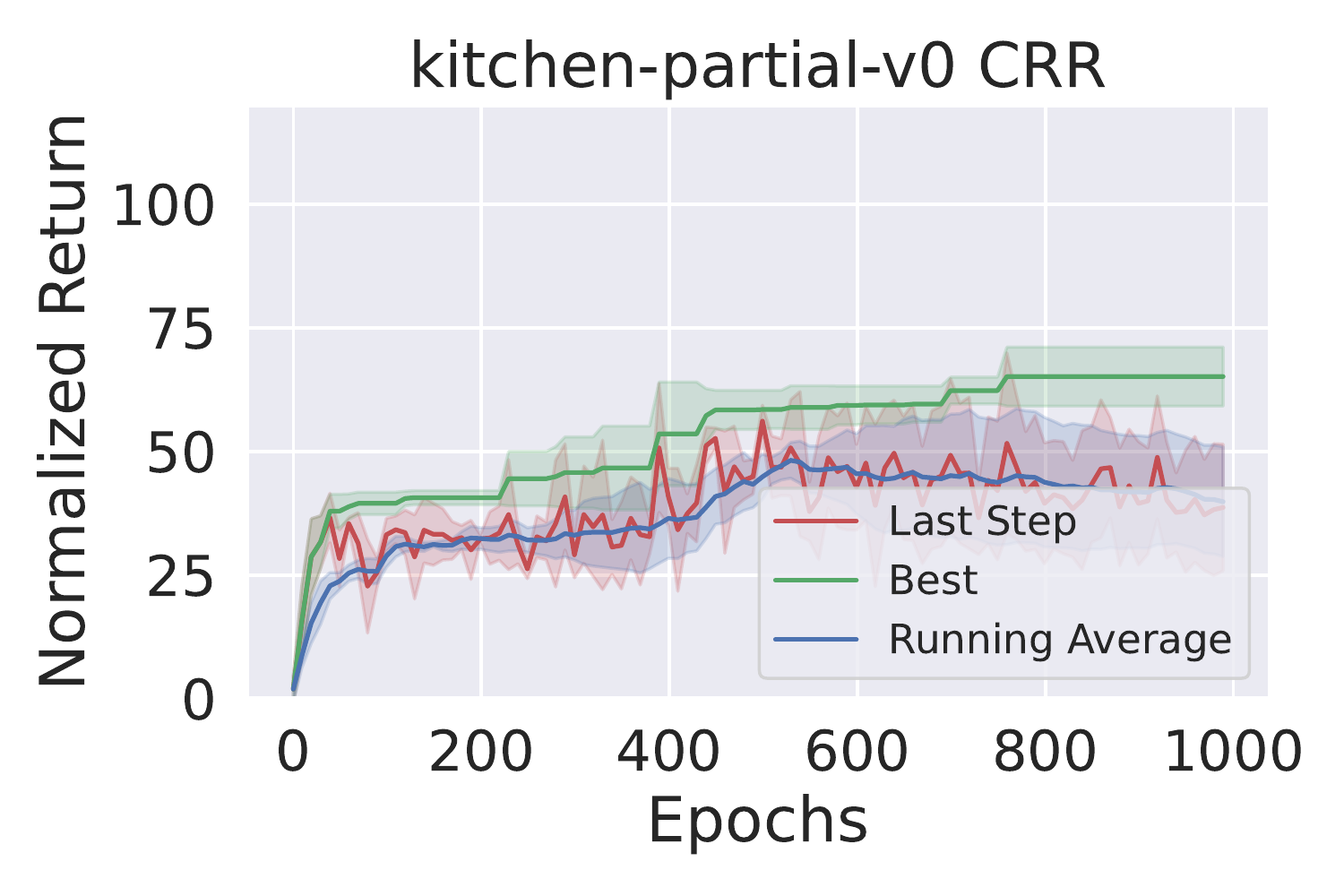}&\includegraphics[width=0.225\linewidth]{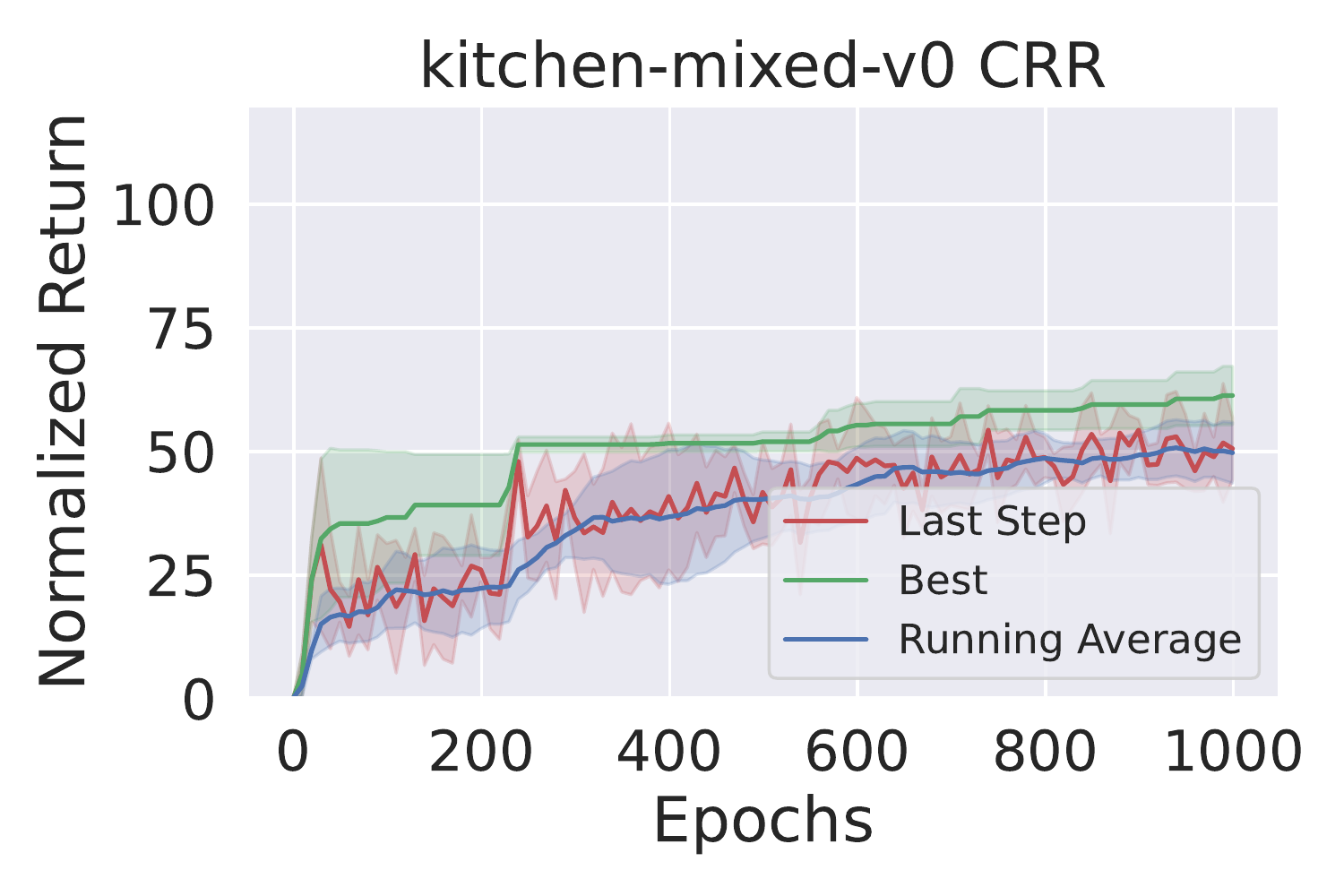}\\\includegraphics[width=0.225\linewidth]{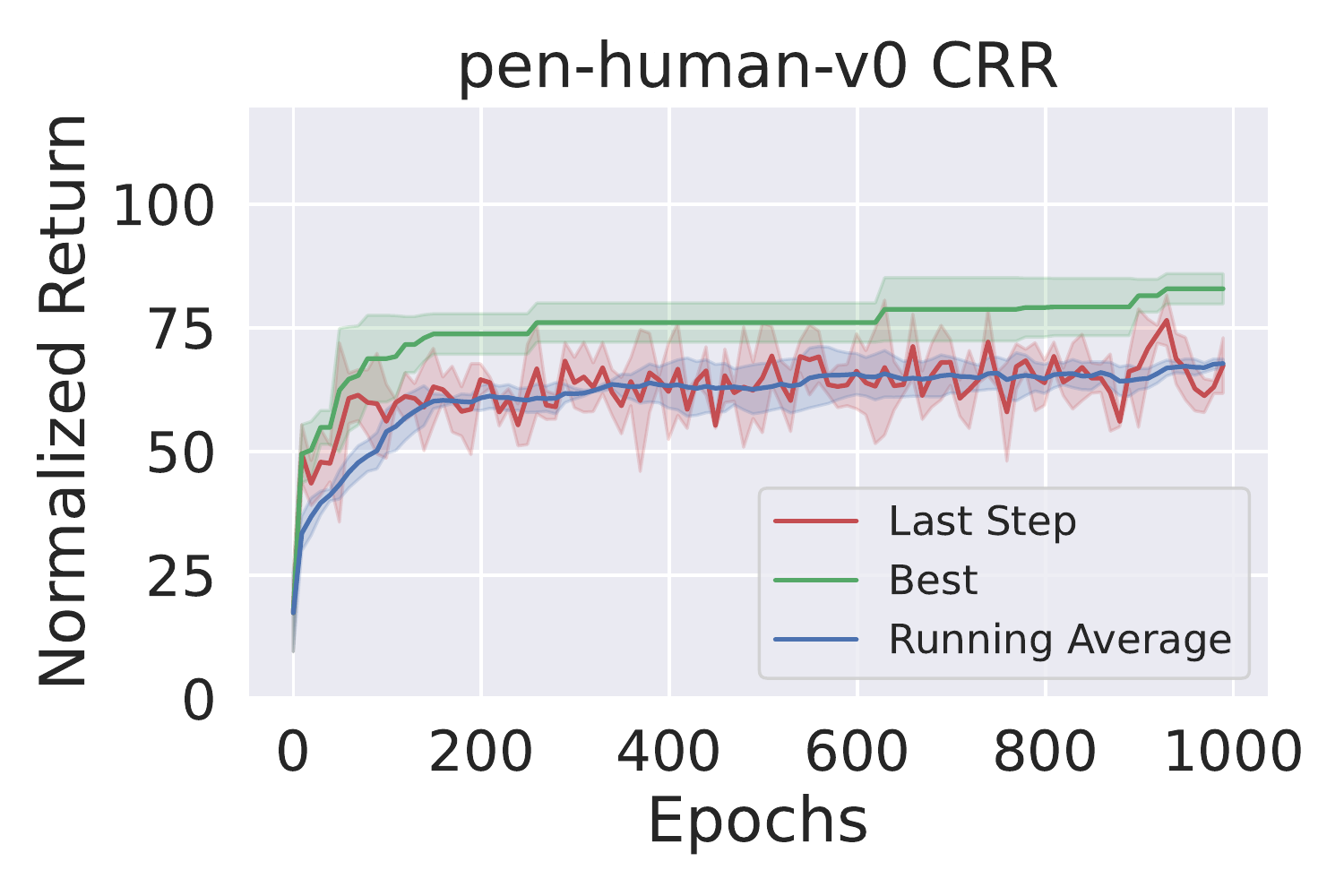}&\includegraphics[width=0.225\linewidth]{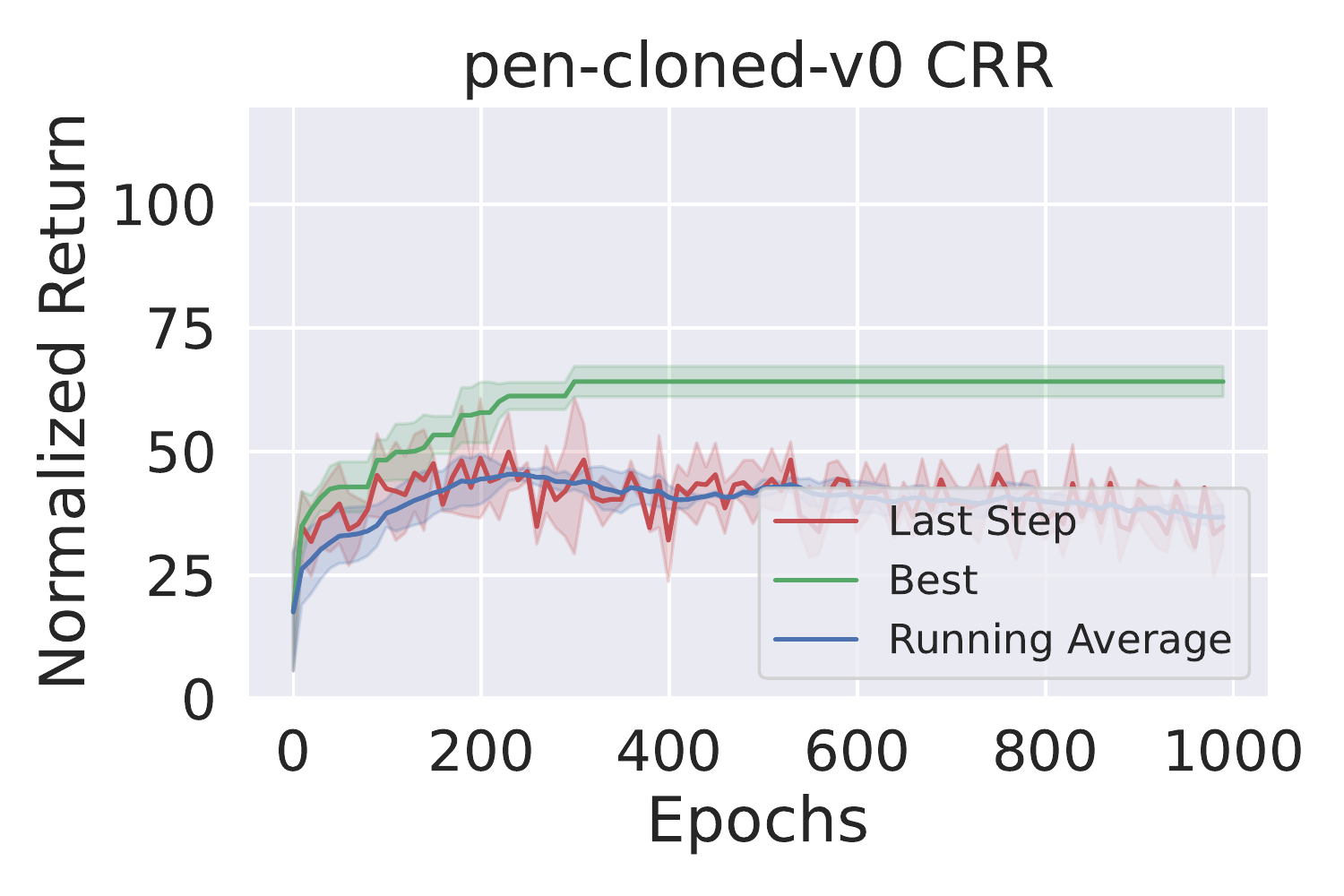}&\includegraphics[width=0.225\linewidth]{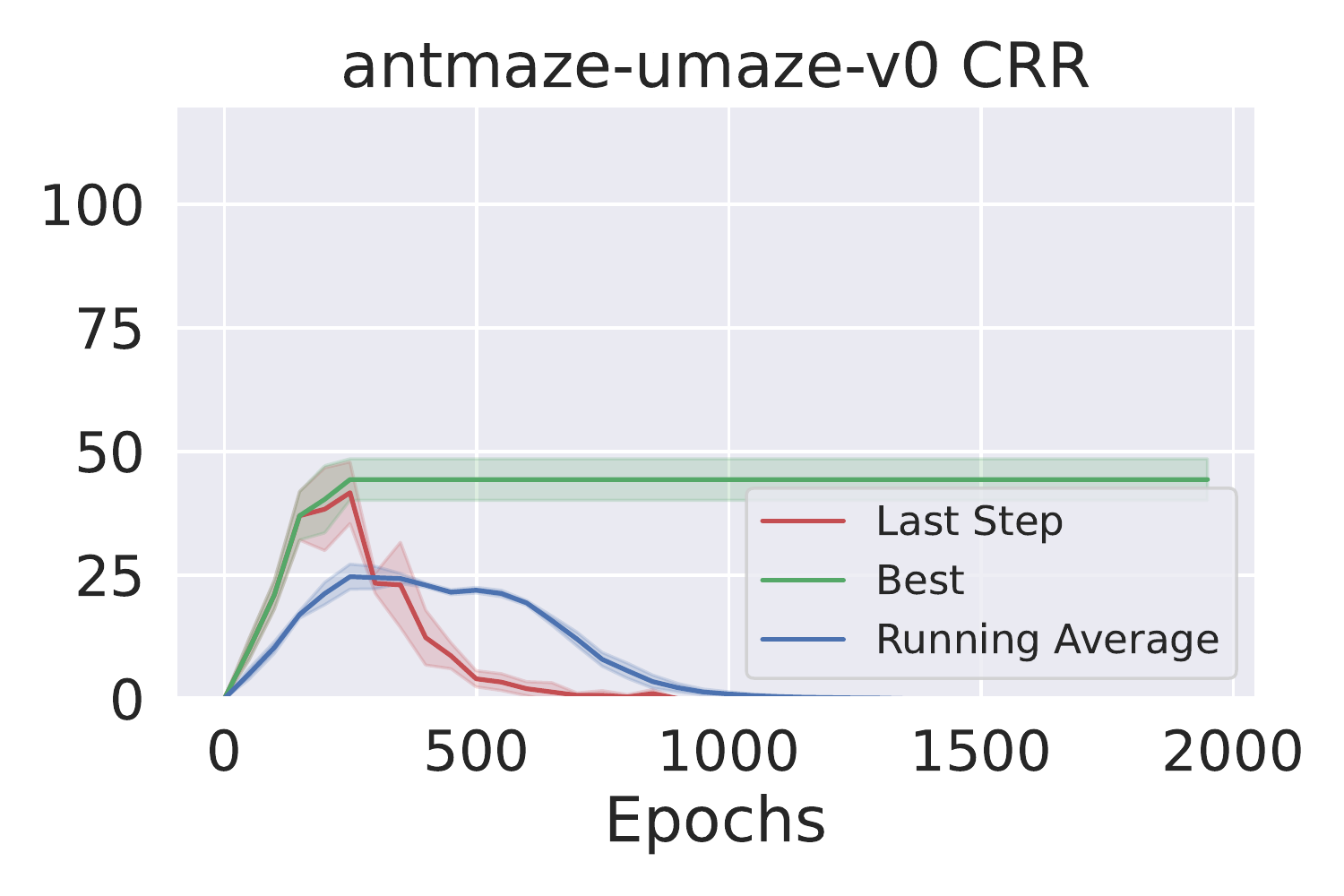}&\includegraphics[width=0.225\linewidth]{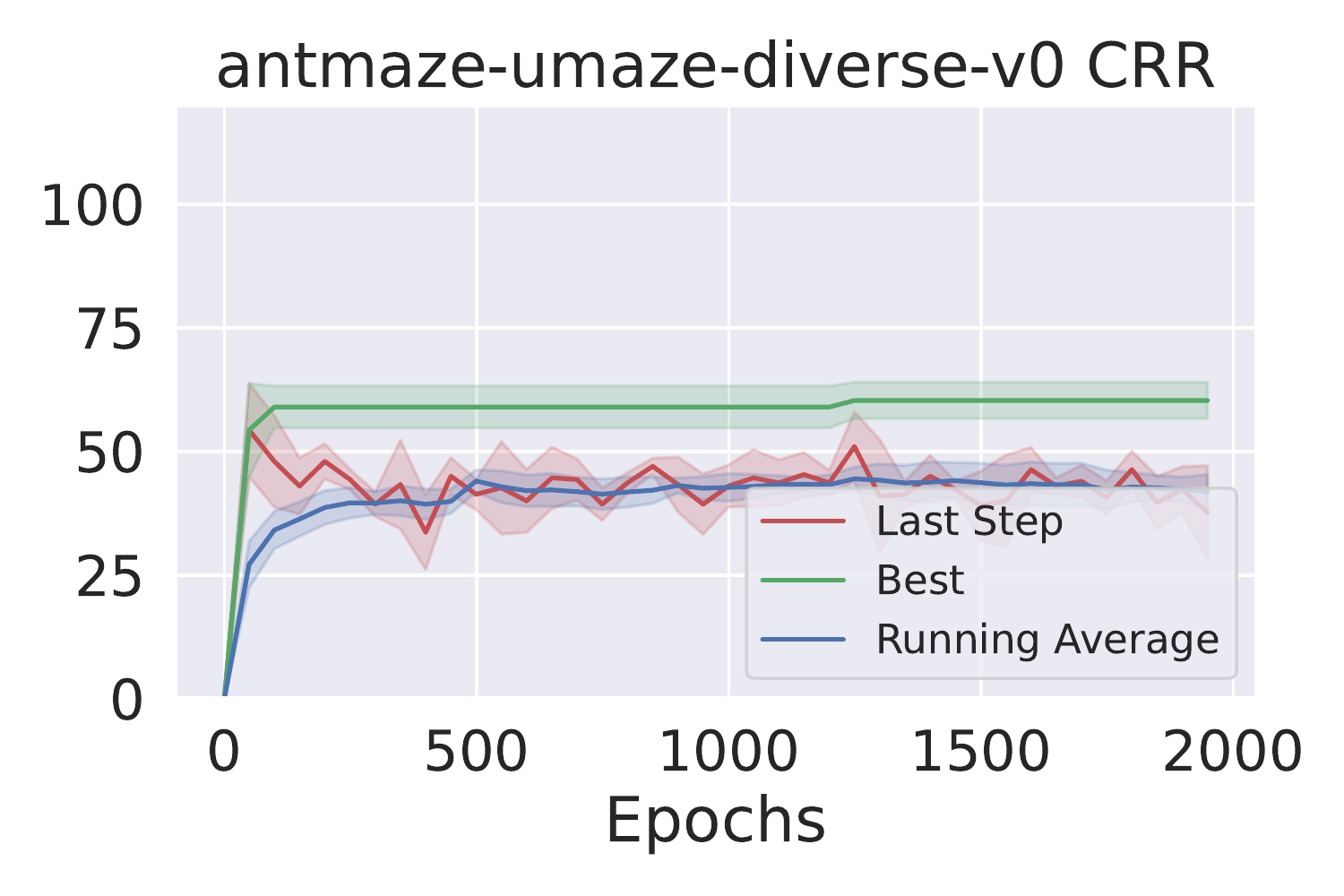}\\\includegraphics[width=0.225\linewidth]{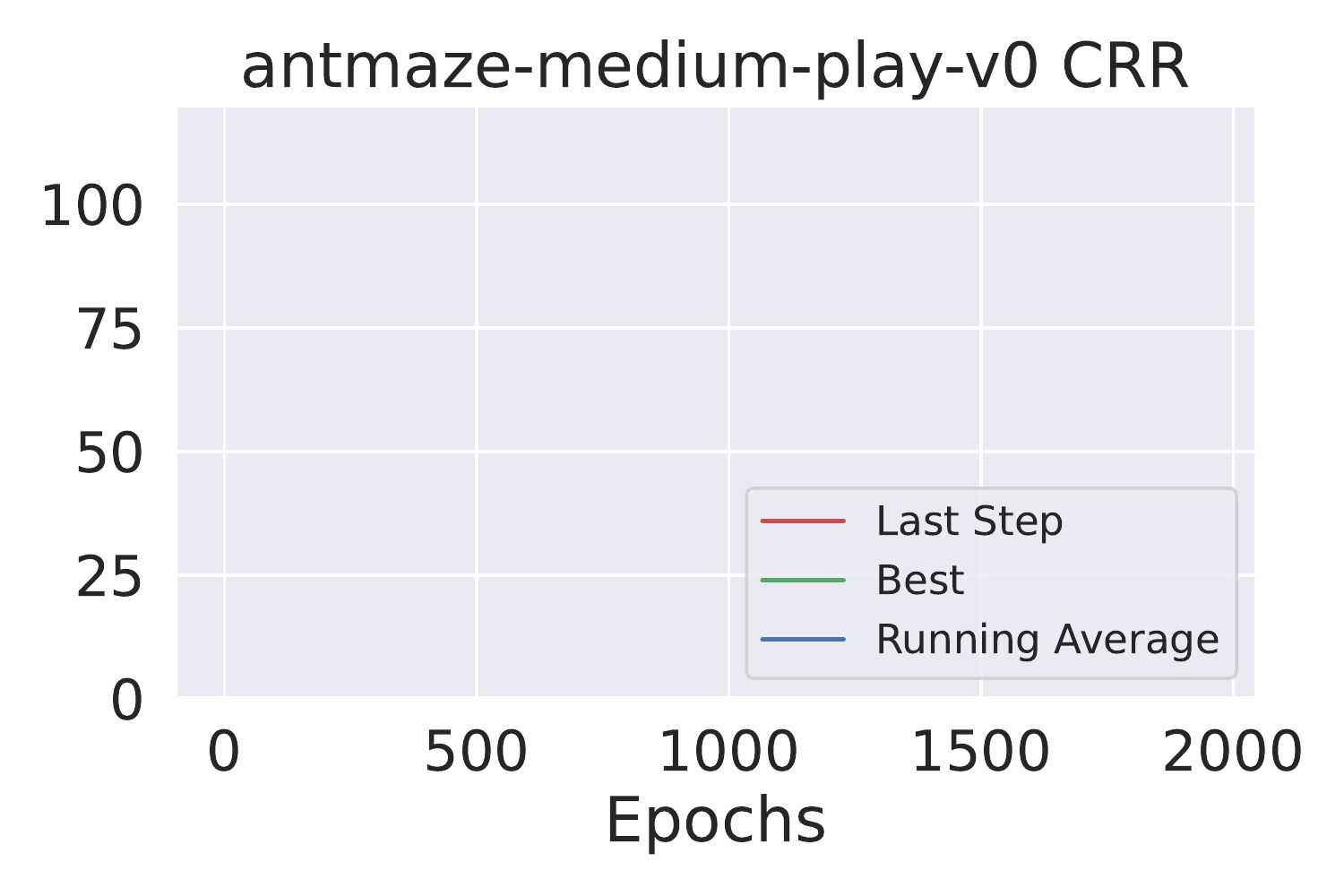}&\includegraphics[width=0.225\linewidth]{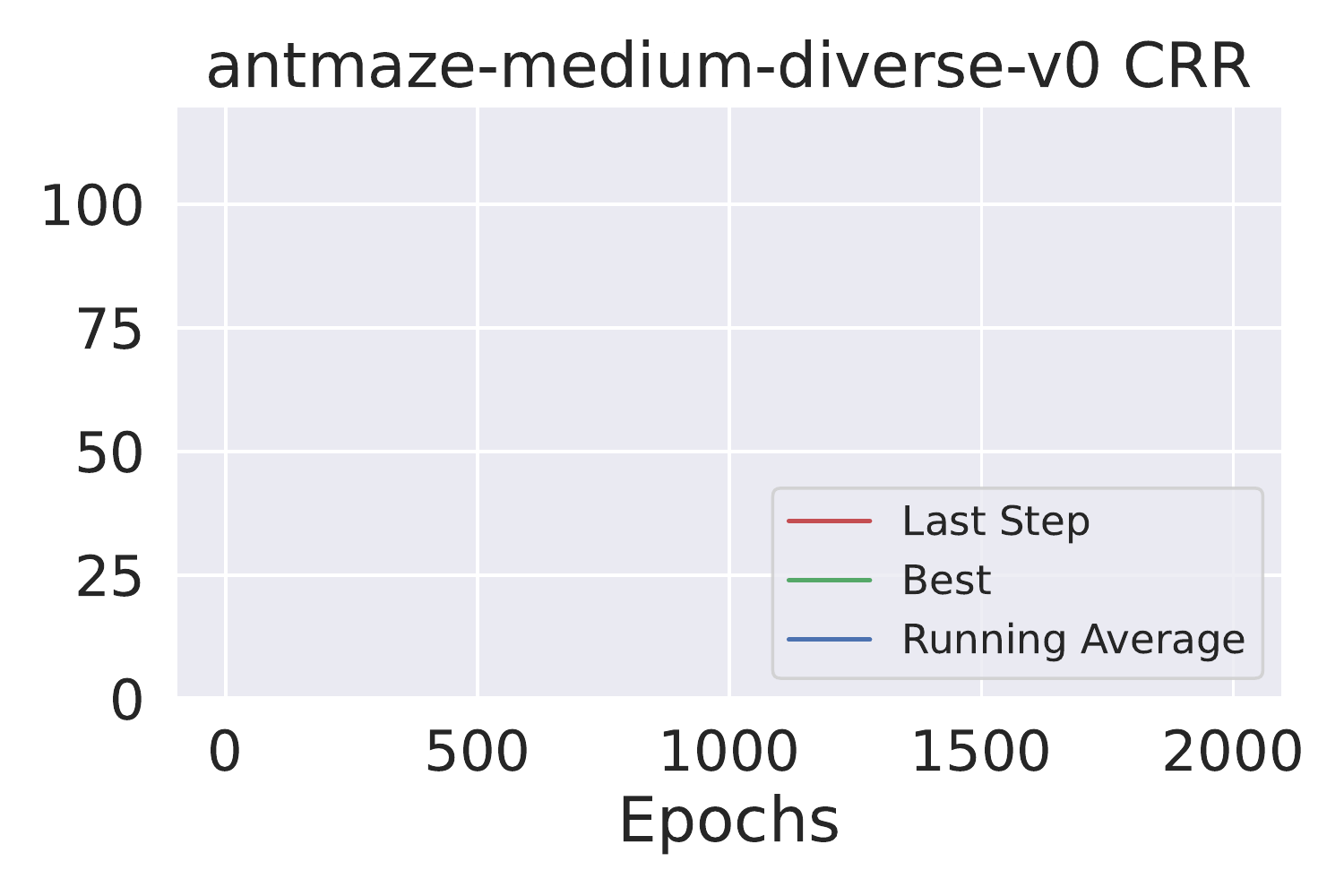}&\includegraphics[width=0.225\linewidth]{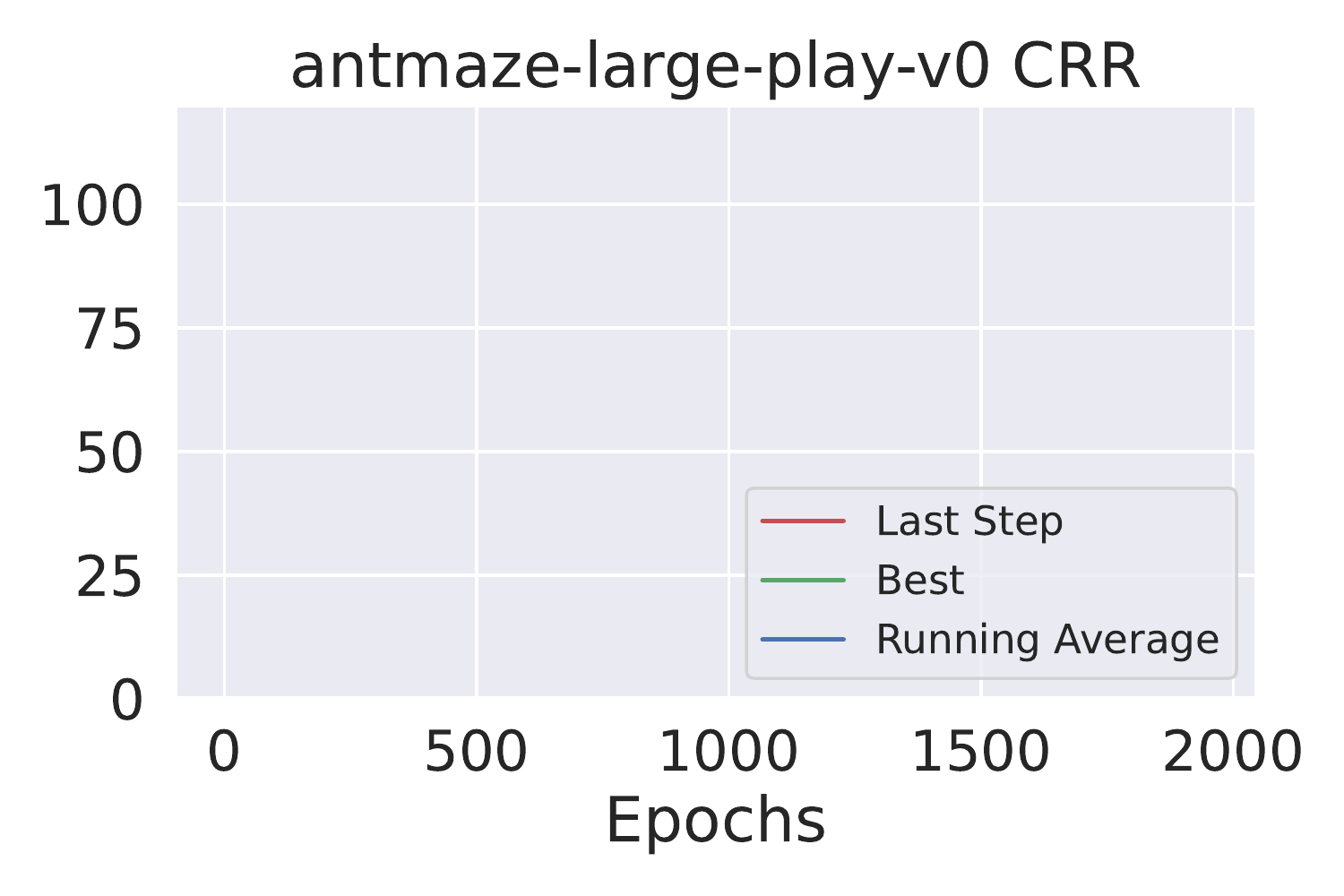}&\includegraphics[width=0.225\linewidth]{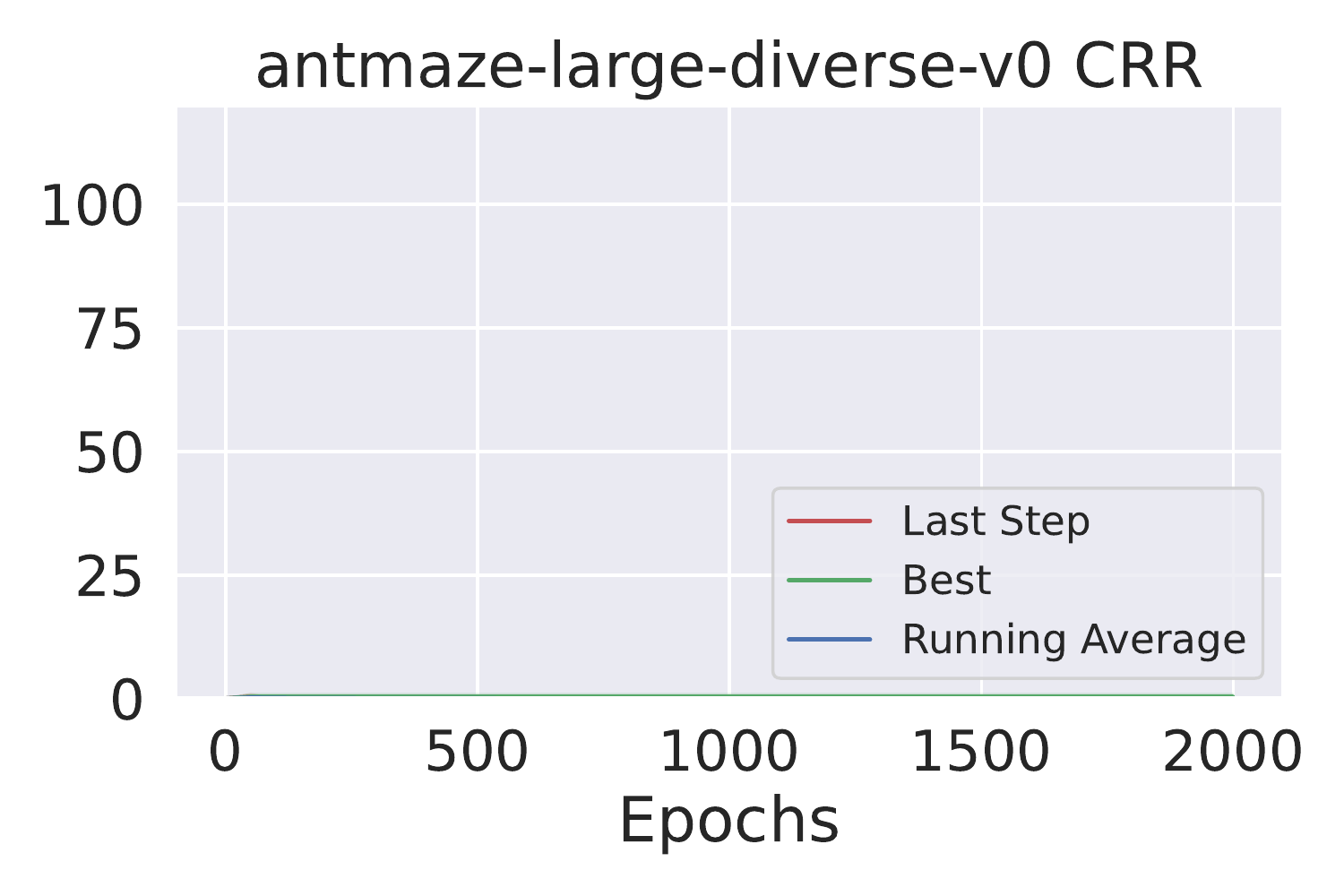}\\\end{tabular}
\centering
\caption{Training Curves of CRR on D4RL}
\end{figure*}
\begin{figure*}[htb]
\centering
\begin{tabular}{cccccc}
\includegraphics[width=0.225\linewidth]{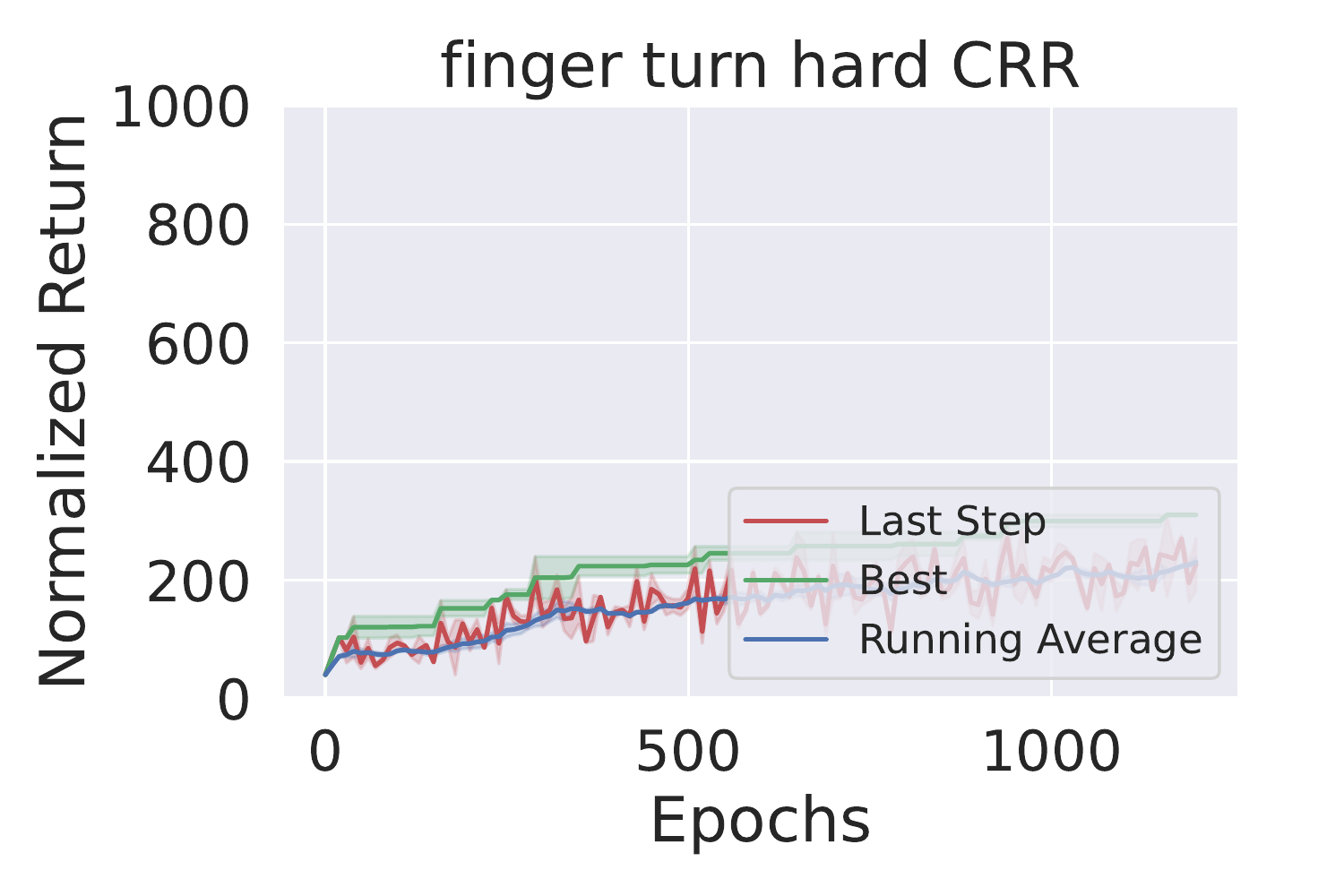}&\includegraphics[width=0.225\linewidth]{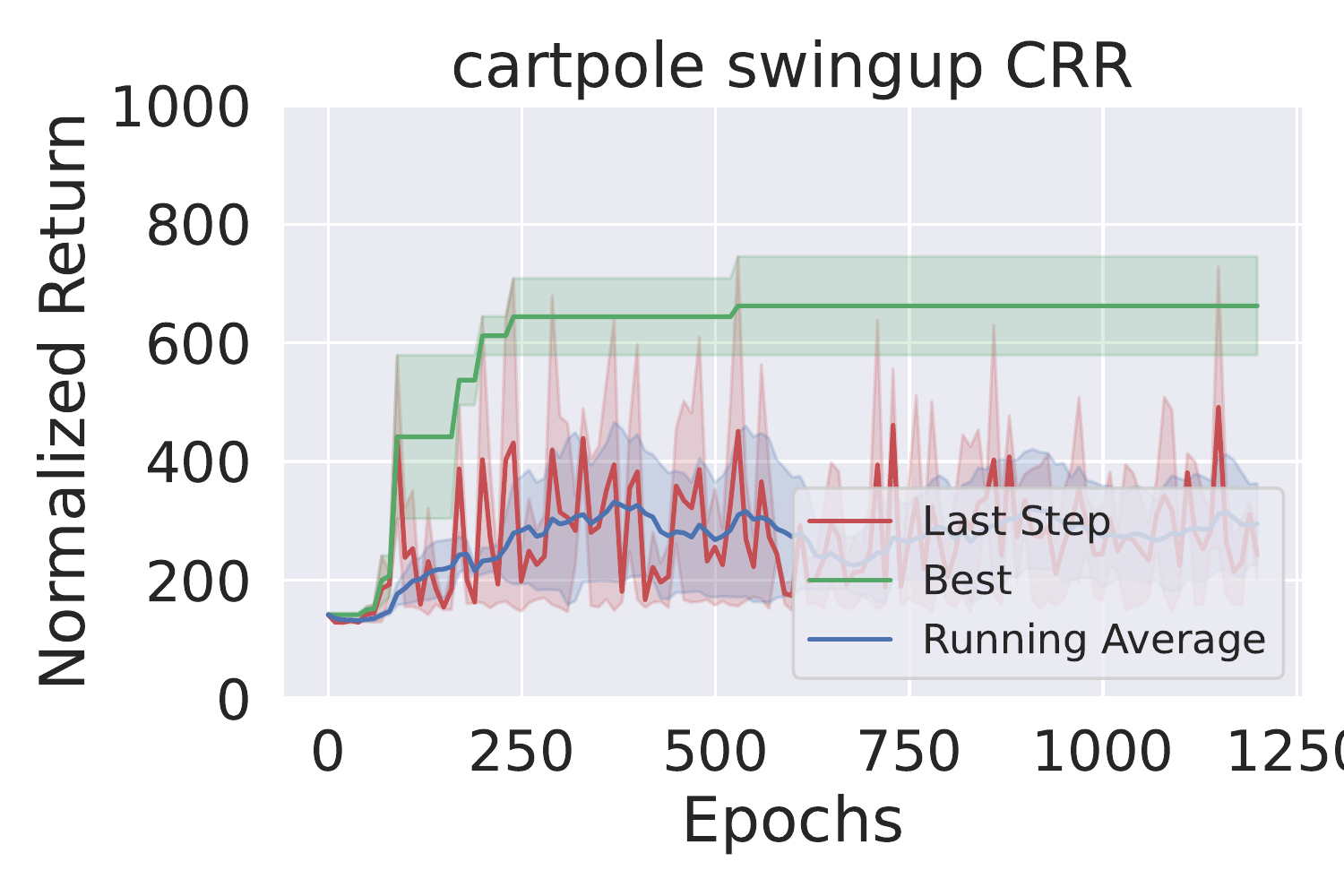}&\includegraphics[width=0.225\linewidth]{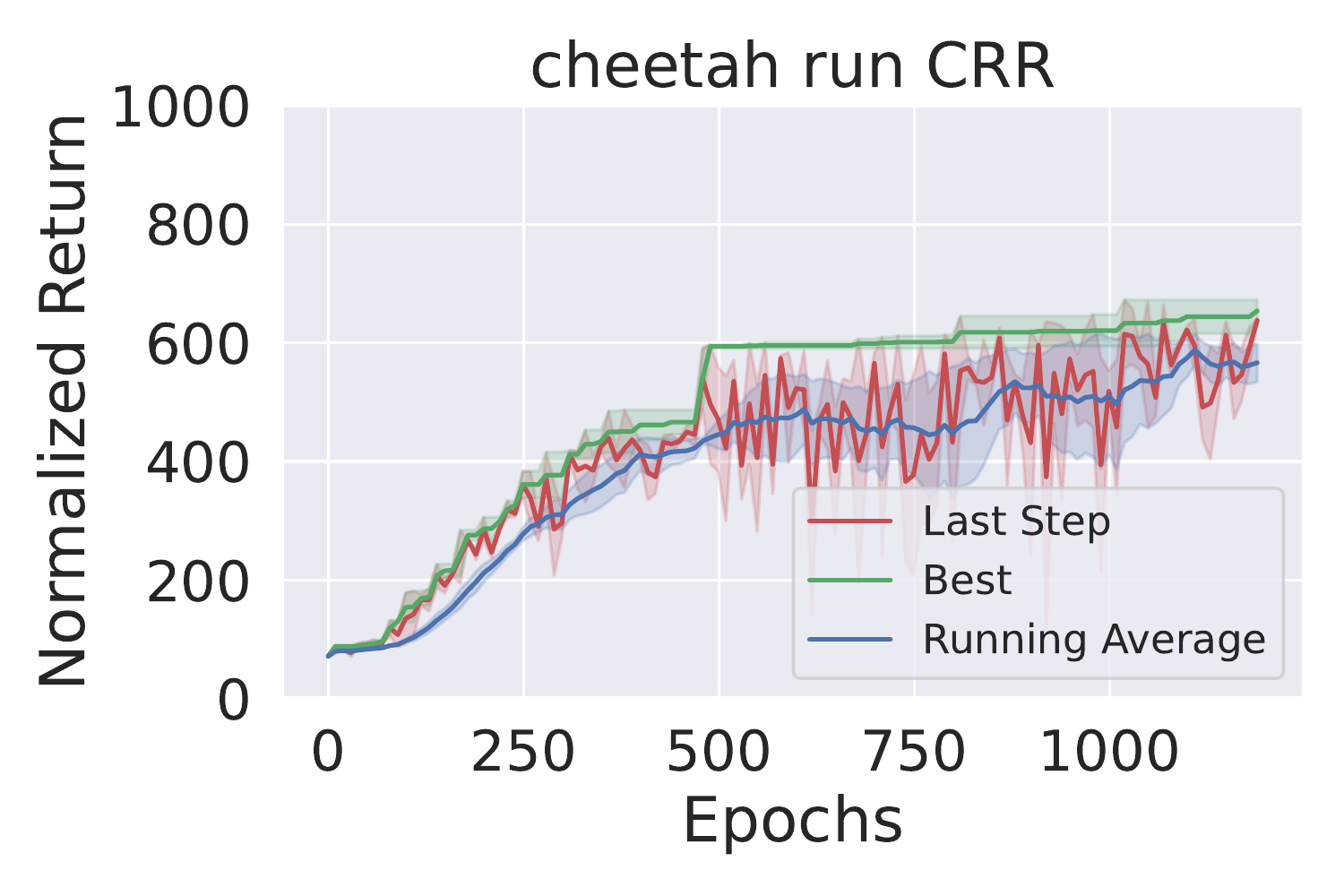}&\includegraphics[width=0.225\linewidth]{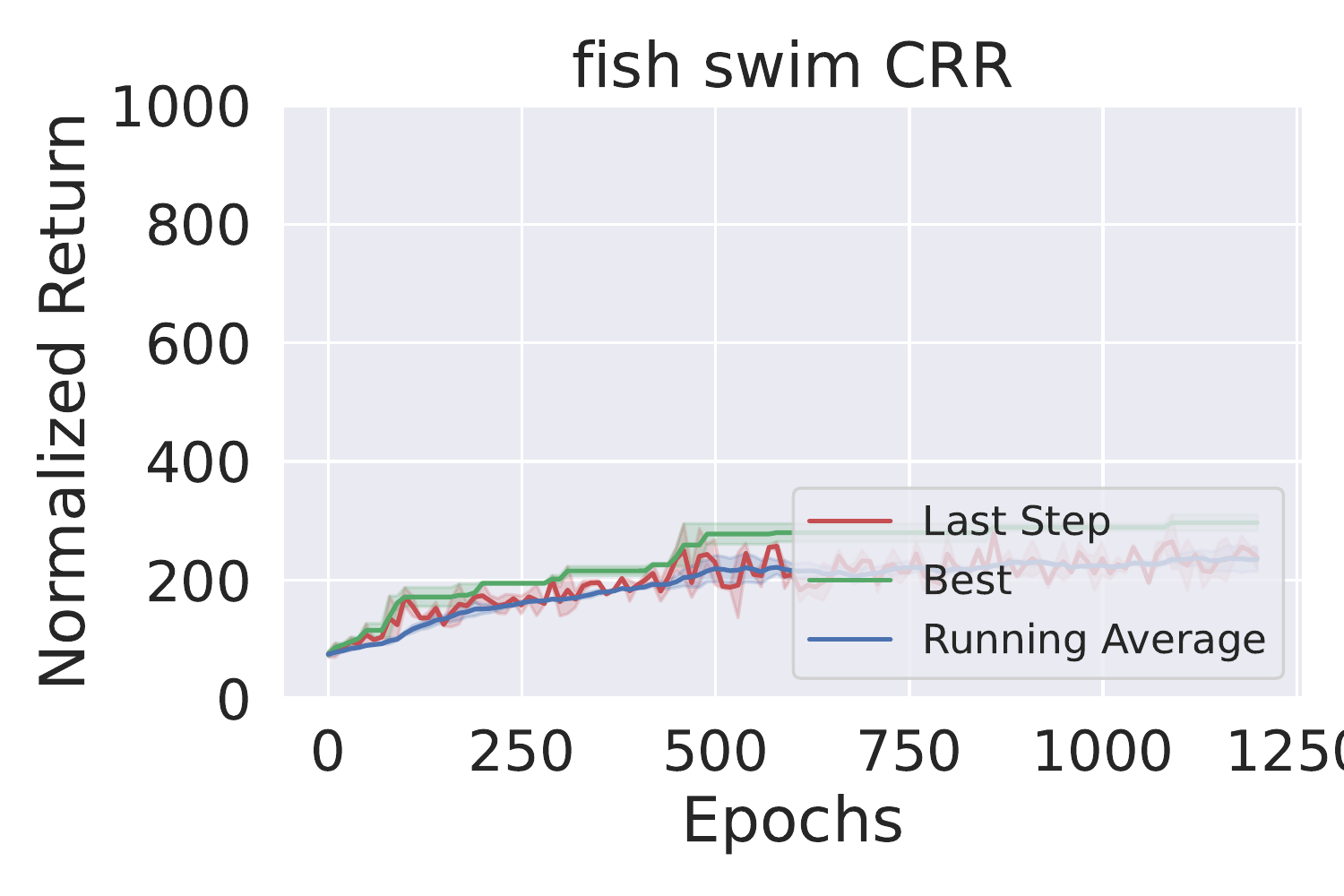}\\\includegraphics[width=0.225\linewidth]{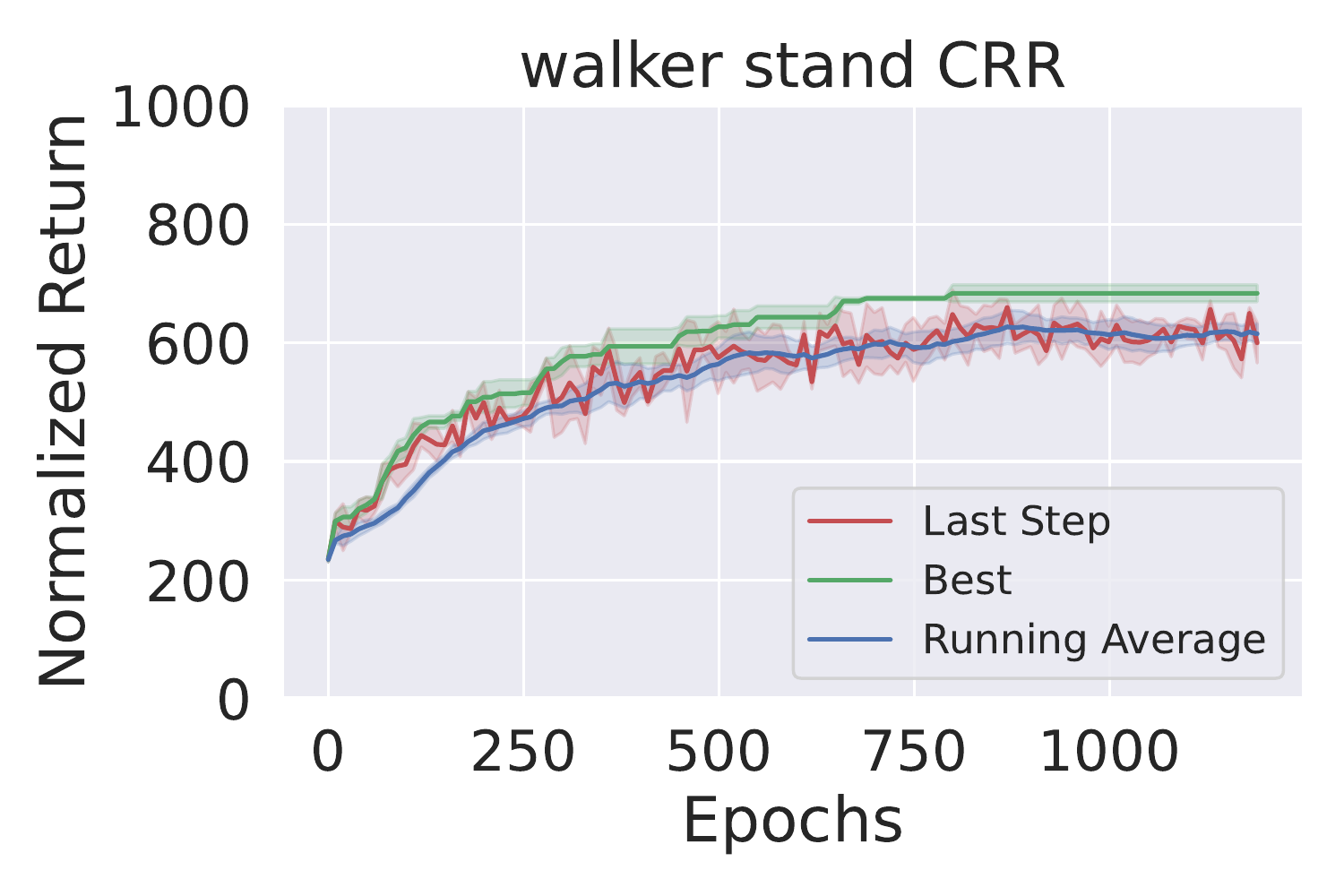}&\includegraphics[width=0.225\linewidth]{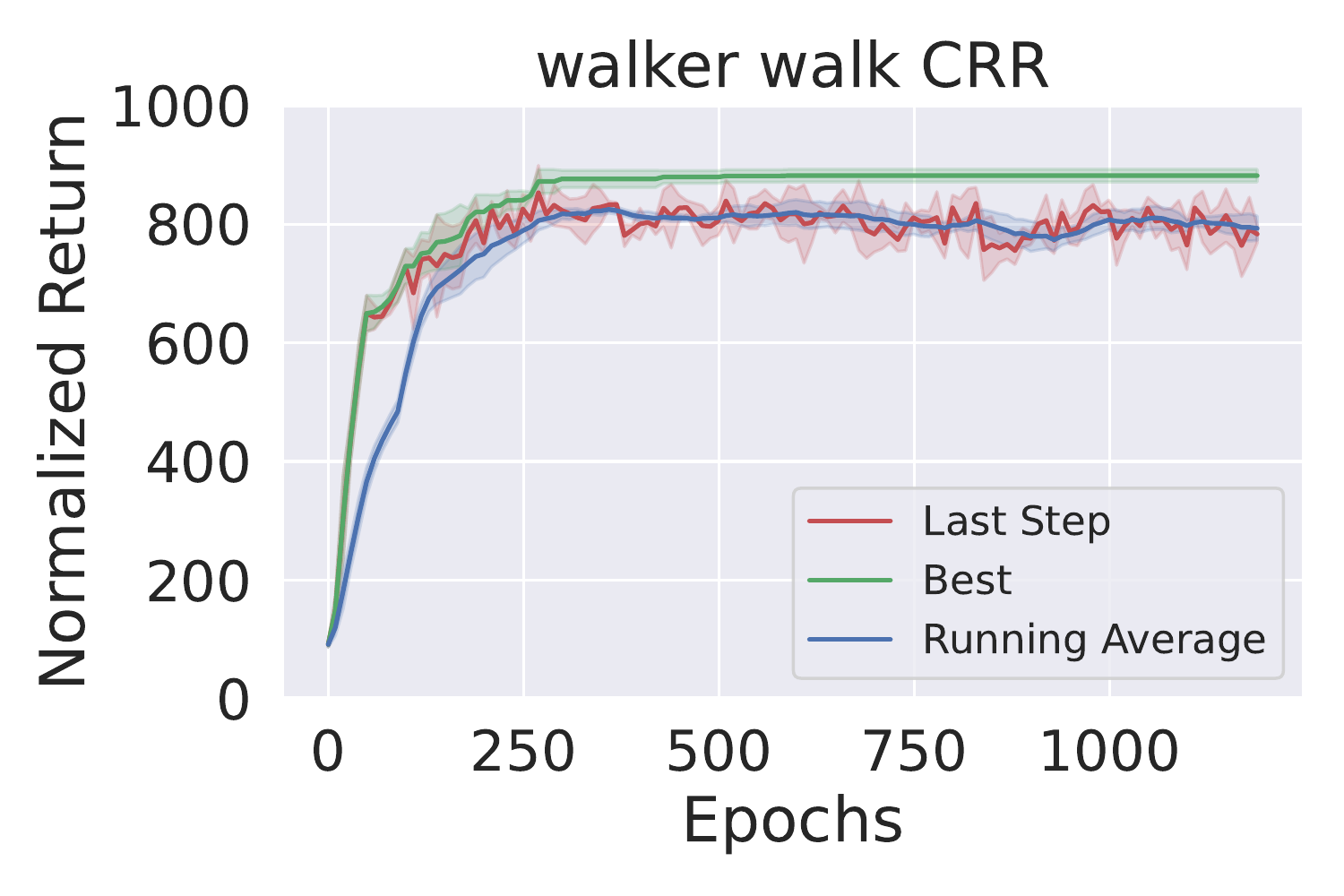}&&\\\end{tabular}
\centering
\caption{Training Curves of CRR on RLUP}
\end{figure*}
\begin{figure*}[htb]
\centering
\begin{tabular}{cccccc}
\includegraphics[width=0.225\linewidth]{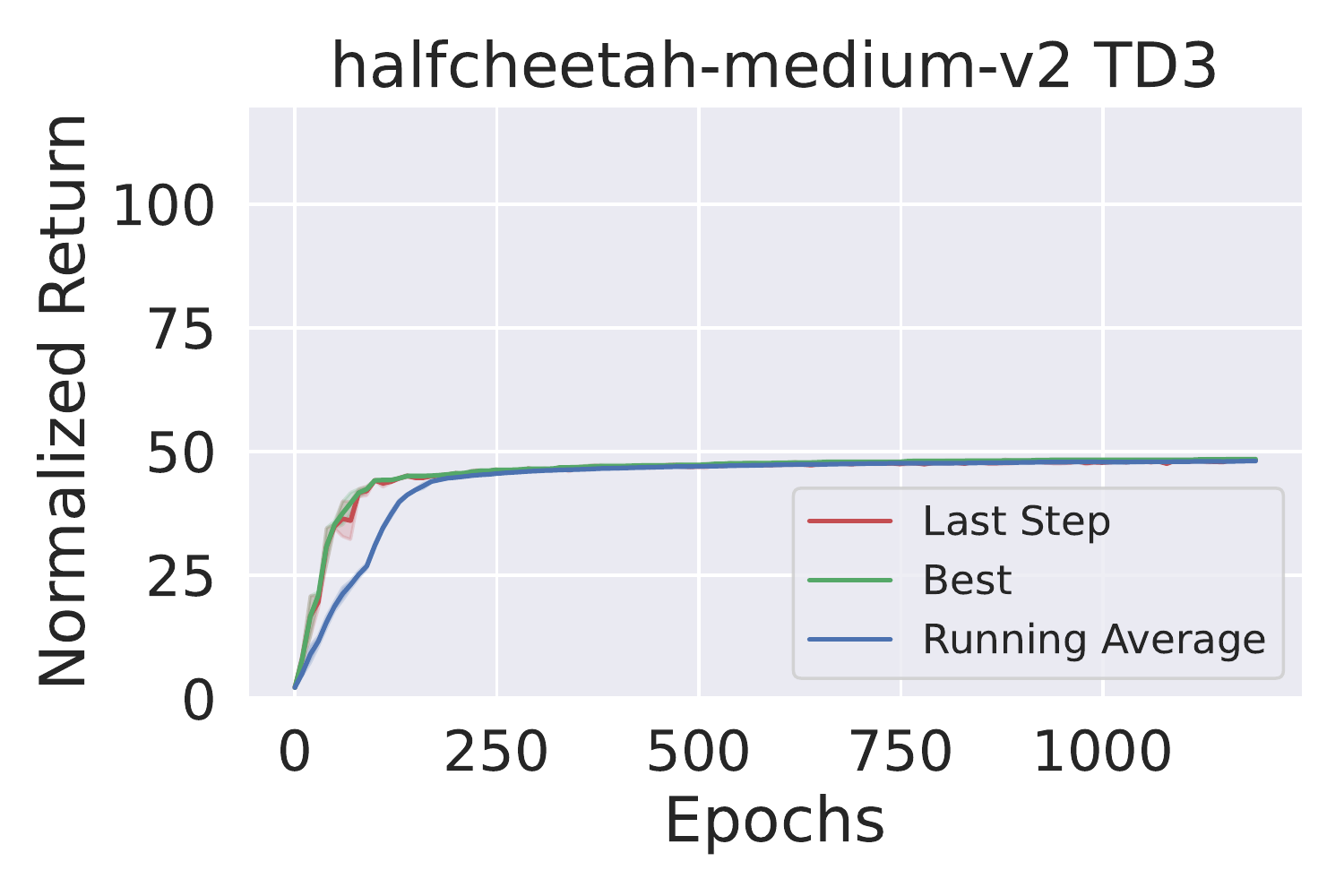}&\includegraphics[width=0.225\linewidth]{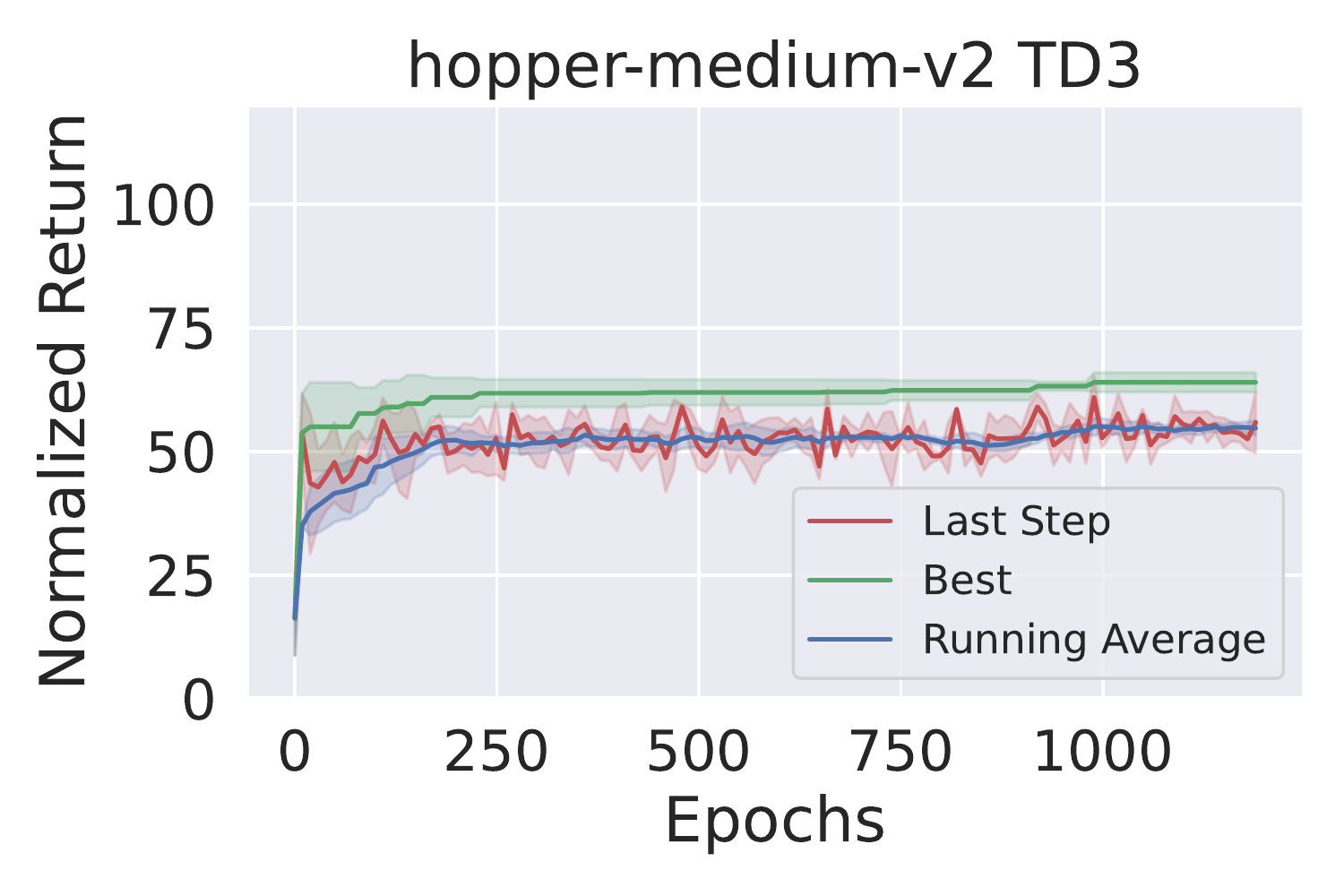}&\includegraphics[width=0.225\linewidth]{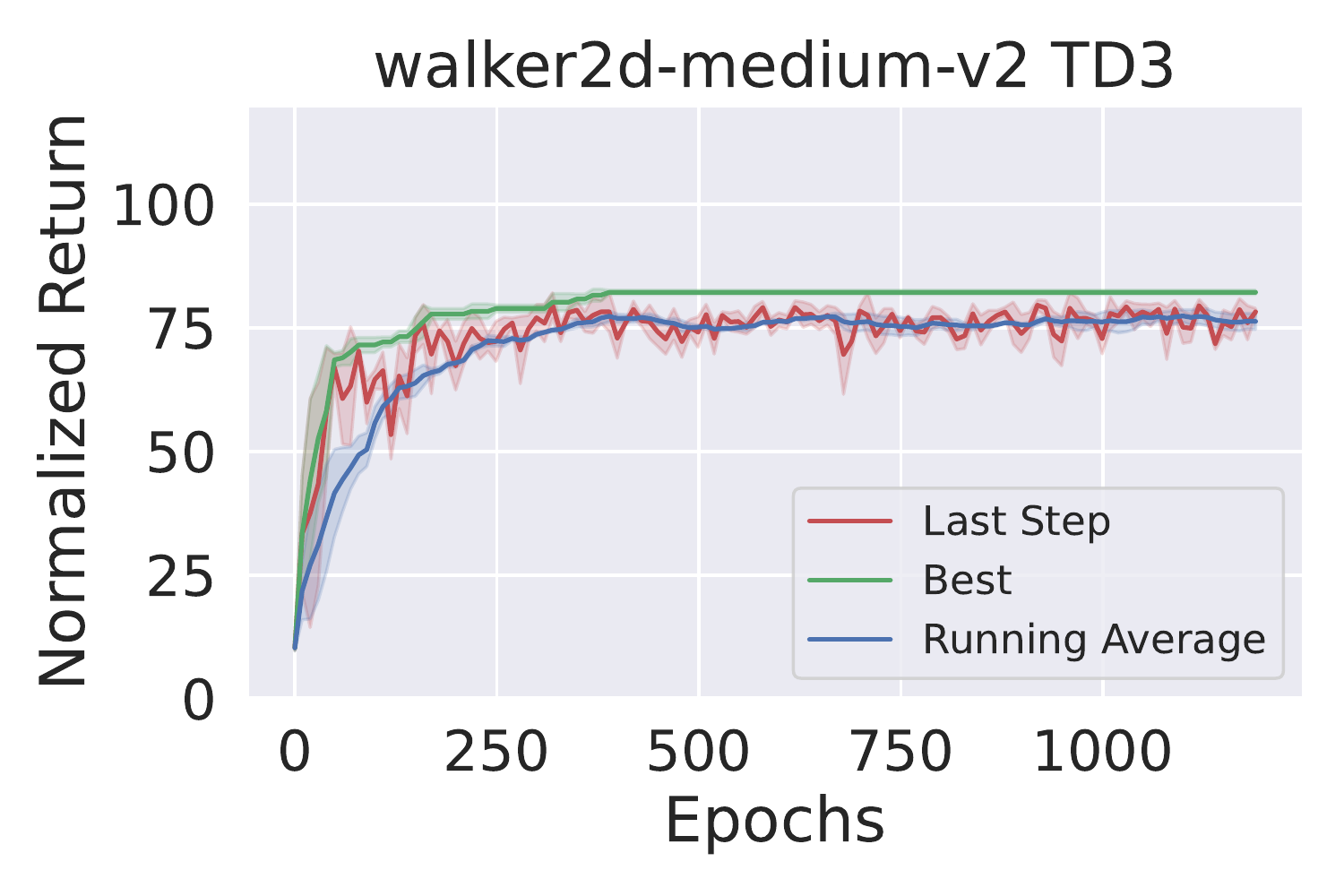}&\includegraphics[width=0.225\linewidth]{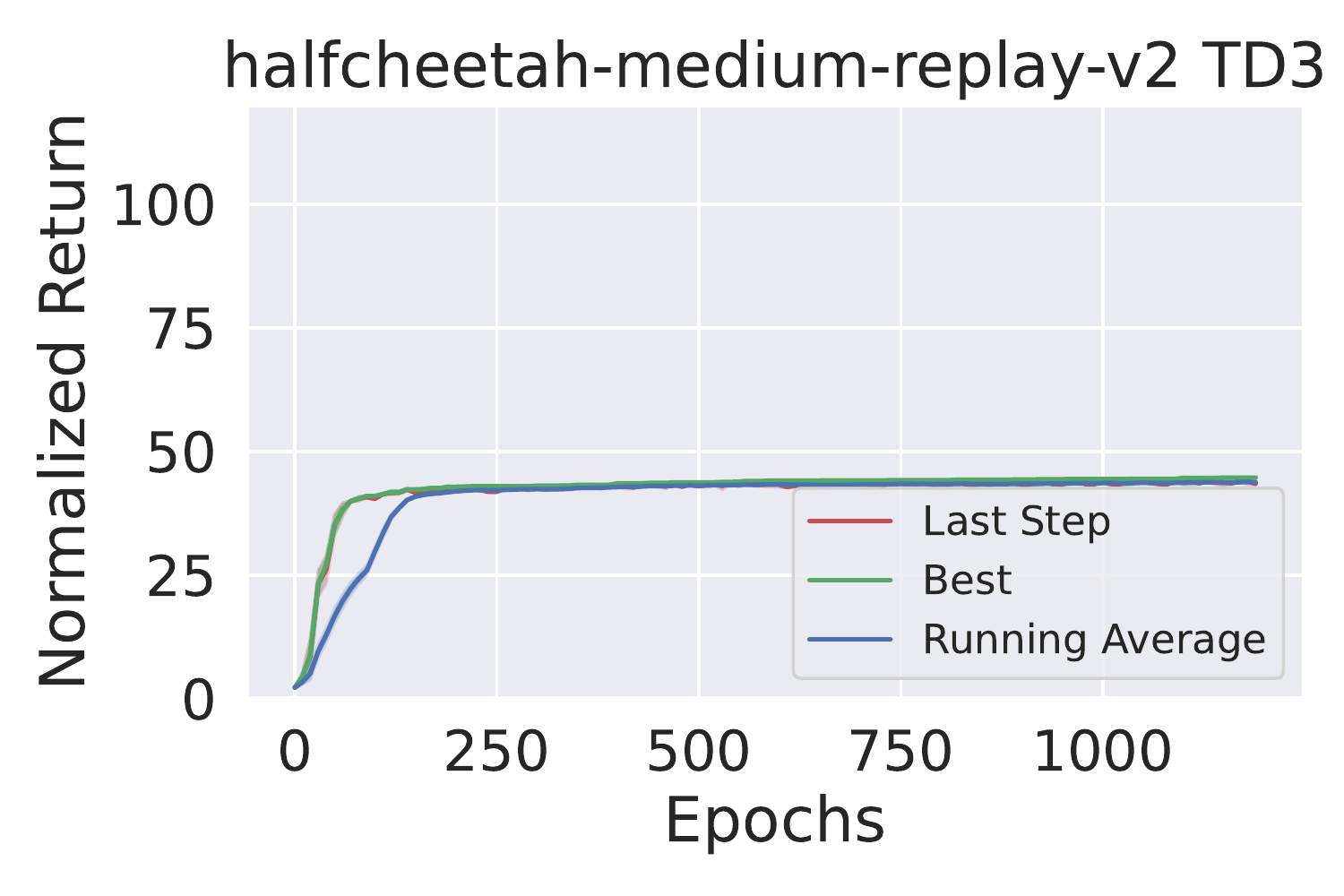}\\\includegraphics[width=0.225\linewidth]{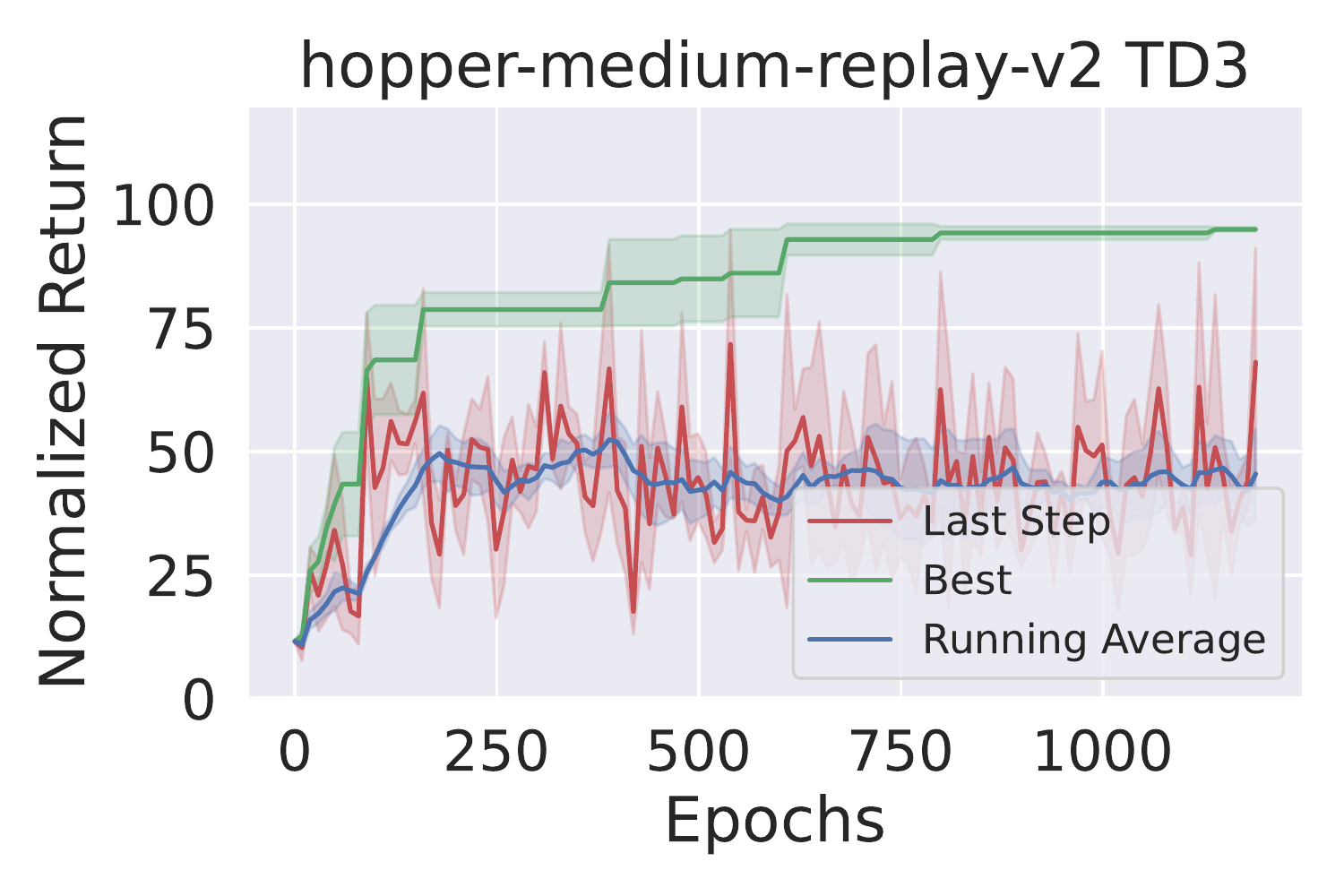}&\includegraphics[width=0.225\linewidth]{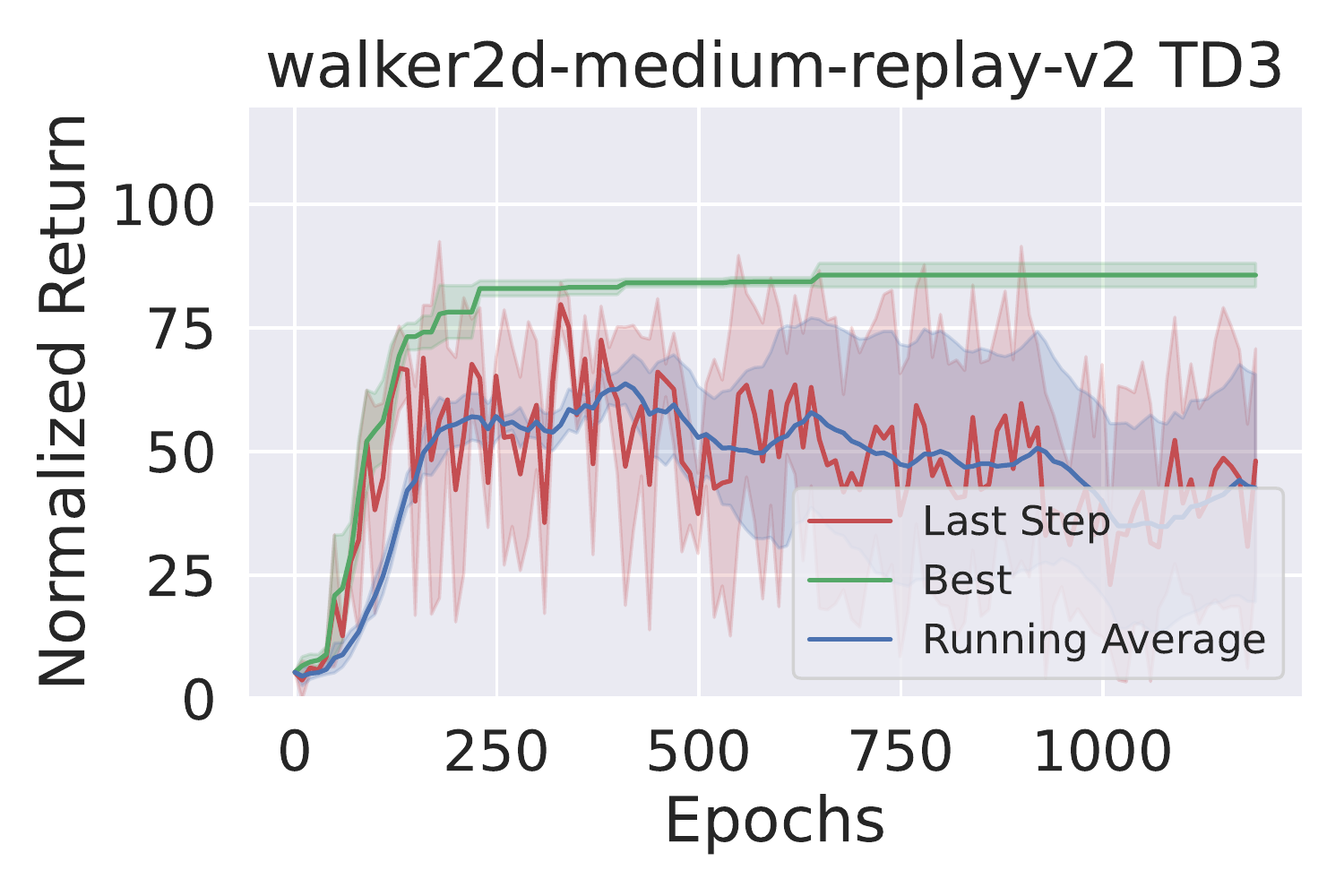}&\includegraphics[width=0.225\linewidth]{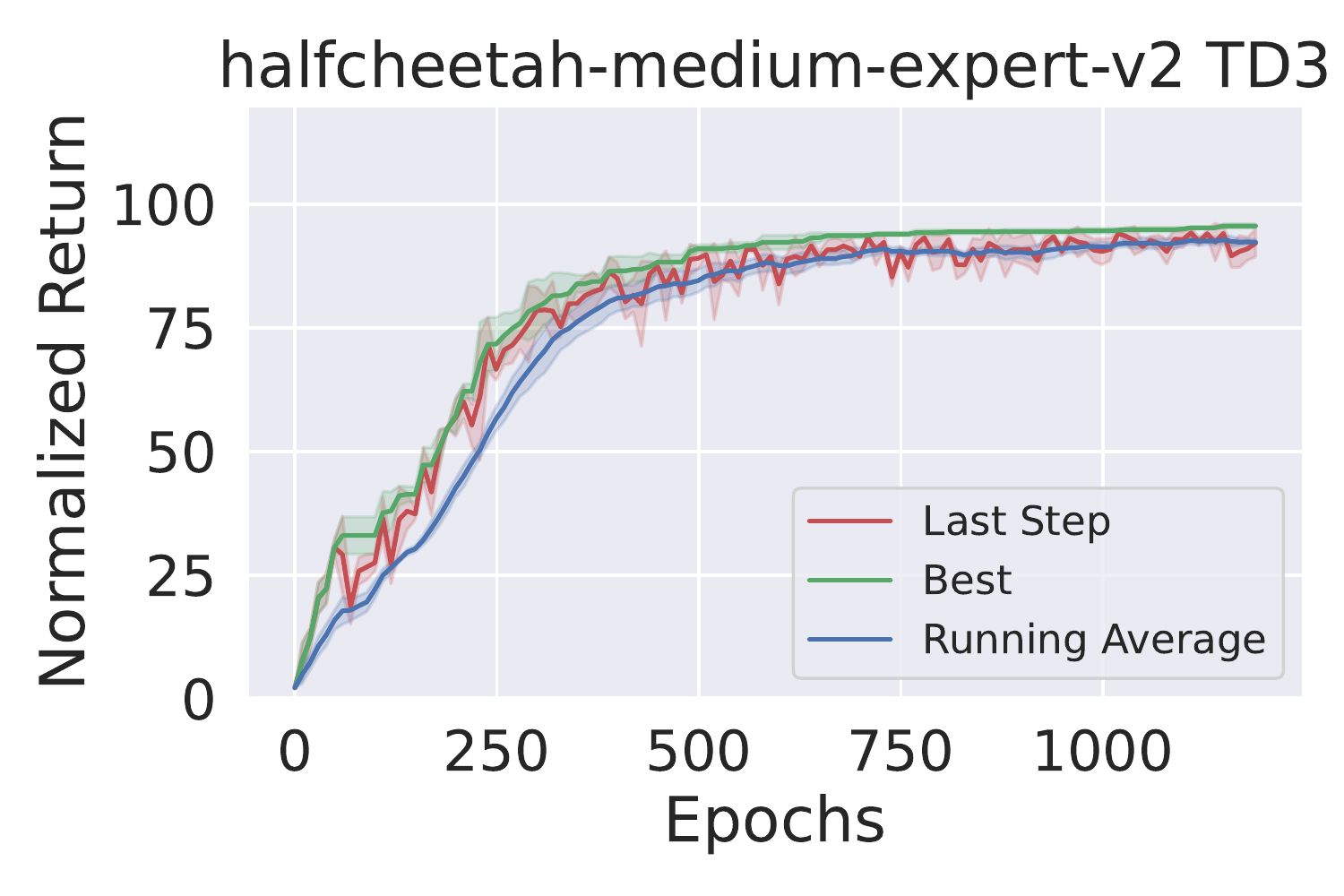}&\includegraphics[width=0.225\linewidth]{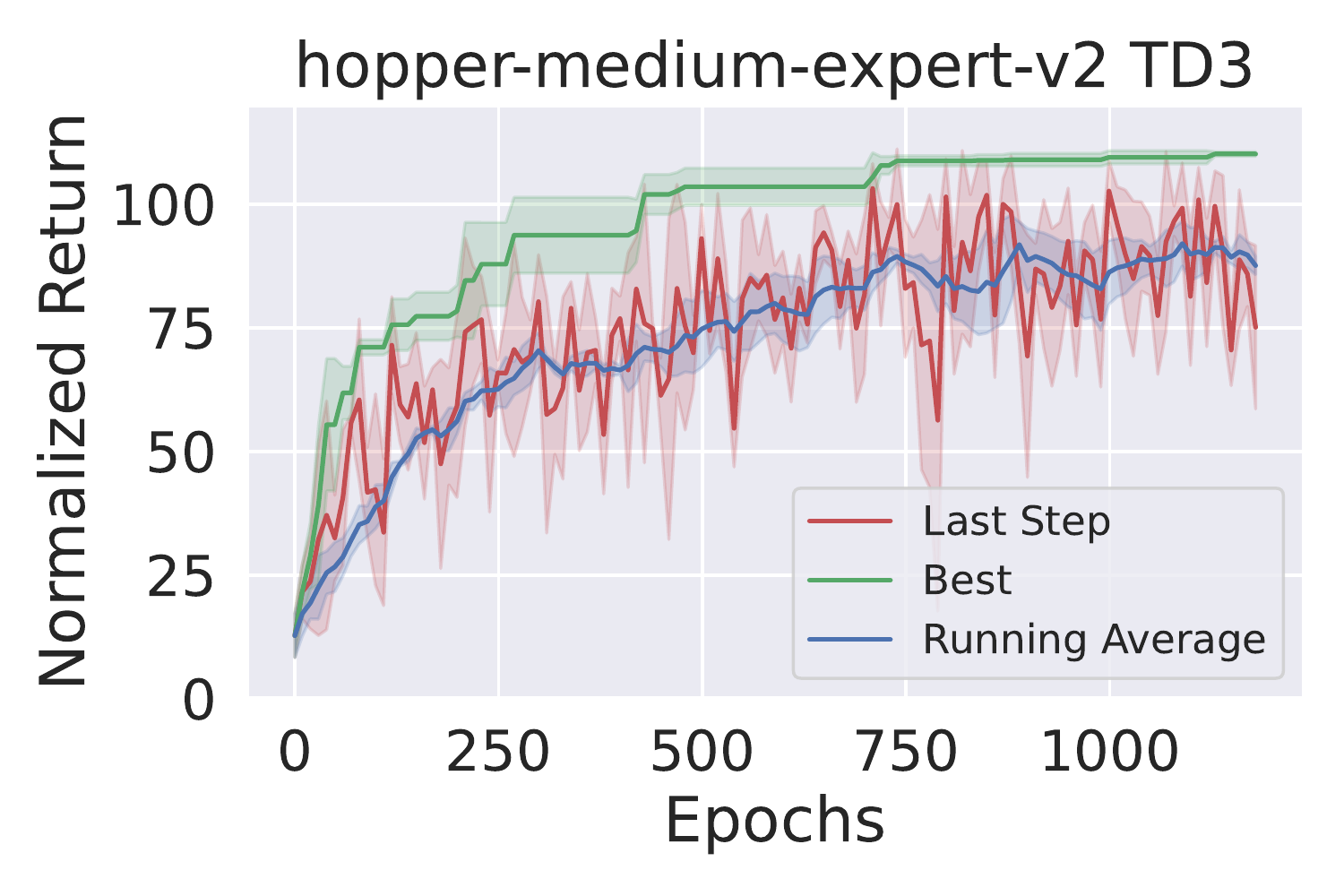}\\\includegraphics[width=0.225\linewidth]{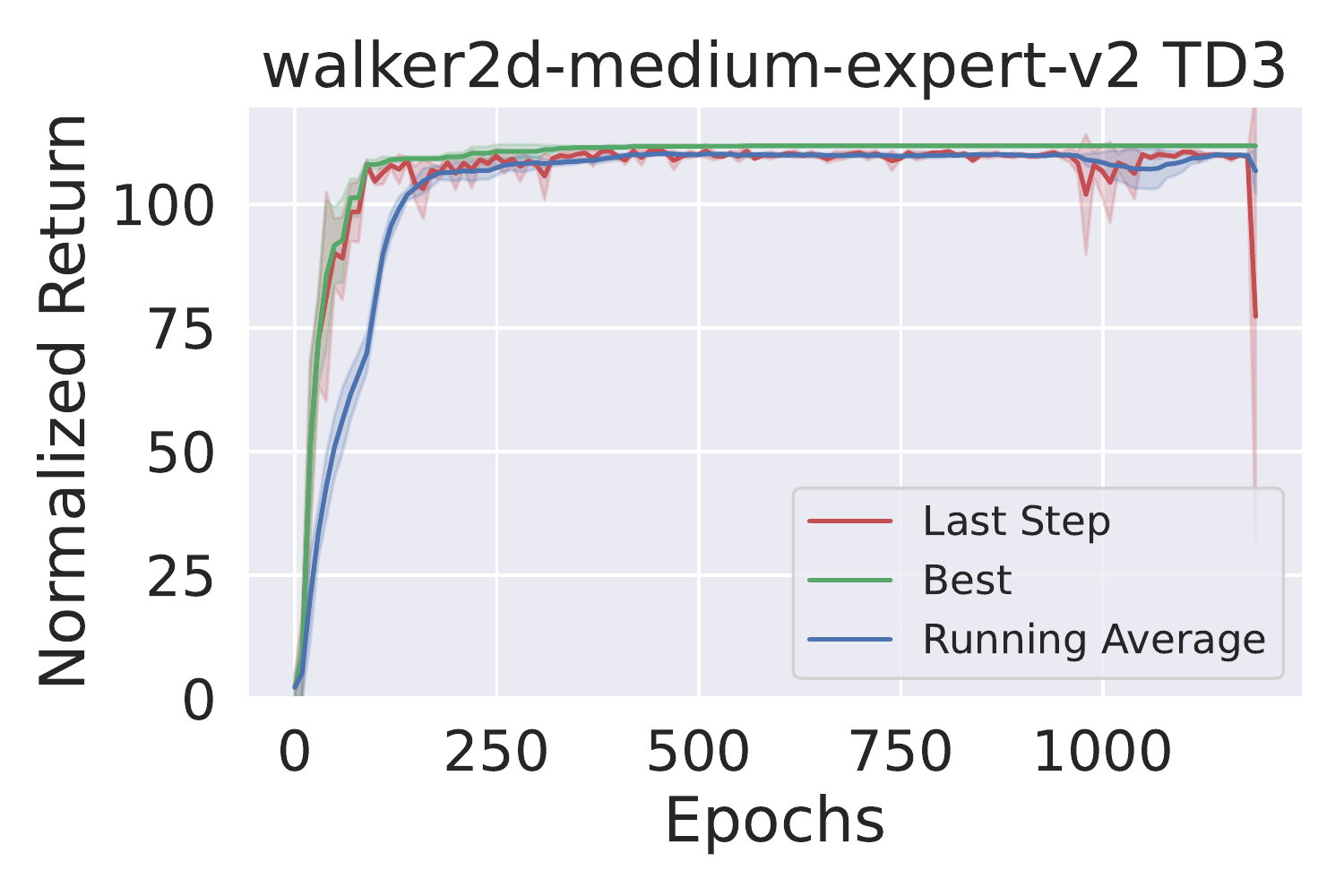}&\includegraphics[width=0.225\linewidth]{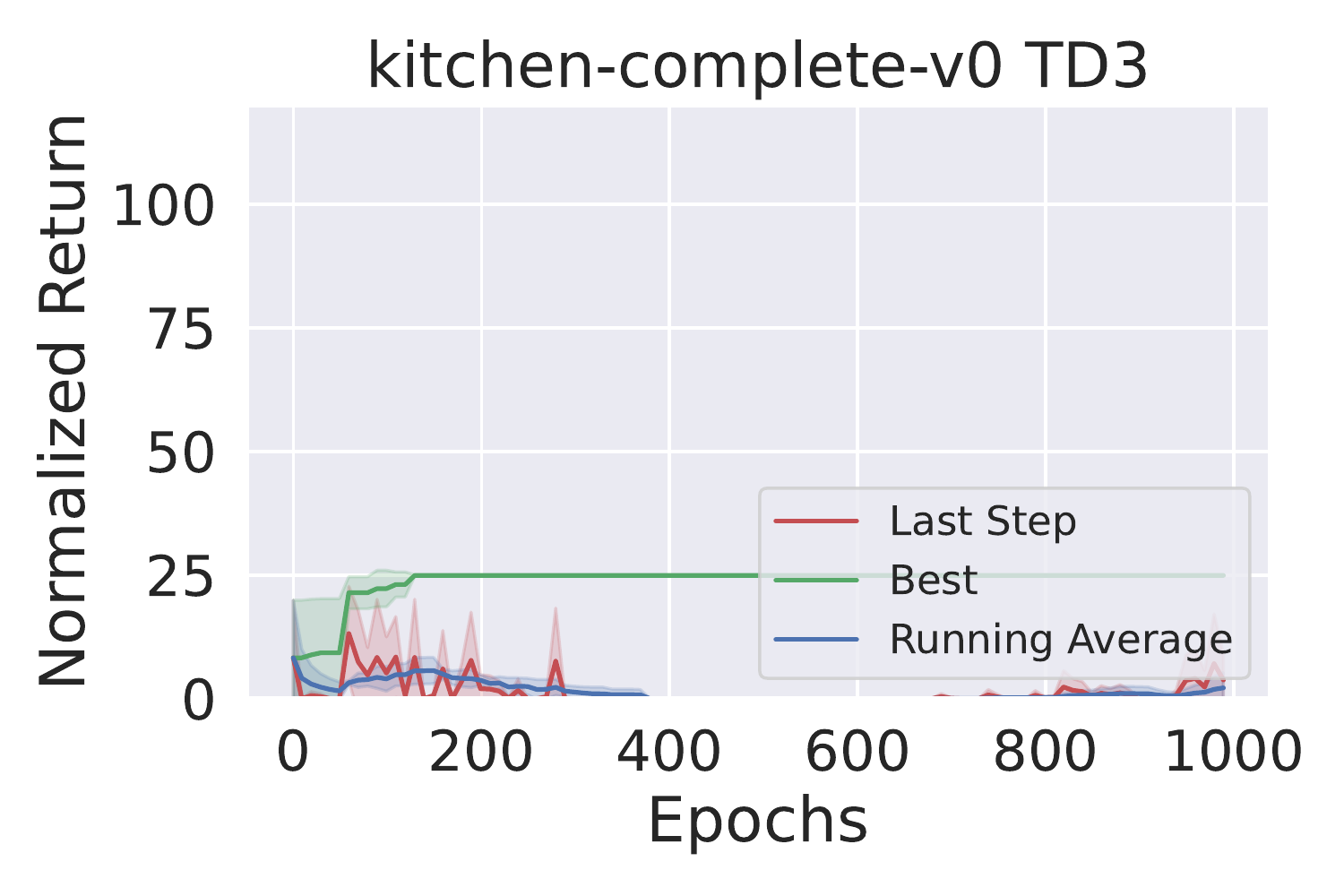}&\includegraphics[width=0.225\linewidth]{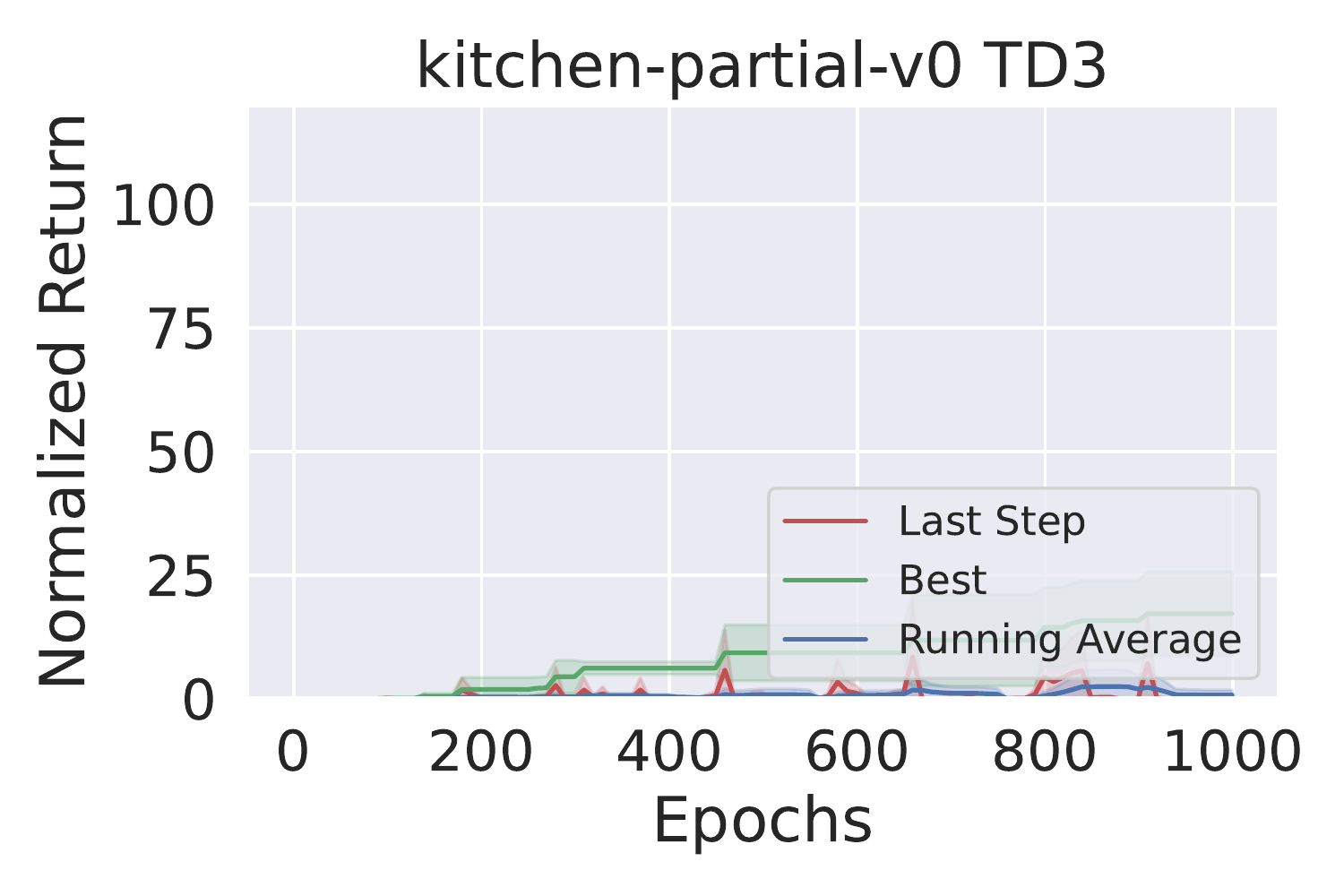}&\includegraphics[width=0.225\linewidth]{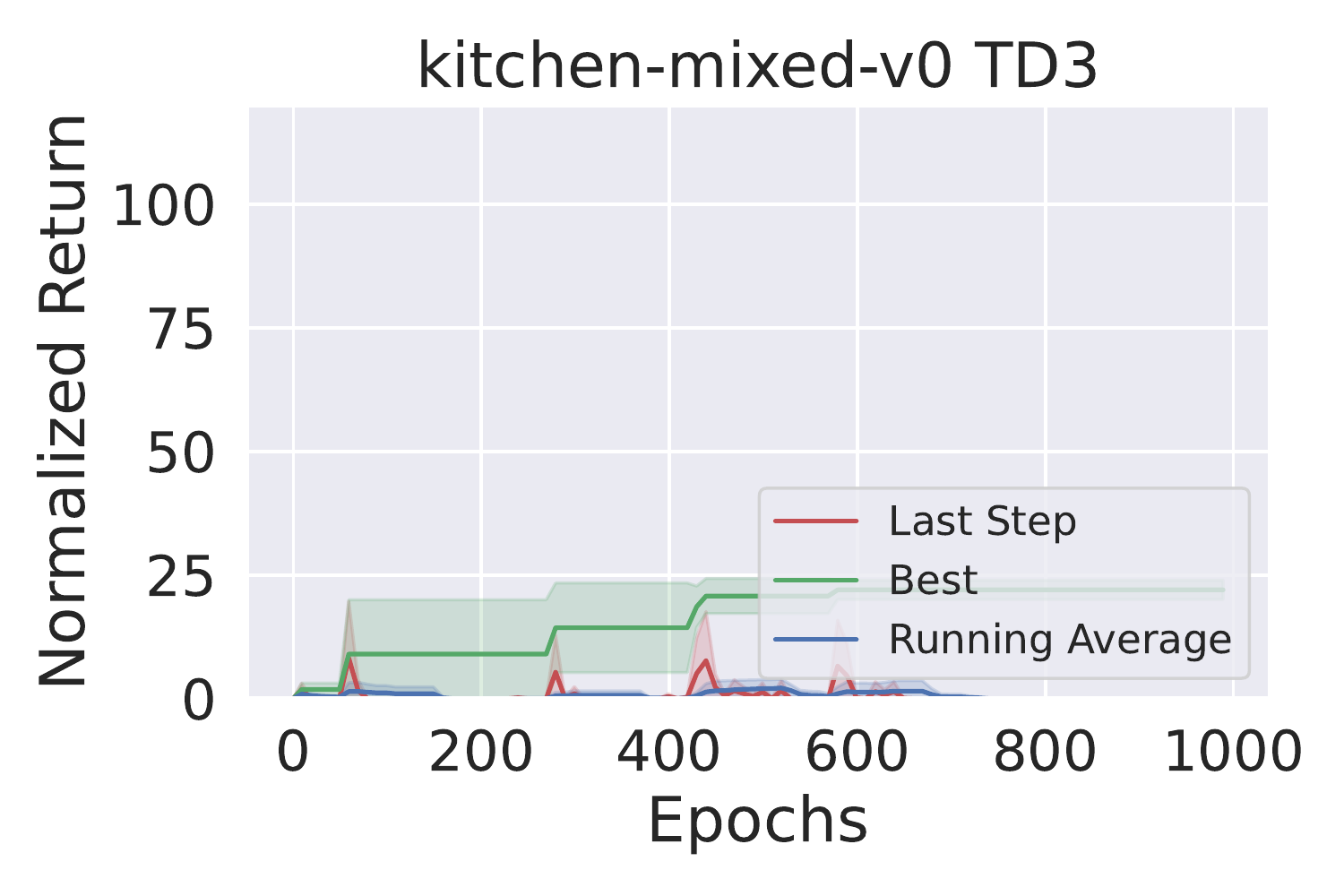}\\\includegraphics[width=0.225\linewidth]{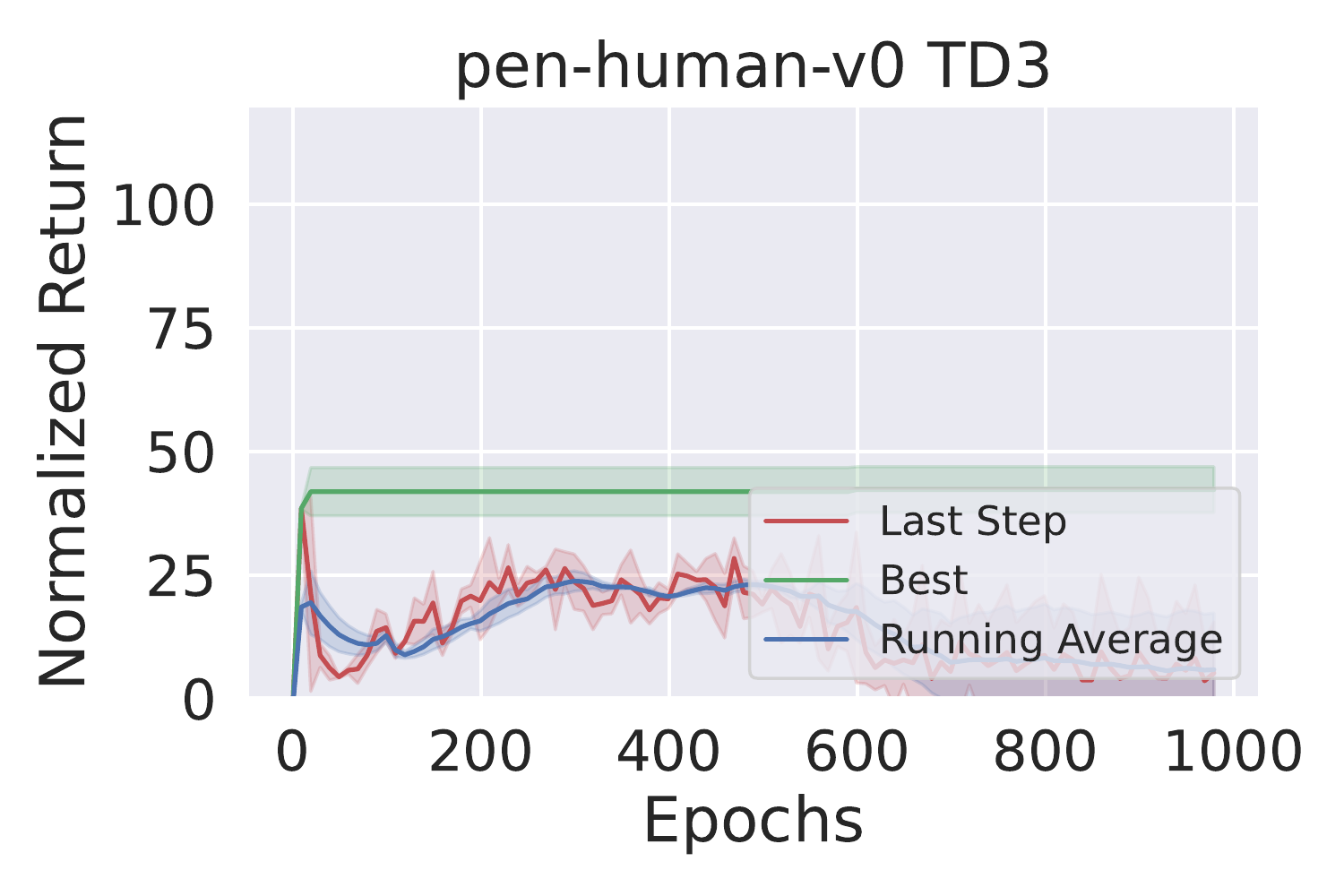}&\includegraphics[width=0.225\linewidth]{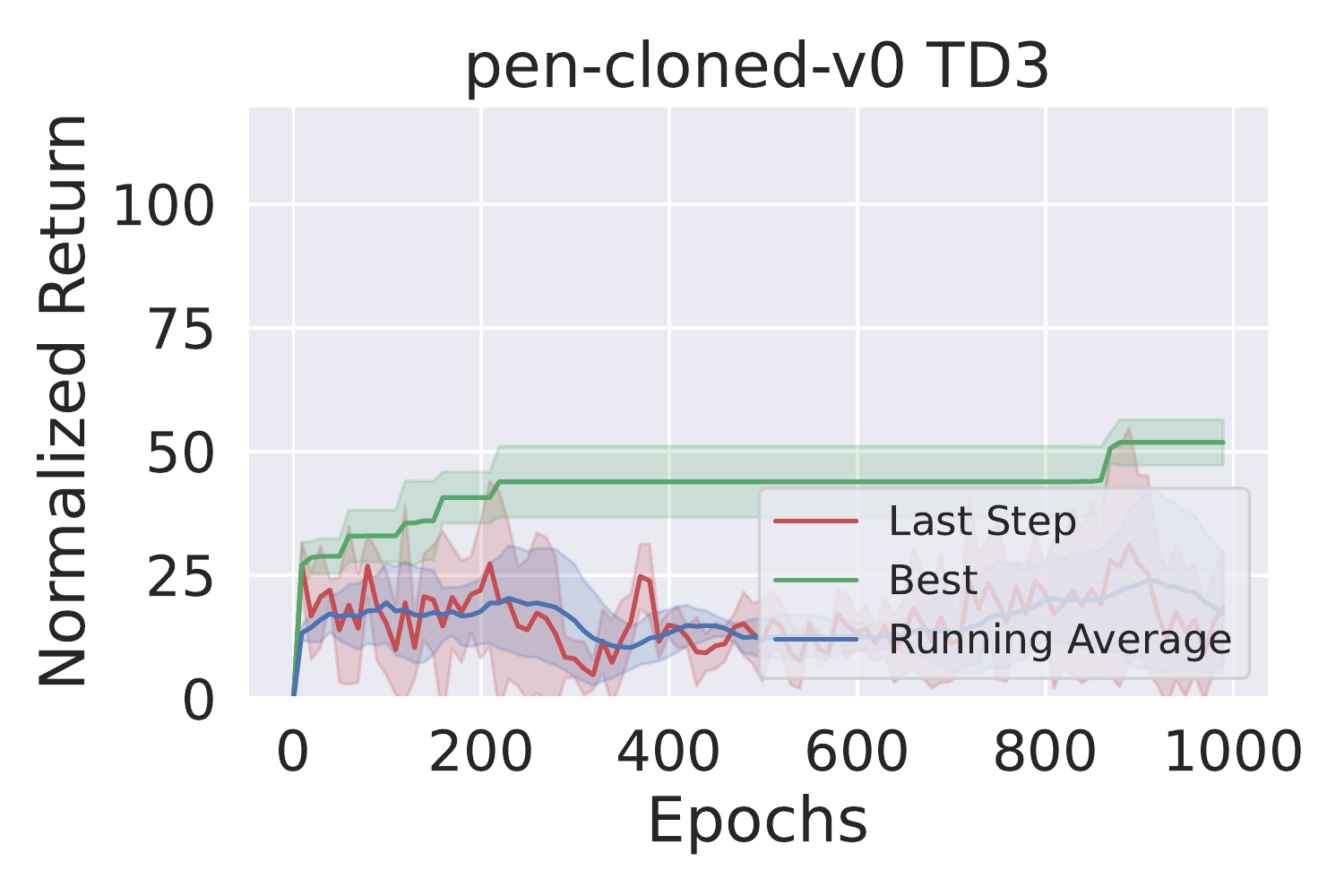}&\includegraphics[width=0.225\linewidth]{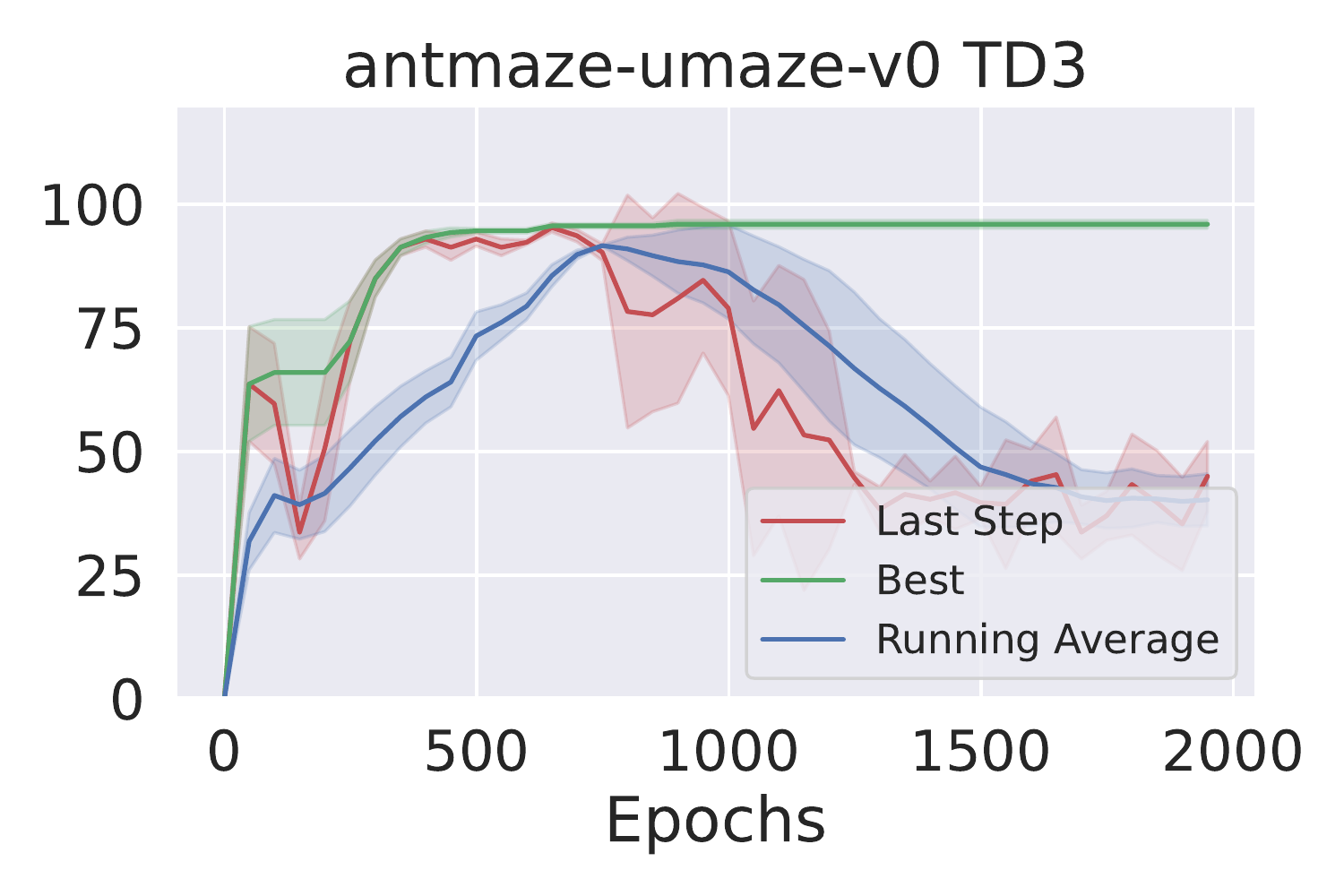}&\includegraphics[width=0.225\linewidth]{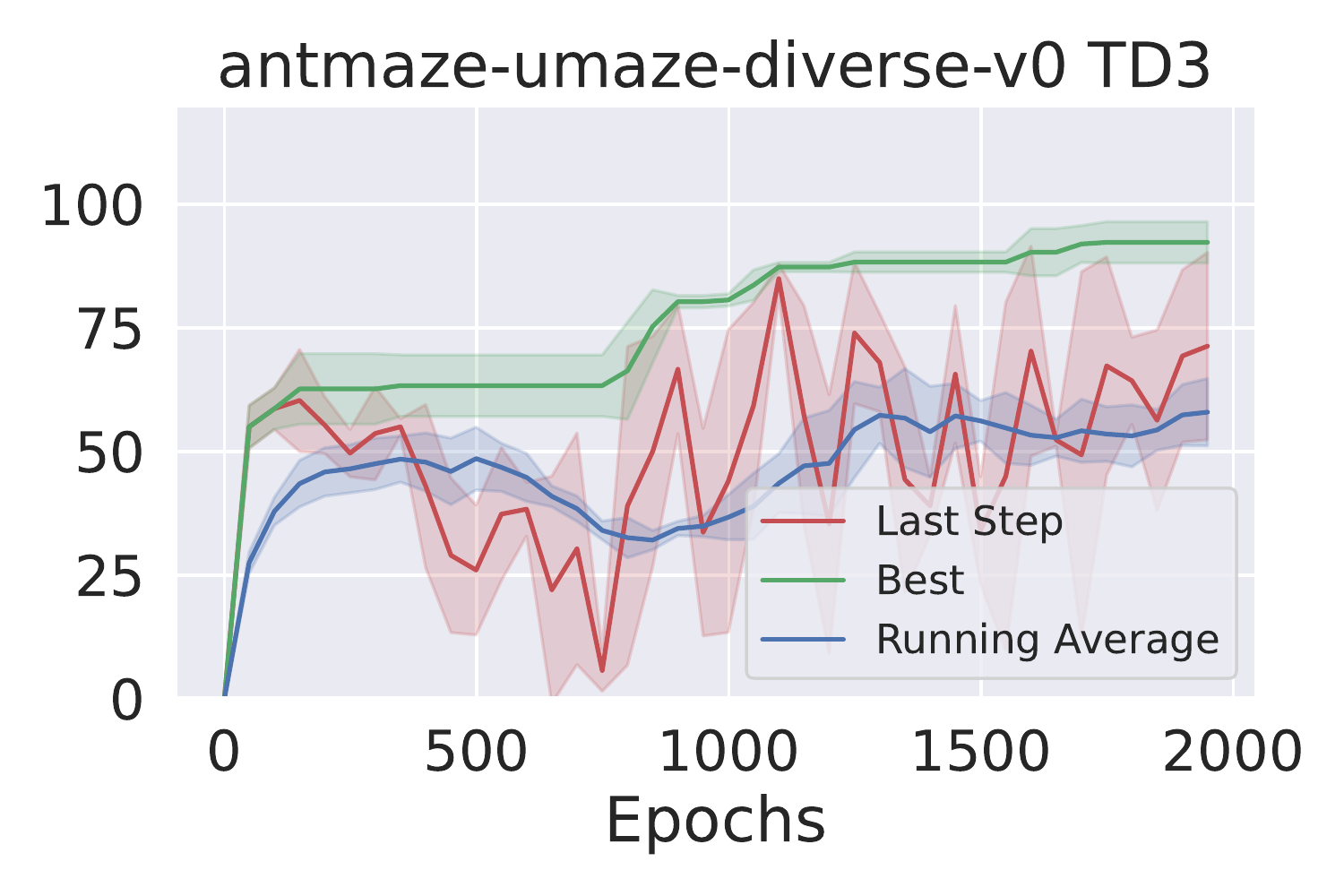}\\\includegraphics[width=0.225\linewidth]{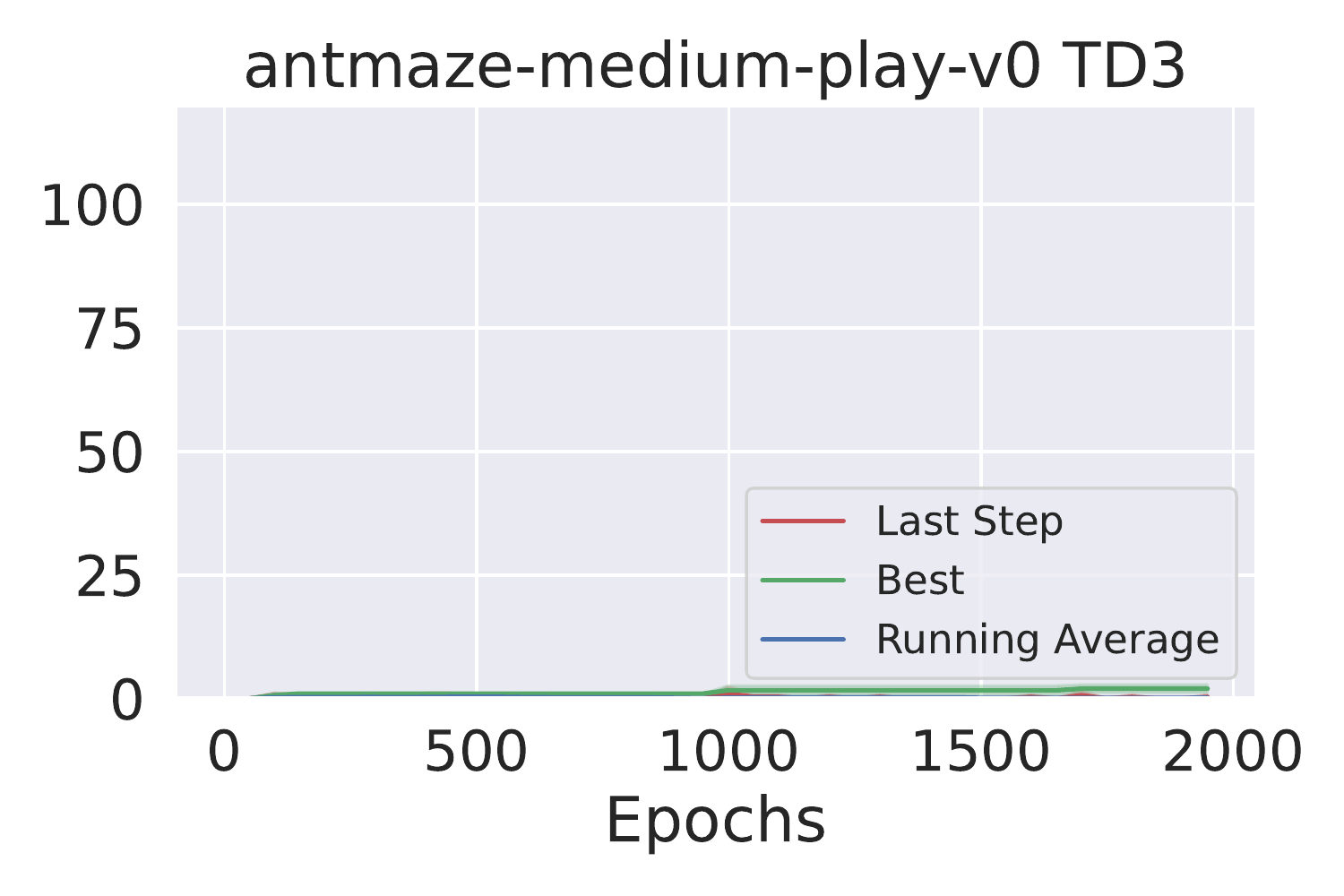}&\includegraphics[width=0.225\linewidth]{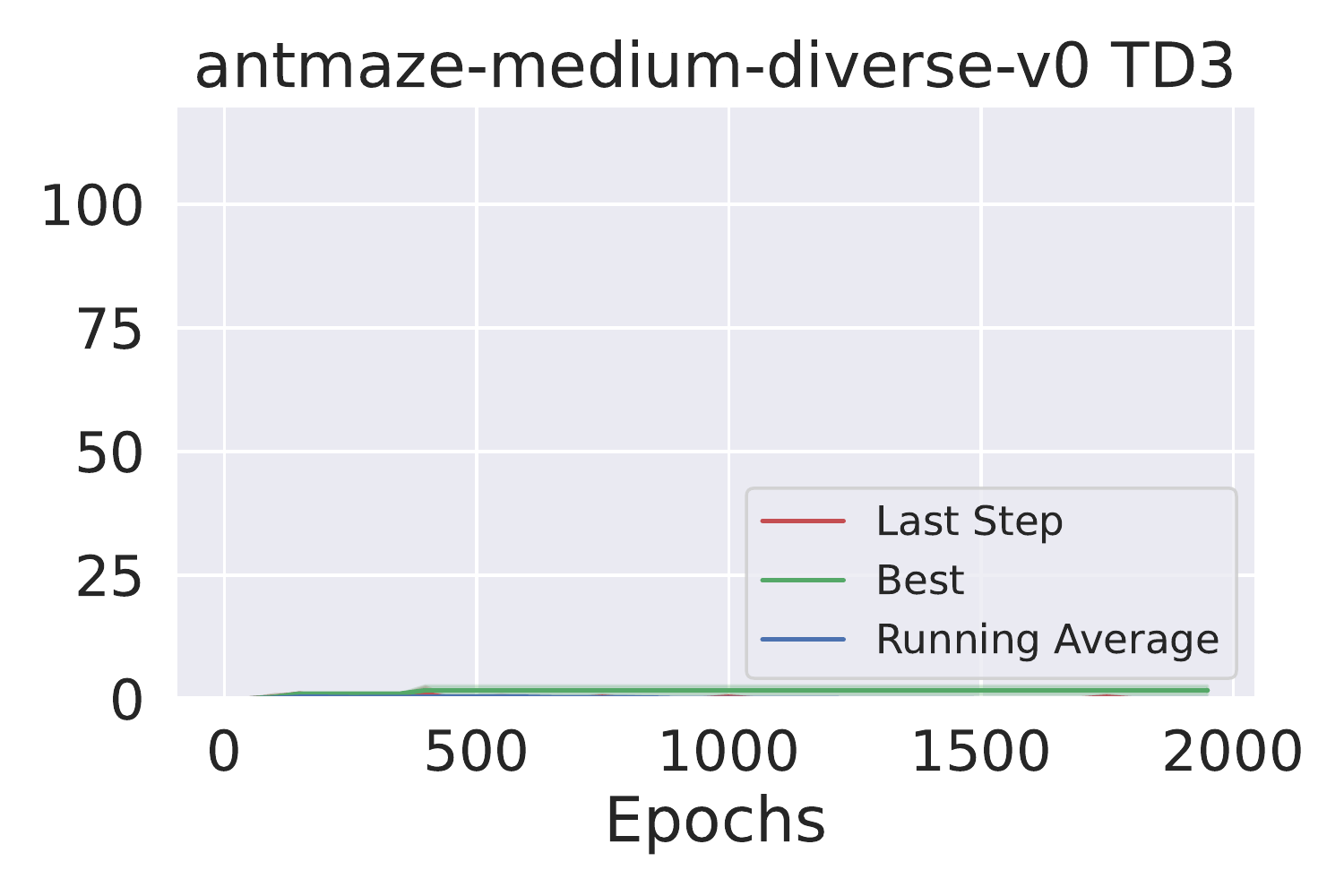}&\includegraphics[width=0.225\linewidth]{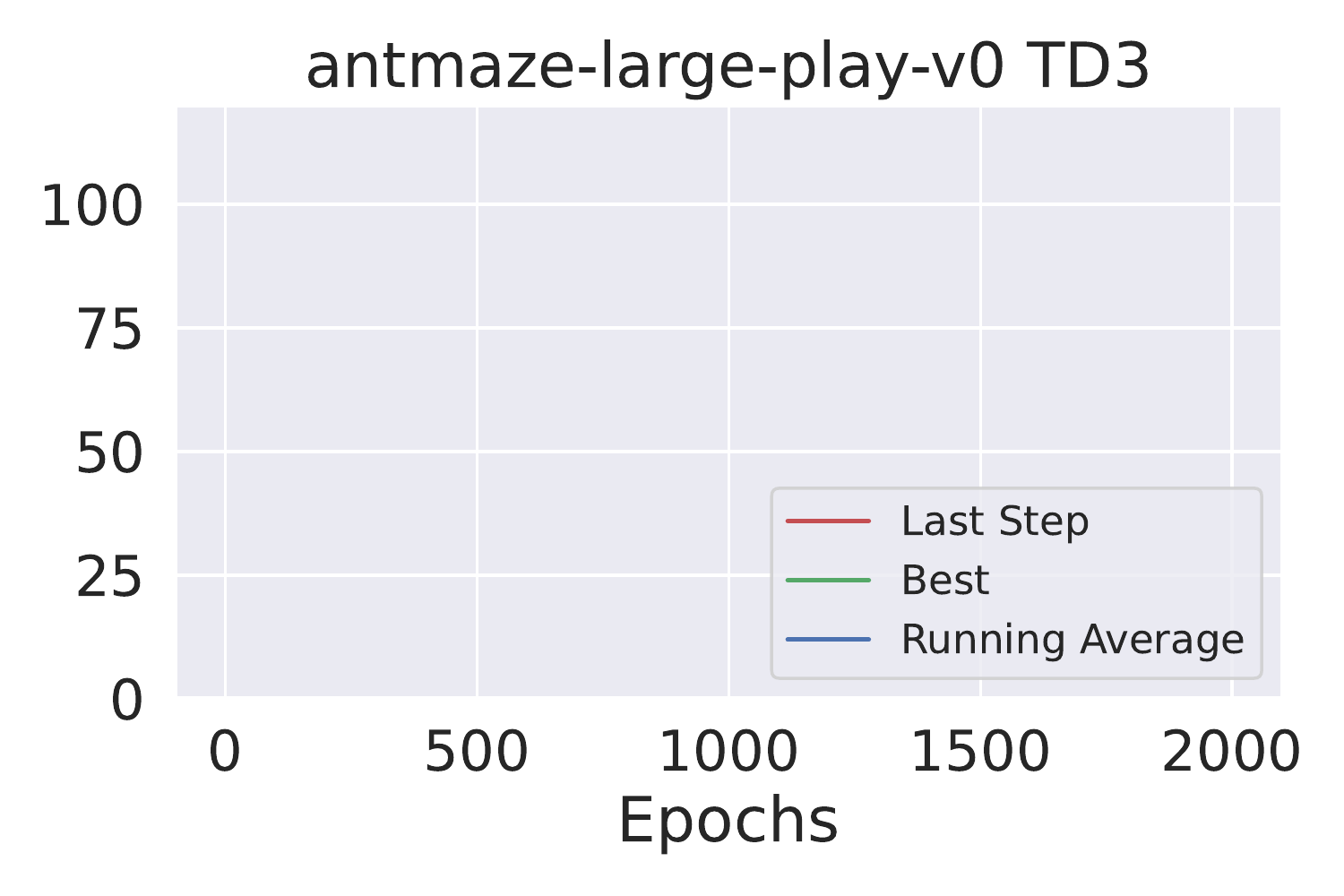}&\includegraphics[width=0.225\linewidth]{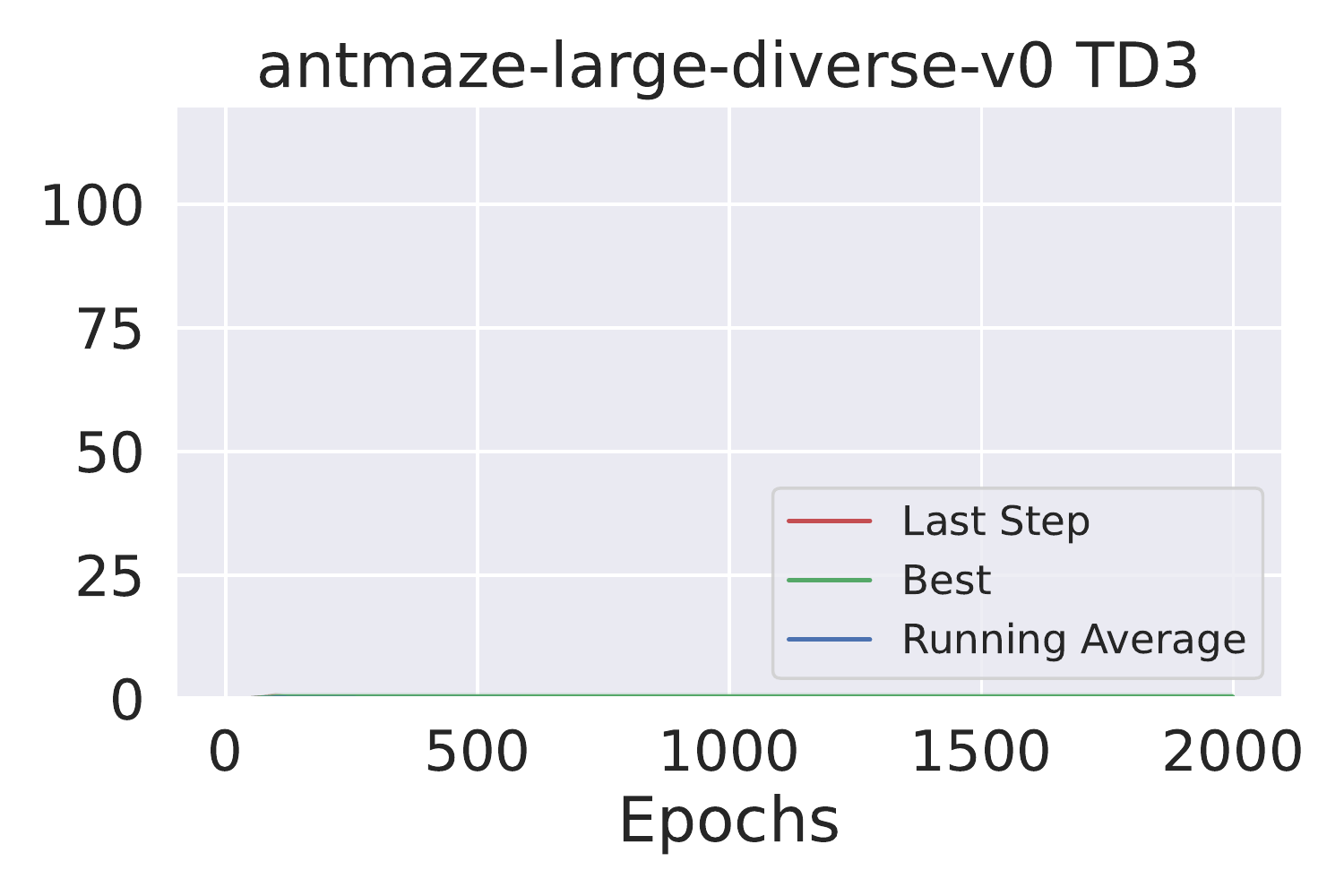}\\\end{tabular}
\centering
\caption{Training Curves of TD3+BC on D4RL}
\end{figure*}
\begin{figure*}[htb]
\centering
\begin{tabular}{cccccc}
\includegraphics[width=0.225\linewidth]{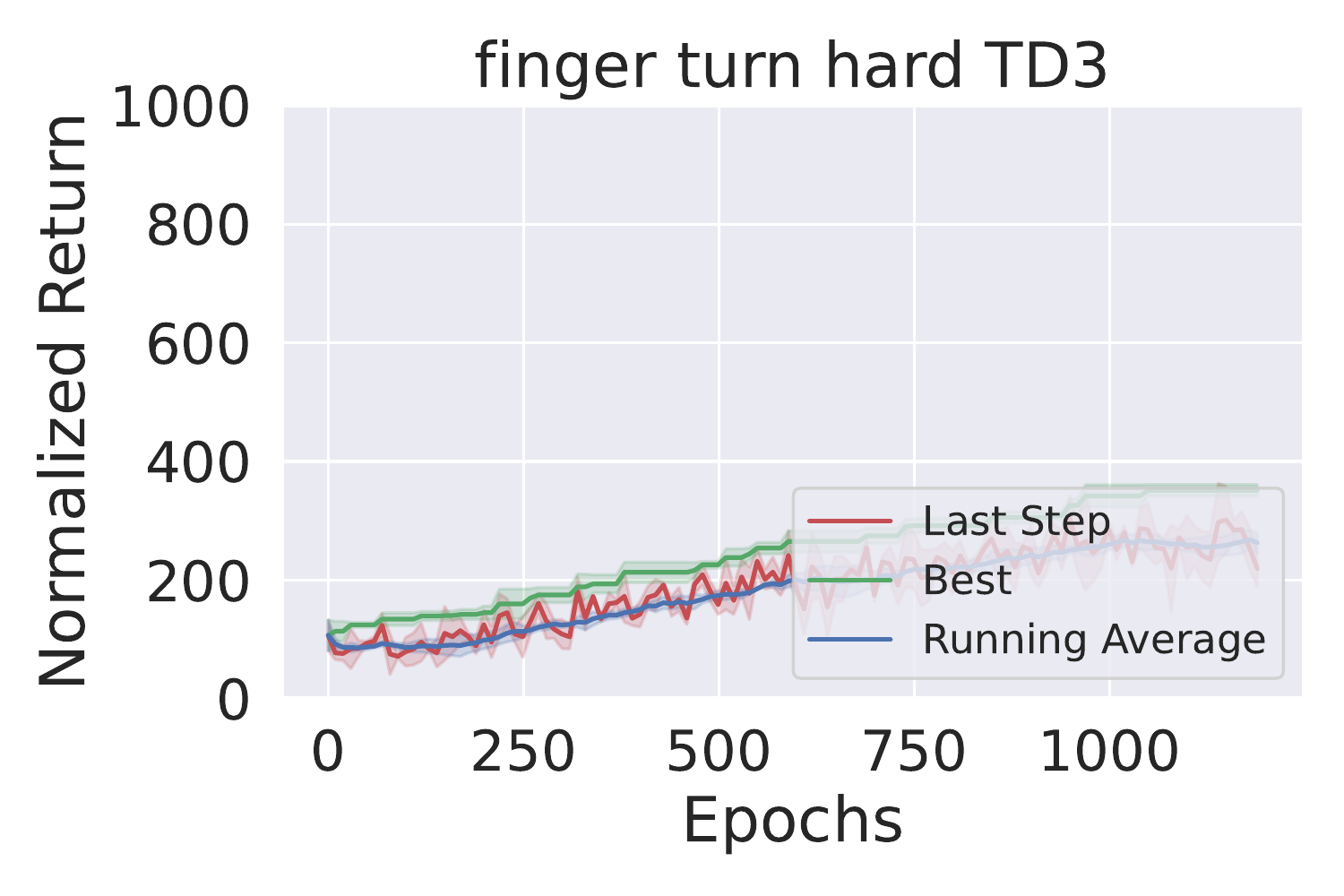}&\includegraphics[width=0.225\linewidth]{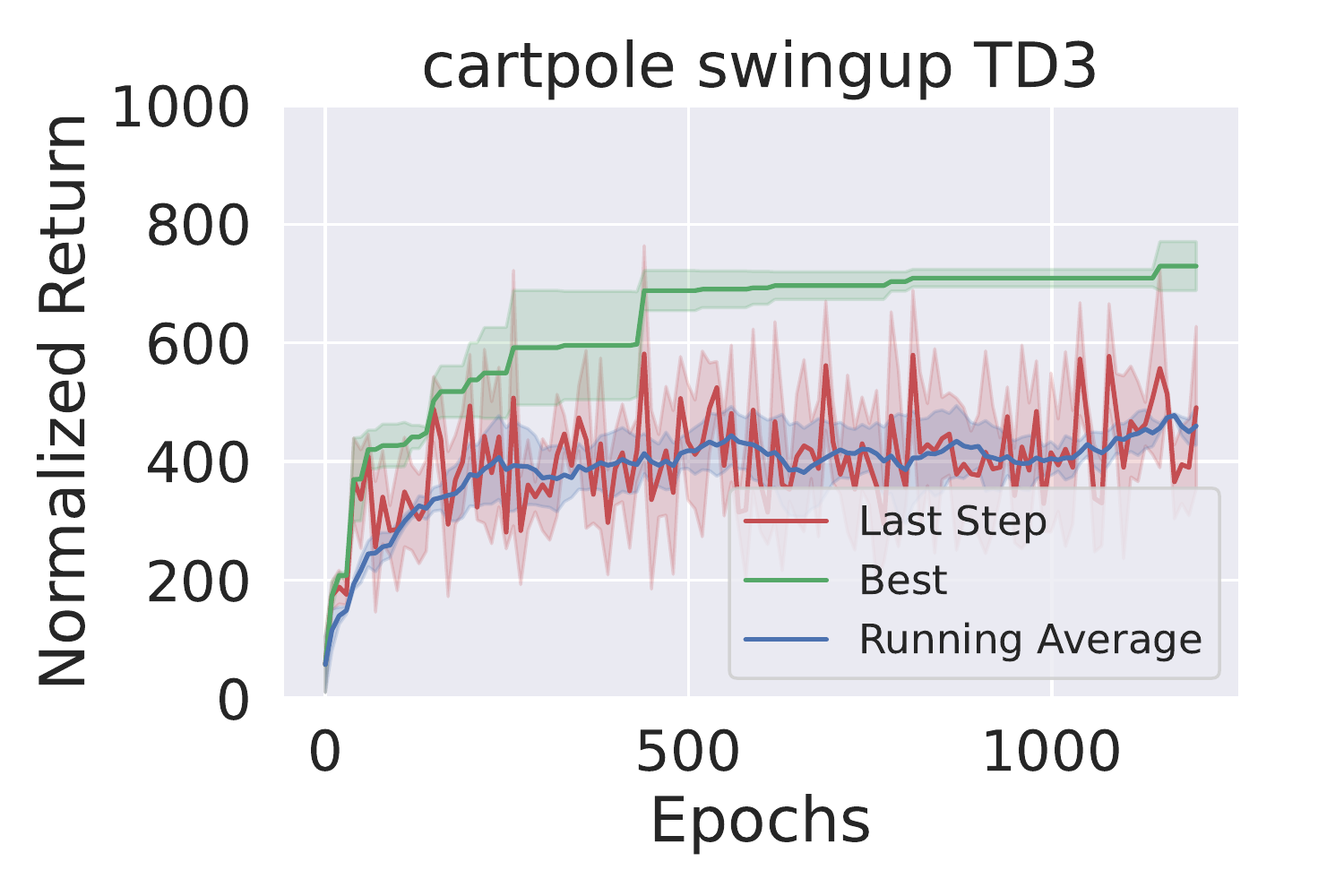}&\includegraphics[width=0.225\linewidth]{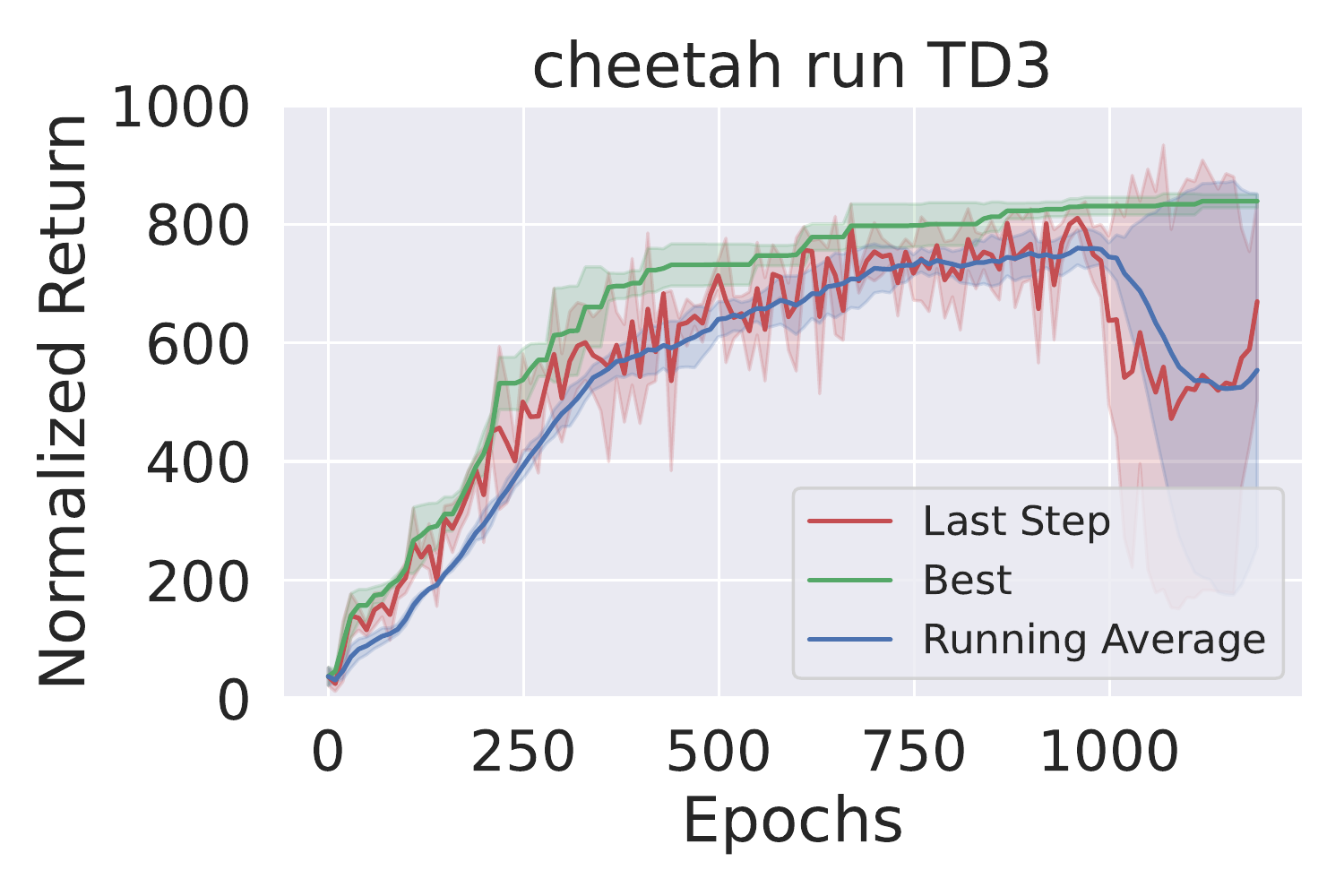}&\includegraphics[width=0.225\linewidth]{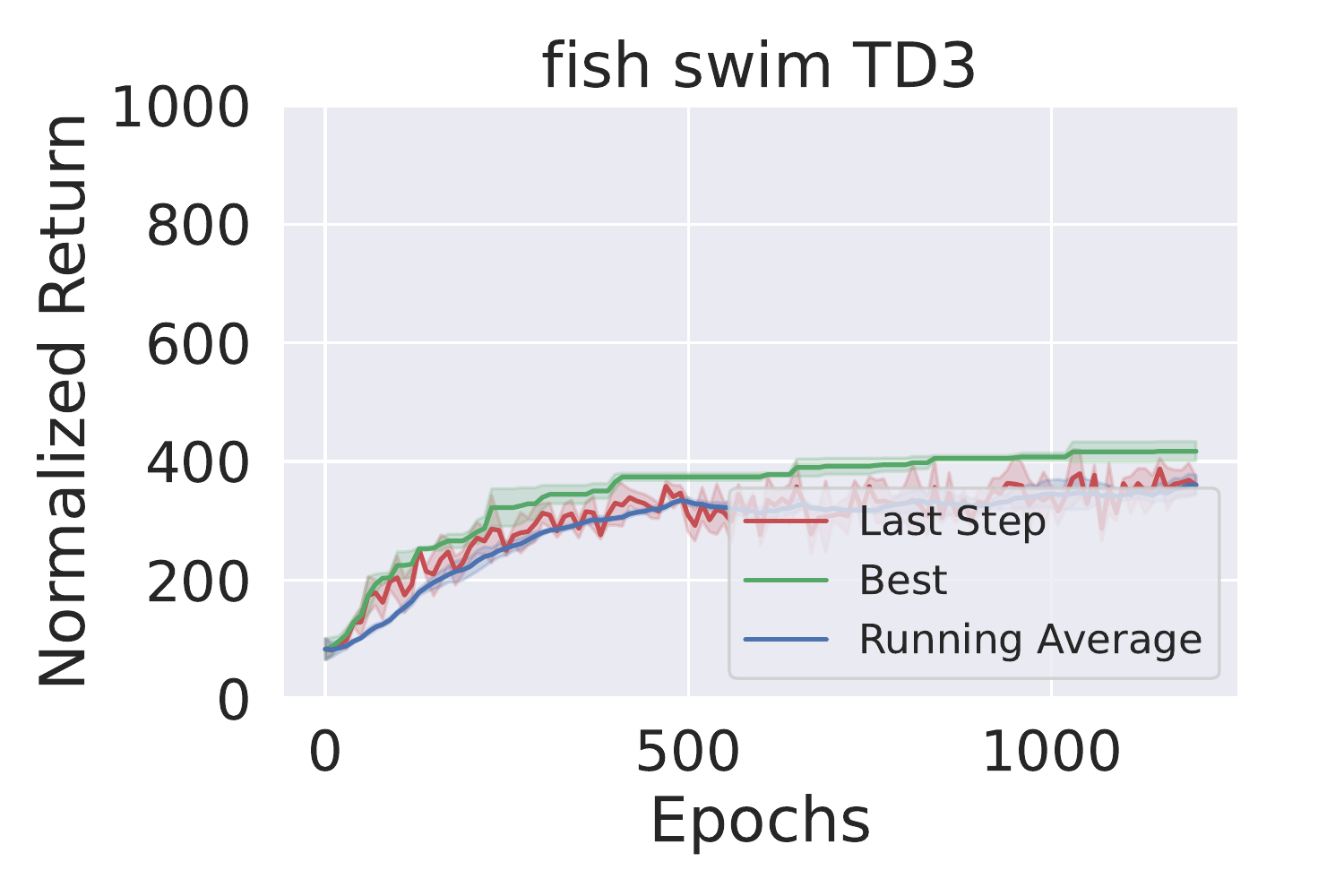}\\\includegraphics[width=0.225\linewidth]{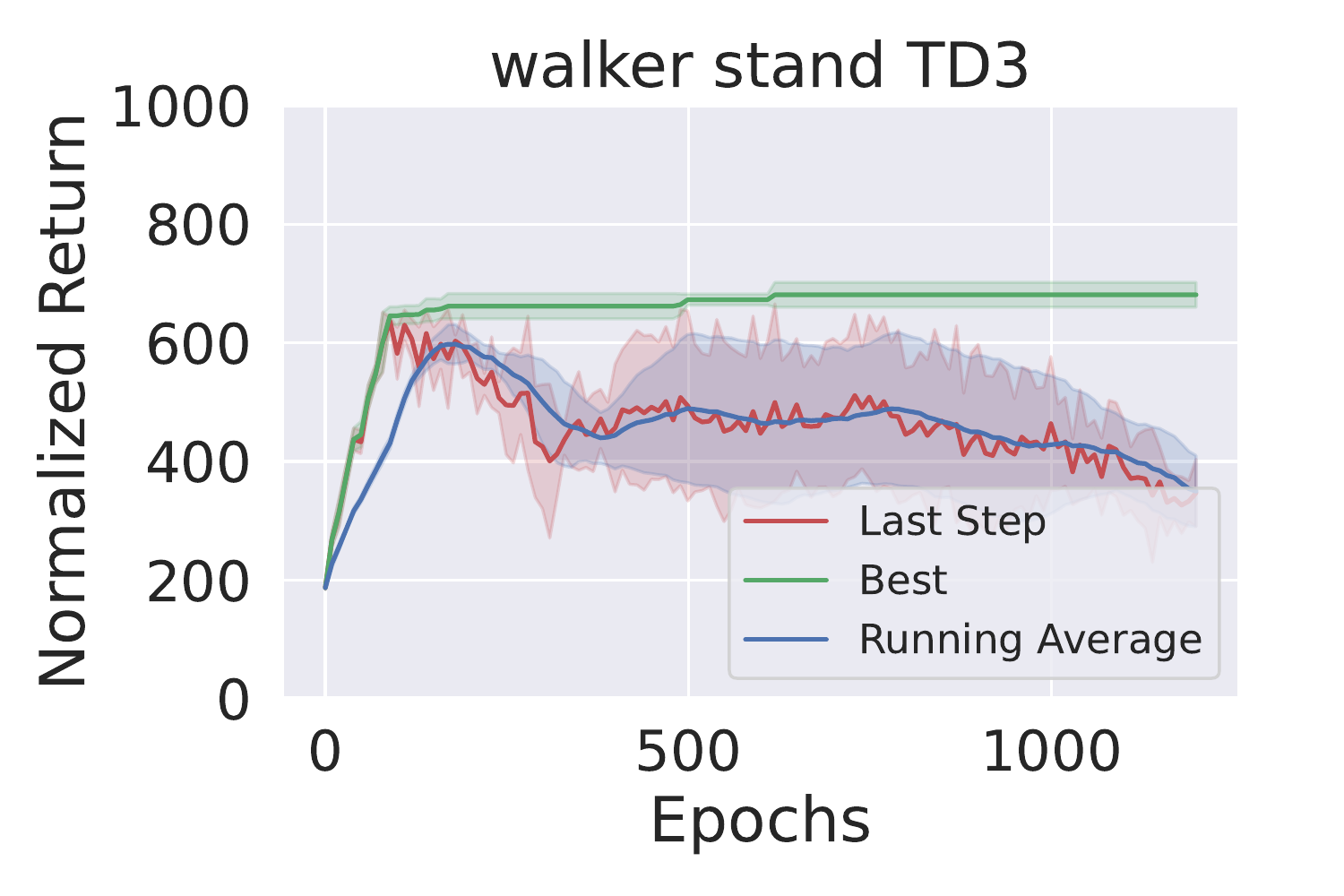}&\includegraphics[width=0.225\linewidth]{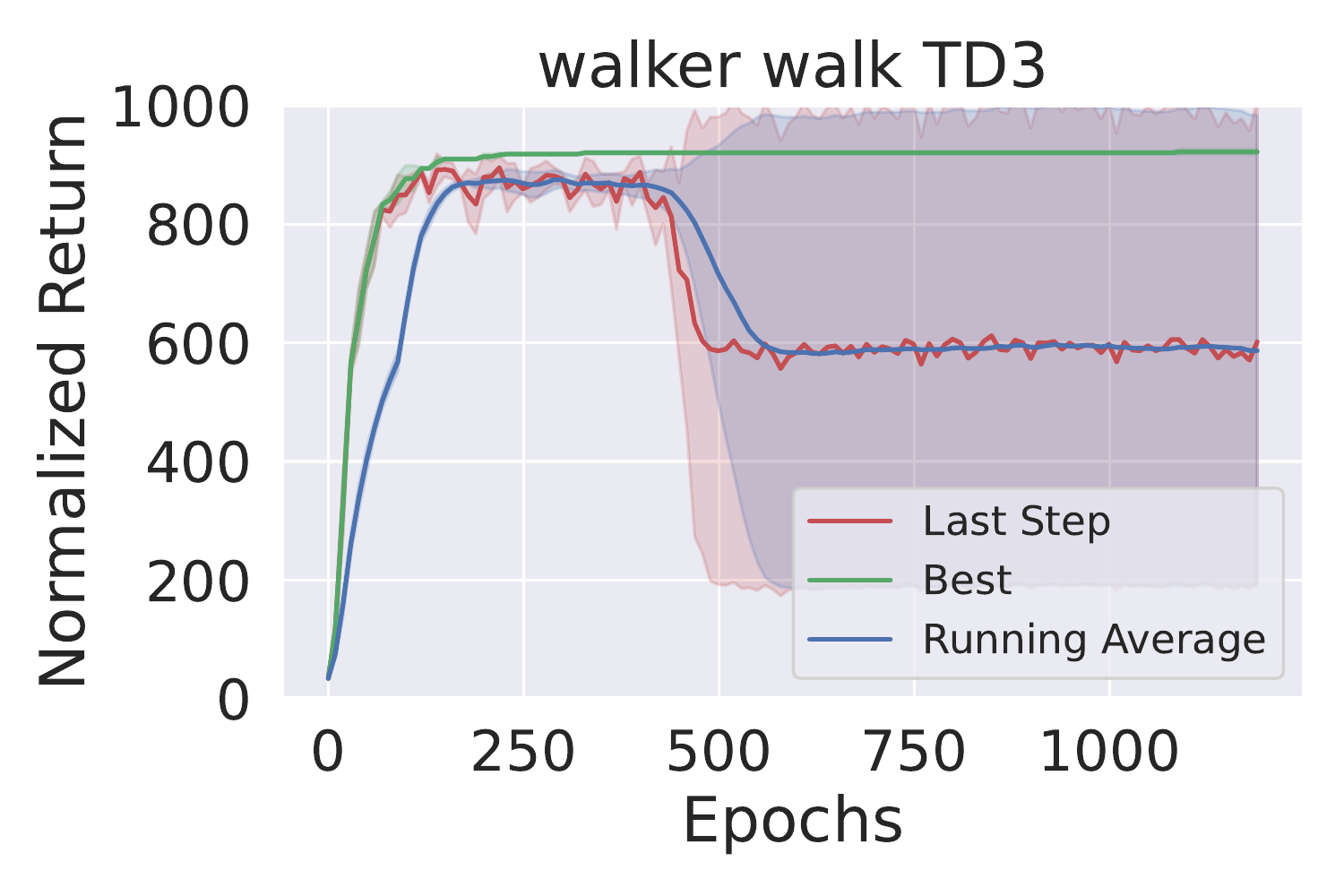}&&\\\end{tabular}
\centering
\caption{Training Curves of TD3+BC on RLUP}
\end{figure*}
\begin{figure*}[htb]
\centering
\begin{tabular}{cccccc}
\includegraphics[width=0.225\linewidth]{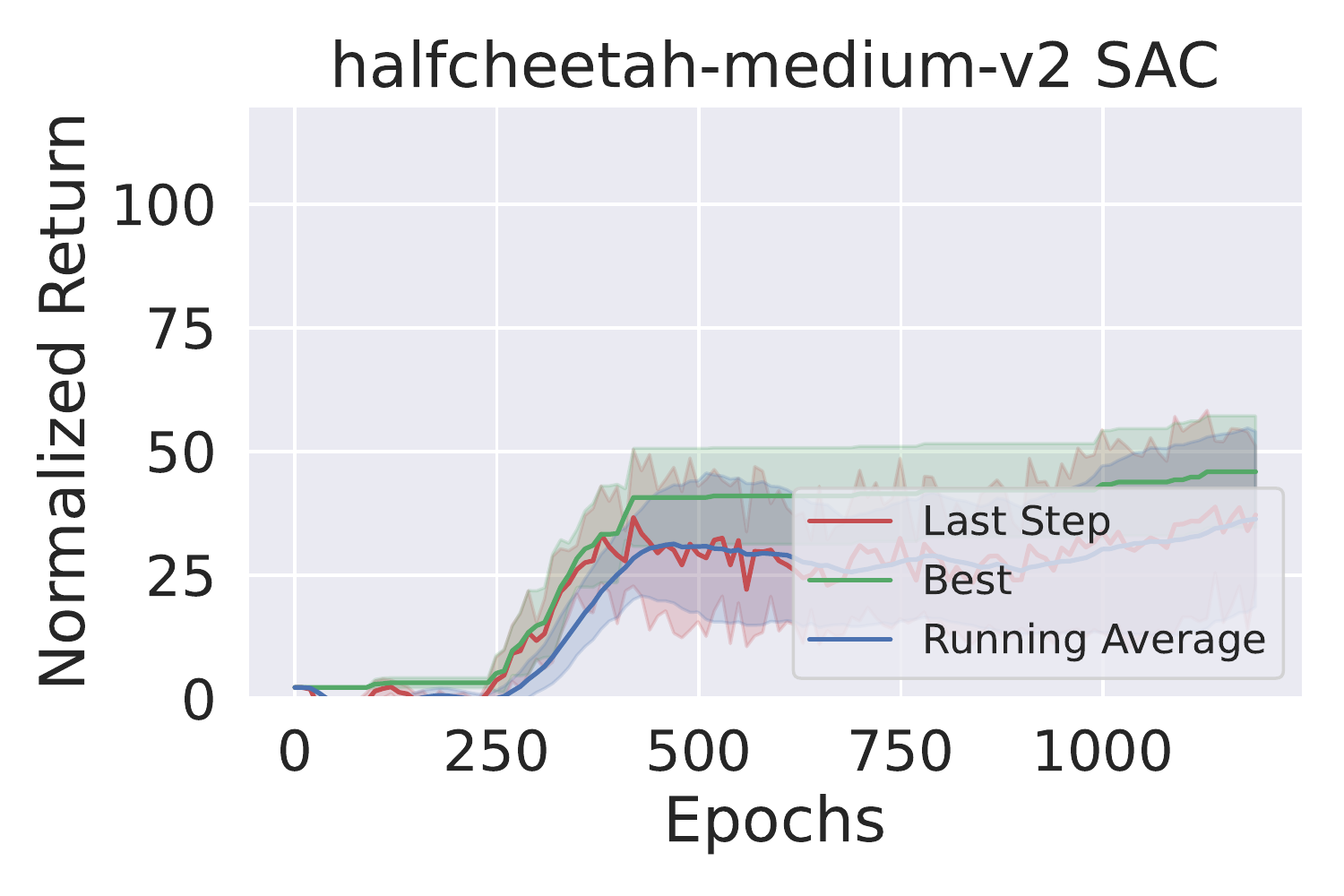}&\includegraphics[width=0.225\linewidth]{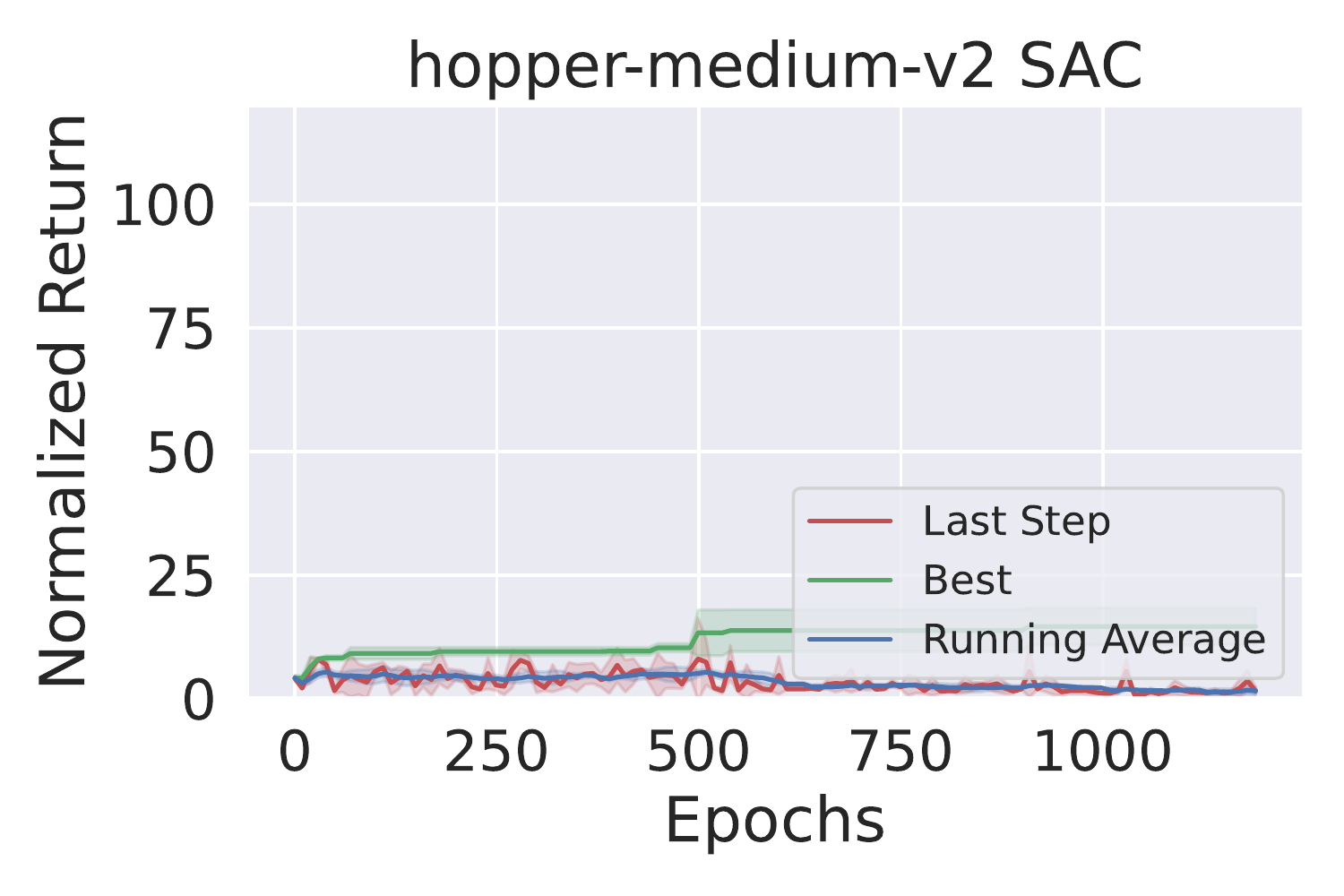}&\includegraphics[width=0.225\linewidth]{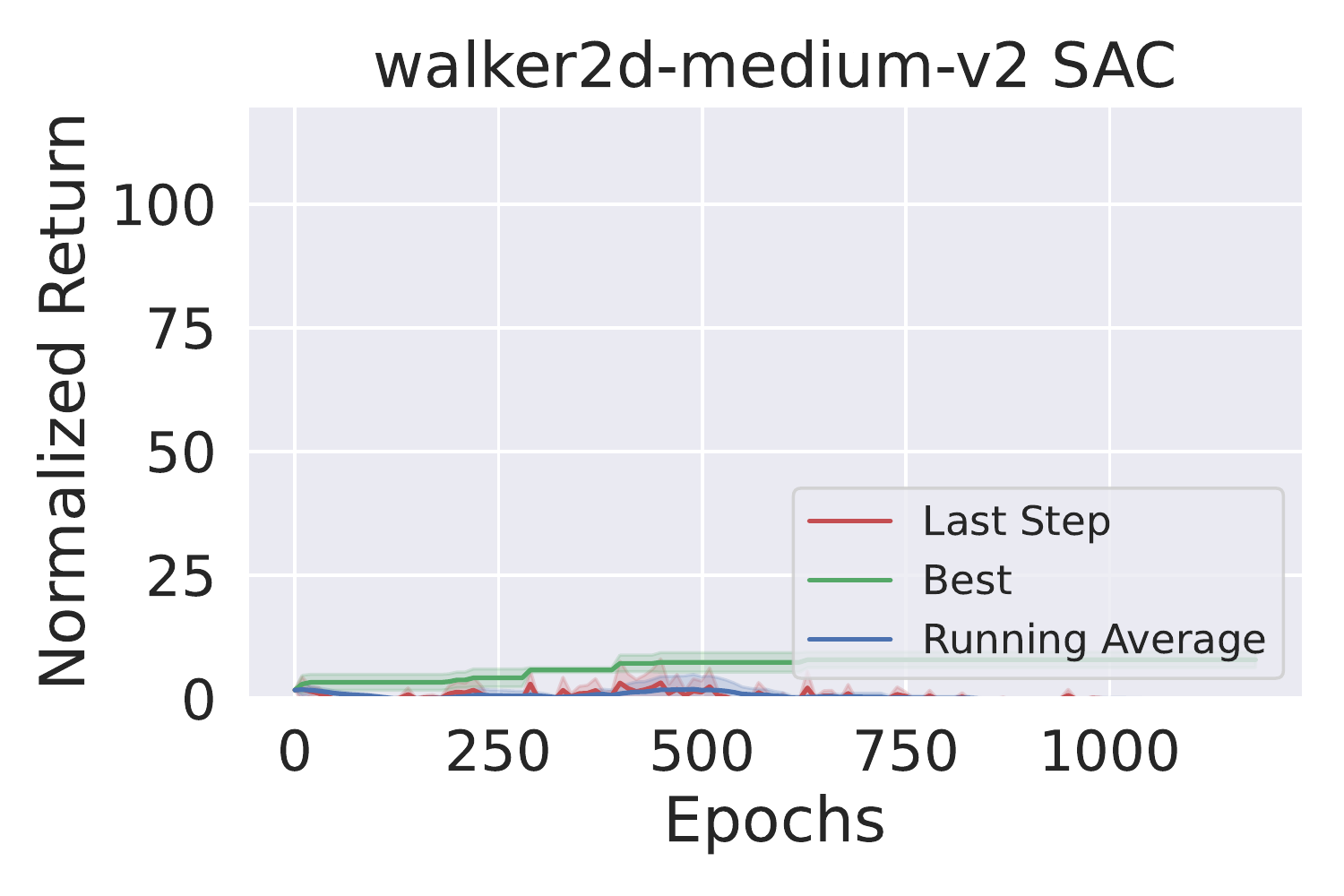}&\includegraphics[width=0.225\linewidth]{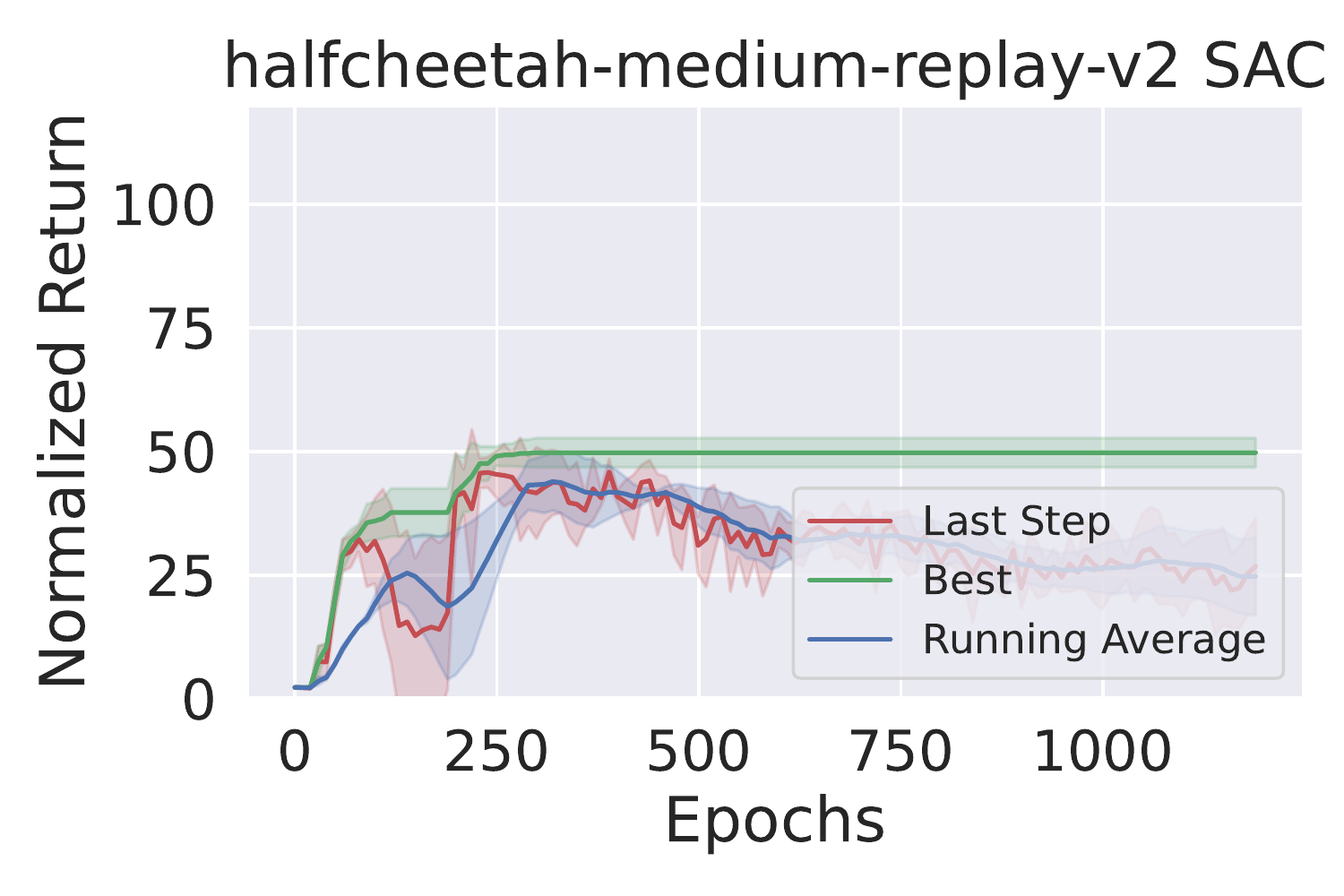}\\\includegraphics[width=0.225\linewidth]{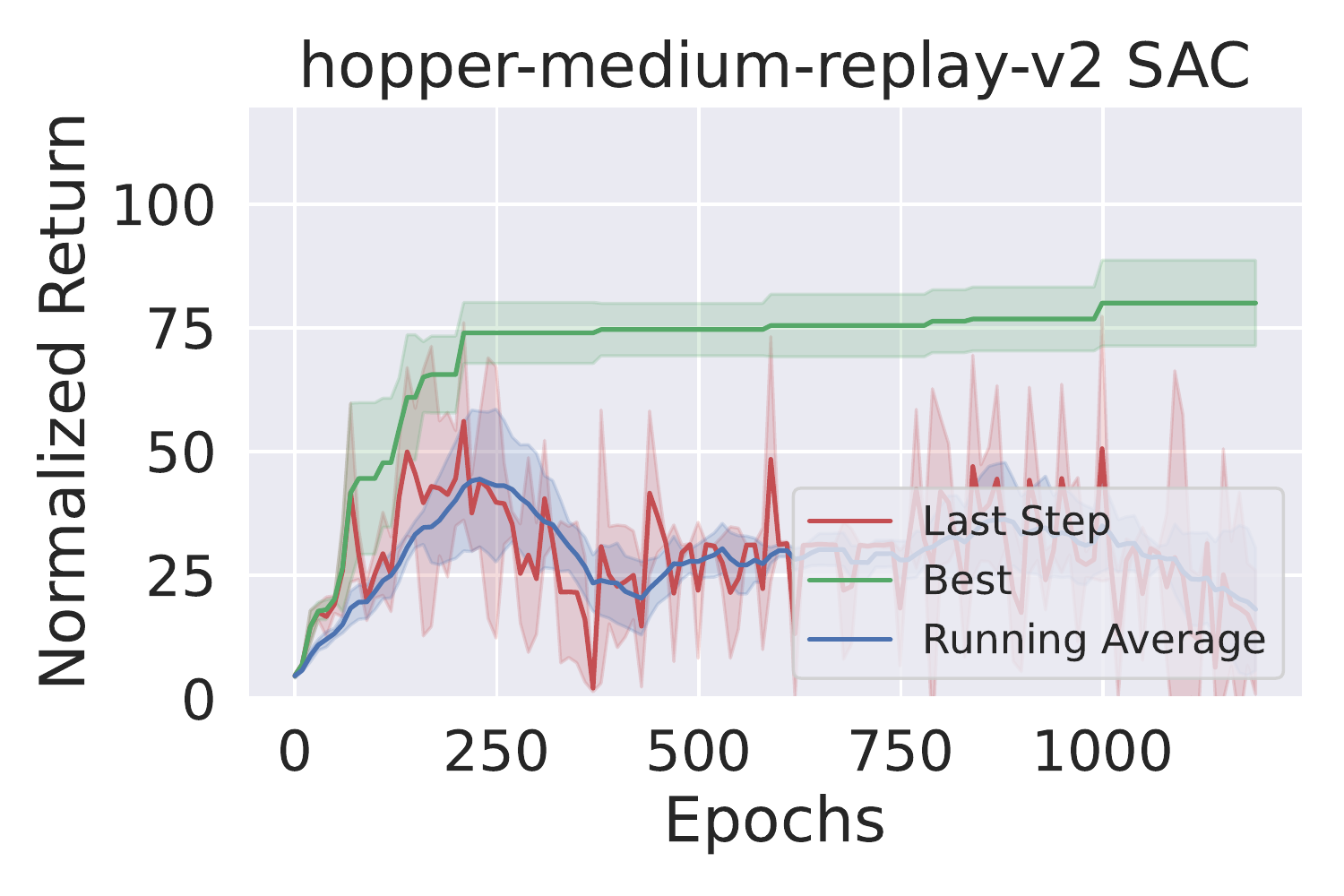}&\includegraphics[width=0.225\linewidth]{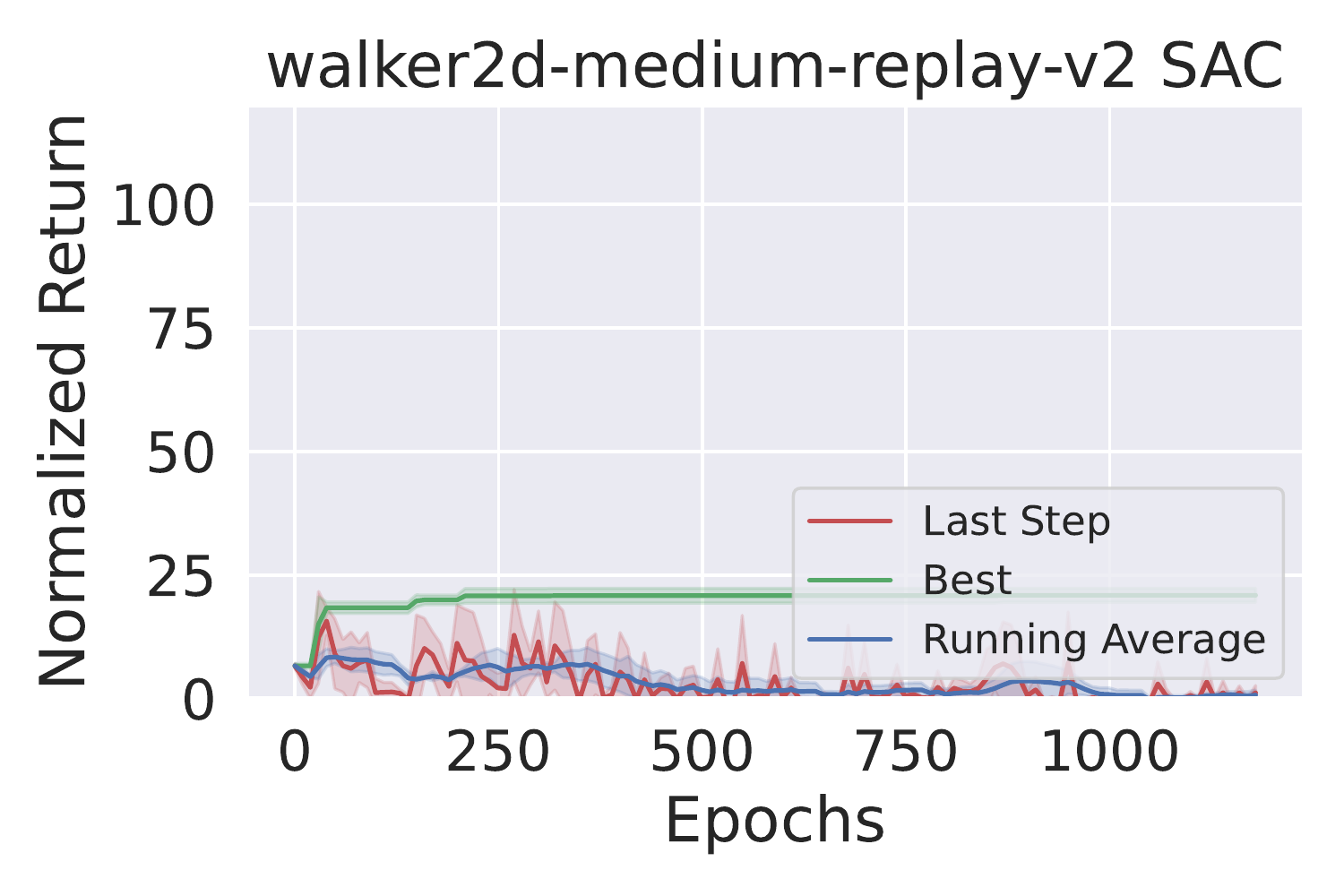}&\includegraphics[width=0.225\linewidth]{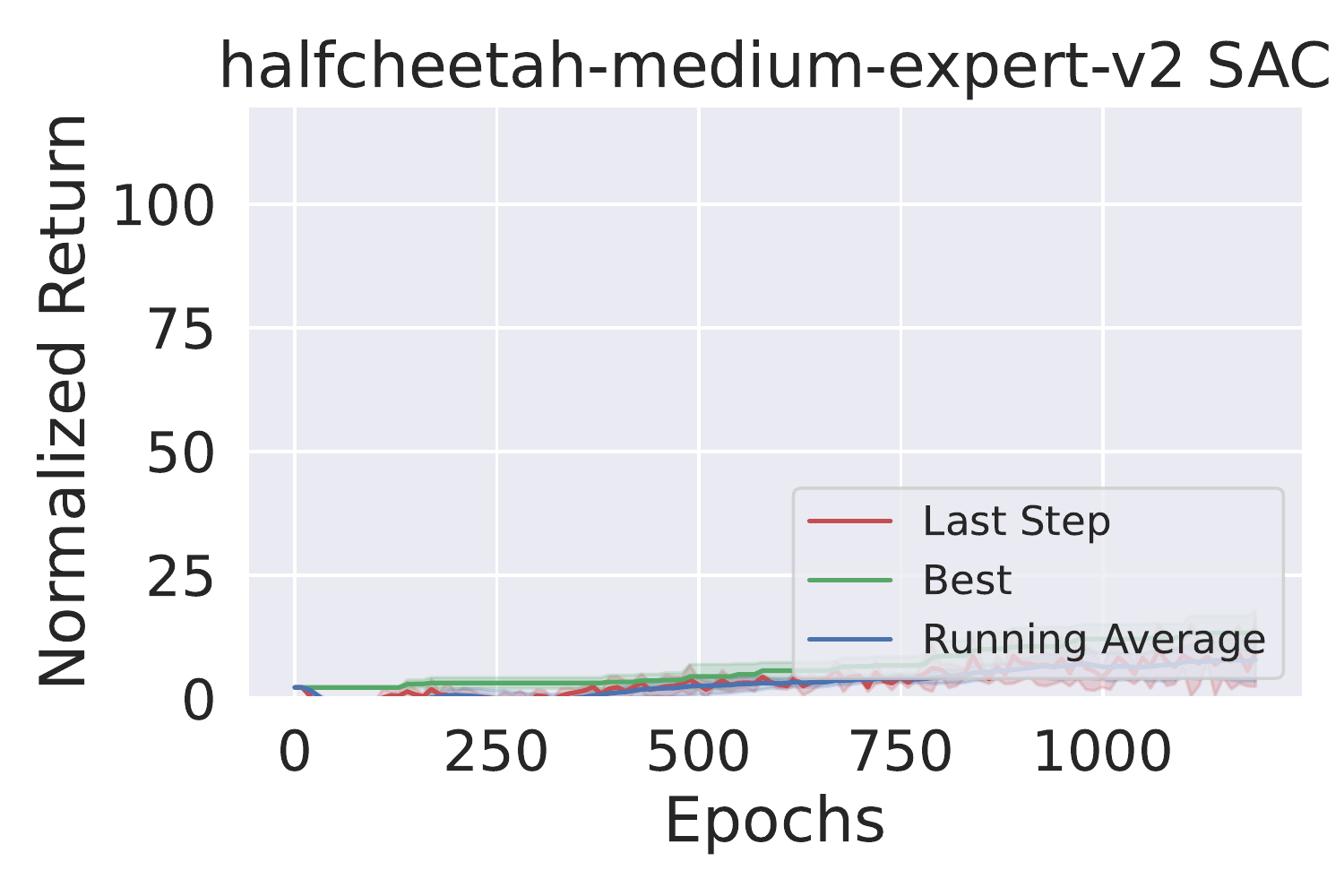}&\includegraphics[width=0.225\linewidth]{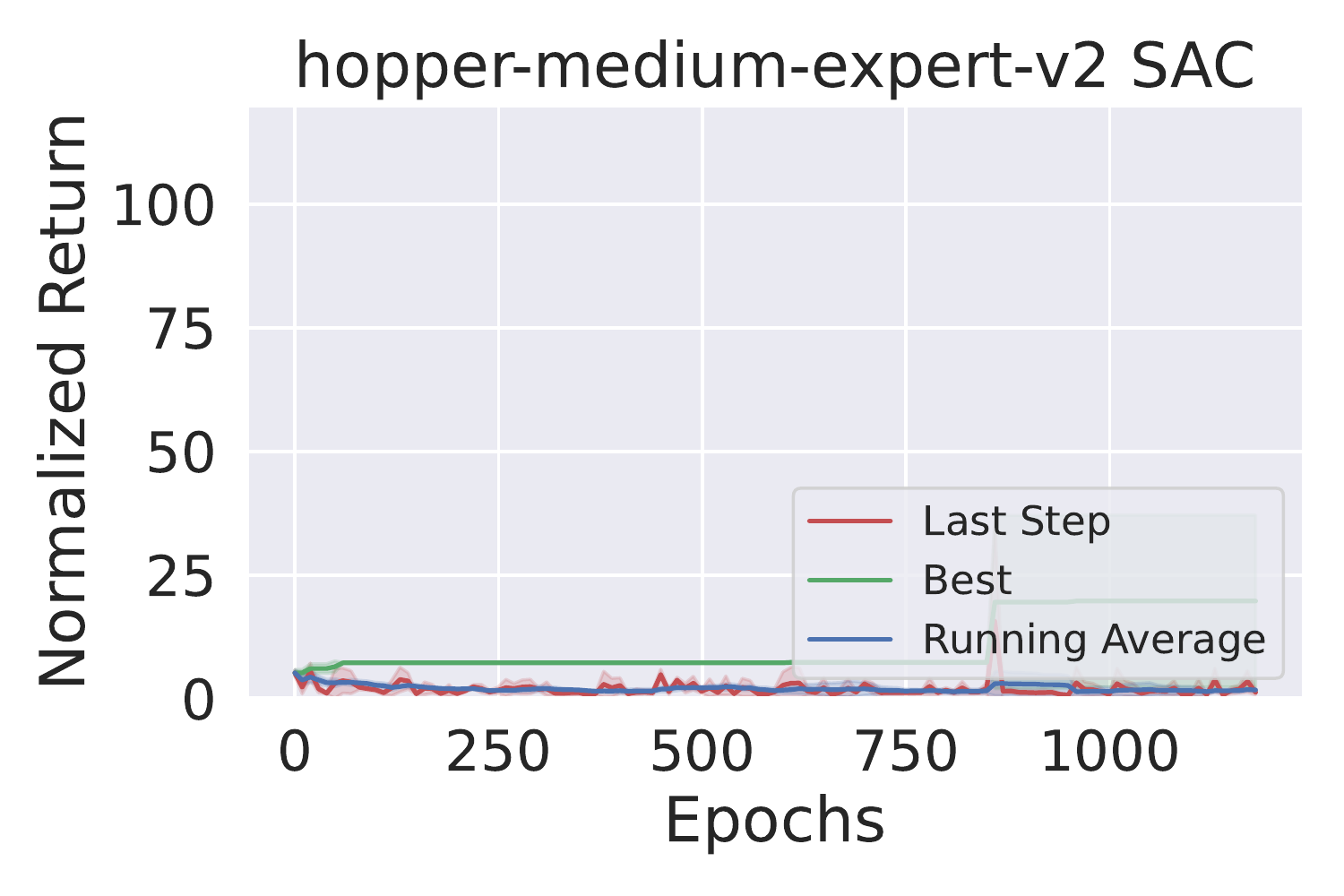}\\\includegraphics[width=0.225\linewidth]{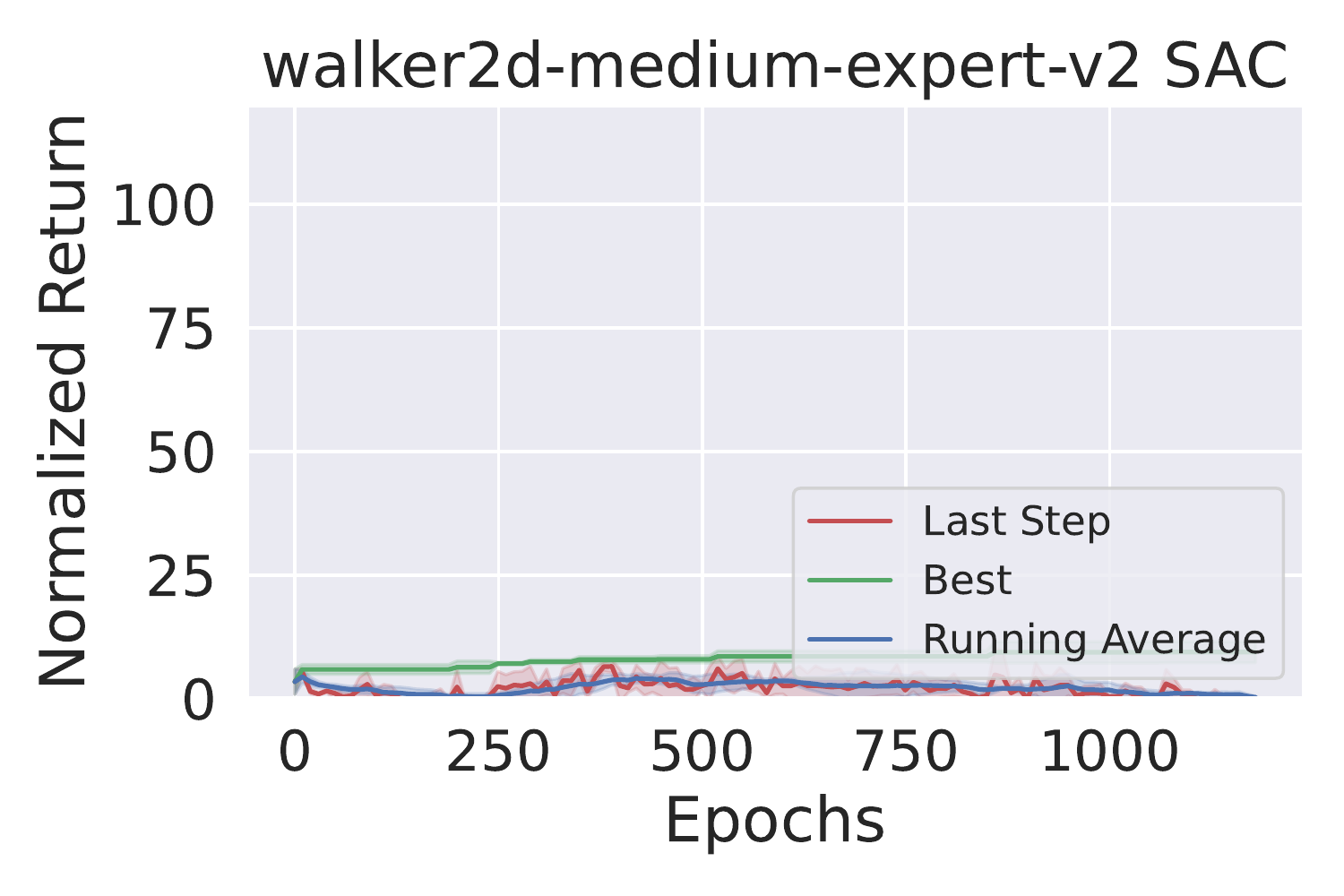}&\includegraphics[width=0.225\linewidth]{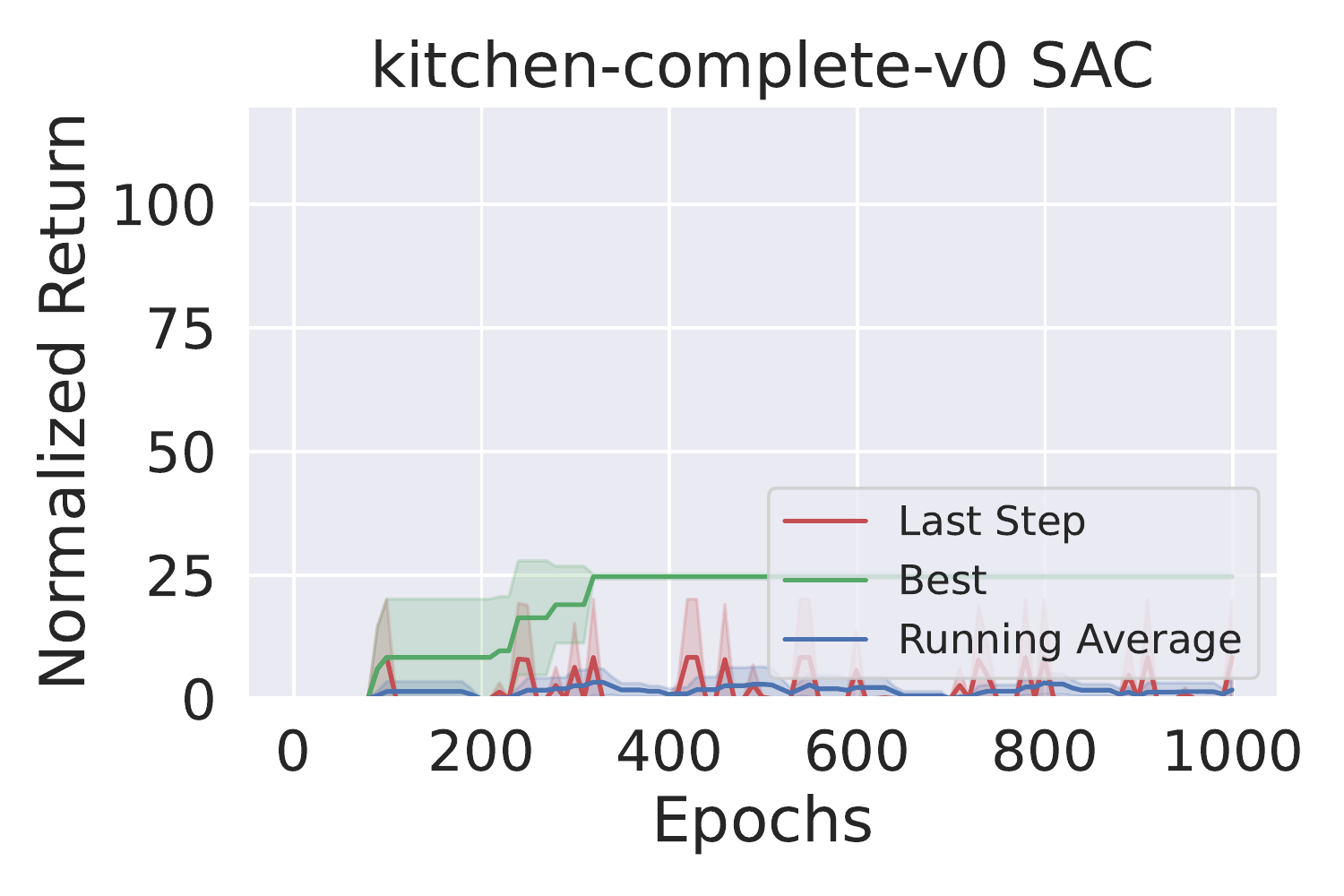}&\includegraphics[width=0.225\linewidth]{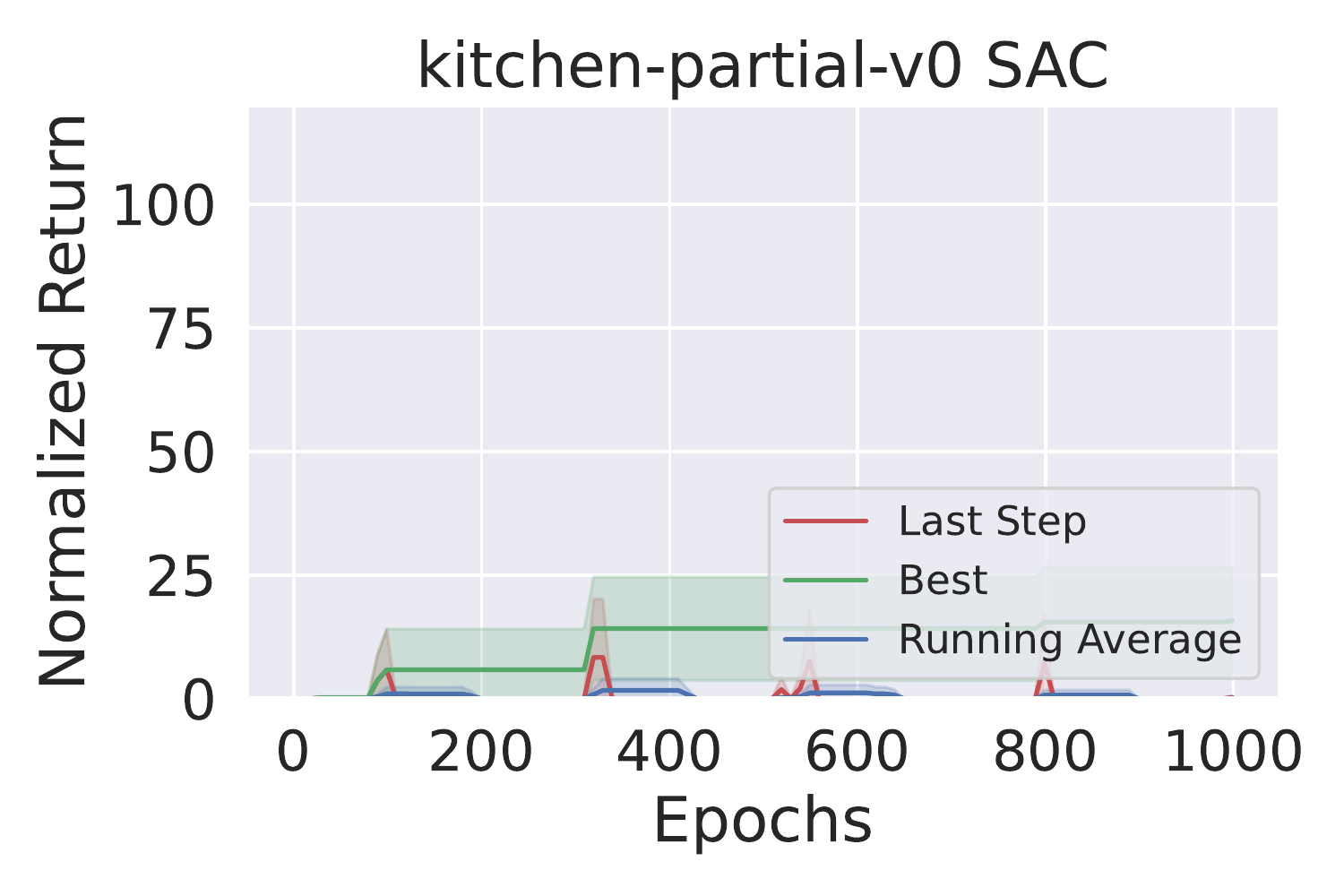}&\includegraphics[width=0.225\linewidth]{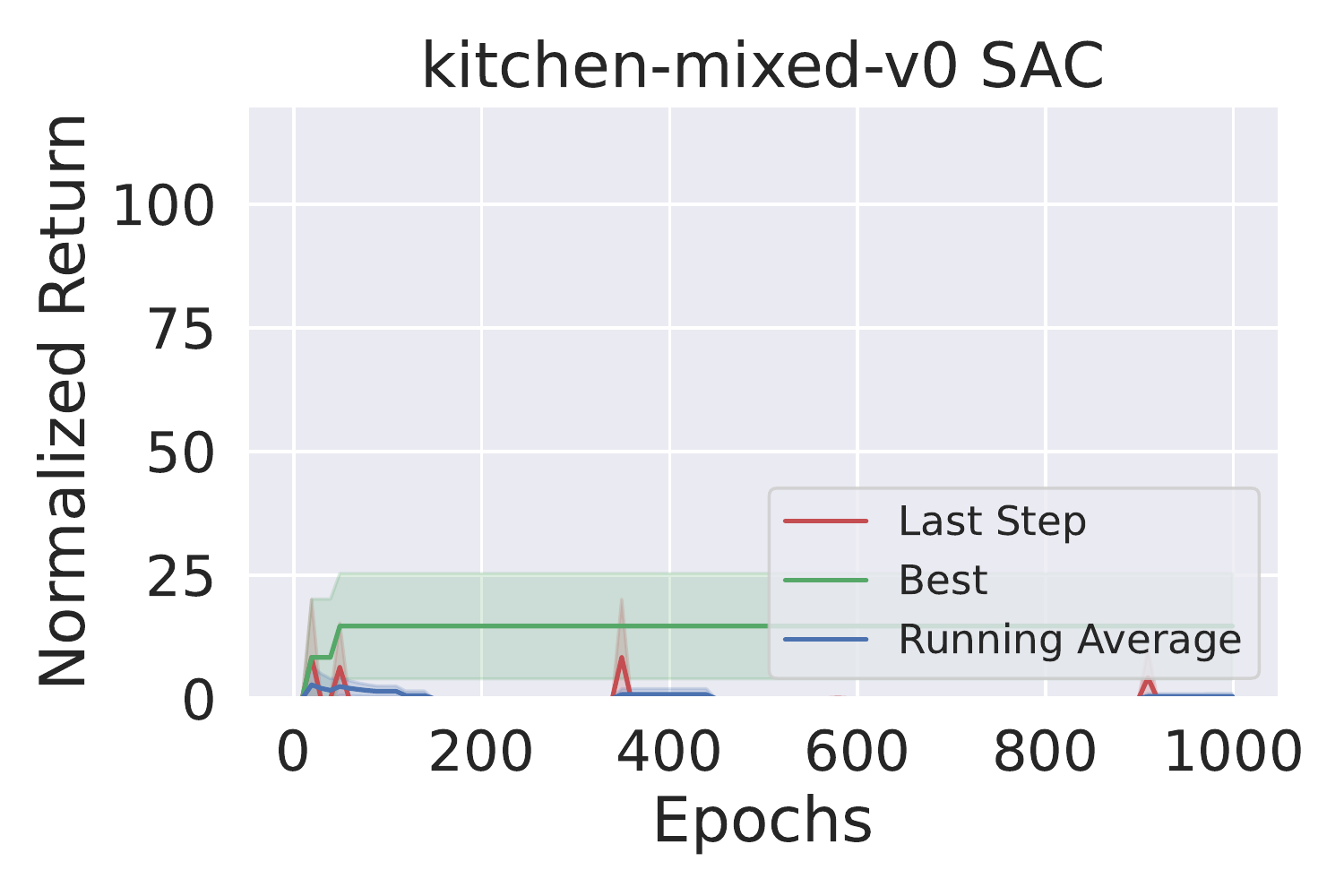}\\\includegraphics[width=0.225\linewidth]{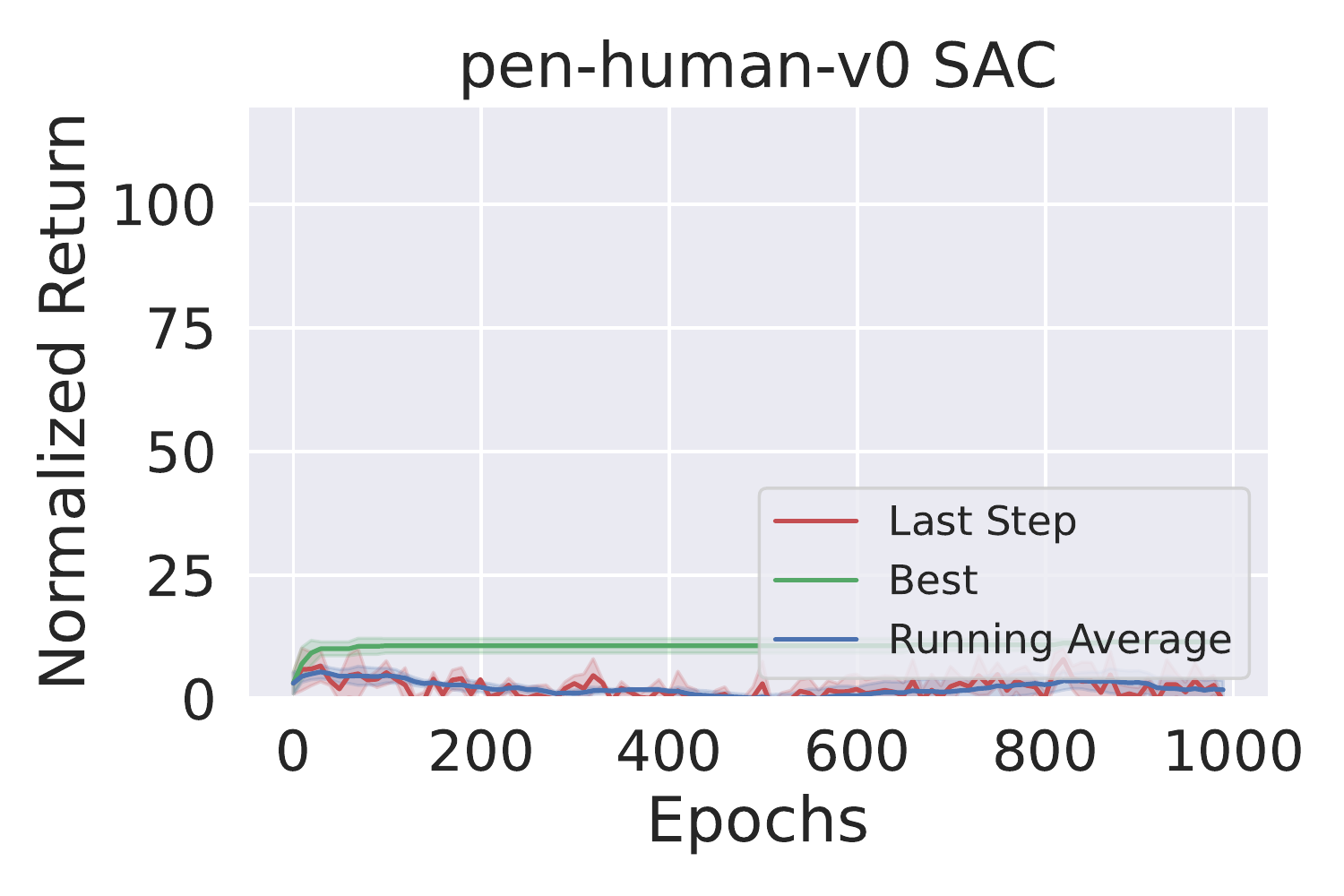}&\includegraphics[width=0.225\linewidth]{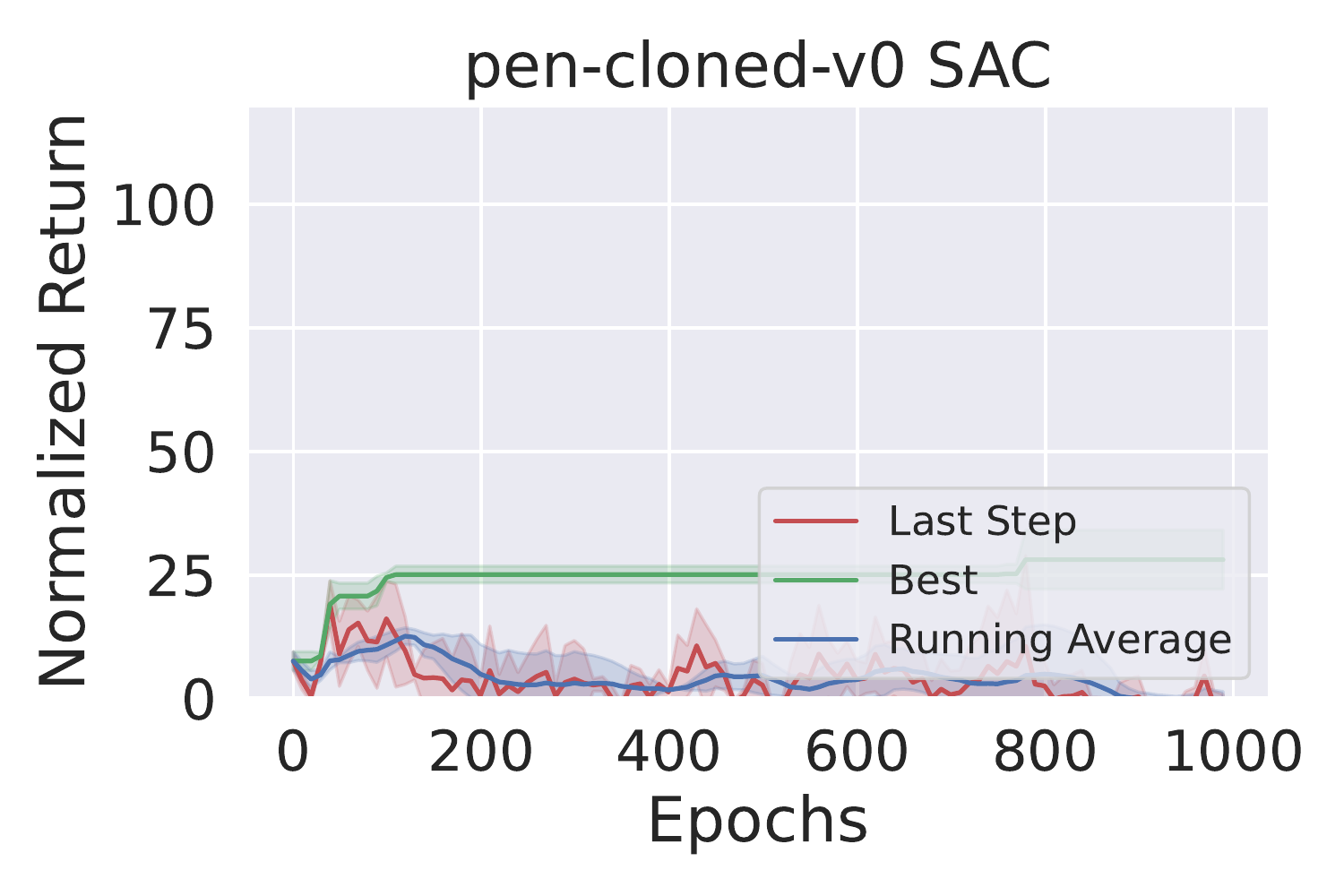}&\includegraphics[width=0.225\linewidth]{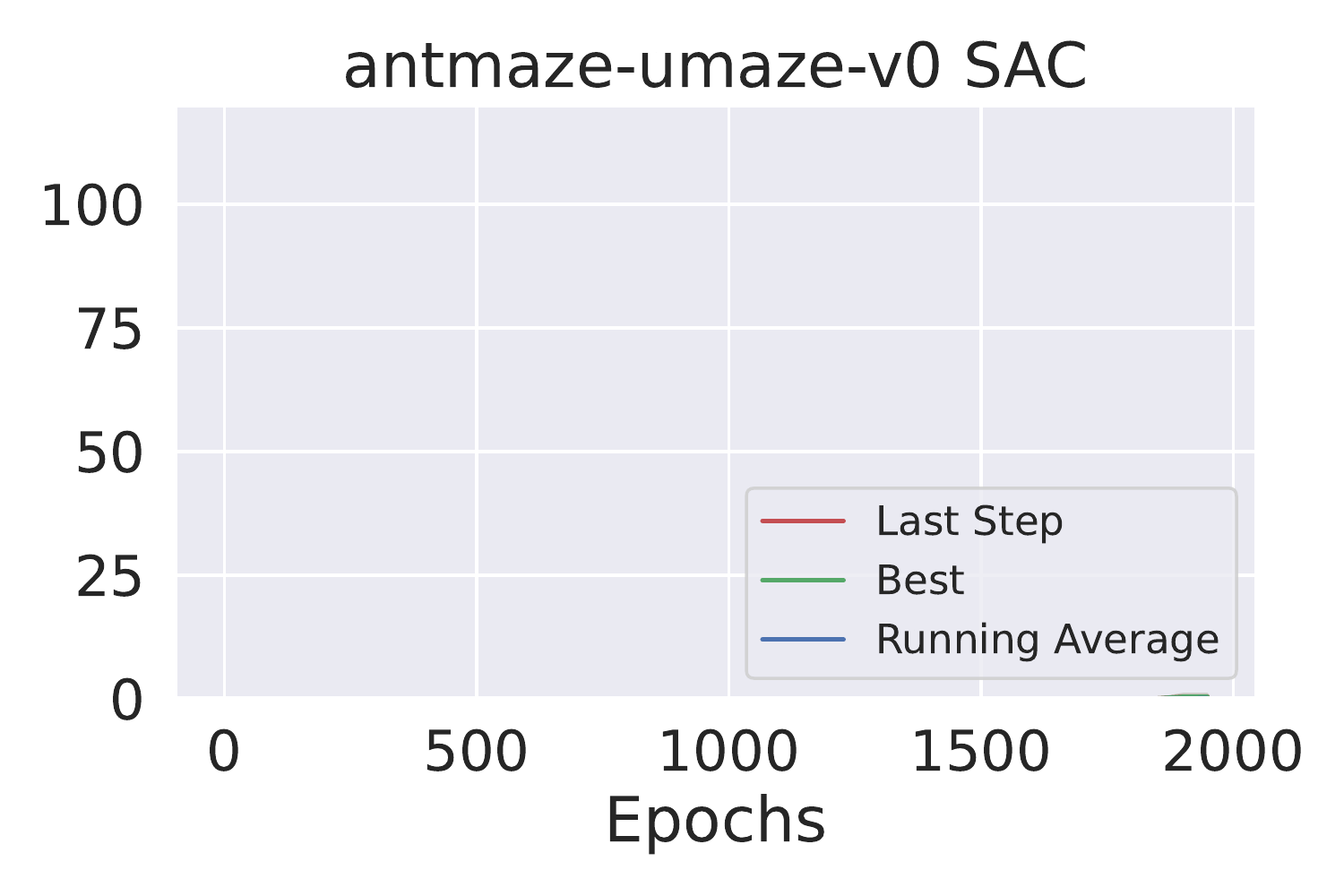}&\includegraphics[width=0.225\linewidth]{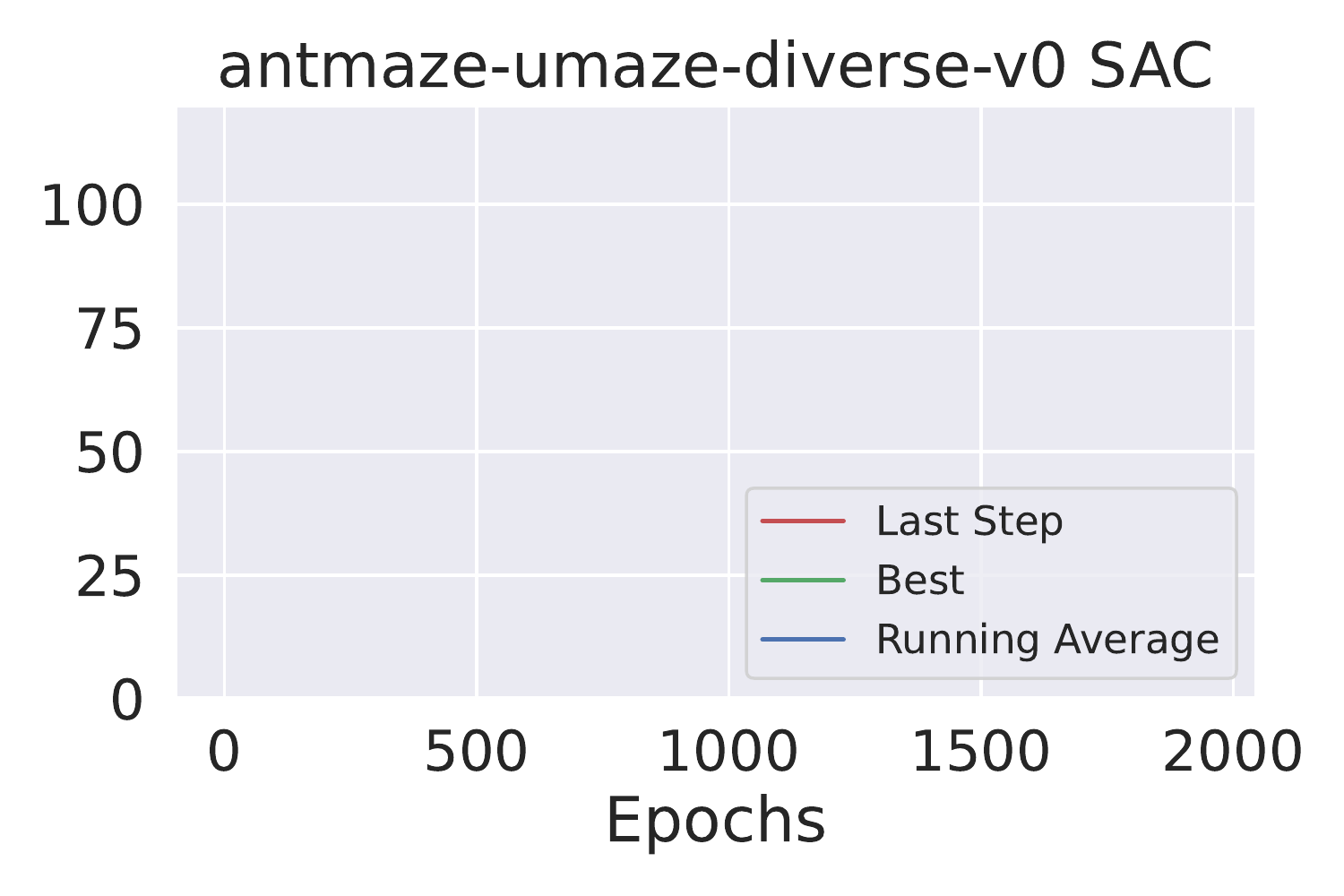}\\\includegraphics[width=0.225\linewidth]{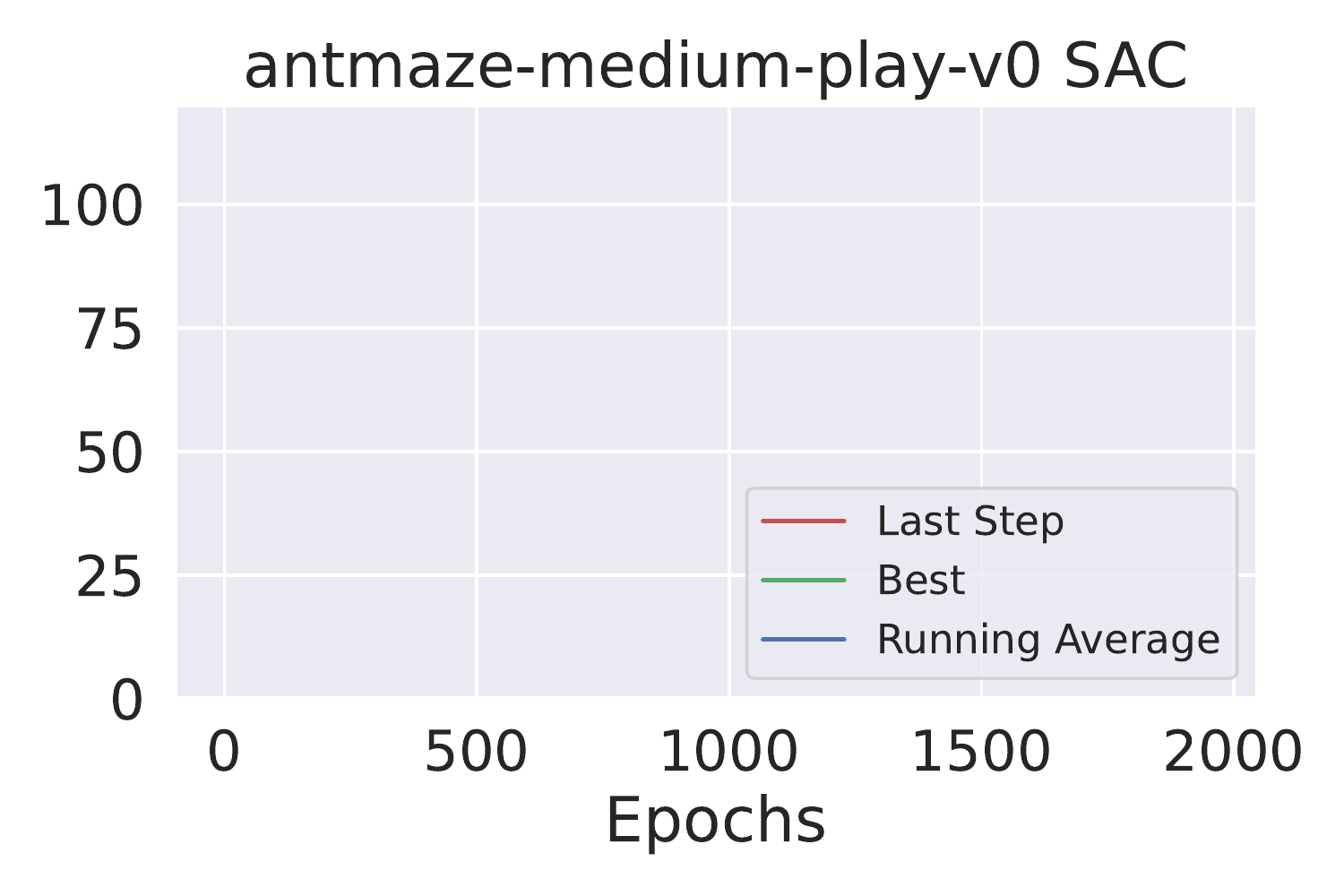}&\includegraphics[width=0.225\linewidth]{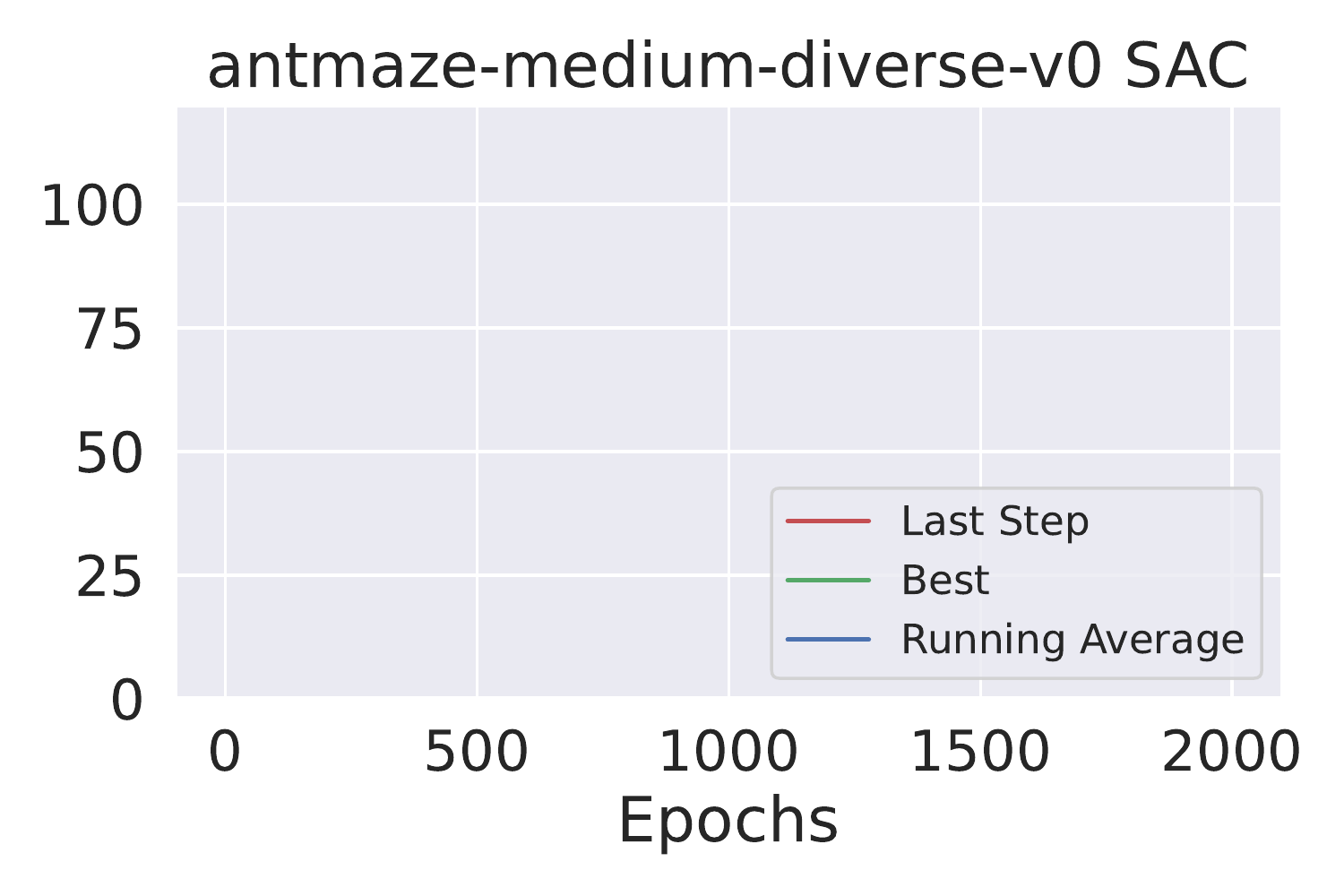}&\includegraphics[width=0.225\linewidth]{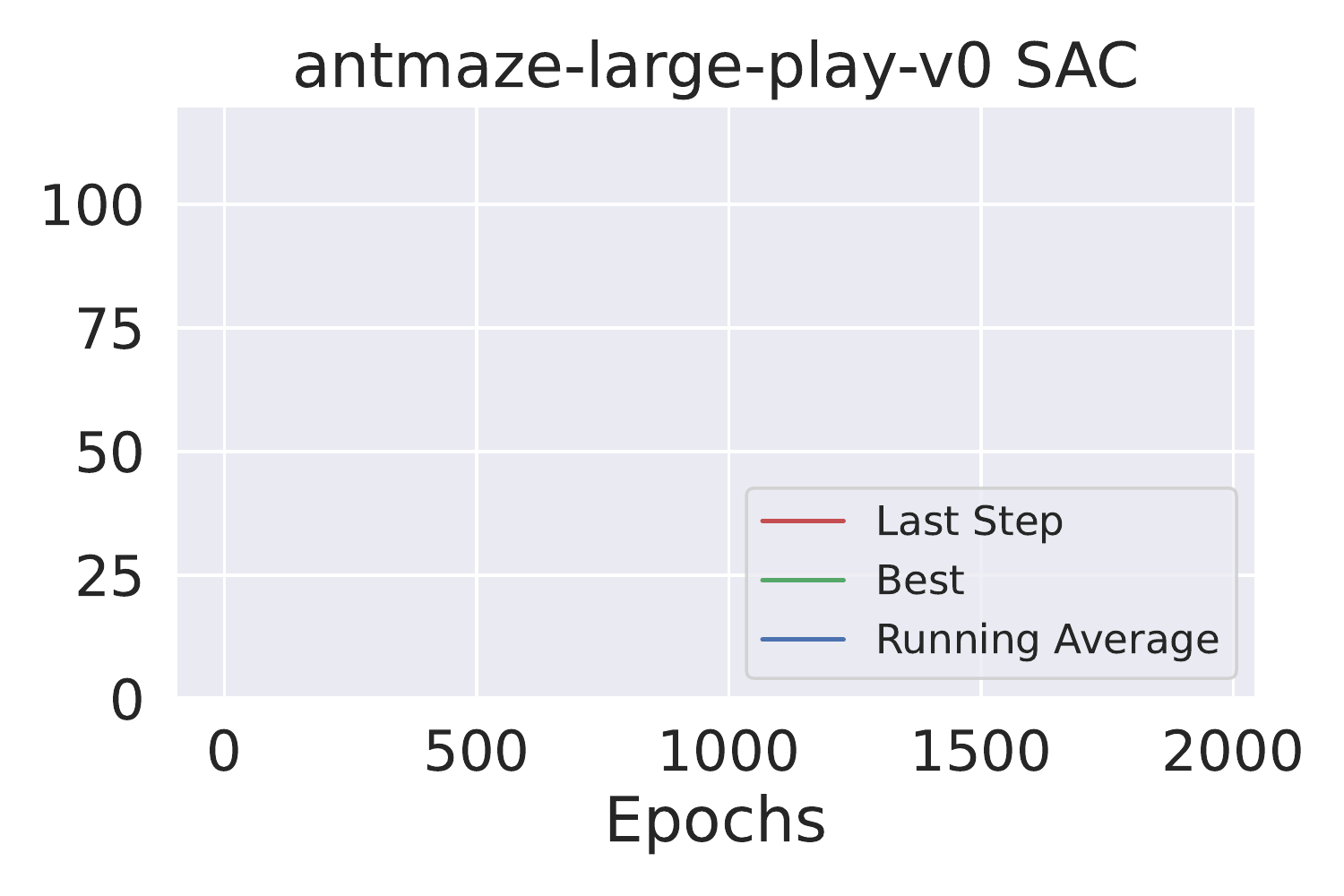}&\includegraphics[width=0.225\linewidth]{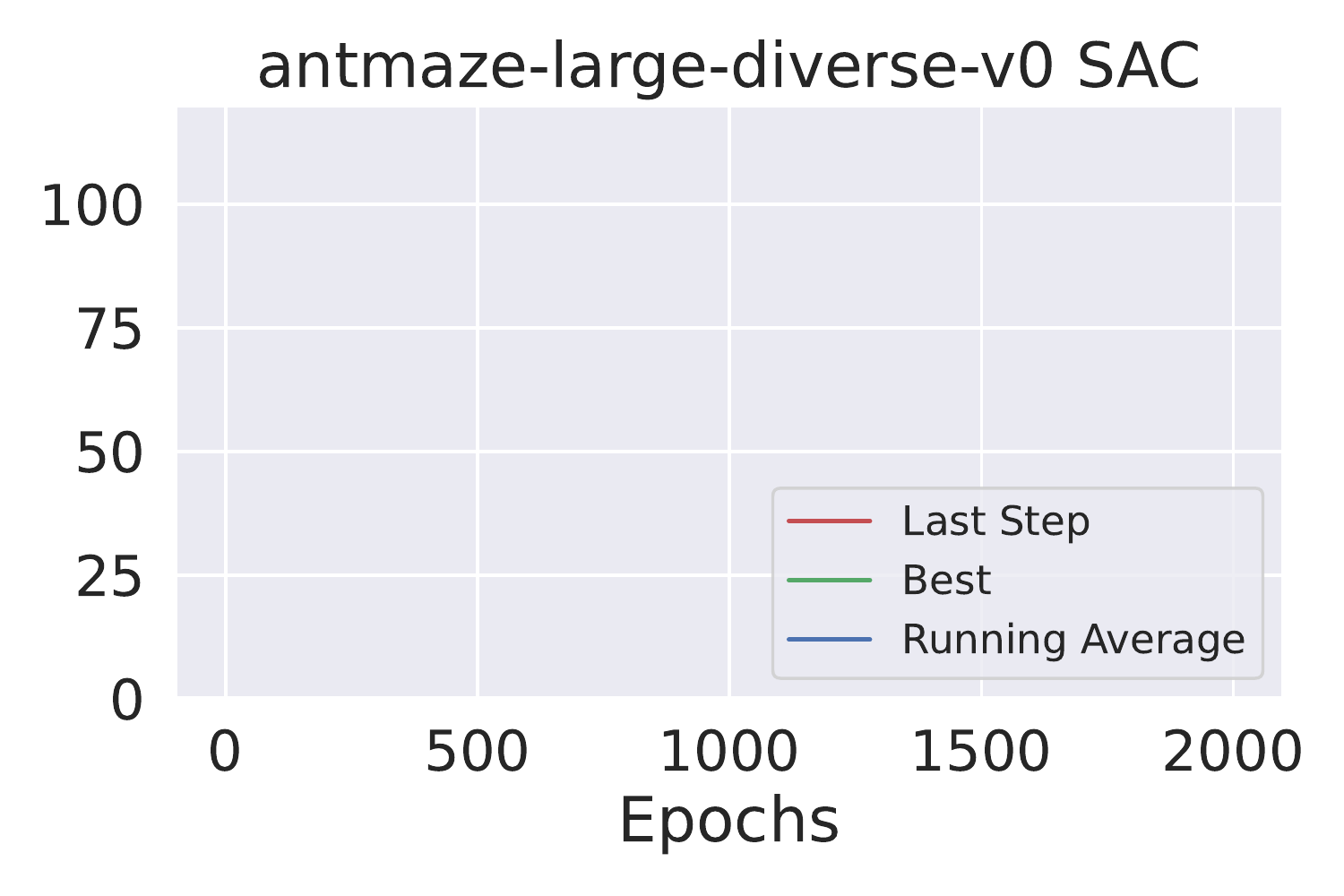}\\\end{tabular}
\centering
\caption{Training Curves of SAC on D4RL}
\end{figure*}
\begin{figure*}[htb]
\centering
\begin{tabular}{cccccc}
\includegraphics[width=0.225\linewidth]{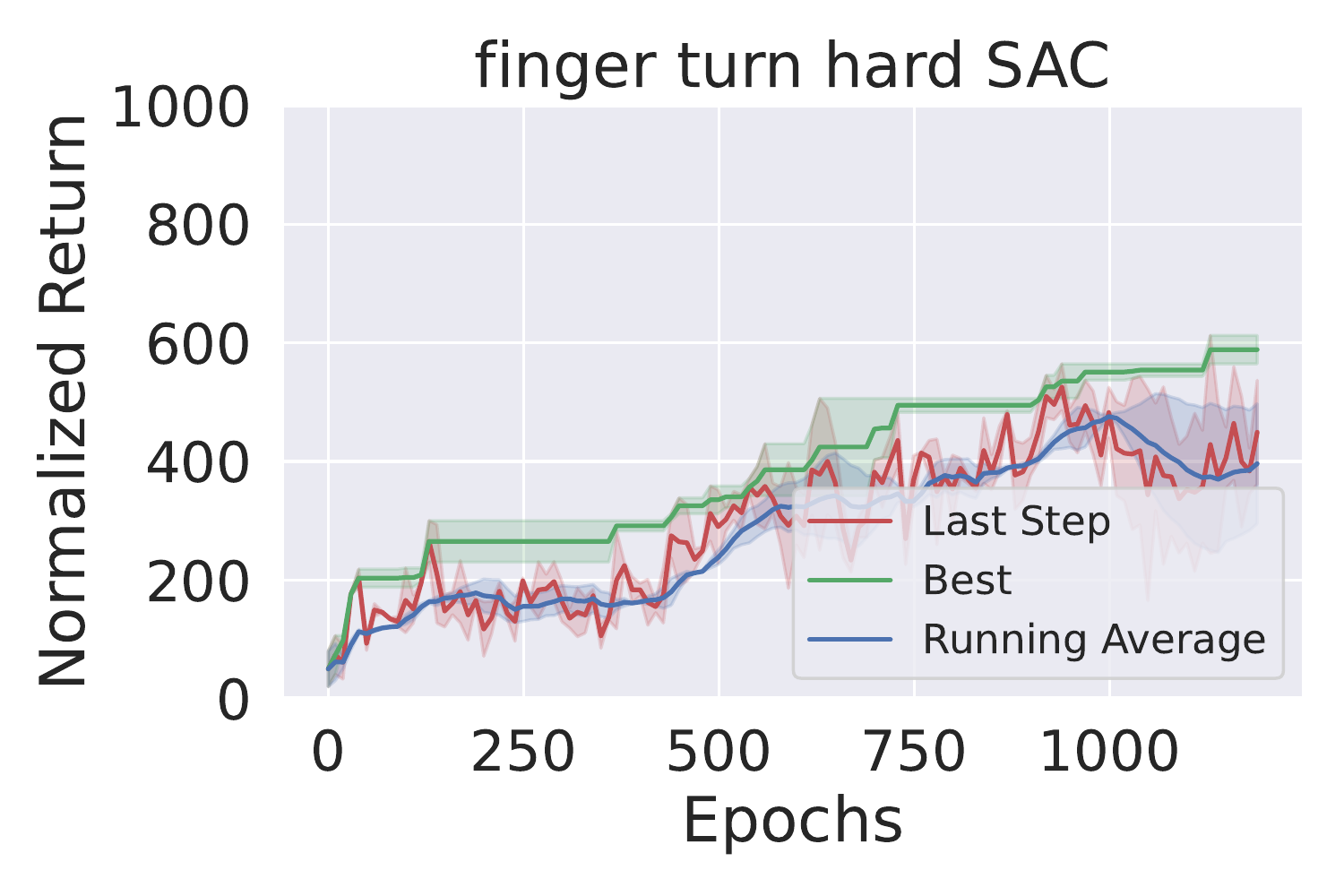}&\includegraphics[width=0.225\linewidth]{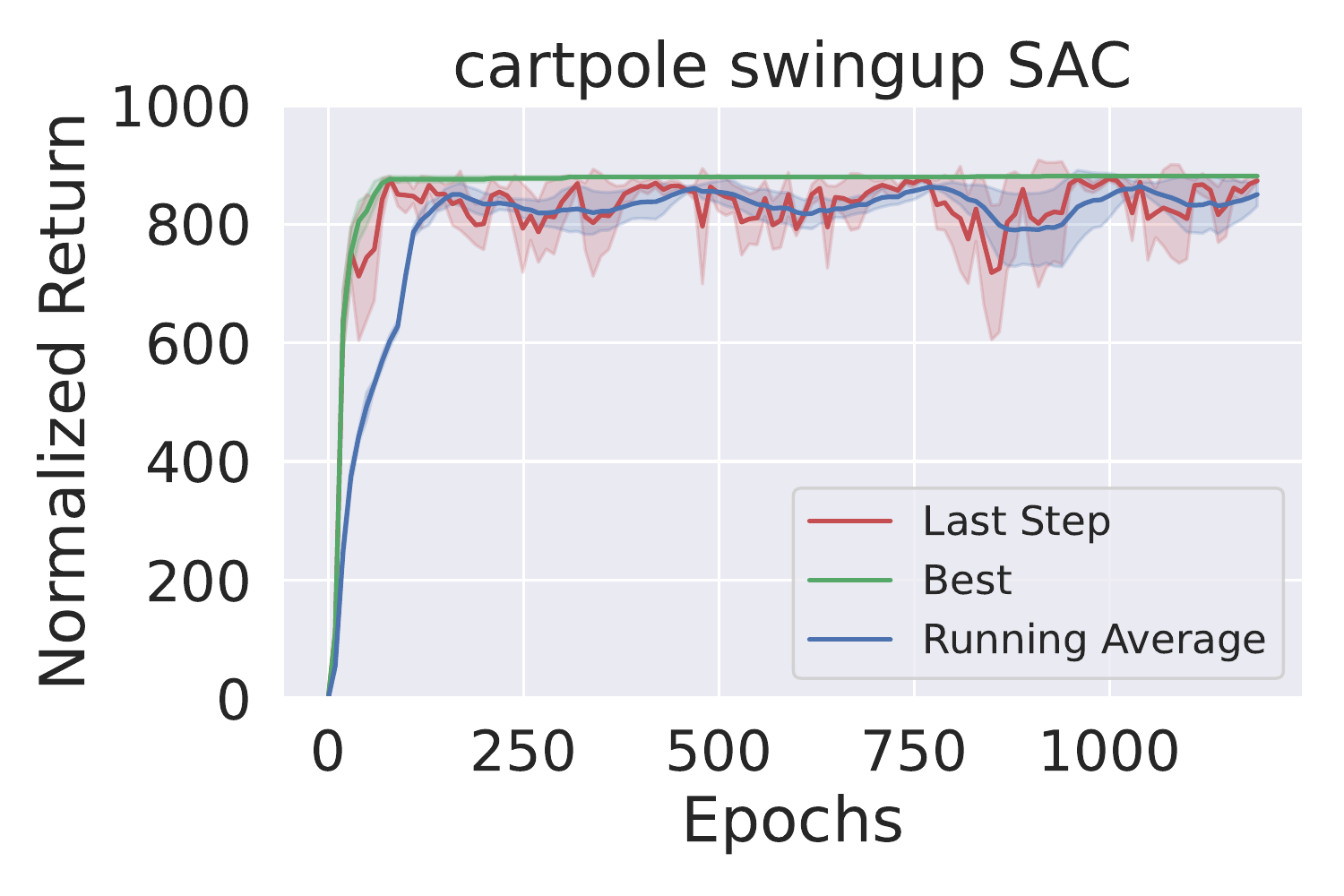}&\includegraphics[width=0.225\linewidth]{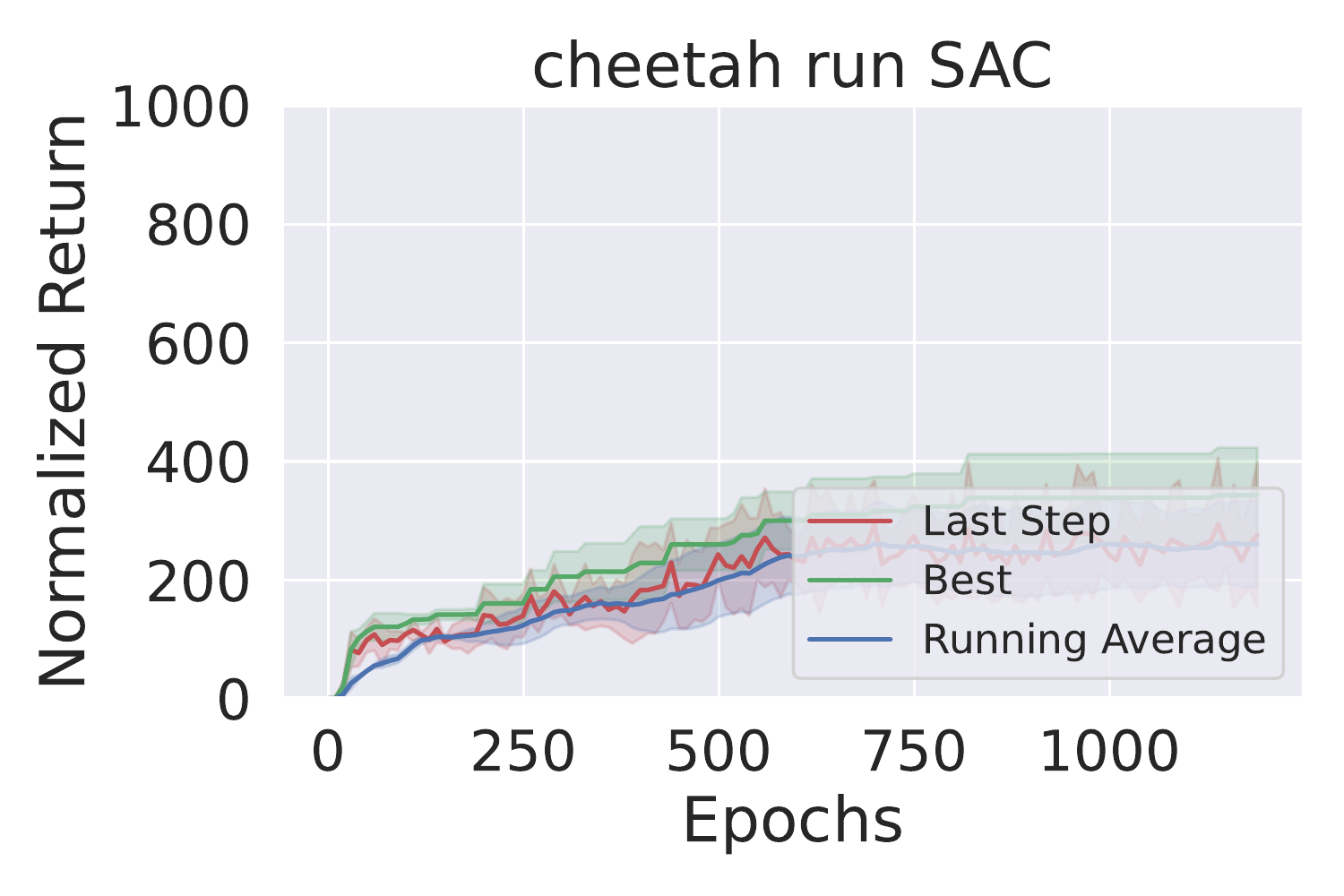}&\includegraphics[width=0.225\linewidth]{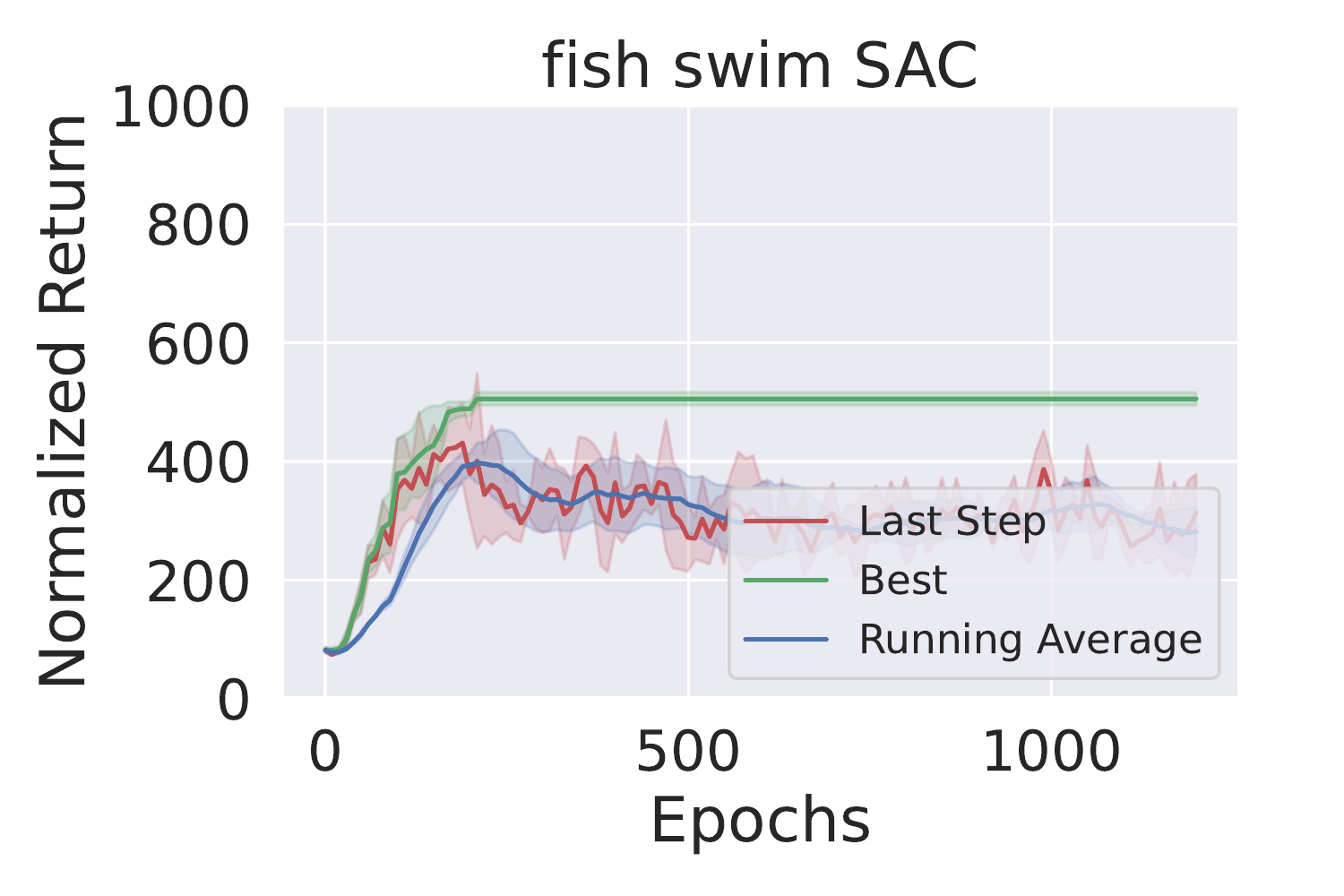}\\\includegraphics[width=0.225\linewidth]{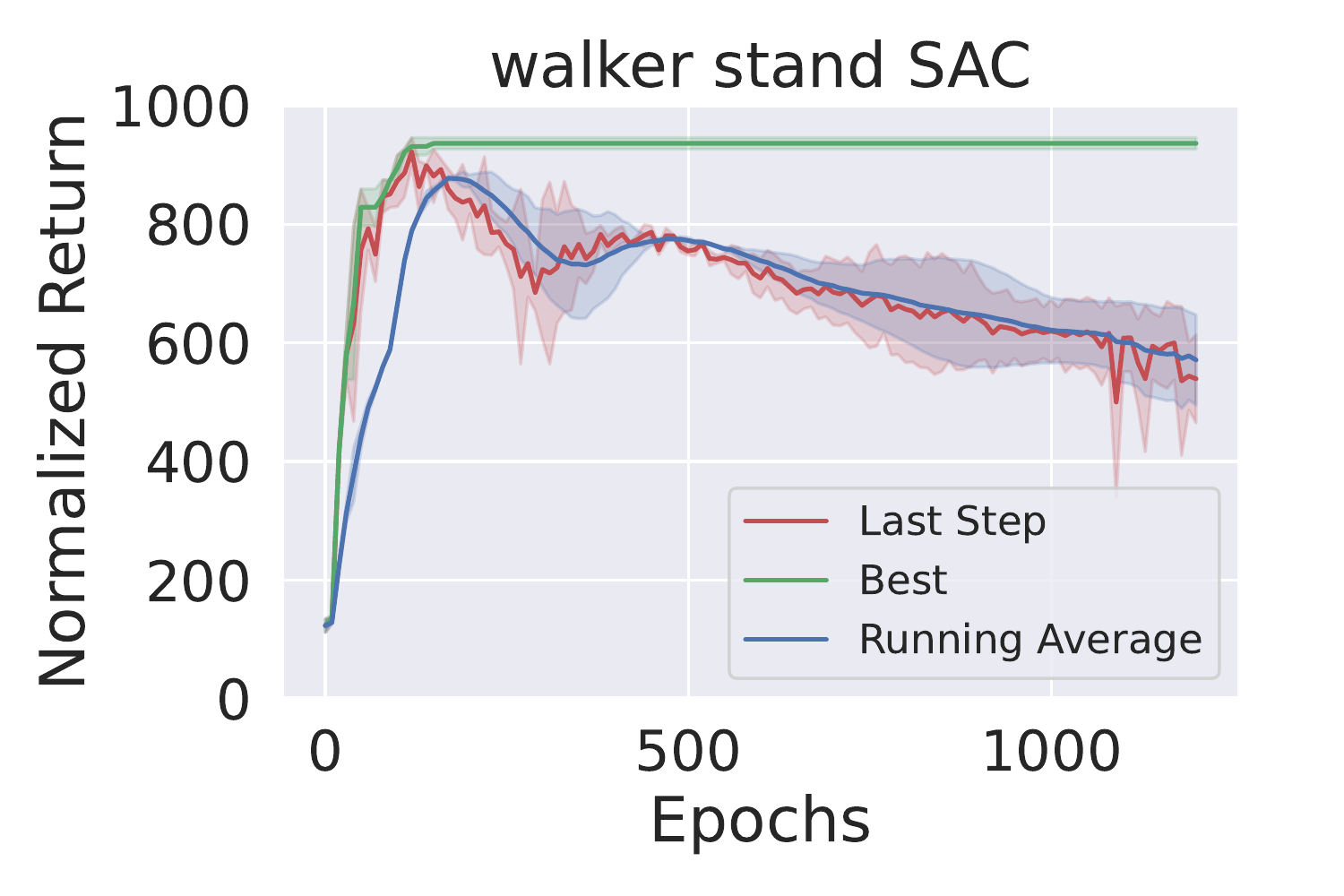}&\includegraphics[width=0.225\linewidth]{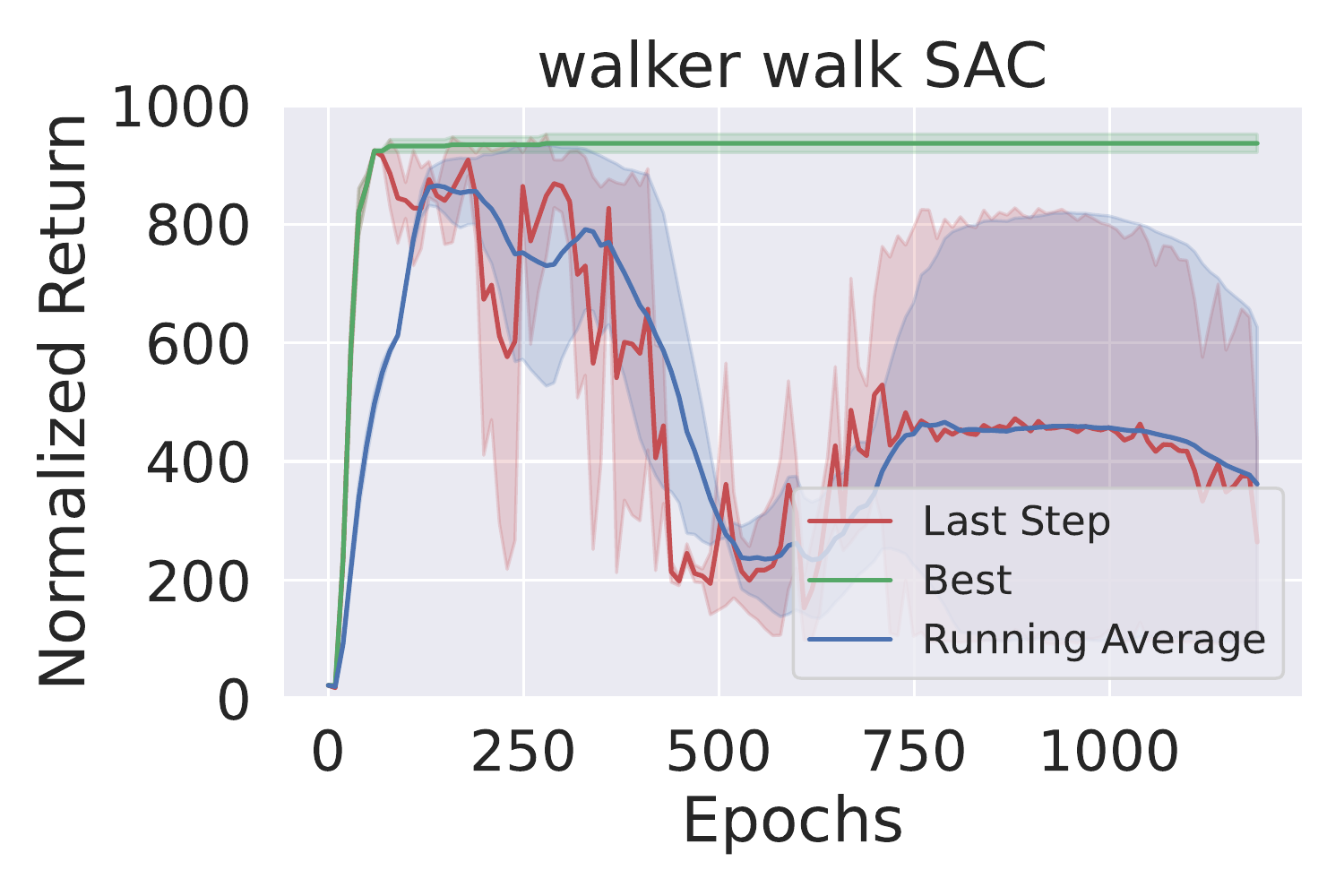}&&\\\end{tabular}
\centering
\caption{Training Curves of SAC on RLUP}
\end{figure*}
\begin{figure*}[htb]
\centering
\begin{tabular}{cccccc}
\includegraphics[width=0.225\linewidth]{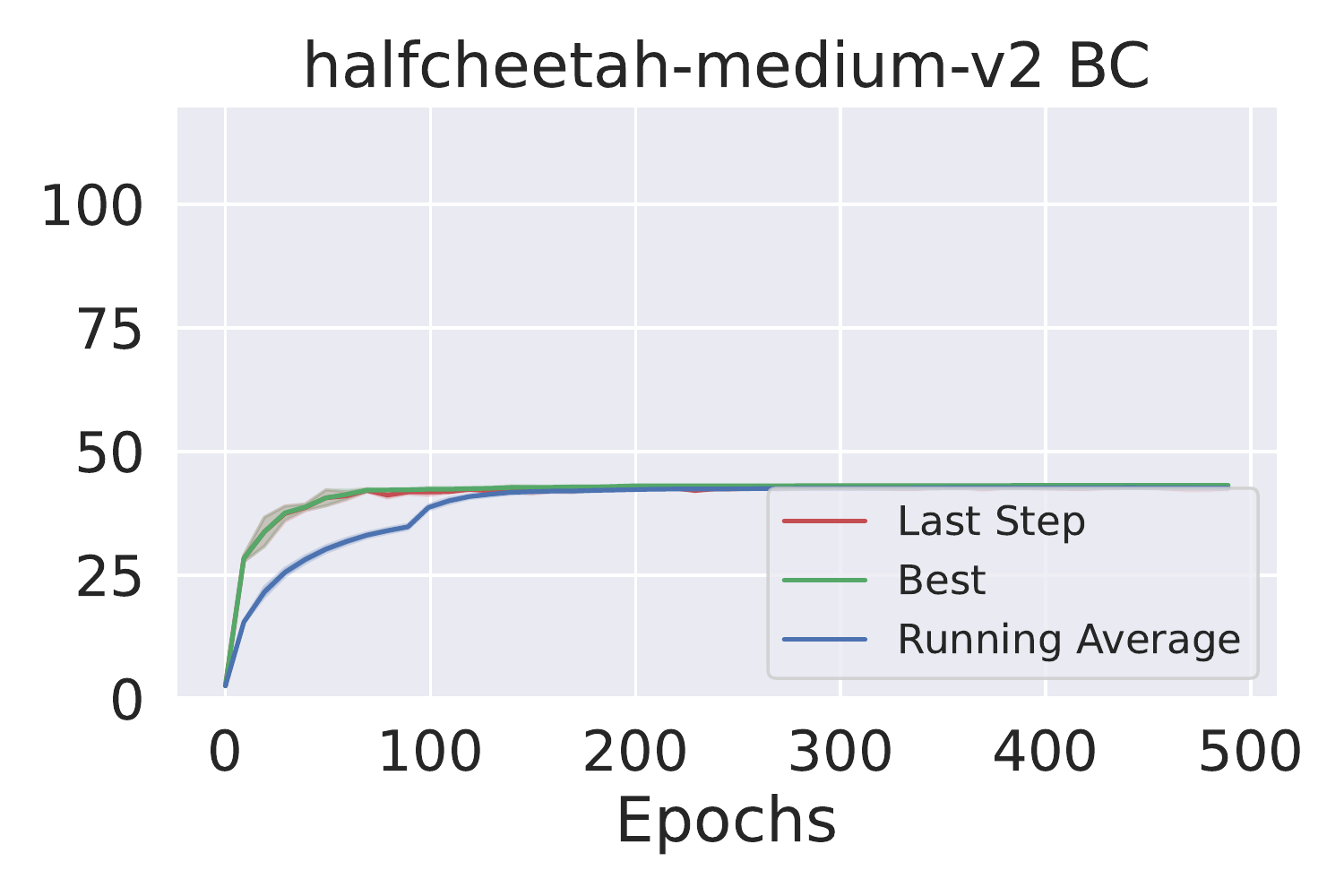}&\includegraphics[width=0.225\linewidth]{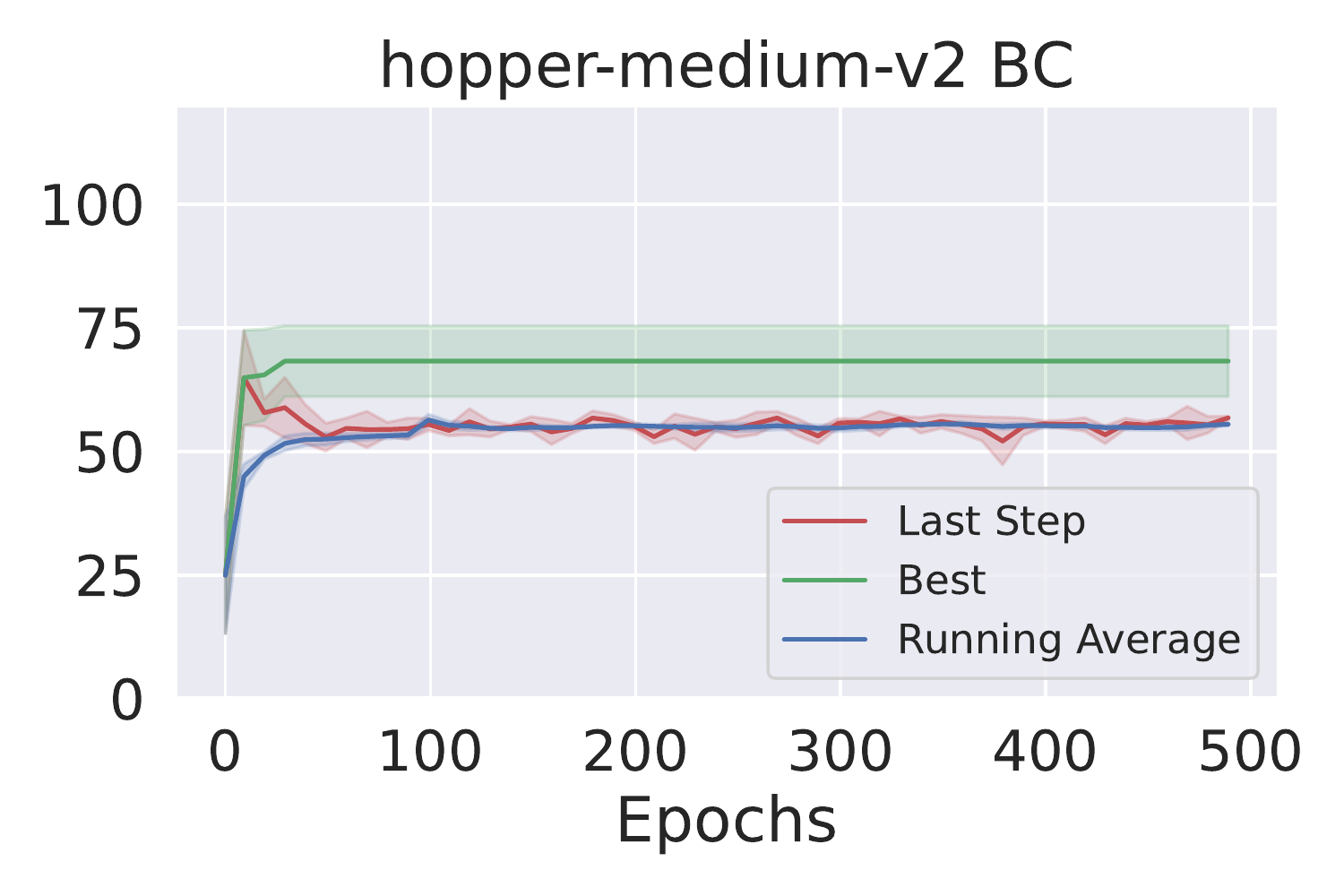}&\includegraphics[width=0.225\linewidth]{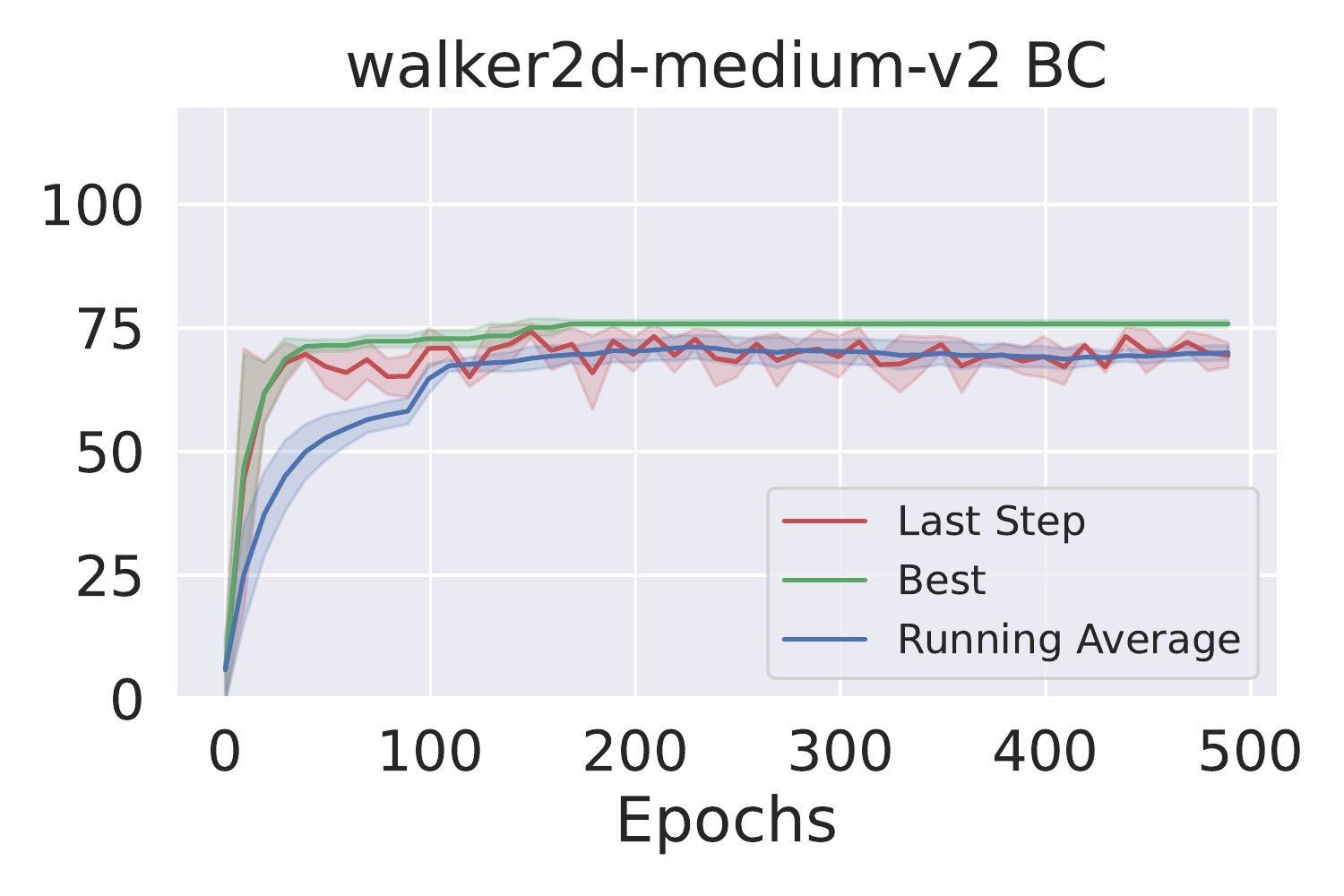}&\includegraphics[width=0.225\linewidth]{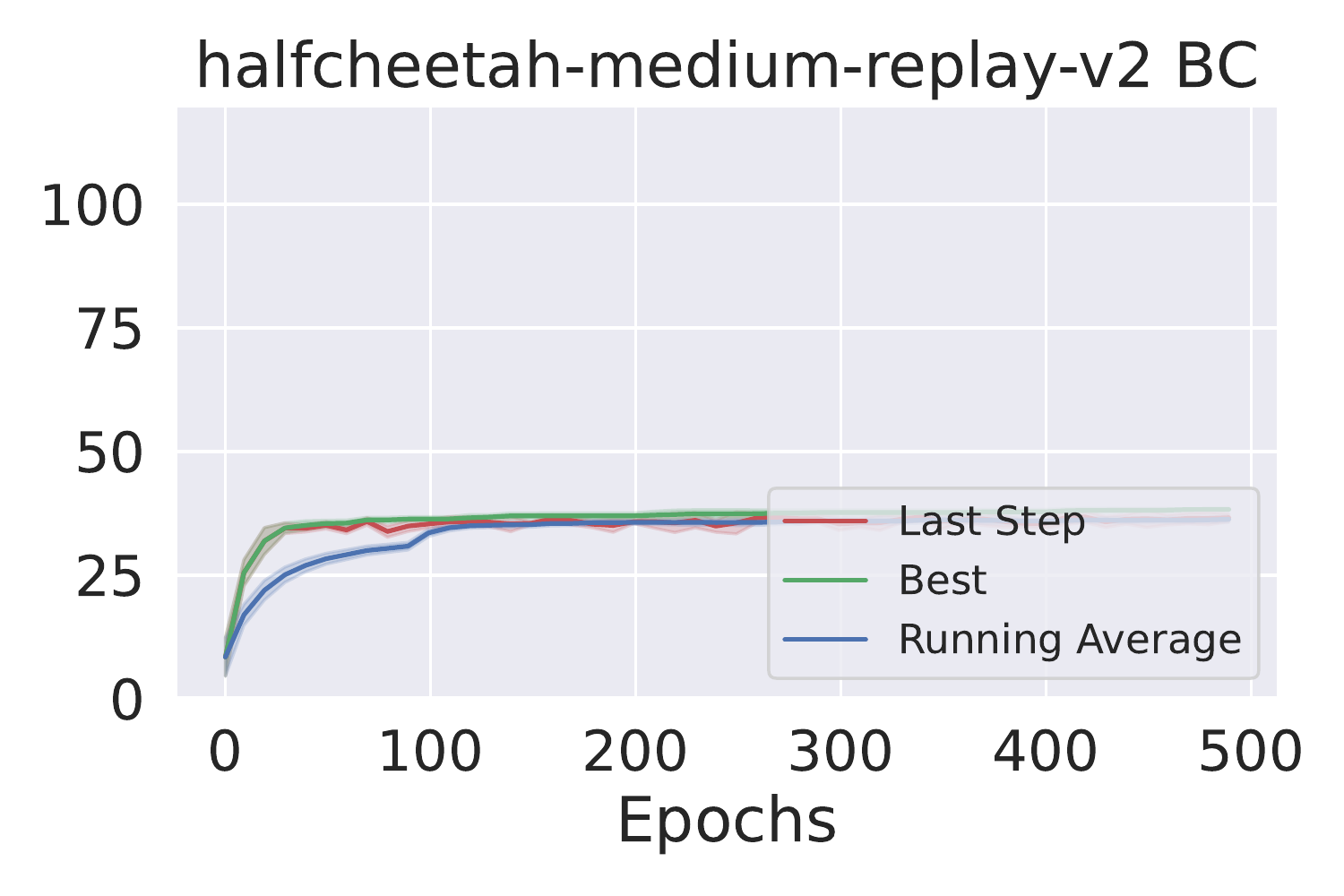}\\\includegraphics[width=0.225\linewidth]{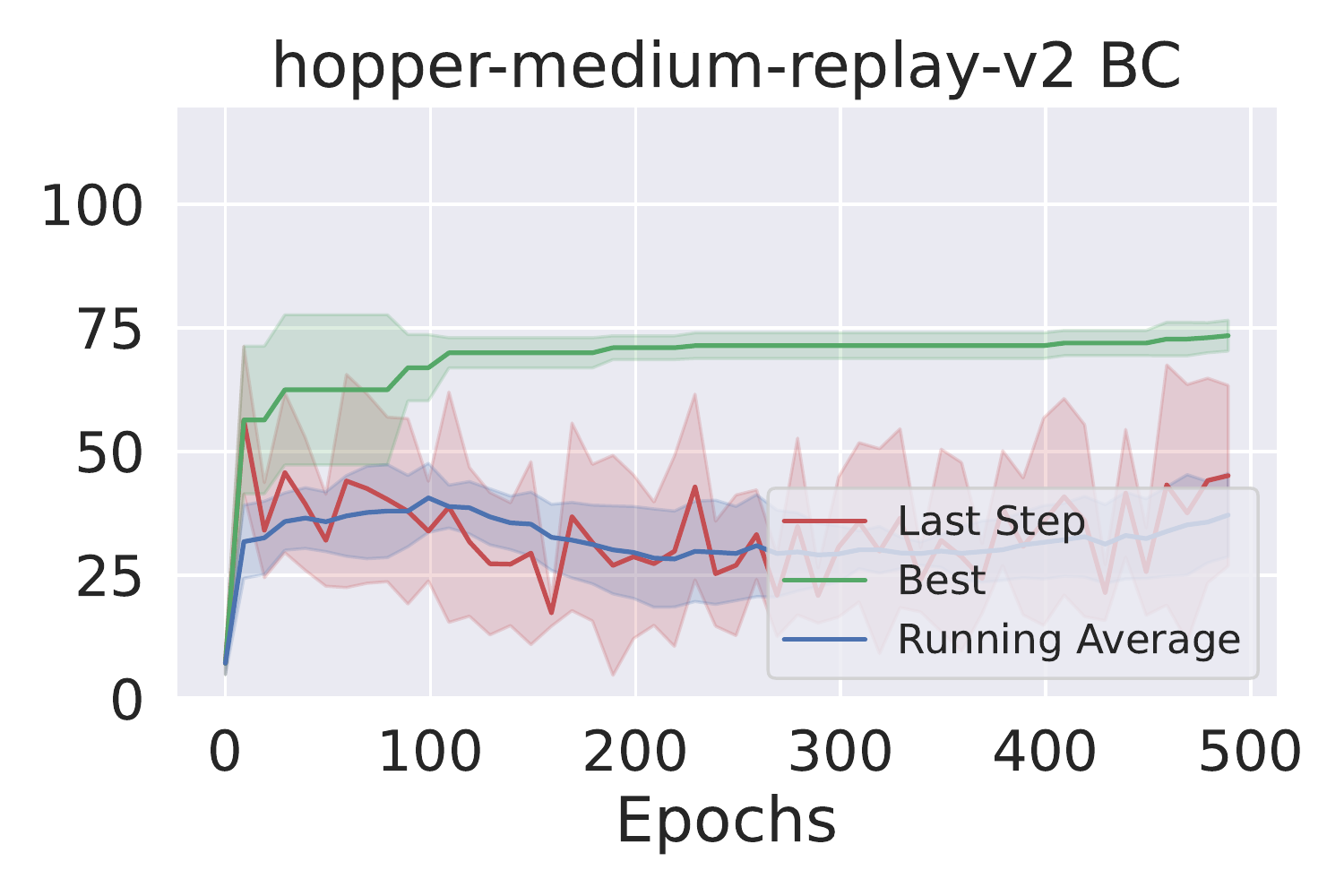}&\includegraphics[width=0.225\linewidth]{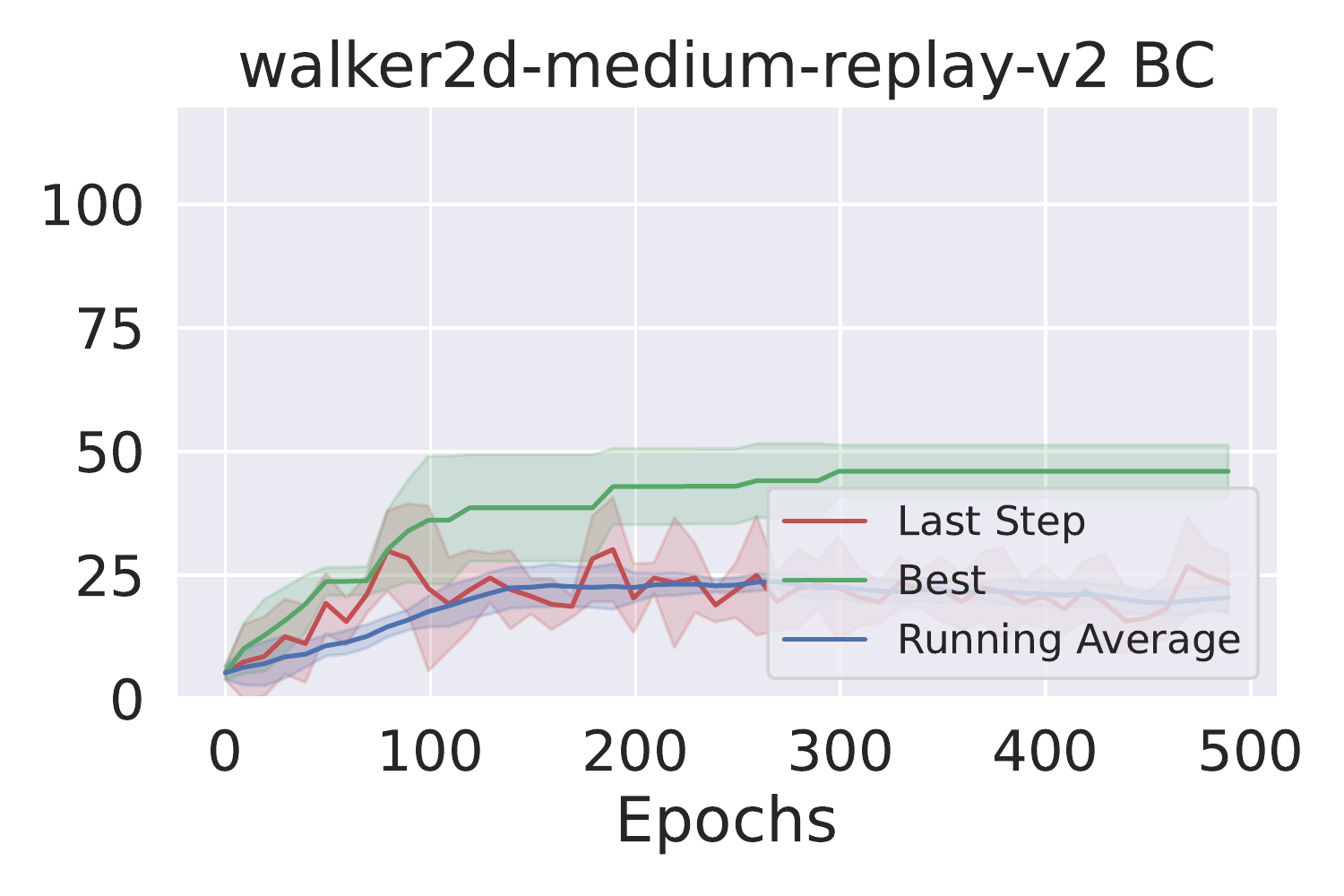}&\includegraphics[width=0.225\linewidth]{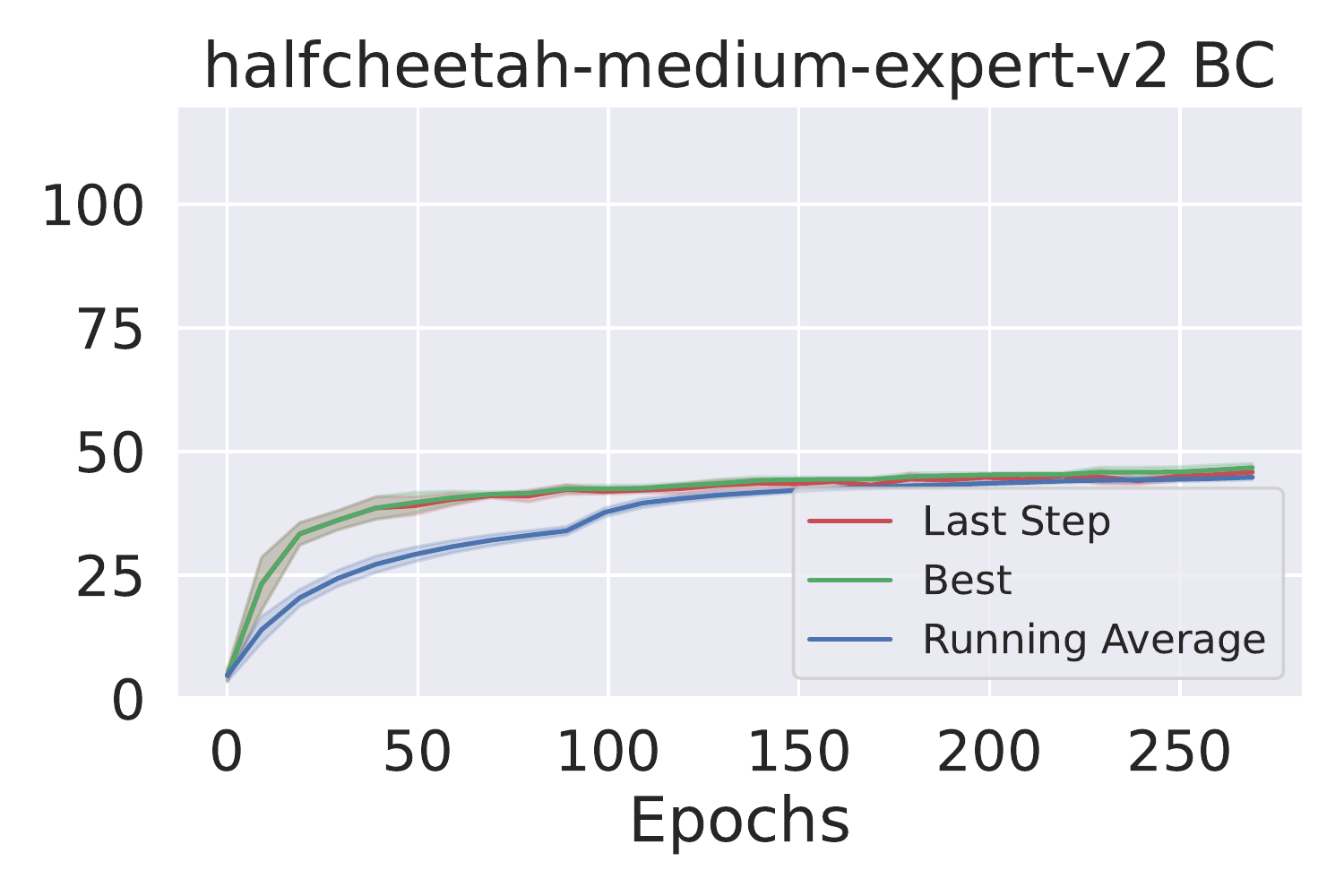}&\includegraphics[width=0.225\linewidth]{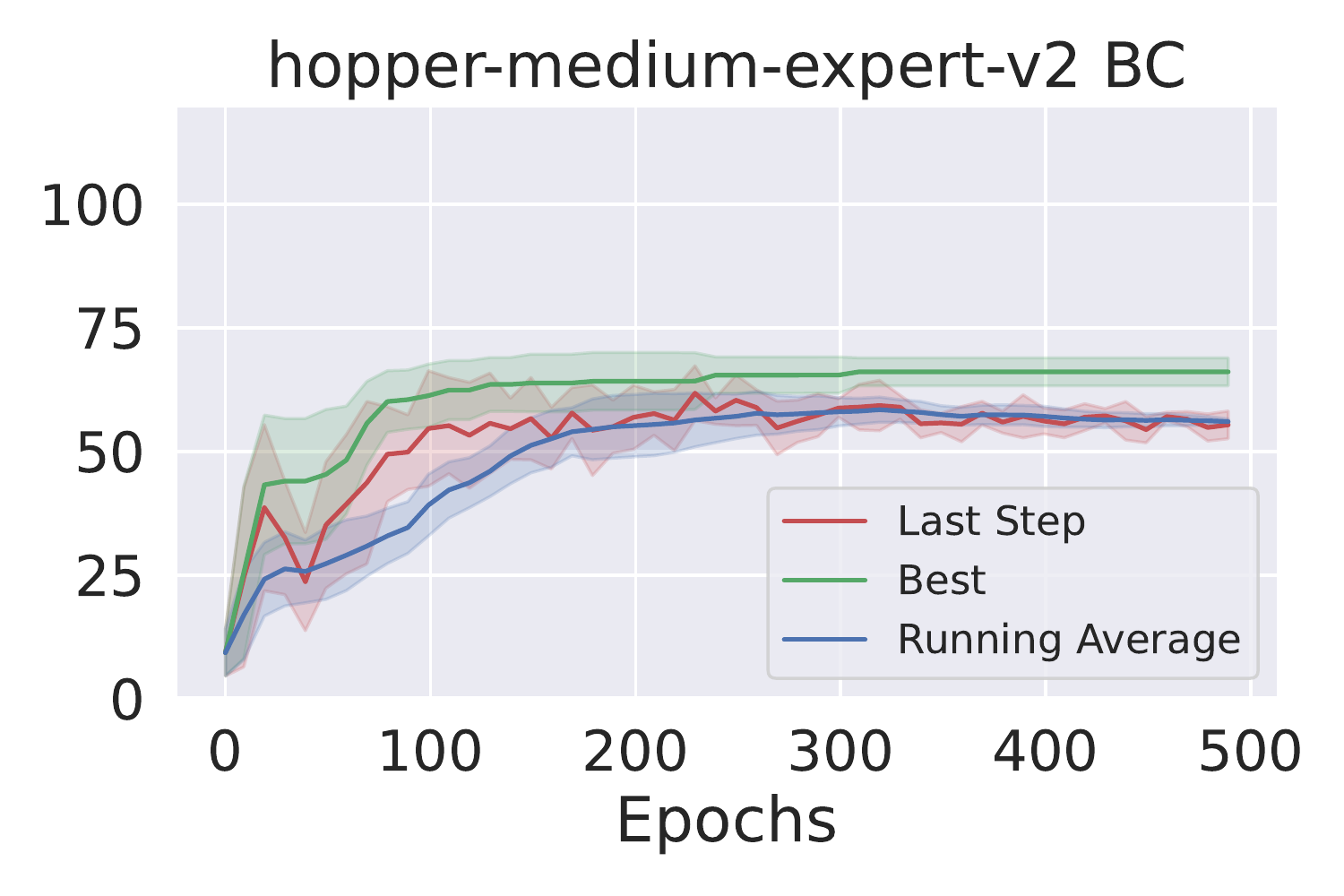}\\\includegraphics[width=0.225\linewidth]{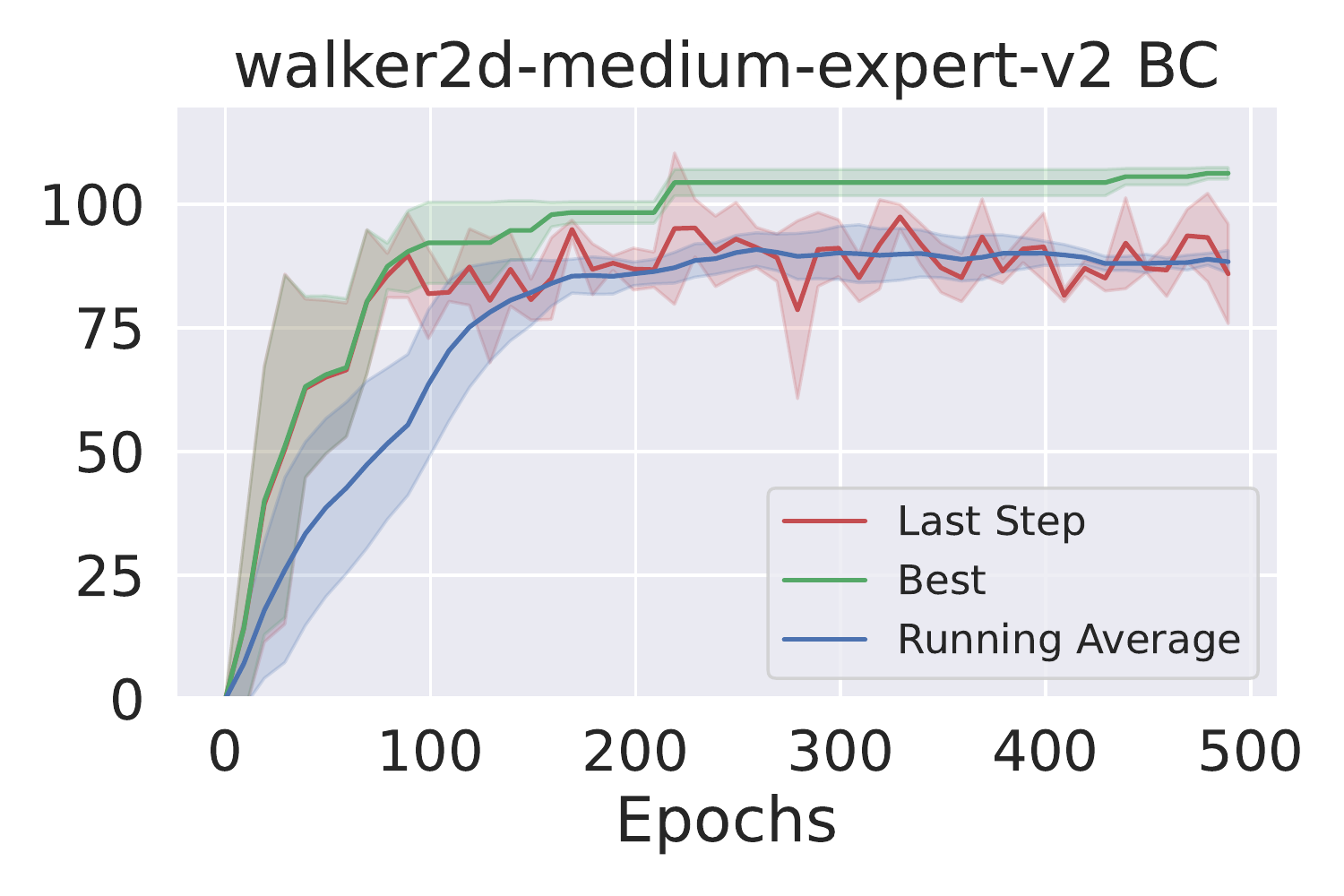}&\includegraphics[width=0.225\linewidth]{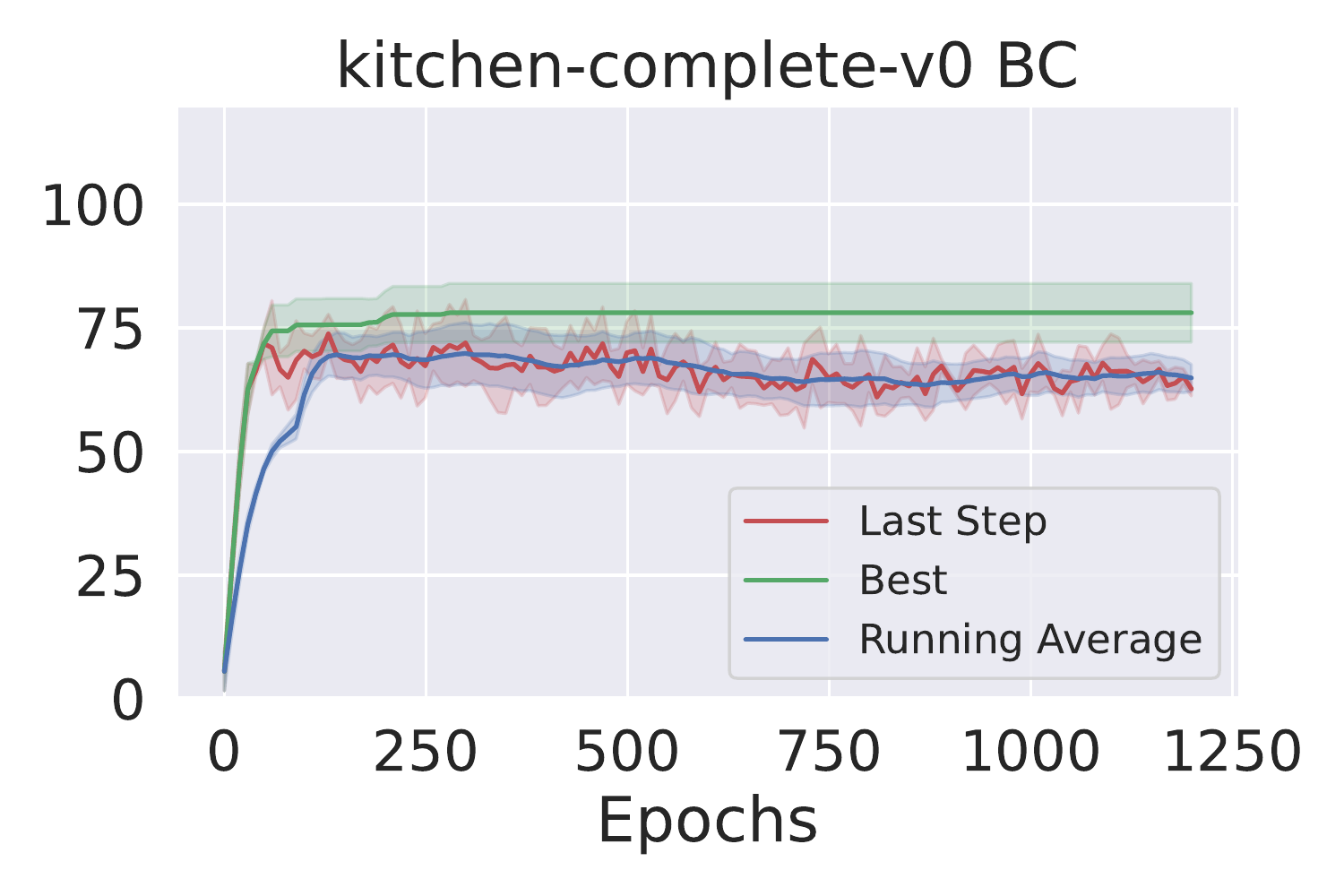}&\includegraphics[width=0.225\linewidth]{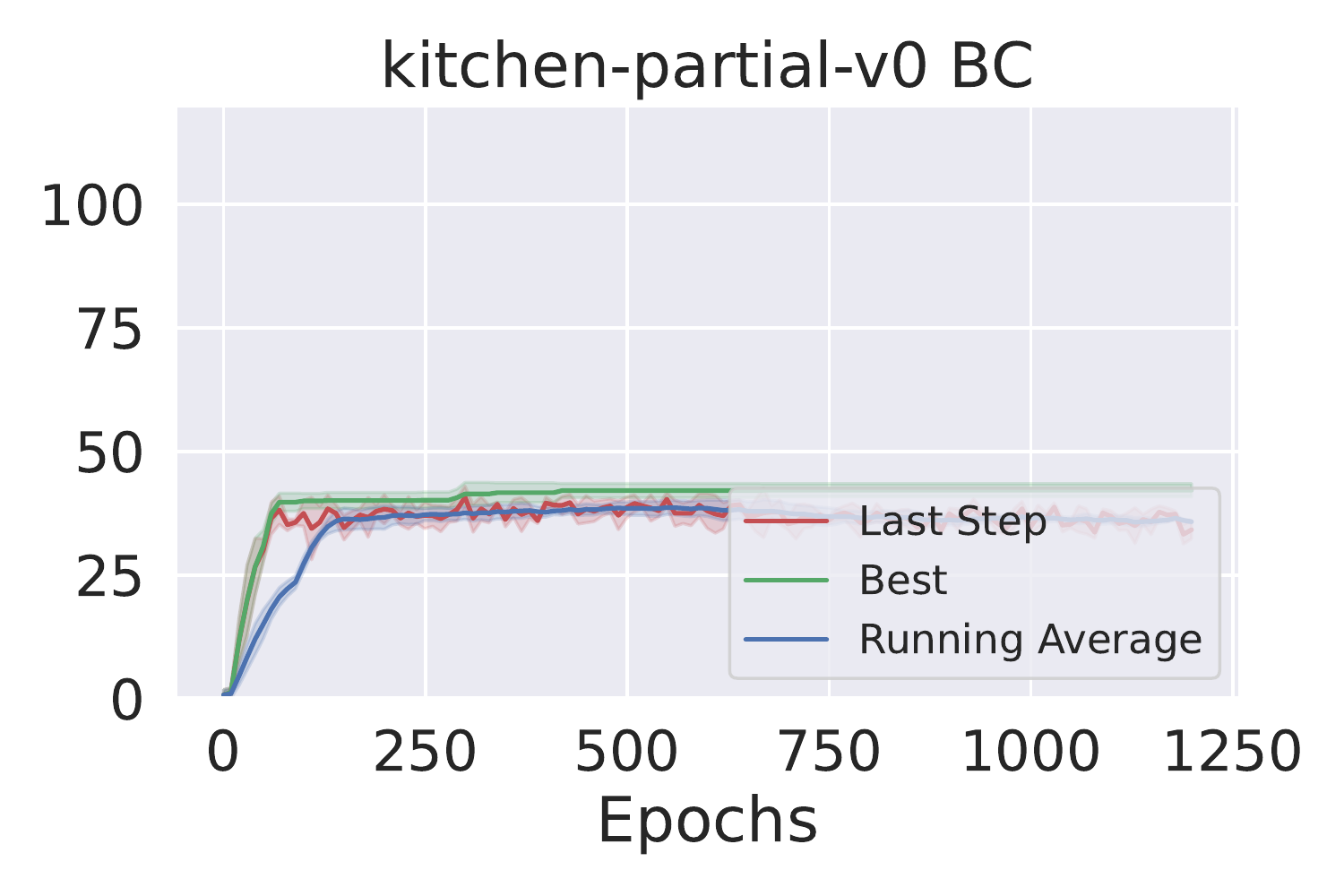}&\includegraphics[width=0.225\linewidth]{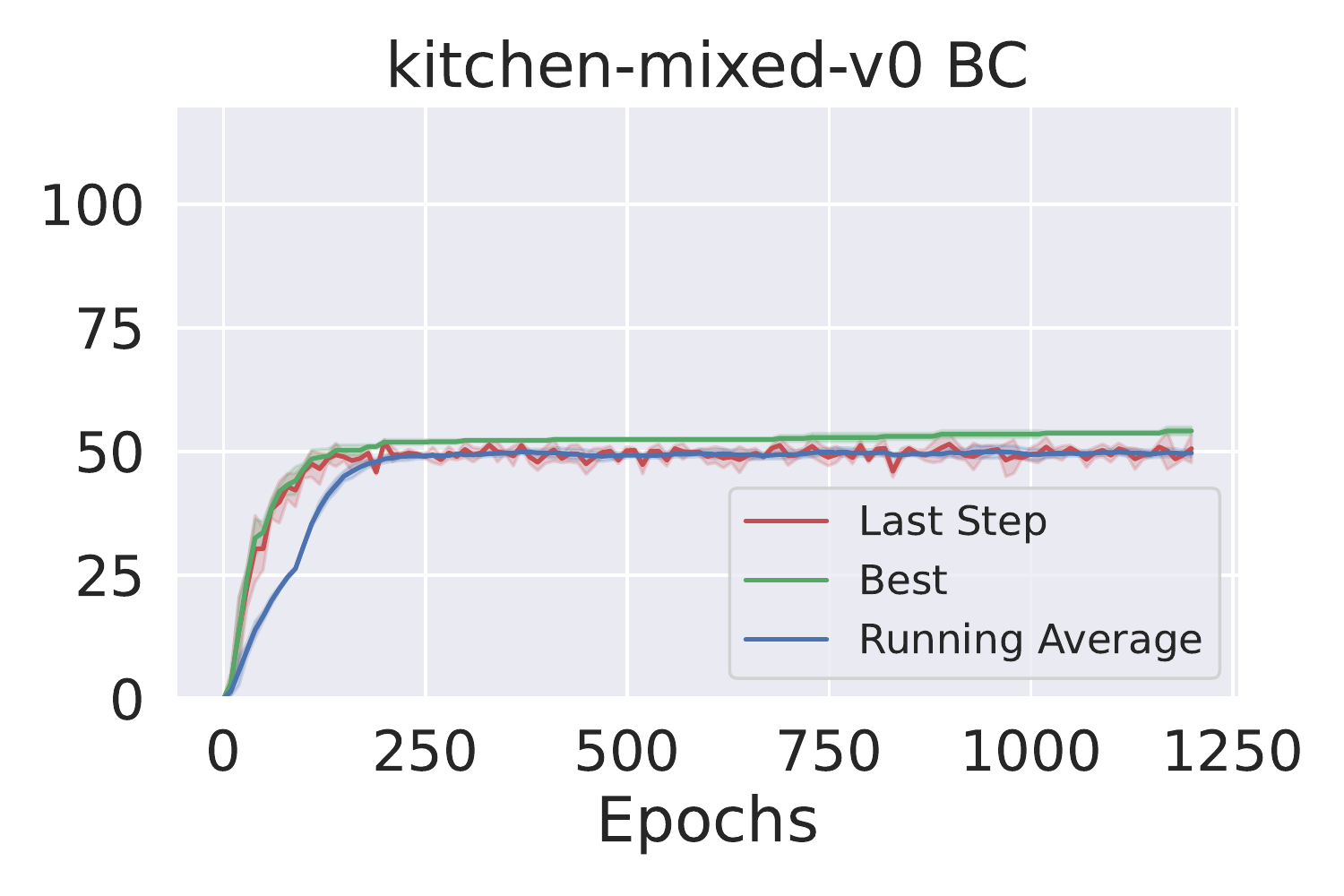}\\\includegraphics[width=0.225\linewidth]{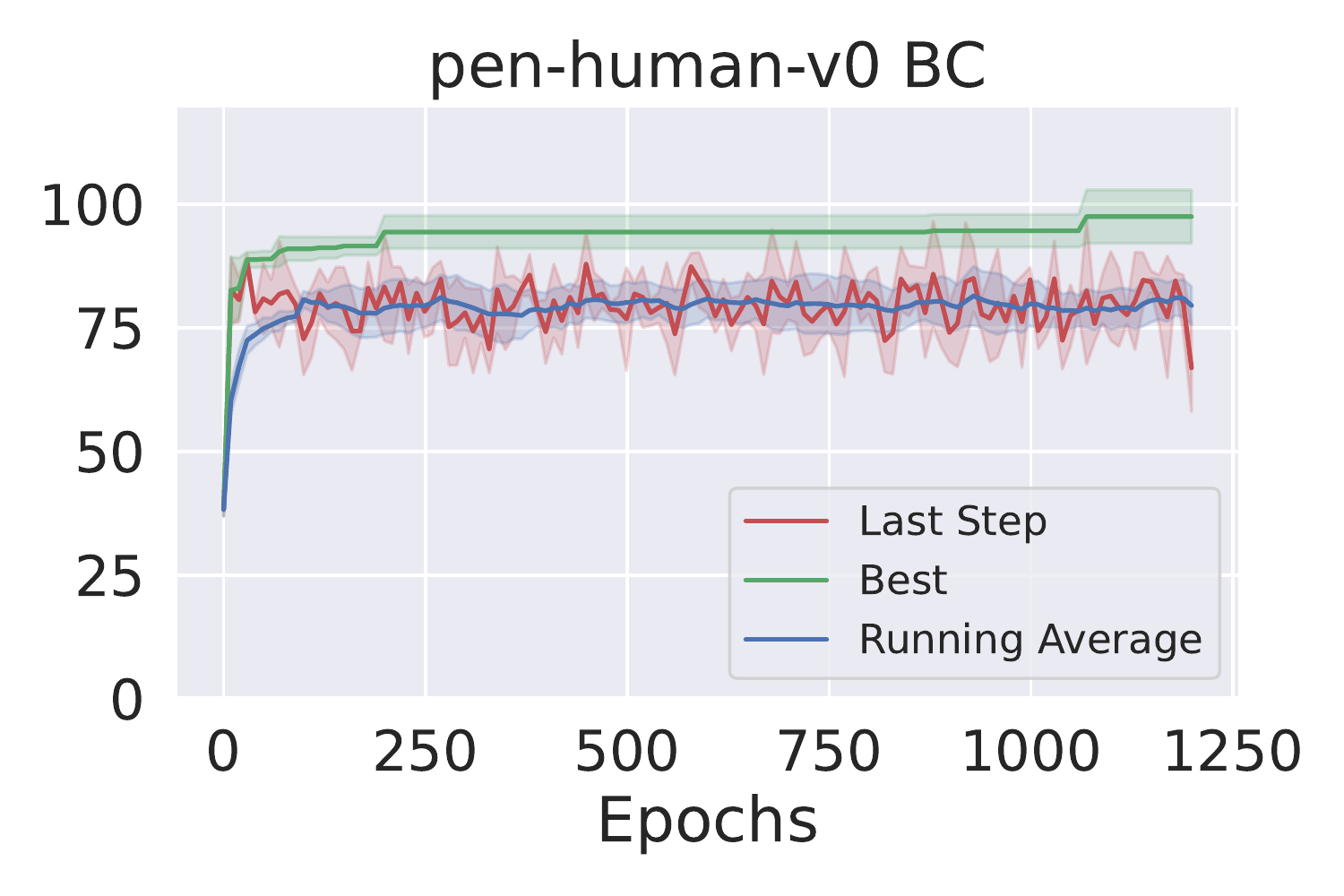}&\includegraphics[width=0.225\linewidth]{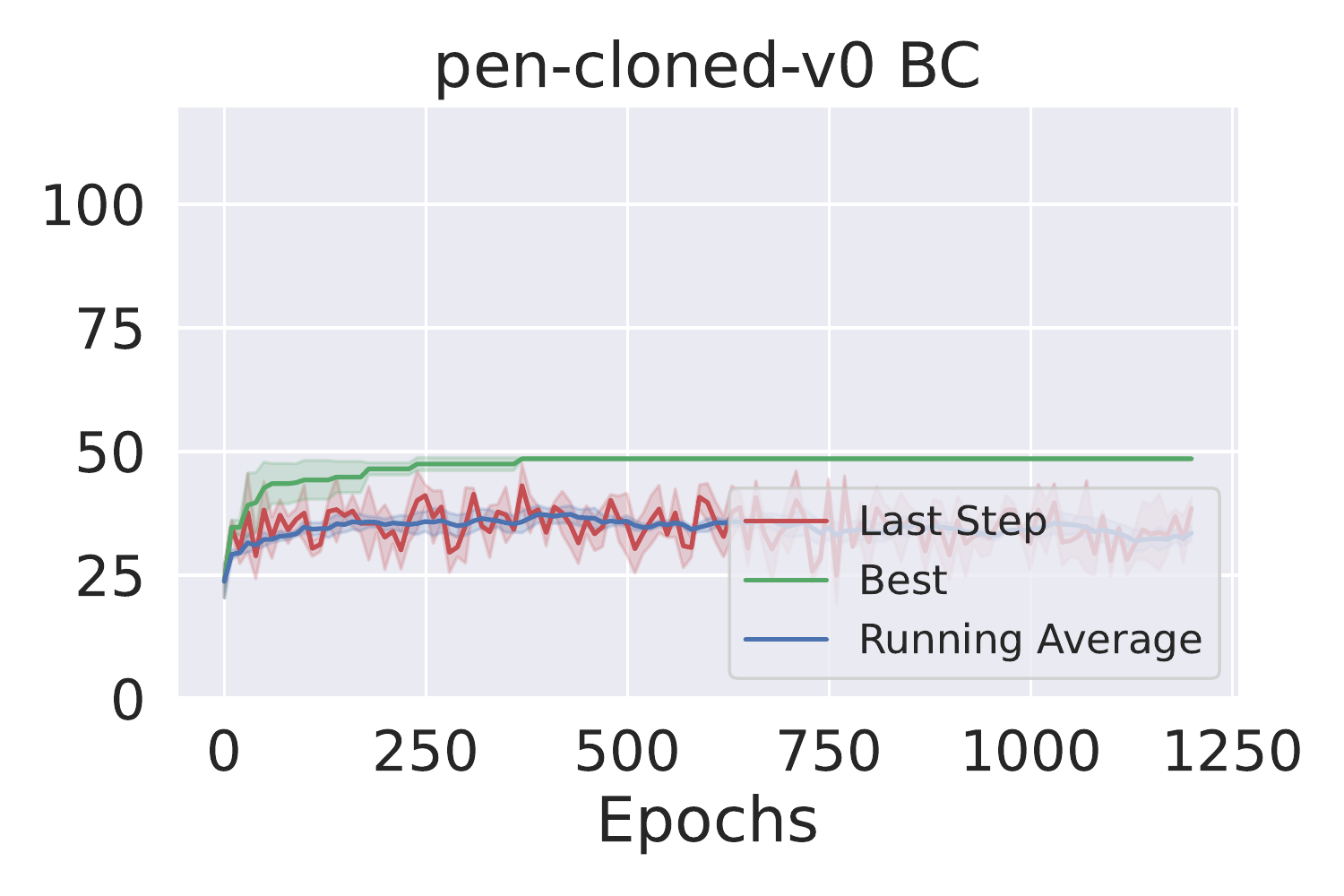}&\includegraphics[width=0.225\linewidth]{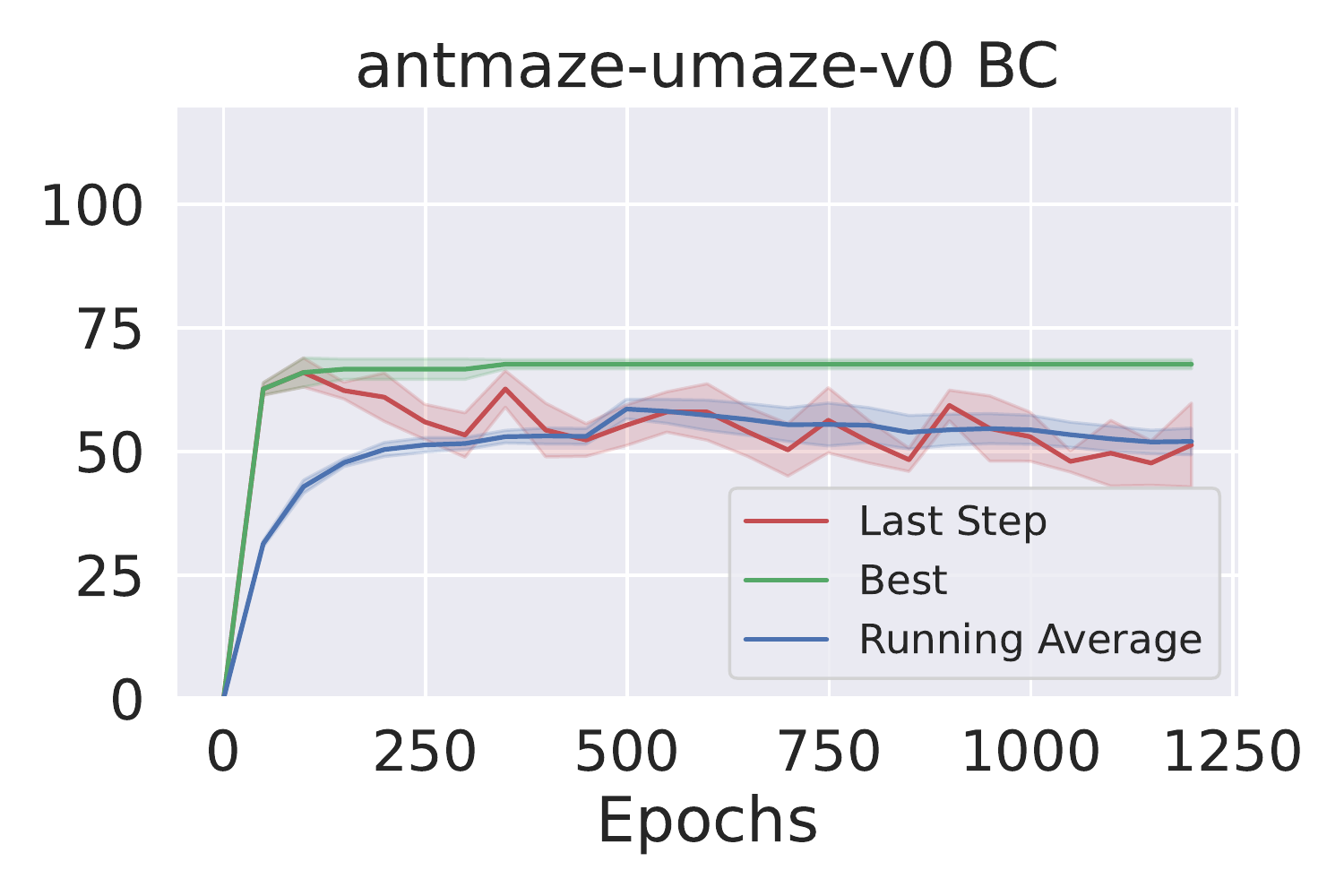}&\includegraphics[width=0.225\linewidth]{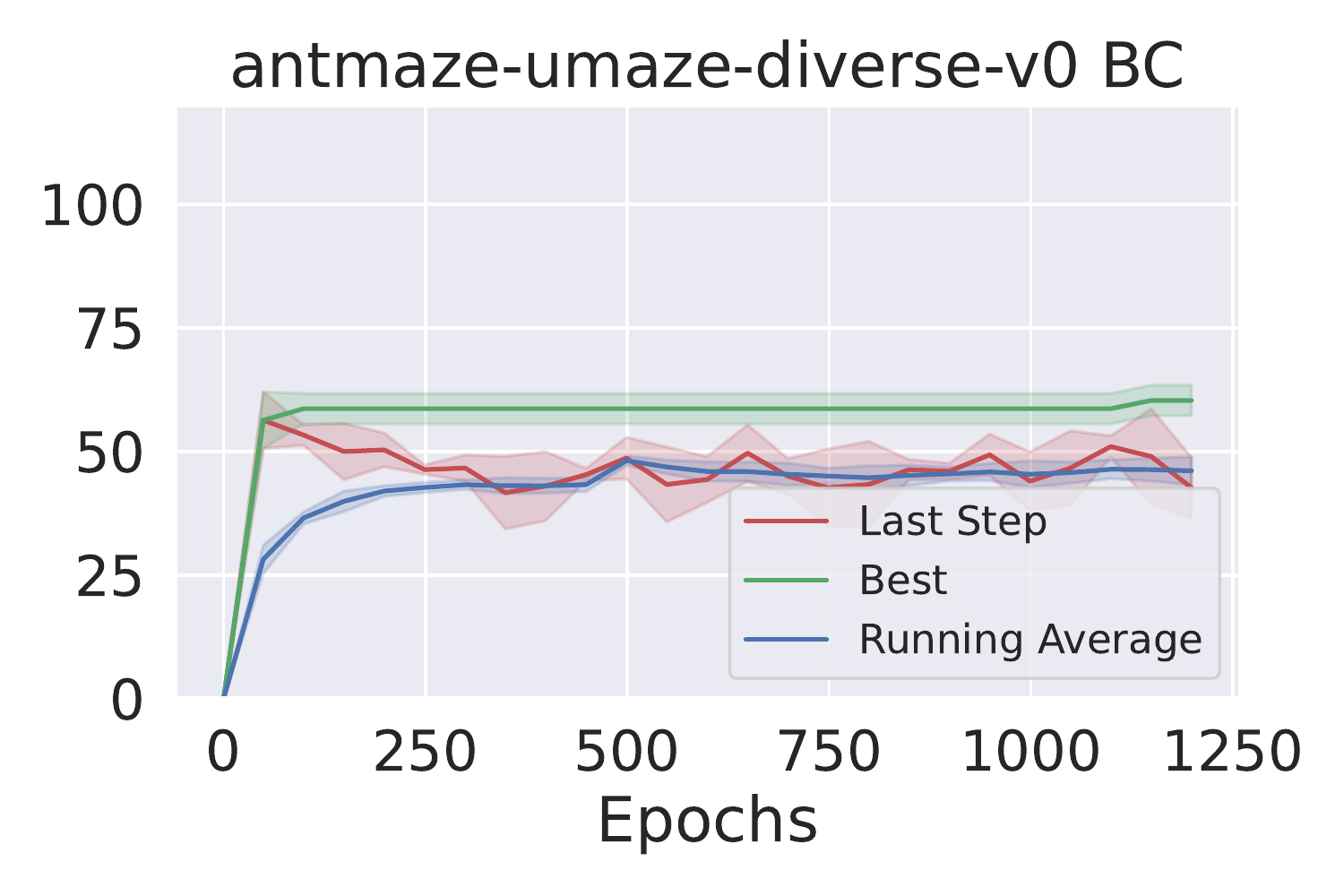}\\\includegraphics[width=0.225\linewidth]{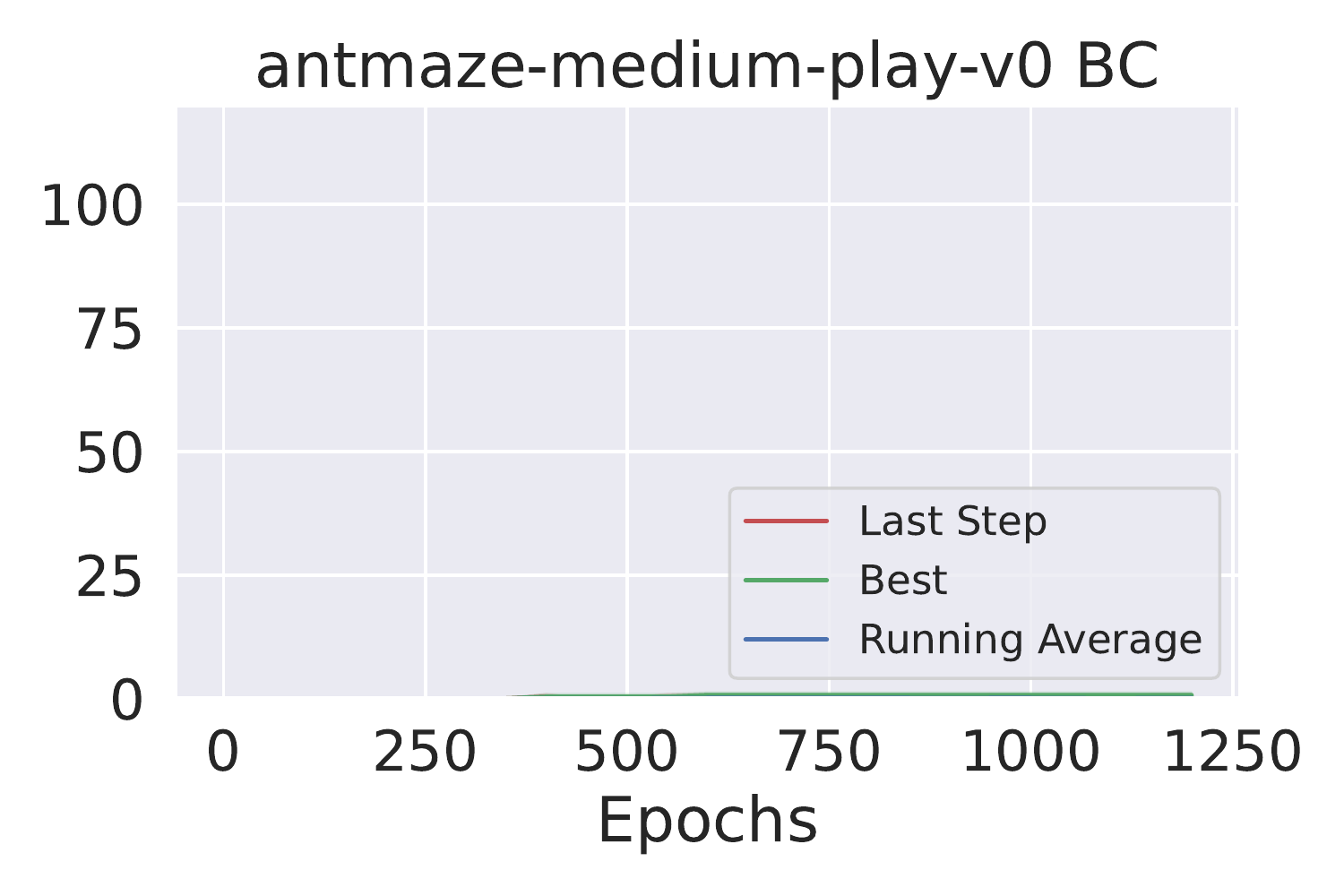}&\includegraphics[width=0.225\linewidth]{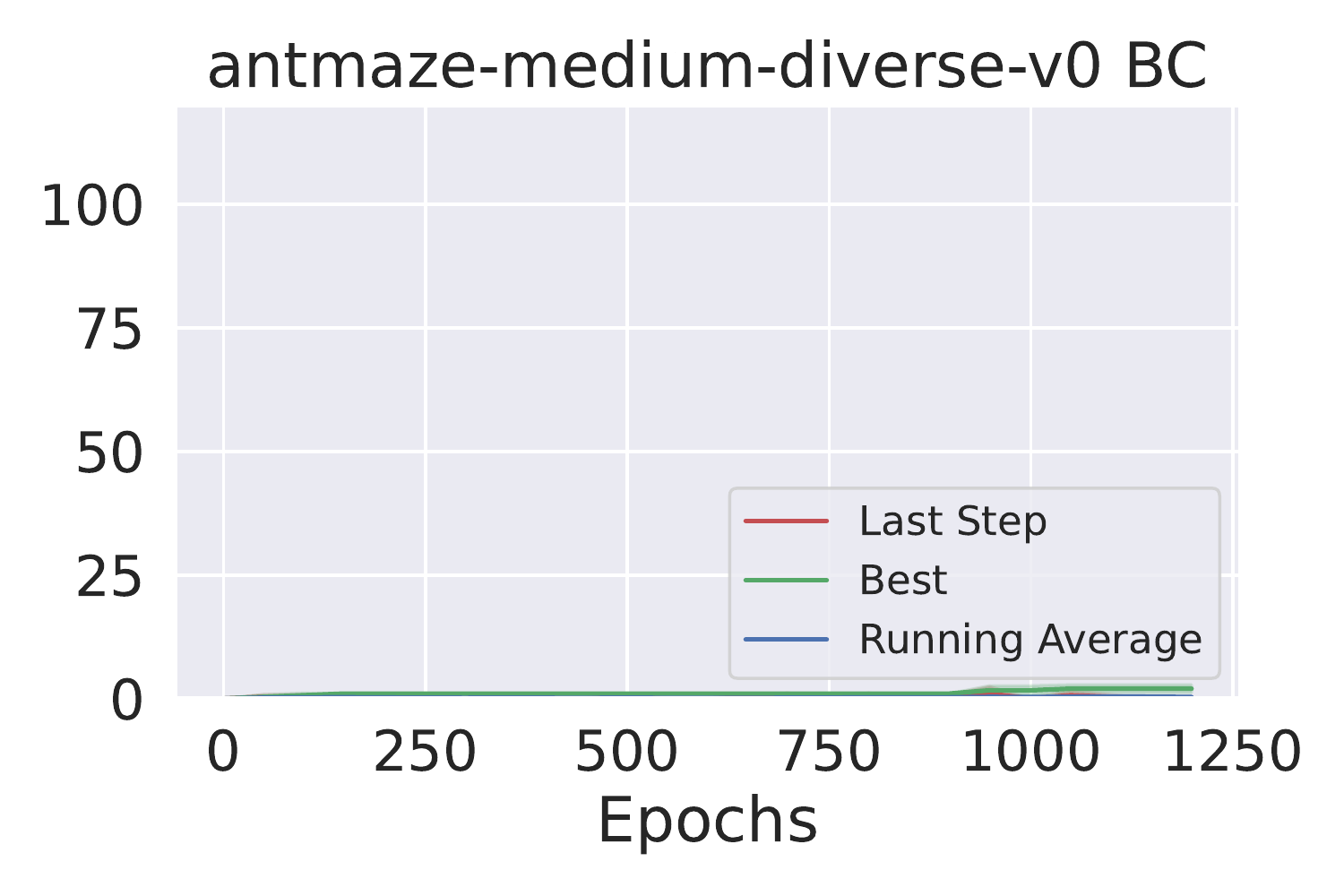}&\includegraphics[width=0.225\linewidth]{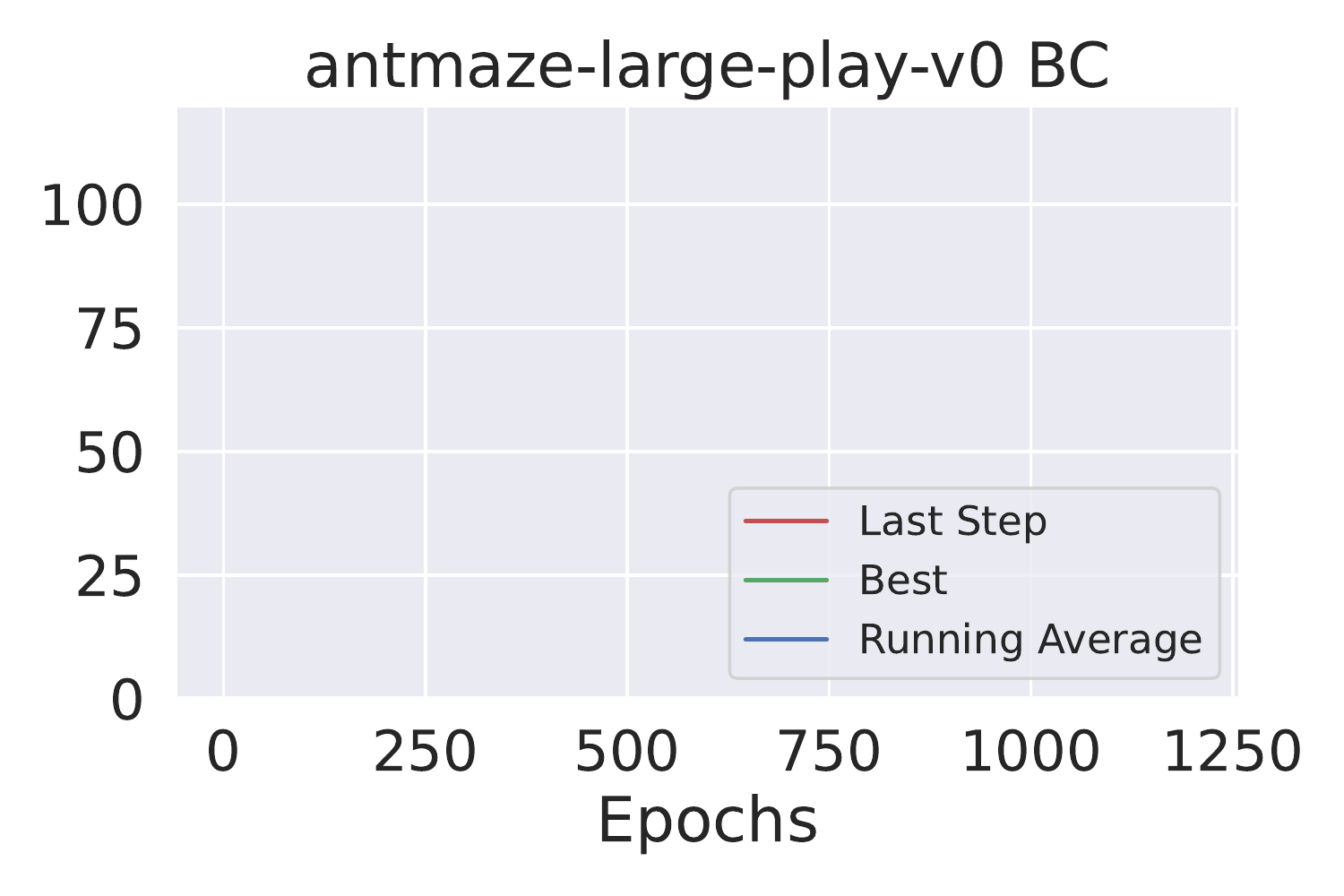}&\includegraphics[width=0.225\linewidth]{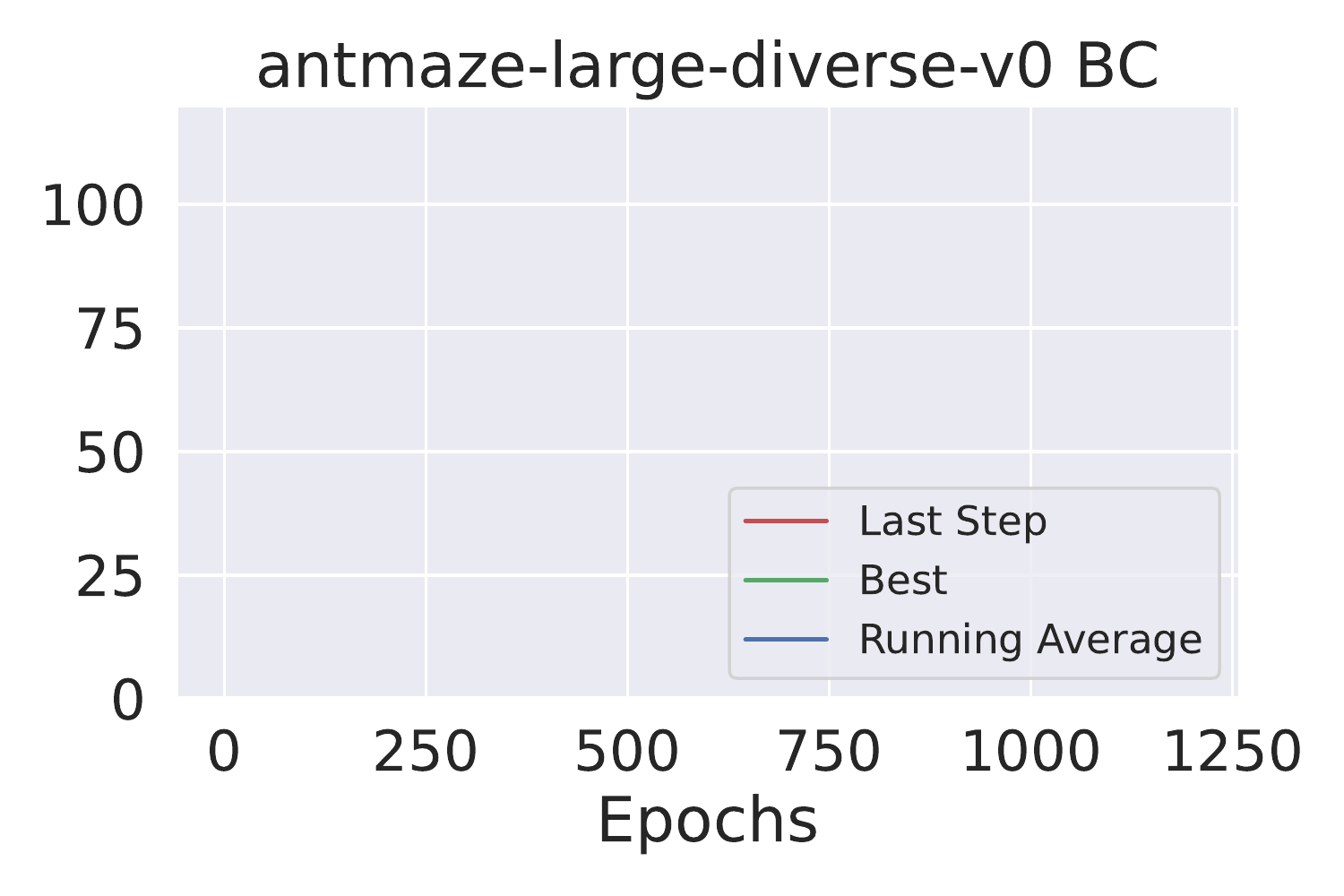}\\\end{tabular}
\centering
\caption{Training Curves of BC on D4RL}
\end{figure*}
\begin{figure*}[htb]
\centering
\begin{tabular}{cccccc}
\includegraphics[width=0.225\linewidth]{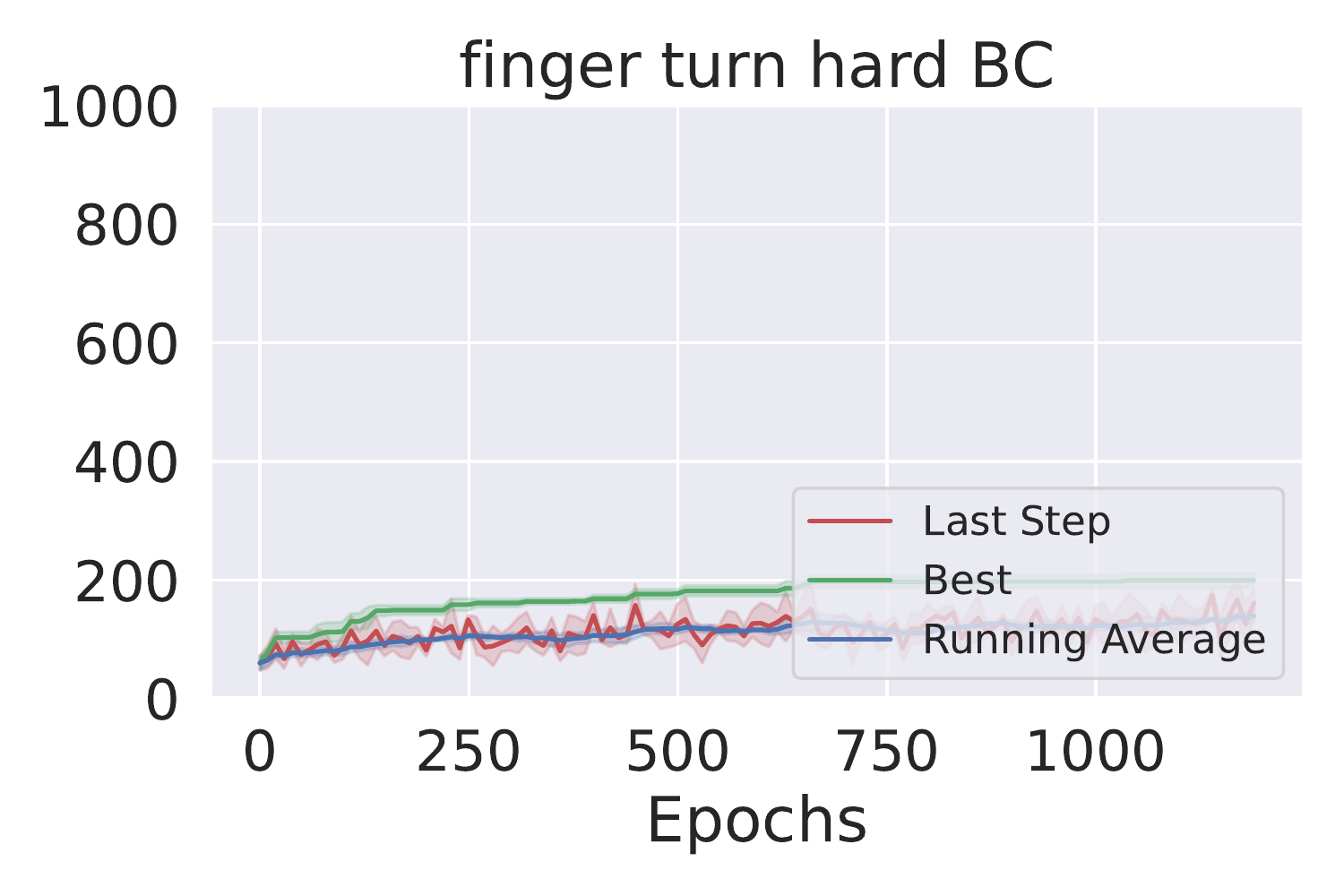}&\includegraphics[width=0.225\linewidth]{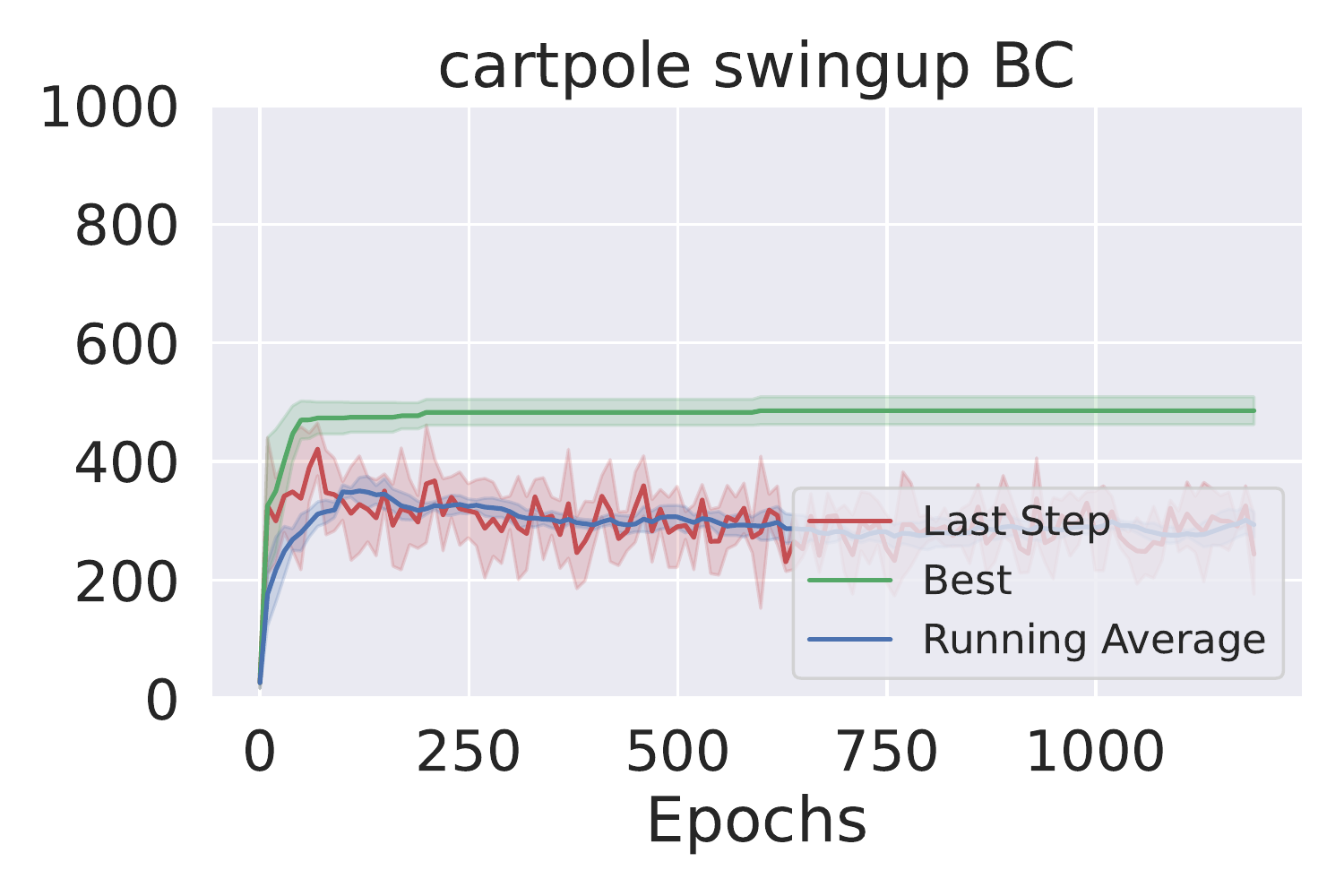}&\includegraphics[width=0.225\linewidth]{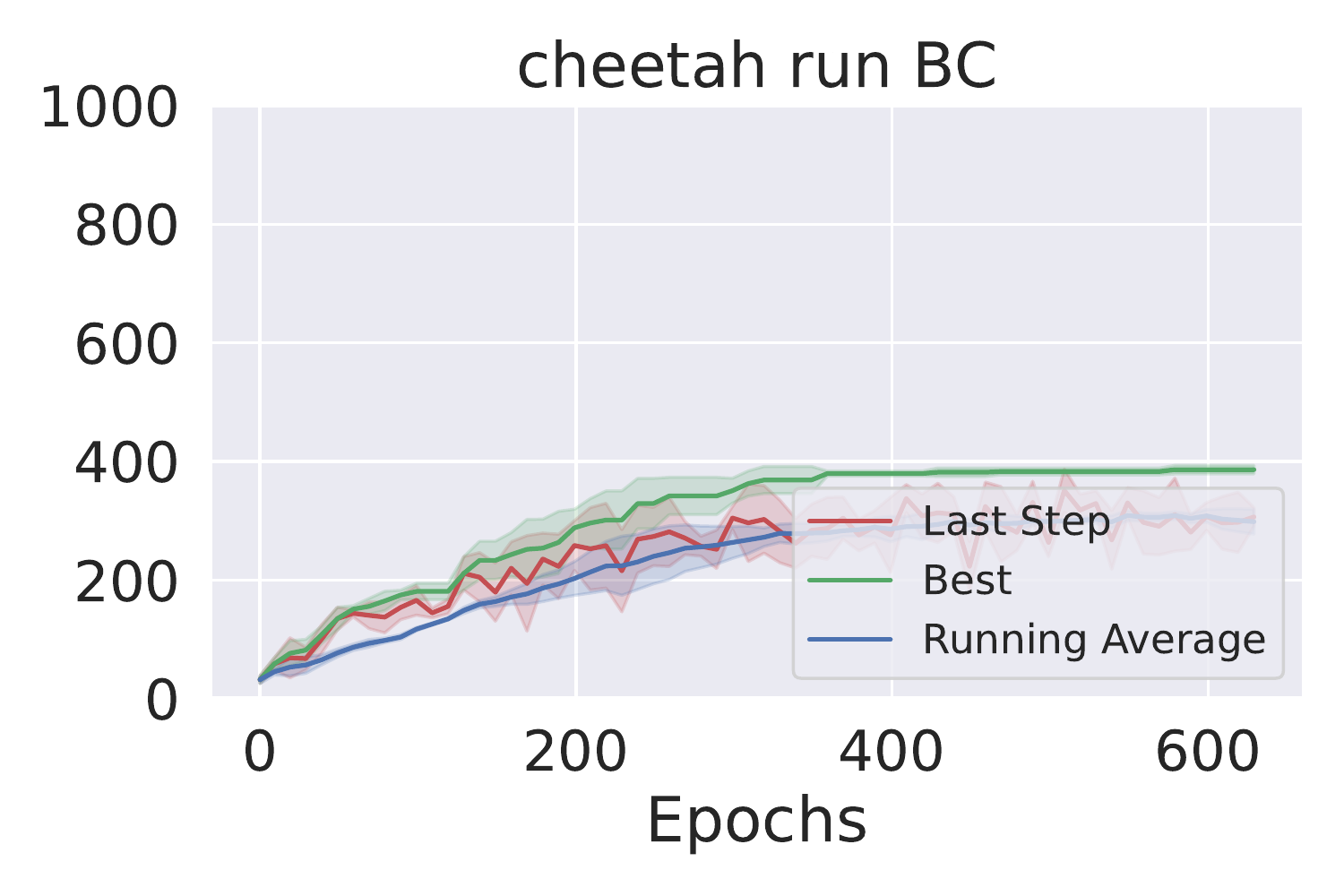}&\includegraphics[width=0.225\linewidth]{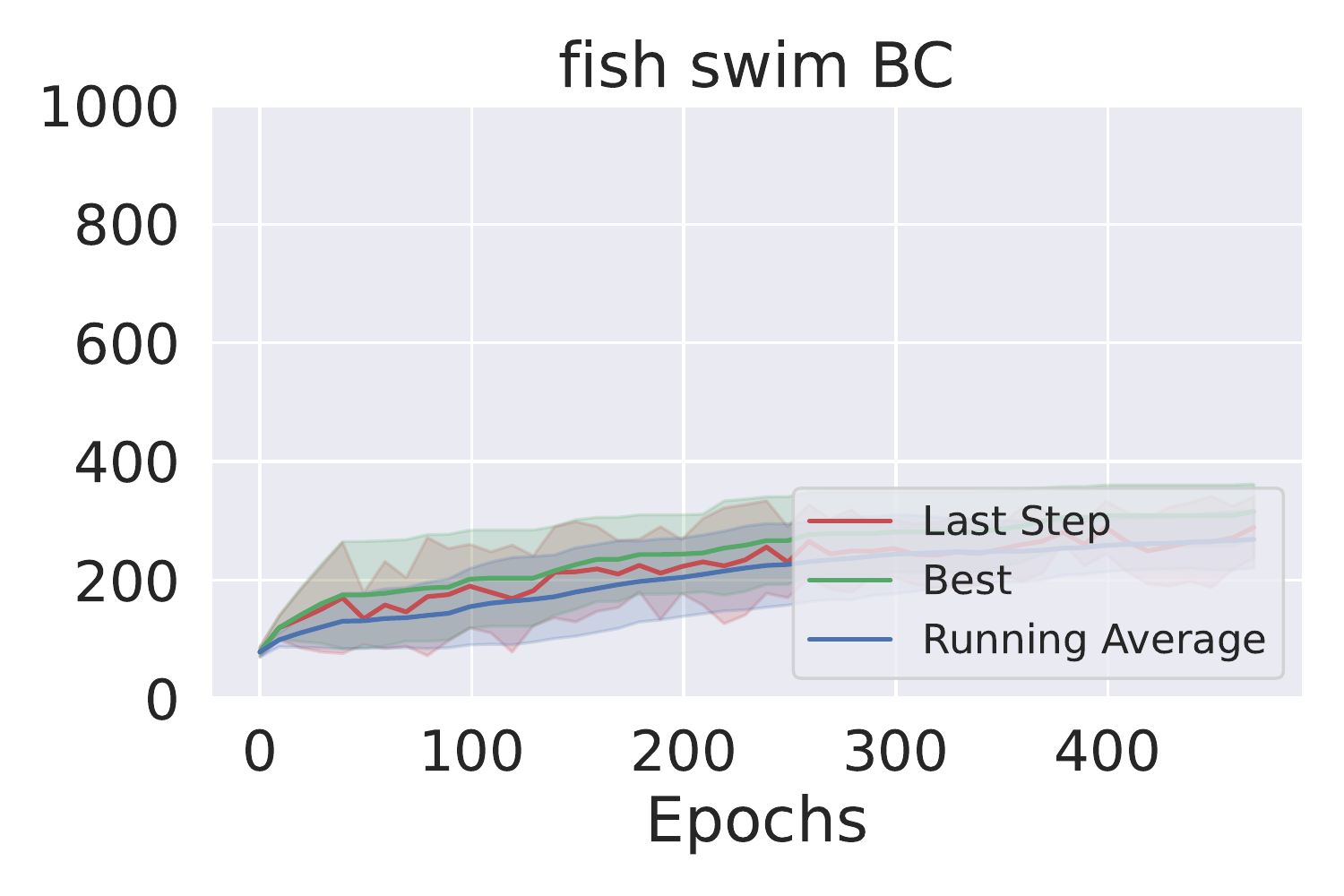}\\\includegraphics[width=0.225\linewidth]{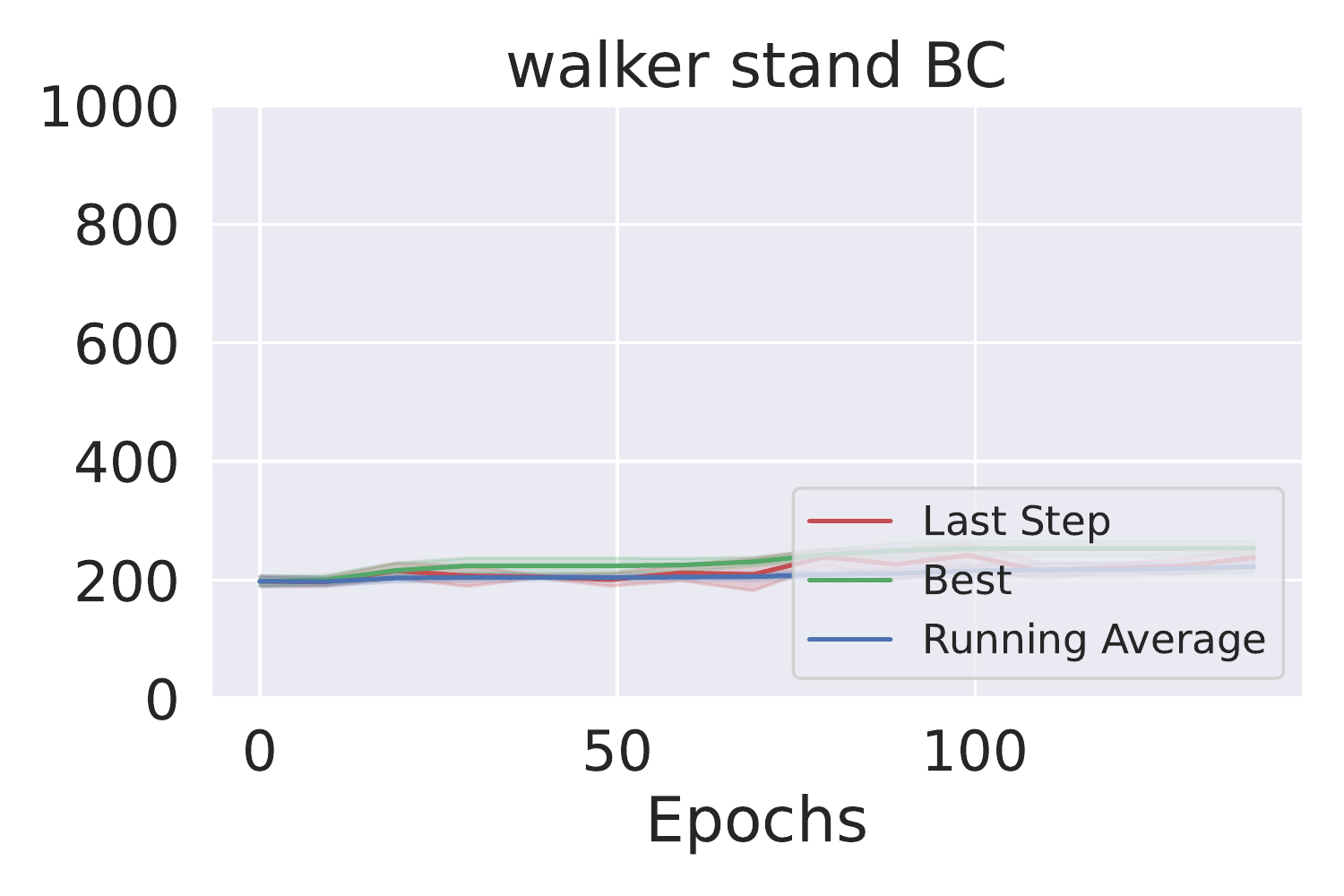}&\includegraphics[width=0.225\linewidth]{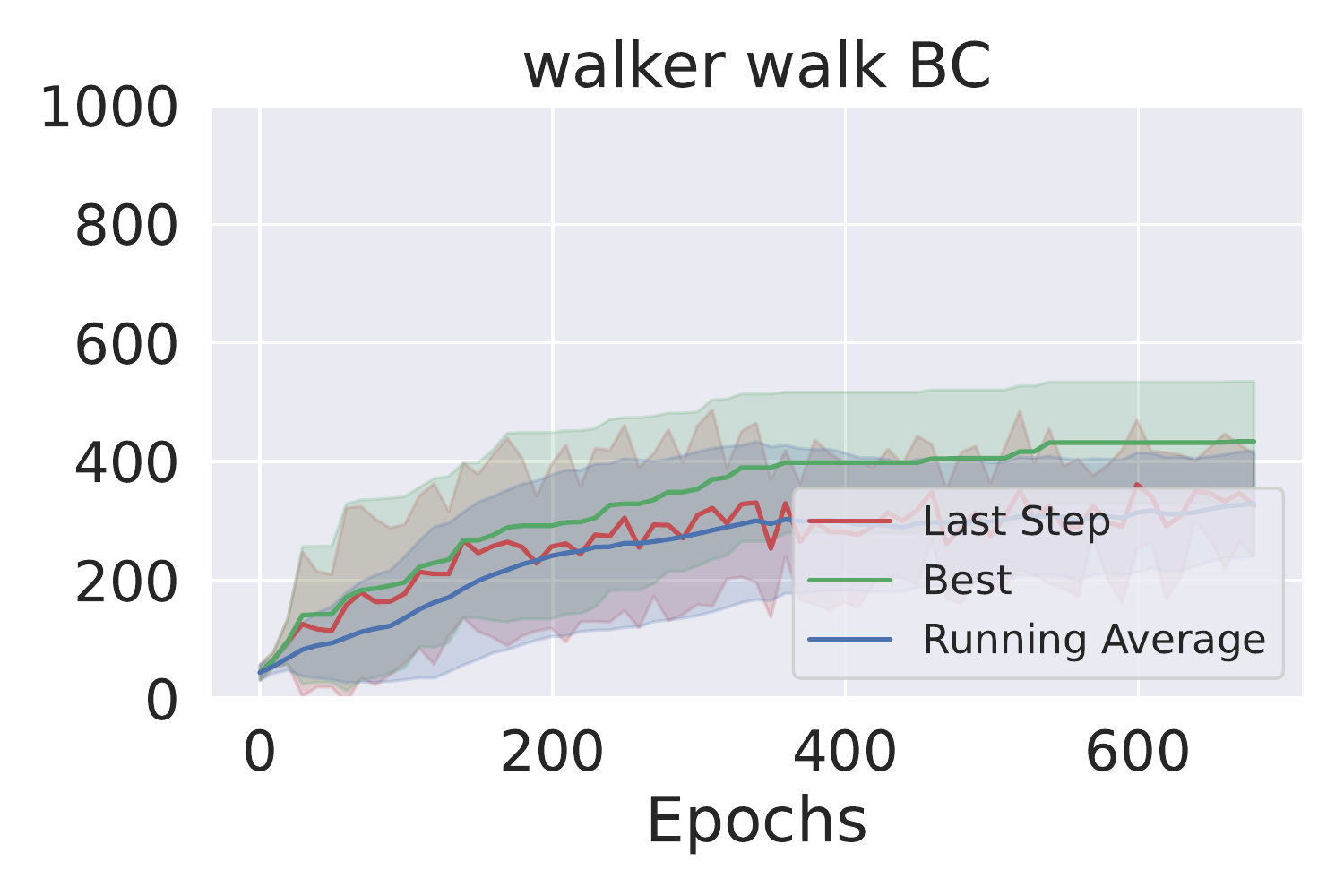}&&\\\end{tabular}
\centering
\caption{Training Curves of BC on RLUP}
\end{figure*}
\begin{figure*}[htb]
\centering
\begin{tabular}{cccccc}
\includegraphics[width=0.225\linewidth]{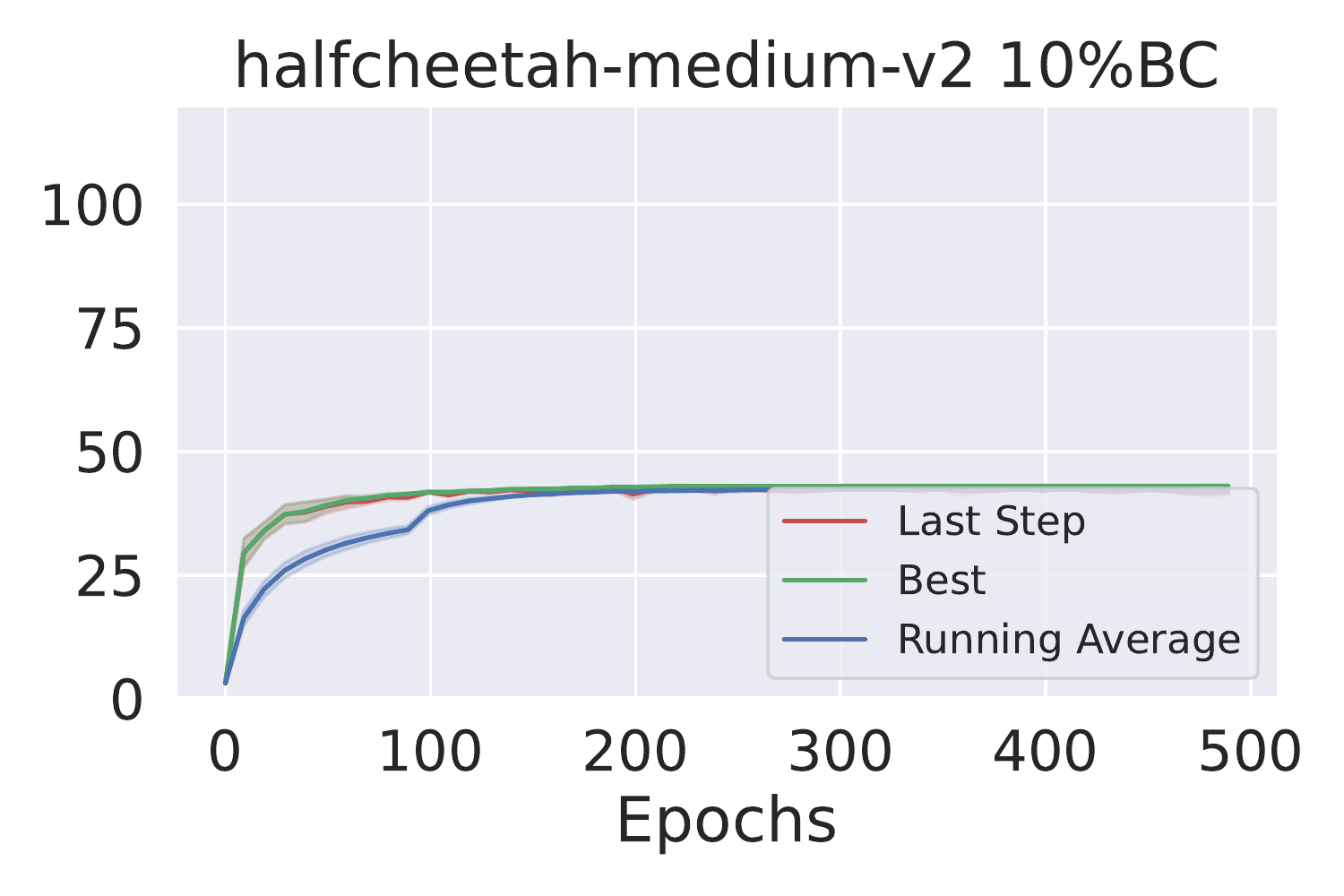}&\includegraphics[width=0.225\linewidth]{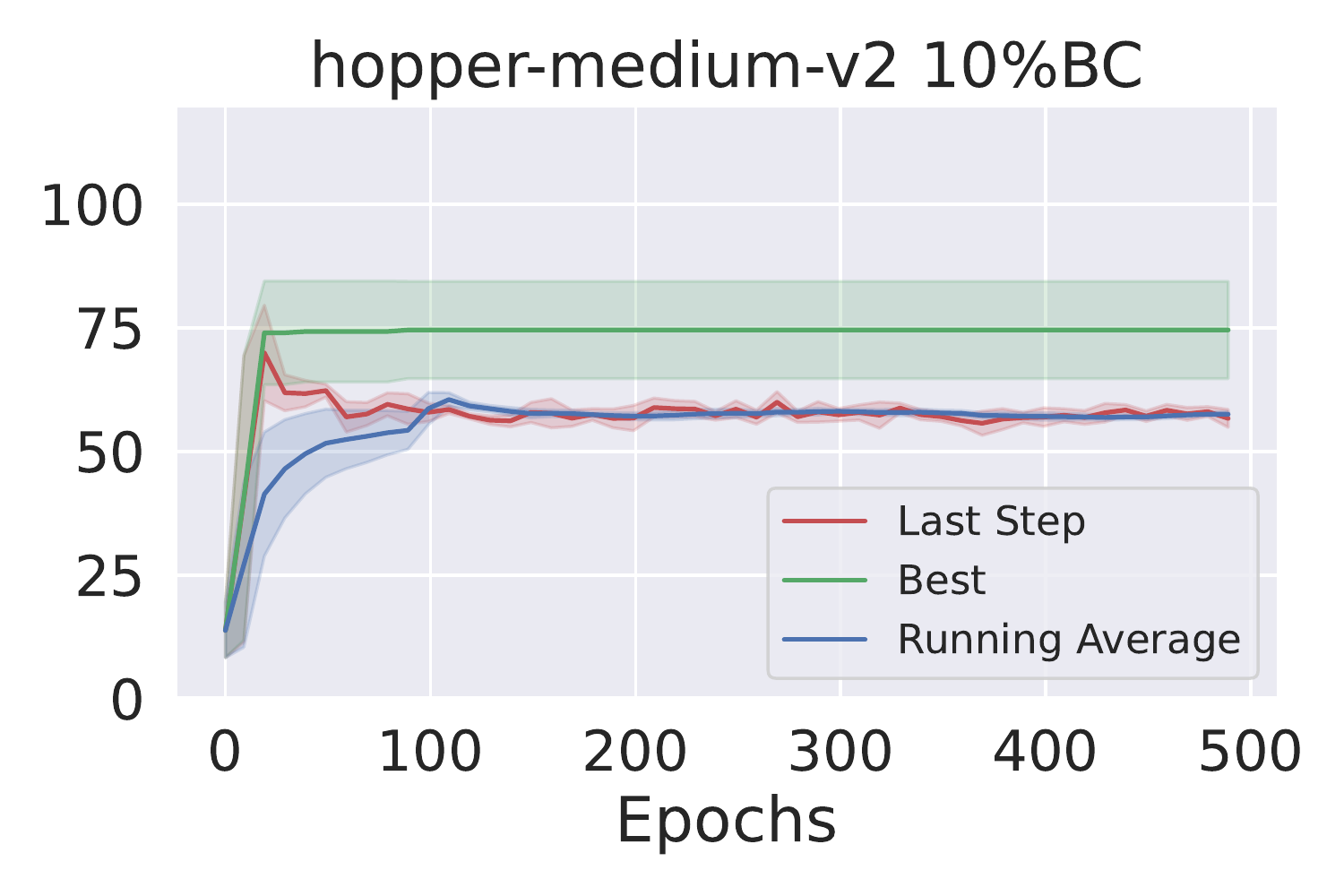}&\includegraphics[width=0.225\linewidth]{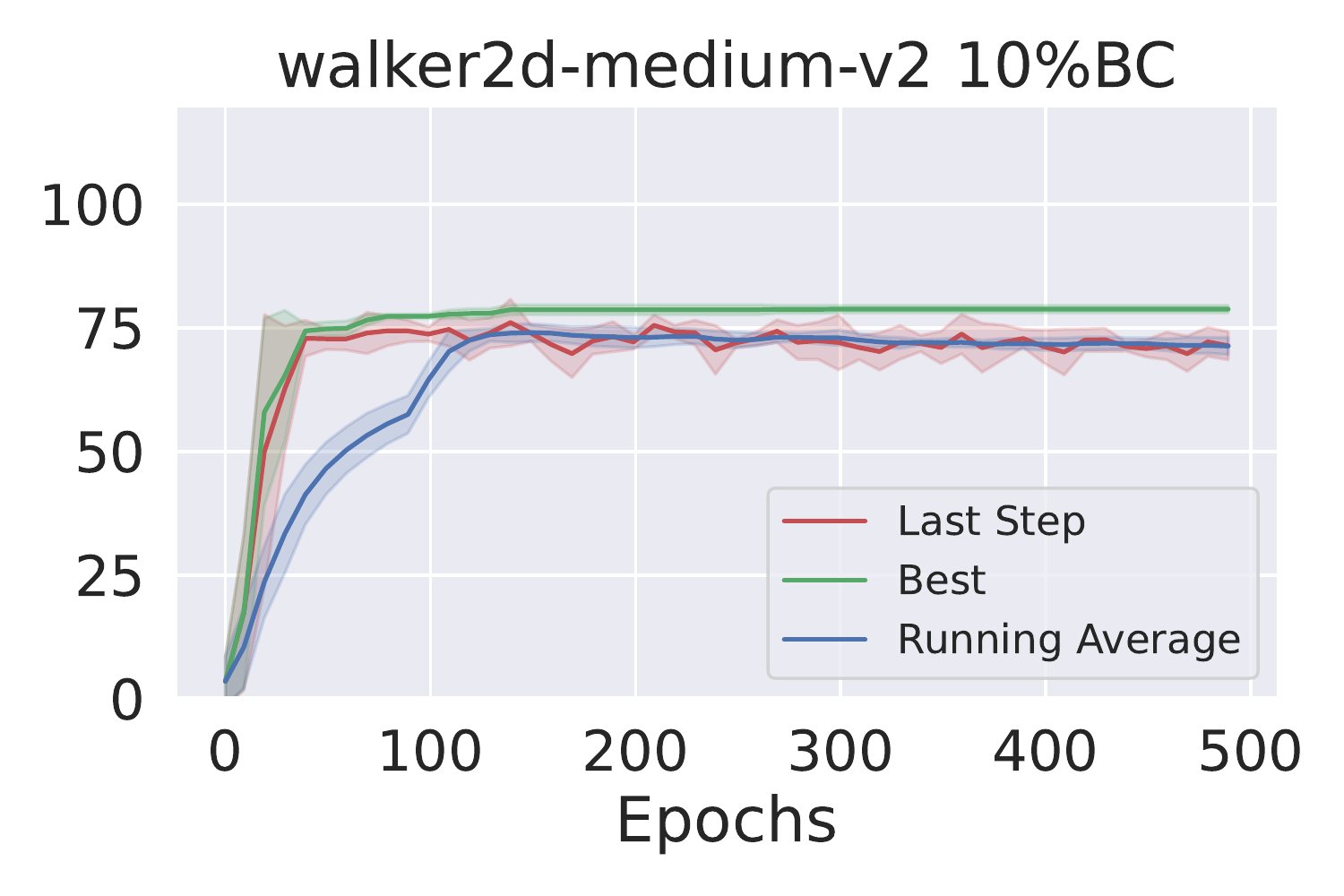}&\includegraphics[width=0.225\linewidth]{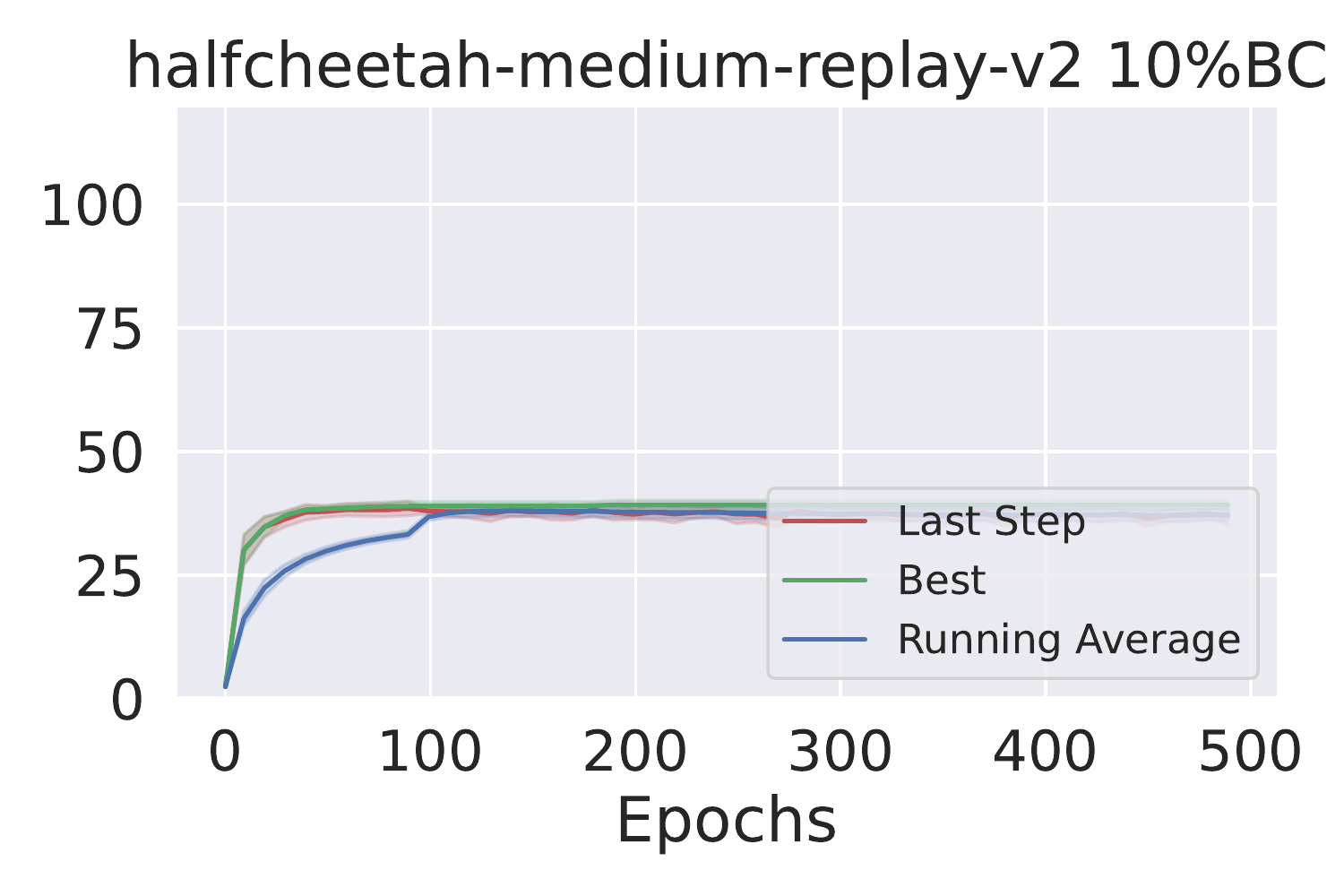}\\\includegraphics[width=0.225\linewidth]{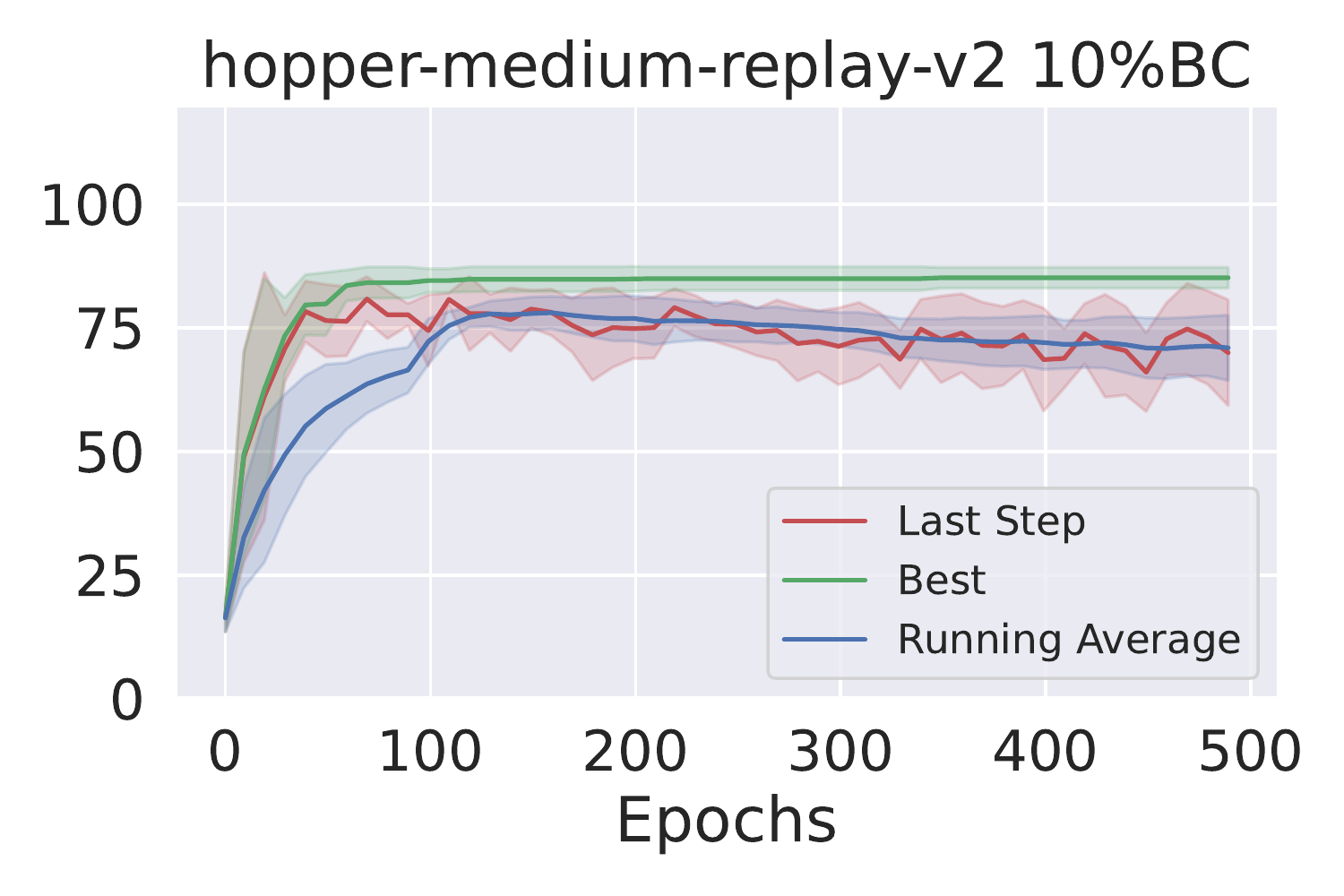}&\includegraphics[width=0.225\linewidth]{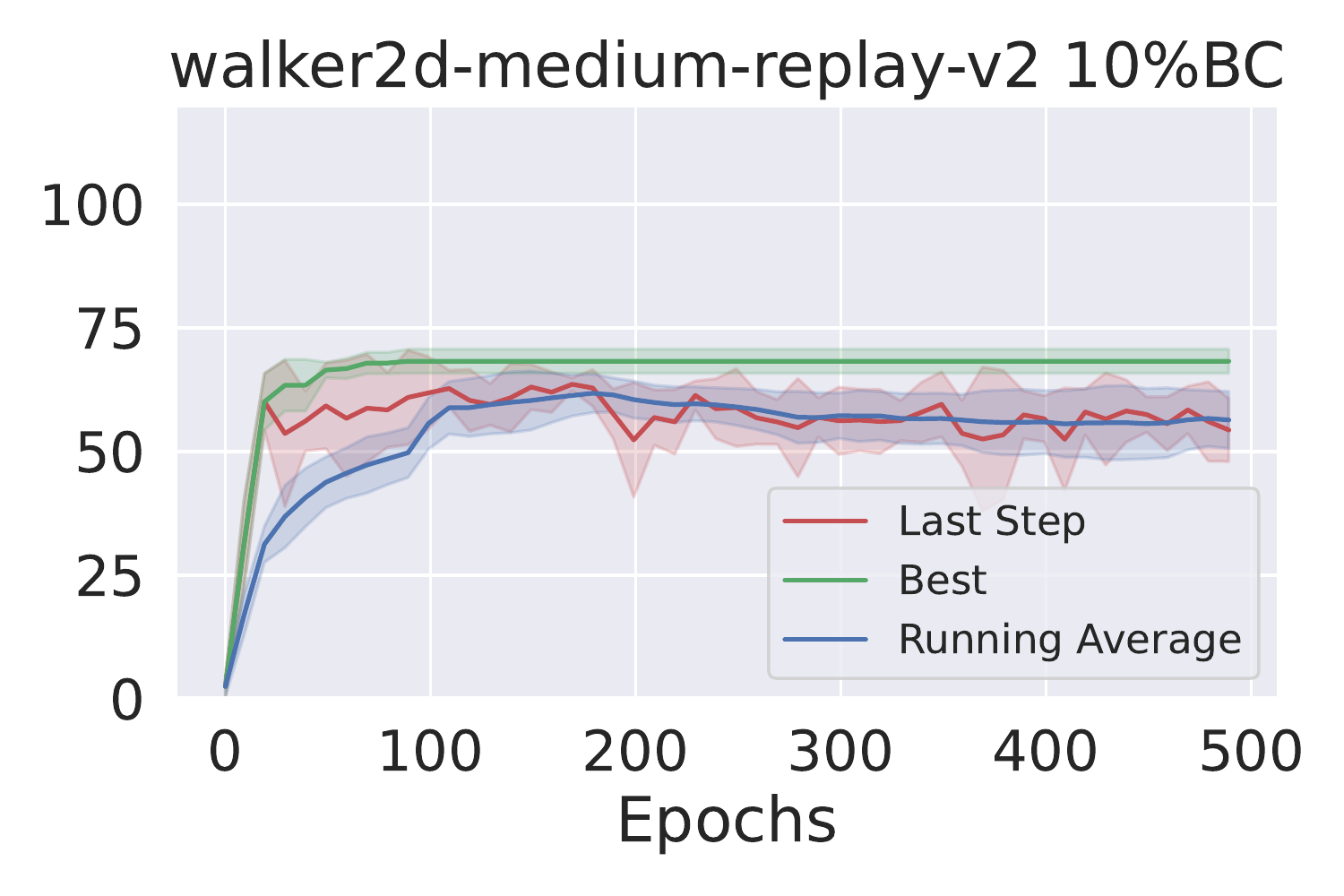}&\includegraphics[width=0.225\linewidth]{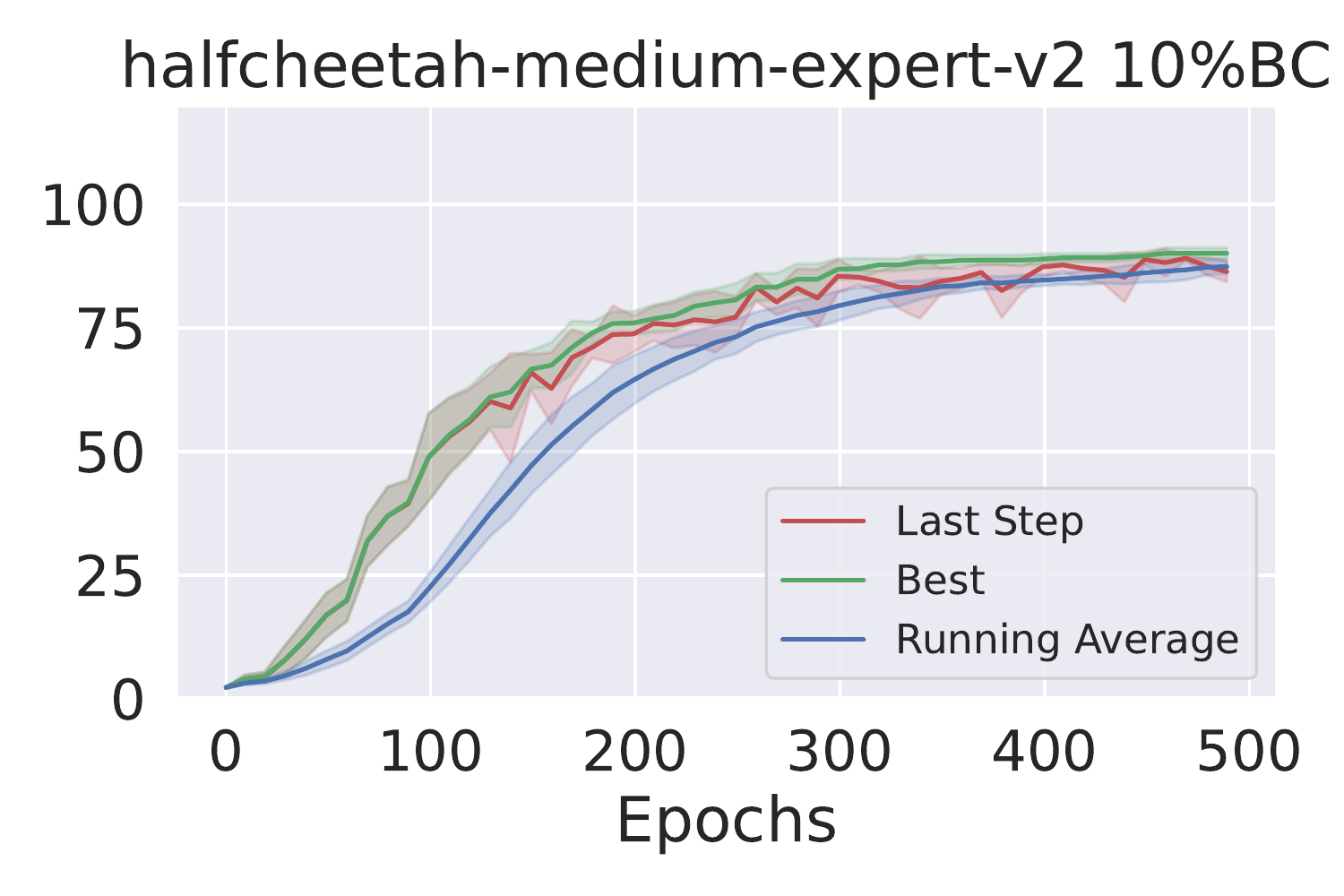}&\includegraphics[width=0.225\linewidth]{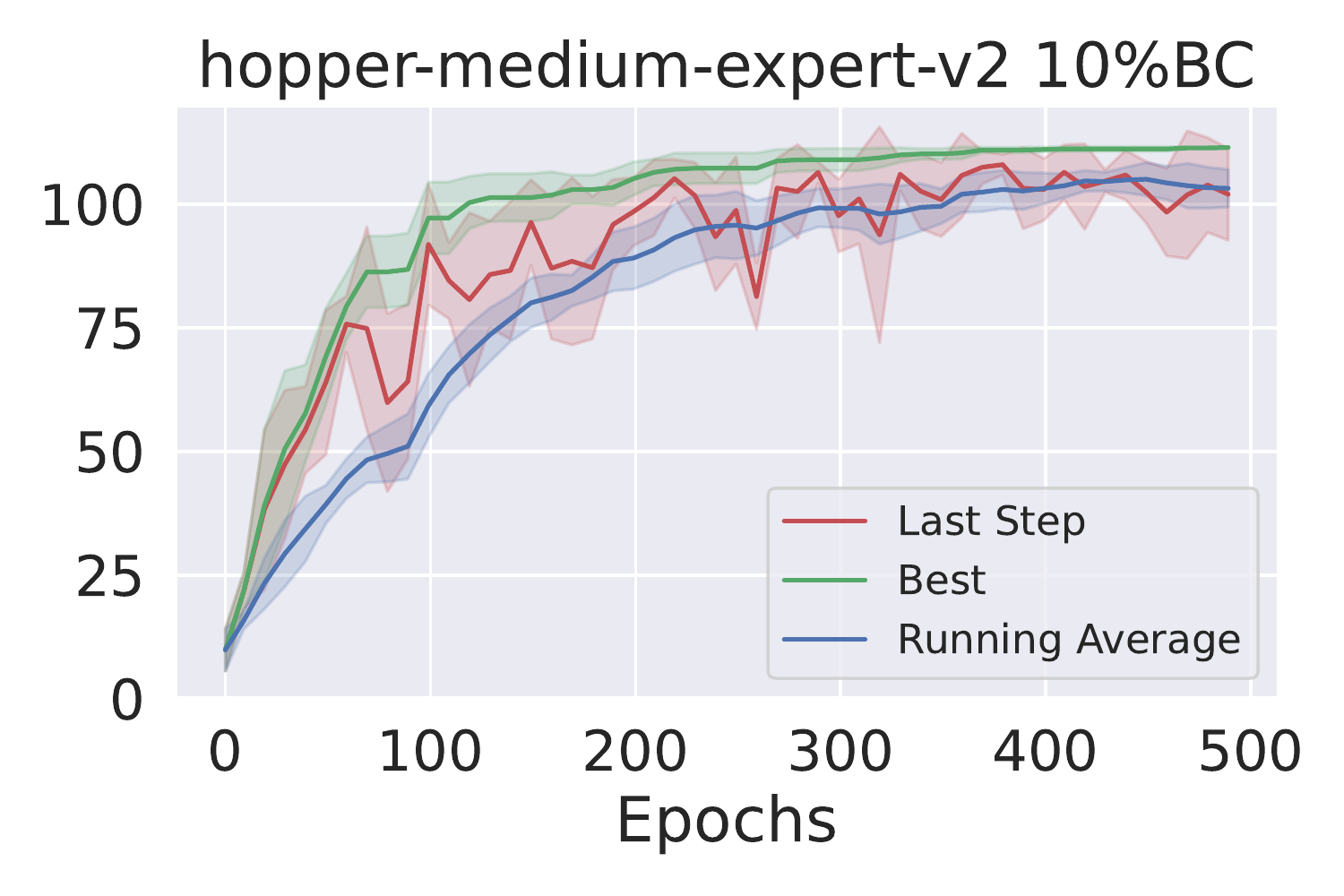}\\\includegraphics[width=0.225\linewidth]{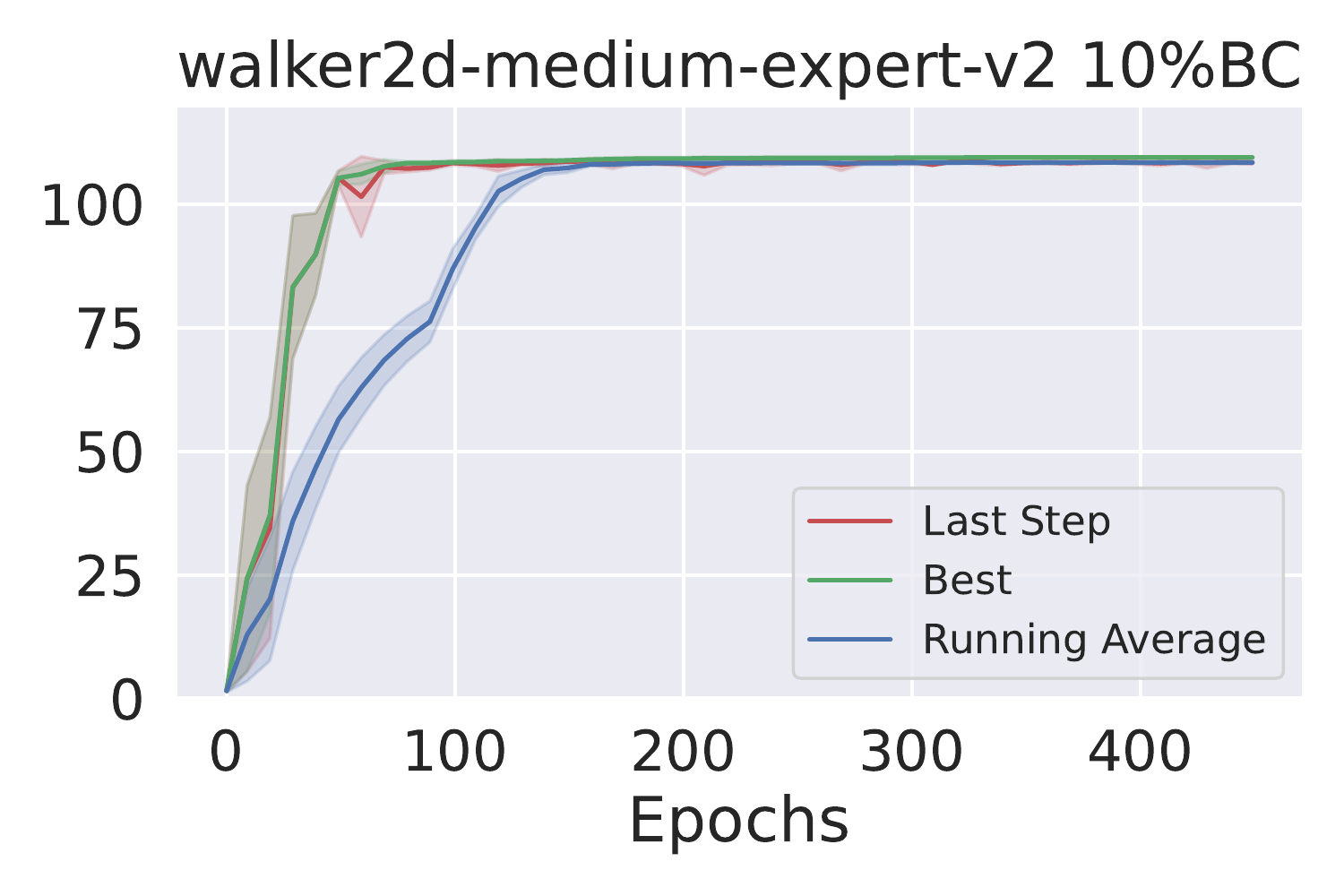}&\includegraphics[width=0.225\linewidth]{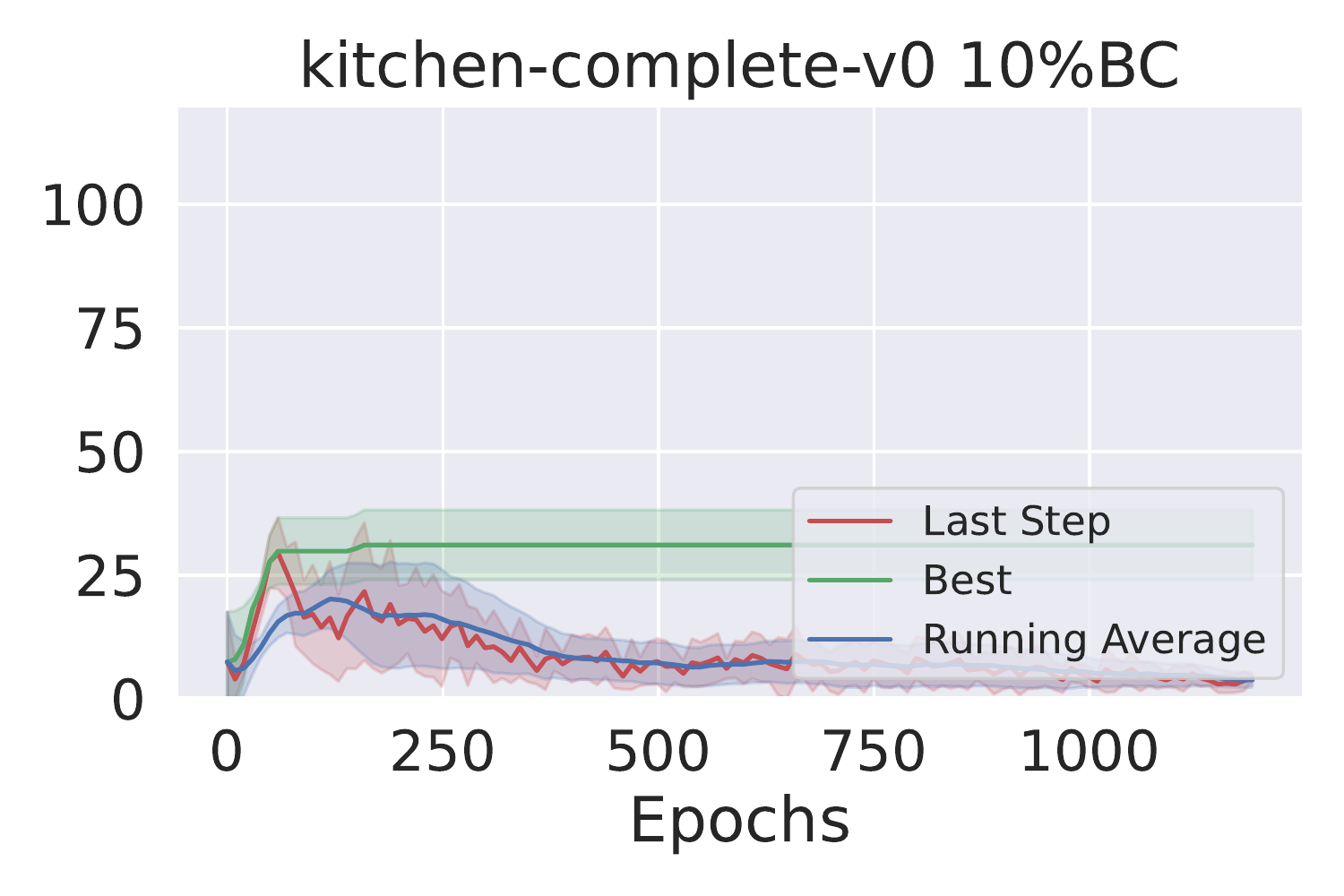}&\includegraphics[width=0.225\linewidth]{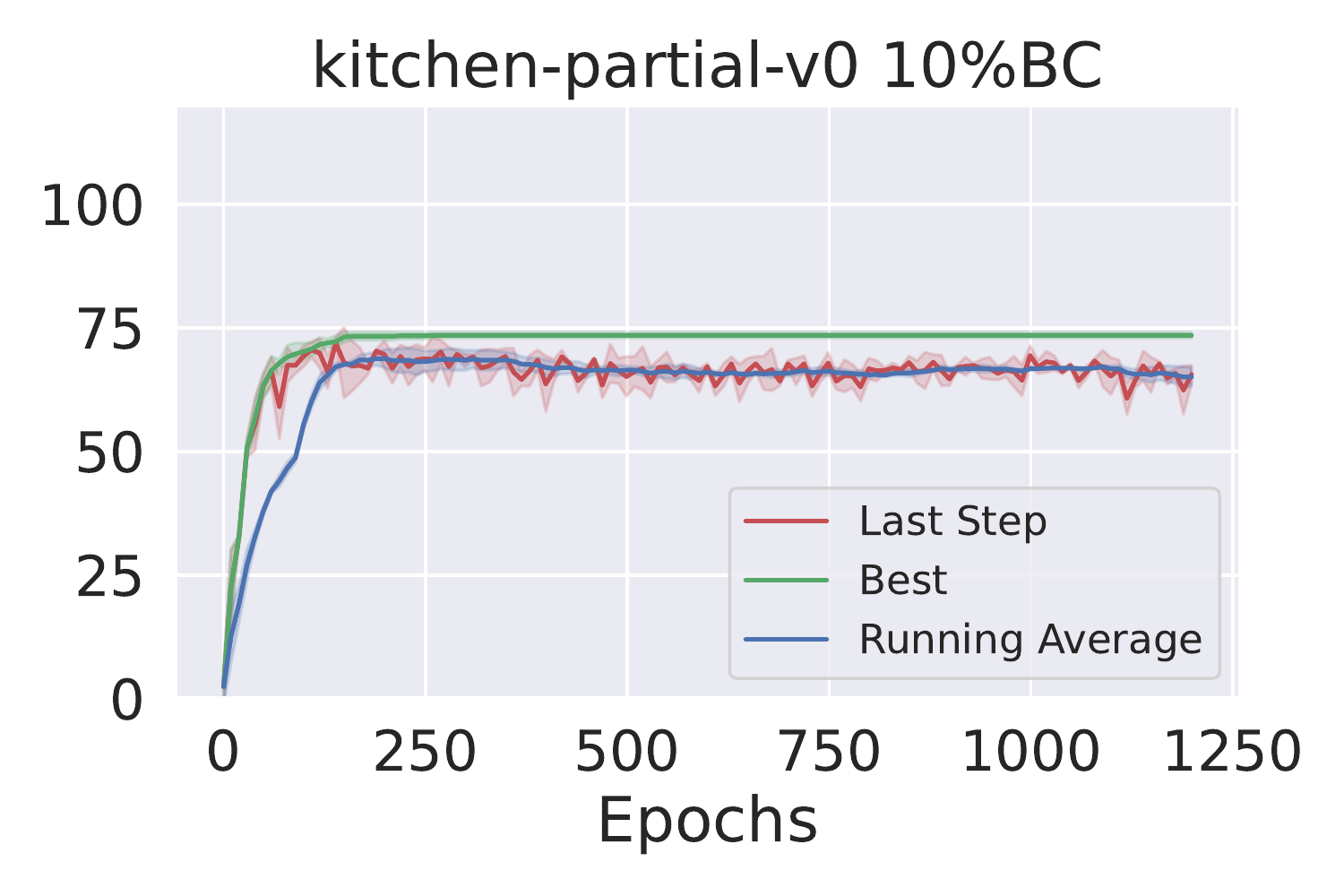}&\includegraphics[width=0.225\linewidth]{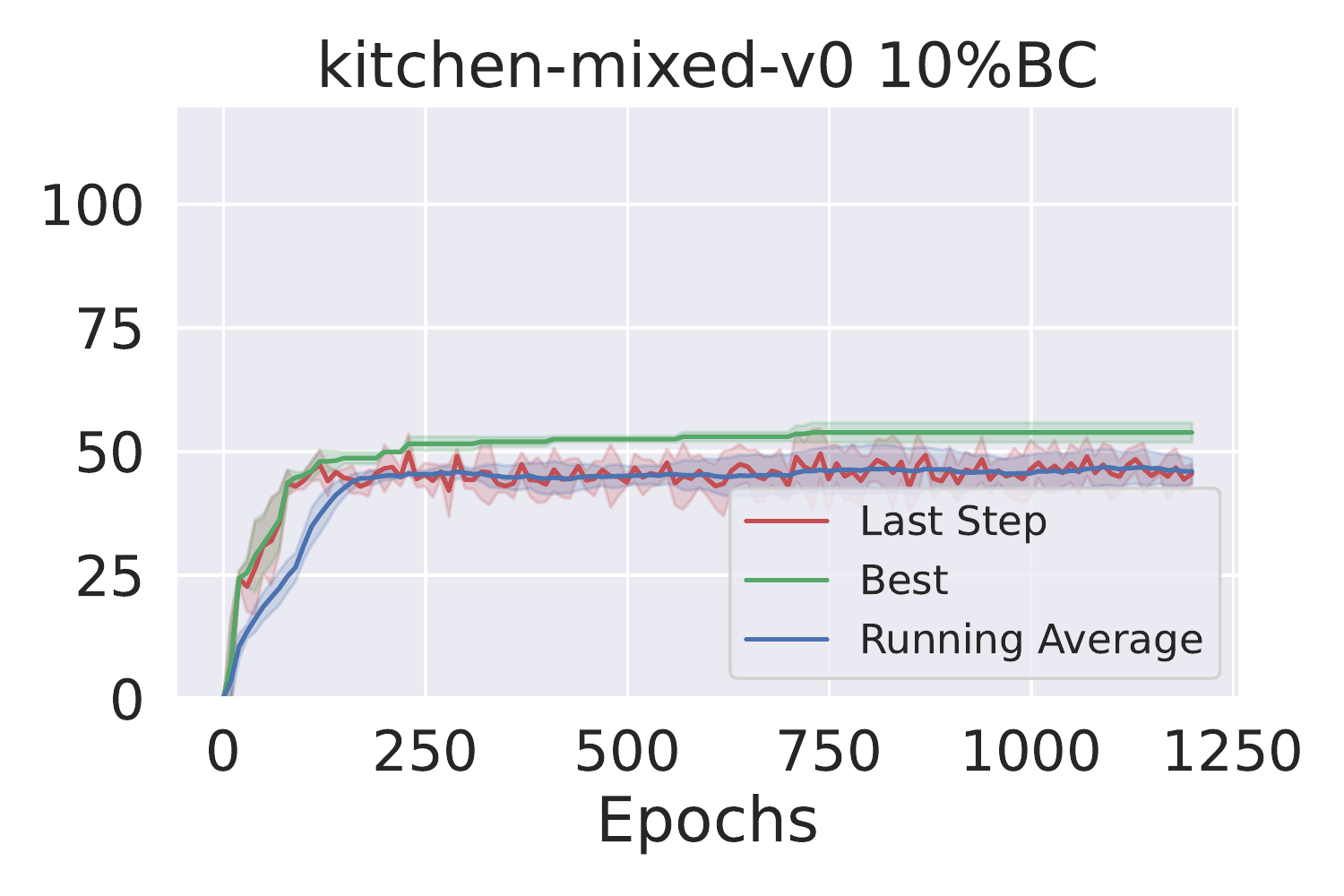}\\\includegraphics[width=0.225\linewidth]{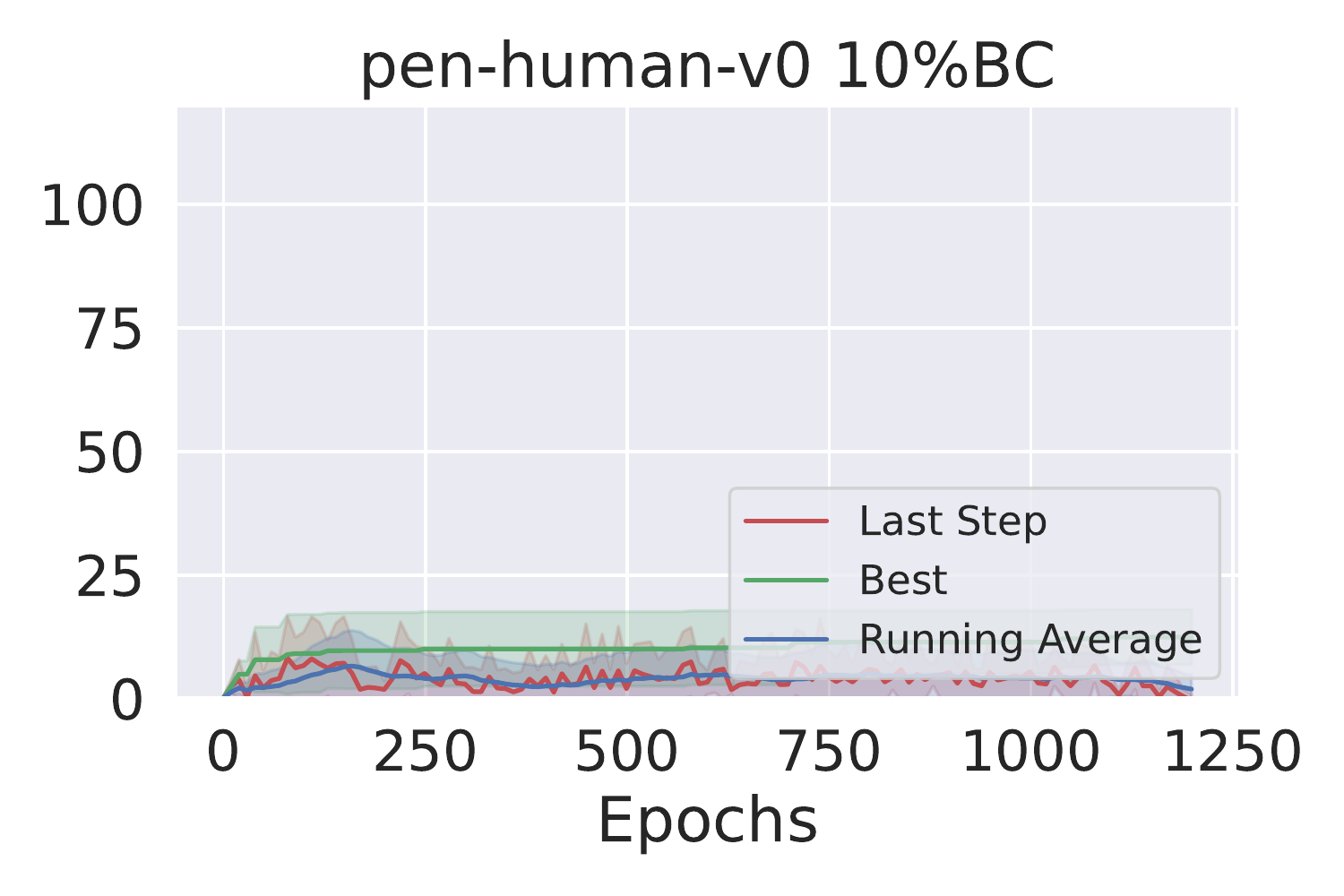}&\includegraphics[width=0.225\linewidth]{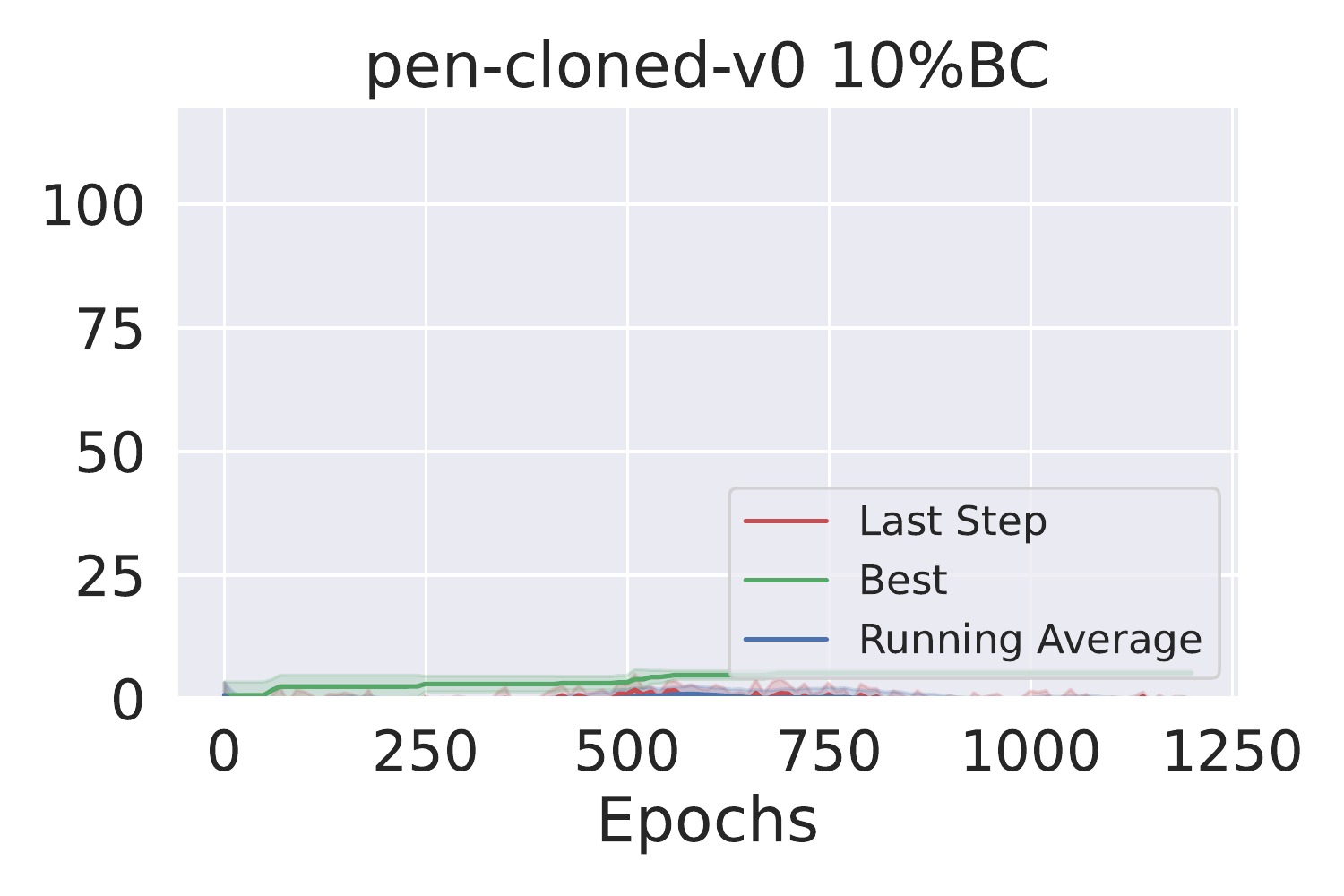}&\includegraphics[width=0.225\linewidth]{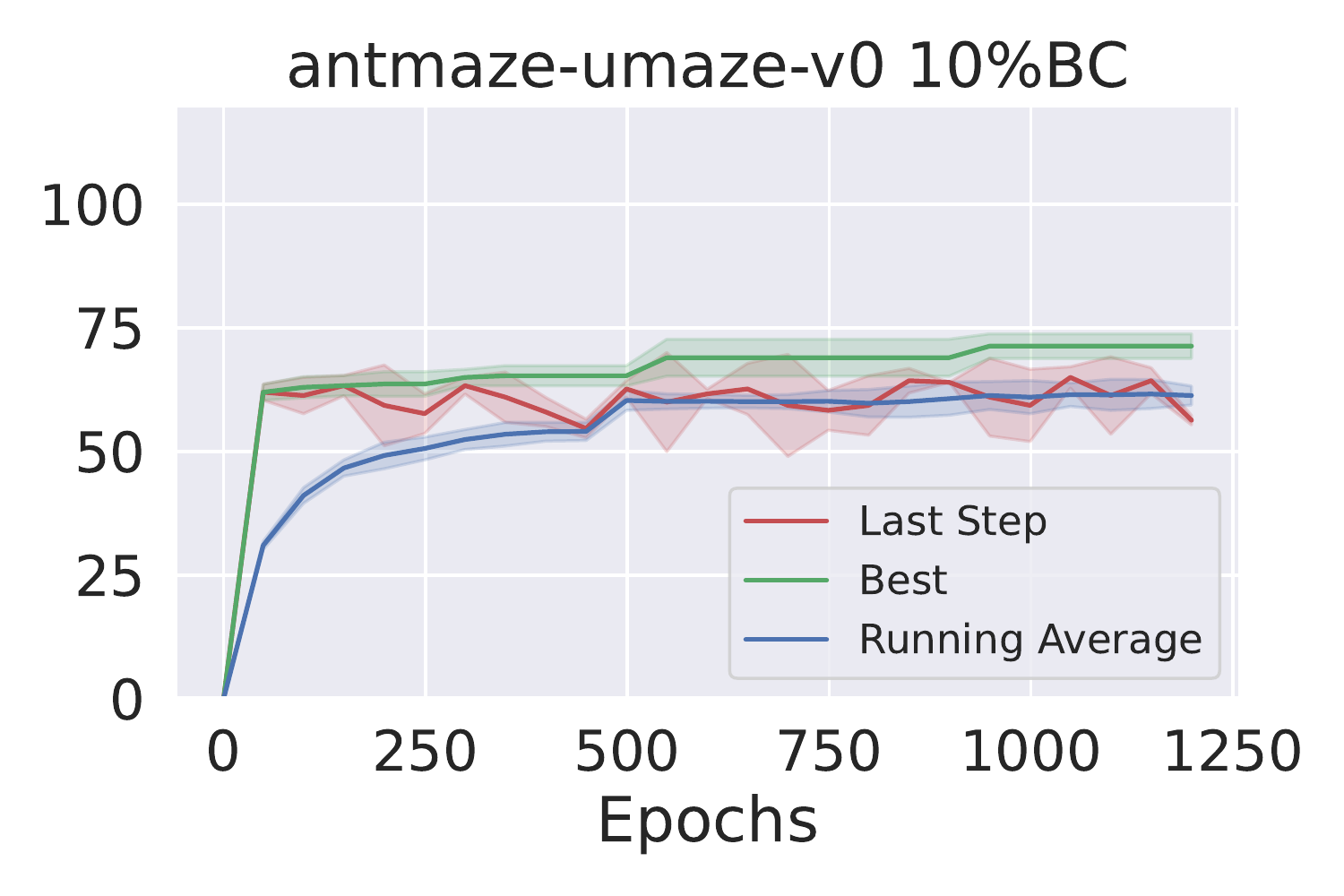}&\includegraphics[width=0.225\linewidth]{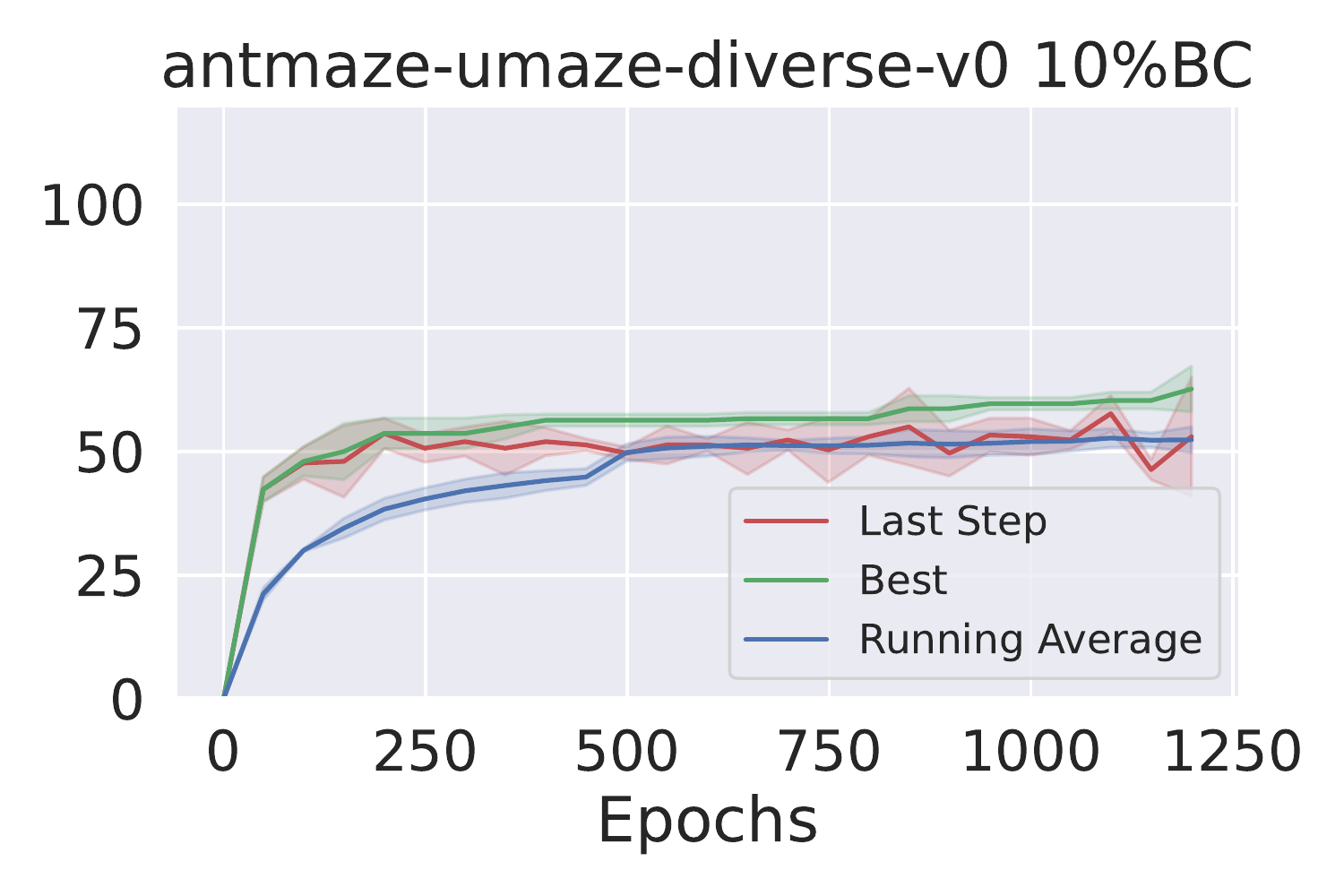}\\\includegraphics[width=0.225\linewidth]{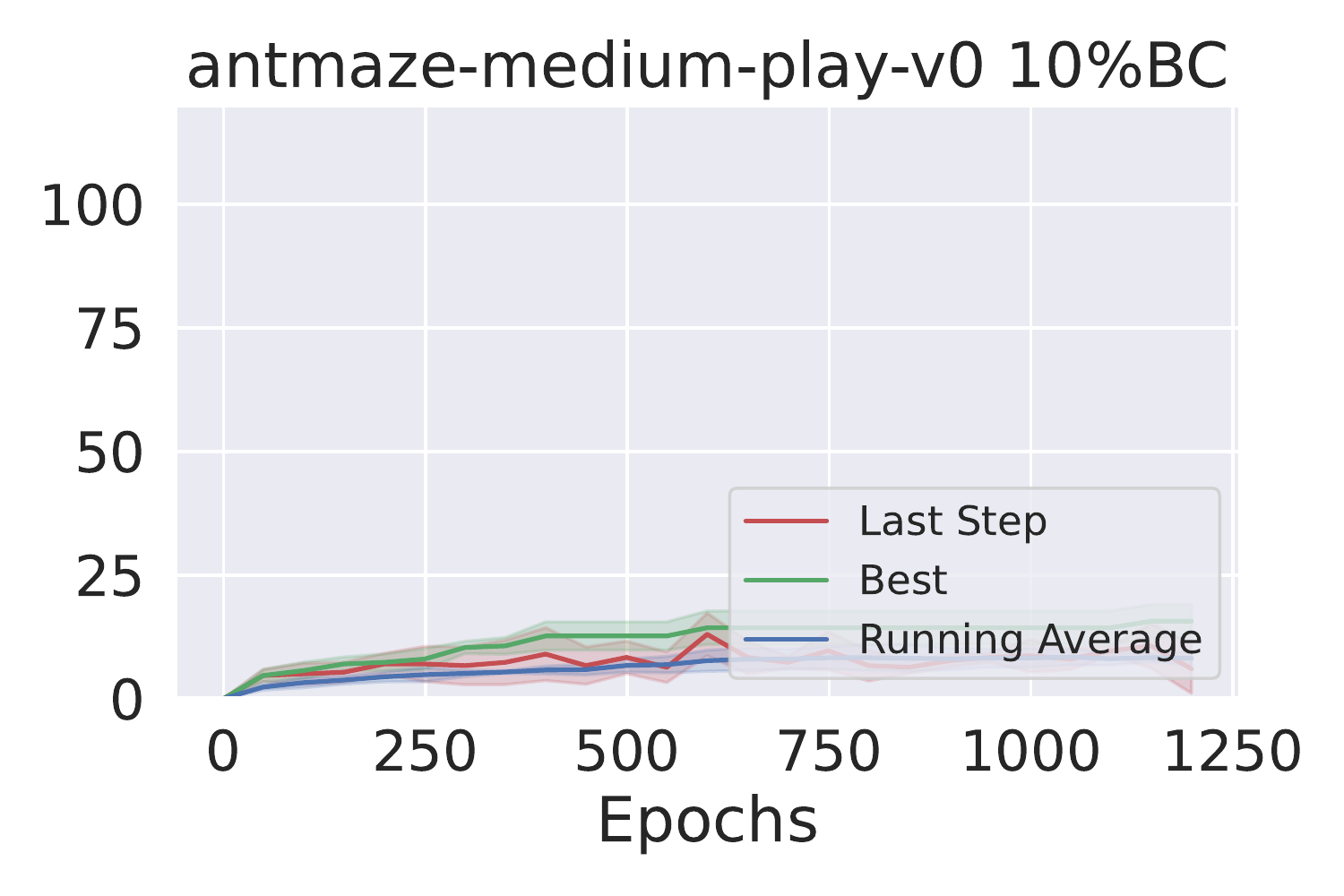}&\includegraphics[width=0.225\linewidth]{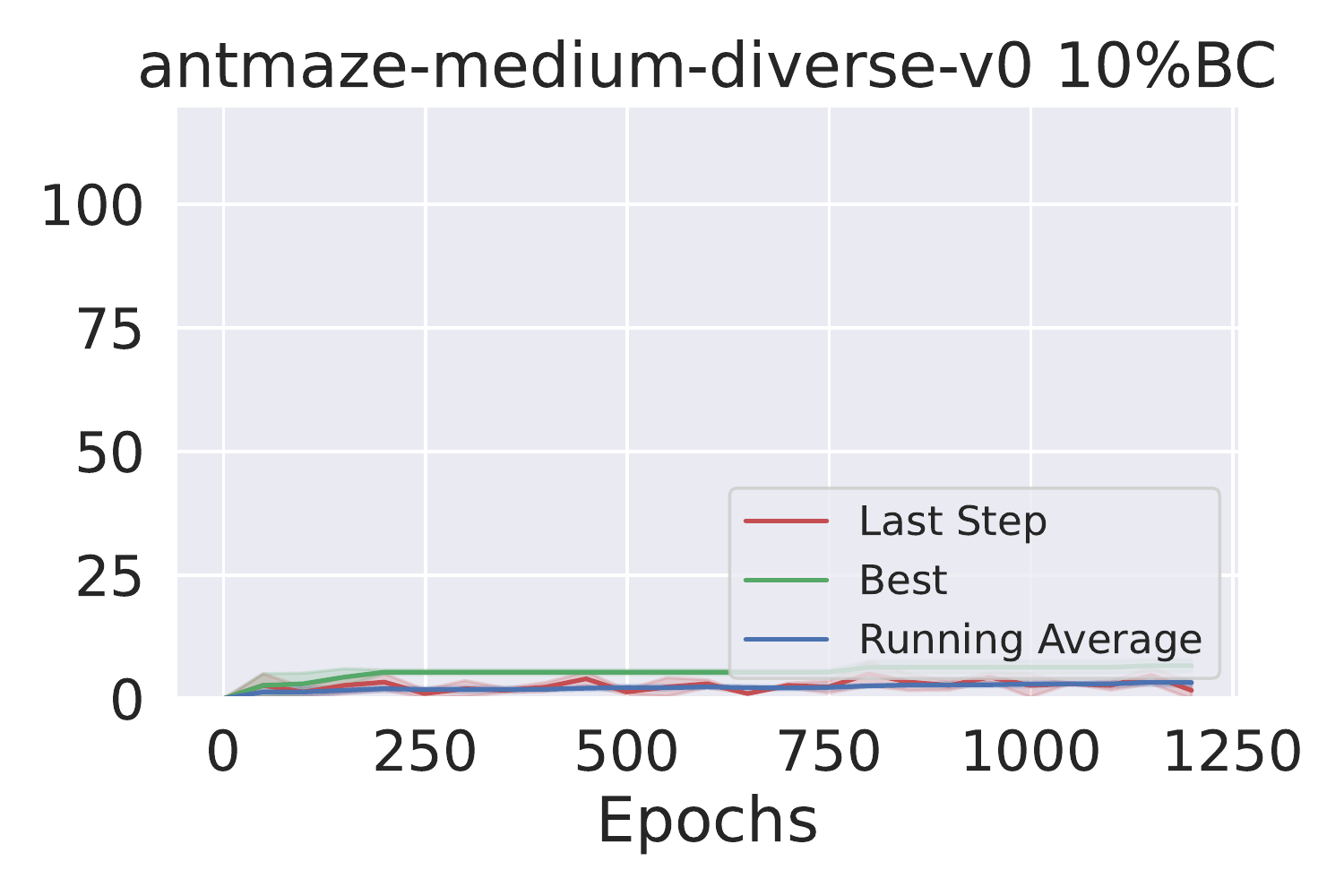}&\includegraphics[width=0.225\linewidth]{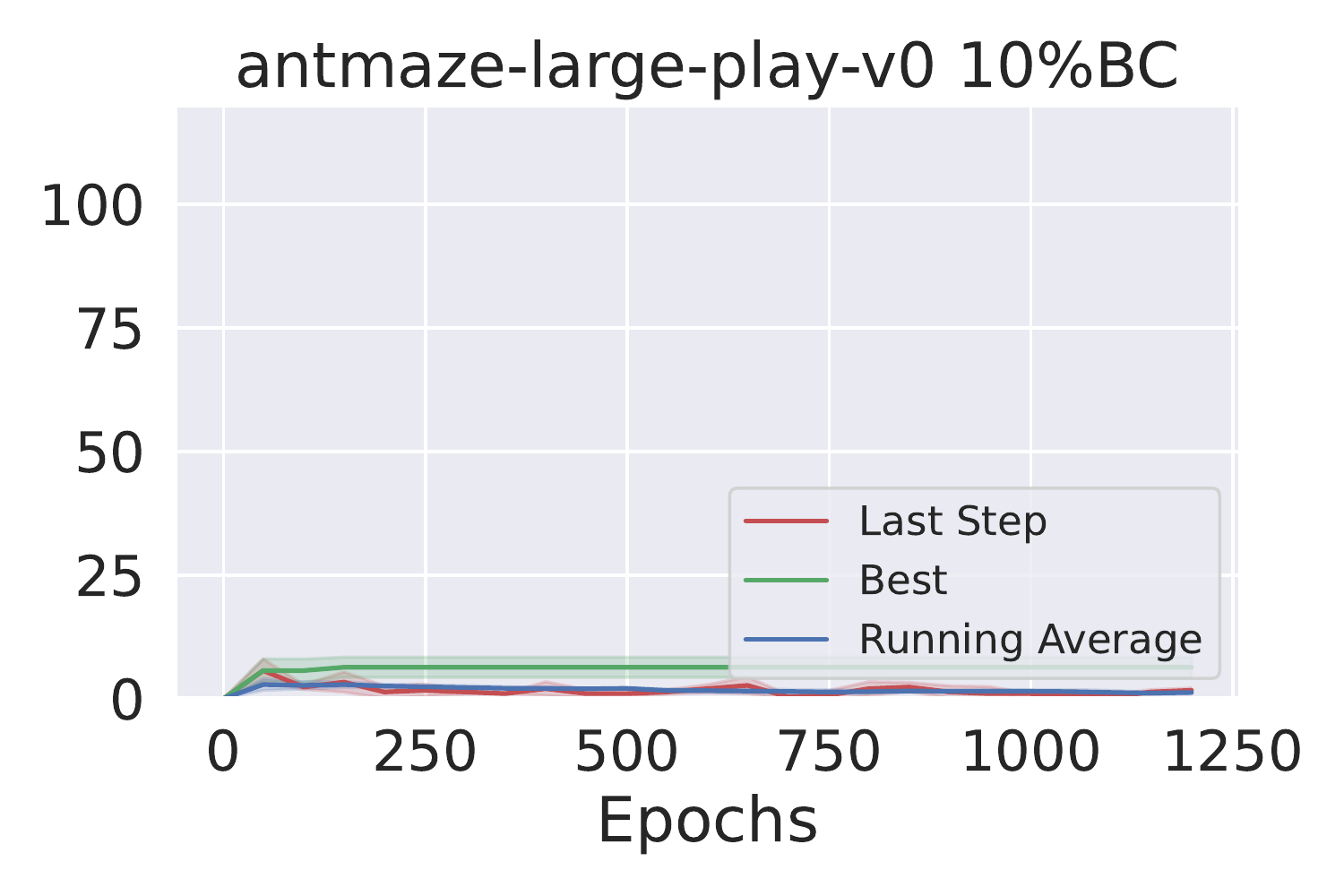}&\includegraphics[width=0.225\linewidth]{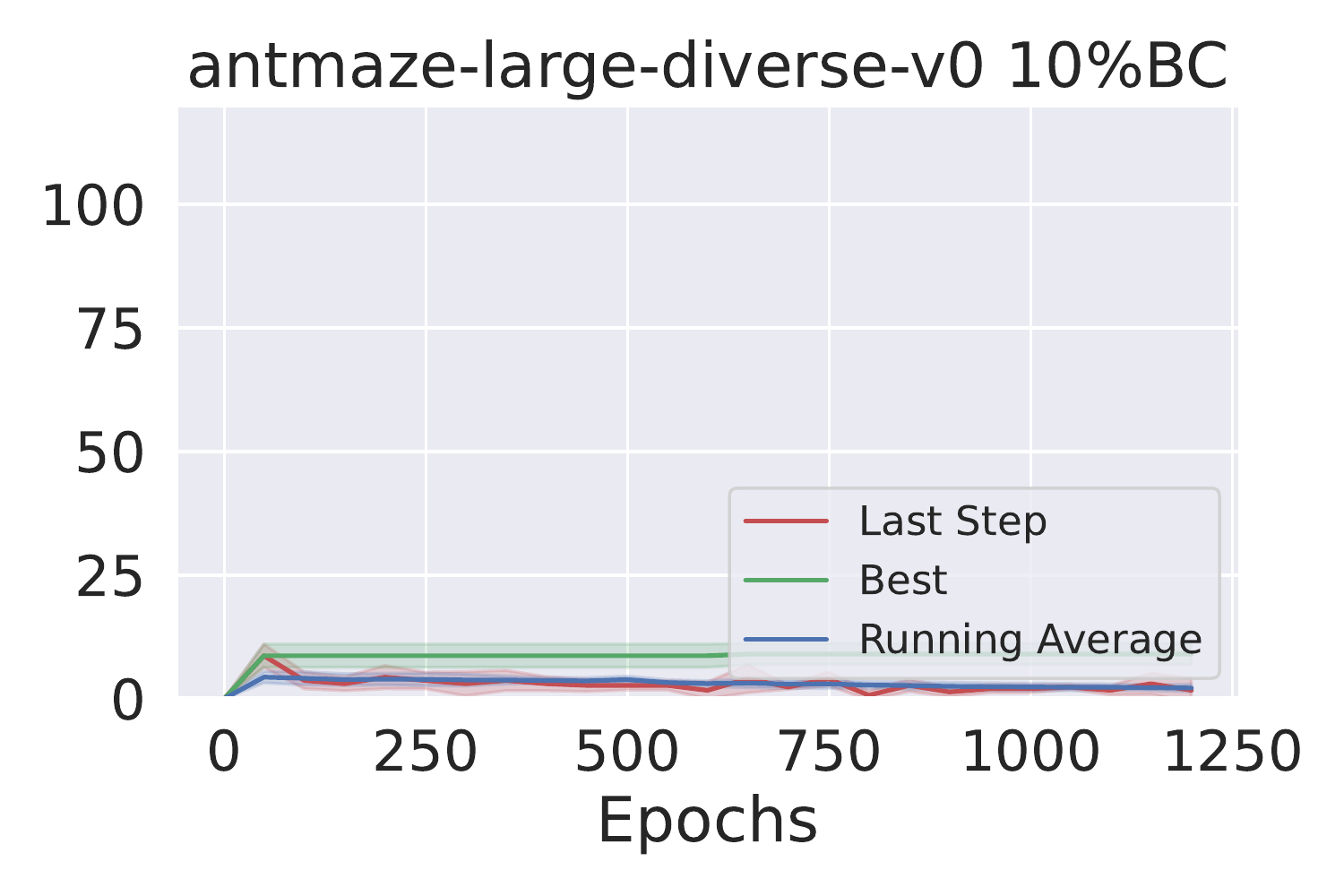}\\\end{tabular}
\centering
\caption{Training Curves of 10\%BC on D4RL}
\end{figure*}
\begin{figure*}[htb]
\centering
\begin{tabular}{cccccc}
\includegraphics[width=0.225\linewidth]{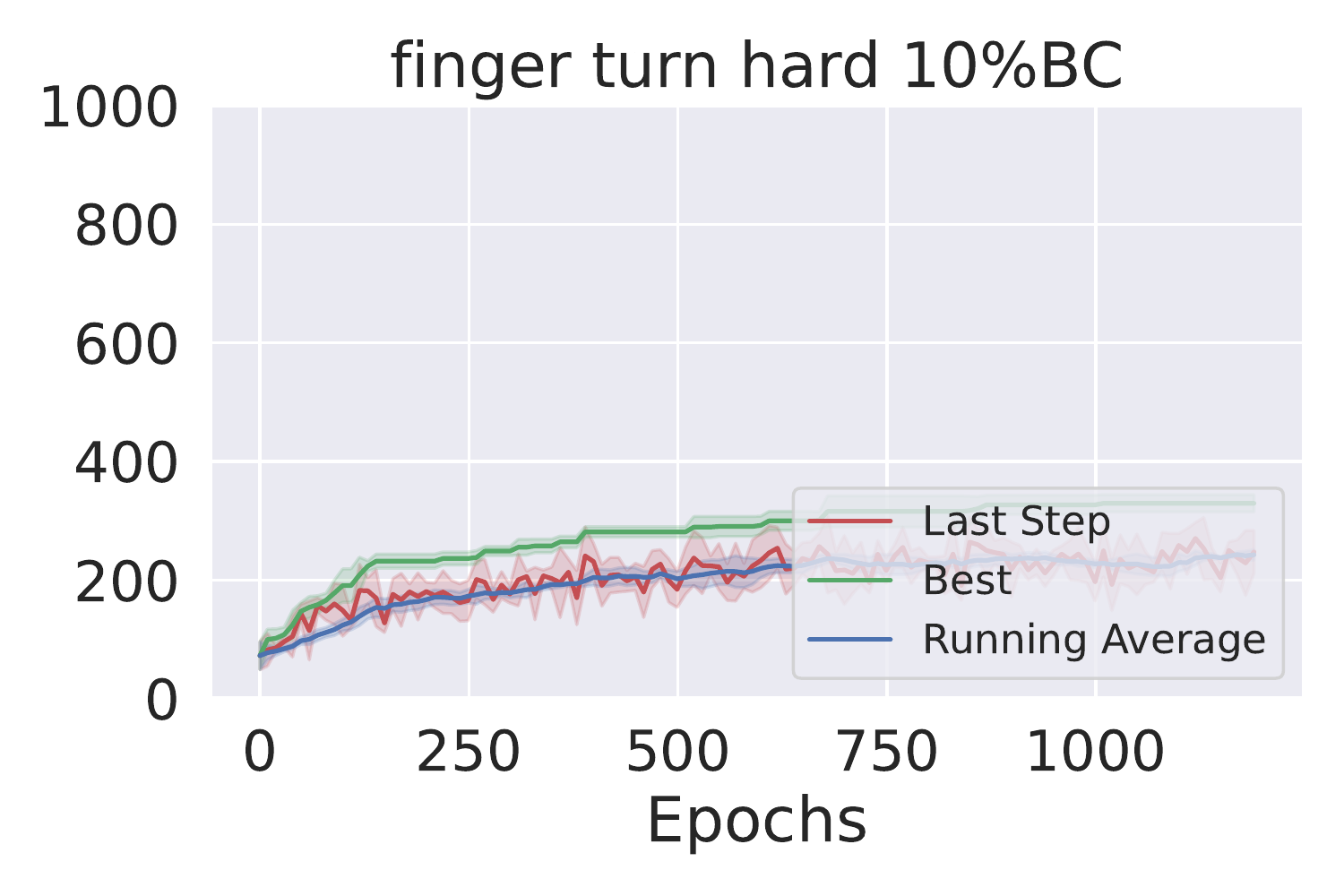}&\includegraphics[width=0.225\linewidth]{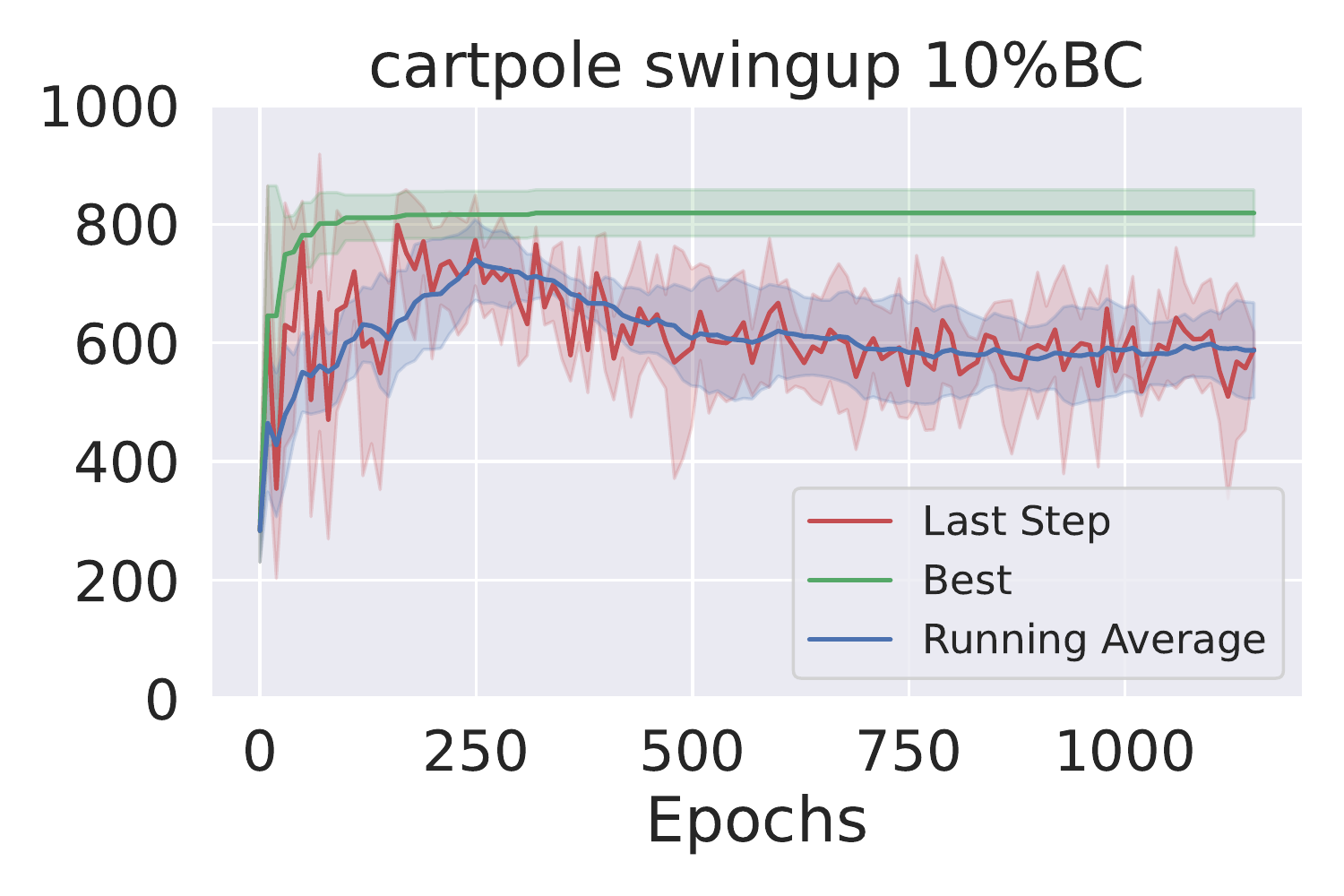}&\includegraphics[width=0.225\linewidth]{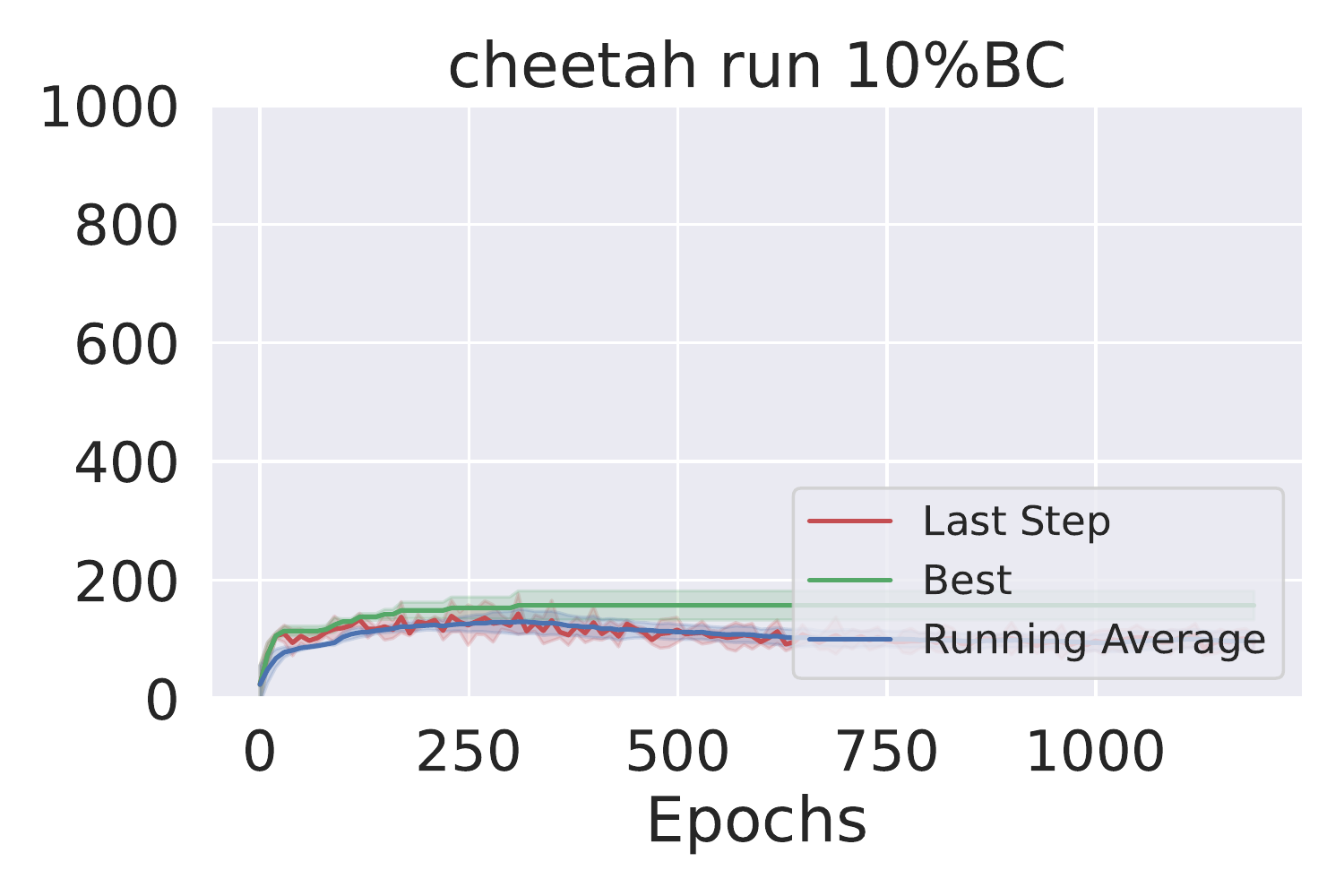}&\includegraphics[width=0.225\linewidth]{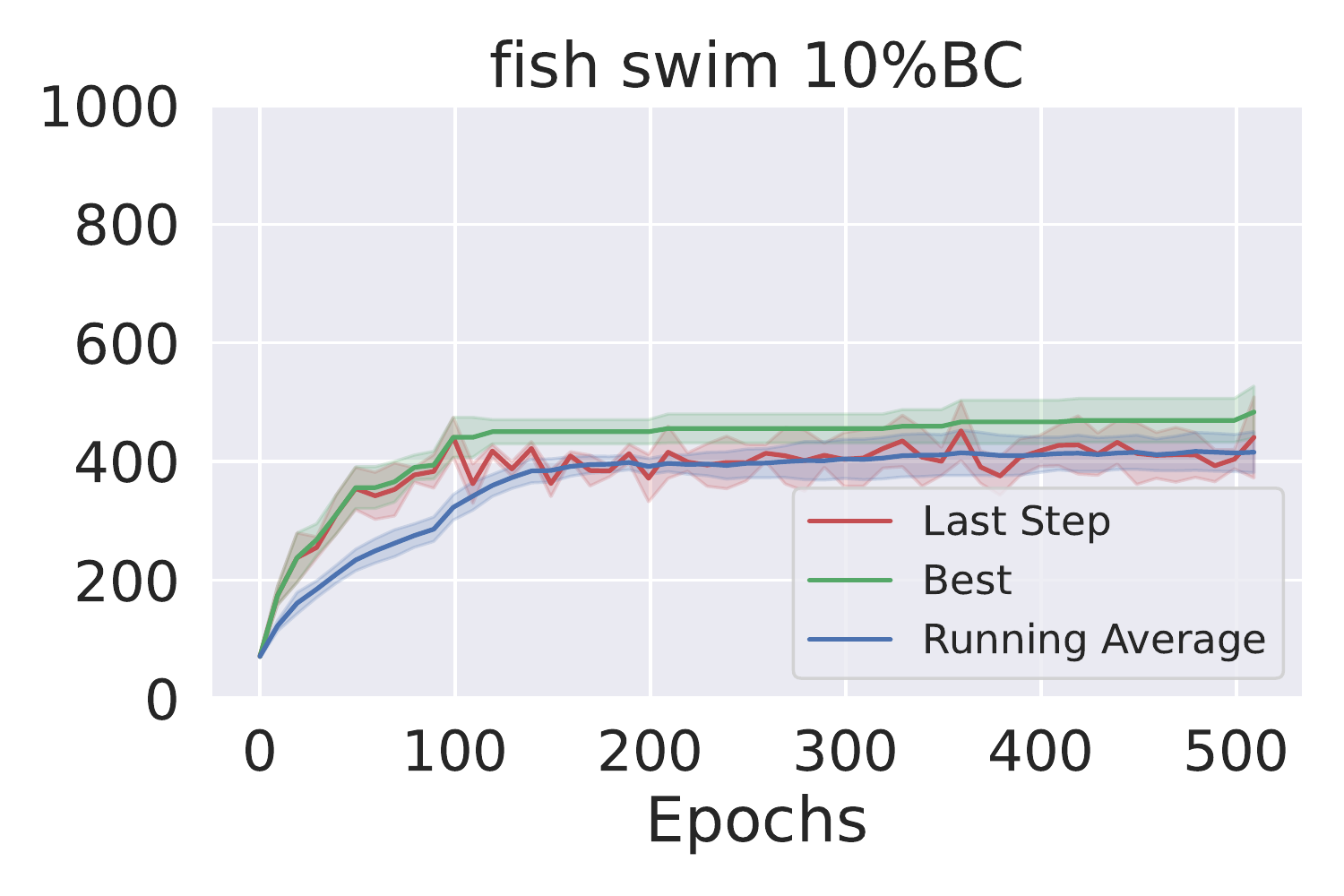}\\\includegraphics[width=0.225\linewidth]{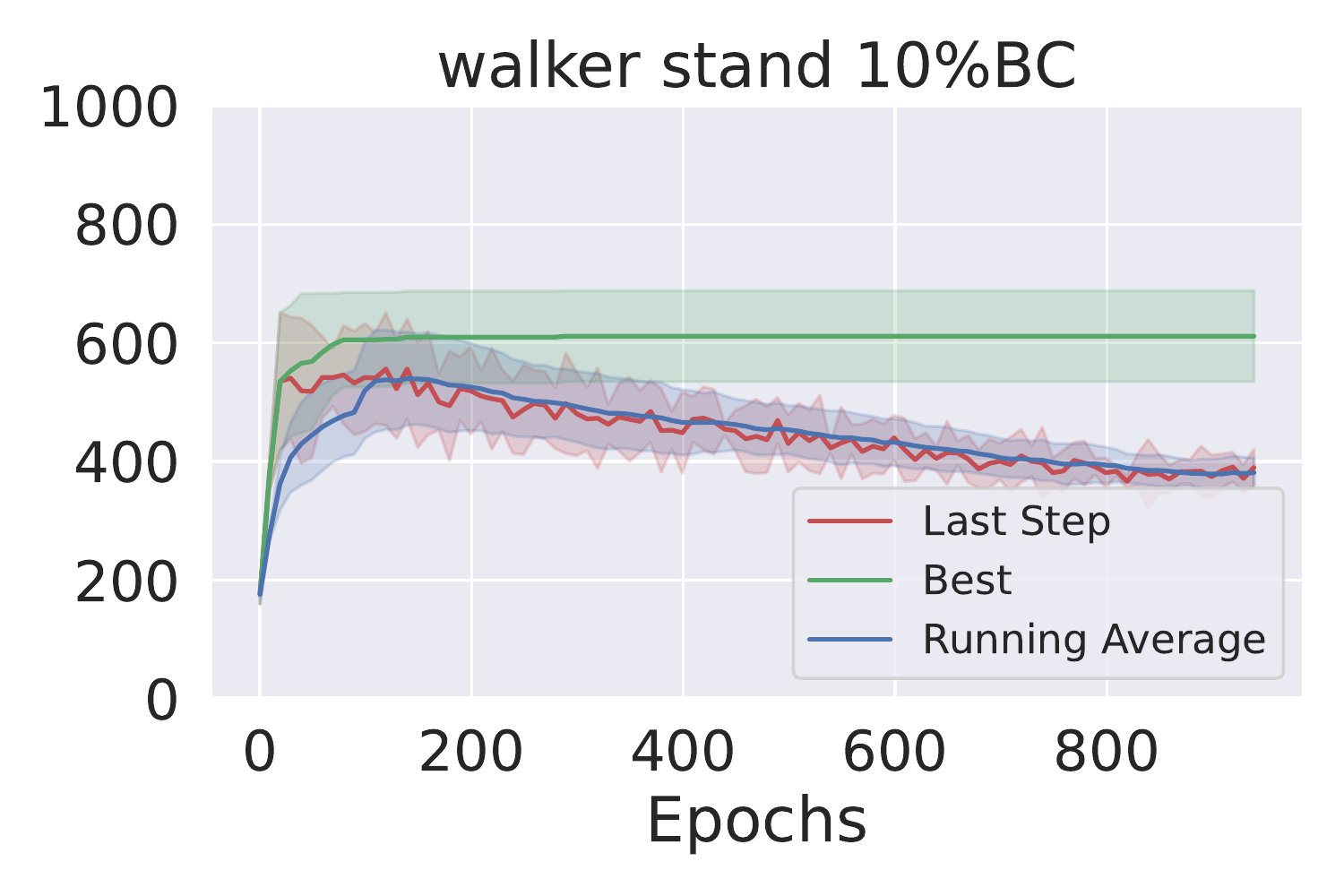}&\includegraphics[width=0.225\linewidth]{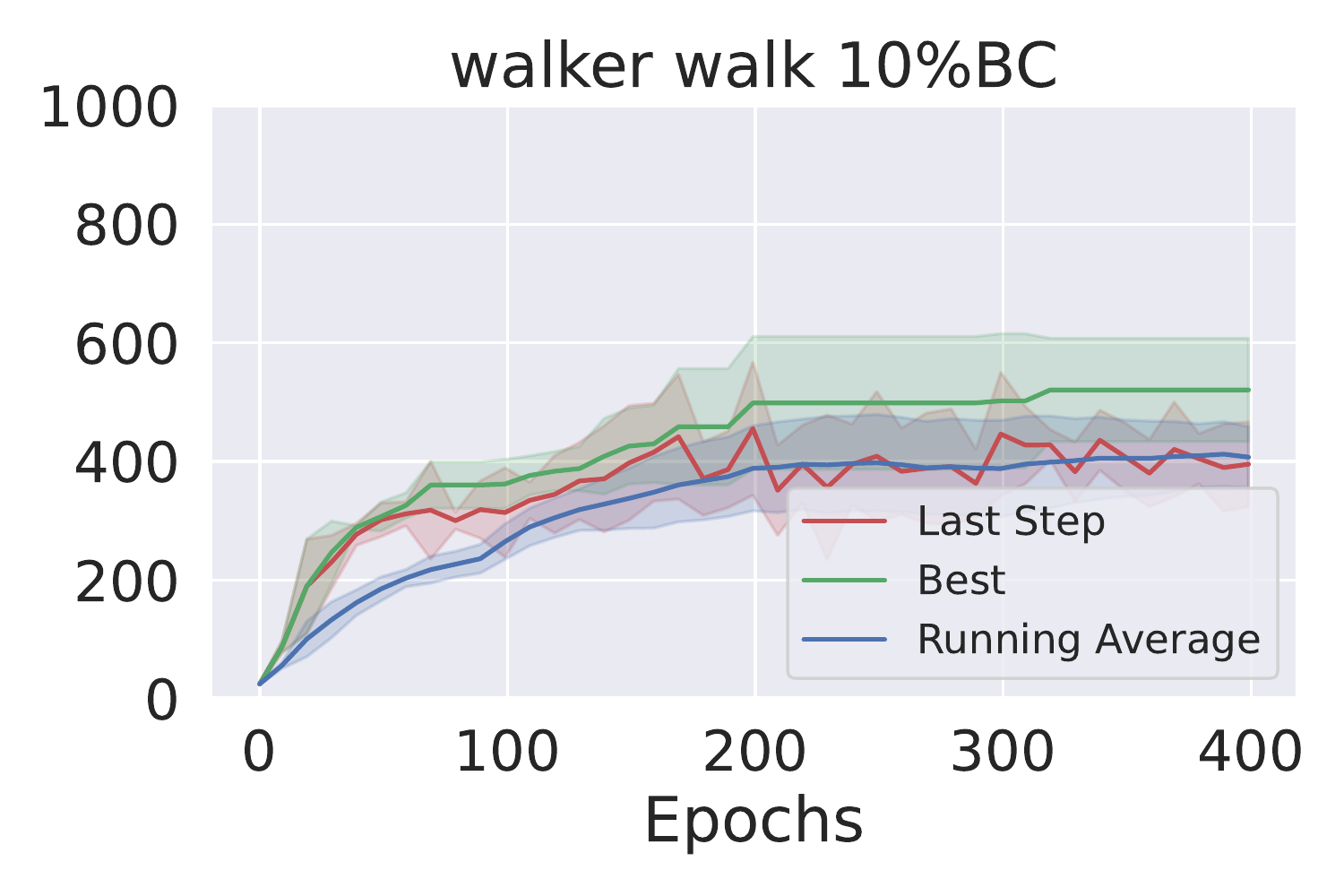}&&\\\end{tabular}
\centering
\caption{Training Curves of 10\%BC on RLUP}
\end{figure*}
\begin{figure*}[htb]
\centering
\begin{tabular}{cccccc}
\includegraphics[width=0.225\linewidth]{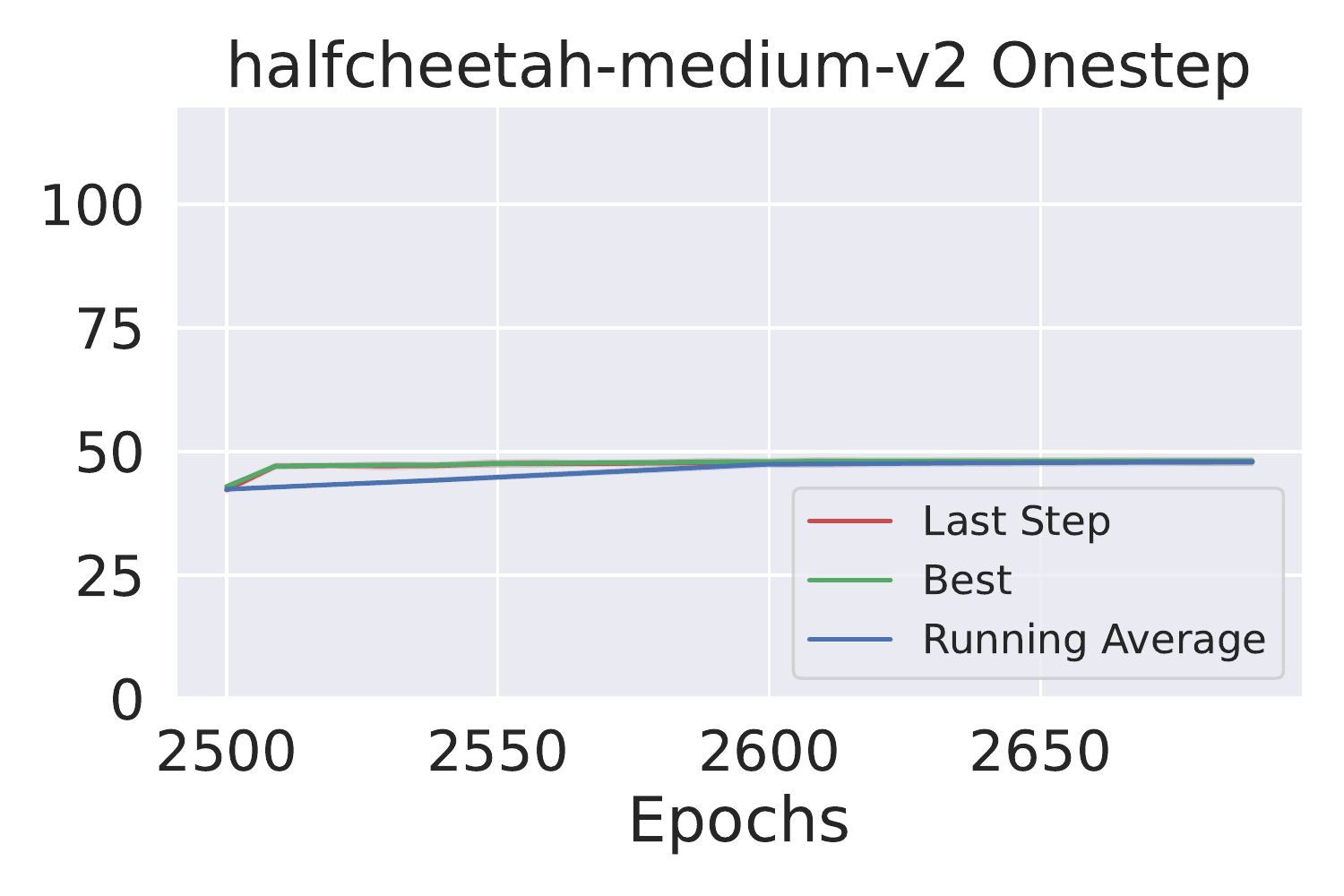}&\includegraphics[width=0.225\linewidth]{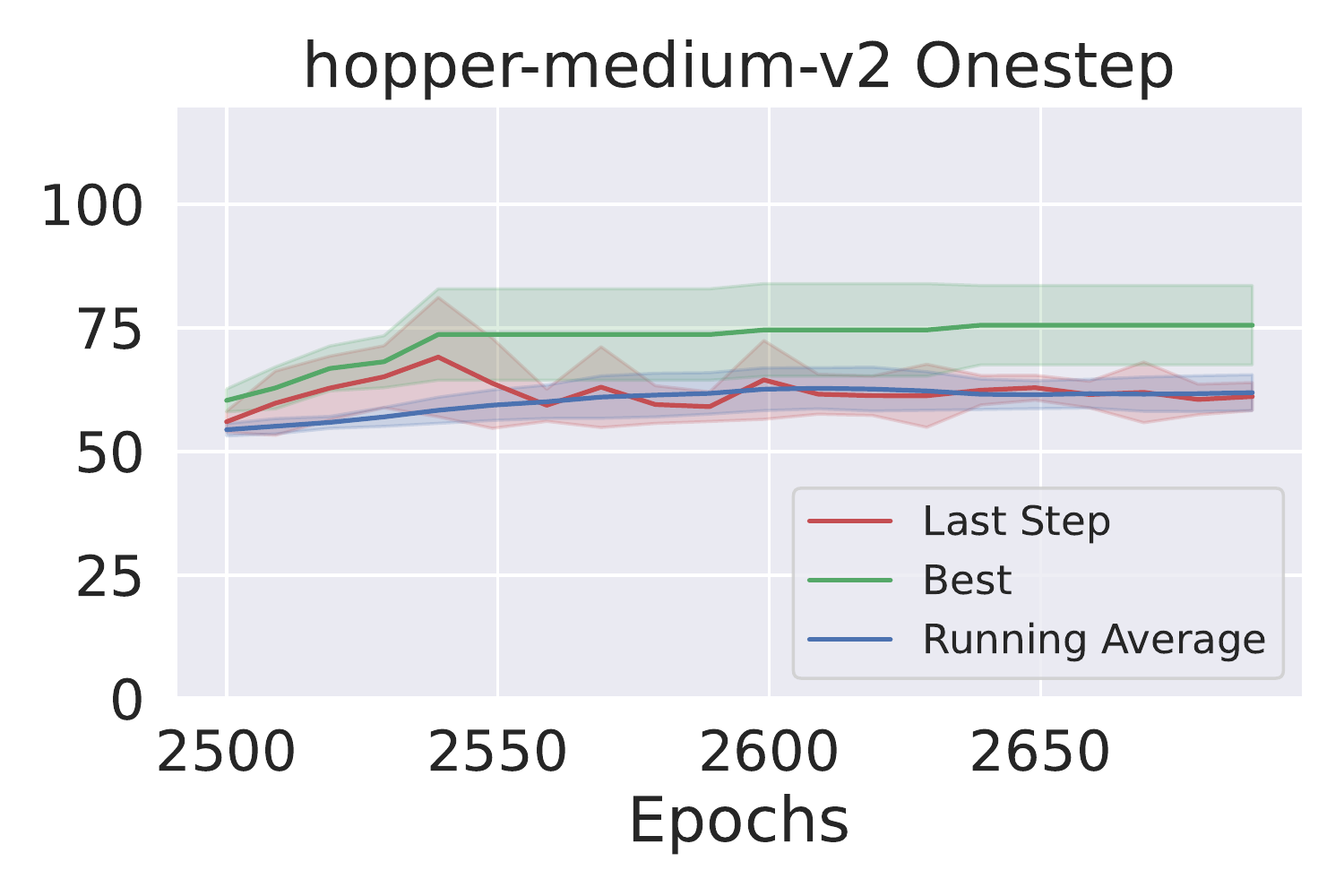}&\includegraphics[width=0.225\linewidth]{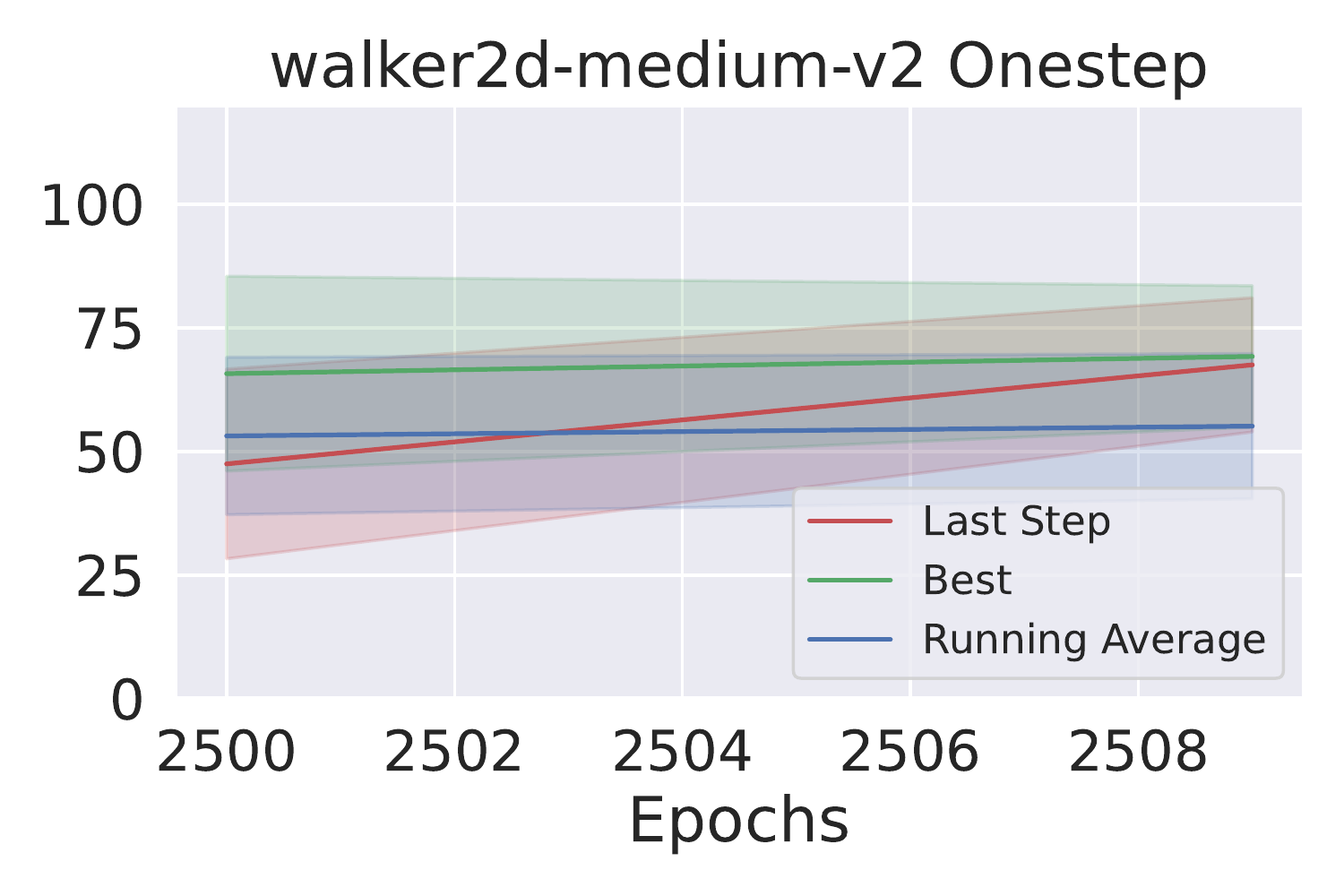}&\includegraphics[width=0.225\linewidth]{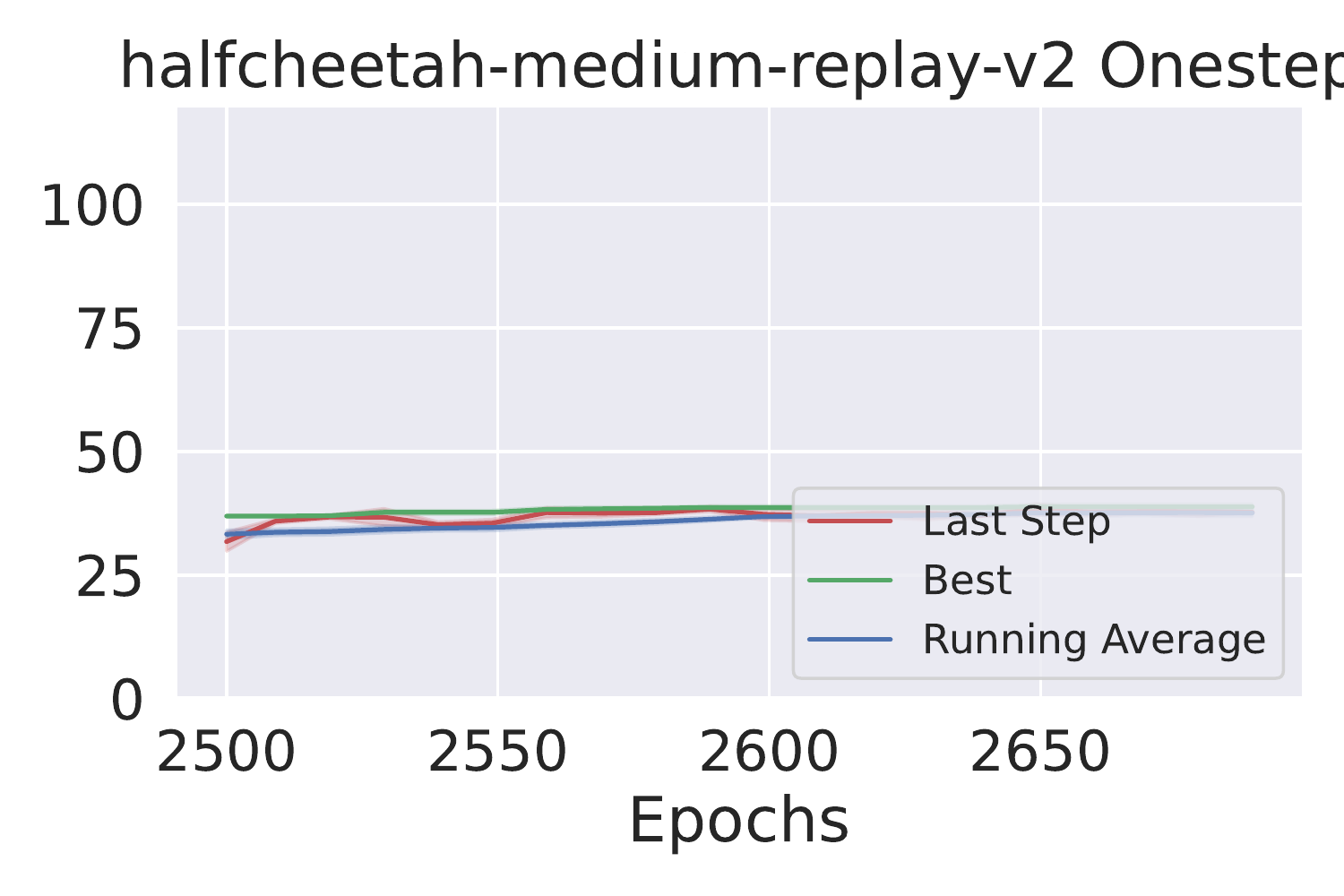}\\\includegraphics[width=0.225\linewidth]{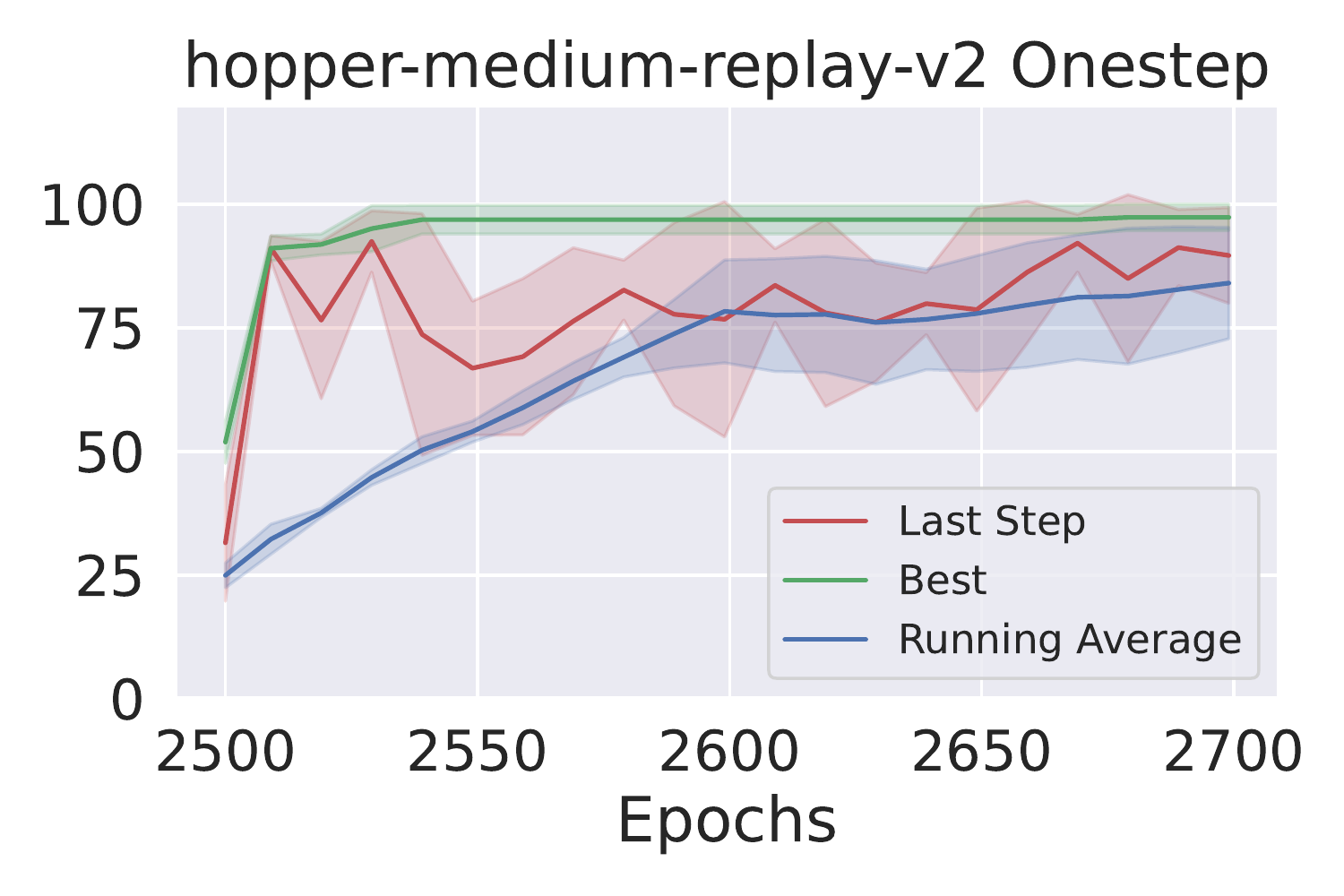}&\includegraphics[width=0.225\linewidth]{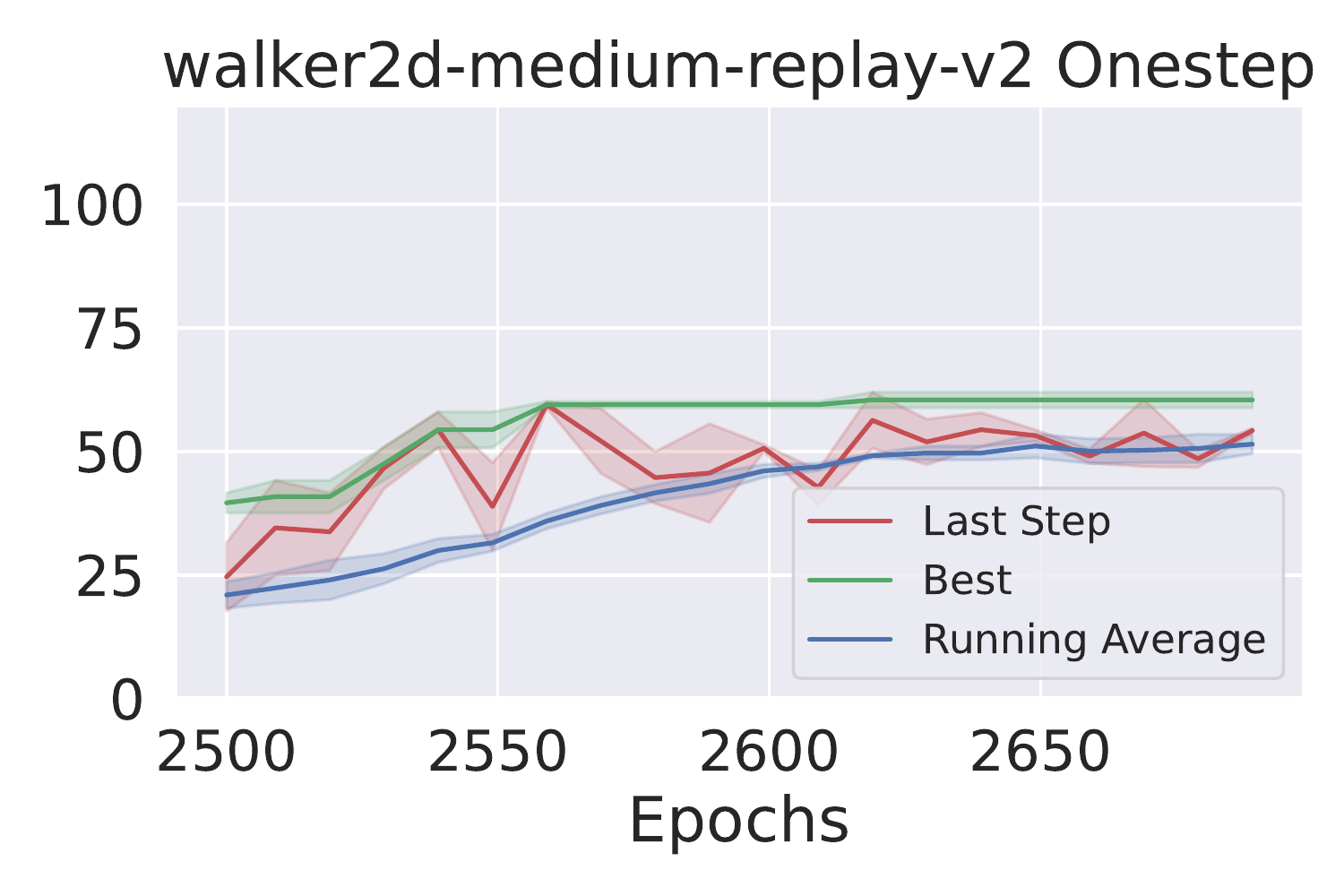}&\includegraphics[width=0.225\linewidth]{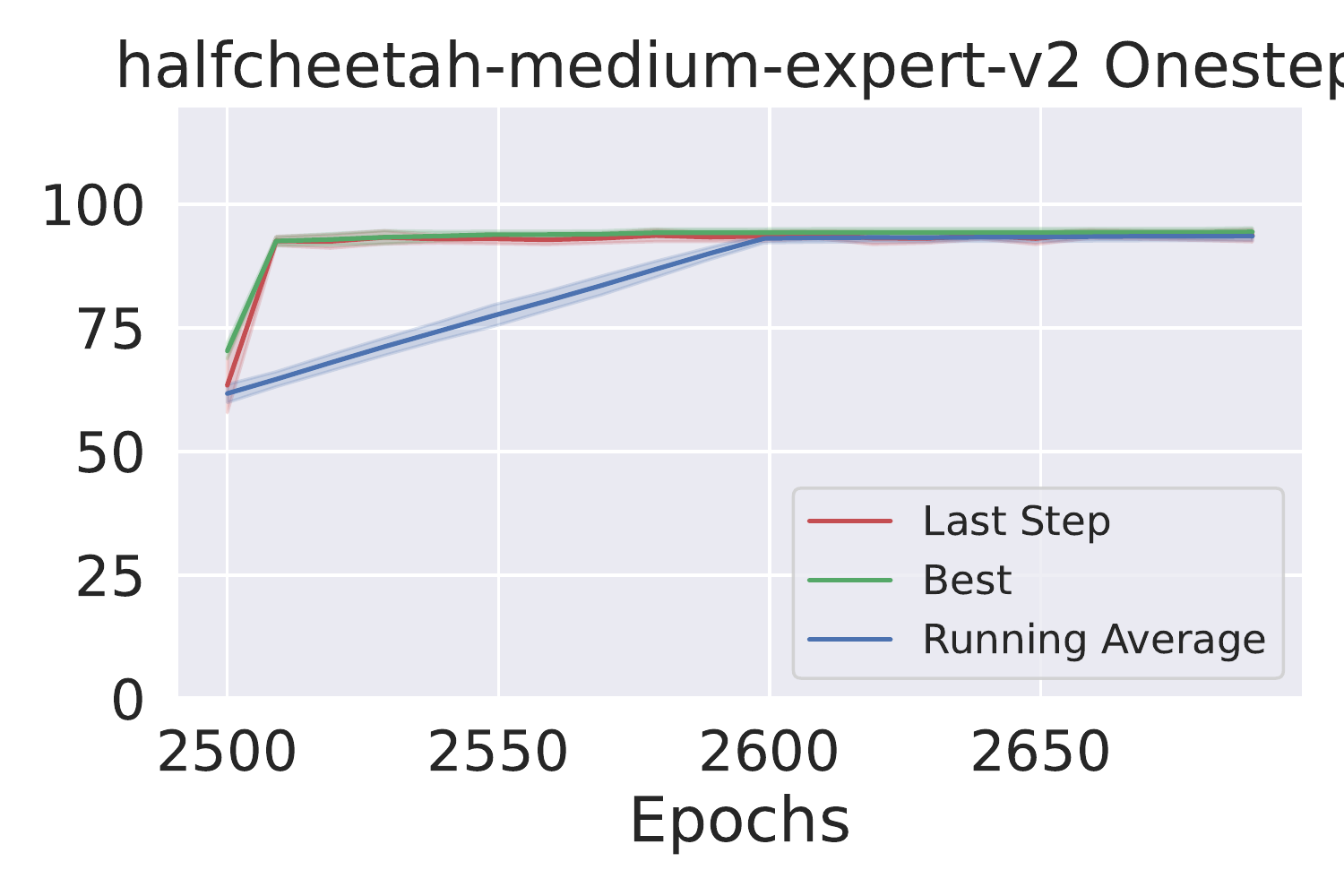}&\includegraphics[width=0.225\linewidth]{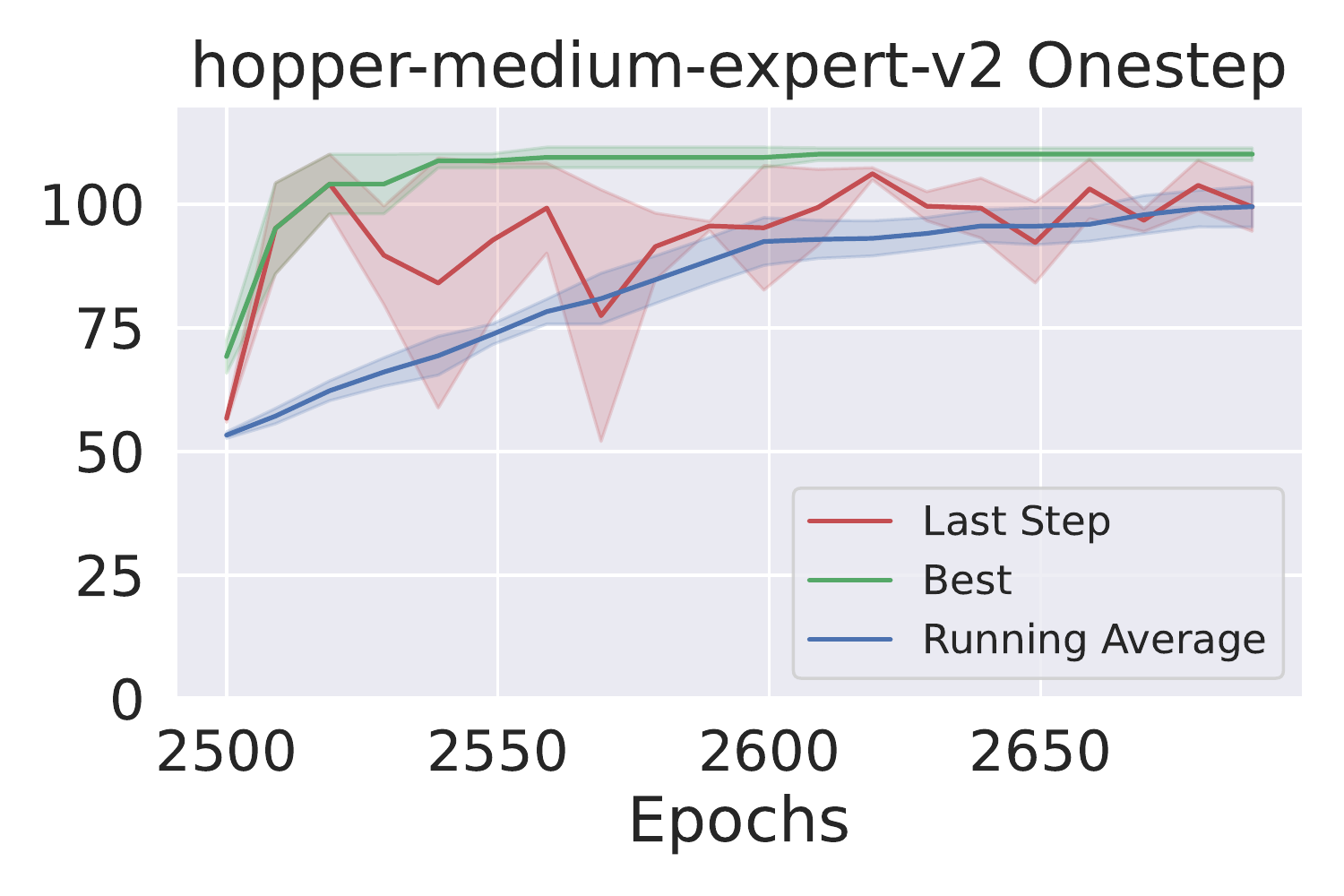}\\\includegraphics[width=0.225\linewidth]{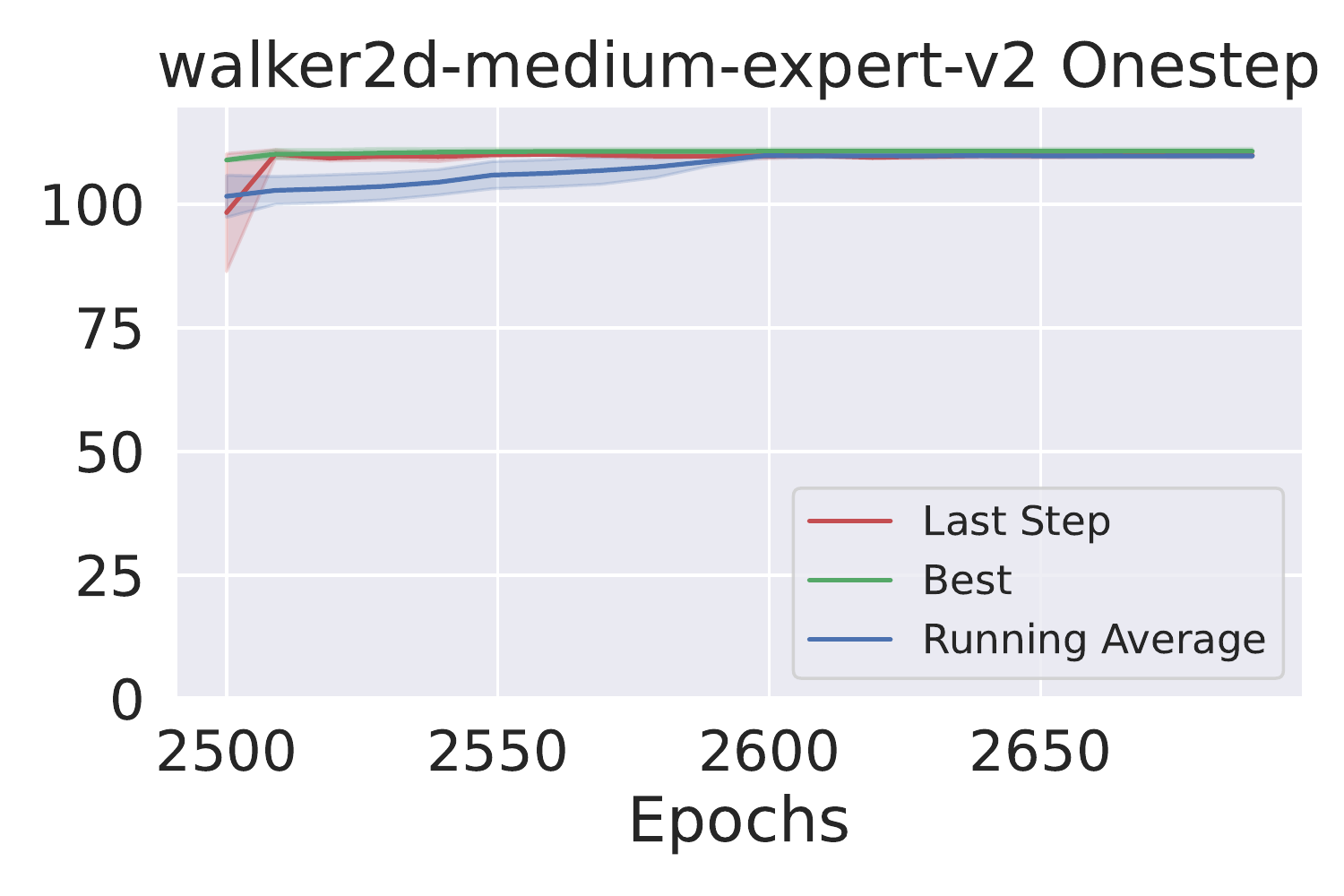}&\includegraphics[width=0.225\linewidth]{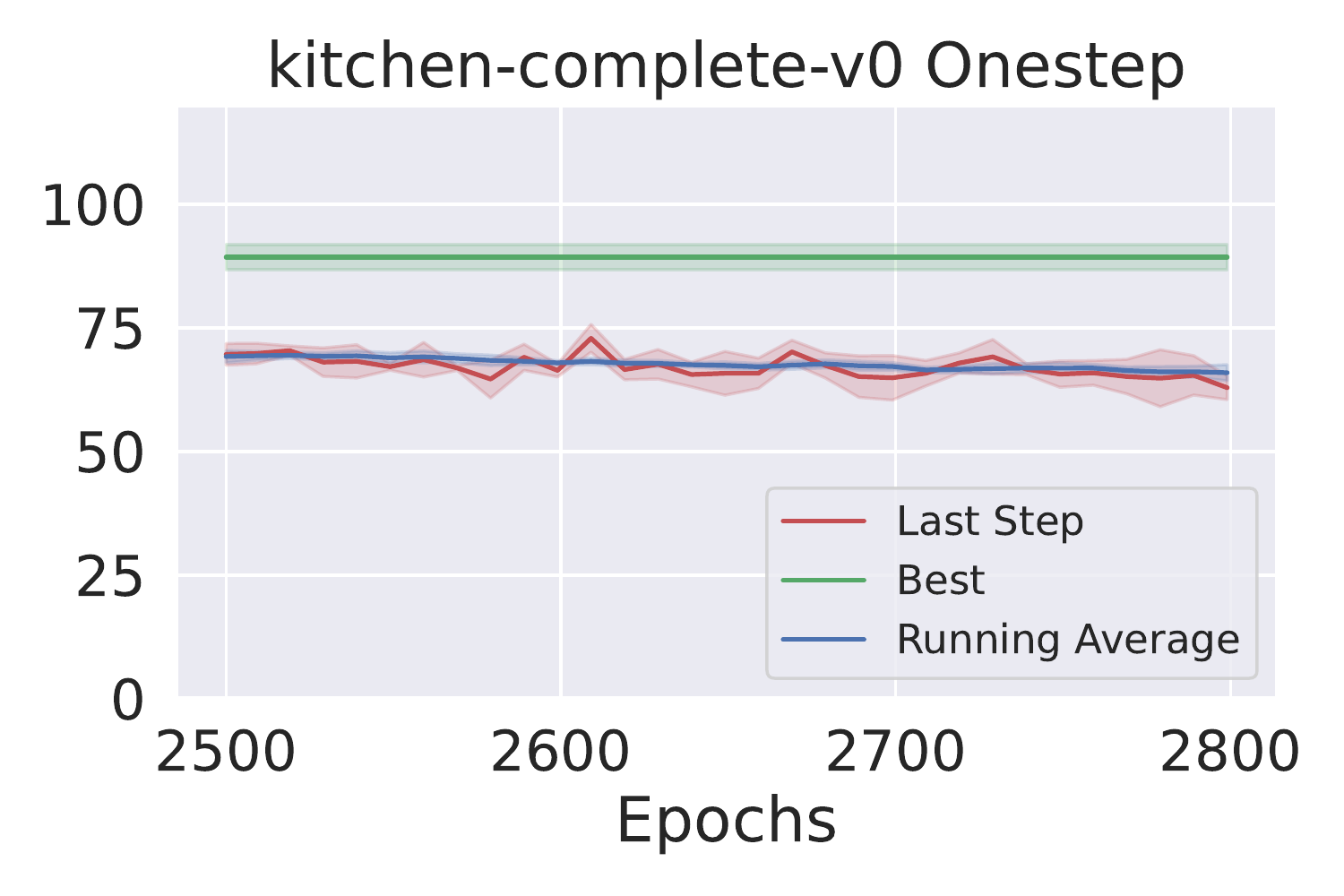}&\includegraphics[width=0.225\linewidth]{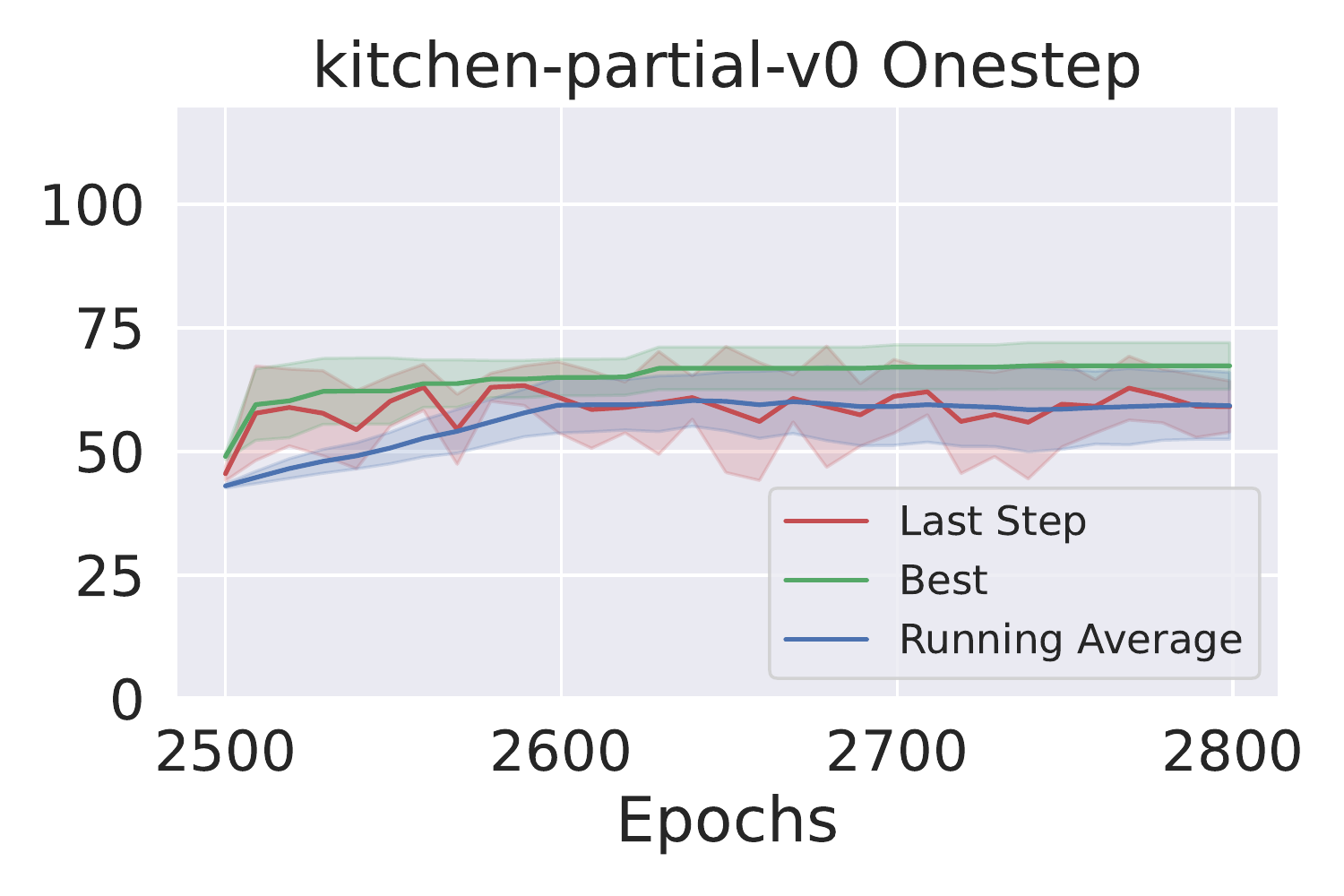}&\includegraphics[width=0.225\linewidth]{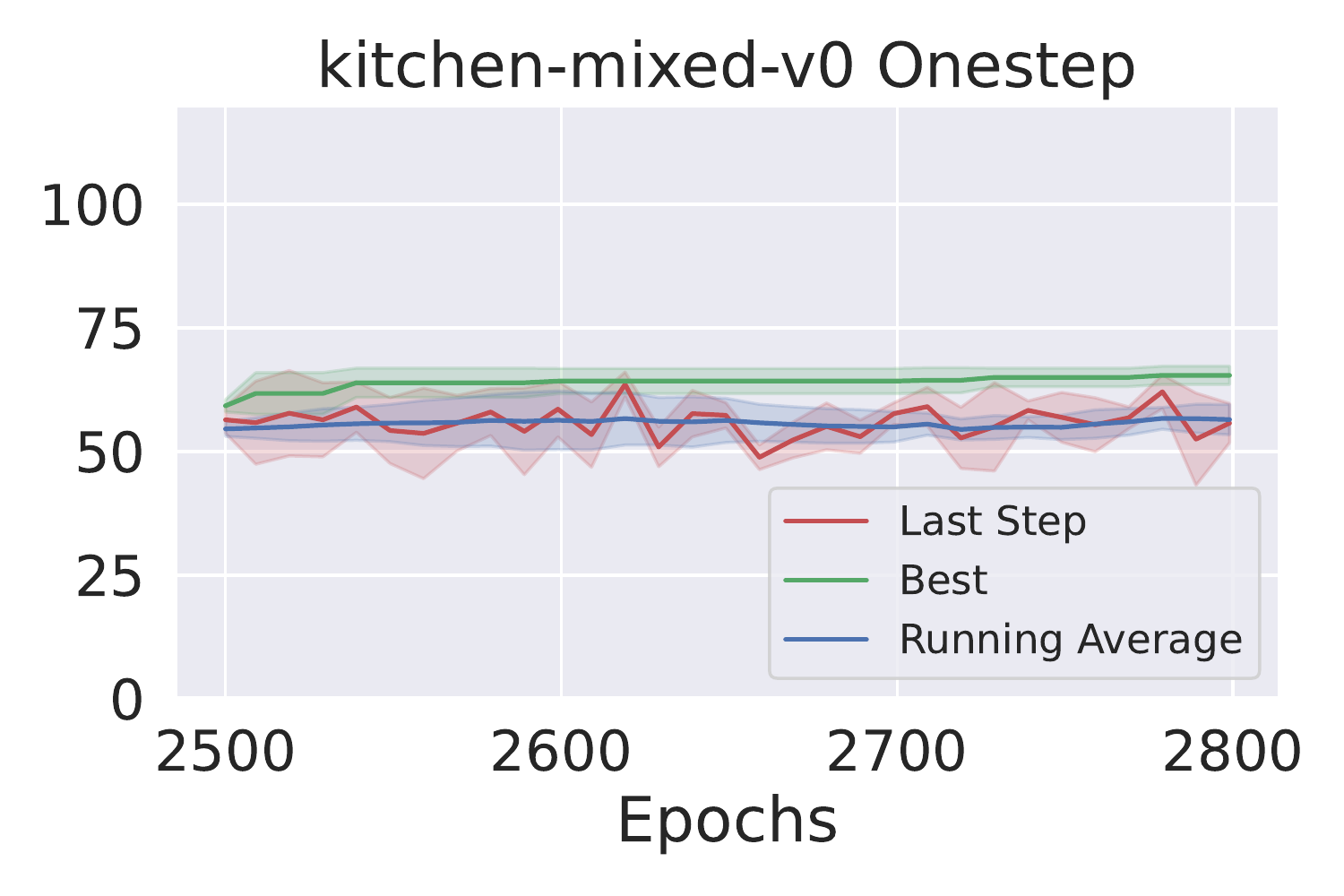}\\\includegraphics[width=0.225\linewidth]{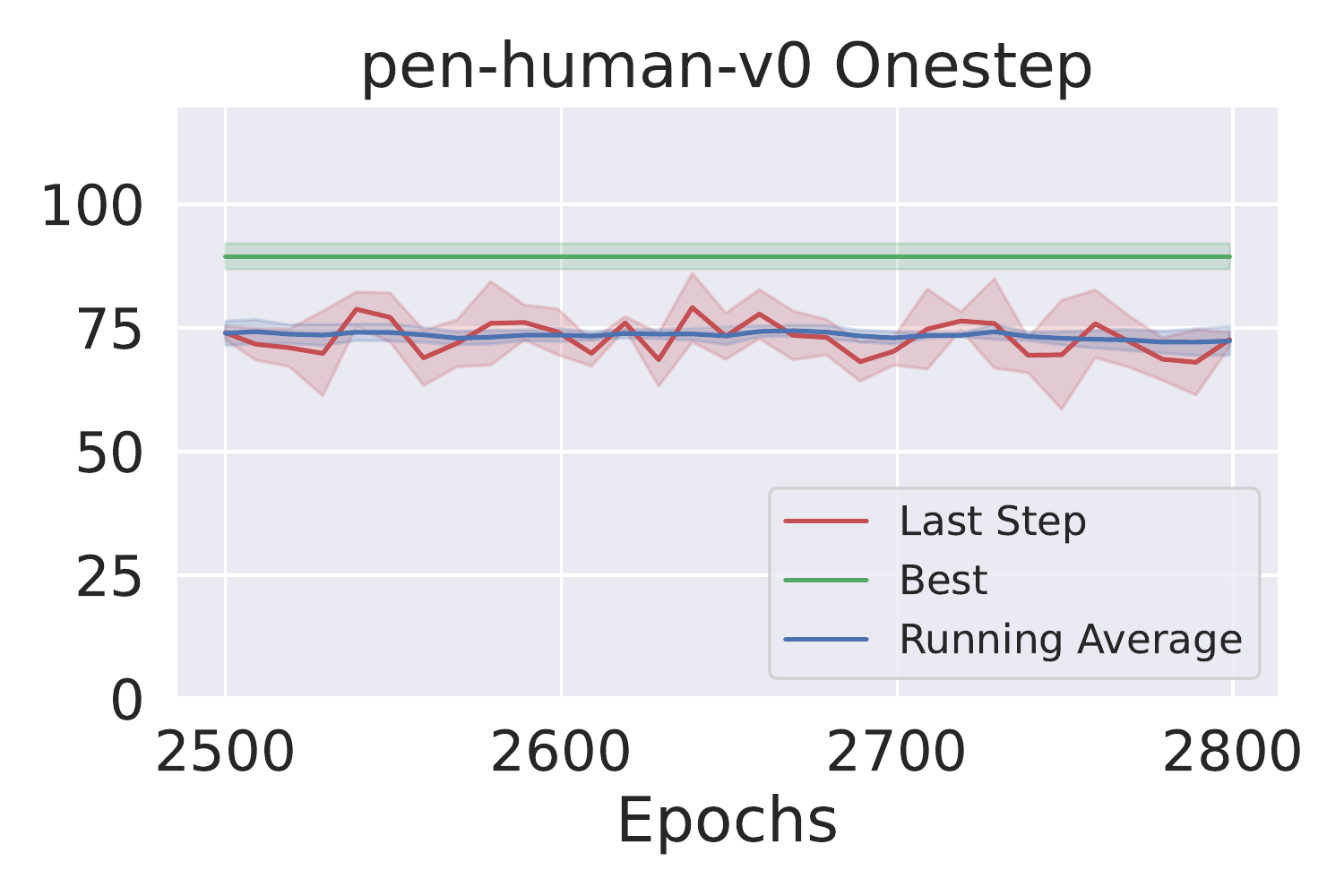}&\includegraphics[width=0.225\linewidth]{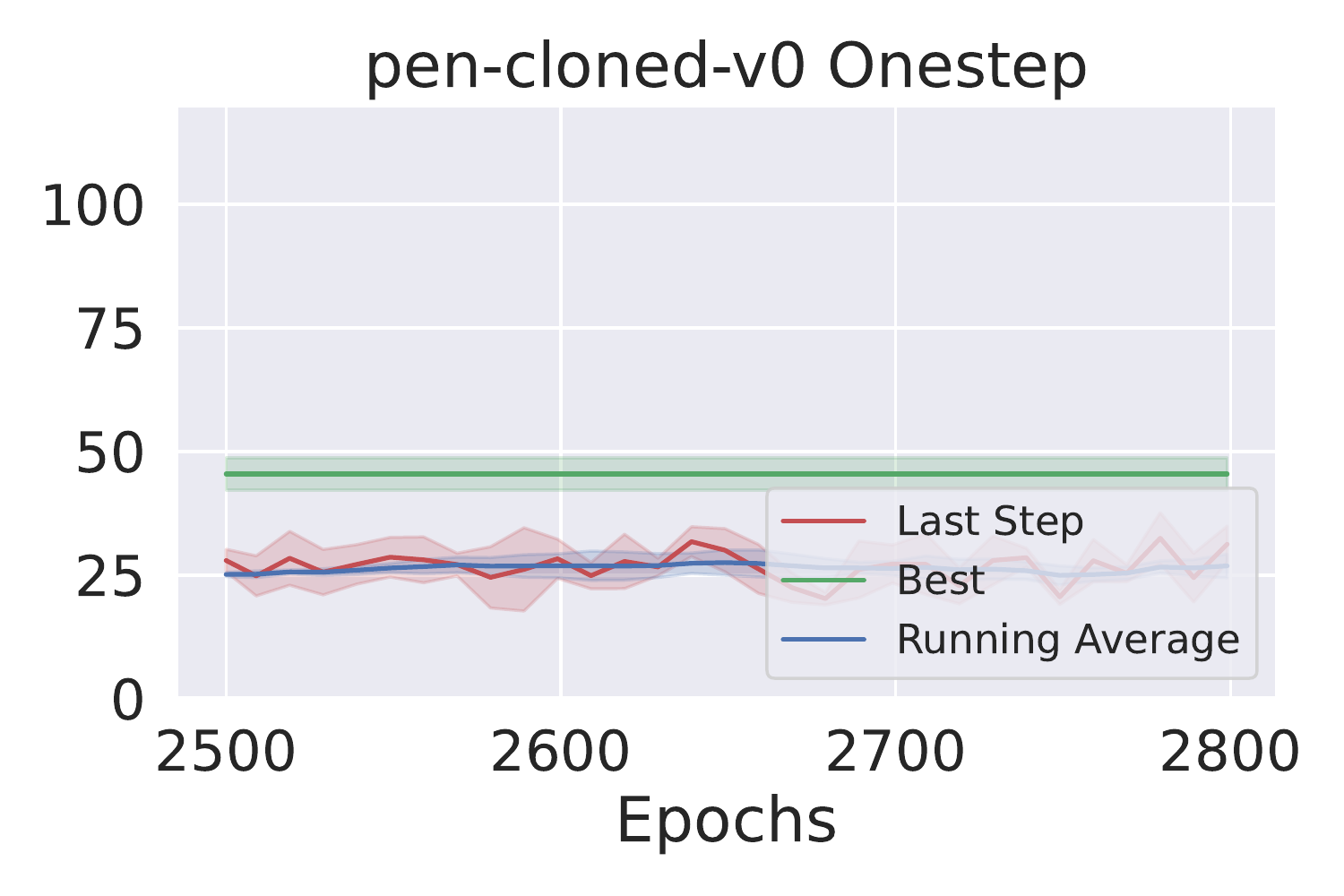}&\includegraphics[width=0.225\linewidth]{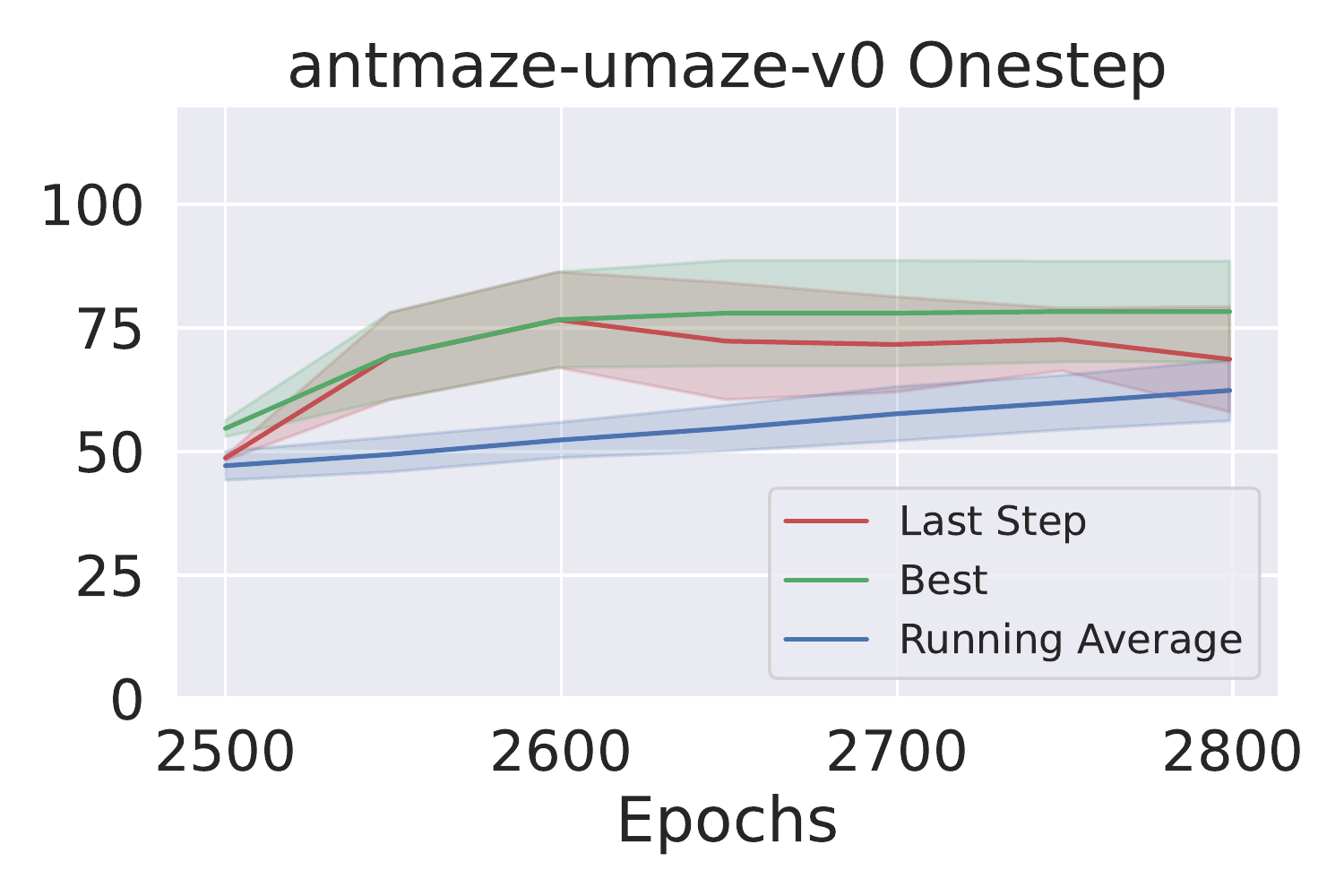}&\includegraphics[width=0.225\linewidth]{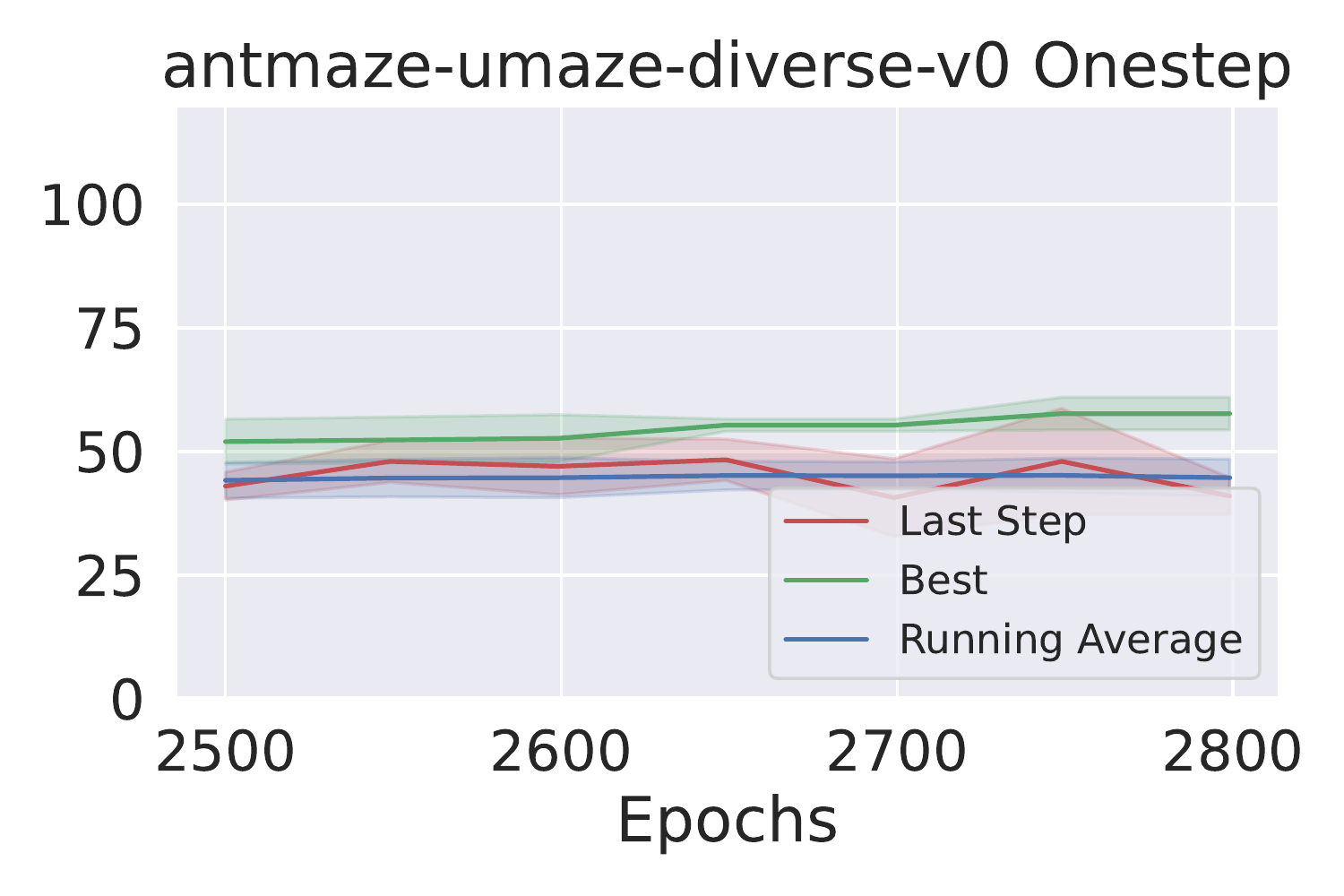}\\\includegraphics[width=0.225\linewidth]{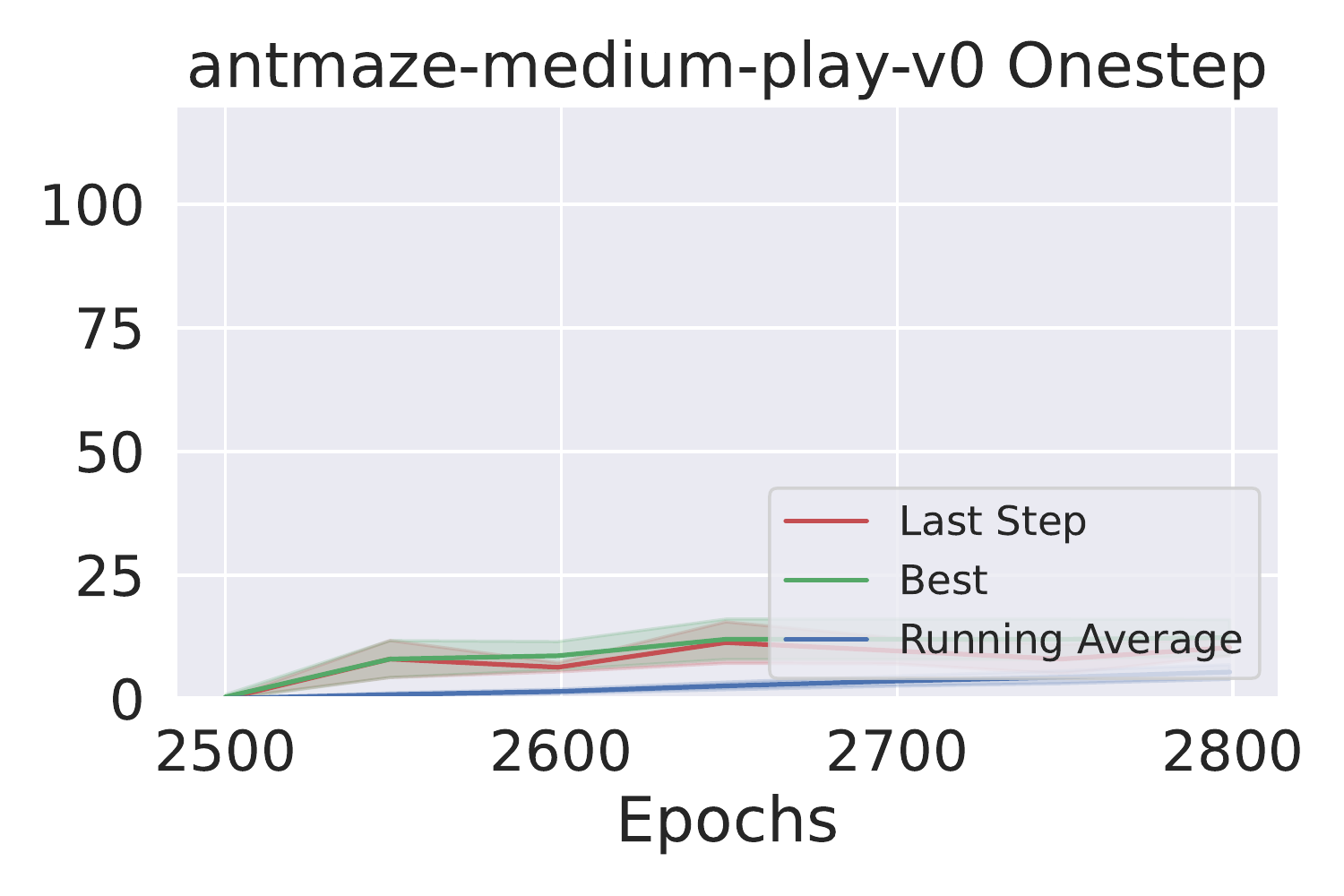}&\includegraphics[width=0.225\linewidth]{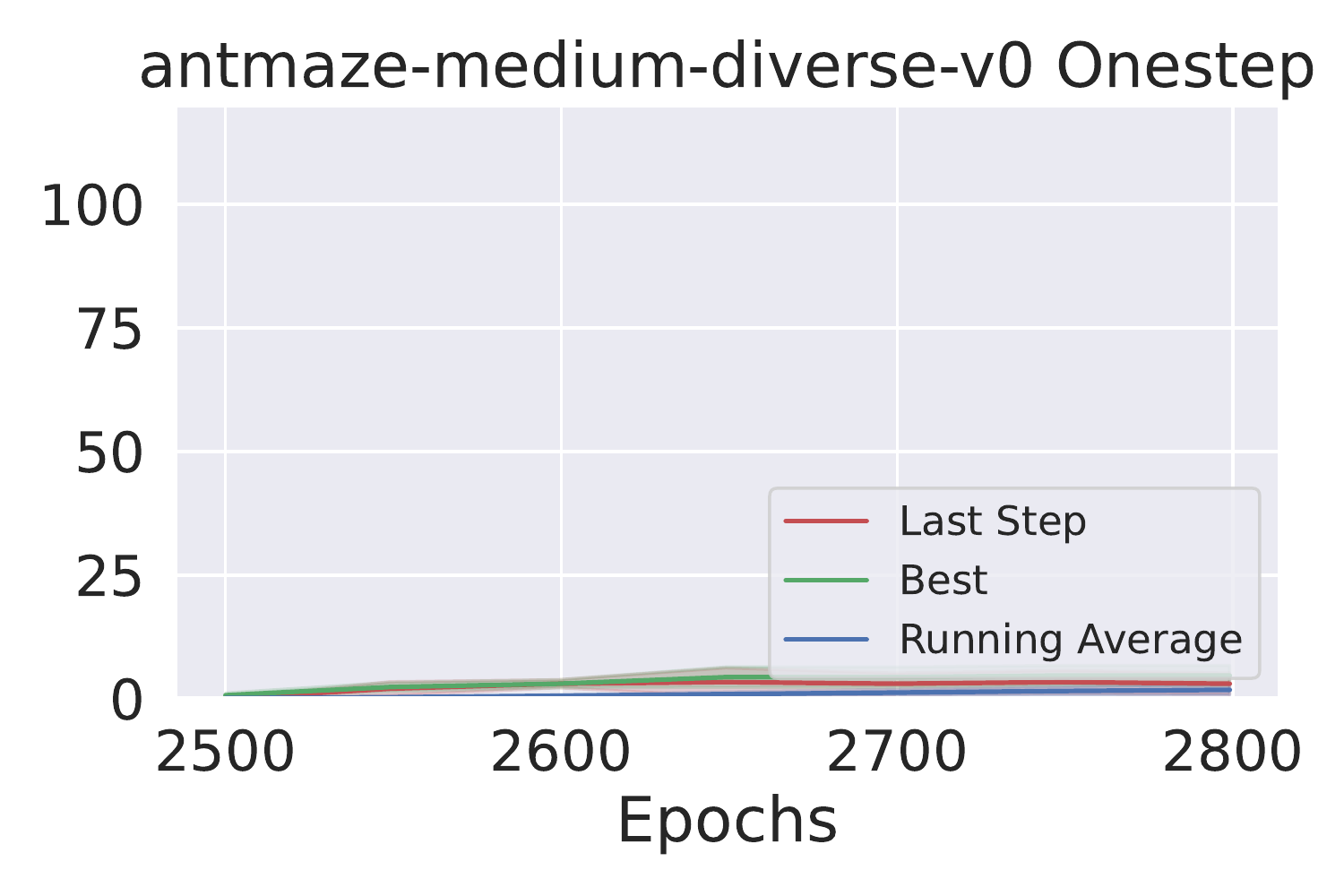}&\includegraphics[width=0.225\linewidth]{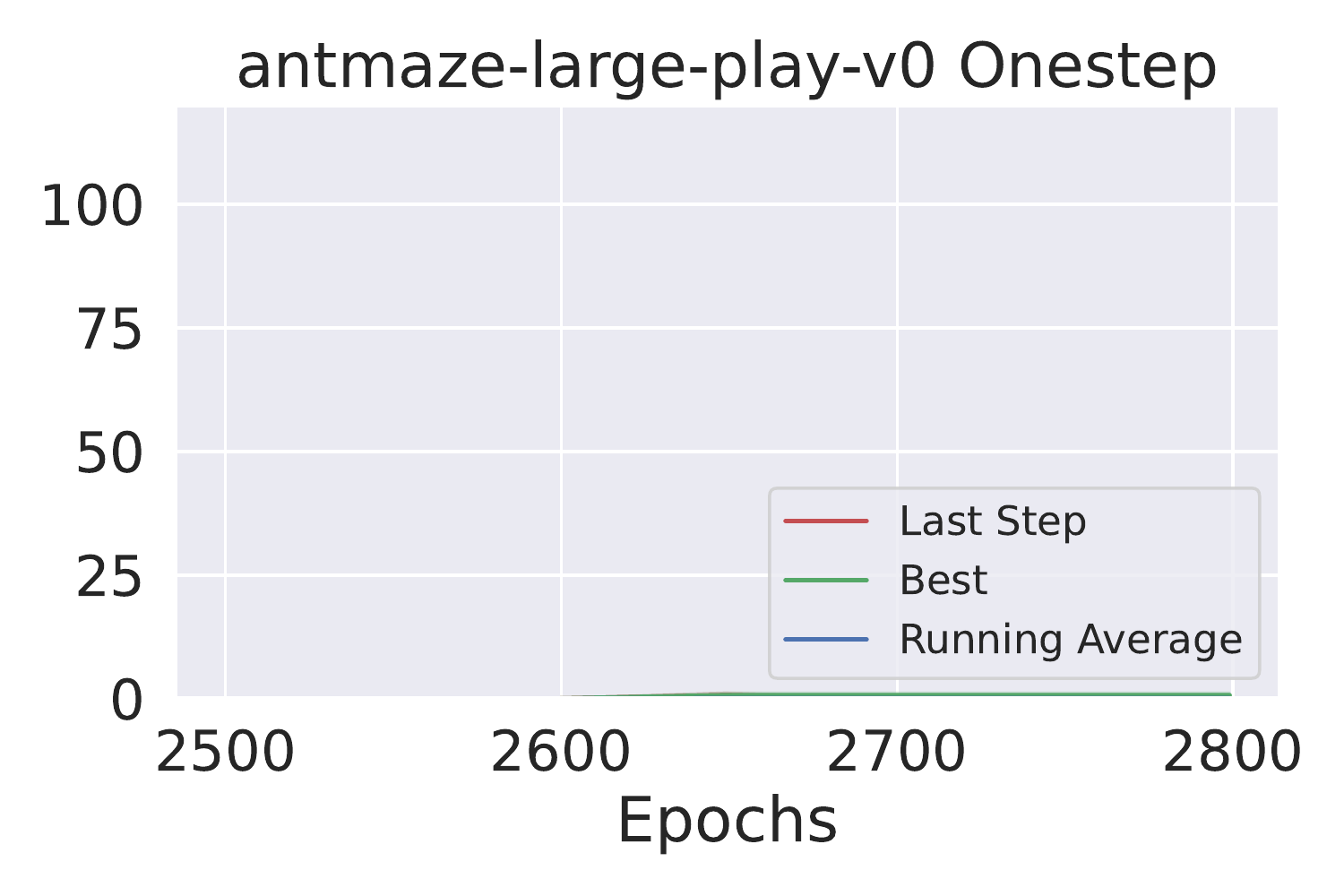}&\includegraphics[width=0.225\linewidth]{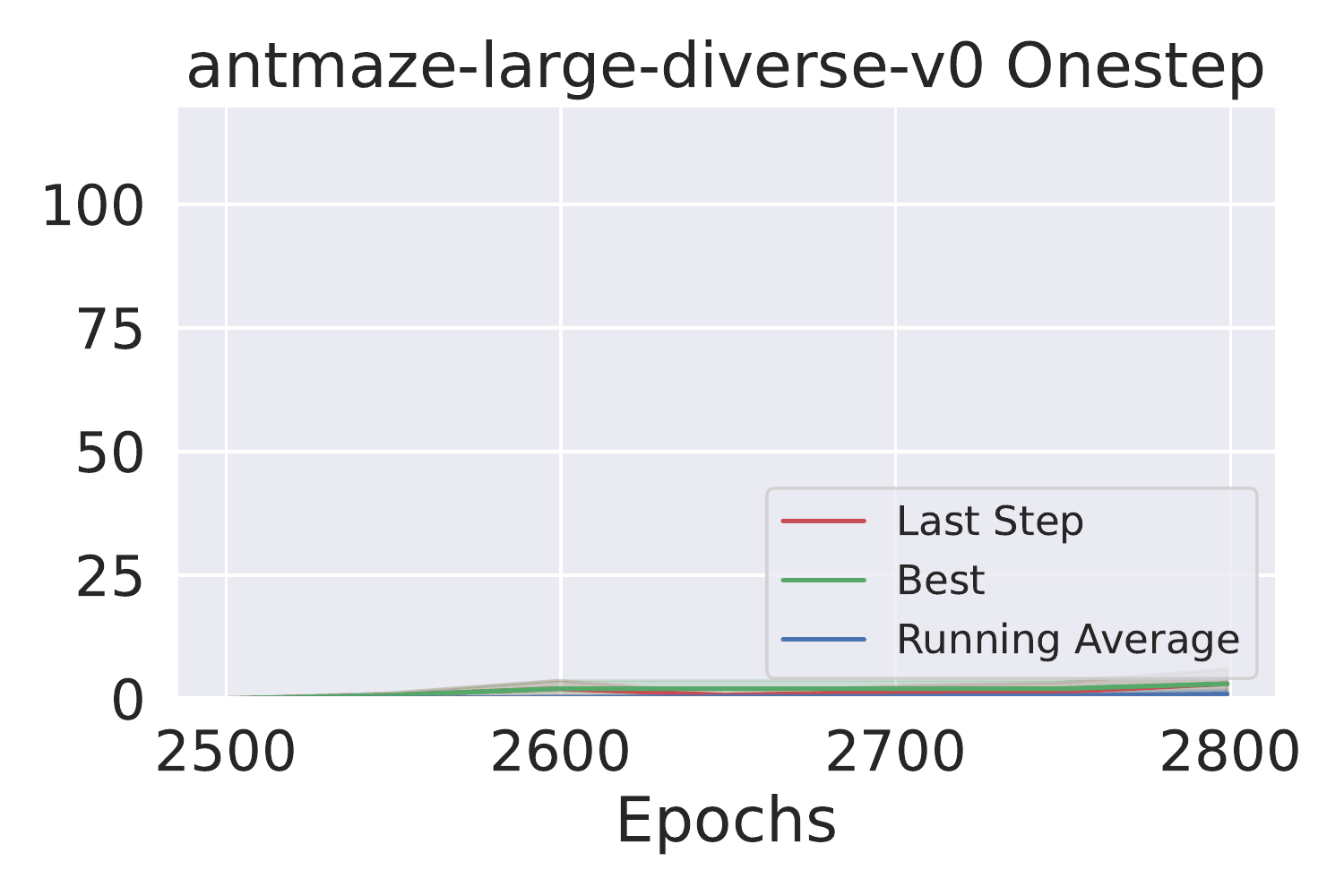}\\\end{tabular}
\centering
\caption{Training Curves of OnestepRL on D4RL}
\end{figure*}
\begin{figure*}[htb]
\centering
\begin{tabular}{cccccc}
\includegraphics[width=0.225\linewidth]{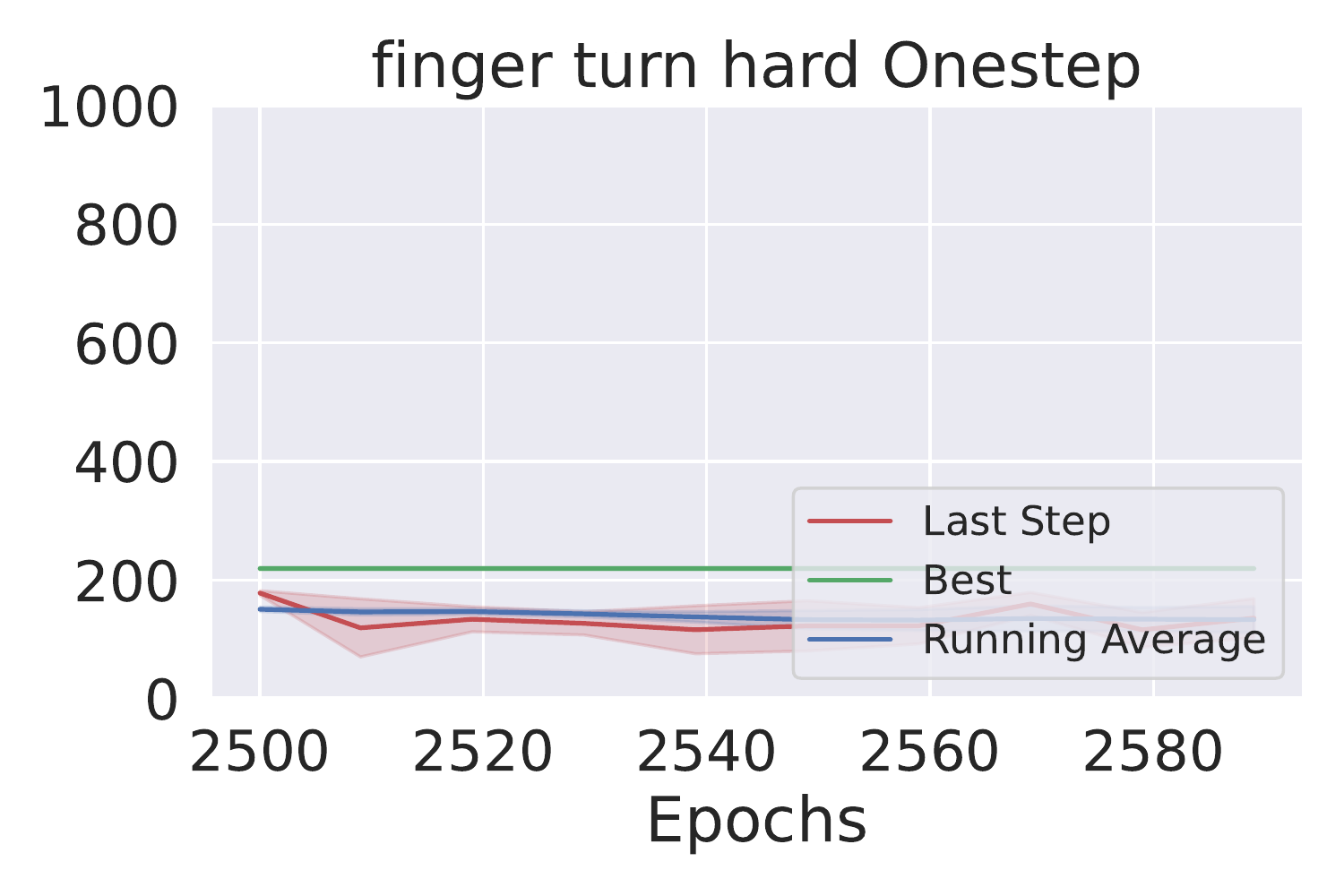}&\includegraphics[width=0.225\linewidth]{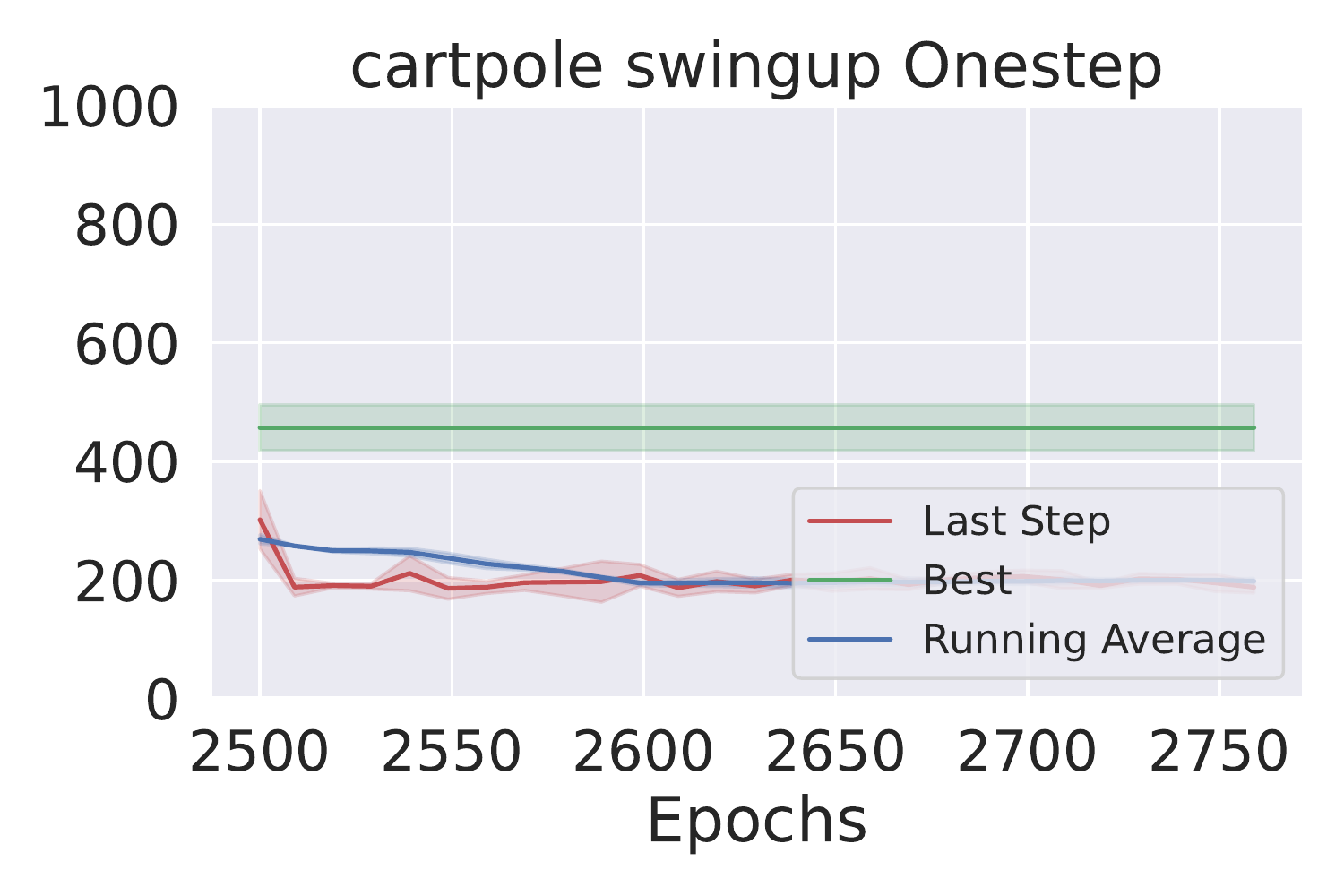}&\includegraphics[width=0.225\linewidth]{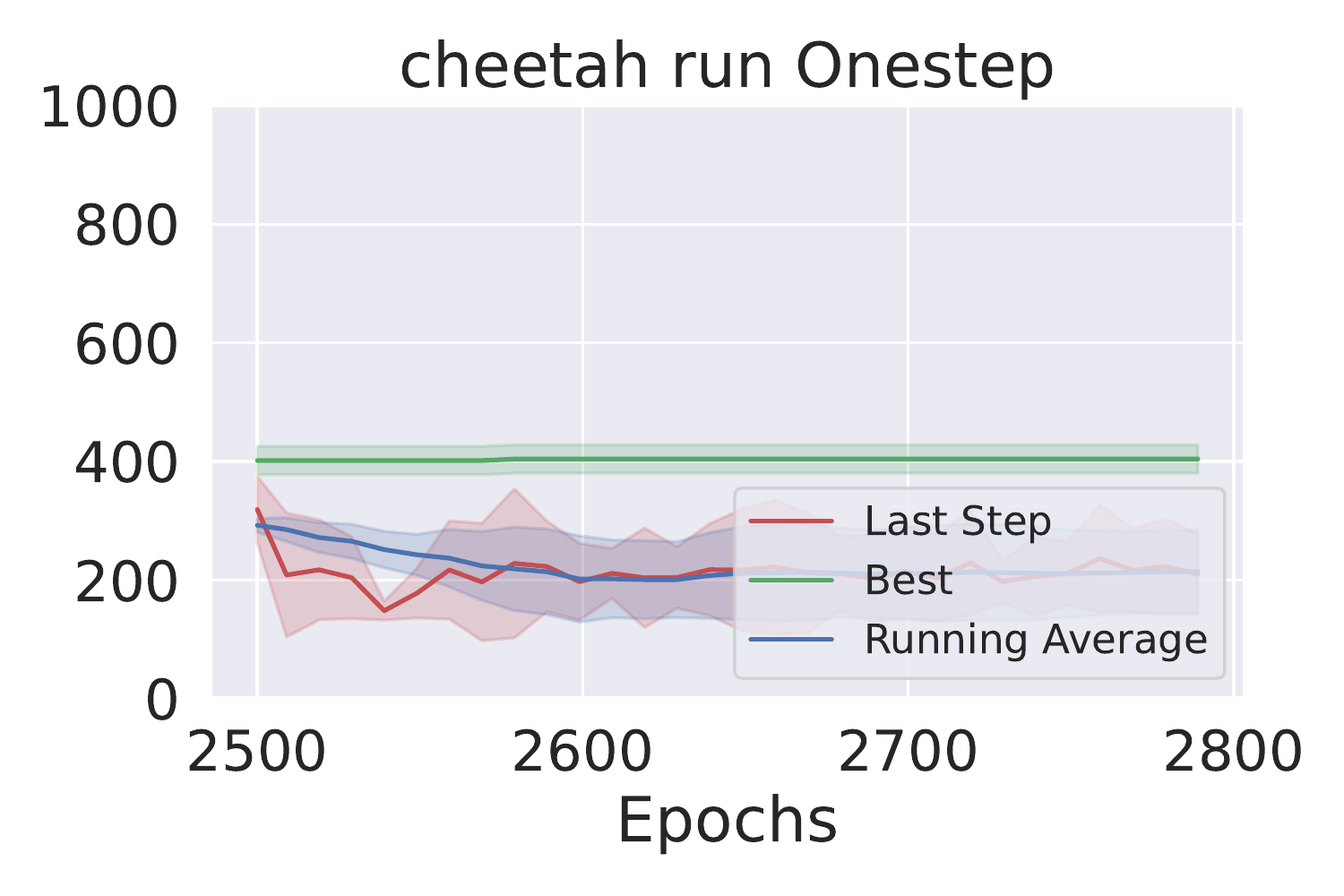}&\includegraphics[width=0.225\linewidth]{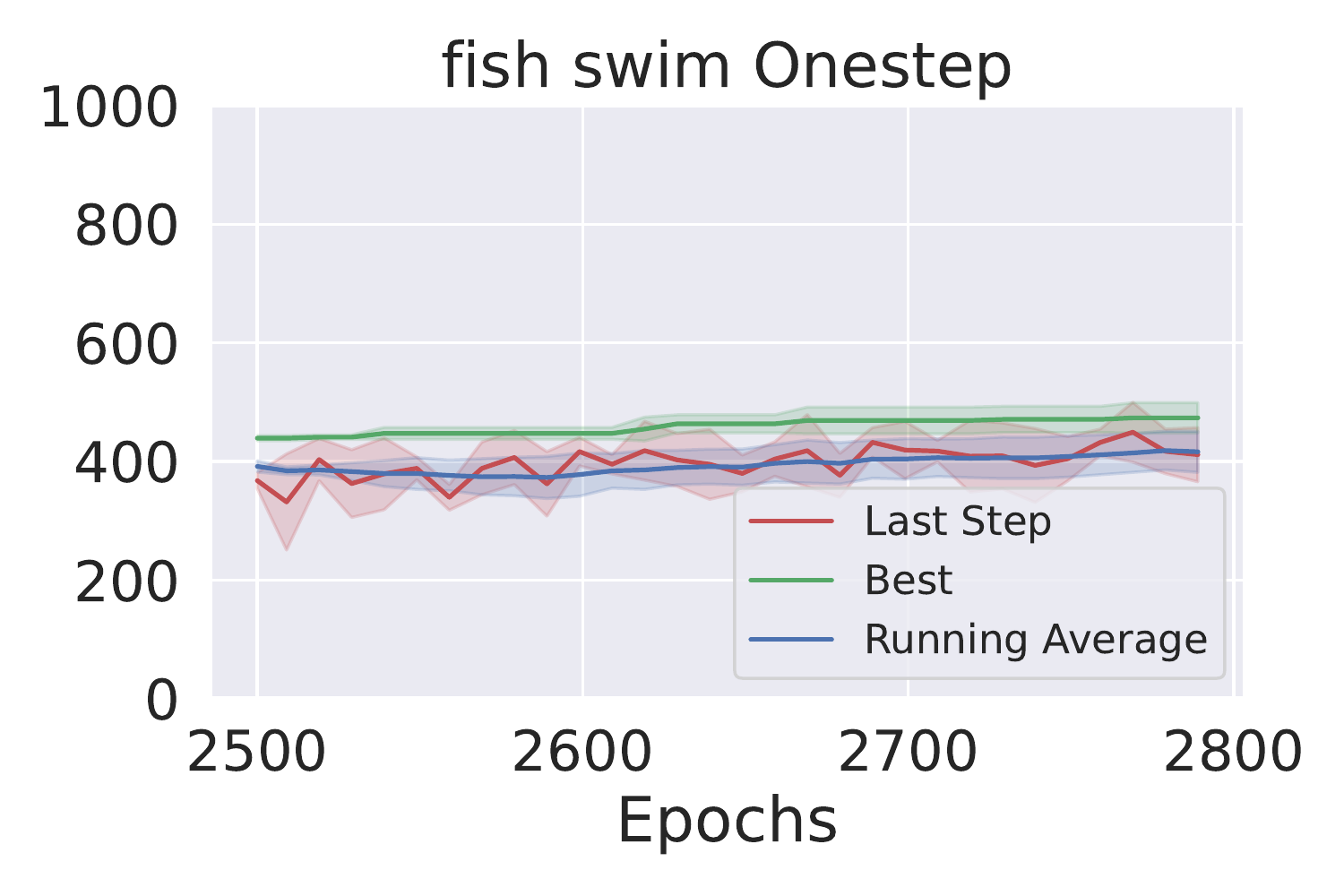}\\\includegraphics[width=0.225\linewidth]{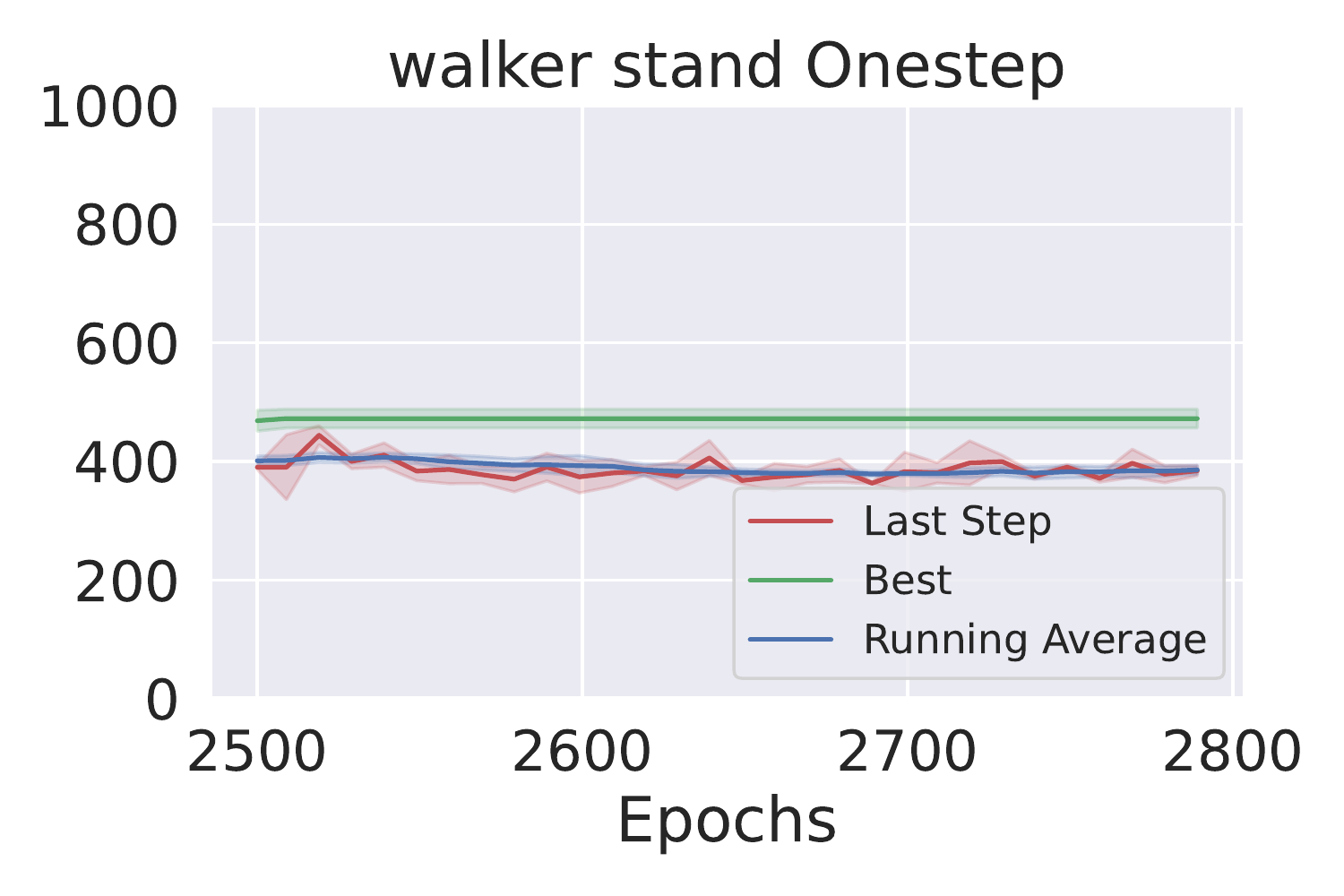}&\includegraphics[width=0.225\linewidth]{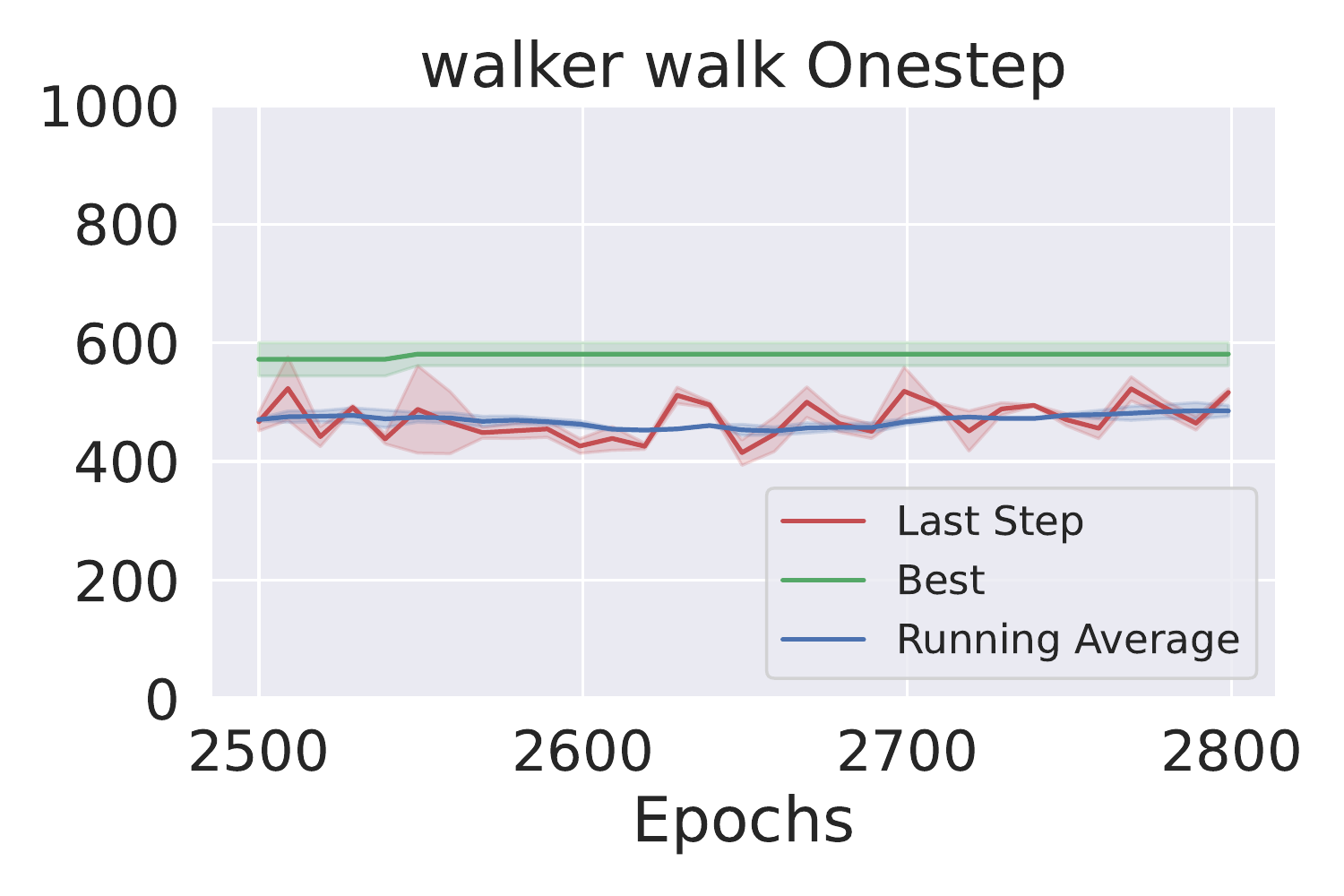}&&\\\end{tabular}
\centering
\caption{Training Curves of OnestepRL on RLUP}
\end{figure*}

\begin{figure*}[htb]
\centering
\begin{tabular}{cccccc}
\includegraphics[width=0.225\linewidth]{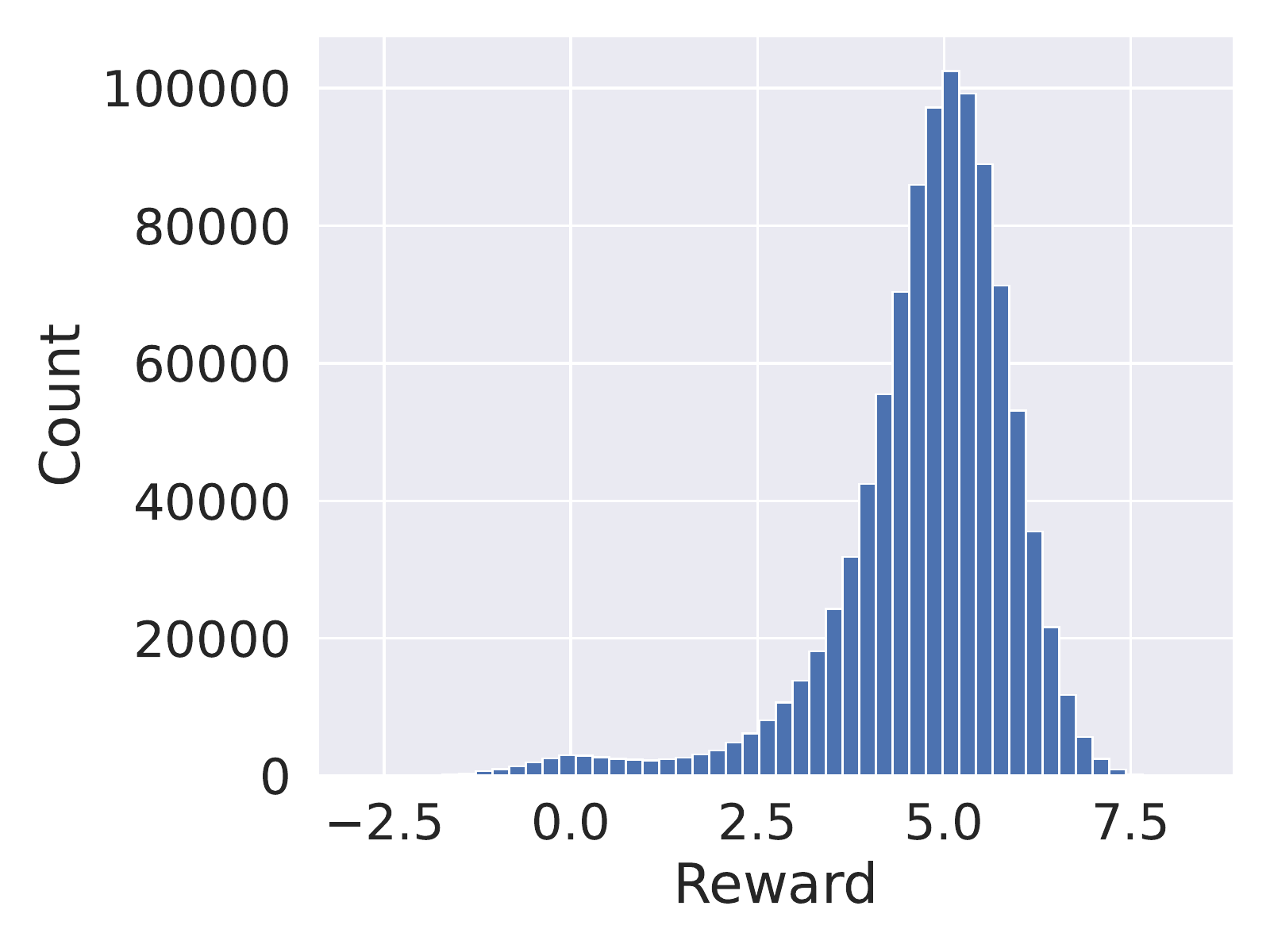}&\includegraphics[width=0.225\linewidth]{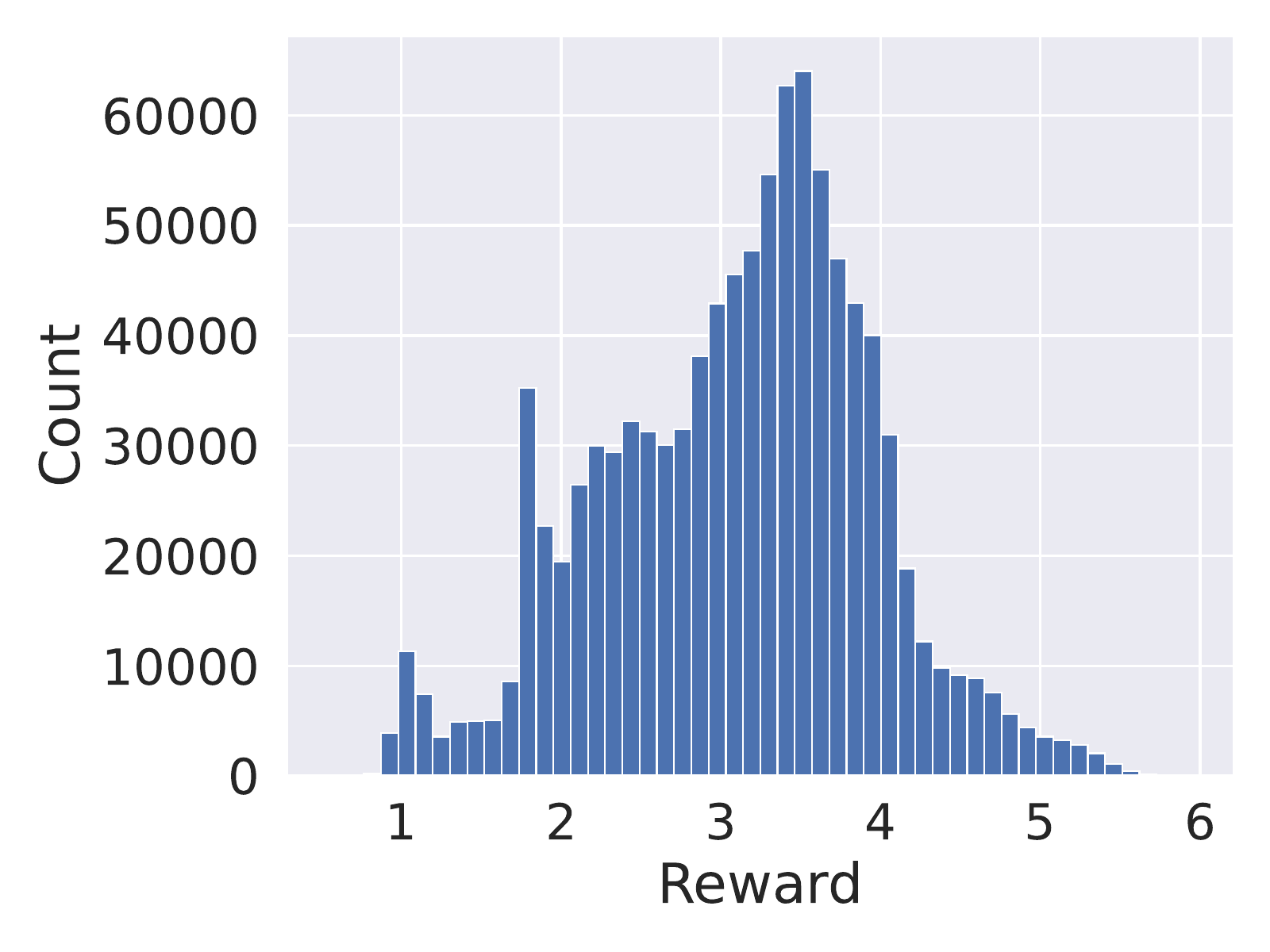}&\includegraphics[width=0.225\linewidth]{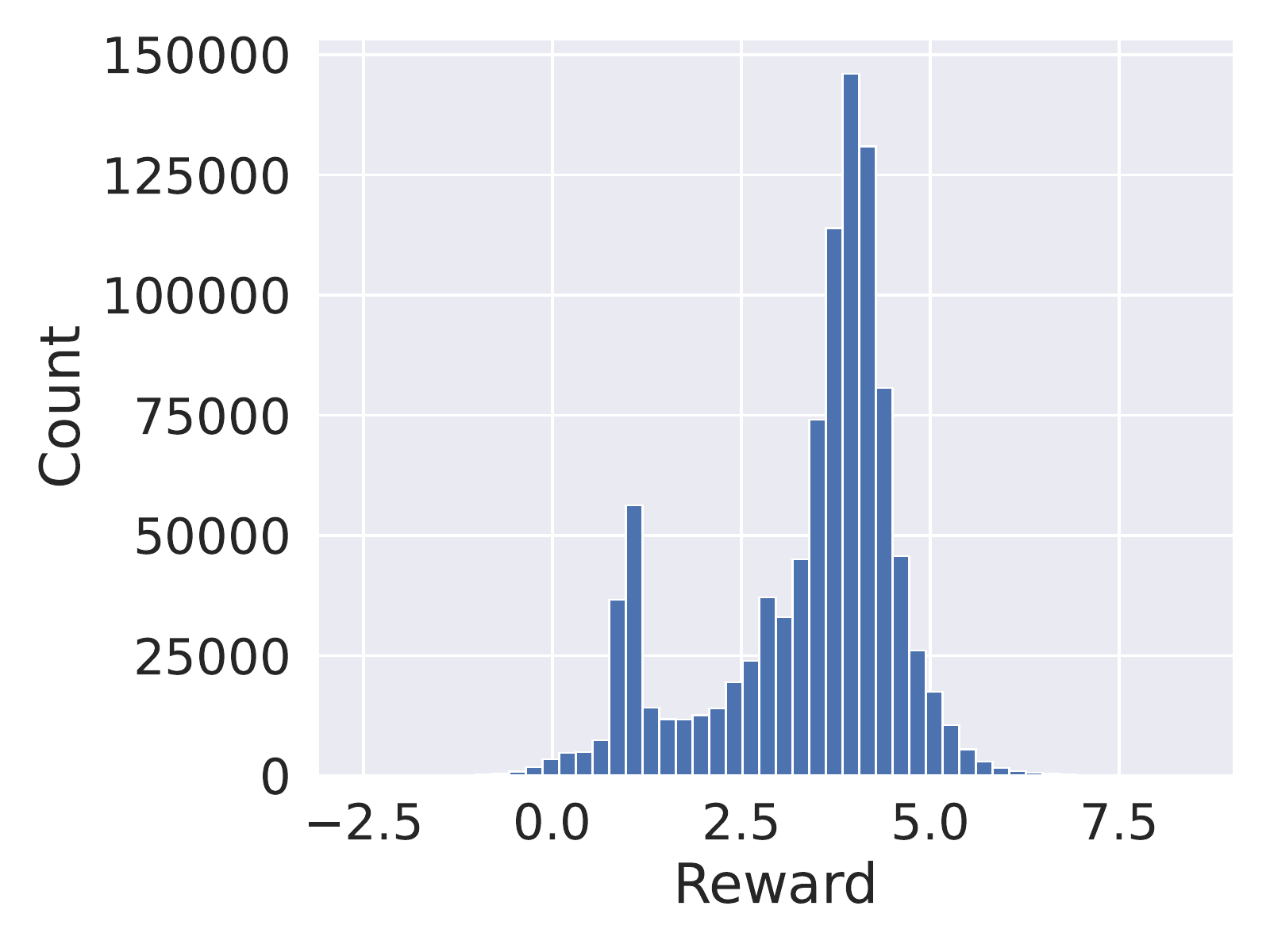}&\includegraphics[width=0.225\linewidth]{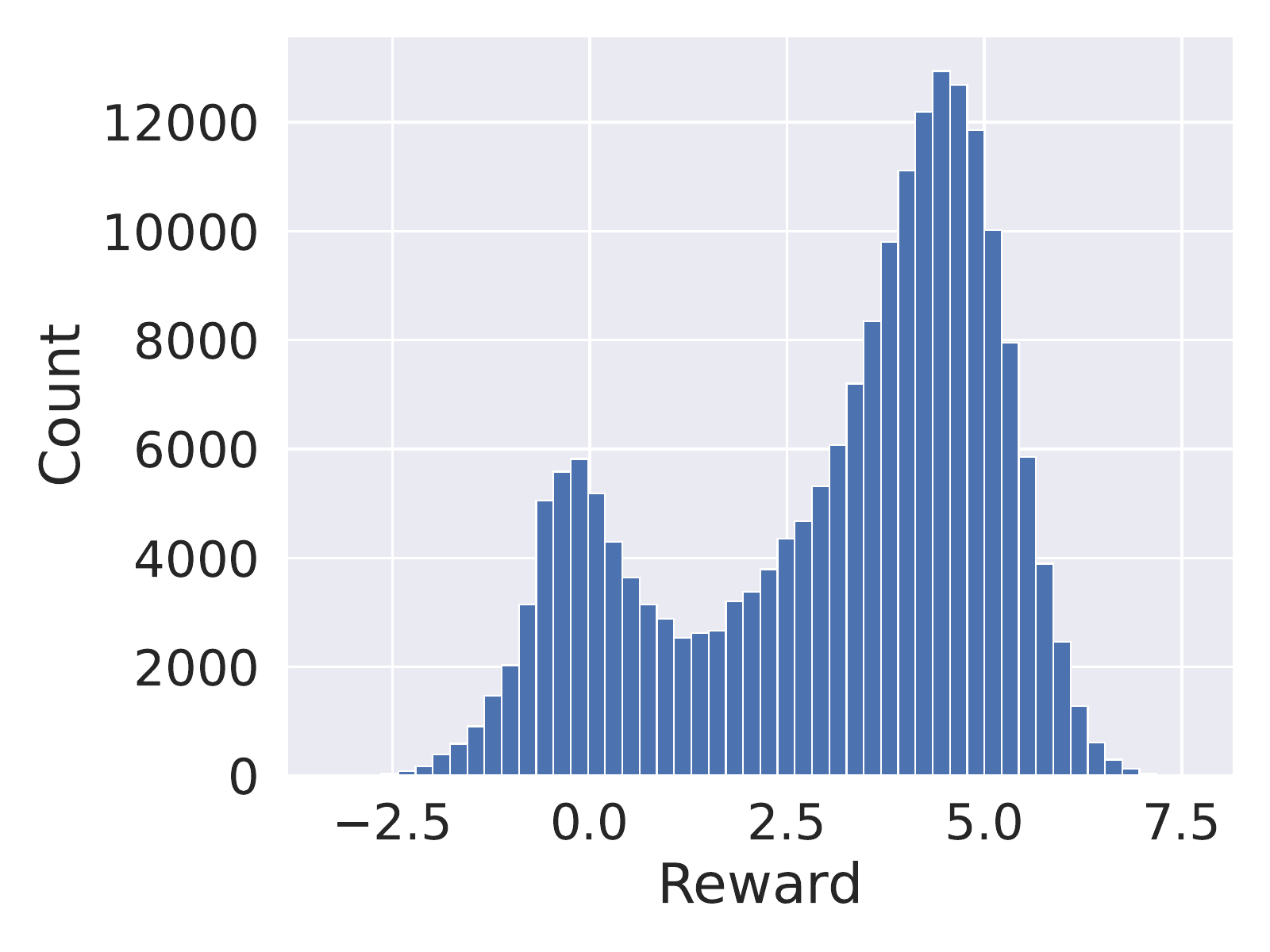}\\\includegraphics[width=0.225\linewidth]{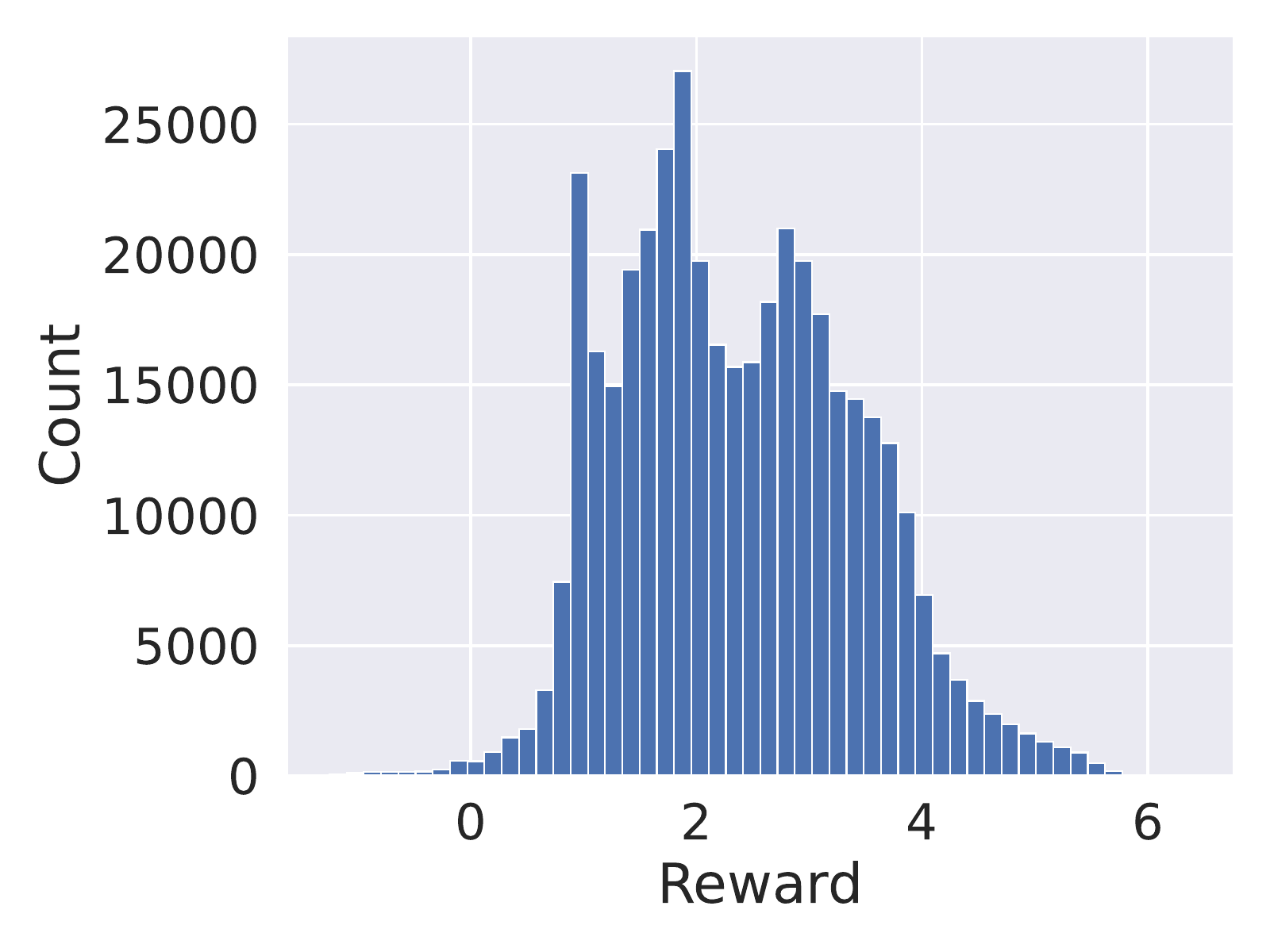}&\includegraphics[width=0.225\linewidth]{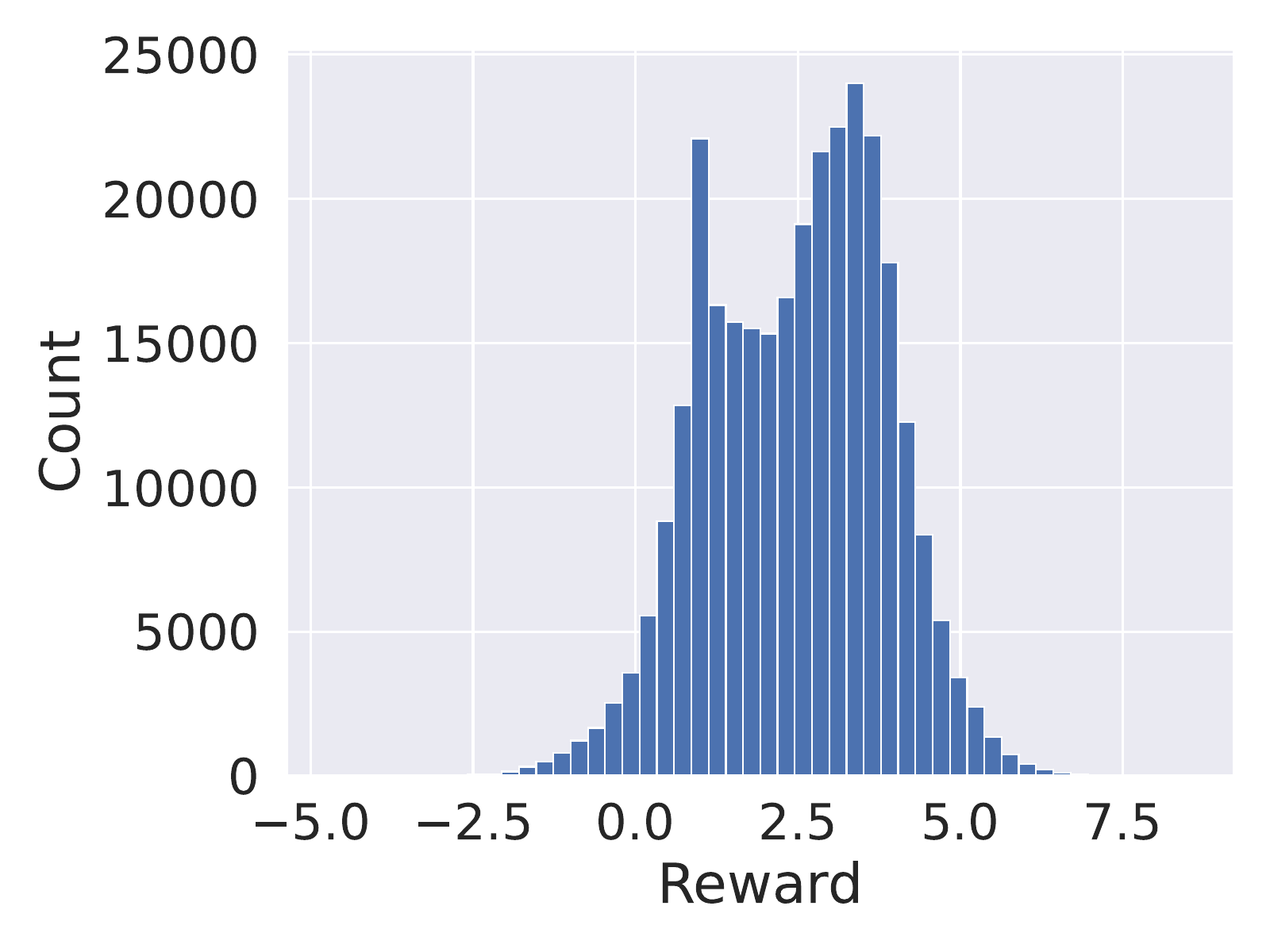}&\includegraphics[width=0.225\linewidth]{dist_plots/halfcheetah-medium-expert-v2}&\includegraphics[width=0.225\linewidth]{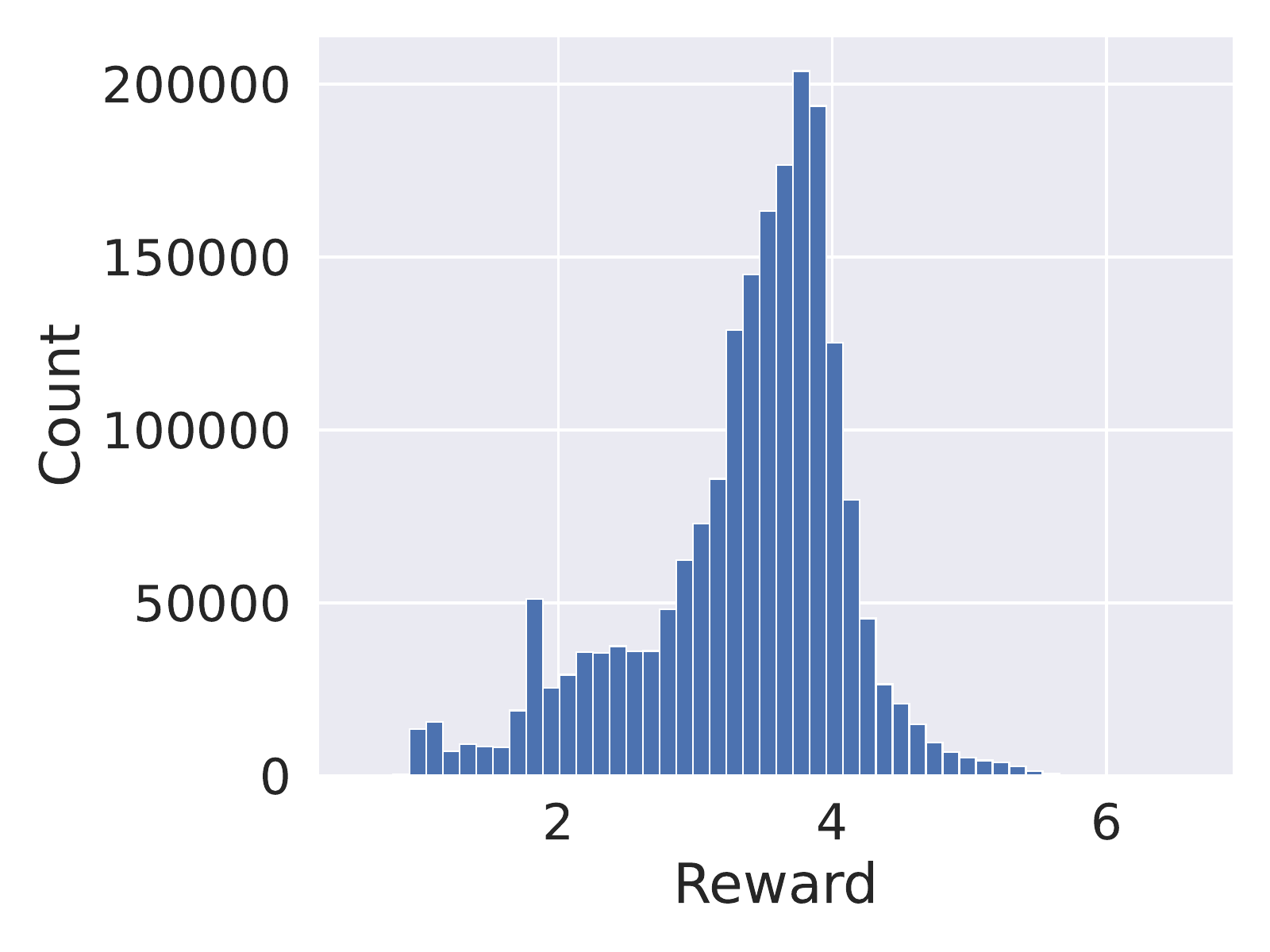}\\\includegraphics[width=0.225\linewidth]{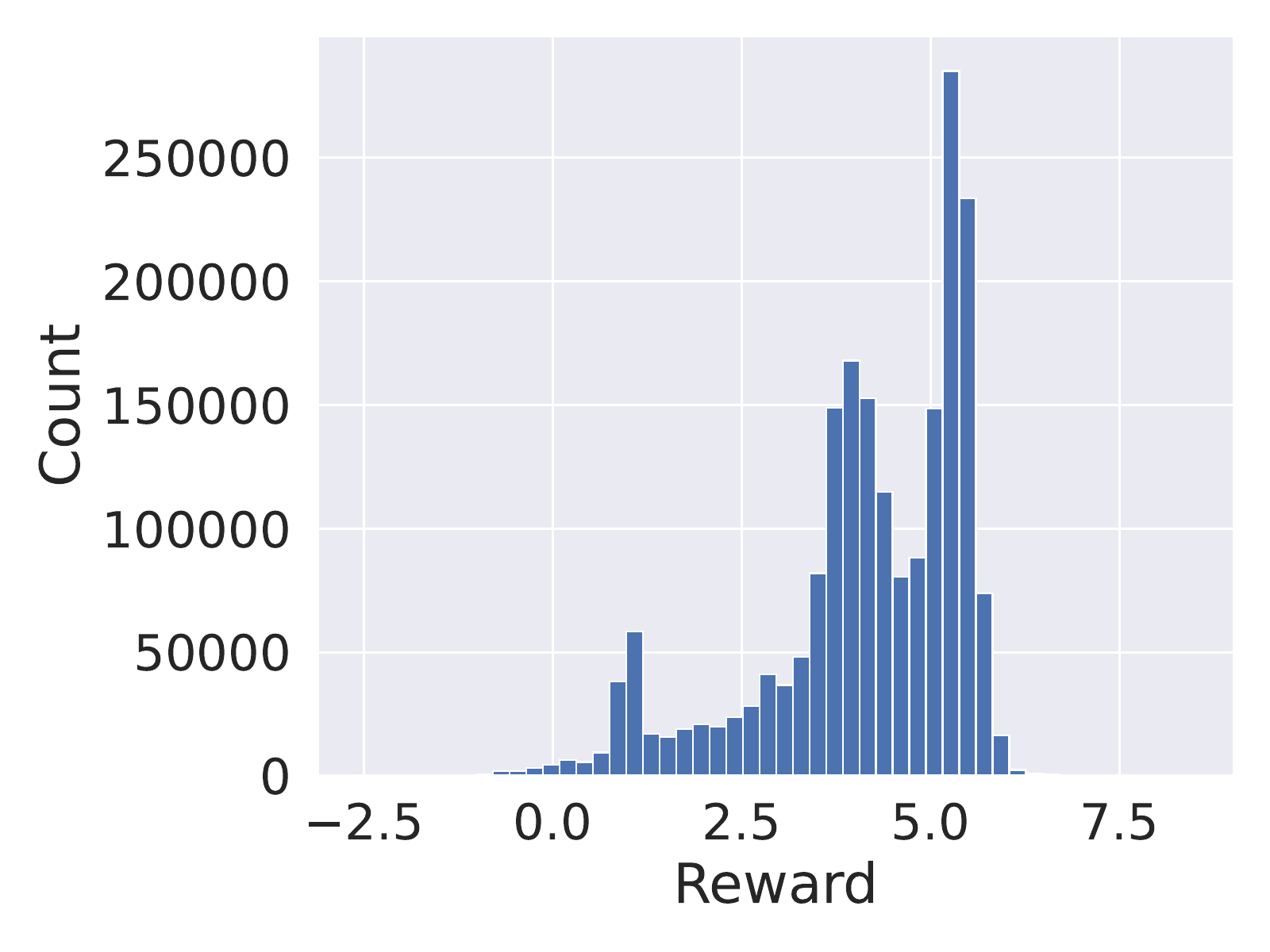}&\includegraphics[width=0.225\linewidth]{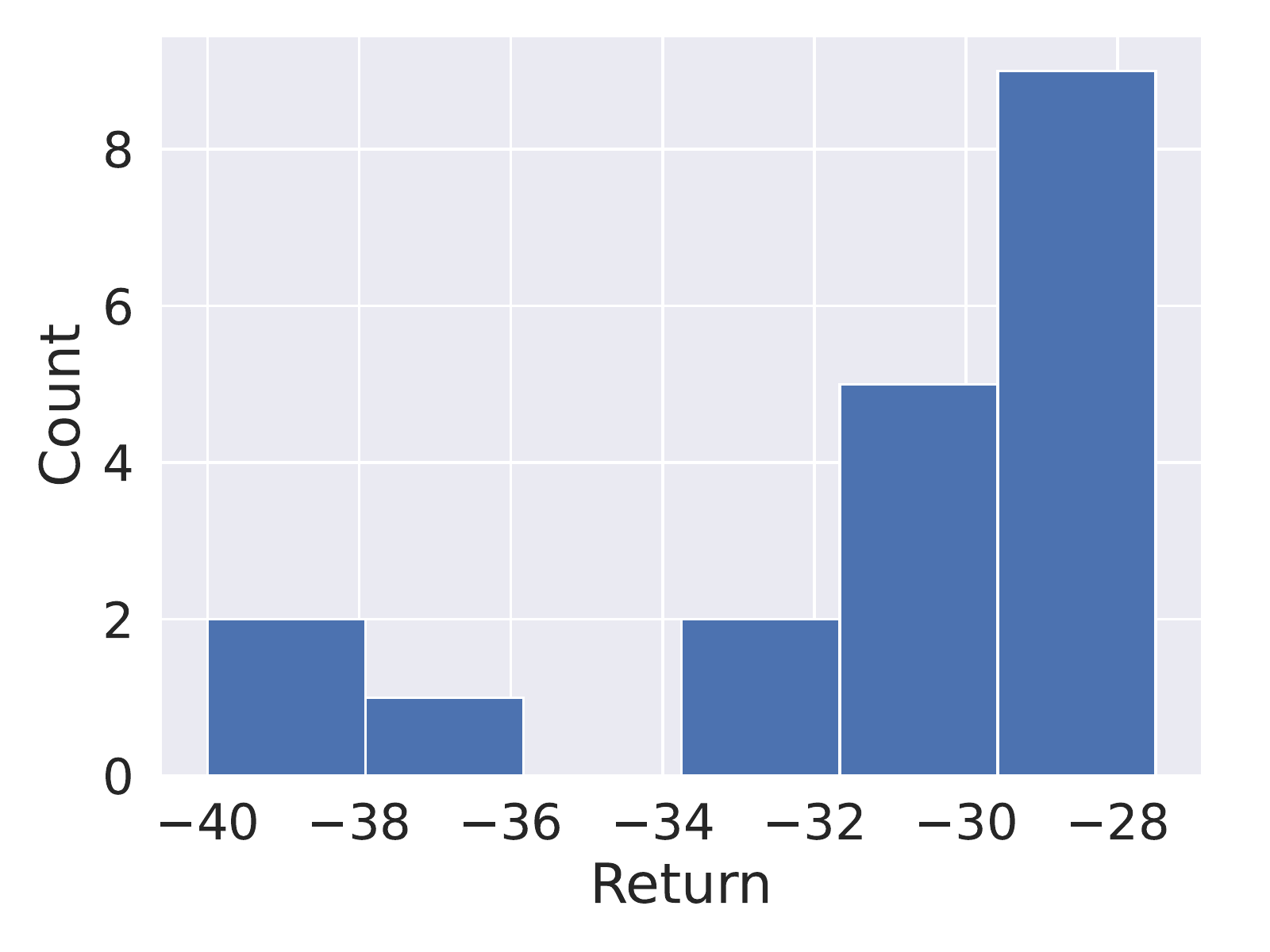}&\includegraphics[width=0.225\linewidth]{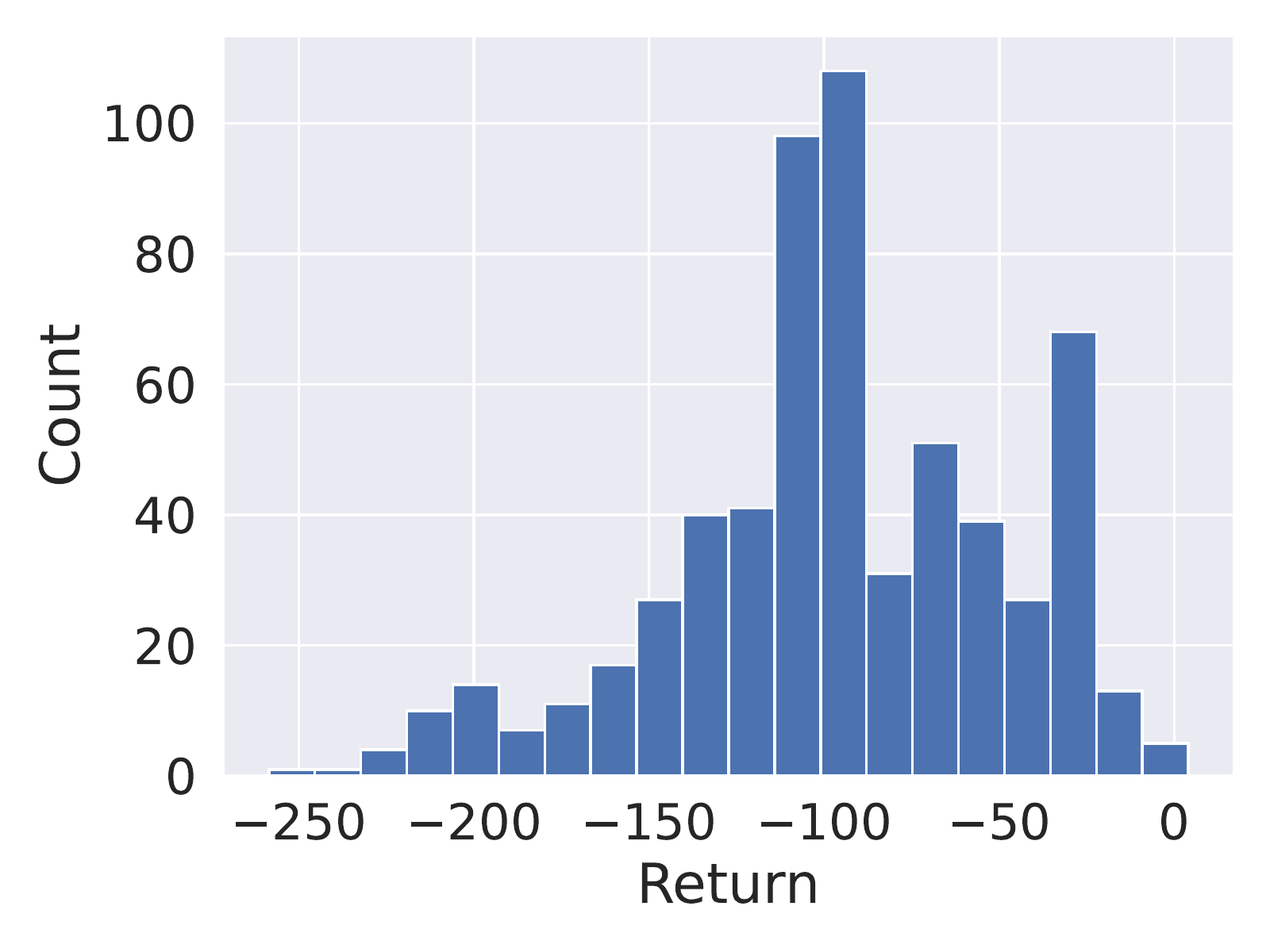}&\includegraphics[width=0.225\linewidth]{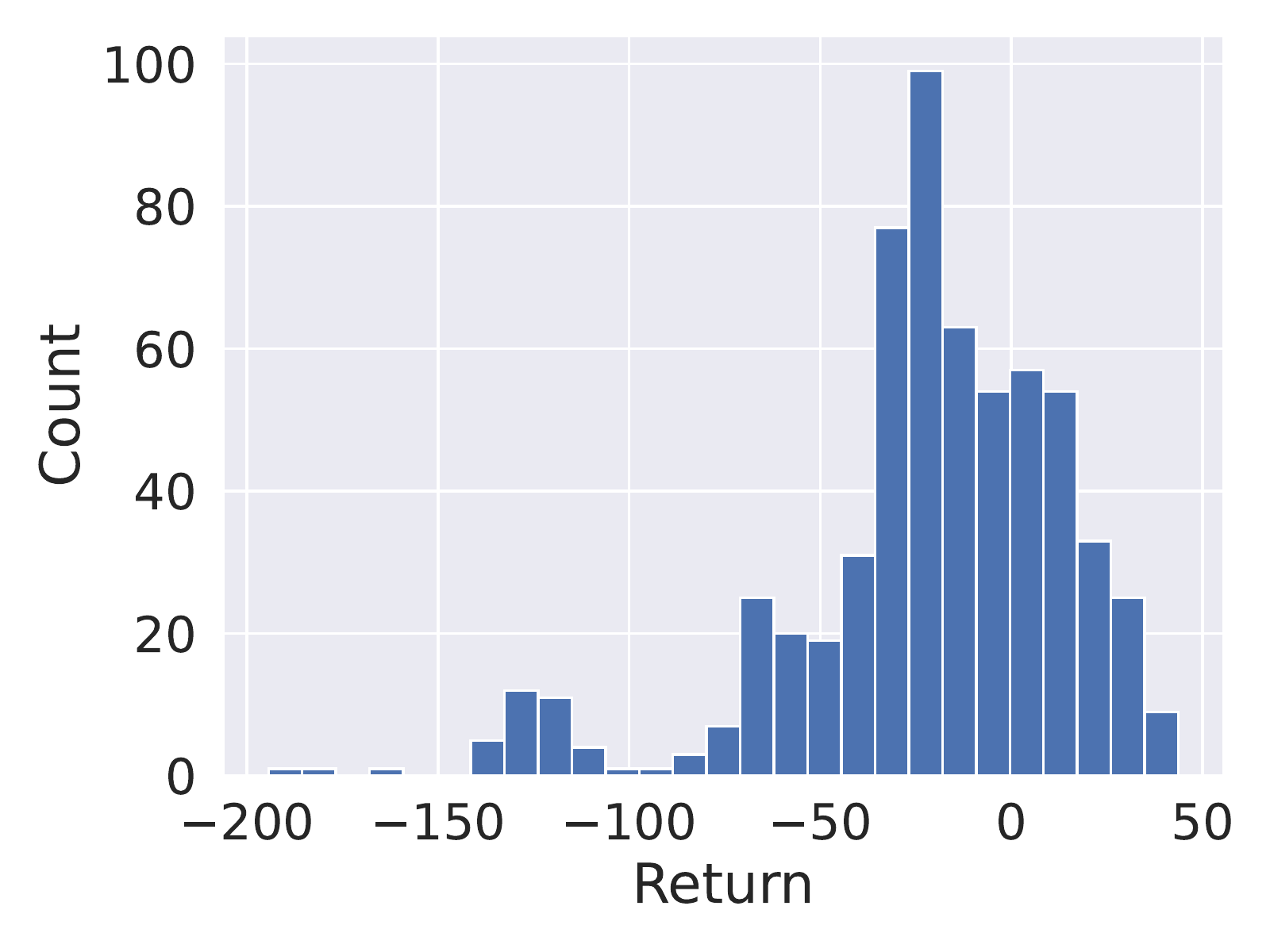}\\\includegraphics[width=0.225\linewidth]{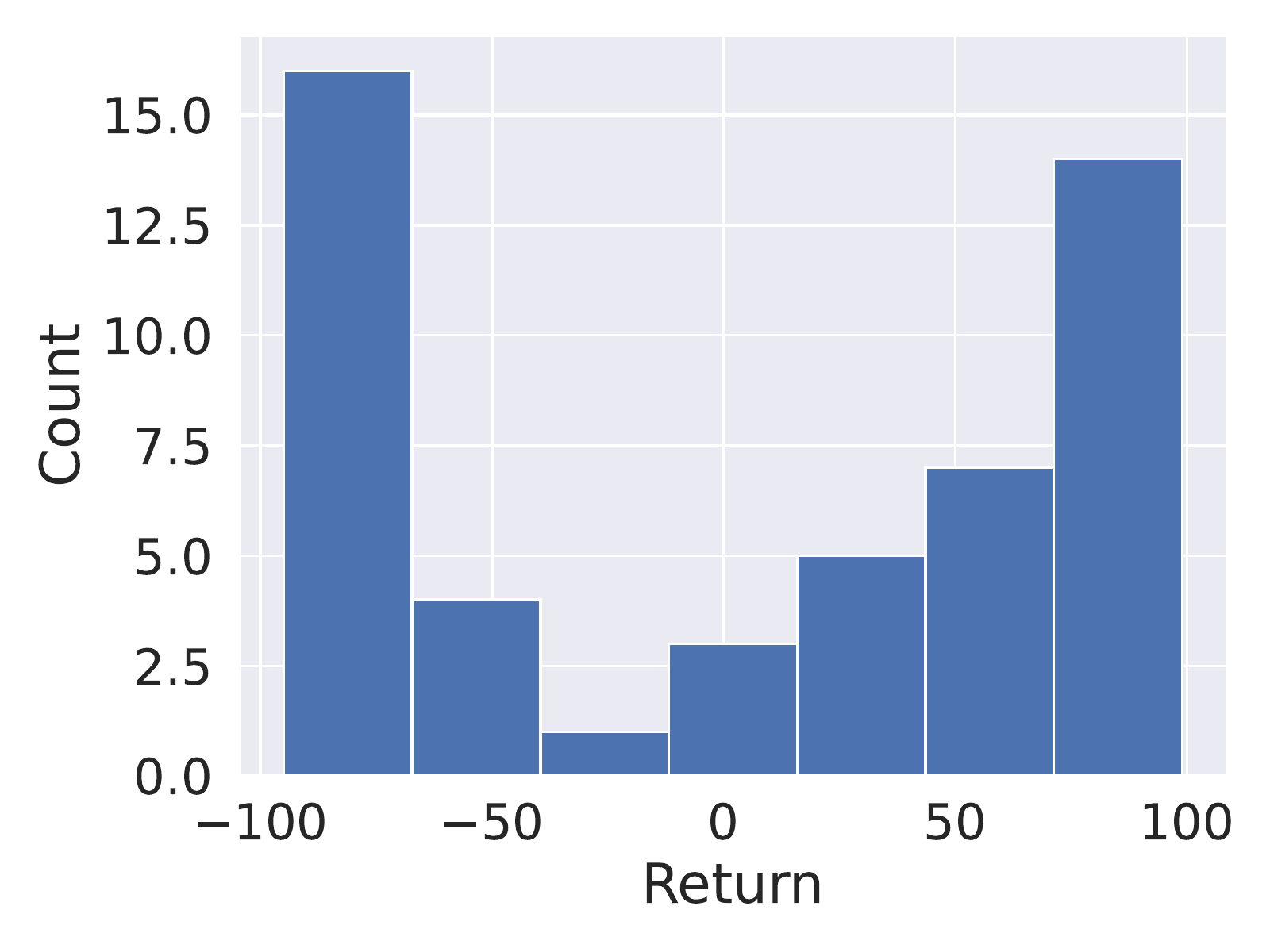}&\includegraphics[width=0.225\linewidth]{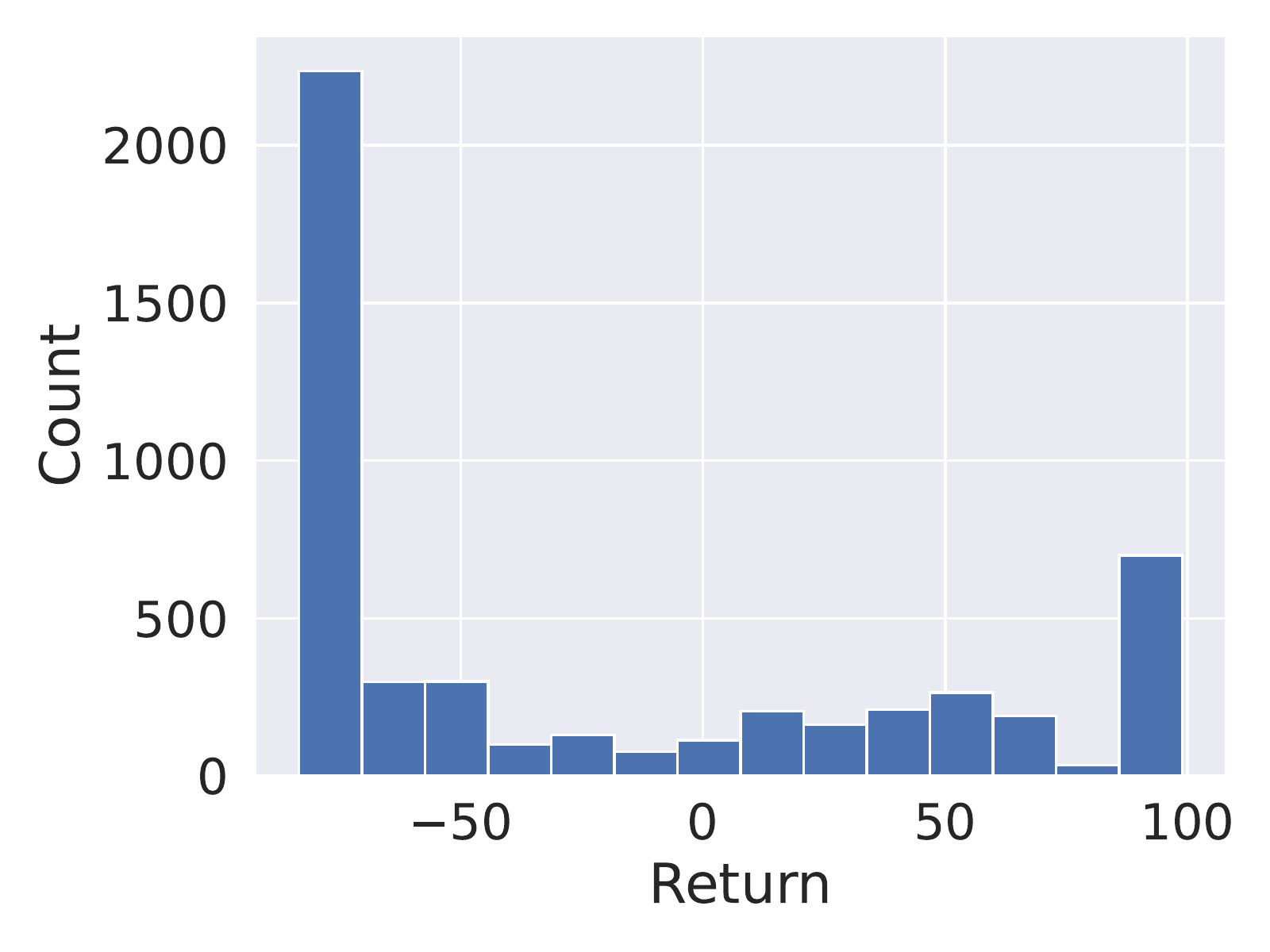}&\includegraphics[width=0.225\linewidth]{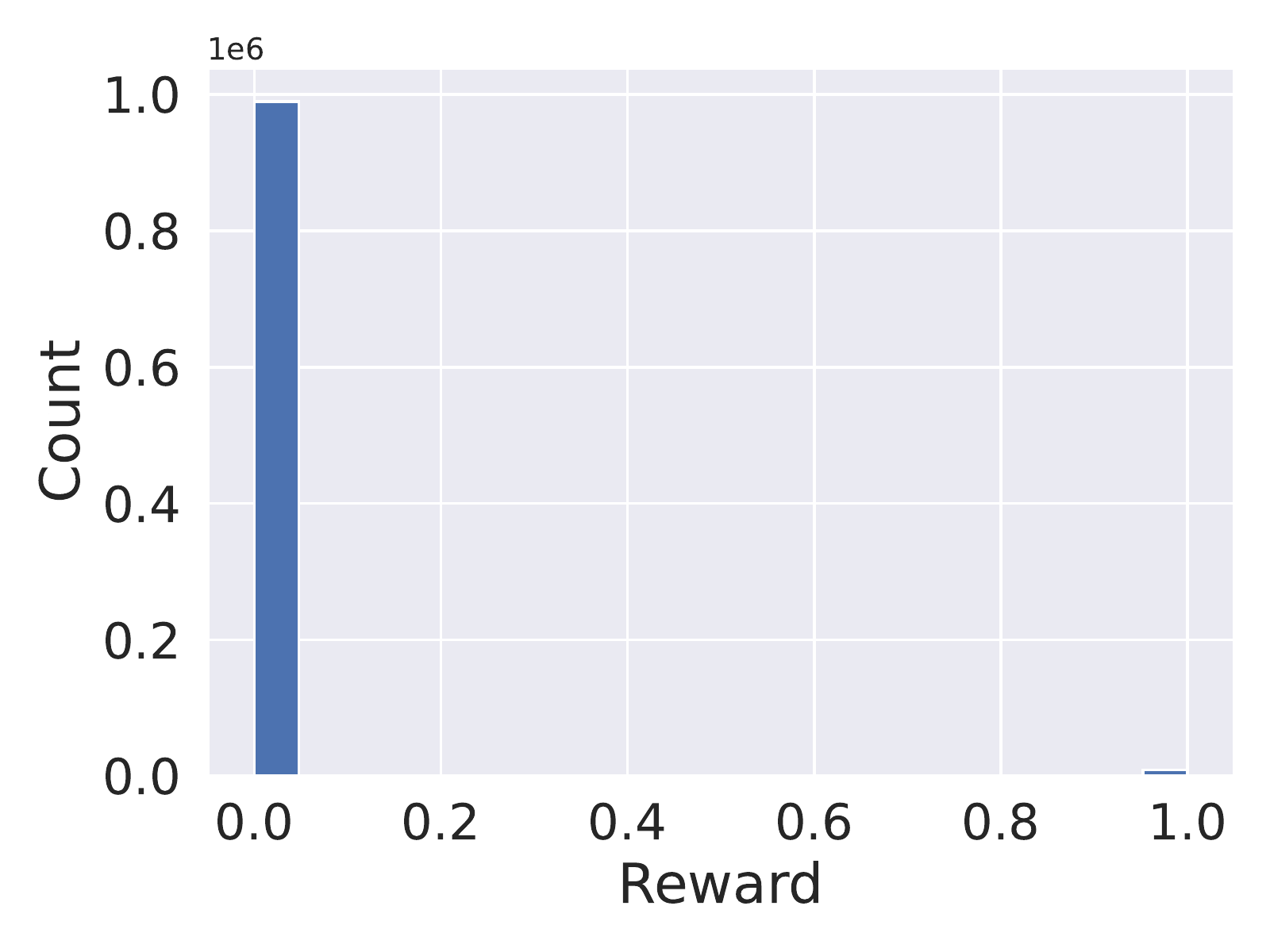}&\includegraphics[width=0.225\linewidth]{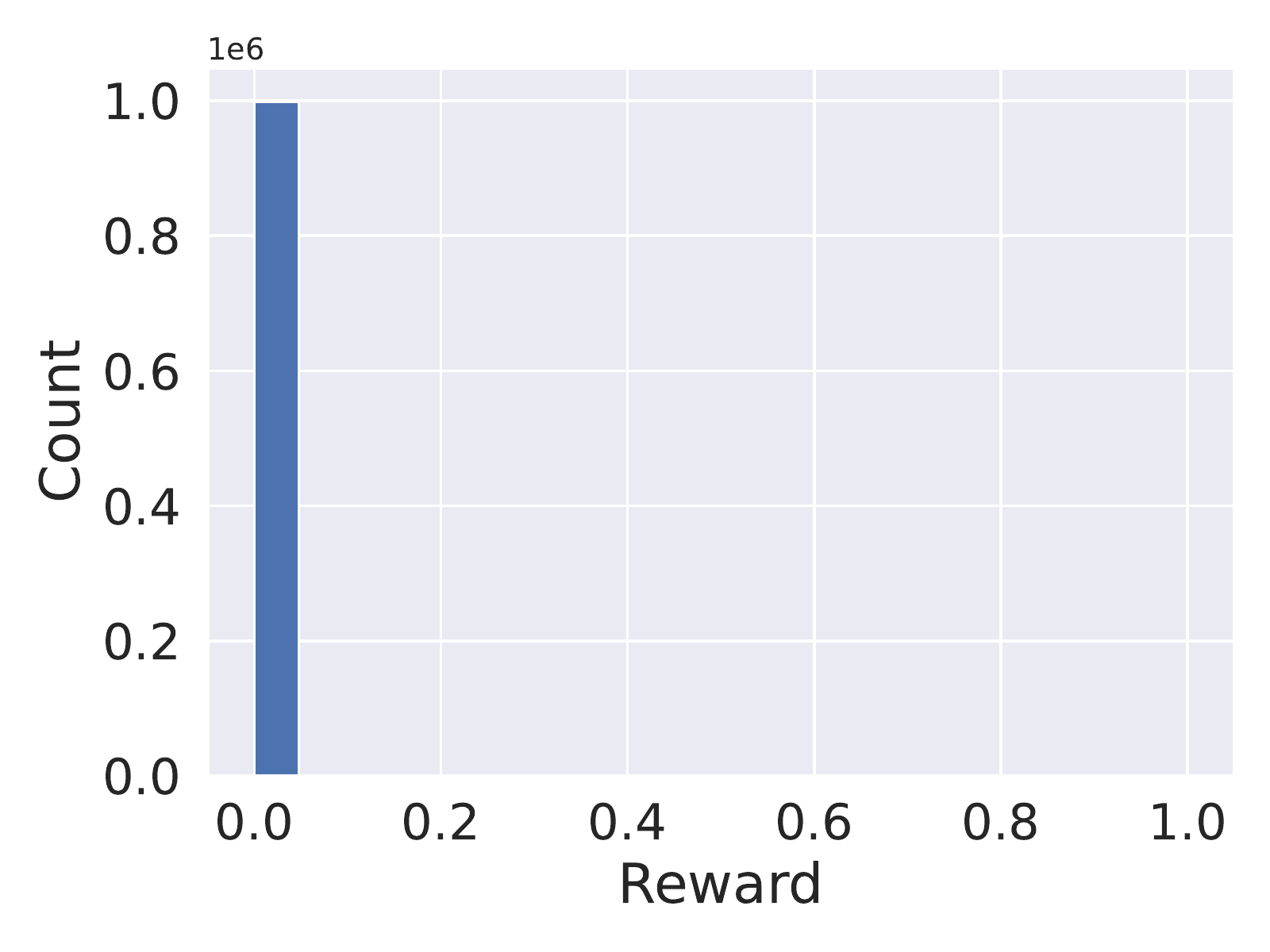}\\\includegraphics[width=0.225\linewidth]{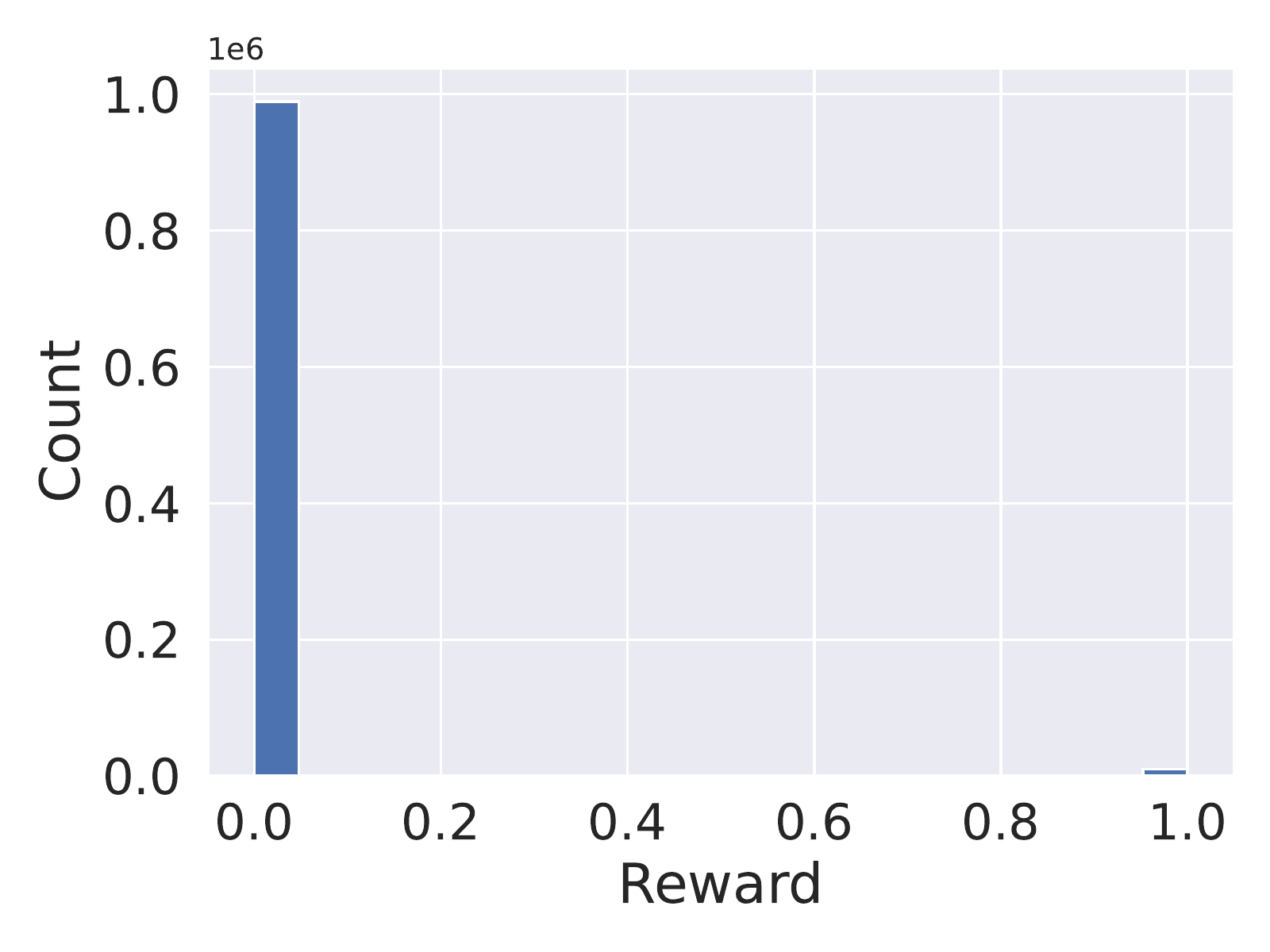}&\includegraphics[width=0.225\linewidth]{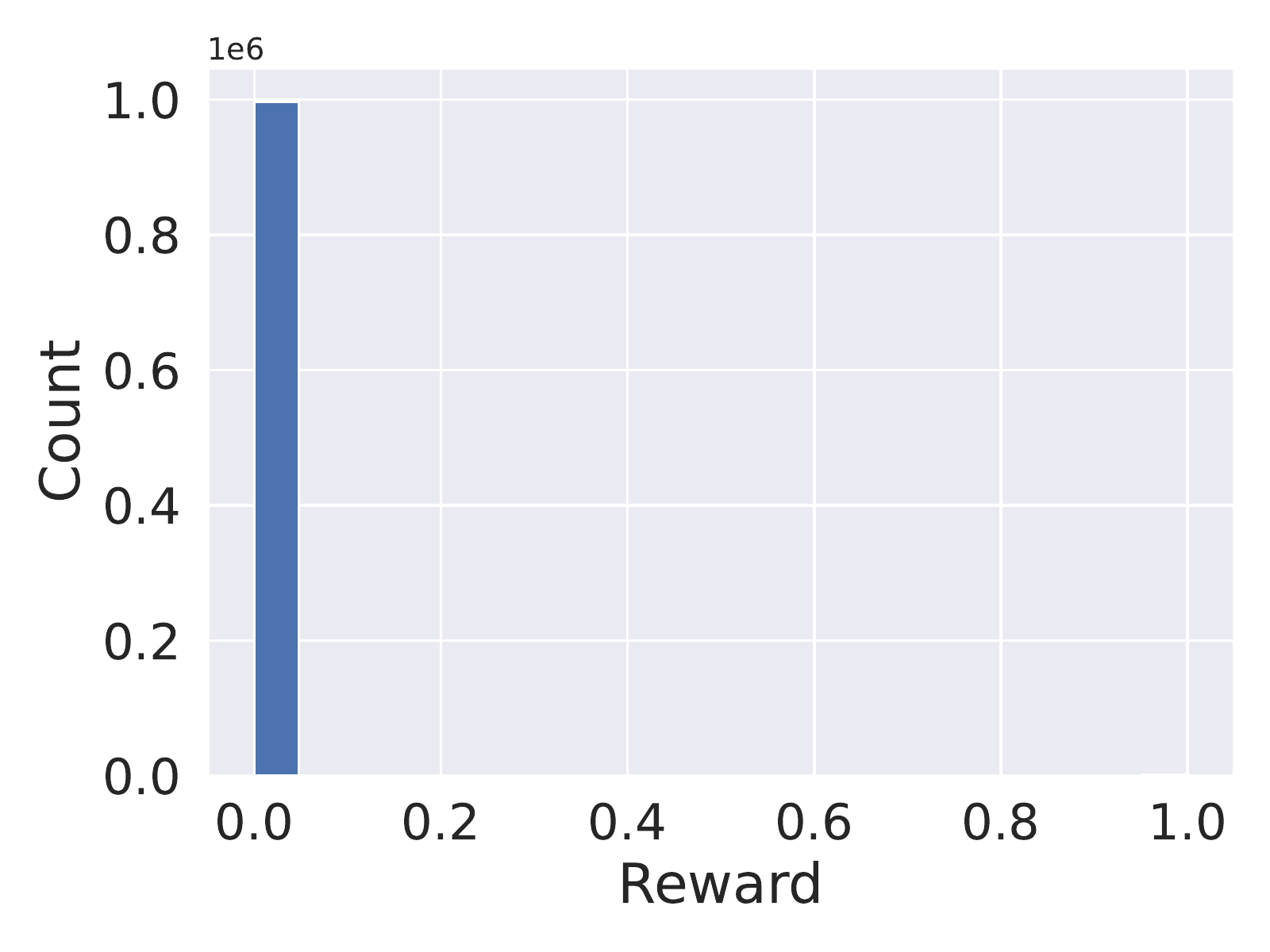}&\includegraphics[width=0.225\linewidth]{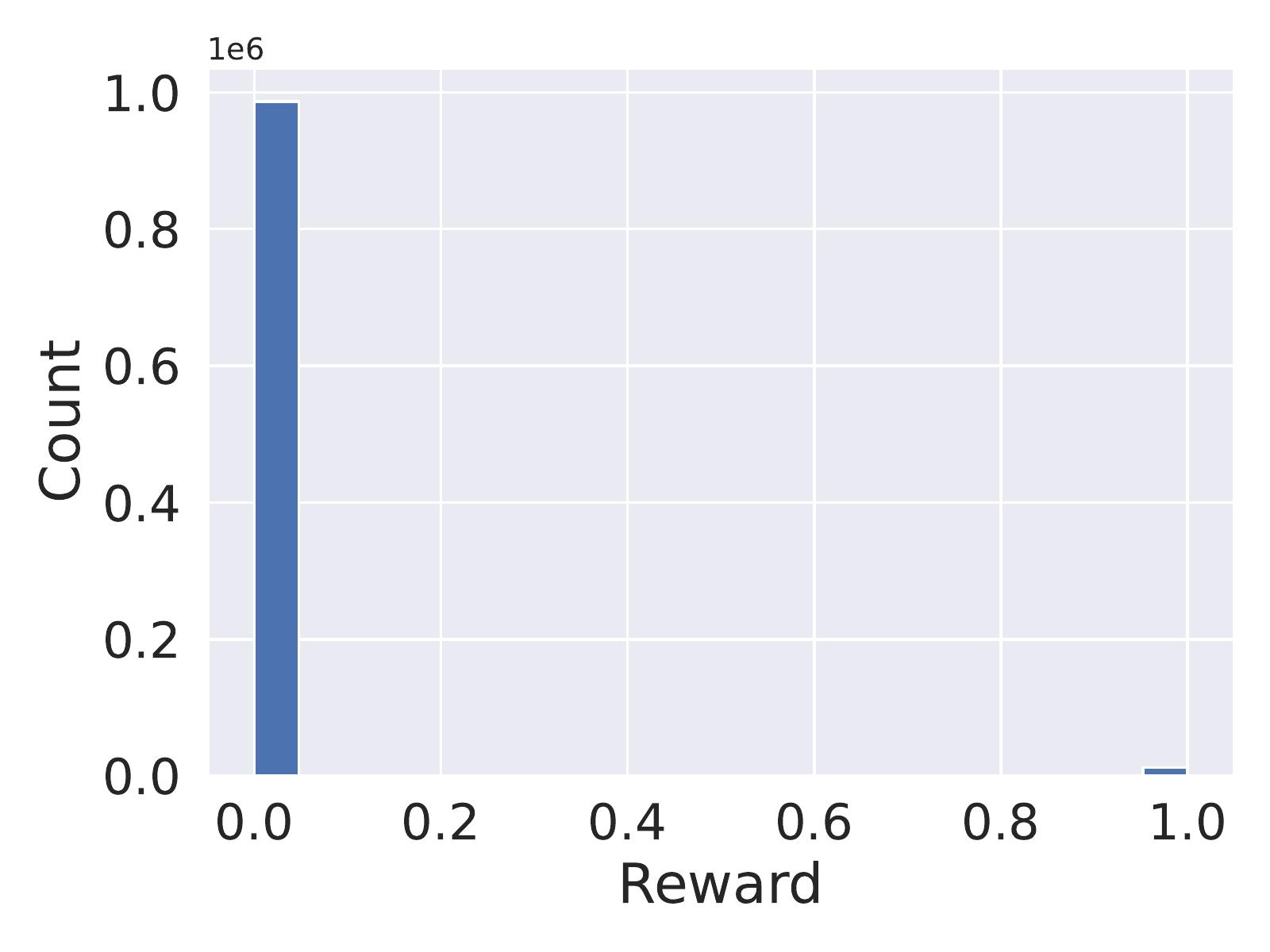}&\includegraphics[width=0.225\linewidth]{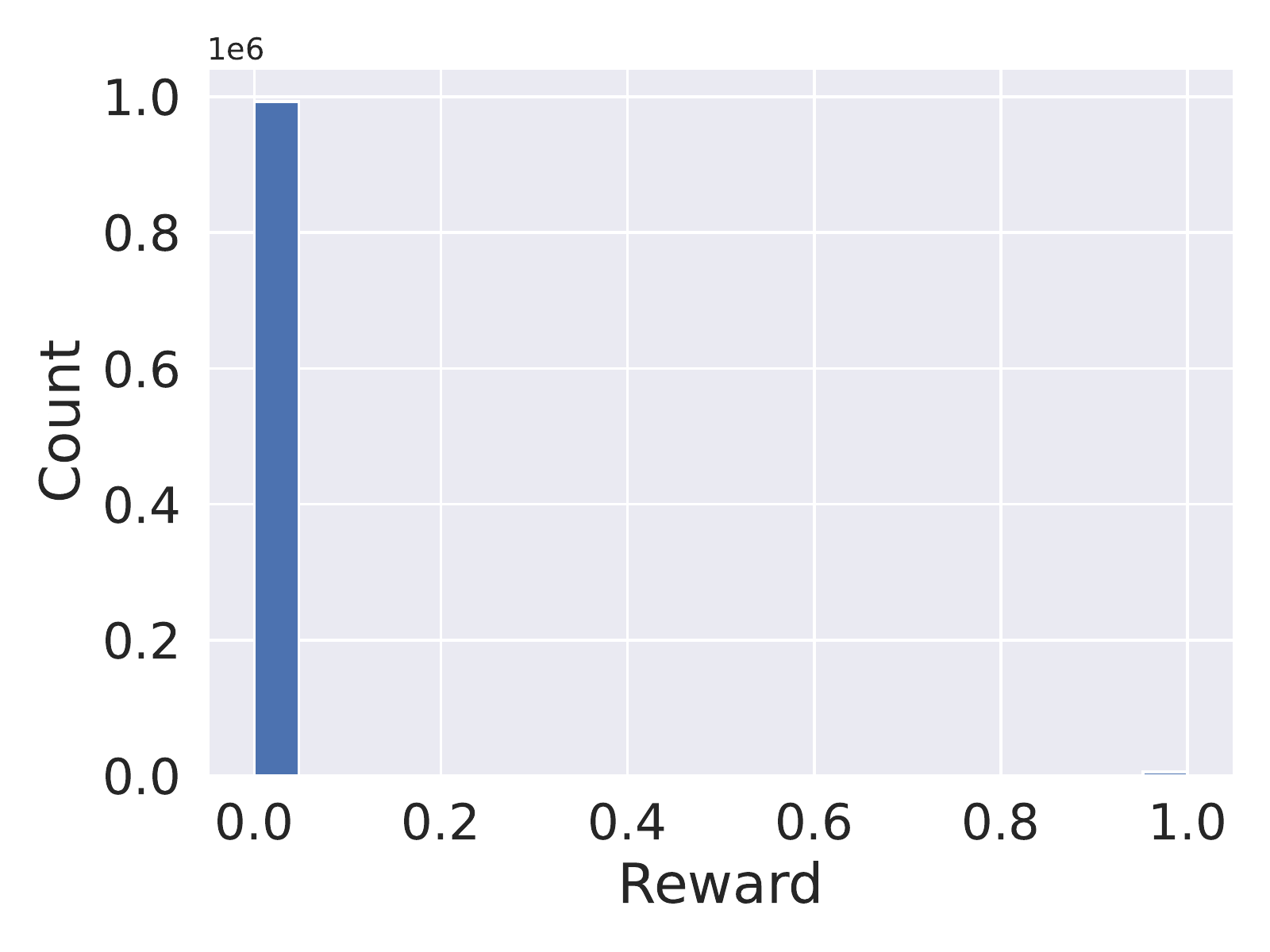}\\\end{tabular}
\centering
\caption{Reward Distribution of D4RL}
\end{figure*}
\begin{figure*}[htb]
\centering
\begin{tabular}{cccccc}
\includegraphics[width=0.225\linewidth]{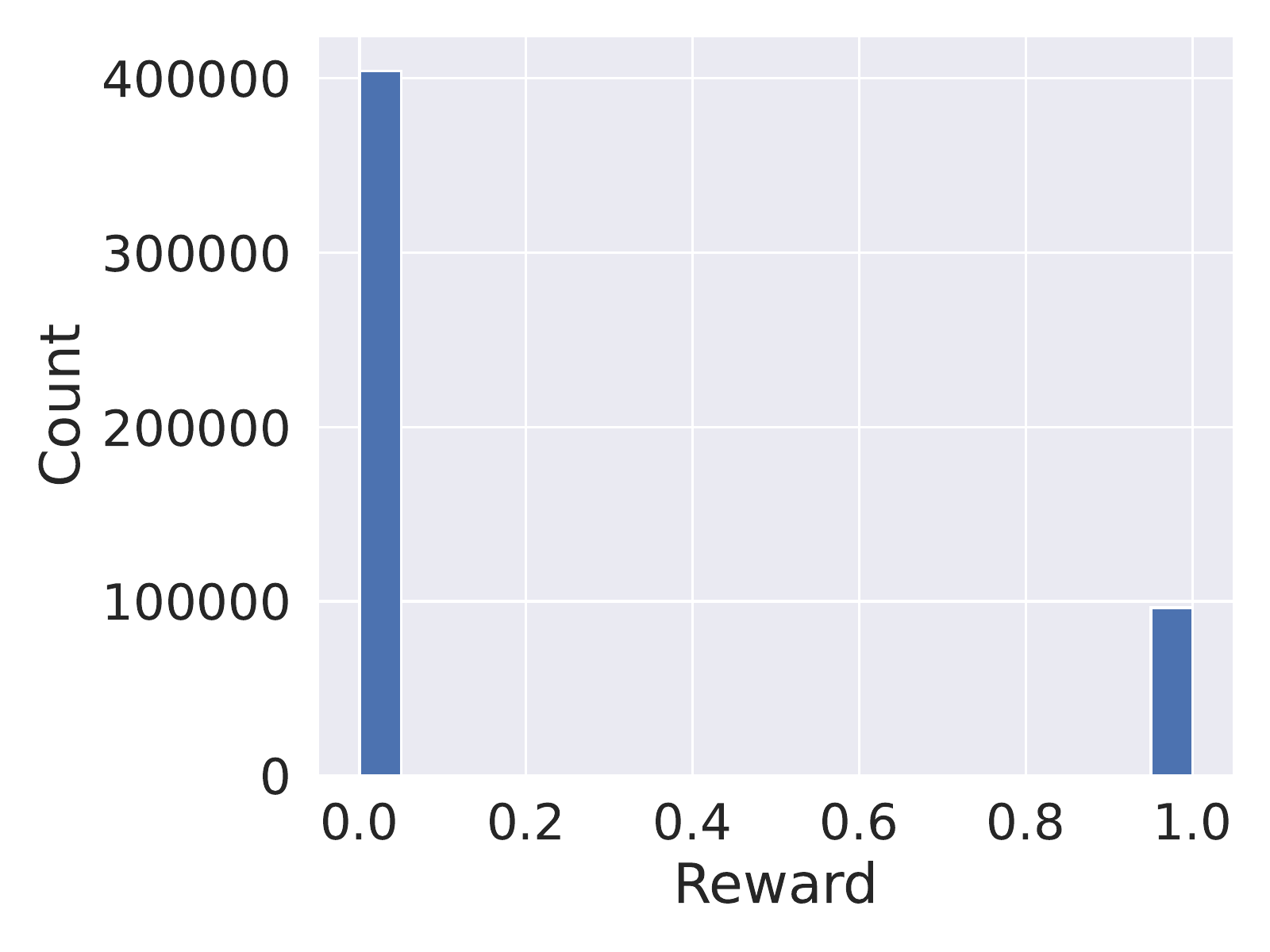}&\includegraphics[width=0.225\linewidth]{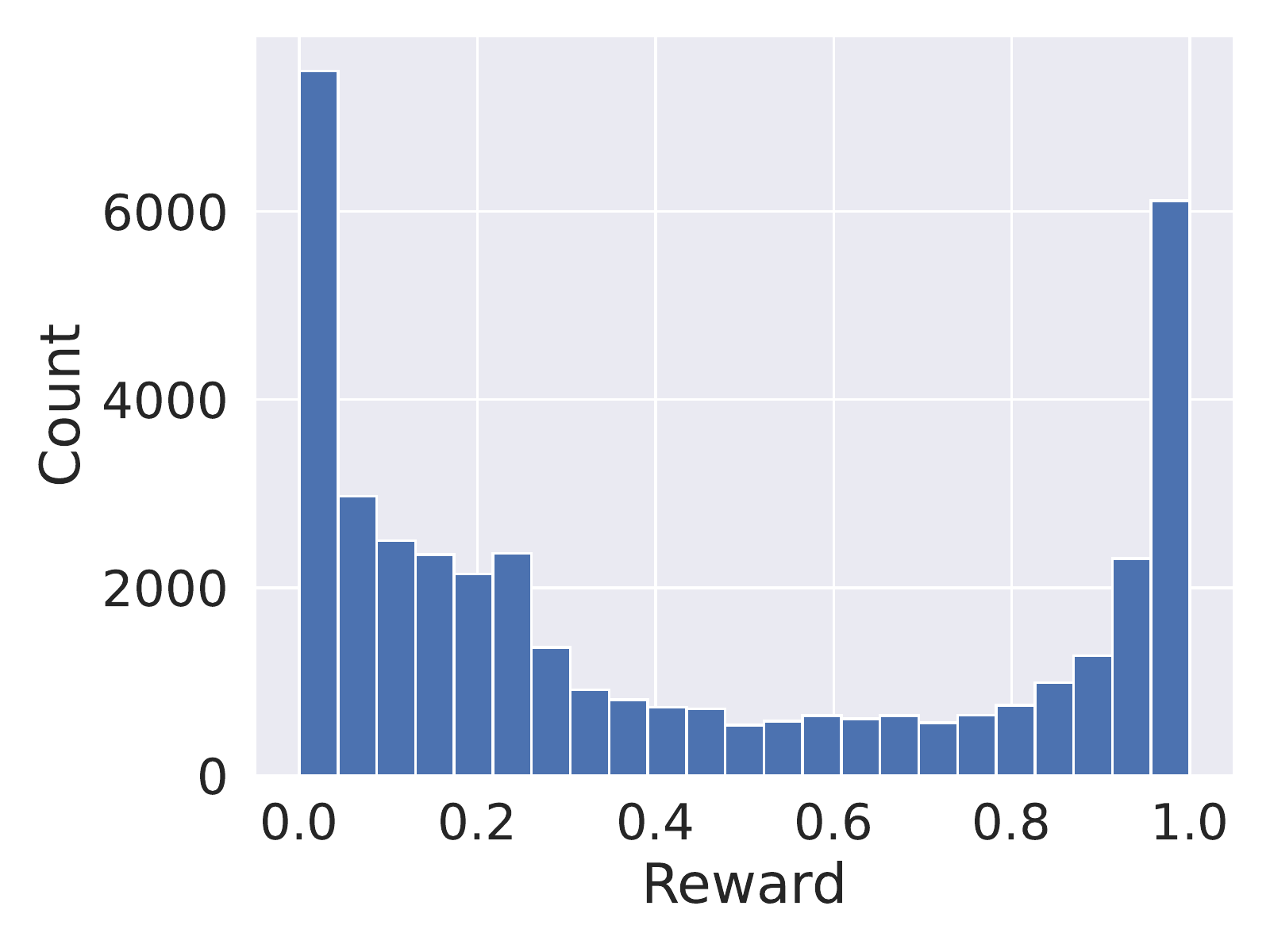}&\includegraphics[width=0.225\linewidth]{dist_plots/cheetah_run}&\includegraphics[width=0.225\linewidth]{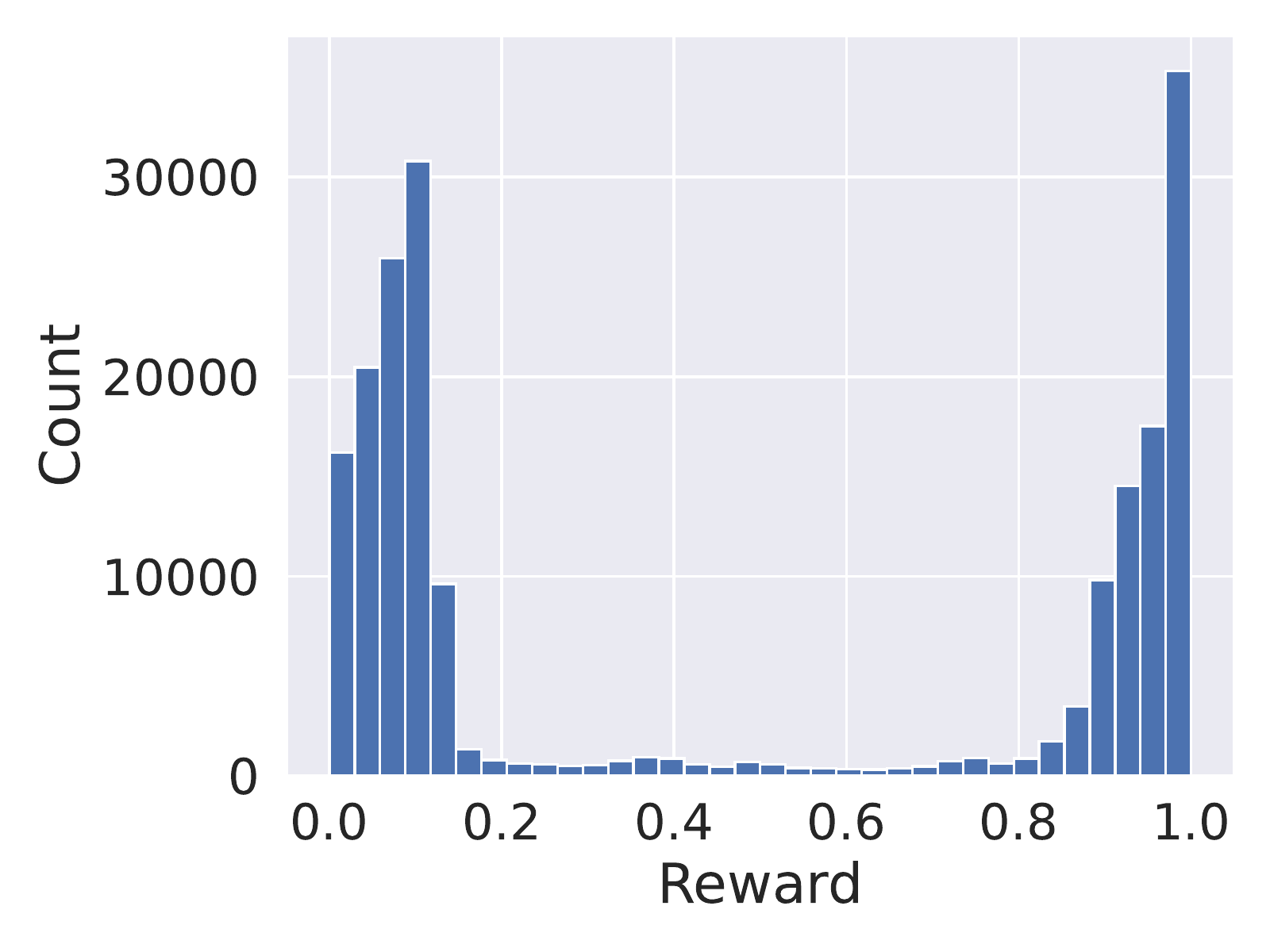}\\\includegraphics[width=0.225\linewidth]{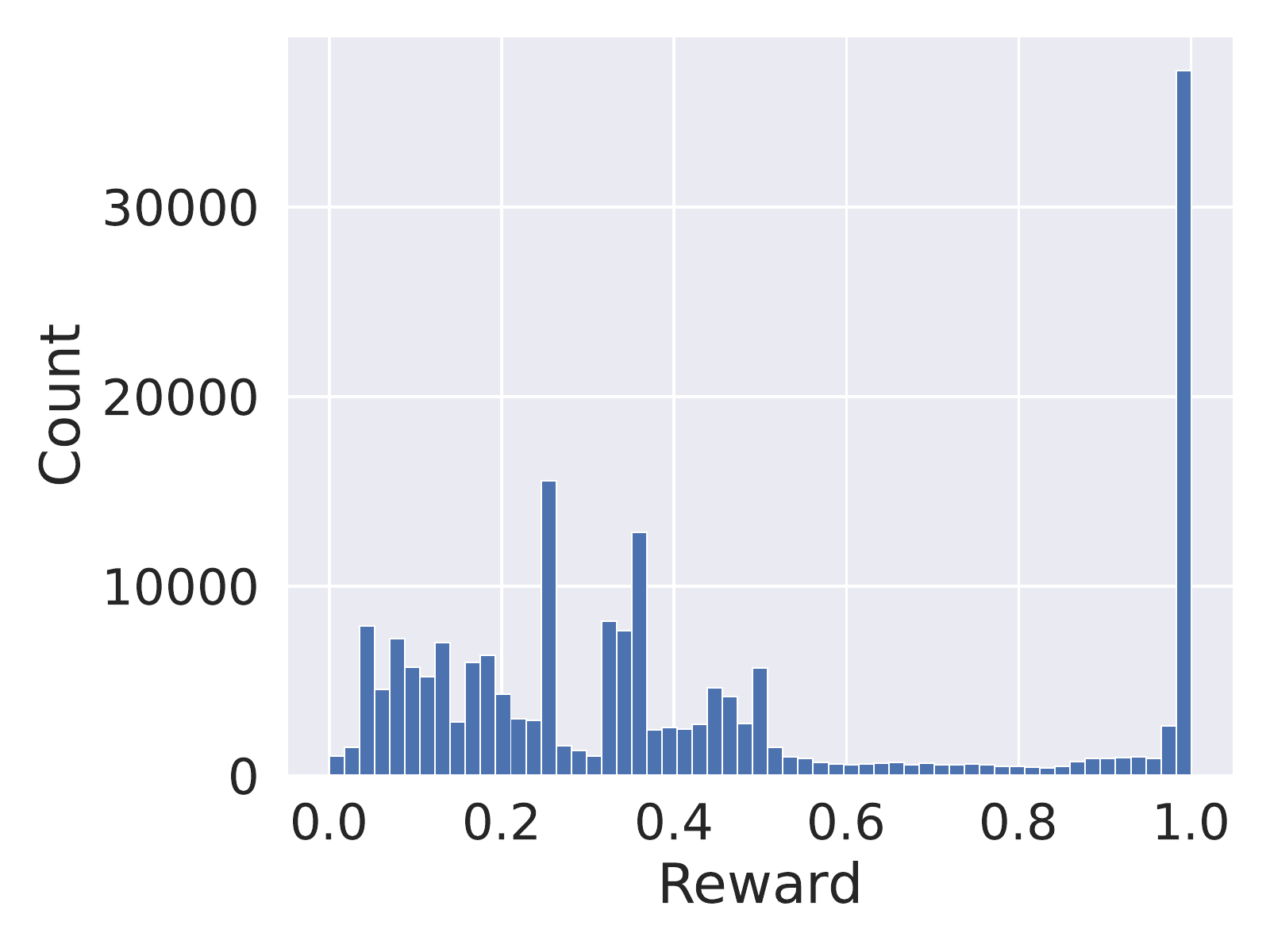}&\includegraphics[width=0.225\linewidth]{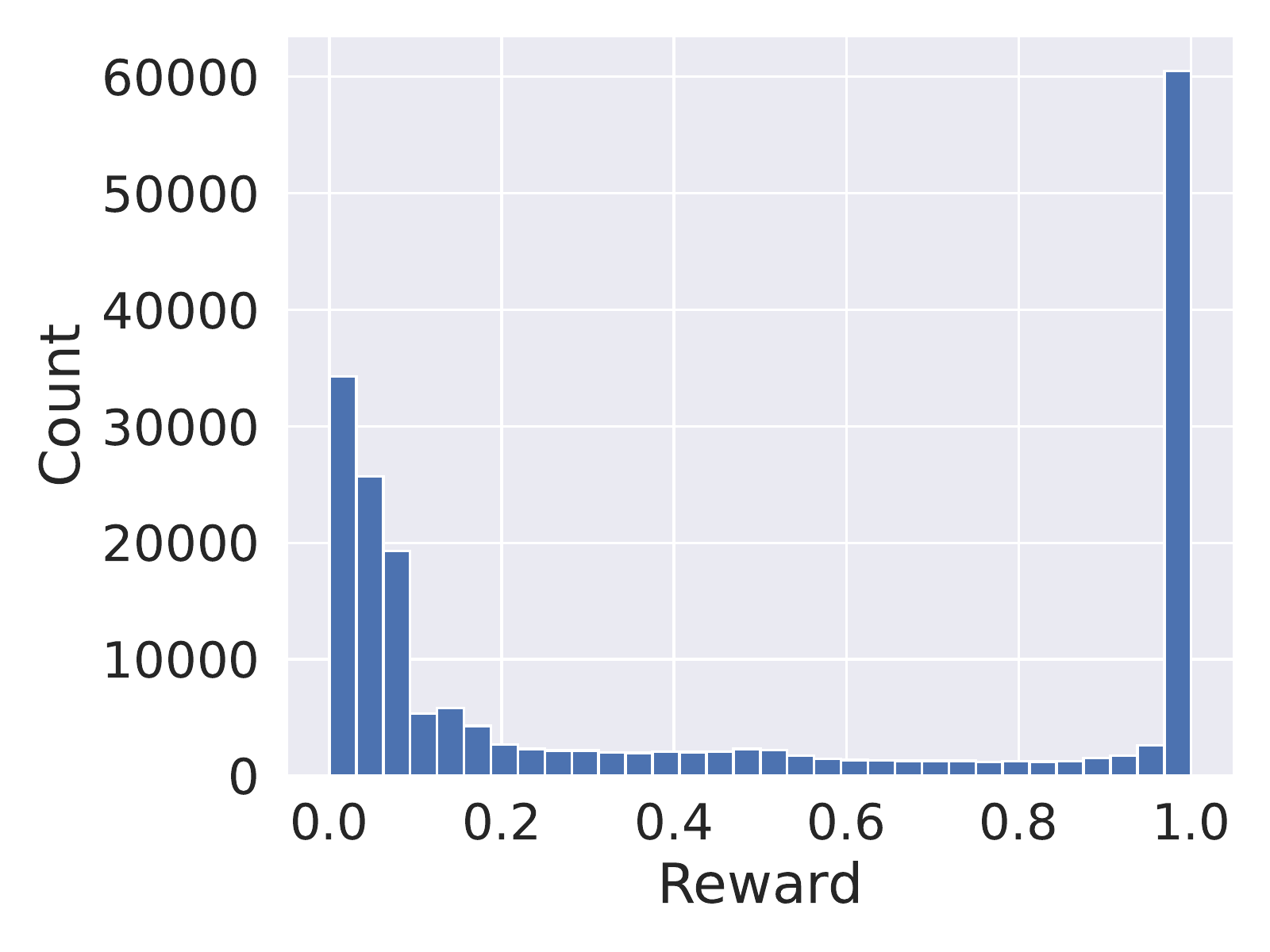}&&\\\end{tabular}
\centering
\caption{Reward Distribution of RLUP}
\end{figure*}

\end{document}